\documentclass{article} 

\pdfoutput=1

\usepackage{amsmath,amsfonts,bm}

\def\eqref#1{equation~\ref{#1}}

\def\1{\bm{1}}

\def\va{{\bm{a}}}
\def\vb{{\bm{b}}}

\def\vx{{\bm{x}}}

\def\mA{{\bm{A}}}

\def\mD{{\bm{D}}}
\def\mE{{\bm{E}}}

\def\mG{{\bm{G}}}

\def\mK{{\bm{K}}}

\def\mP{{\bm{P}}}
\def\mQ{{\bm{Q}}}

\def\mV{{\bm{V}}}
\def\mW{{\bm{W}}}
\def\mX{{\bm{X}}}
\def\mY{{\bm{Y}}}

\DeclareMathAlphabet{\mathsfit}{\encodingdefault}{\sfdefault}{m}{sl}
\SetMathAlphabet{\mathsfit}{bold}{\encodingdefault}{\sfdefault}{bx}{n}

\newcommand{\R}{\mathbb{R}}

\usepackage{hyperref}
\usepackage{url}
\usepackage{graphicx}
\usepackage{braket}
\usepackage{booktabs}
\usepackage{caption}
\usepackage{subcaption}
\usepackage{changepage}

\graphicspath{{plgt_paper_figures_pdf/}}

\title{Power Law Graph Transformer for Machine Translation and Representation Learning}

\author{Burc Gokden \\
Fromthesky Research Labs LLC\\
Oregon, USA \\
\texttt{burc@fromtheskyresearchlabs.com} \\
}

\begin{document}

\maketitle

\begin{abstract}

We present the Power Law Graph Transformer, a transformer model with well defined deductive and inductive tasks for prediction and representation learning. The deductive task learns the dataset level (global) and instance level (local) graph structures in terms of learnable power law distribution parameters. The inductive task outputs the prediction probabilities using the deductive task output, similar to a transductive model. We trained our model with Turkish-English and Portuguese-English datasets from TED talk transcripts for machine translation and compared the model performance and characteristics to a transformer model with scaled dot product attention trained on the same experimental setup. We report BLEU scores of $17.79$ and $28.33$ on the Turkish-English and Portuguese-English translation tasks with our model, respectively. We also show how a duality between a quantization set and N-dimensional manifold representation can be leveraged to transform between local and global deductive-inductive outputs using successive application of linear and non-linear transformations end-to-end.
   
\end{abstract}

\section{Introduction}

Statistically distributed representations of language models\cite{hinton1986, bengio2003} and application of attention models \cite{bahdanau2014, vasvani2017} resulted in breakthrough improvements in Natural Language Processing (NLP) tasks using deep neural networks. These approaches can also be used to design a graph transformer that has deductive and inductive components more clearly established than a transductive model. The deductive functionality can be achieved by expanding the data representation to learn generic representations for a vocabulary $\mathcal{V}$ (of tokens), which is a quantization set of $V$ discrete graph states ($x_i \in \mathcal{V}$) that is a superset of a sentence $\vx=\{x_1, x_2,...,x_S\}$. A sentence that is syntactically and semantically valid in a language model (LM) represents a graph instance of tokens from the quantization set, each represented with statistically distributed dense embedding vectors with N feature dimensions. A graph transformer model can be developed if we can learn metric tensor instances of language model manifold from graph instances and derive an accompanying energy-curvature tensor that can be used to propagate the language model vectors across the encoder-decoder network. A big challenge in defining such a model is the need for expert domain knowledge to define connections between graph states in terms of a weighted adjacency matrix or a more abstract metric tensor that can generalize to an N-dimensional manifold where N can be very large. In our previous work \cite{burc2019}, we showed that it is possible to predict molecular properties in a simple one-hot encoding setting where metric tensor was a hand-engineered inverse-distance weighted adjacency matrix of size $W \times W$ with $W$ being a pre-set maximum number of nodes for each graph. The energy-curvature tensor was a matrix of same size derived as part of a learnable coulomb attention model applied on the adjacency matrix and hidden states. We also proposed in our previous work that this attention model can be improved and generalized by using distributed embedding representations and transformer architecture. 

In this paper, we present a generalized form of our power law attention model that is scalable to any graph size for a given quantization base set $\mathcal{V}$ of size V and a non-linear manifold of N feature dimensions. Specifically, we develop an end-to-end deductive-inductive power law graph transformer (PLGT) model for machine translation task by using a set of linear embedding vectors from source and target languages. For deductive task, the model learns generalized power law coefficients, metric tensor and energy-curvature tensor instances for a language model manifold. For the inductive task, the attention model learned in deductive task is used to predict probabilities from source input autoregressively, producing same output as a transductive transformer.

In the next sections, we will briefly go over background work, and present the details of the power law attention model and the graph transformer architecture. Then, we will show our results for Turkish-English (TR-EN) and Portuguese-English (PT-EN) translation tasks from ted\_hrlr\_translate dataset \cite{qi2018, tedhrlrtranslate}. 

\section{Background}

A key understanding in data representation that significantly improved the performance of neural machine translation (NMT) models was distributed representation of data first introduced in \cite{hinton1986}. The distributed representations of statistical language models developed in \cite{bengio2003} demonstrated that the joint probability distribution of discrete random variables can be used to represent each token (e.g.\ word, subword) in a sentence as a dense vector. These vectors can provide the model with information that grows exponentially within an embedding vector space to reduce the curse of dimensionality. Each embedding vector is composed of a representation with fixed number of feature dimensions for each token in a vocabulary. Then a joint probability distribution for a word sequence can be learned from these vectors which are conveniently called word embeddings. The ability to represent a language model statistically with word embeddings was further improved for large scale data in \cite{mikolov2013a, mikolov2013b} that introduced projection only training for CBOW and skip-gram word2vec models with additional optimizations for the objective function. A key achievement of word2vec was their efficient linear representation of syntactic and semantic relationships with embedding vectors demonstrating improved analogical reasoning \cite{mikolov2013b}. Another word embedding model, GloVe used global corpus statistics to demonstrate similar analogical reasoning capabilities \cite{pennington2014}.

State of the art in statistical machine translation (SMT) was further improved by using recurrent neural networks (RNN) with Long-Short Term Memory (LSTM) cells \cite{sutskever2014} and Gated Recurrent Units (GRU) \cite{cho2014} in an encoder-decoder architecture. The RNN encoder-decoder architectures suffered from inability to translate longer sentences, where a fixed sized vector formed a bottleneck to represent all the data from source sentence into the decoder. The introduction of attention models that can attend to different parts of the source sentence by learning an additive alignment model improved these pioneering SMT models to predict longer sentences more accurately. An attention model provides a weighted context vector learned from source sentence to the decoder to predict the next target word \cite{bahdanau2014} to overcome the bottleneck from a fixed-length vector. Efficient methods using dot-product attention with global and local approaches were explored and compared in \cite{luong2015}. The concept of self attention that used linear combination of hidden states to achieve representation of variable length sentences into a fixed sized embedding was demonstrated in \cite{lin2017}. The transformer model \cite{vasvani2017} that relies on dot product based self attention and encoder-decoder architecture without recurrent networks demonstrated results better than RNN based SMT models with reduced training cost. The transformer model forms the basis for advanced NLP architectures today  \cite{devlin2019, brown2020}. These models utilize mainly a transductive learning approach \cite{vapnik}.

\section{Model Architecture}

\begin{figure}
\centering
\begin{subfigure}[b]{\textwidth}
	\centering
	\includegraphics[width=\textwidth]{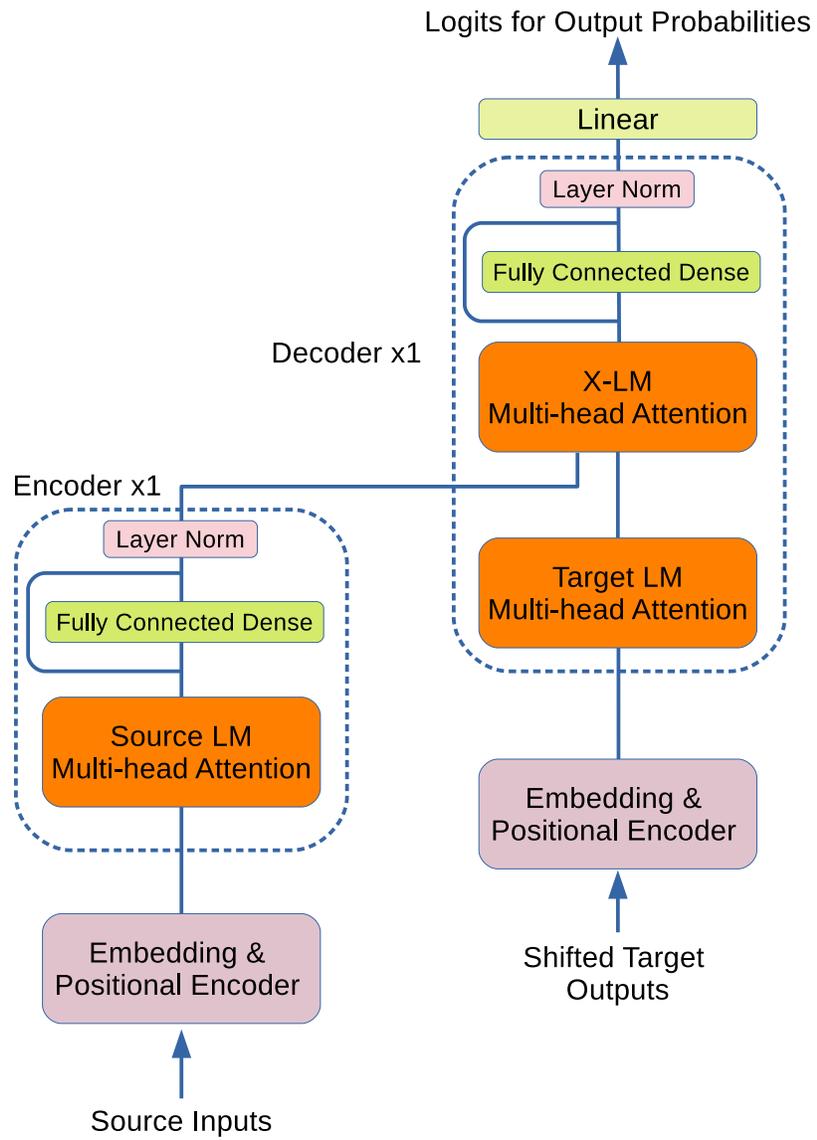}
\end{subfigure}
\caption{Single layer encoder-decoder architecture of the graph transformer}
\label{fig1a}
\end{figure}

Our model follows the general design principles in \cite{vasvani2017} that uses a scaled dot product attention (SDPA) based encoder-decoder architecture to represent and translate data within the model. A key difference is in the attention model which utilizes both linear self attention and power law attention together with deeply connected neural layers. The model is autoregressive, consisting of single layer encoder-decoder configuration that takes in a sequence of input sentences and target sentence is formed one token at a time, where earlier predicted tokens are fed as input to the decoder. The encoder and decoder first learn a metric tensor instance through a deep neural network accepting linear self attention of source and target inputs, respectively. This metric tensor is then used to learn the energy-curvature tensor that can facilitate localized linear transformations between source and target languages. The general layout of graph transformer is shown in fig. \ref{fig1a}.

\begin{figure}
\centering
\begin{subfigure}[b]{\textwidth}
	\centering
	\includegraphics[width=\textwidth]{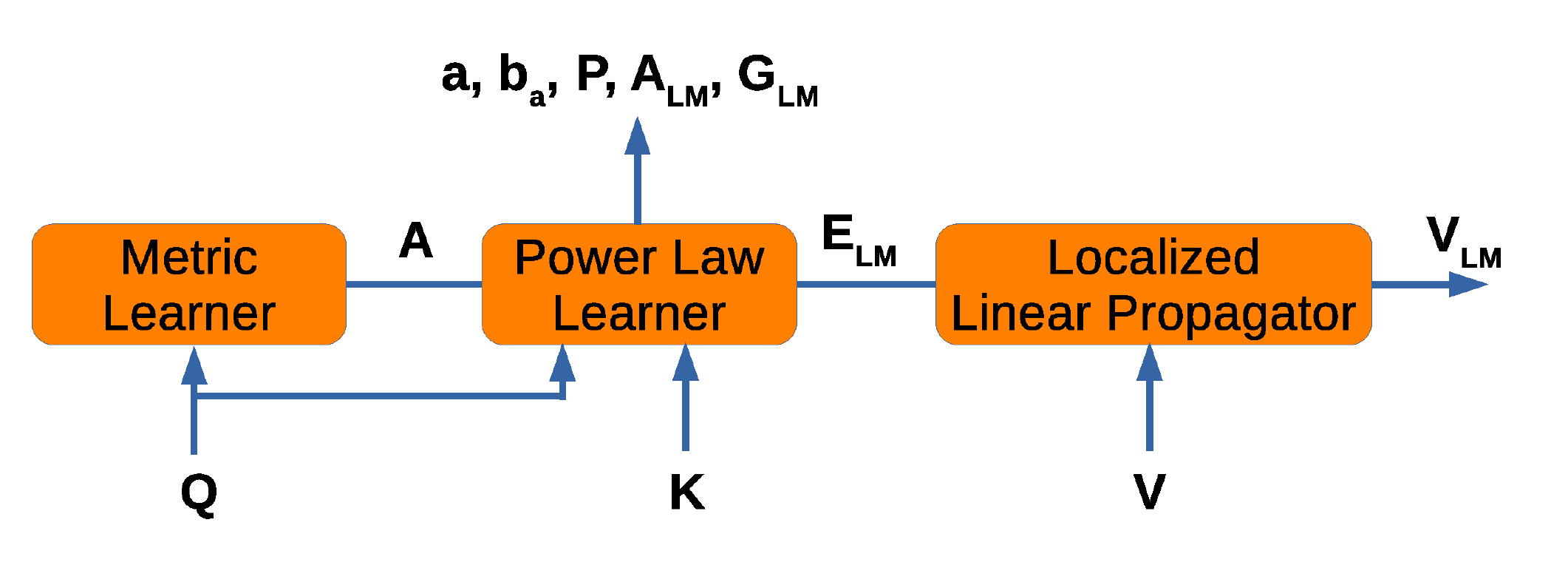}
\end{subfigure}
\caption{Functional diagram of the attention block}
\label{fig1b}
\end{figure}  

\subsection{Encoder and Decoder Layers}

Encoder first converts the tokenized input sentence $\vx=\{x_{1}, x_{2}, x_{3},\ldots, x_{S}\}$ into a learned embedding matrix $\mX$ with vector space dimension $d_{emb}$ for each token: $\mX \in \R^{S \times d_{emb}}$. A positional encoding is added to $\mX$ to inform the model of the sequence of tokens in input sentence after multiplying the embedding matrix with $\sqrt{d_{emb}}$ \cite{vasvani2017}. The encoder consists of a multi-head attention layer and deep fully connected dense layers with residual connections and layer normalization  \cite{ba2016} at their output. The power law attention design used in encoder is a combination of linear transformations and deep residual neural network layers that learn the source language model (SLM) representation from an ensemble of source input sentences.

The decoder takes as input an embedding matrix $\mY_{shifted}$ prepared in the same way as the input embedding to the encoder layer. The decoder encodes the right shifted target sentence into a target language model (TLM) representation in the first attention layer. The second attention layer takes as input the encoder output for the source language and output of the first attention layer to form the cross-language model (XLM) representation from projections of source and target language model representations of input and target sentence. The last stage of decoder is a fully-connected dense layer same as in encoder. Each attention layer and dense layer has residual connections followed by layer normalization at the output. The attention output in the decoder is masked to ensure that the prediction can only depend on known outputs that occur earlier in the sequence.

We use single encoder and decoder layer for the graph transformer implemented in this work, although it is possible to scale model with identical encoder and decoder stacks.

\subsection{The Power Law Graph Attention (PLGA) Layer}

The attention layer for graph transformer consists of three stages as shown in fig. \ref{fig1b}. In the first stage, a metric tensor is inferred from an input graph which is a matrix formed by concatenating embedding vectors with feature dimension of $d_{emb}$ for $S$ tokens (graph nodes) in the input sentence. Metric tensor for a language model manifold is a generalized, abstract form of a weighted adjacency matrix learned through a deep neural network. For many types of unstructured data that can be represented as a graph with large number of dimensions as features and many connections between nodes, it is not straightforward to define a distance metric between each node. In our earlier study \cite{burc2019}, the inverse of three dimensional eucledian norm between each node was used as a hand-engineered distance metric to define a weighted adjacency matrix for each graph (molecules) to demonstrate a reasonable level of prediction capacity for the graph attention model. The first stage learns the metric tensor in an end-to-end fashion using self attention and a deep residual network without the need to define a distance heuristic from domain knowledge.

The second stage uses the metric tensor as an input to learn power law relationships and coupling coefficients for a generalized energy-curvature (EC) tensor for the language model manifold. Thus, each element of an EC tensor is a superposition of exponentiated metric tensor elements weighted by a coupling coefficient. EC tensor corresponds to the generalized form of a language model represented  by a manifold with $d_{LM}$ dimensions. We refer to this tensor the Energy-Curvature tensor for two reasons: First, it is derived entirely from the metric tensor in a similar fashion the Ricci tensor that defines the curvature of a Riemannian manifold is derived. Secondly, imposing a power law relationship through metric tensor elements approximates a sum of abstract potentials that manipulates the curvature of the manifold.

The third stage is a linear transformation which evolves the embedding space representation of the input sentence to an instance of a language model representation as output of the attention layer.

The attention layer has multi-head support where input is split into $h$ heads with depth of each attention sublayer defined as $d_k=d_{LM}/h$. Each head learns its own subspace of metric tensor and energy-curvature tensor for a subset of language model dimensions. The attention layer architecture is shown in fig. \ref{fig2}.

\subsubsection{Learning Metric Tensor from self-attention}

The input to attention model is a dense matrix $\mX$ of size $S \times d_{emb}$ superposed with positional encoding. The language model dimension $d_{LM}$ is set to be equal to embedding dimension $d_{emb}$ in our implementation. We define a localized graph operator for a single input graph instance represented by $\mQ(=\mX)$ using self-attention:

\begin{equation}
\mD_{Q}=\mQ^{T}\mQ \equiv \Ket{\mQ^T} \Bra{\mQ^T} \label{eq1}
\end{equation}

$\mD_{Q}$ is a $d_{LM} \times d_{LM}$ density matrix operator for a graph (sentence) with mixed statistically distributed representations. We also introduce the bra-ket notation for $\mQ$ which is a concatenated, well-defined sequence of embedding vectors that carry linear syntactic and semantic relationships and distributed probabilistic representations of elements in a Vocabulary. To align the size defined in model implementation ($S \times d_{LM}$ for $\mQ$) with bra-ket notation, $\mQ^T$ is used as the bra-ket state. Thus $\Ket{\mQ^T}$ state has same size of the matrix $\mQ^T$ ($d_{LM} \times S$) and $\Bra{\mQ^T}$ is the transpose. Each element of $\mQ$ attends onto other elements of the same graph and this operator can be used to get the degree of $\mQ$-ness in another graph $\mV$ such that $\Ket{\mQ^T} \Braket{\mQ^T|\mV^T}$. The inner product $\Braket{\mQ^T|\mV^T}$ is a matrix where each entry is dot product of token vectors and is a measure of similarity between tokens.

\begin{figure}
\centering
\begin{subfigure}[b]{\textwidth}
	\centering
	\includegraphics[width=\textwidth]{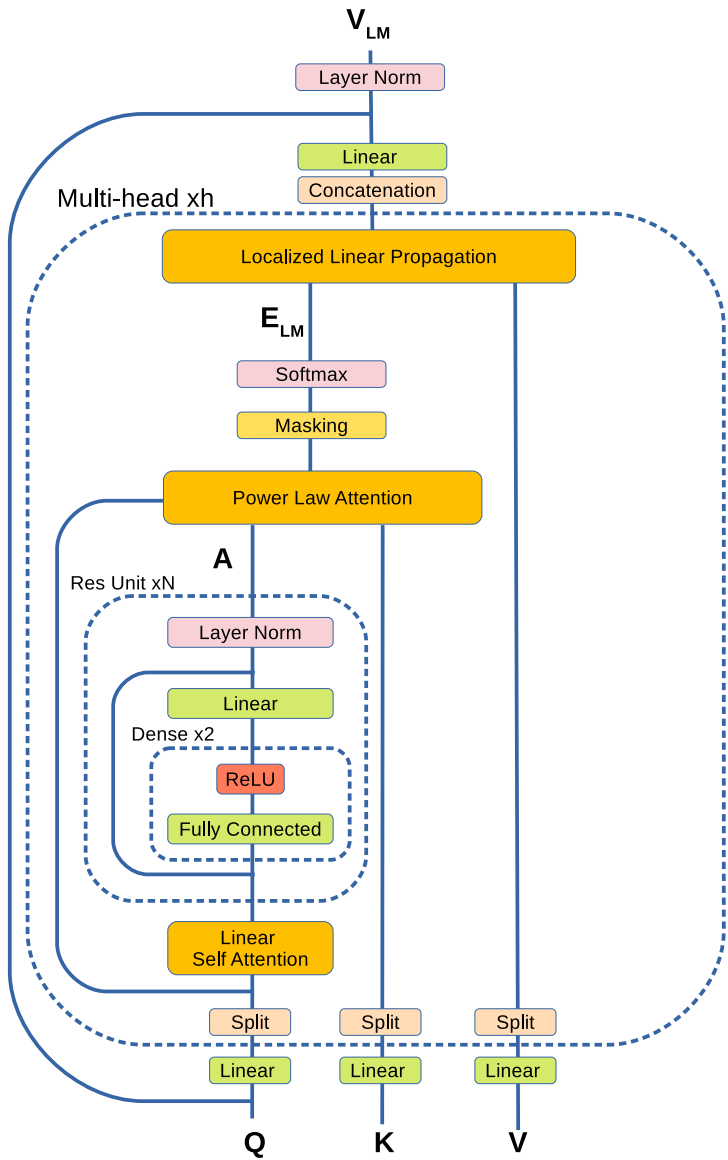}
\end{subfigure}
\caption{Model architecture of attention block implemented within encoder and decoder of the power law graph transformer.}
\label{fig2}
\end{figure}

The metric tensor $\mA$ is learned from the self-attention of training instances through a deep residual network where each residual unit is composed of two fully-connected layers of size \textit{A-dff} with ReLU activation for each layer followed by a linear fully-connected layer of size $d_k$ and layer normalization as shown in fig. \ref{fig2}.

The generalized metric tensor $\mA$ is finally wrapped through a fully-connected layer with learnable weights $\mW$ and bias $\vb_{W}$:

\begin{equation}
\mA_{LM}=ReLU\left(\mW\mA+\vb_{W}\right)+\epsilon \label{eq2}
\end{equation}

The use of ReLU activation with a small value of $\epsilon=1 \times 10^{-9}$ ensures that $\mA_{LM}$ is a tensor with positive non-zero elements. We found that the model converges robustly with this configuration since we also randomly initialize the learnable elementwise power matrix $\mP$ defined in next section with glorot initialization \cite{glorot2010}.

\subsubsection{Learning Energy-Curvature Tensor for the Language Model}

The energy-curvature tensor for the language model $\mG_{LM}$ is derived from metric tensor as:

\begin{equation}
\mG_{LM} = \va\mA_{LM}^{\odot{\mP}}+\vb_{a}  \label{eq3}
\end{equation}

$\va$ and $\vb_{a}$ are the learnable coupling weights and bias for potentials generated from metric tensor $\mA_{LM}$. $\mP$ is a learnable power matrix that is applied to $\mA_{LM}$ elementwise.

The deductive task infers a generalized metric tensor, energy-curvature tensor and learns coupling and power coefficients for the language model characterized with $d_{LM}$-dimensional manifold and a quantization set of the Vocabulary size $V$. To achieve the inductive task, it is necessary to obtain a localized instance projection of the energy-curvature tensor that can transform the representation of an input graph to a language model representation. The EC tensor is first projected onto the graph instance by finding the expected value of the EC tensor weighted over query and key inputs in eq. \ref{eq41}. The localized EC operator is scaled by $\sqrt{d_k}$ to avoid small gradient regions. The scaled EC operator is then run through a Leaky ReLU activation step followed by softmax. Masking is applied before softmax by setting values to be ignored close to $-\infty$. The resulting localized graph operator $\mE_{LM}$ is then applied onto input value vector $\mV$ as a linear transformation (eq. \ref{eq44}):

\begin{eqnarray}
\label{eq4}
E_{QK}[ \mG_{LM}] &=& \mQ \mG_{LM} \mK^T \equiv \Braket{\mQ^T|\mG_{LM}|\mK^T} \label{eq41}\\
\mE &=& LeakyReLU\left(E_{QK}[ \mG_{LM}]/\sqrt{d_k}\right) \label{eq42}\\
\mE_{LM} &=& softmax\left[mask(\mE)\right] \label{eq43}\\
\mV_{LM} &=& \mE_{LM} \mV \equiv \mE_{LM} \Ket{\mV} \label{eq44}
\end{eqnarray}  

The inductive task output of the attention layer is the language model representation of input $\mV_{LM}$ and the deductive task outputs are:  $\mE_{LM}$, $\mP$, $\va$, $\vb_{a}$, $\mG_{LM}$, $\mA_{LM}$.

For the source language model encoder, there is a single stage attention layer where query, key and value entries are all equal to the source sentence embedding (SE) vector sequence, $\mQ_{SE}=\mK_{SE}=\mV_{SE}=\mX$. For the decoder, the first attention layer has its query, key and value as the target sentence embedding (TE) vector sequence shifted right, $\mQ_{TE}=\mK_{TE}=\mV_{TE}=\mY_{shifted}$. For the second attention layer (XLM) of decoder for cross-language model transformation, query is the output of the first attention layer, $\mQ_{XLM}=\mV_{TLM}$ and key and value are the output of source encoder layer, $\mK_{XLM}=\mV_{XLM}=\mV_{SLM}$.

\section{The Dataset}

The dataset is used in this study is a parallel corpus created from TED Talk transcripts for two different language families: Portuguese-English (PT-EN) and Turkish-English (TR-EN) machine translation tasks \cite{qi2018}. The PT-EN dataset is composed of 51785 sentence pairs for train, 1193 sentence pairs for development and 1803 sentence pairs for test. The TR-EN dataset is composed of 182450 sentence pairs for train, 4045 sentence pairs for development and 5029 sentence pairs for test. We used the dataset as prepared in Tensorflow Datasets Catalog \cite{tedhrlrtranslate}. The training dataset was shuffled before each training run. The datasets were used to create a subword vocabulary of maximum 15k tokens using the wordpiece approach used in BERT implementation \cite{shuster2012, devlin2019}. The tokenizer used for the source and target were Bert Tokenizer implementation in Tensorflow-Text package \cite{tftext}. Vocabulary is generated separately for Portuguese and English ($\sim8k$ tokens each), as well as Turkish and English ($\sim15k$ tokens each) from their respective paired train datasets.

\section{Experimental Setup}

We trained graph transformers that have single encoder-decoder layer as shown in fig. \ref{fig1a}. The embedding and language model manifolds have dimension sizes of $d_{emb}=d_{LM}=512$. We ran models with different number of attention heads and scaled dense connections for the metric tensor and pointwise feedforward connections accordingly to maintain same $\textit{A-dff}/d_k$ ratio for all models. We tried multi-head attention with 1,2,4,8 and 16 heads and scaled residual layer/residual dense/point-wise feed-forward parameters as shown in table \ref{table1}. We also trained a transformer model with scaled dot-product attention (SDPA) that has 4 encoder-decoder layers, 8 heads and a drop-out rate of 0.1 for comparison. SDPA transformer has same $d_{LM}$ and $\textit{dff}$ values as the power law graph transformer models. The number of residual layers for graph transformers and the number of encoder-decoder layers for SDPA model were chosen to be maximum values a single GPU in our setup can handle for each model without running out of memory except for model \#3 in table \ref{table1}.

The training is performed by using Adam optimizer \cite{kingma2015} at a custom scheduled learning rate \cite{vasvani2017} with warm-up steps of $15000$ for PLGA and $4000$ for SDPA transformer models and used a batch size of $64$ for both. Attention layer weights, fully connected layer weights and biases were initialized with glorot normal, glorot uniform and zeros, respectively. During training, we kept track of the training and validation cross-entropy loss (log perplexity), and accuracy at the end of every epoch. Outside the attention layer, drop-out \cite{srivastava14} is applied to embedding inputs after positional encodings are added as well as before residual sum and layer normalization at the attention layer and encoder-decoder outputs. Outside drop-out rate was set at $0.4$. Inside the attention layer, a drop-out rate of $0.1$ is also applied at output of every residual unit before summing and layer normalization for metric tensor learning. The drop-out rate within attention layer for inputs $\mQ$, $\mK$ were kept at zero and dropout rate for $\mE_{LM}$ was set at $0.1$. We found these values to give good compromise to avoid overfit of loss curve quantified by log perplexity over $120$ epochs. We kept checkpoints for the model parameters at $10$ epochs after the minimum validation loss is observed \cite{cho2014}, at highest validation accuracy and a number of checkpoints sampled over $120$ epochs for comparison. Training took $\sim10-36$ hours for each model based on hyperparameters and the dataset.  BLEU metric \cite{papineni2002} was used to evaluate the test dataset. BLEU score was calculated using sacrebleu package \cite{sacrebleu2018}. For evaluation of our models, we run predictions using beam search with beam length of $1$ (greedy search) and $4$ with length normalization \cite{wu2019}. The maximum number of iterations carried out for evaluation was set at 50 iterations above input sentence length.  Model variations were implemented using Tensorflow \cite{tfwp2015}. Implementation of Power Law Graph Transformer can be found at \url{https://github.com/burcgokden/Power-Law-Graph-Transformer}. 

\begin{table}[ht]
\caption{Set of model hyperparameters used for training the dataset. The unfilled sections have the same value as model \#1. Models \#1-6 are power law graph transformers. SDPA is transformer with scaled dot product attention.}
\label{table1}
\centering
\resizebox{\textwidth}{!}{

\begin{tabular}{c c c c c c c c }
\toprule[1.2pt] 
Model & \# Layers & \# Heads & A-dff &  \# Res. Dense Layers & \# Res. Units & $d_{LM}$ & dff \\ 
\midrule[1.2pt]
\#1 & 1 & 16 & 128 & 2 & 10 & 512 & 2048 \\  
\midrule 
\#2 &  & 8 & 256 &  & 9 &  & \\
\midrule 
\#3 &  & 8 & 256 &  & 8 &  & \\
\midrule 
\#4 &  & 4 & 512 &  & 5 & & \\ 
\midrule  
\#5 &  & 2 & 1024 &  & 2 & &   \\ 
\midrule 
\#6 &  & 1 & 2048 &  & 1 &  &  \\
\midrule 
SDPA & 4 & 8 & n/a & n/a & n/a & 512 & 2048 \\ 
\bottomrule[1.2pt]
\end{tabular}} 
\end{table}

\section{Results}

\textbf{Evaluation with beam length=1}. The results for inductive task are shown in table \ref{table2} for graph transformers and SDPA transformer model. We ran PT-EN translation task on all model variations for comparison. The model \#2 with 8 heads and 9 residual layers gave the best BLEU score among graph transformer variations. Model \#1 with 16 heads and 10 residual layers exhibited reduced BLEU scores for PT-EN task compared to model \#2 with 8 heads. The reduction of head size and residual layer size have a big impact on the model capacity. Model \#6 with single head and single residual unit had lowest BLEU score of $16.02$ even if the fully connected layers in residual units have the largest number of neurons per layer. This suggests that the number of deep residual connections that learn the metric tensor and number of heads exploring alternate versions of graph manifold are important to represent unstructured data such as language datasets, that usually employ ambiguous relationships between nodes. The SDPA model was also trained using PT-EN dataset and had a slightly better BLEU score of $27.97$ vs $27.79$ for model \#2 variation of the graph transformer. The models \#2 and \#3 trained on larger TR-EN dataset gave similar results of $17.58$ and $17.61$ which was $\sim0.8$ higher than SDPA BLEU score of $16.82$ on the same dataset.

\begin{figure}
\centering
\begin{subfigure}[b]{0.45\textwidth}
	\centering
	\includegraphics[width=1.2\textwidth]{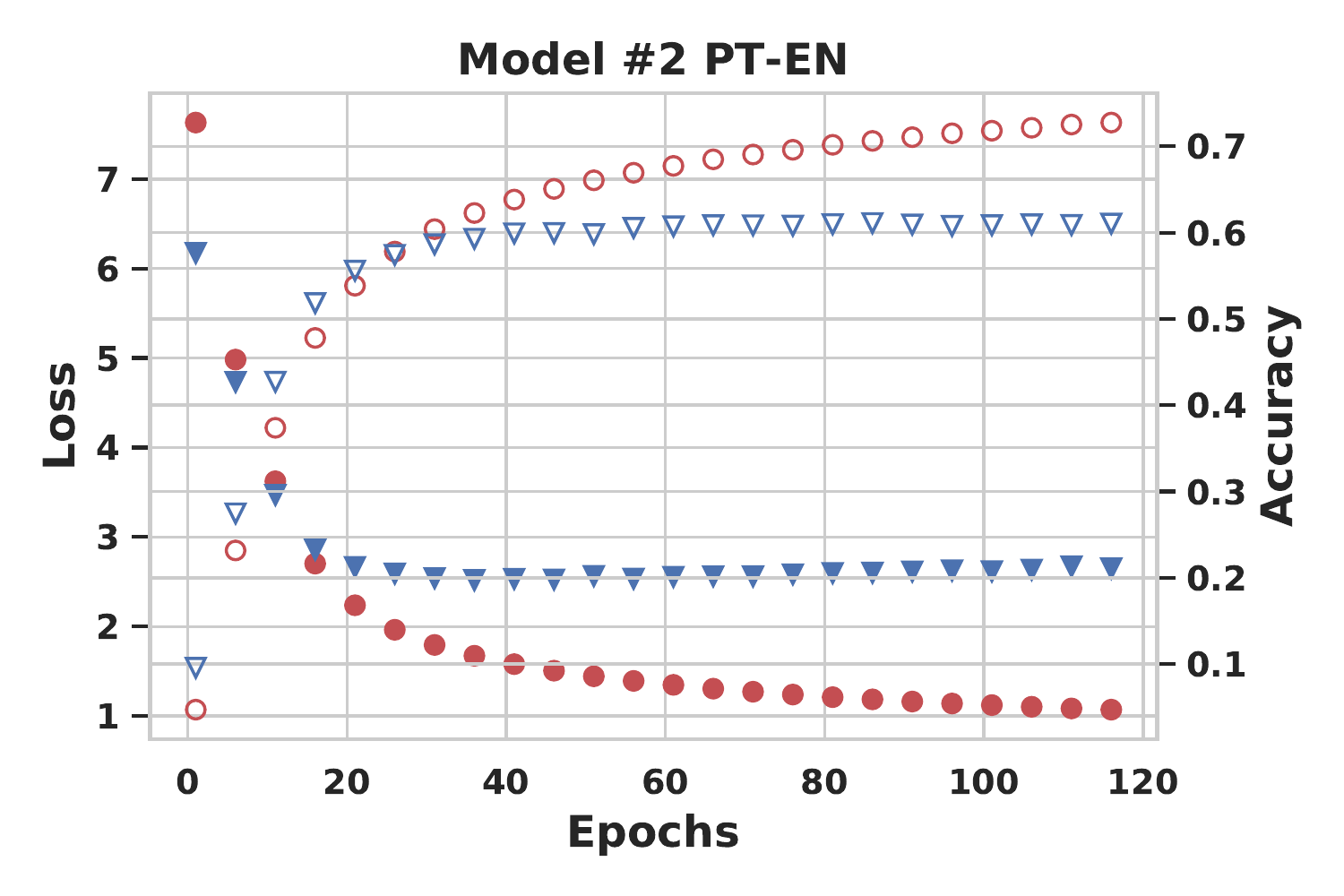}
	\caption{}
	\label{fig3a}
\end{subfigure}
\hfill
\begin{subfigure}[b]{0.45\textwidth}
	\centering
	\includegraphics[width=1.2\textwidth]{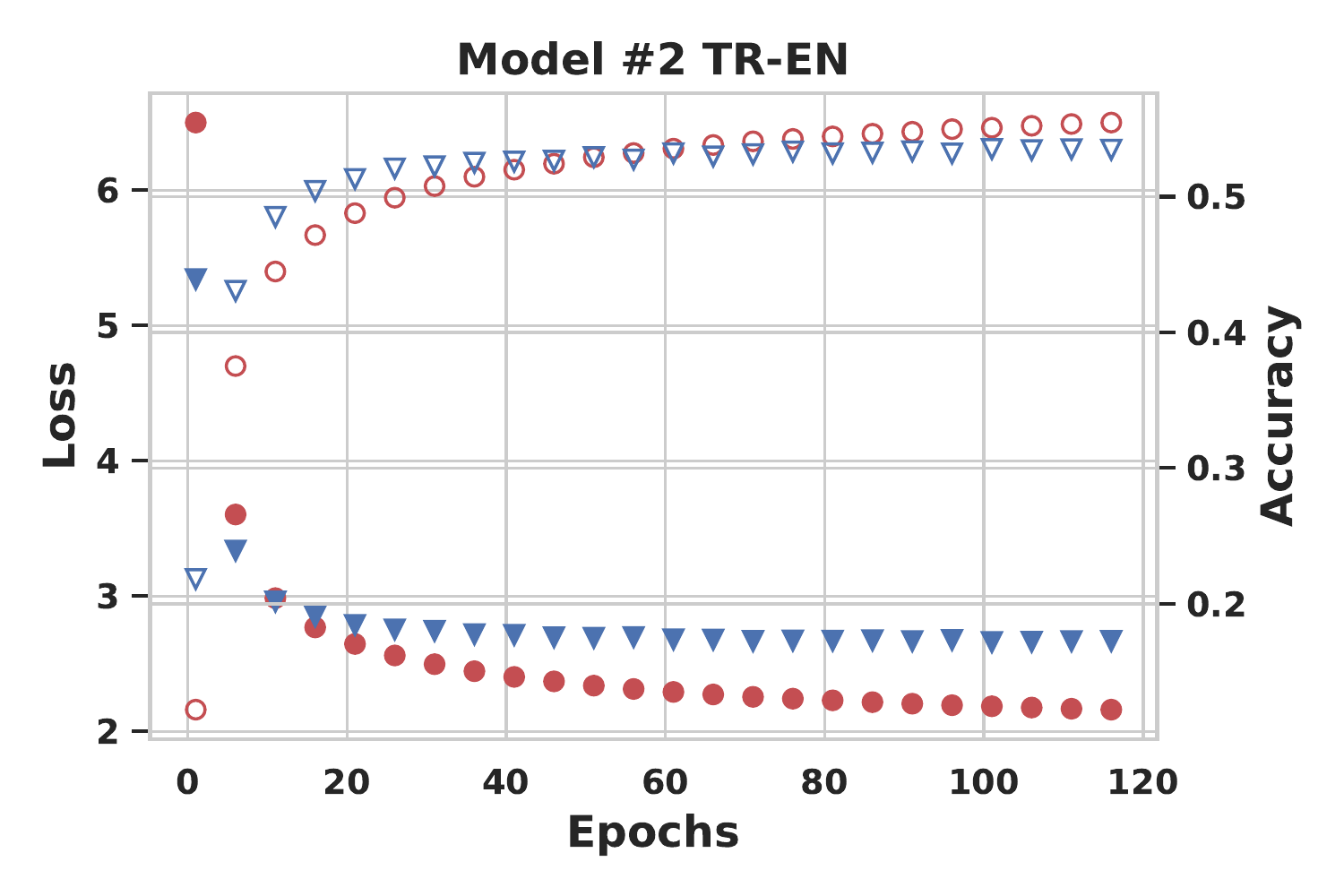}
	\caption{}
	\label{fig3b}
\end{subfigure}
\hfill
\begin{subfigure}[b]{0.45\textwidth}
	\centering
	\includegraphics[width=1.2\textwidth]{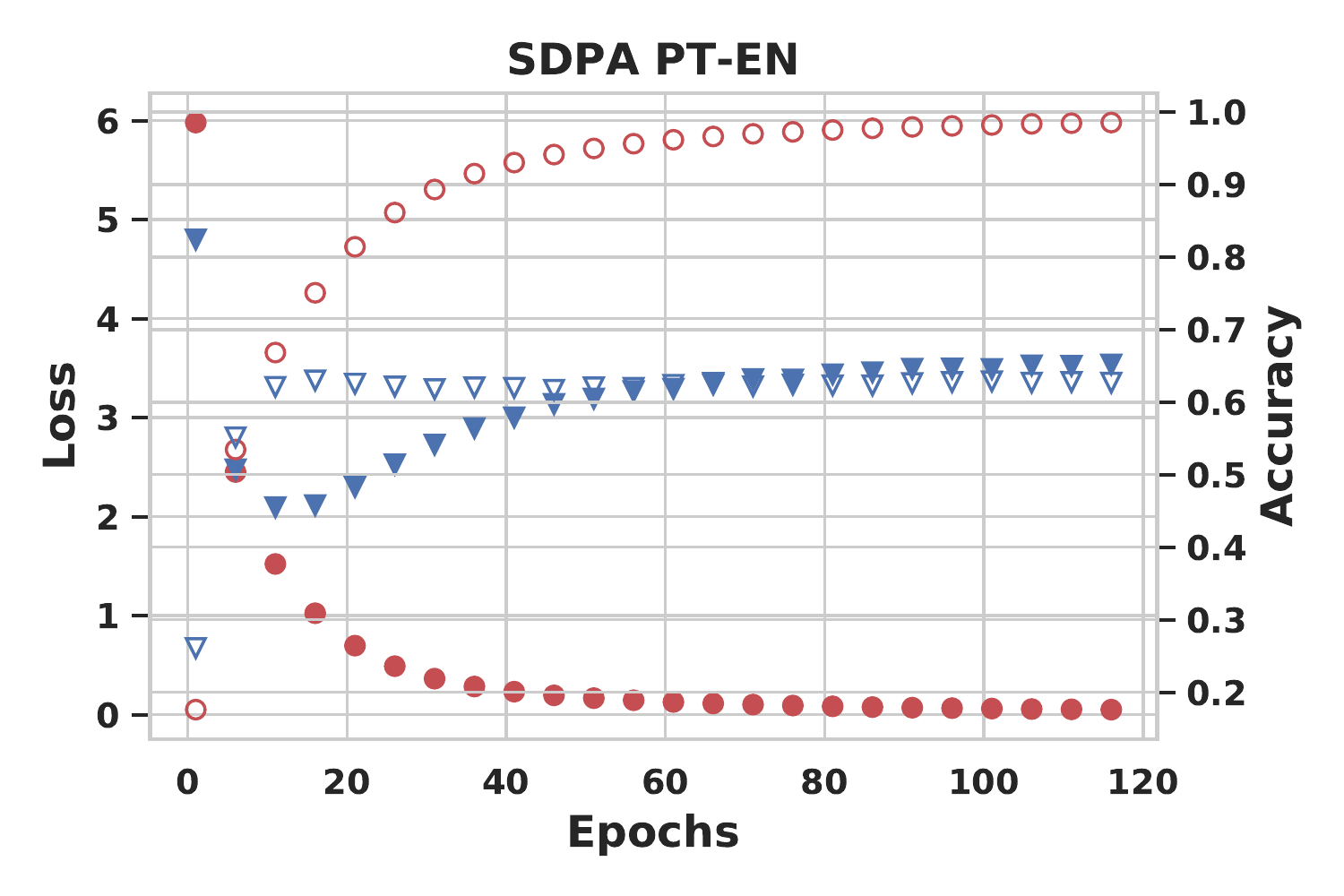}
	\caption{}
	\label{fig3c}
\end{subfigure}
\hfill
\begin{subfigure}[b]{0.45\textwidth}
	\centering
	\includegraphics[width=1.2\textwidth]{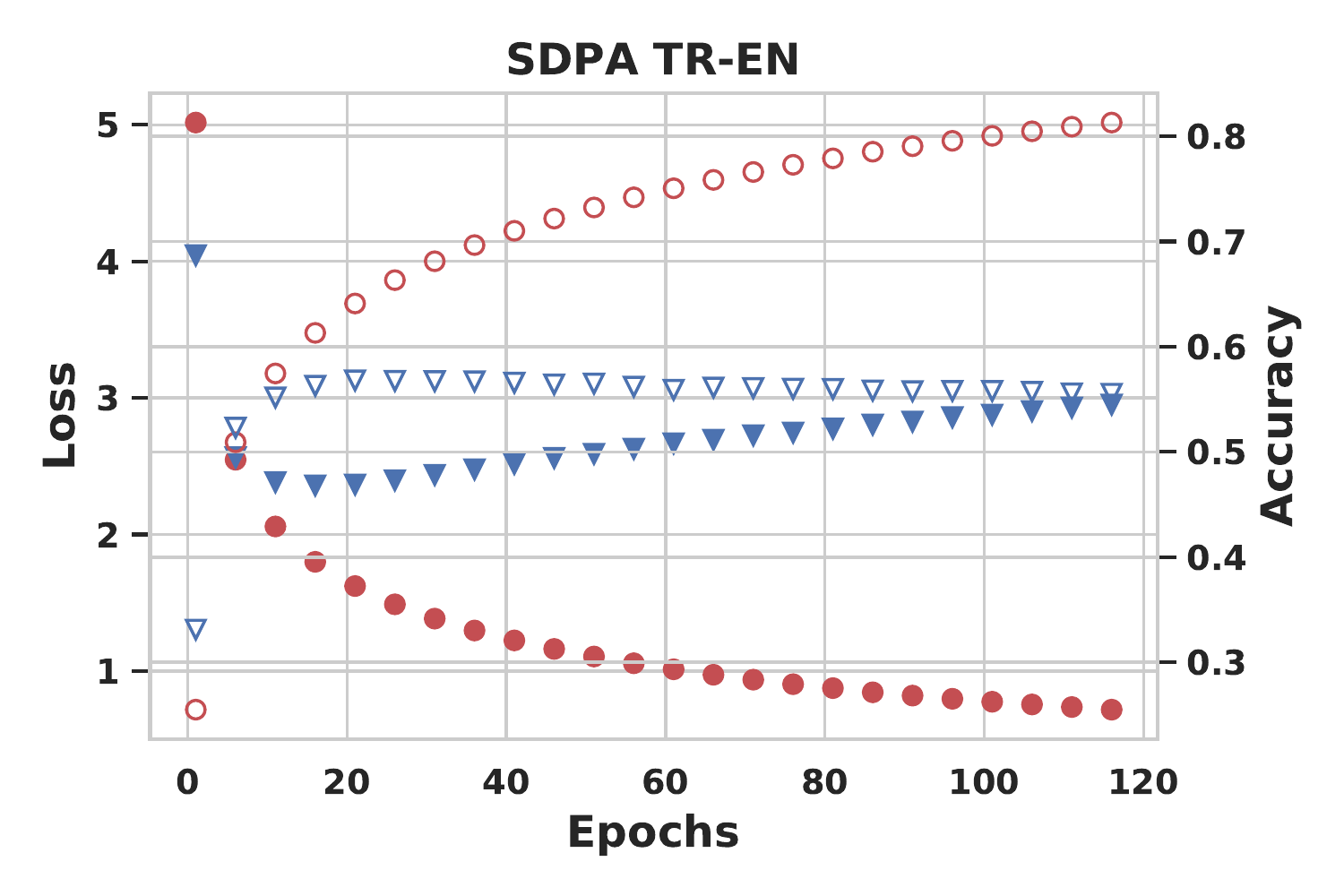}
	\caption{}
	\label{fig3d}
\end{subfigure}
\caption{Loss and accuracy curves for model \#2 (a,b) and SDPA (c,d) architectures trained on PT-EN and TR-EN datasets. Full (hollow) circles in red are for train set loss (accuracy) values and full (hollow) triangles in blue are validation set loss (accuracy) values. For clarity values at every 5 epochs are plotted.}
\label{fig3}
\end{figure}

We also compared the loss and accuracy curves over 120 epochs for PLGA model \#2 and SDPA model. The results are shown in fig. \ref{fig3} for models trained on PT-EN and TR-EN tasks. The SDPA transformer converges to a minimum loss earlier during training. The validation loss curve starts to overfit at longer training times, therefore an early stopping strategy is expected to give the best case BLEU score for SDPA model in this work. For the PLGA model, the overfit is much less and validation and training accuracy have a smaller gap. The best case results and highest accuracy points occur at later epochs for the PLGA model. This suggests that the PLGA and SDPA architectures explore the model space differently.

\noindent\textbf{Evaluation with beam length=4}. We evaluated the model \#2 and SDPA model with highest BLEU scores using beam search at beam length=4 to compare. The results are shown in table \ref{table3}. The PLGA model results in better BLEU score than RNN model \cite{bahdanau2014} with attention  evaluated in \cite{qi2018} for PT-EN and TR-EN tasks with standard (randomly initialized) embeddings. When evaluated at beam length=4, the SDPA model fared better in BLEU score than the PLGA model for PT-EN and TR-EN tasks.

\begin{table}[ht]
\caption{Greedy search BLEU results for models trained using TR-EN and PT-EN datasets for 120 epochs and evaluated at various intervals. Maximum BLEU scores are shown. (HA: model was evaluated at highest validation accuracy.)}
\label{table2}
\centering
\resizebox{0.8\textwidth}{!}{
\begin{tabular}{c c c c c } 
\toprule[1.3pt]
Model & Dataset & BLEU & Log Perplexity & Epoch \\ 
\midrule[1.1pt]  
 \#1 & PT-EN &  26.96 & 2.74 & 110 \\ 
\midrule 
 \#2 & PT-EN & 27.79 & 2.64 & 110(HA)  \\ 
\midrule  
 \#2 & TR-EN  & 17.58 & 2.66  & 118(HA) \\
\midrule  
 \#3 & PT-EN & 27.66 & 2.68 & 120 \\  
\midrule  
 \#3 & TR-EN & 17.61 & 2.67 & 120 \\ 
\midrule  
 \#4 & PT-EN  & 27.55 & 2.64 & 110 \\  
\midrule  
 \#5 & PT-EN  & 16.74 & 3.42 & 80 \\
\midrule  
 \#6 & PT-EN  & 16.02 & 3.40 & 118(HA) \\
\midrule  
 SDPA & PT-EN  & 27.97 & 2.15 & 17(HA) \\
\midrule   
 SDPA & TR-EN  & 16.82 & 2.43 & 30 \\   
\bottomrule[1.2pt]
\end{tabular}} 
\end{table}

\begin{table}[ht]
\caption{BLEU score comparison for PLGA transformer, SDPA transformer and the RNN model trained in ref. \cite{qi2018} using same datasets with standard and pre-trained embeddings. PLGA and SDPA transformers were evaluated using beam length of 4 and 1 (shown in parentheses).}
\label{table3}
\centering
\resizebox{0.6\textwidth}{!}{
\begin{tabular}{c c c c }
\toprule
& & \multicolumn{2}{c}{BLEU} \\
\cline{3-4} \addlinespace
Model & Dataset & $(std-std)$ & $(pre-pre)$ \\
\midrule
\#2 & TR-EN & 17.79(17.58) & -- \\
\midrule
SDPA & TR-EN & 18.31(16.82) & -- \\
\midrule
from \cite{qi2018} & TR-EN & 14.9 & 17.9 \\
\midrule
\#2 & PT-EN & 28.33(27.79) & -- \\
\midrule
SDPA & PT-EN & 29.57(27.97) & -- \\
\midrule
from \cite{qi2018} & PT-EN & 26.2 & 30.8 \\
\bottomrule
\end{tabular}
}
\end{table}

\begin{figure}
\begin{adjustwidth}{-5em}{-5em}
\centering
\begin{subfigure}[b]{0.6\textwidth}
	\centering
	\includegraphics[width=1.1\textwidth]{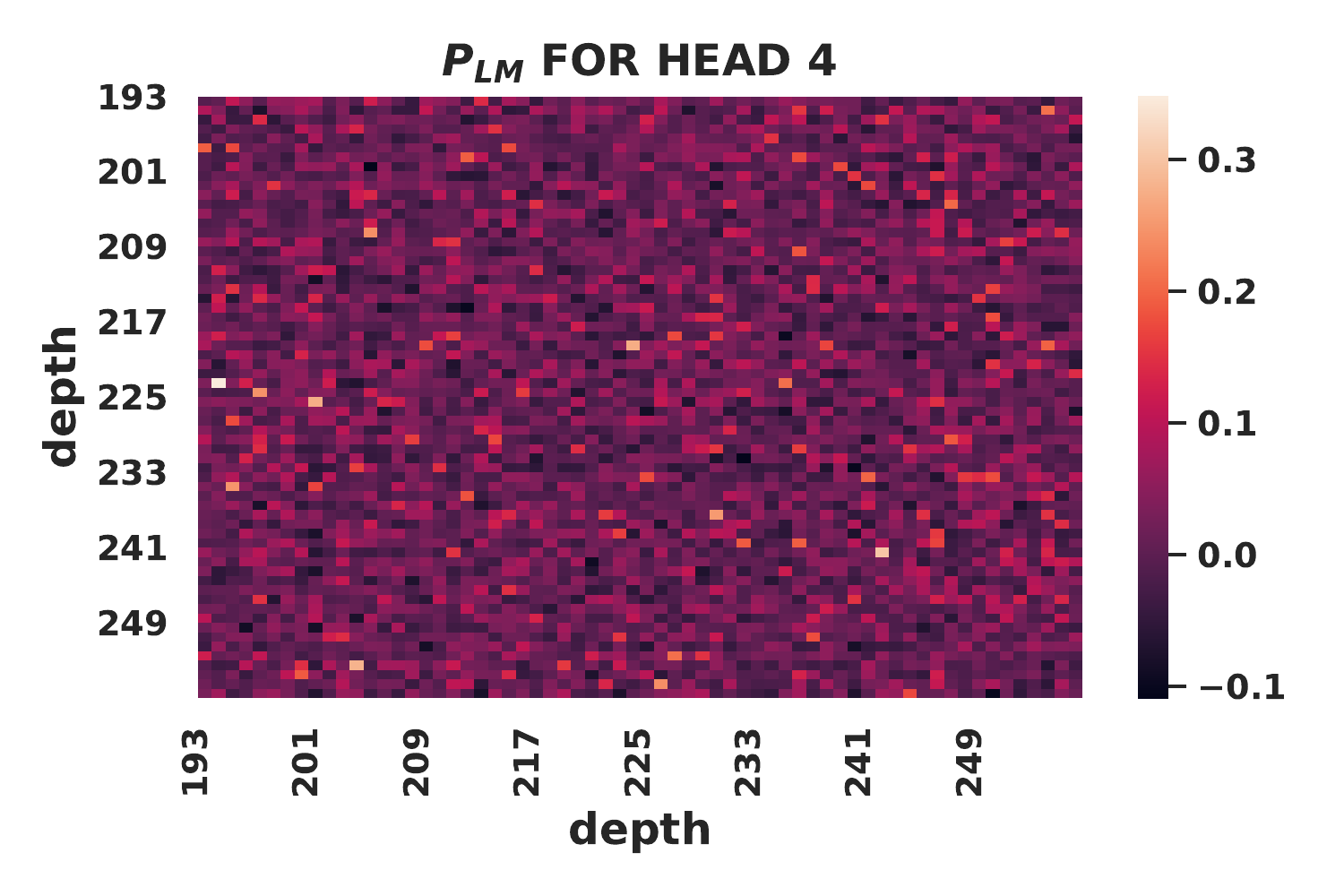}
	\caption{}
	\label{fig4a}
\end{subfigure}
\hfill
\begin{subfigure}[b]{0.6\textwidth}
	\centering
	\includegraphics[width=1.1\textwidth]{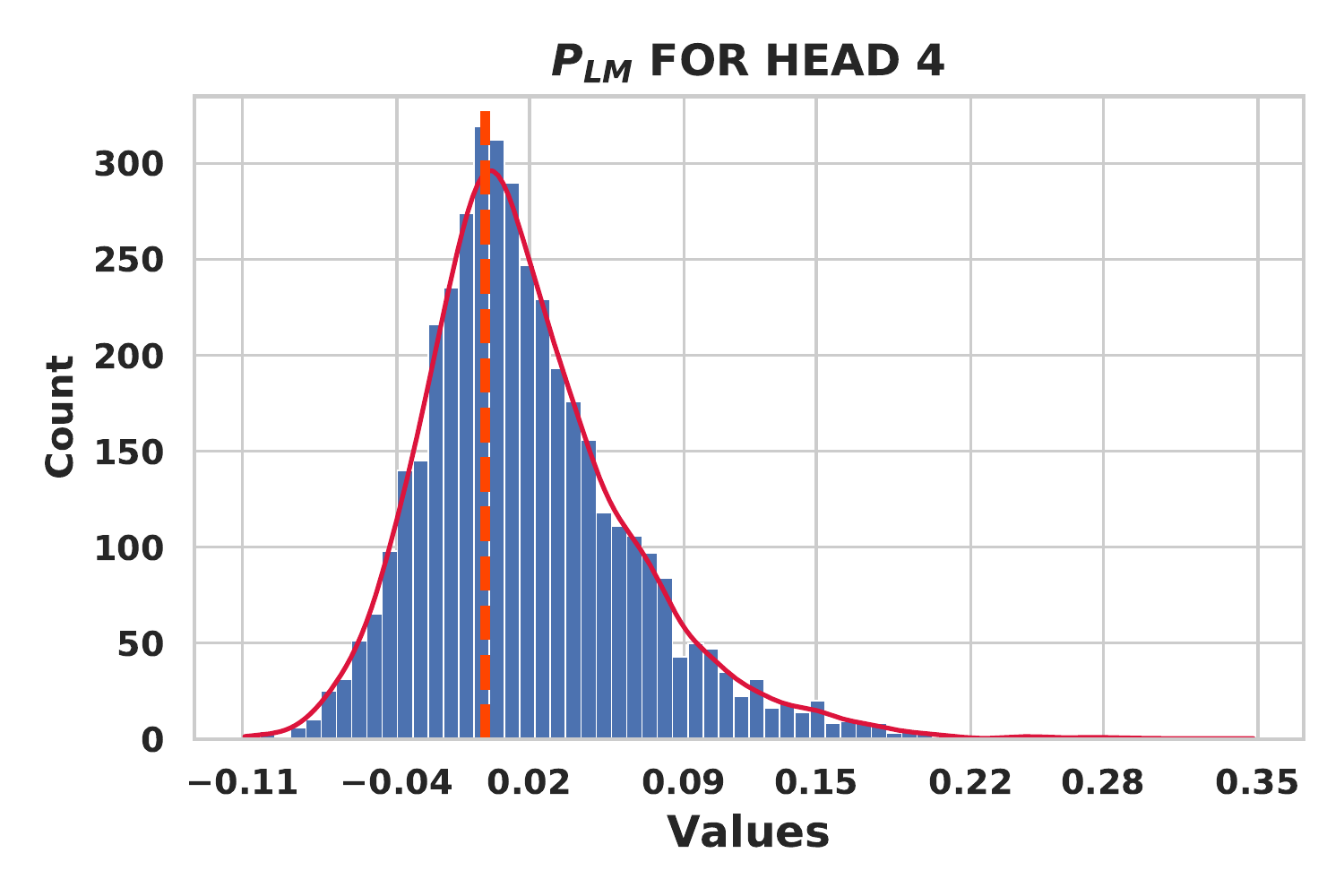}
	\caption{}
	\label{fig4b}
\end{subfigure}
\hfill
\begin{subfigure}[b]{0.6\textwidth}
	\centering
	\includegraphics[width=1.1\textwidth]{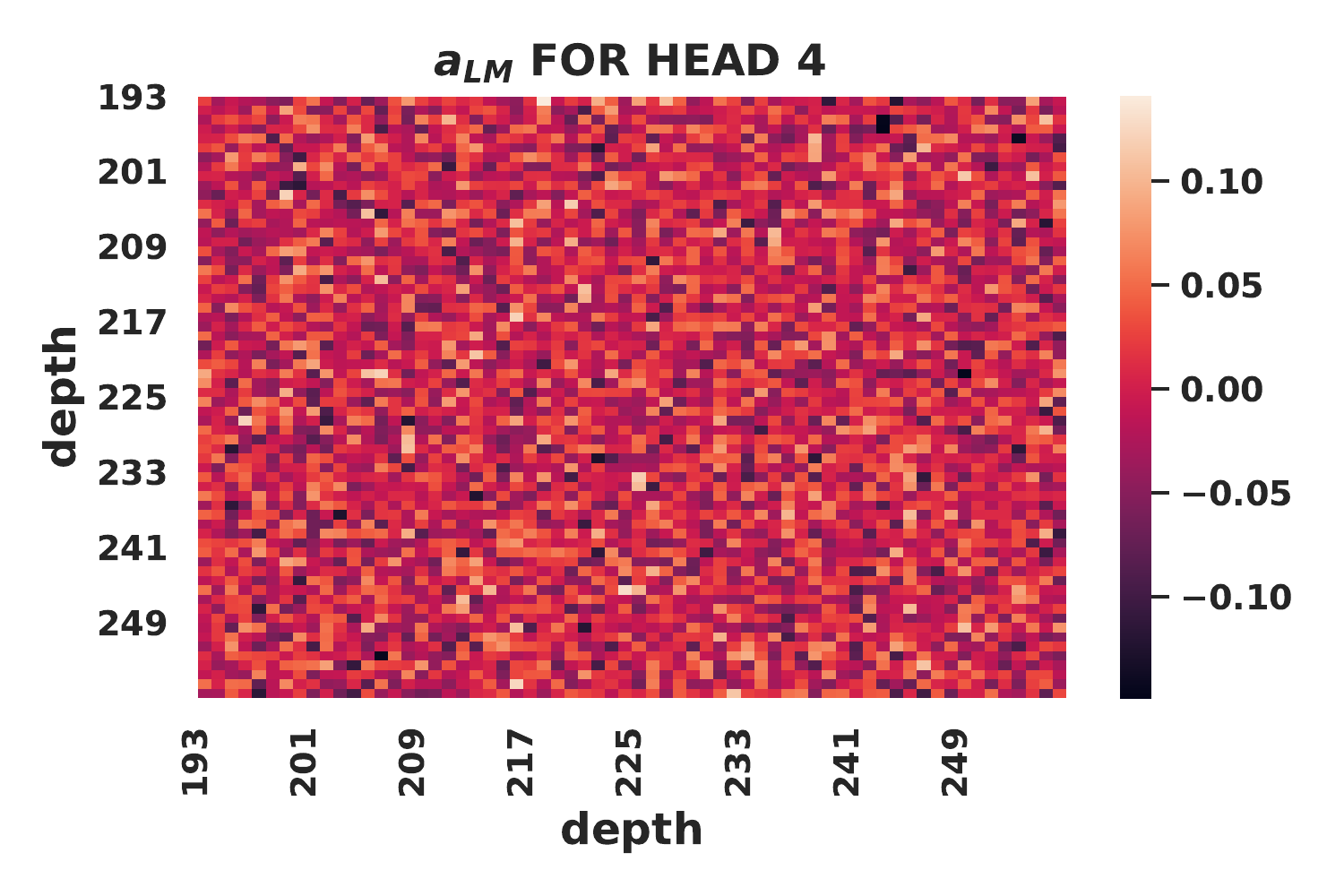}
	\caption{}
	\label{fig4c}
\end{subfigure}
\hfill
\centering
\begin{subfigure}[b]{0.6\textwidth}
	\centering
	\includegraphics[width=1.1\textwidth]{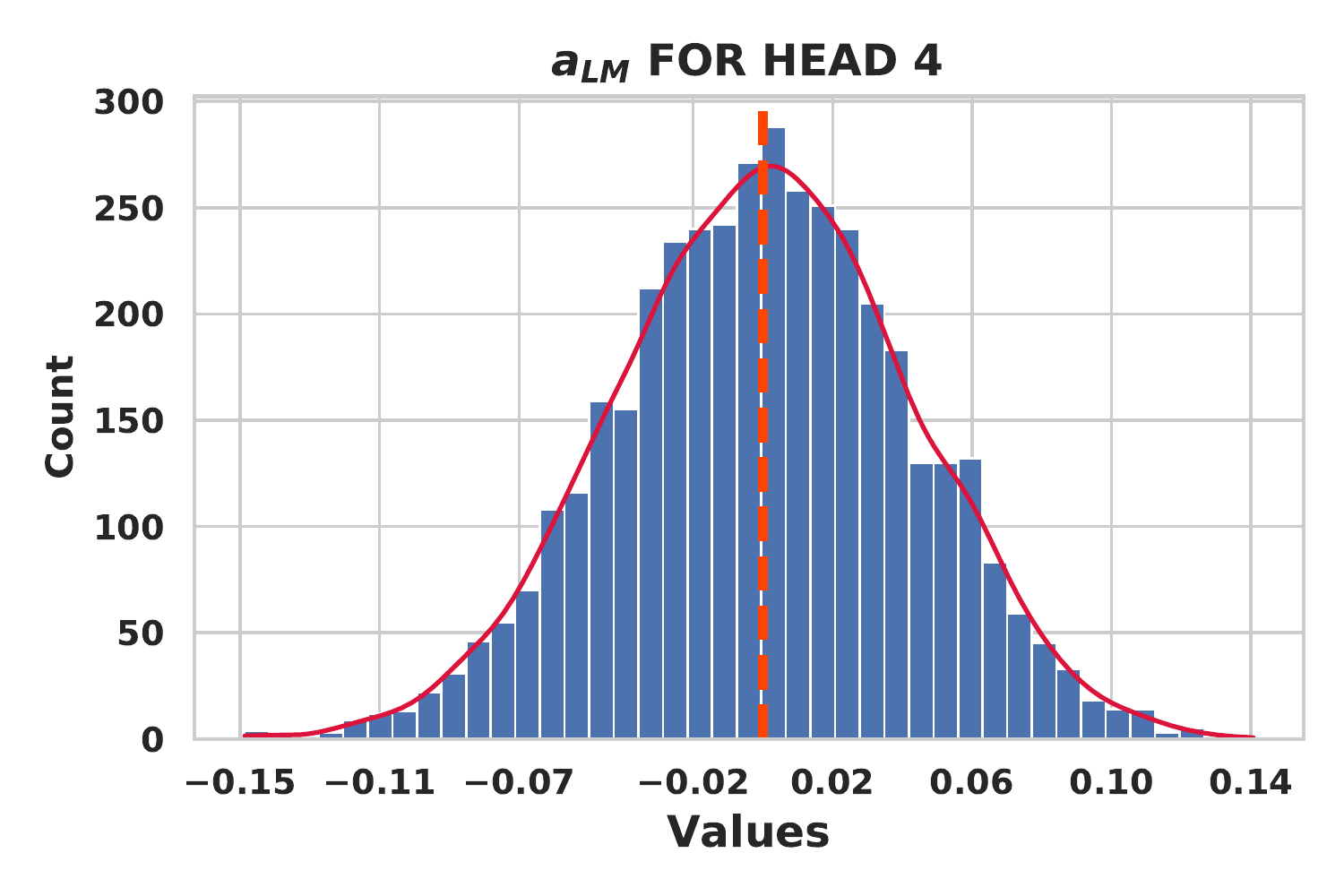}
	\caption{}
	\label{fig4d}
\end{subfigure}
\hfill
\begin{subfigure}[b]{0.6\textwidth}
	\centering
	\includegraphics[width=1.1\textwidth]{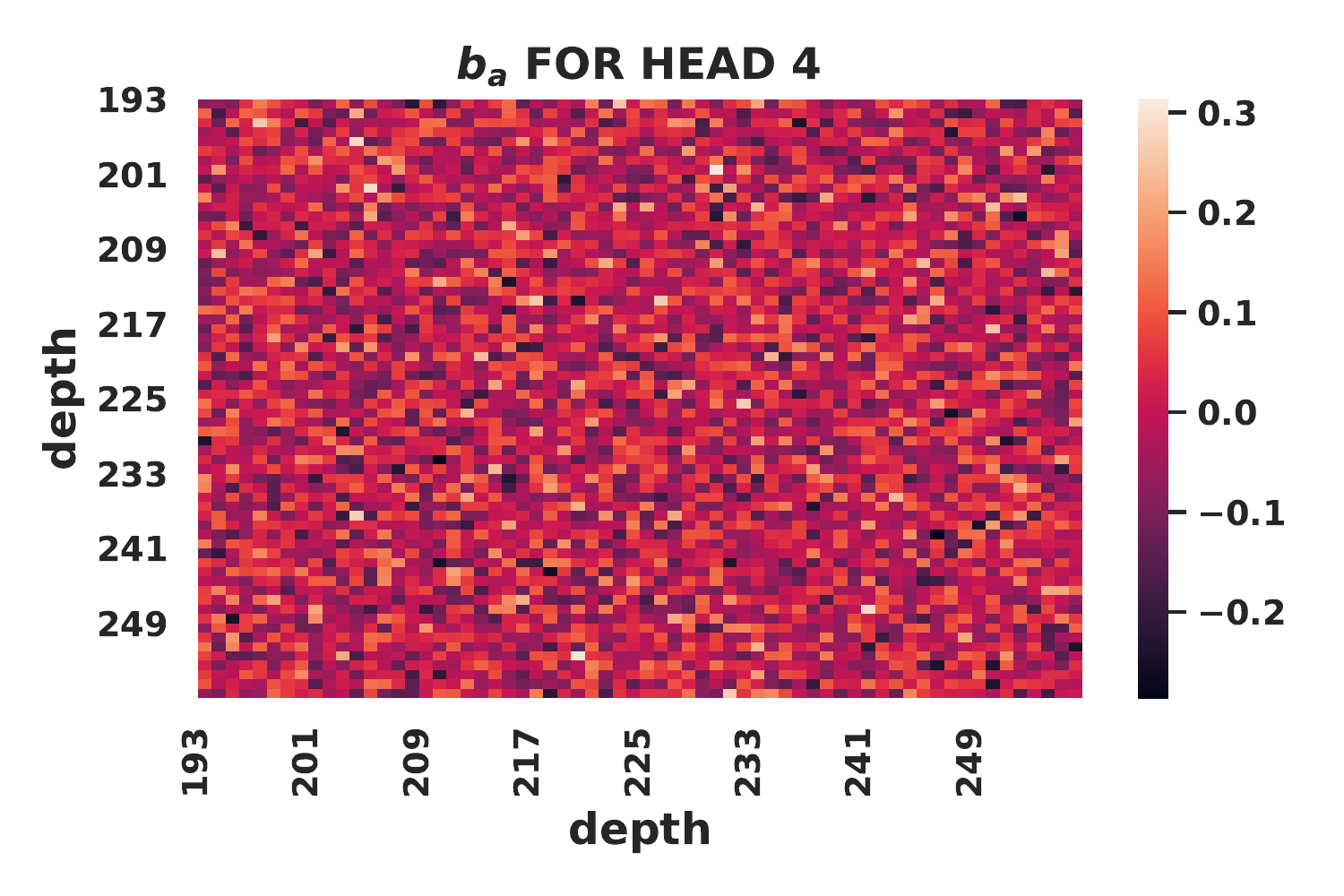}
	\caption{}
	\label{fig4e}
\end{subfigure}
\hfill
\begin{subfigure}[b]{0.6\textwidth}
	\centering
	\includegraphics[width=1.1\textwidth]{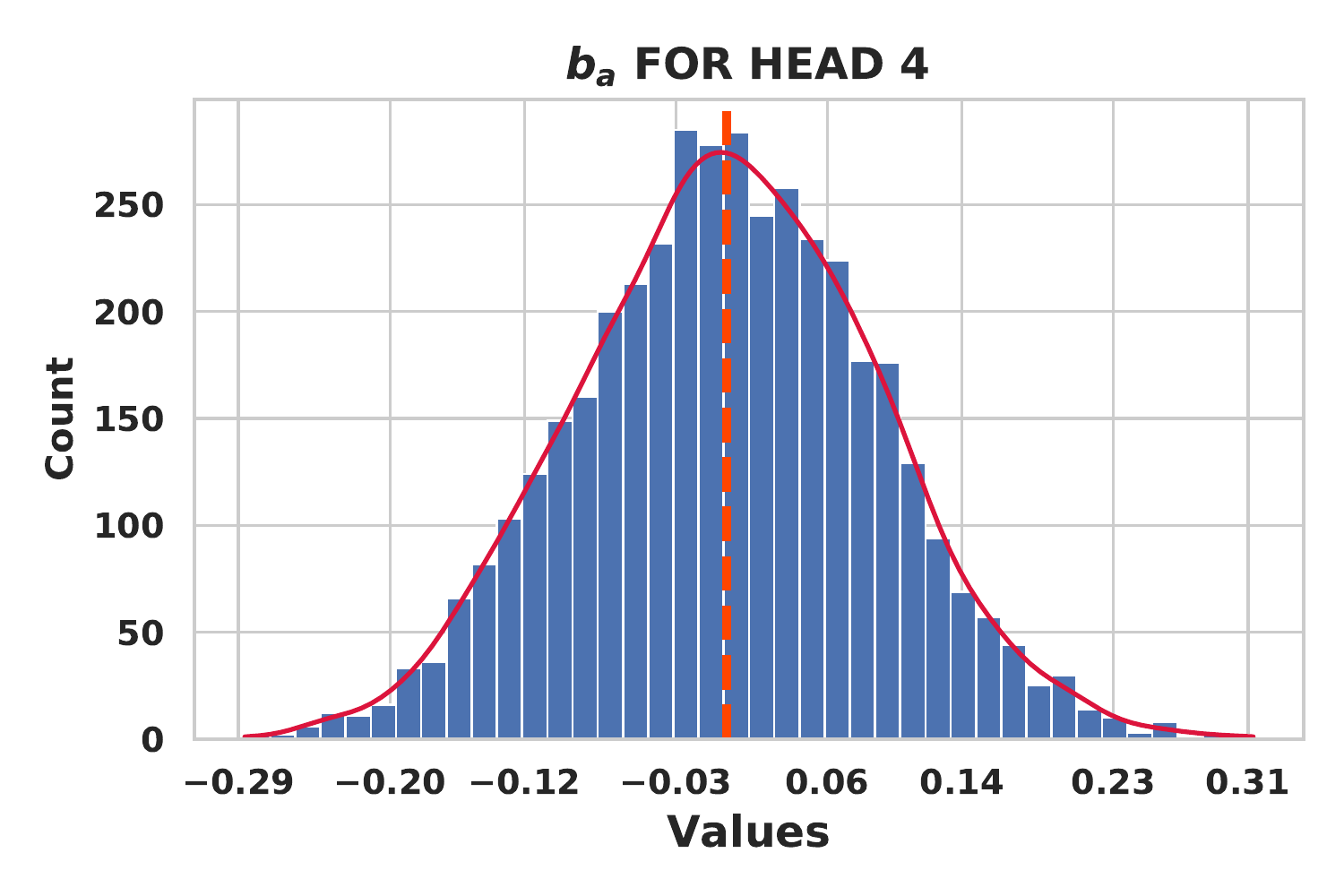}
	\caption{}
	\label{fig4f}
\end{subfigure}
\hfill
\caption{Heatmap and histogram distributions of deductive task outputs for head 4 of XLM Attention from model \#2. (a,b): $\mP_{LM}$, (c,d): $\va_{LM}$, (e,f): $\vb_{a}$. Dashed line in orange marks zero value.}
\label{fig4}
\end{adjustwidth}
\end{figure}
\begin{figure}
\begin{adjustwidth}{-5em}{-5em}
\centering

\begin{subfigure}[b]{0.6\textwidth}
	\centering
	\includegraphics[width=1.1\textwidth]{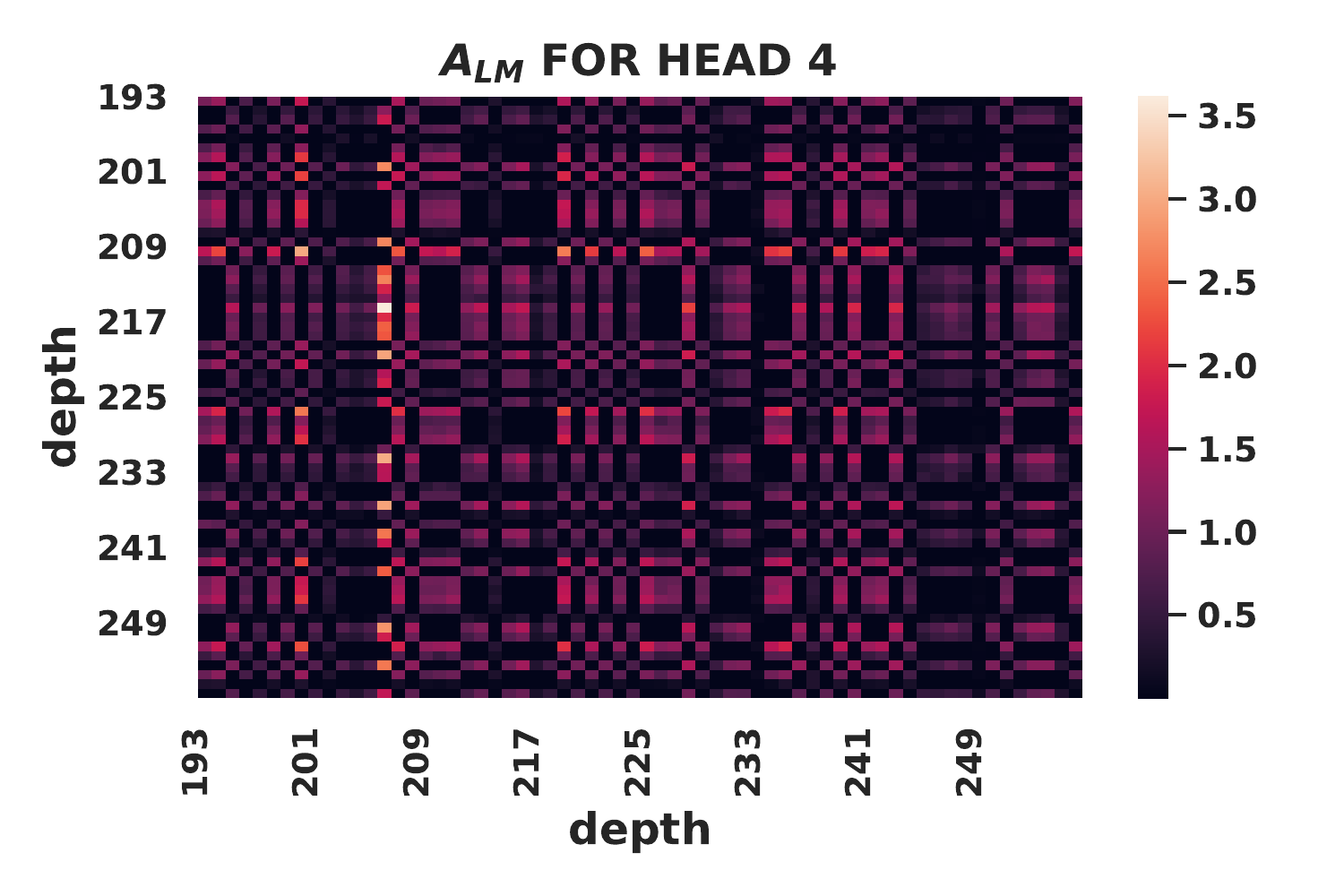}
	\caption{}
	\label{fig5a}
\end{subfigure}
\hfill
\begin{subfigure}[b]{0.6\textwidth}
	\centering
	\includegraphics[width=1.1\textwidth]{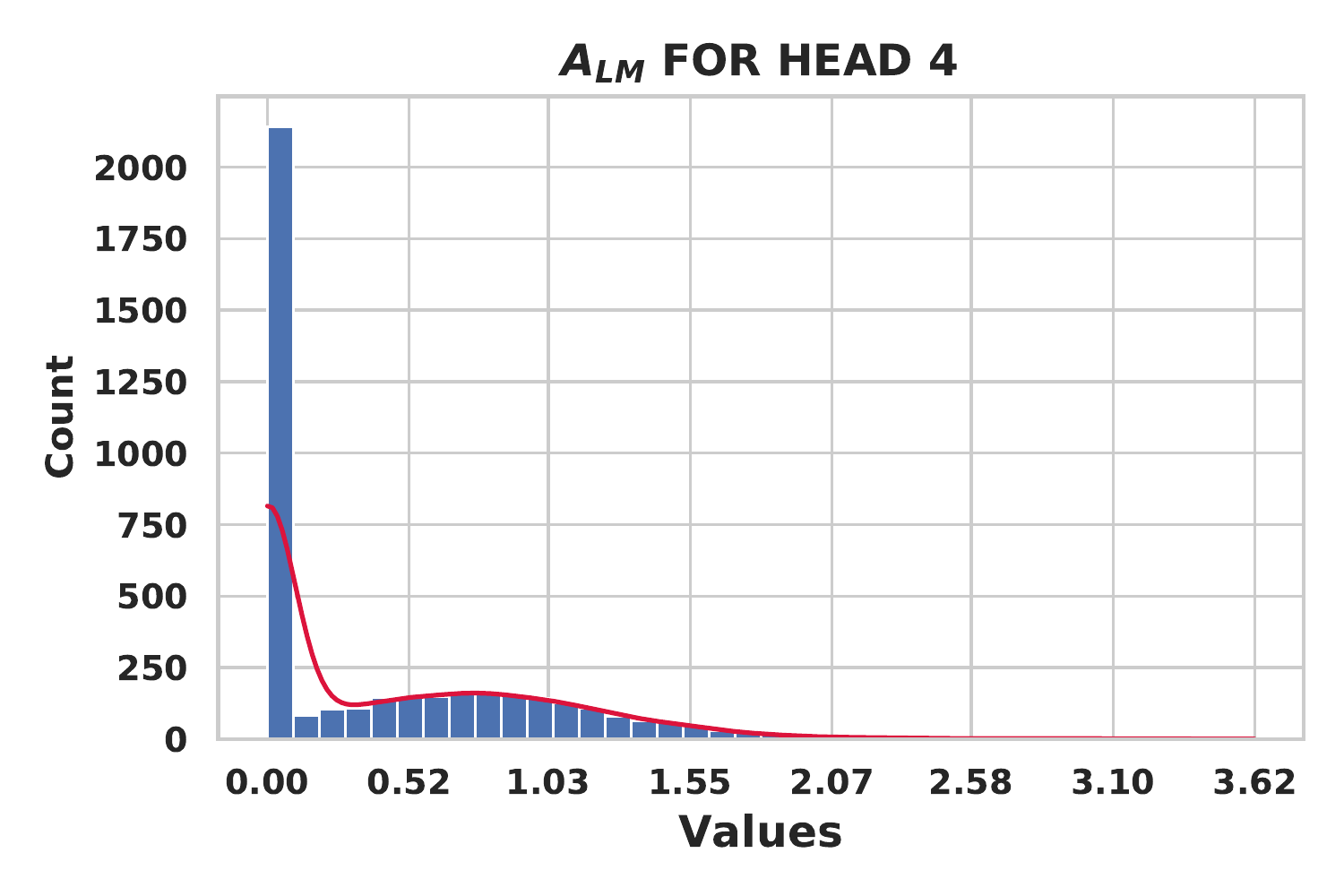}
	\caption{}
	\label{fig5b}
\end{subfigure}
\hfill
\begin{subfigure}[b]{0.6\textwidth}
	\centering
	\includegraphics[width=1.1\textwidth]{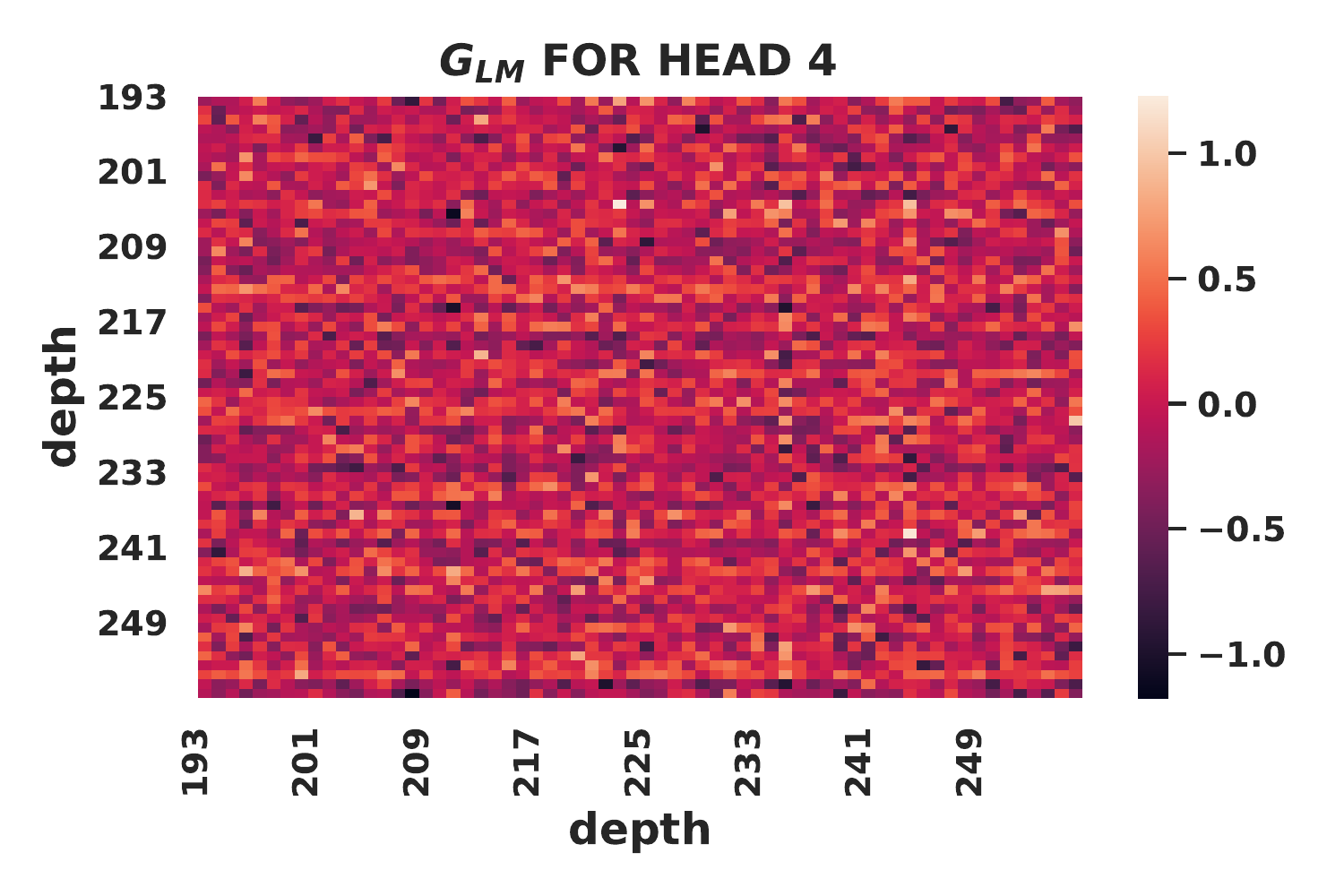}
	\caption{}
	\label{fig5c}
\end{subfigure}
\hfill
\begin{subfigure}[b]{0.6\textwidth}
	\centering
	\includegraphics[width=1.1\textwidth]{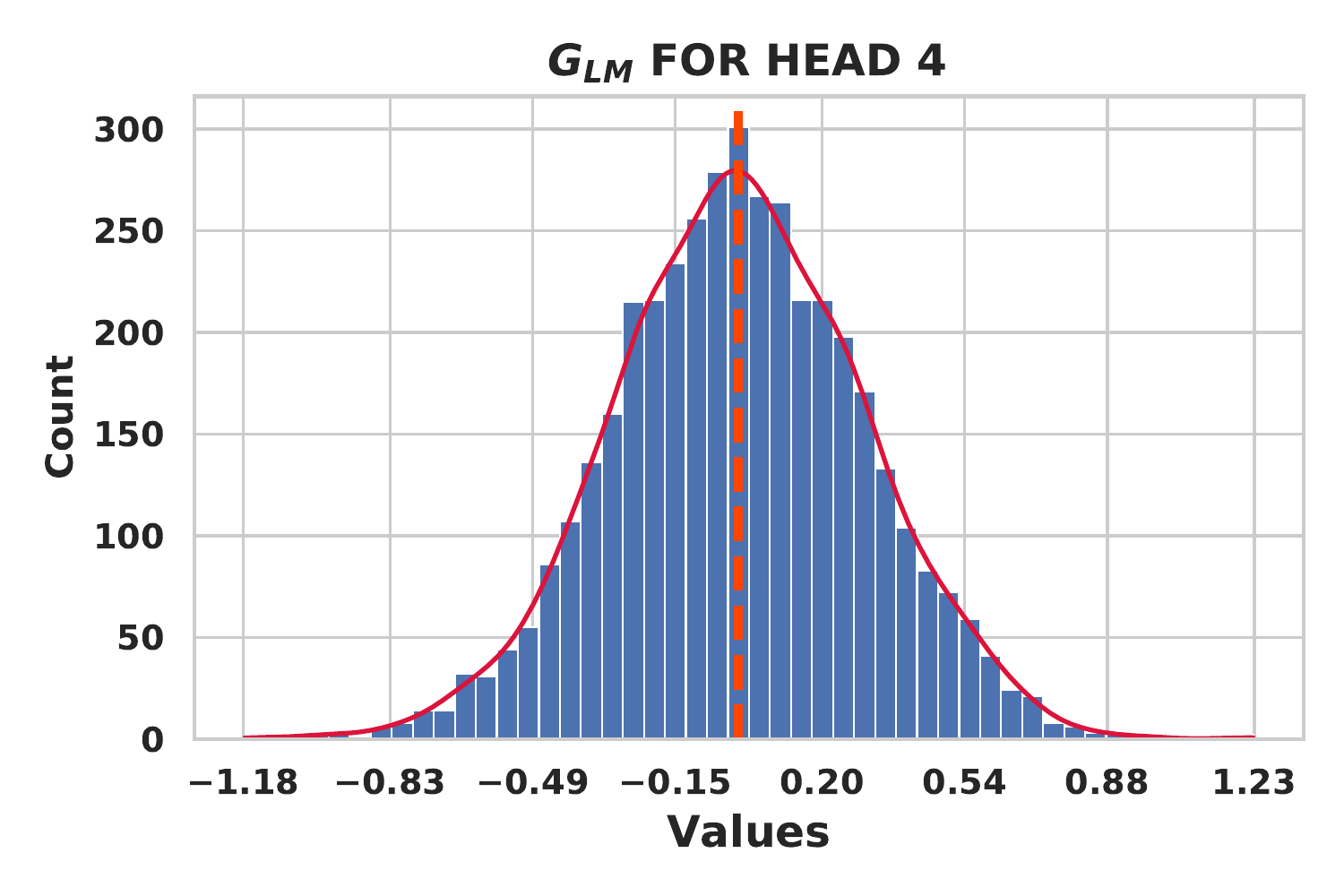}
	\caption{}
	\label{fig5d}
\end{subfigure}
\hfill
\begin{subfigure}[b]{0.6\textwidth}
	\centering
	\includegraphics[width=1.1\textwidth]{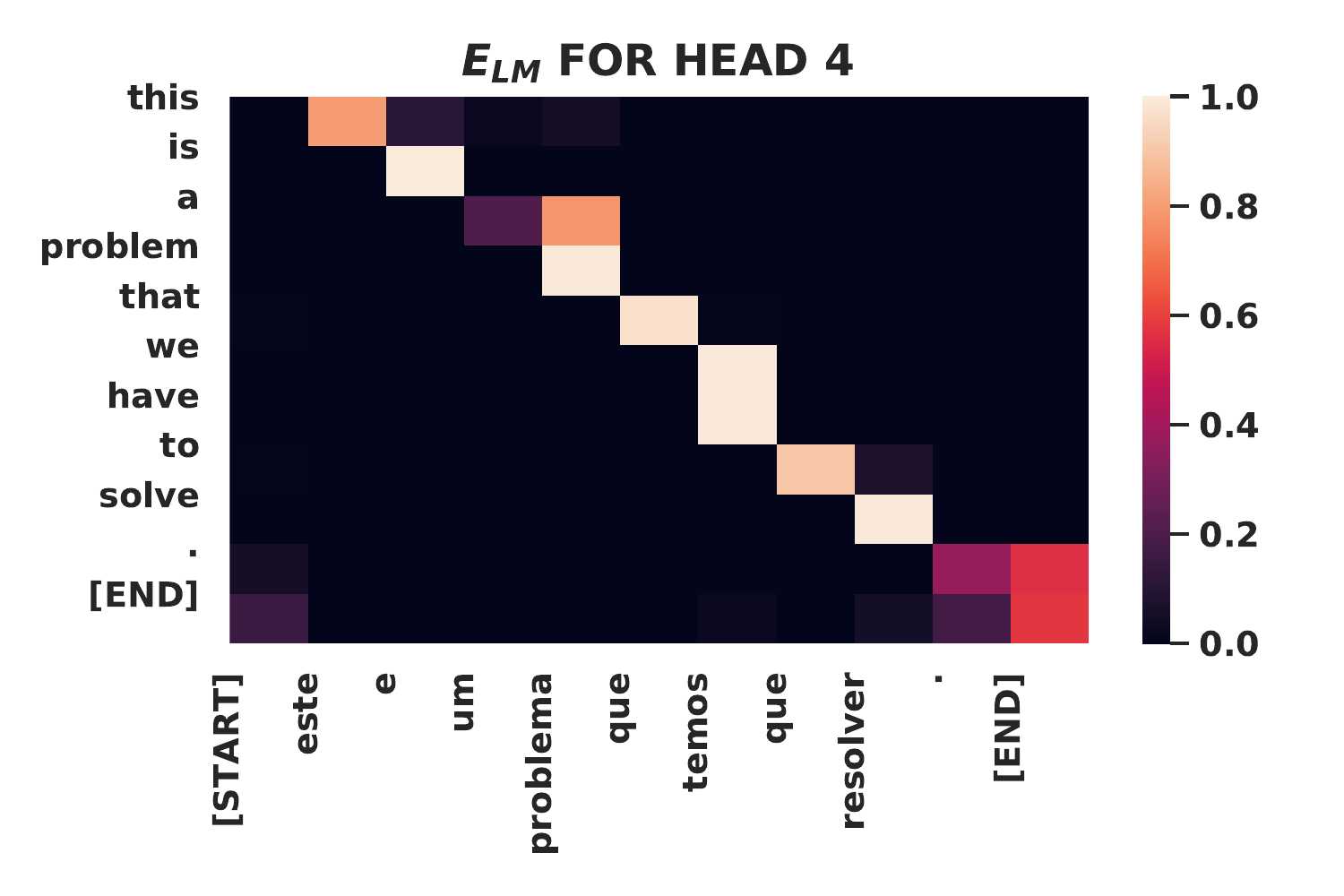} 
	\caption{}
	\label{fig5e}
\end{subfigure}

\caption{Heatmap and histogram distribution of deductive task outputs for head 4 of XLM Attention from model \#2. (a,b): $\mA_{LM}$, (c,d): $\mG_{LM}$, (e): $\mE_{LM}$. Dashed line in orange marks zero value.}
\label{fig5}
\end{adjustwidth}
\end{figure}

For the deductive task of the model, we analyzed the 2D heatmap and histogram distributions of the set $ ( \mE_{LM}, \mP_{LM}, \va_{LM}, \vb_{a}, \mA_{LM}, \mG_{LM} )$. Out of these parameters, $(\mP_{LM}, \va_{LM}, \vb_{a})$ are learned for the entire dataset and are generalized for the language model. The rest of the outputs  $(\mE_{LM}, \mA_{LM}, \mG_{LM})$ are inferred instances for an input sentence (graph instance). We show in figs. \ref{fig4} and \ref{fig5}, the heatmaps and histograms from head $4$ of last attention stage (X-LM attention) of model \#2 trained using PT-EN dataset and evaluated with greedy search. The outputs from all heads for X-LM, source LM and target LM attention models are included in the appendix. Following input sentence from PT-EN dataset was evaluated to generate the deductive task outputs:

\begin{quote}
\begin{tabular}{ l l l }
 Input       &:& ``este é um problema que temos que resolver ."\\
 Prediction  &:& ``this is a problem that we have to solve ."\\
 Ground truth&:& ``this is a problem we have to solve ."
 \end{tabular}
\end{quote}

The heatmaps for $(\mP_{LM}, \va_{LM}, \vb_{a})$ show approximately gaussian distribution without any clear pattern visible. The histogram distribution for $\mP_{LM}$ (fig. \ref{fig4b}) has slightly longer tail above zero. The coefficients $\va_{LM}$ and  bias $\vb_a$ are less skewed and more broadly distributed around zero value. $\vb_a$ profile indicates that the model will have a non-zero background distribution if the metric tensor for an input instance were all zero.

The metric tensor $\mA_{LM}$ (fig. \ref{fig5a}) for input sentence is positive definite and heat map is intriguingly not approximately gaussian, where the dark regions close to zero and non-zero ``active" regions are grouped and connected similar to a loosely knit straw basket pattern. The histogram shows a long tail distribution with a sharp peak at number of values close to zero.  This indicates that the metric tensor is still sparse for $d_{emb}=512$ chosen as embedding feature dimension in our models. The EC tensor $\mG_{LM}$ is shown in fig. \ref{fig5c} and was derived using eq. \ref{eq3} in the attention model. Unlike $\mA_{LM}$, the EC tensor distribution has no clear pattern on the heatmap and is approximately gaussian and fairly centered around zero. Fig. \ref{fig5e} shows the attention weights between source and predicted target sentences which is similar to attention weights observed from other attention based models \cite{bahdanau2014, vasvani2017}. We observe that the range of attention weights for $\mE_{LM}$ varies on a scale from 0 to 1. The deductive output values are well behaved, given that the input is not normalized and learned parameters are randomly initialized. We suspect that the layer normalization and positive-definite $\mA_{LM}$ condition applied in eq. \ref{eq2} help improve the interpretability of the deductive task outputs. 

Other head outputs show similar distributions for heatmap and histogram we presented here for X-LM attention outputs. For source and target attention blocks, the attention weights $\mE_{LM}$ show more varied patterns among heads.

\section{Discussion and Further Work}

We can make several key observations about the graph transformer architecture explained in this work. The model uses a quantization set (subword Vocabulary generated from dataset) and its dense vector representation for each element in the set (embedding space vectors with $d_{emb}$ feature dimensions) for linear transformations. A language model manifold (defined by $d_{LM}$ feature dimensions) is  obtained by non-linear transformations through a deep neural network that learns from a large ensemble of local instances of sentences (graph instances). The manifold and quantization set define a duality where we can statistically build global relationships from a large ensemble of local instances and similarly infer single local instances from global relationships. The deductive task builds the global relationships by learning $(\mP_{LM}, \va_{LM}, \vb_{a})$ which are parameters characterizing the relationships for the entire language model. These parameters are not associated with a single sentence but language defined by the entire dataset available to the model, unlike attention weights that are obtained from a SDPA transformer model for a single sentence. The inductive task builds the localized relationships by propagating through $(\mE_{LM}, \mA_{LM}, \mG_{LM})$ from a single instance of input sentence where the output is same as that of a transductive transformer model. The duality transformation between local and global relationships is highly non-linear, defined by a deep residual network. On the other hand, we take advantage of linear transformations whenever local inferences were made from an input instance. 

The number of heads also have a large impact on locality condition besides model capacity. The multi-head configuration allows the model to consider multiple interpretations of an input sentence in sub-spaces. By splitting the input into smaller size heads, we also enforce the model to consider feature dimensions in the same head split to interact more closely. Therefore, a constraint on locality is also introduced within each head split.

Several possible configurations were not explored in this study due to limitations of scope and our experimental setup. The model architecture can be also explored for deeper residual networks with less head splits and multi-layer encoder-decoder networks. The number of embedding dimensions is another hyperparameter that could have an impact on both deductive and inductive task outputs. The hyperparameters we used in this study were optimizations that were reported to work well for the SDPA transformer model in the literature. A better hyperparameter set optimized for graph transformer architecture could improve the BLEU score of this model further. Other optimizations reported for large scale NLP systems \cite{britz2017, wu2016} could be used to scale graph transformer to larger datasets and multi-GPU setups. The deductive outputs provide a rich set of statistical information of the language model and neural network itself. A more in depth analysis of the deductive outputs can provide better understanding of the dataset domain and model architecture.

\section{Conclusion}

We presented a generalized power law graph transformer architecture with well defined deductive and inductive tasks. The deductive task learns the global characteristics of the dataset using a power law attention model. The inductive task uses the global characteristics to predict the output probabilities for an input instance through encoder-decoder architecture. We applied our model for TR-EN and PT-EN machine translation tasks and compared its performance and characteristics to a SDPA transformer model evaluated on same experimental setup. The graph transformer developed in this work used many of the optimizations that SDPA transformer was shown to benefit from in the literature and we believe that hyperparameters better optimized for graph transformer architecture can result in higher BLEU scores.

Our model empirically takes advantage of a duality between a subword Vocabulary represented by $d_{emb}$ embedding feature dimensions and a language model represented by $d_{LM}$ feature dimensions to define local and global statistics from a machine translation dataset. While a single instance of a sentence can be considered as a graph instance exploring a local region of the language model manifold, a large ensemble of such localized instances can be used to learn an abstract, statistical representation for the entire manifold.

In more general terms, graph samples generated from a linear quantization set are used to build a statistical representation for a non-linear manifold using deep residual networks and attention based on a power-law relationship. The power law relationship is inherently scale invariant and we expect that it will be particularly interesting to apply the model to datasets with varying scale and features from domains beyond NLP tasks such as graph databases, communication networks, and many-body problems in quantum mechanics and astronomy.

\subsubsection*{Acknowledgments}

I thank my parents for their support and patience. I would like to acknowledge numerous contributions of machine learning research community on NLP tasks and graph networks in recent years. This research was conducted independently without support from a grant or corporation.

\bibliography{plgt_paper}

\begin{thebibliography}{10}

\bibitem{hinton1986}
G.~E. Hinton, J.~L. Mcclelland, and D.~E. Rumelhart, ``Distributed
  representations,'' in {\em Parallel Distributed Processing: Explorations in
  the Microstructure of Cognition, {V}olume 1: {F}oundations} (D.~E. Rumelhart
  and J.~L. Mcclelland, eds.), pp.~77--109, Cambridge, MA: MIT Press, 1986.

\bibitem{bengio2003}
Y.~Bengio, R.~Ducharme, P.~Vincent, and C.~Janvin, ``A neural probabilistic
  language model,'' {\em J. Mach. Learn. Res.}, vol.~3, p.~1137–1155, Mar.
  2003.

\bibitem{bahdanau2014}
D.~Bahdanau, K.~Cho, and Y.~Bengio, ``Neural machine translation by jointly
  learning to align and translate,'' 2014.
\newblock cite arxiv:1409.0473Comment: Accepted at ICLR 2015 as oral
  presentation.

\bibitem{vasvani2017}
A.~Vaswani, N.~Shazeer, N.~Parmar, J.~Uszkoreit, L.~Jones, A.~N. Gomez,
  u.~Kaiser, and I.~Polosukhin, ``Attention is all you need,'' in {\em
  Proceedings of the 31st International Conference on Neural Information
  Processing Systems}, NIPS'17, (Red Hook, NY, USA), pp.~6000--6010, Curran
  Associates Inc., 2017.

\bibitem{burc2019}
B.~Gokden, ``Coulgat: An experiment on interpretability of graph attention
  networks,'' {\em CoRR}, vol.~abs/1912.08409, 2019.

\bibitem{qi2018}
Y.~Qi, D.~Sachan, M.~Felix, S.~Padmanabhan, and G.~Neubig, ``When and why are
  pre-trained word embeddings useful for neural machine translation?,'' in {\em
  Proceedings of the 2018 Conference of the North {A}merican Chapter of the
  Association for Computational Linguistics: Human Language Technologies,
  Volume 2 (Short Papers)}, (New Orleans, Louisiana), pp.~529--535, Association
  for Computational Linguistics, June 2018.

\bibitem{tedhrlrtranslate}
``Tensorflow dataset for ted talk transcripts.''
  \url{https://www.tensorflow.org/datasets/catalog/ted_hrlr_translate}.

\bibitem{mikolov2013a}
T.~Mikolov, K.~Chen, G.~Corrado, and J.~Dean, ``Efficient estimation of word
  representations in vector space,'' {\em CoRR}, vol.~abs/1301.3781, 2013.

\bibitem{mikolov2013b}
T.~Mikolov, I.~Sutskever, K.~Chen, G.~S. Corrado, and J.~Dean, ``Distributed
  representations of words and phrases and their compositionality,'' in {\em
  NIPS}, pp.~3111--3119, Curran Associates, Inc., 2013.

\bibitem{pennington2014}
J.~Pennington, R.~Socher, and C.~D. Manning, ``Glove: Global vectors for word
  representation.,'' in {\em EMNLP}, vol.~14, pp.~1532--1543, 2014.

\bibitem{sutskever2014}
I.~Sutskever, O.~Vinyals, and Q.~V. Le, ``Sequence to sequence learning with
  neural networks,'' in {\em Advances in neural information processing
  systems}, pp.~3104--3112, 2014.

\bibitem{cho2014}
K.~Cho, B.~van Merri{\"e}nboer, C.~Gulcehre, D.~Bahdanau, F.~Bougares,
  H.~Schwenk, and Y.~Bengio, ``Learning phrase representations using {RNN}
  encoder{--}decoder for statistical machine translation,'' in {\em Proceedings
  of the 2014 Conference on Empirical Methods in Natural Language Processing
  ({EMNLP})}, (Doha, Qatar), pp.~1724--1734, Association for Computational
  Linguistics, Oct. 2014.

\bibitem{luong2015}
T.~Luong, H.~Pham, and C.~D. Manning, ``Effective approaches to attention-based
  neural machine translation,'' in {\em Proceedings of the 2015 Conference on
  Empirical Methods in Natural Language Processing}, (Lisbon, Portugal),
  pp.~1412--1421, Association for Computational Linguistics, Sept. 2015.

\bibitem{lin2017}
Z.~Lin, M.~Feng, C.~N. dos Santos, M.~Yu, B.~Xiang, B.~Zhou, and Y.~Bengio, ``A
  structured self-attentive sentence embedding,'' {\em CoRR},
  vol.~abs/1703.03130, 2017.

\bibitem{devlin2019}
J.~Devlin, M.-W. Chang, K.~Lee, and K.~Toutanova, ``{BERT}: Pre-training of
  deep bidirectional transformers for language understanding,'' in {\em
  Proceedings of the 2019 Conference of the North {A}merican Chapter of the
  Association for Computational Linguistics: Human Language Technologies,
  Volume 1 (Long and Short Papers)}, (Minneapolis, Minnesota), pp.~4171--4186,
  Association for Computational Linguistics, June 2019.

\bibitem{brown2020}
T.~B. Brown, B.~Mann, N.~Ryder, M.~Subbiah, J.~Kaplan, P.~Dhariwal,
  A.~Neelakantan, P.~Shyam, G.~Sastry, A.~Askell, S.~Agarwal, A.~Herbert-Voss,
  G.~Krueger, T.~Henighan, R.~Child, A.~Ramesh, D.~M. Ziegler, J.~Wu,
  C.~Winter, C.~Hesse, M.~Chen, E.~Sigler, M.~Litwin, S.~Gray, B.~Chess,
  J.~Clark, C.~Berner, S.~McCandlish, A.~Radford, I.~Sutskever, and D.~Amodei,
  ``Language models are few-shot learners,'' 2020.

\bibitem{vapnik}
V.~N. Vapnik, {\em The Nature of Statistical Learning Theory}, ch.~9, p.~291.
\newblock Springer, 2nd~ed., November 1999.

\bibitem{ba2016}
L.~J. Ba, J.~R. Kiros, and G.~E. Hinton, ``Layer normalization,'' {\em CoRR},
  vol.~abs/1607.06450, 2016.

\bibitem{glorot2010}
X.~Glorot and Y.~Bengio, ``Understanding the difficulty of training deep
  feedforward neural networks,'' in {\em Proceedings of the Thirteenth
  International Conference on Artificial Intelligence and Statistics} (Y.~W.
  Teh and M.~Titterington, eds.), vol.~9 of {\em Proceedings of Machine
  Learning Research}, (Chia Laguna Resort, Sardinia, Italy), pp.~249--256,
  PMLR, 13--15 May 2010.

\bibitem{shuster2012}
M.~Schuster and K.~Nakajima, ``Japanese and korean voice search,'' in {\em
  International Conference on Acoustics, Speech and Signal Processing},
  pp.~5149--5152, 2012.

\bibitem{tftext}
``Tensorflow text library.'' \url{https://github.com/tensorflow/text/}.

\bibitem{kingma2015}
D.~P. Kingma and J.~Ba, ``Adam: {A} method for stochastic optimization,'' in
  {\em 3rd International Conference on Learning Representations, {ICLR} 2015,
  San Diego, CA, USA, May 7-9, 2015, Conference Track Proceedings} (Y.~Bengio
  and Y.~LeCun, eds.), 2015.

\bibitem{srivastava14}
N.~Srivastava, G.~Hinton, A.~Krizhevsky, I.~Sutskever, and R.~Salakhutdinov,
  ``Dropout: A simple way to prevent neural networks from overfitting,'' {\em
  J. Mach. Learn. Res.}, vol.~15, pp.~1929--1958, Jan. 2014.

\bibitem{papineni2002}
K.~Papineni, S.~Roukos, T.~Ward, and W.-J. Zhu, ``{B}leu: a method for
  automatic evaluation of machine translation,'' in {\em Proceedings of the
  40th Annual Meeting of the Association for Computational Linguistics},
  (Philadelphia, Pennsylvania, USA), pp.~311--318, Association for
  Computational Linguistics, July 2002.

\bibitem{sacrebleu2018}
M.~Post, ``A call for clarity in reporting {BLEU} scores,'' in {\em Proceedings
  of the Third Conference on Machine Translation: Research Papers}, (Brussels,
  Belgium), pp.~186--191, Association for Computational Linguistics, Oct. 2018.

\bibitem{wu2019}
Y.~Wu, M.~Schuster, Z.~Chen, Q.~V. Le, M.~Norouzi, W.~Macherey, M.~Krikun,
  Y.~Cao, Q.~Gao, K.~Macherey, J.~Klingner, A.~Shah, M.~Johnson, X.~Liu,
  L.~Kaiser, S.~Gouws, Y.~Kato, T.~Kudo, H.~Kazawa, K.~Stevens, G.~Kurian,
  N.~Patil, W.~Wang, C.~Young, J.~Smith, J.~Riesa, A.~Rudnick, O.~Vinyals,
  G.~Corrado, M.~Hughes, and J.~Dean, ``Google's neural machine translation
  system: Bridging the gap between human and machine translation,'' {\em CoRR},
  vol.~abs/1609.08144, 2016.

\bibitem{tfwp2015}
M.~Abadi, A.~Agarwal, P.~Barham, E.~Brevdo, Z.~Chen, C.~Citro, G.~S. Corrado,
  A.~Davis, J.~Dean, M.~Devin, S.~Ghemawat, I.~Goodfellow, A.~Harp, G.~Irving,
  M.~Isard, Y.~Jia, R.~Jozefowicz, L.~Kaiser, M.~Kudlur, J.~Levenberg,
  D.~Man\'{e}, R.~Monga, S.~Moore, D.~Murray, C.~Olah, M.~Schuster, J.~Shlens,
  B.~Steiner, I.~Sutskever, K.~Talwar, P.~Tucker, V.~Vanhoucke, V.~Vasudevan,
  F.~Vi\'{e}gas, O.~Vinyals, P.~Warden, M.~Wattenberg, M.~Wicke, Y.~Yu, and
  X.~Zheng, ``{TensorFlow}: Large-scale machine learning on heterogeneous
  systems,'' 2015.
\newblock Software available from tensorflow.org.

\bibitem{britz2017}
D.~Britz, A.~Goldie, M.-T. Luong, and Q.~Le, ``Massive exploration of neural
  machine translation architectures,'' in {\em Proceedings of the 2017
  Conference on Empirical Methods in Natural Language Processing}, (Copenhagen,
  Denmark), pp.~1442--1451, Association for Computational Linguistics, Sept.
  2017.

\bibitem{wu2016}
Y.~Wu, M.~Schuster, Z.~Chen, Q.~V. Le, M.~Norouzi, W.~Macherey, M.~Krikun,
  Y.~Cao, Q.~Gao, K.~Macherey, J.~Klingner, A.~Shah, M.~Johnson, X.~Liu,
  Łukasz Kaiser, S.~Gouws, Y.~Kato, T.~Kudo, H.~Kazawa, K.~Stevens, G.~Kurian,
  N.~Patil, W.~Wang, C.~Young, J.~Smith, J.~Riesa, A.~Rudnick, O.~Vinyals,
  G.~Corrado, M.~Hughes, and J.~Dean, ``Google's neural machine translation
  system: Bridging the gap between human and machine translation,'' {\em CoRR},
  vol.~abs/1609.08144, 2016.

\end{thebibliography}

\bibliographystyle{ieeetr}

\pagebreak

\appendix

\begin{titlepage}
\centering
\vspace*{\fill}
\LARGE APPENDIX
\vspace*{\fill}
\end{titlepage}

\setcounter{figure}{0}
\renewcommand{\thefigure}{A.\arabic{figure}}
\renewcommand*{\thepage}{A\arabic{page}} 

\thispagestyle{headings}

\begin{figure}
\begin{adjustwidth}{-5em}{-5em}
\centering
\begin{subfigure}[b]{0.6\textwidth}
	\centering
	\includegraphics[width=1.1\textwidth]{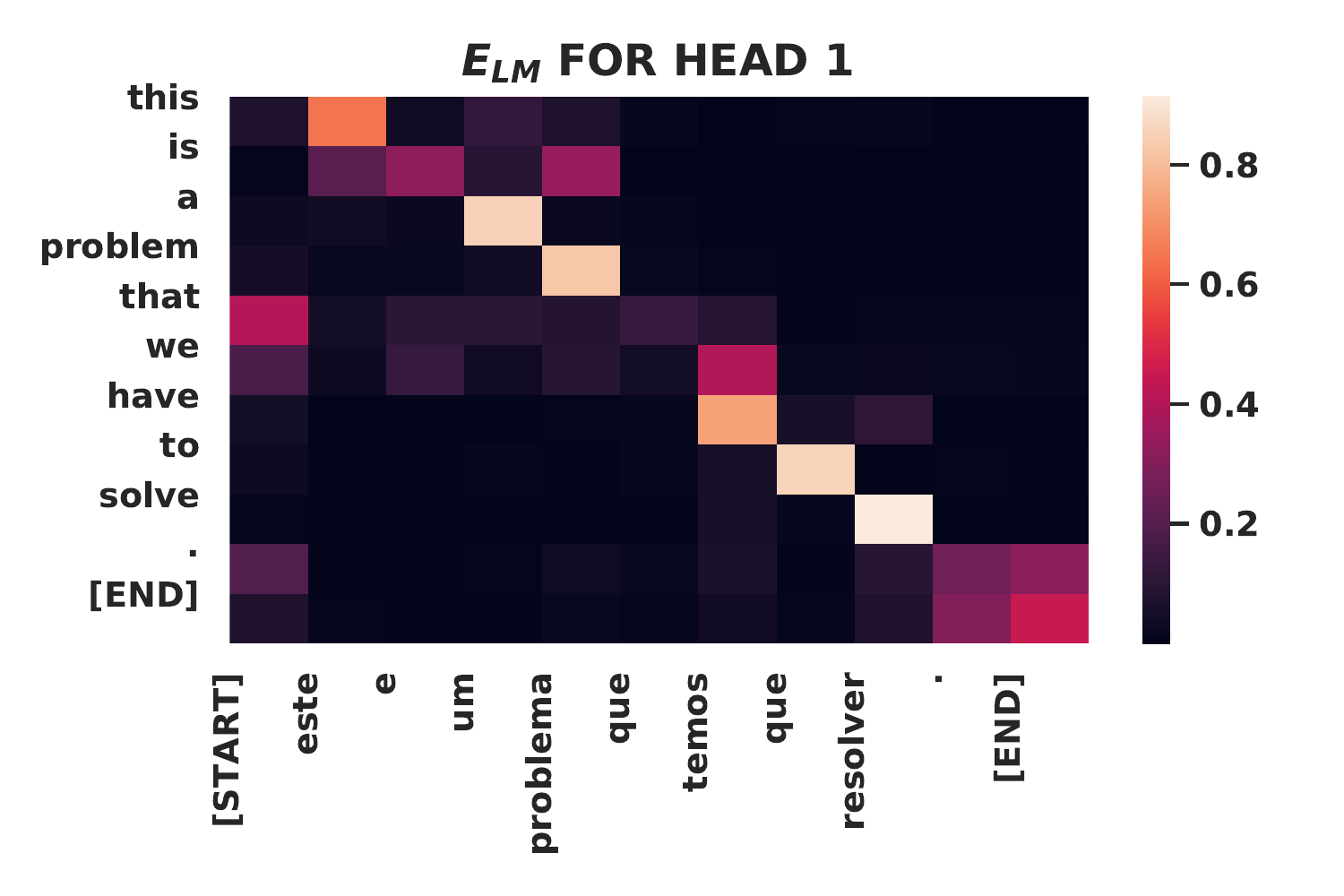}
\end{subfigure}
\hfill
\begin{subfigure}[b]{0.6\textwidth}
	\centering
	\includegraphics[width=1.1\textwidth]{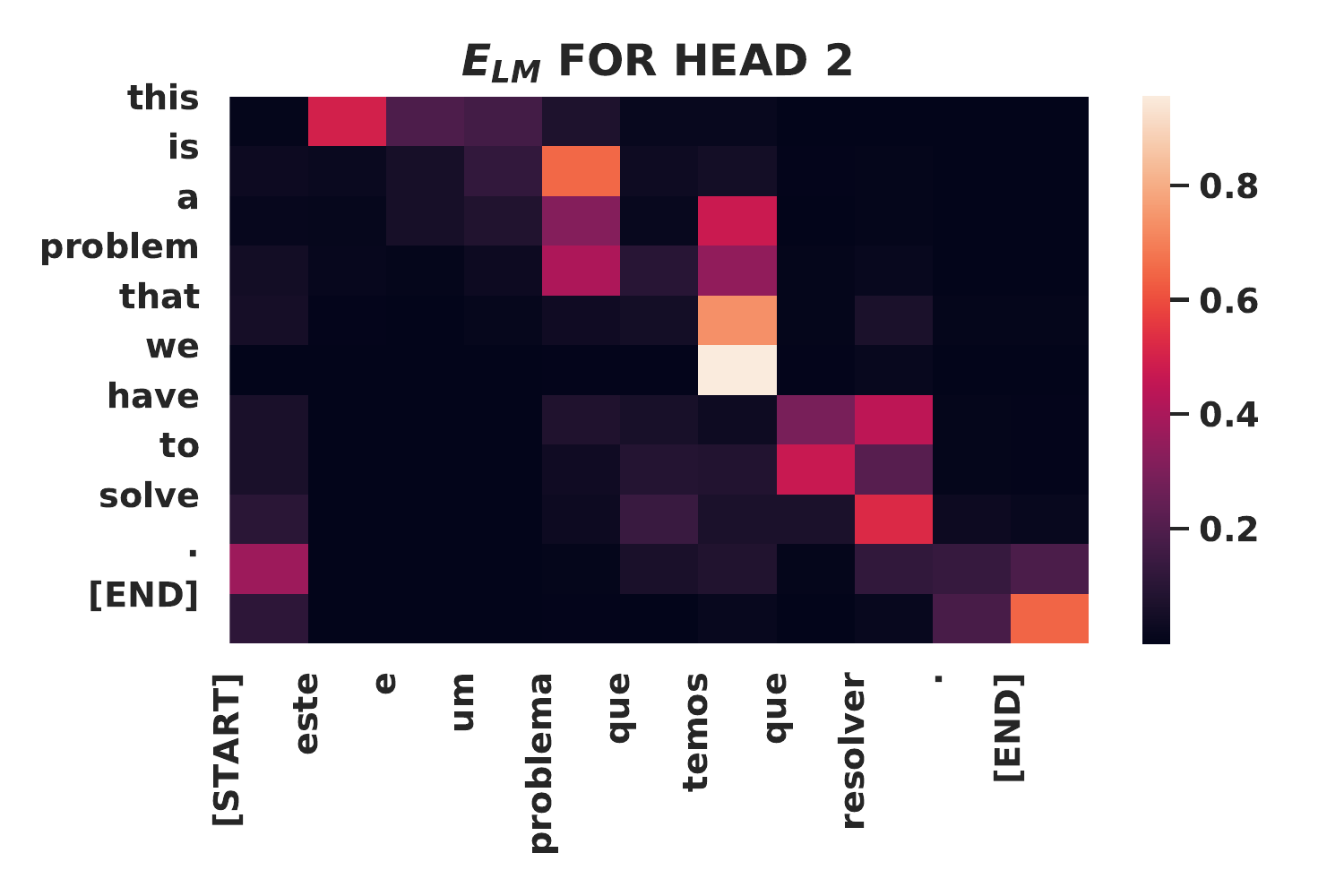}
\end{subfigure}
\hfill
\begin{subfigure}[b]{0.6\textwidth}
	\centering
	\includegraphics[width=1.1\textwidth]{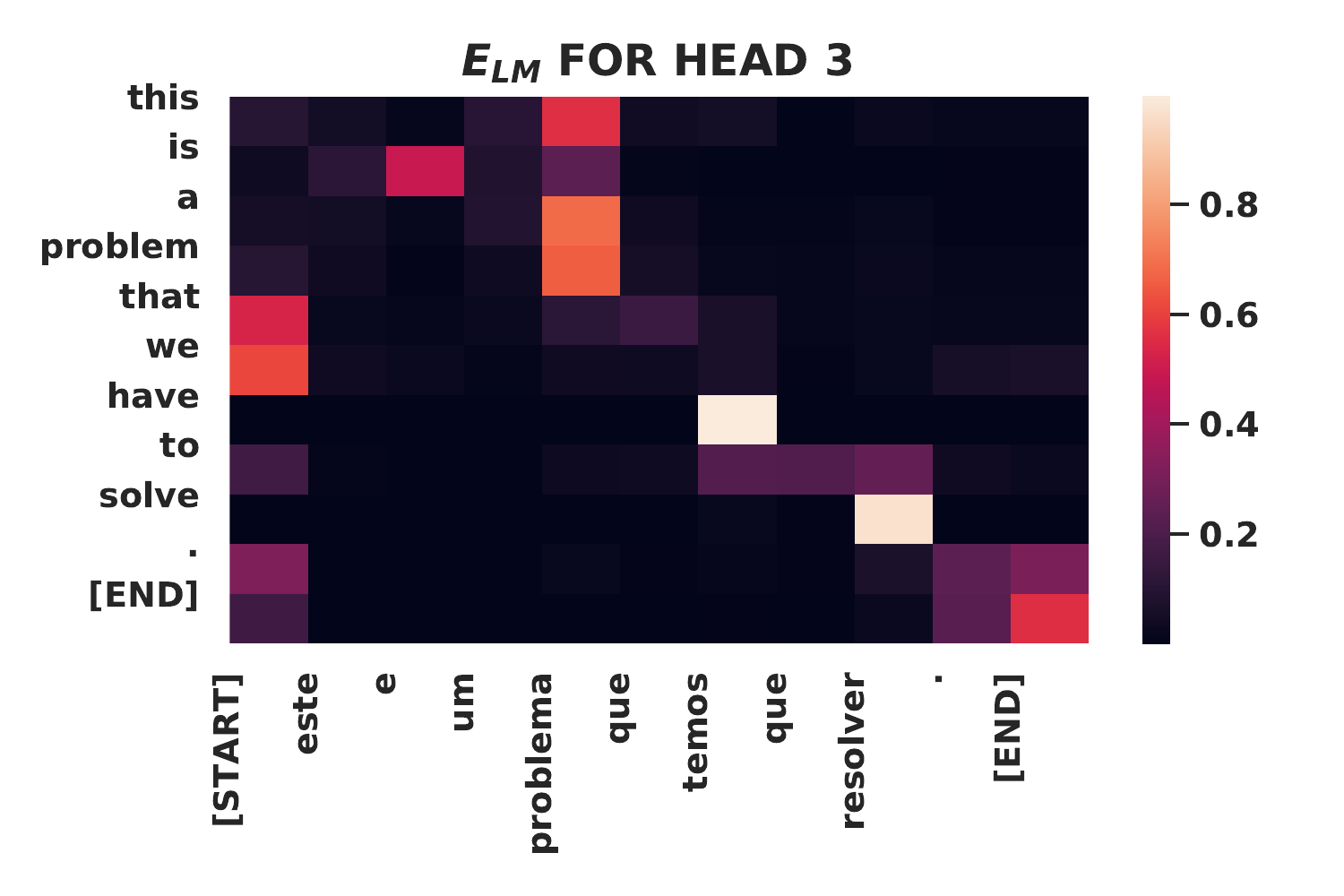}
\end{subfigure}
\hfill
\begin{subfigure}[b]{0.6\textwidth}
	\centering
	\includegraphics[width=1.1\textwidth]{Xheatmapmodel2elmh4}
\end{subfigure}
\centering
\begin{subfigure}[b]{0.6\textwidth}
	\centering
	\includegraphics[width=1.1\textwidth]{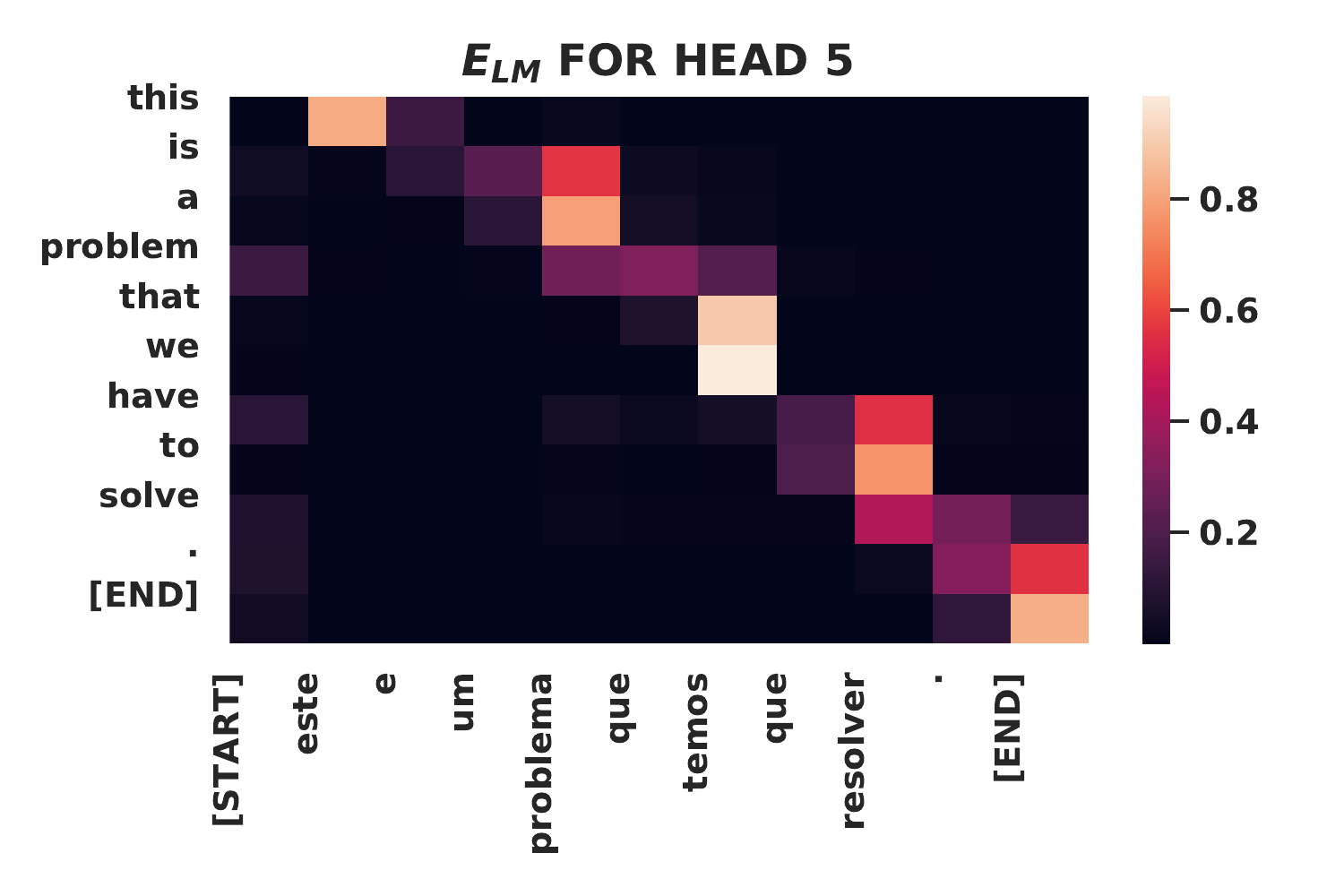}
\end{subfigure}
\hfill
\begin{subfigure}[b]{0.6\textwidth}
	\centering
	\includegraphics[width=1.1\textwidth]{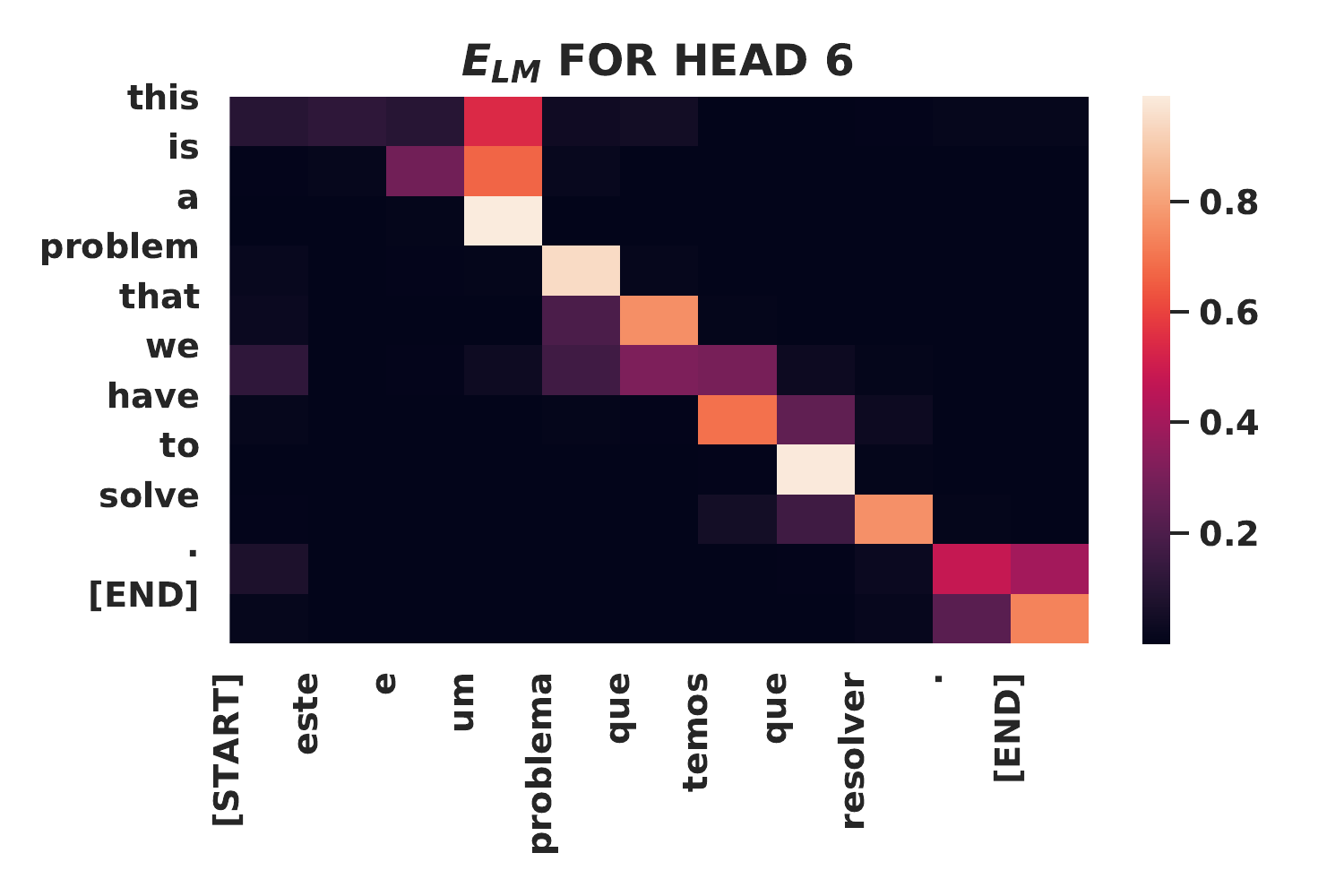}
\end{subfigure}
\hfill
\begin{subfigure}[b]{0.6\textwidth}
	\centering
	\includegraphics[width=1.1\textwidth]{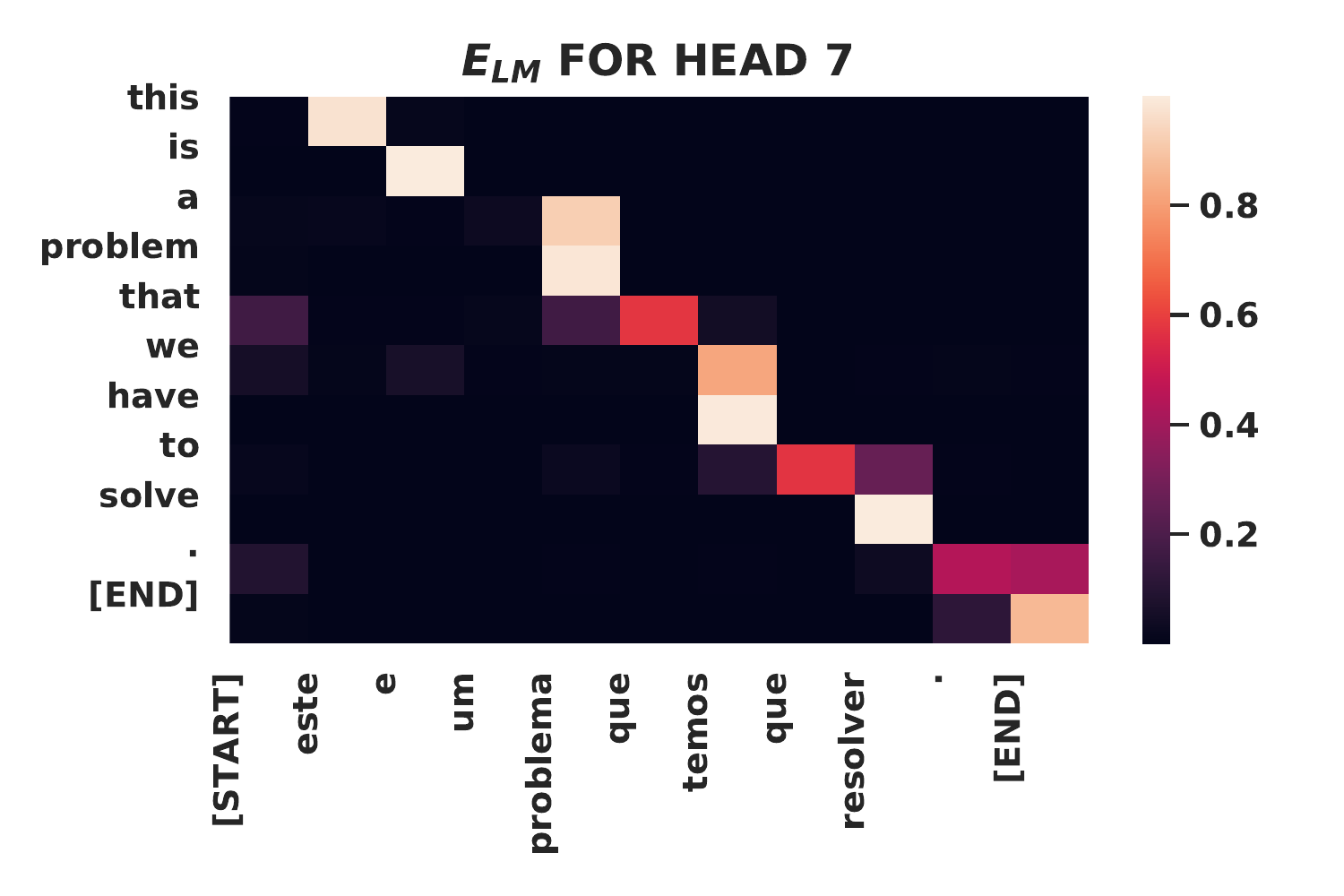}
\end{subfigure}
\hfill
\begin{subfigure}[b]{0.6\textwidth}
	\centering
	\includegraphics[width=1.1\textwidth]{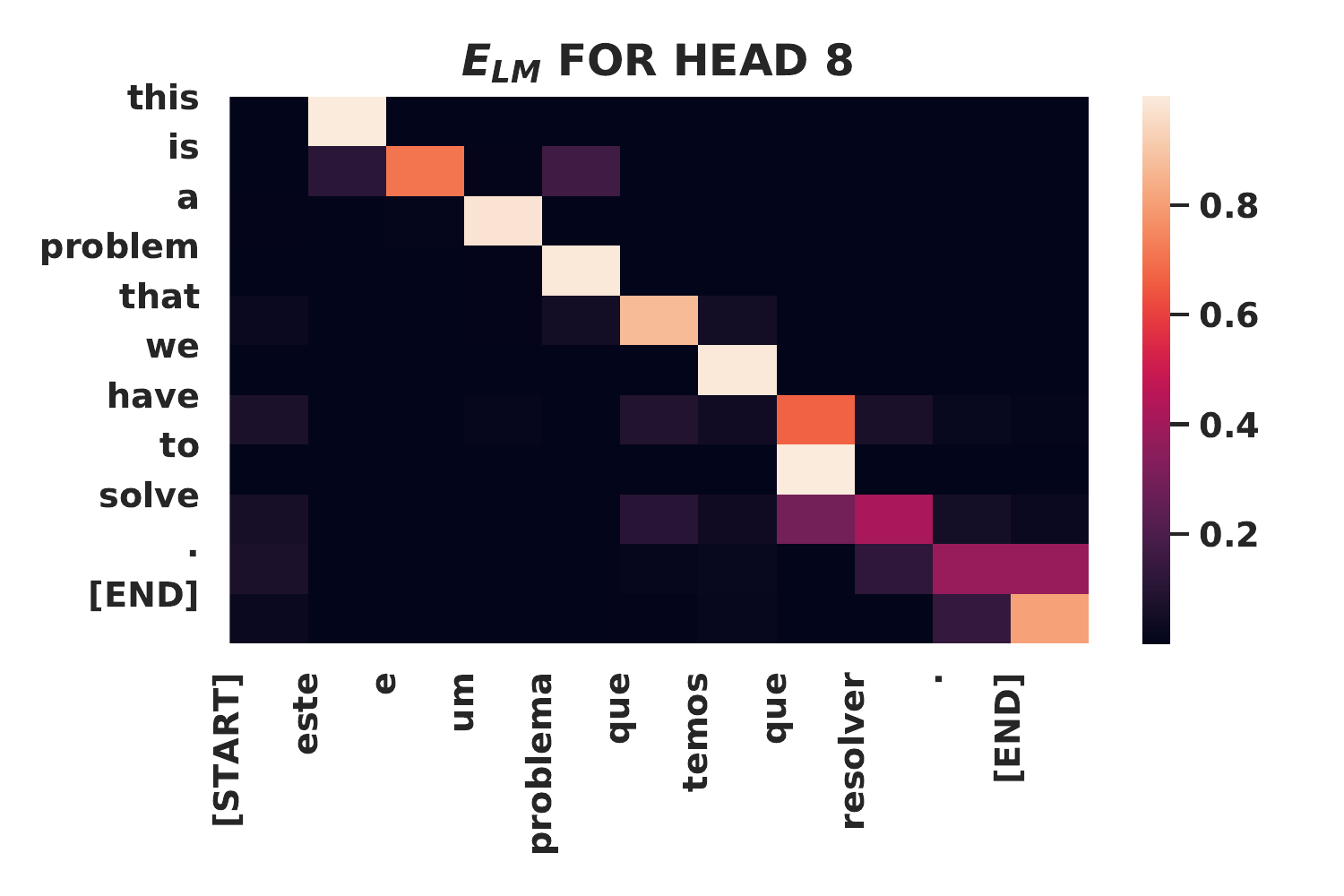}
\end{subfigure}
\caption{$\mE_{LM}$ heatmap plots for all heads from XLM attention stage from graph transformer model \#2.}
\label{fig1apx}
\end{adjustwidth}
\end{figure}  

\clearpage
\thispagestyle{headings}

\begin{figure}
\begin{adjustwidth}{-5em}{-5em}
\centering
\begin{subfigure}[b]{0.6\textwidth}
	\centering
	\includegraphics[width=1.1\textwidth]{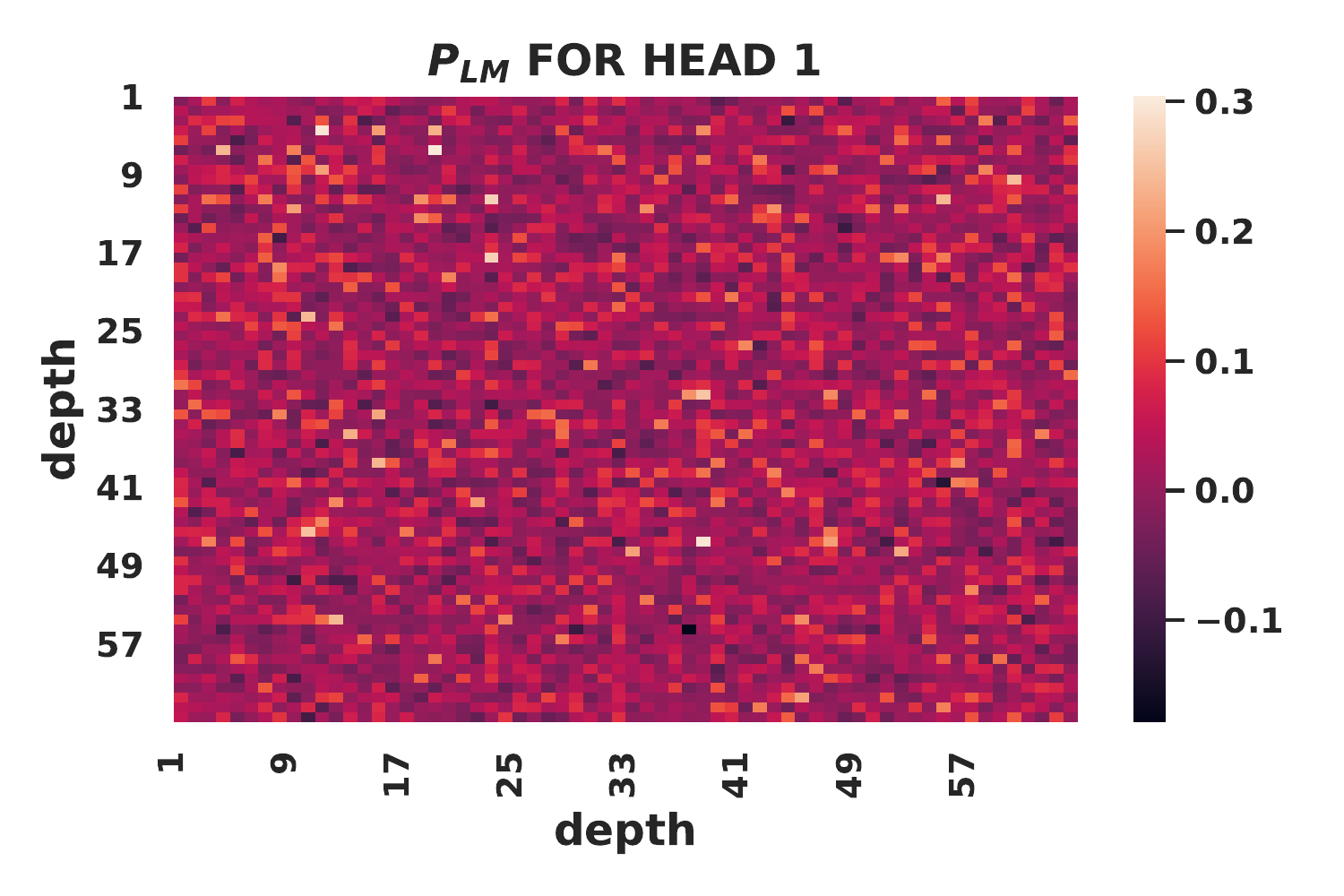}
	
\end{subfigure}
\hfill
\begin{subfigure}[b]{0.6\textwidth}
	\centering
	\includegraphics[width=1.1\textwidth]{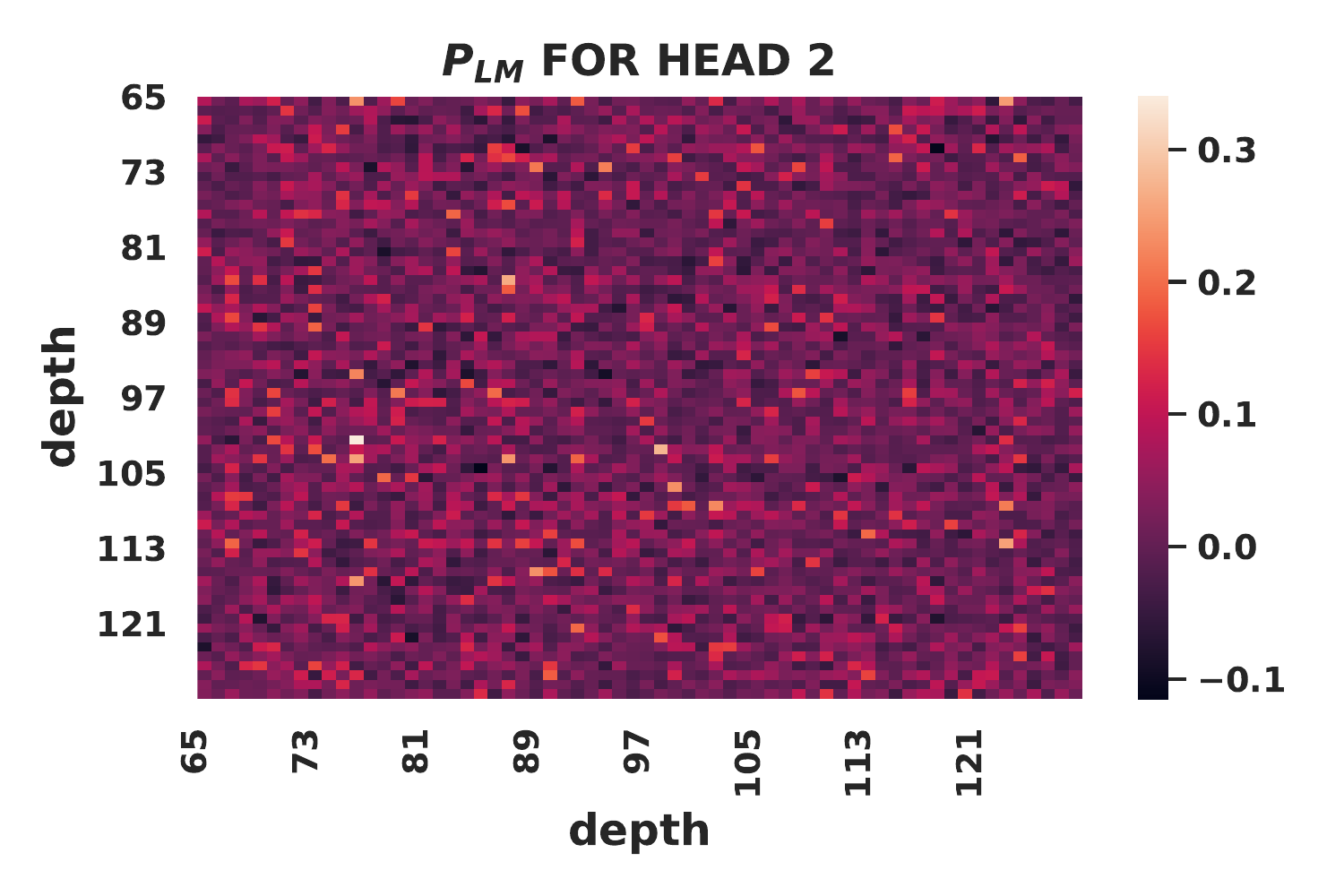}

\end{subfigure}
\hfill
\begin{subfigure}[b]{0.6\textwidth}
	\centering
	\includegraphics[width=1.1\textwidth]{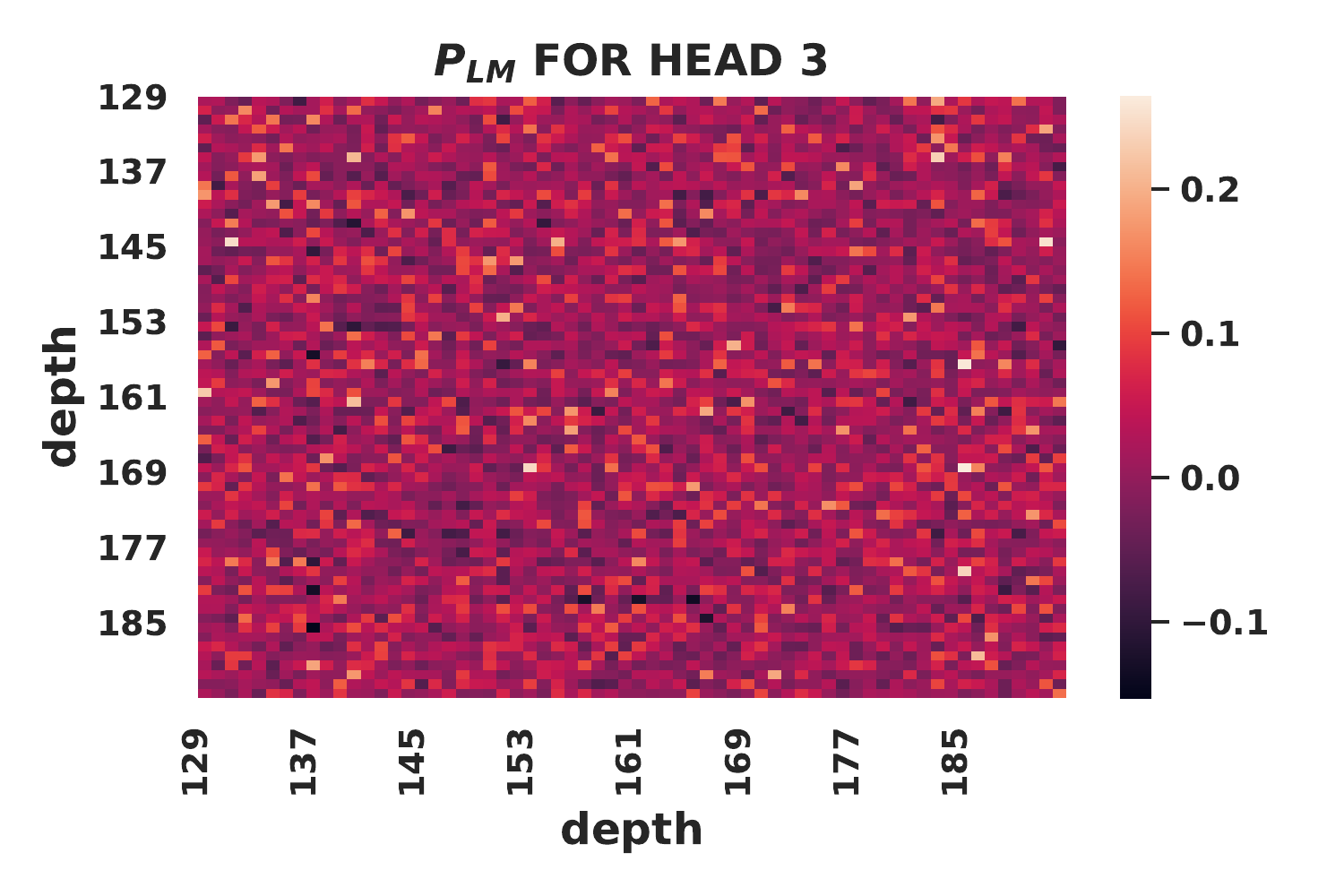}

\end{subfigure}
\hfill
\begin{subfigure}[b]{0.6\textwidth}
	\centering
	\includegraphics[width=1.1\textwidth]{Xheatmapmodel2plmh4}

\end{subfigure}
\centering
\begin{subfigure}[b]{0.6\textwidth}
	\centering
	\includegraphics[width=1.1\textwidth]{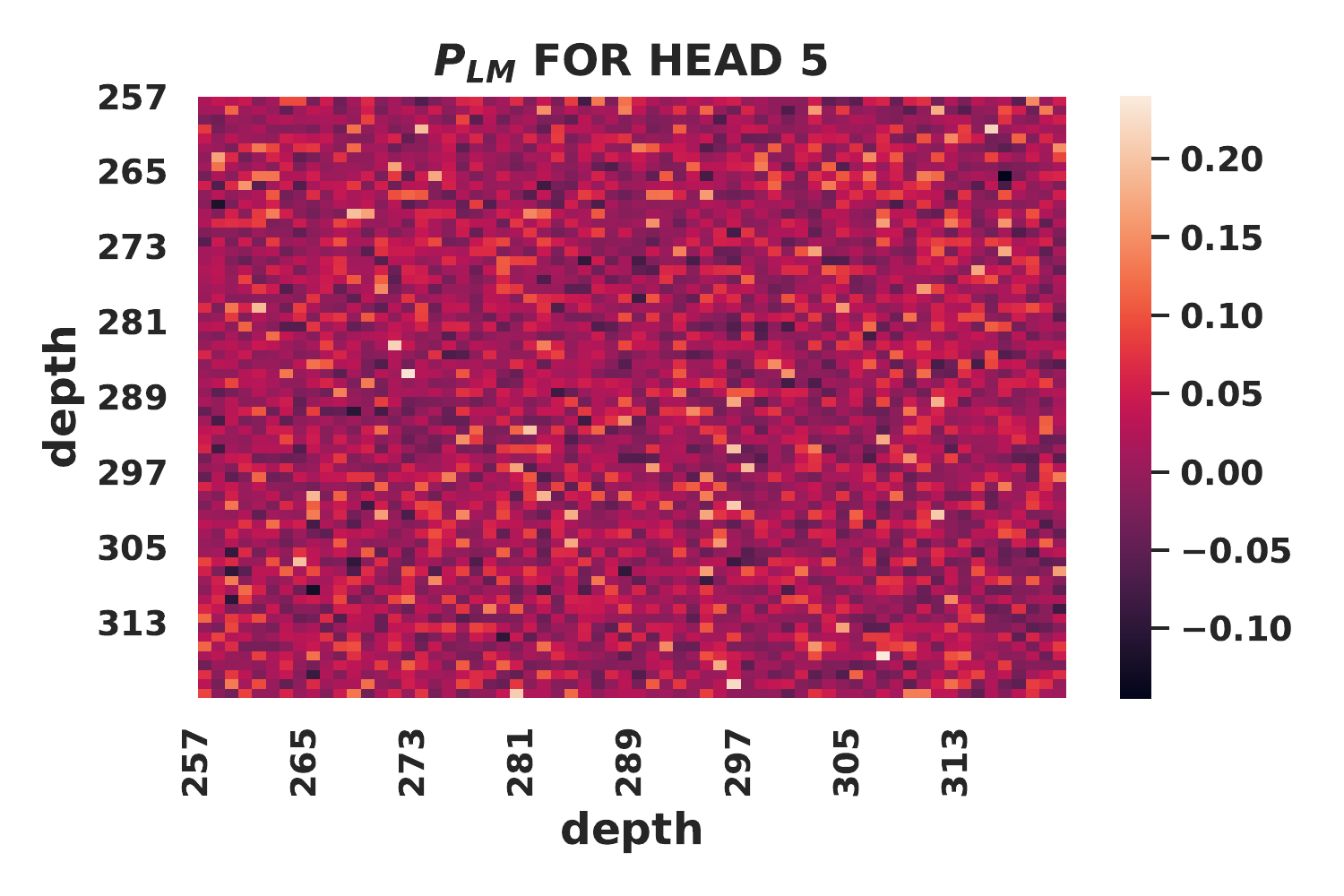}

\end{subfigure}
\hfill
\begin{subfigure}[b]{0.6\textwidth}
	\centering
	\includegraphics[width=1.1\textwidth]{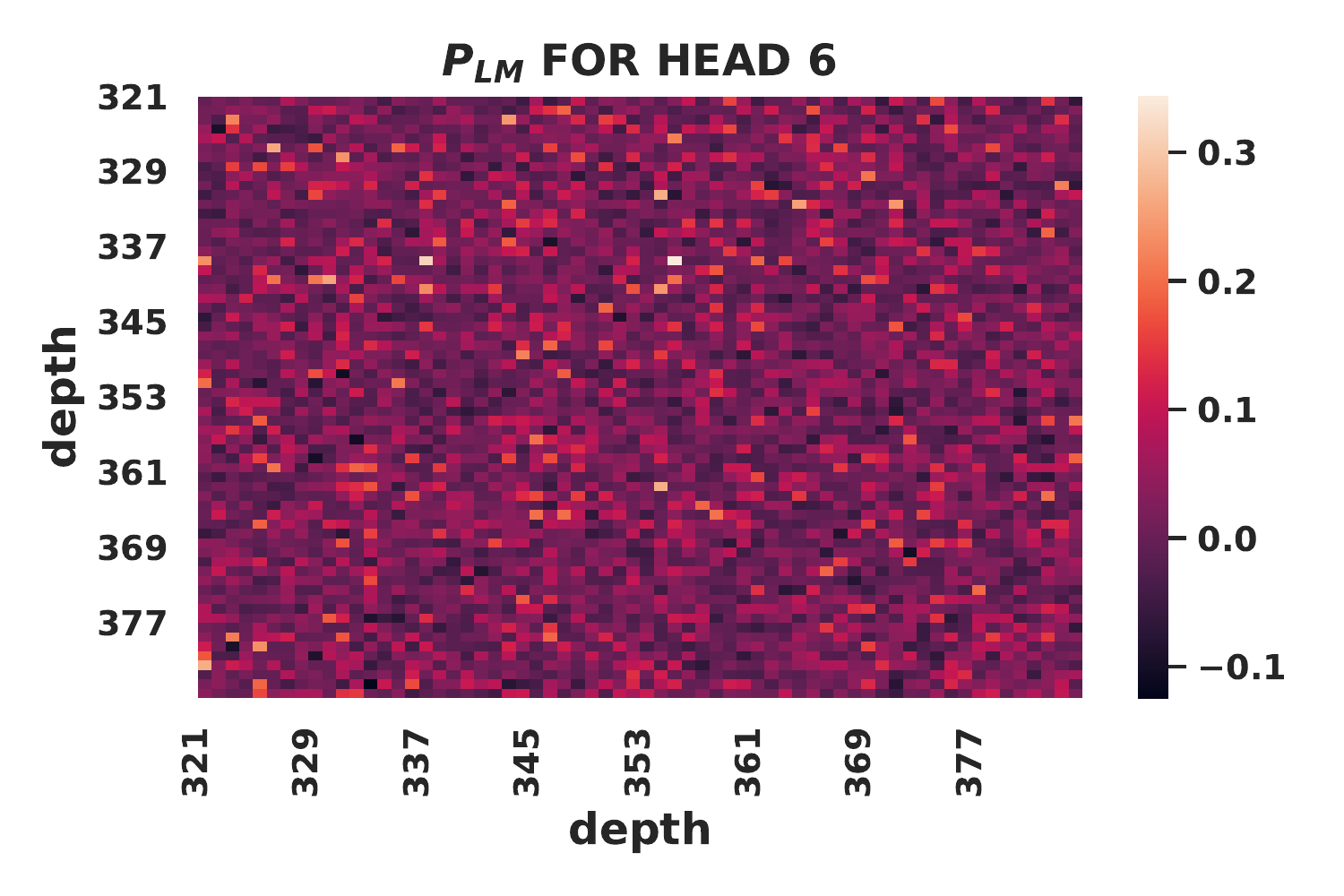}

\end{subfigure}
\hfill
\begin{subfigure}[b]{0.6\textwidth}
	\centering
	\includegraphics[width=1.1\textwidth]{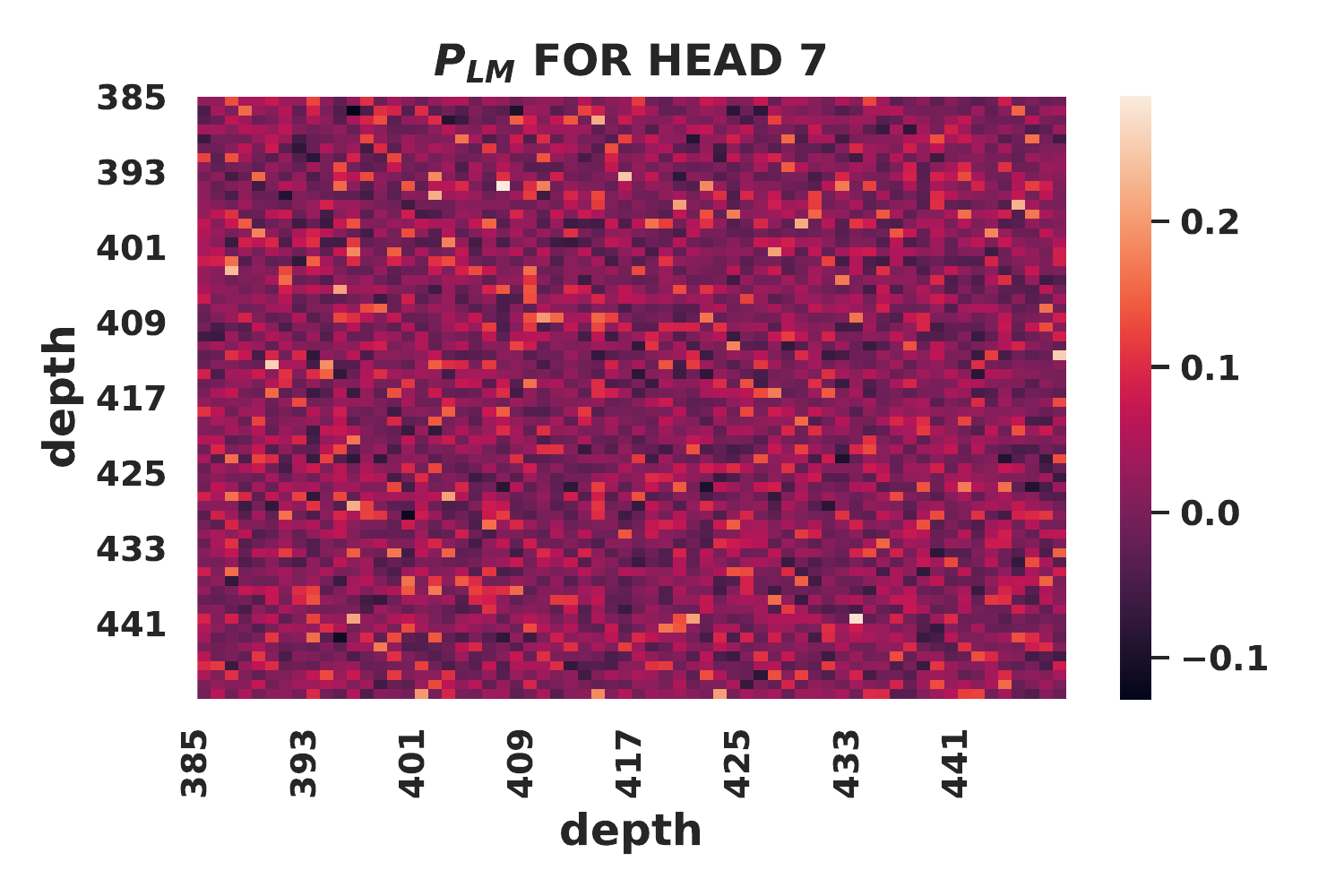}

\end{subfigure}
\hfill
\begin{subfigure}[b]{0.6\textwidth}
	\centering
	\includegraphics[width=1.1\textwidth]{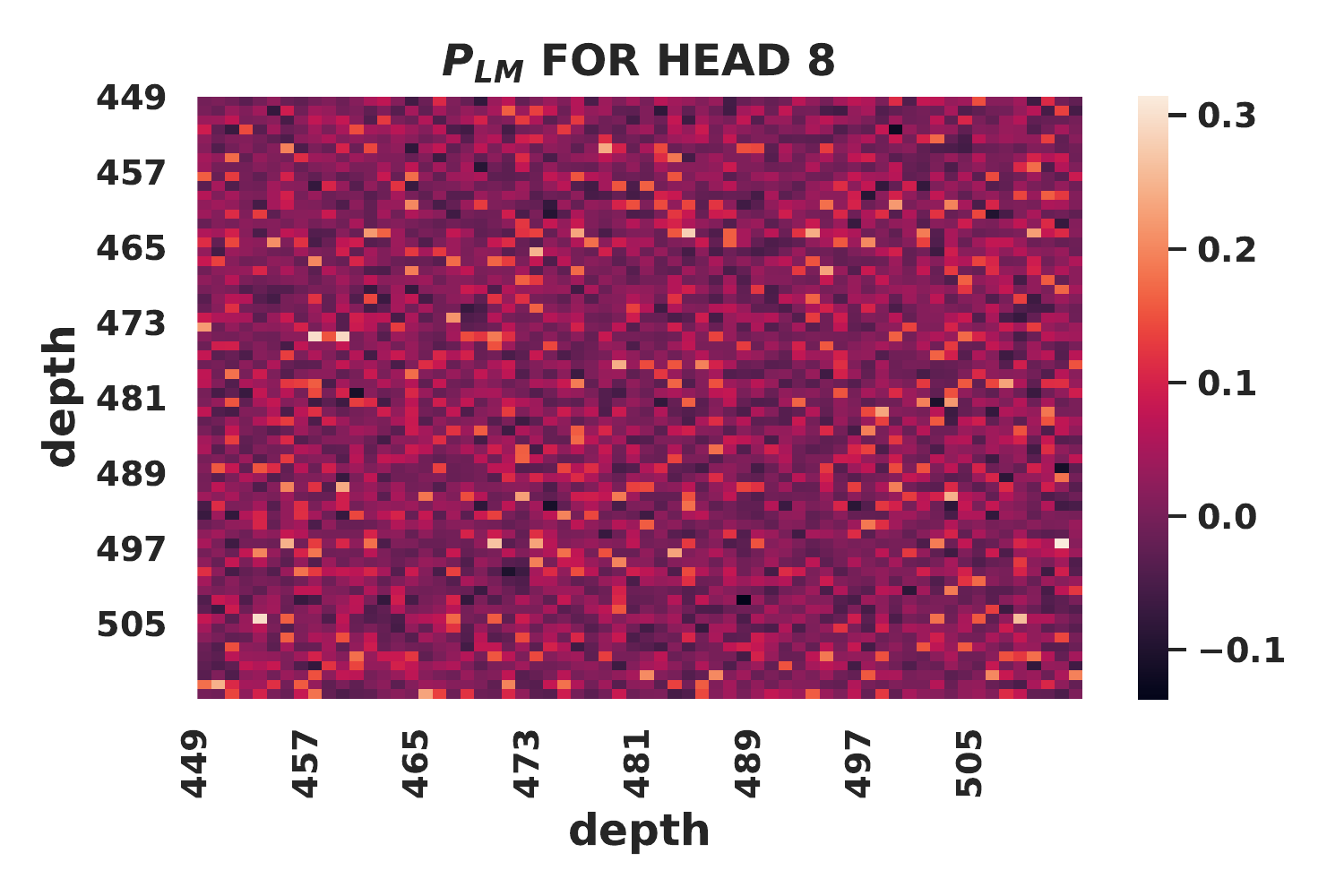}

\end{subfigure}
\caption{$\mP_{LM}$ heatmap plots for all heads from XLM attention stage from graph transformer model \#2. }
\label{fig2apx}
\end{adjustwidth}
\end{figure}

\clearpage
\thispagestyle{headings}

\begin{figure}
\begin{adjustwidth}{-5em}{-5em}
\centering
\begin{subfigure}[b]{0.6\textwidth}
	\centering
	\includegraphics[width=1.1\textwidth]{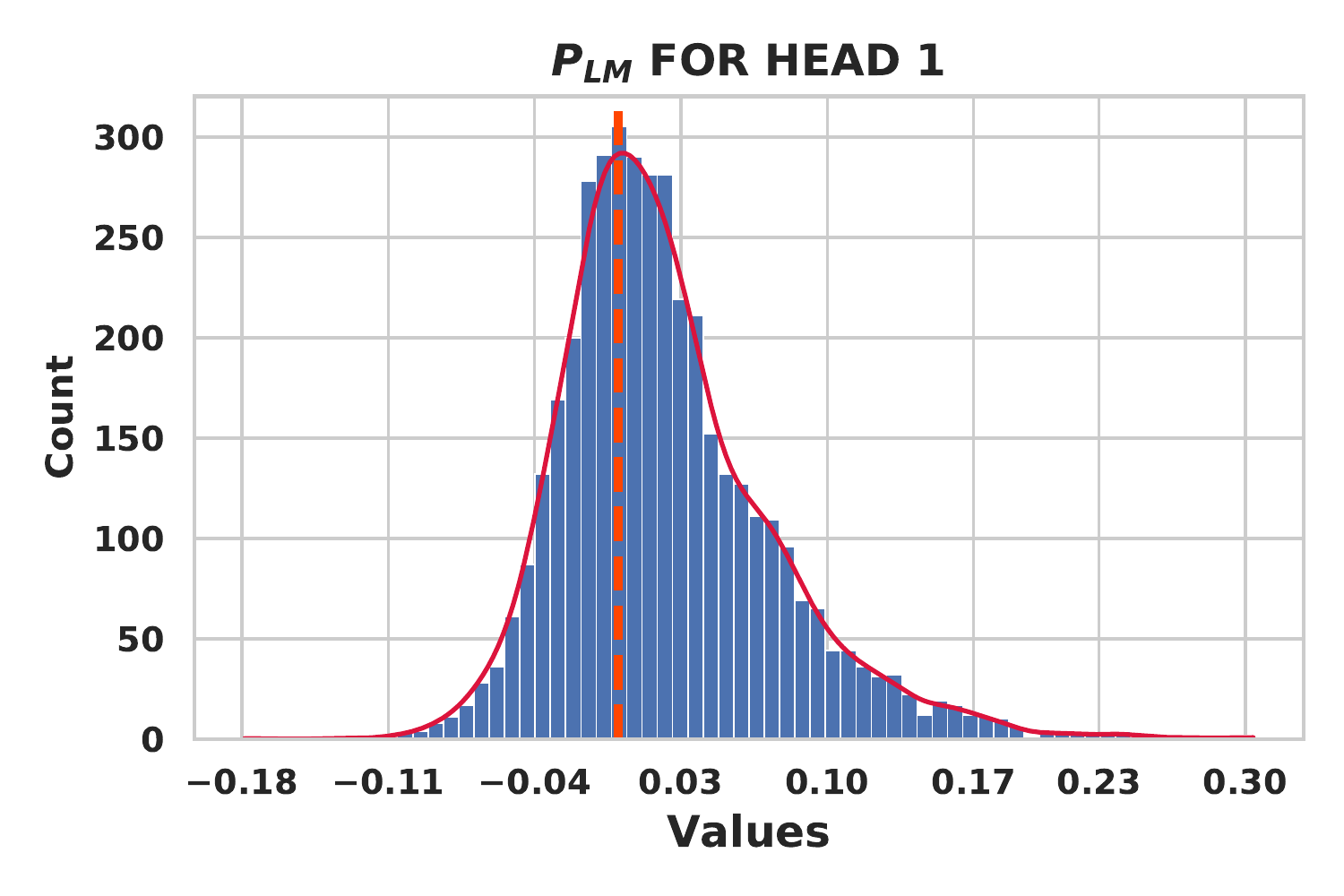}

\end{subfigure}
\hfill
\begin{subfigure}[b]{0.6\textwidth}
	\centering
	\includegraphics[width=1.1\textwidth]{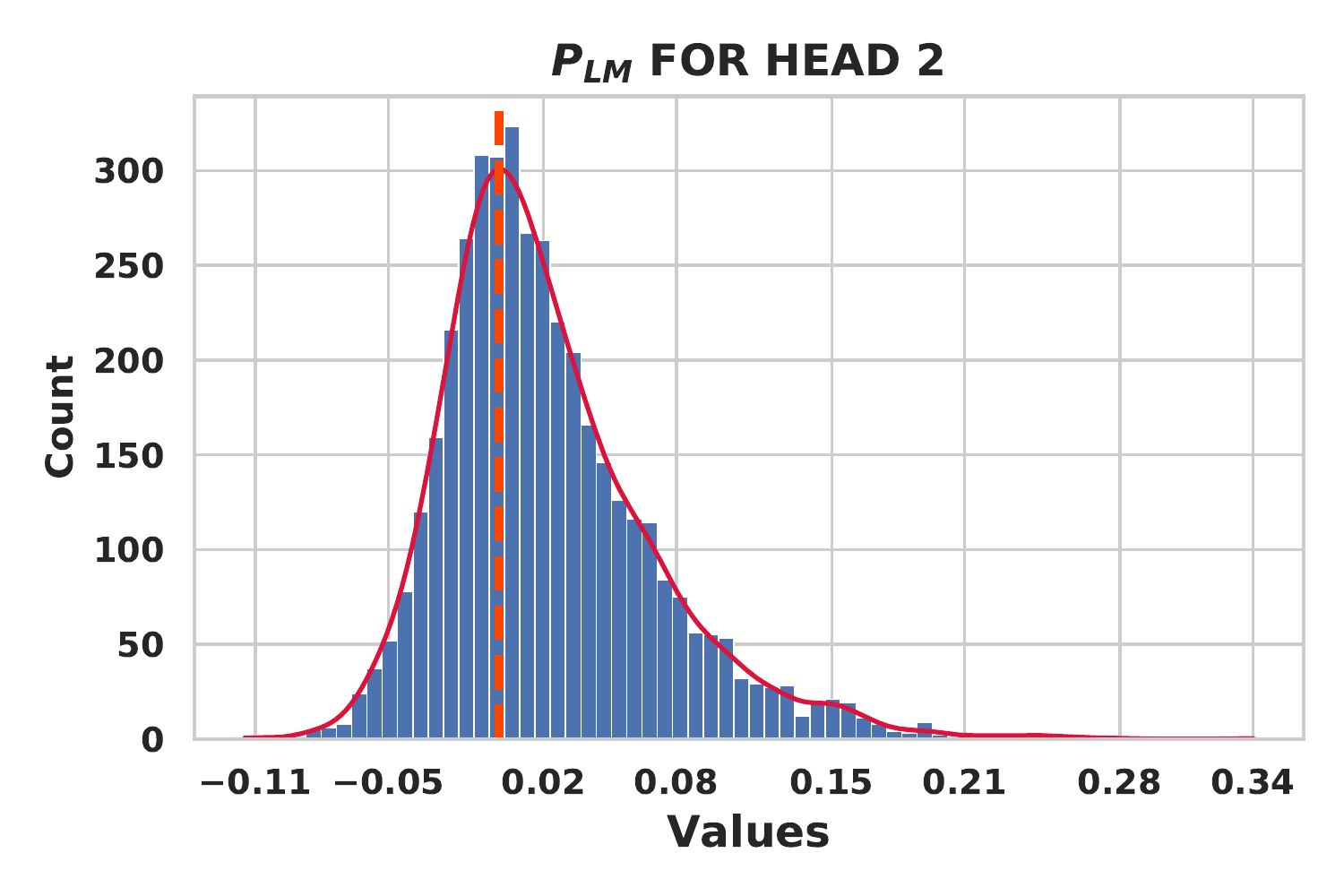}

\end{subfigure}
\hfill
\begin{subfigure}[b]{0.6\textwidth}
	\centering
	\includegraphics[width=1.1\textwidth]{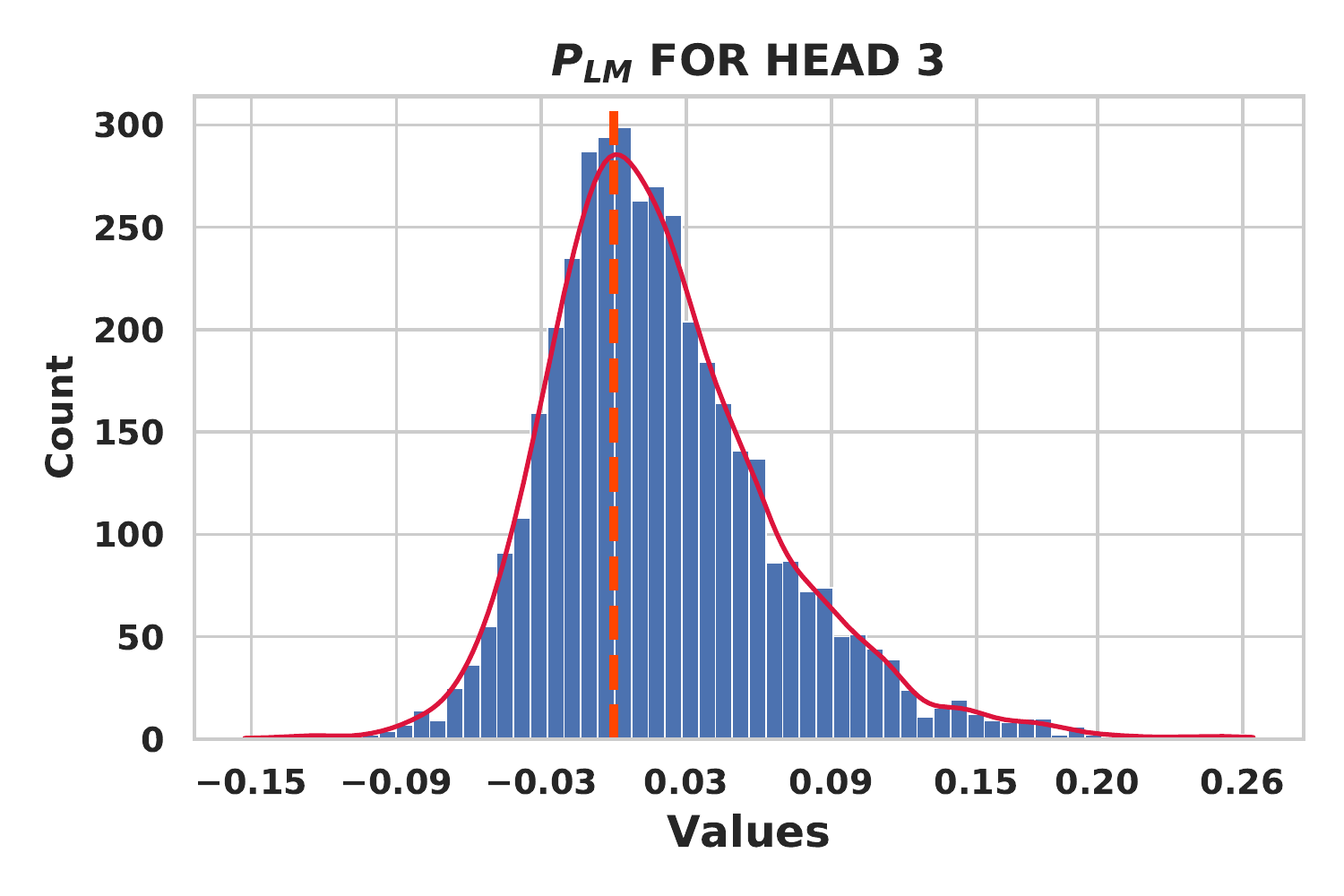}

\end{subfigure}
\hfill
\begin{subfigure}[b]{0.6\textwidth}
	\centering
	\includegraphics[width=1.1\textwidth]{Xhistmodel2plmh4}

\end{subfigure}
\centering
\begin{subfigure}[b]{0.6\textwidth}
	\centering
	\includegraphics[width=1.1\textwidth]{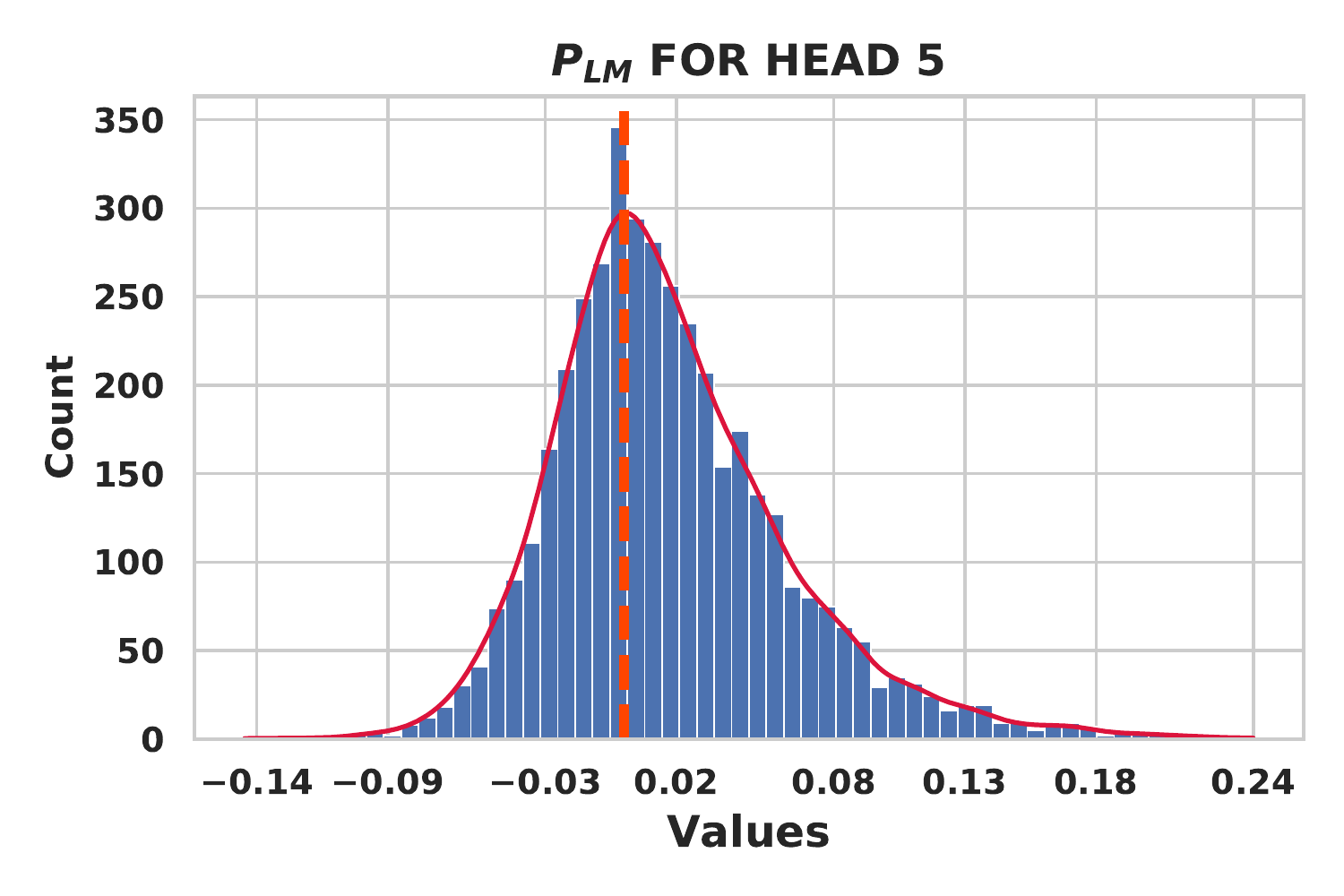}

\end{subfigure}
\hfill
\begin{subfigure}[b]{0.6\textwidth}
	\centering
	\includegraphics[width=1.1\textwidth]{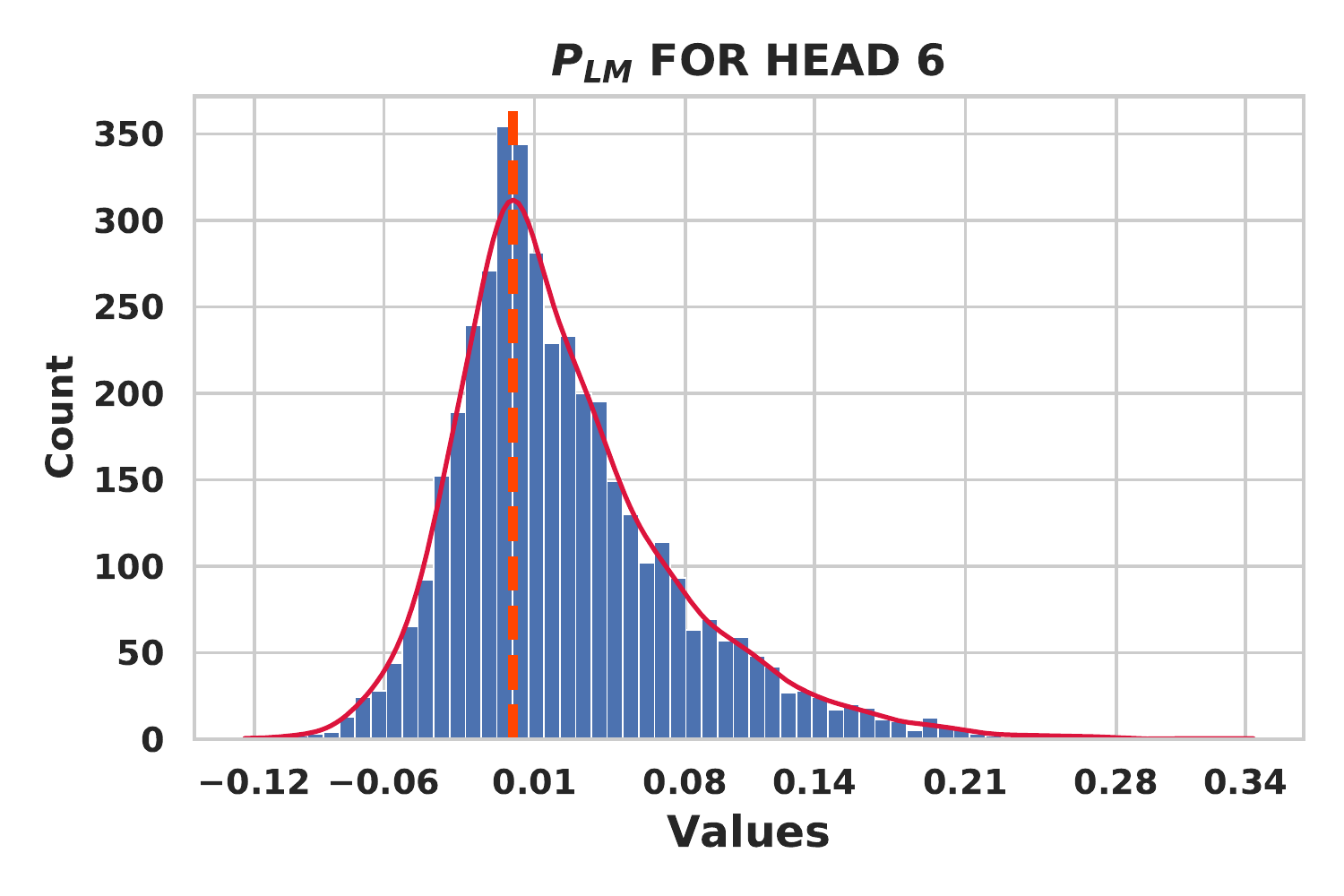}

\end{subfigure}
\hfill
\begin{subfigure}[b]{0.6\textwidth}
	\centering
	\includegraphics[width=1.1\textwidth]{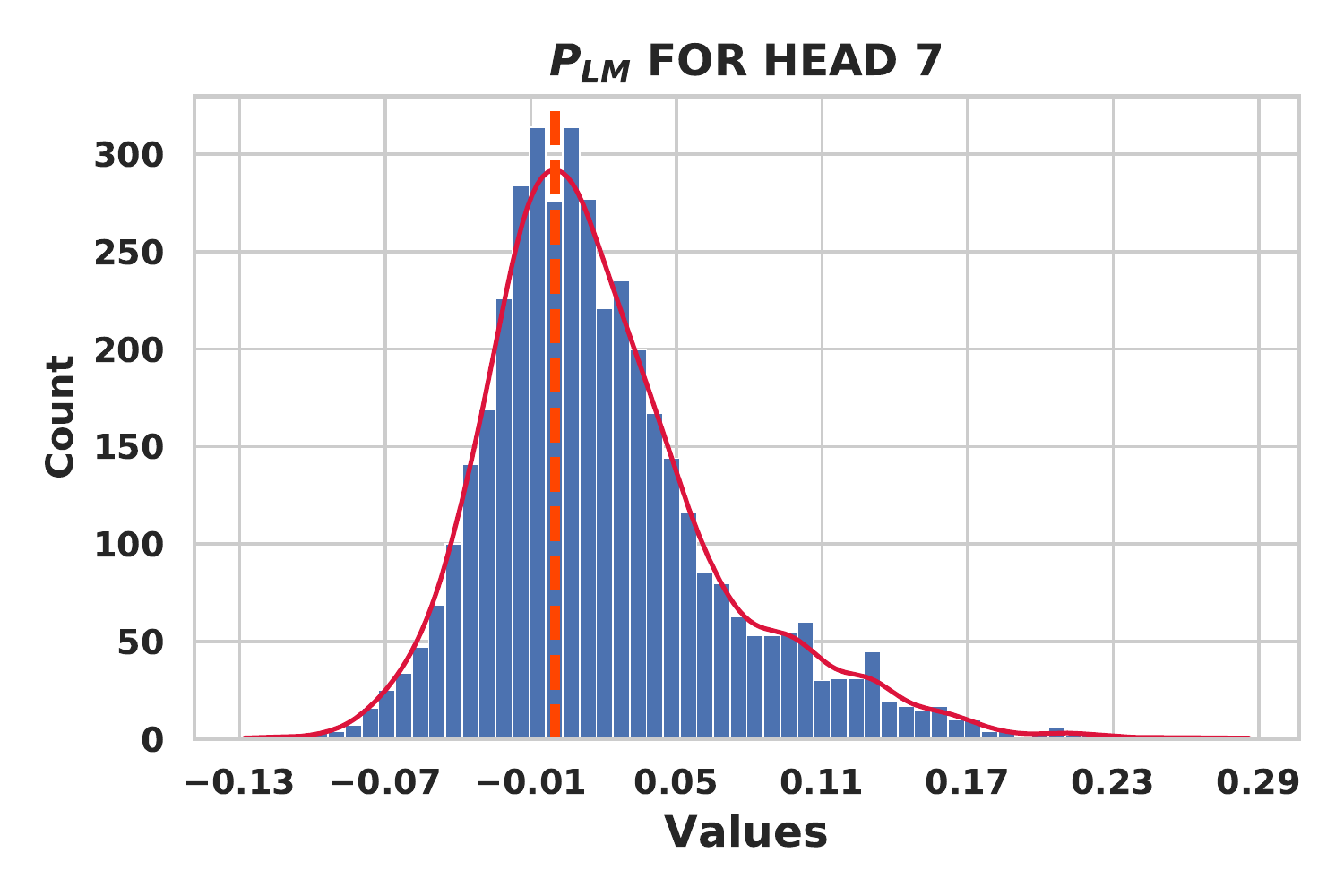}

\end{subfigure}
\hfill
\begin{subfigure}[b]{0.6\textwidth}
	\centering
	\includegraphics[width=1.1\textwidth]{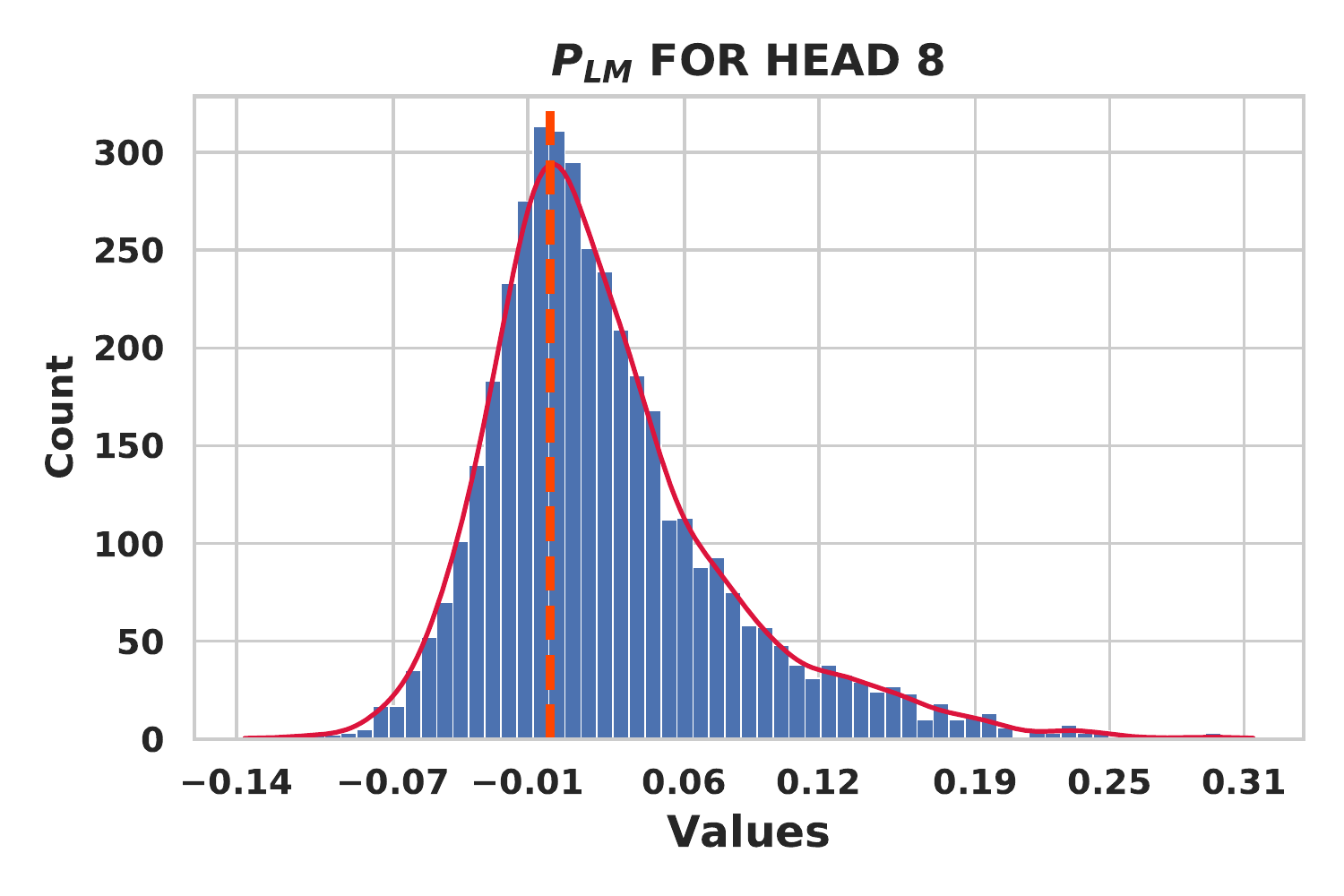}

\end{subfigure}
\caption{$\mP_{LM}$ histogram plots for all heads from XLM attention stage from graph transformer model \#2. Dashed line in orange marks zero value.}
\label{fig3apx}
\end{adjustwidth}
\end{figure}    

\clearpage
\thispagestyle{headings}

\begin{figure}
\begin{adjustwidth}{-5em}{-5em}
\centering
\begin{subfigure}[b]{0.6\textwidth}
	\centering
	\includegraphics[width=1.1\textwidth]{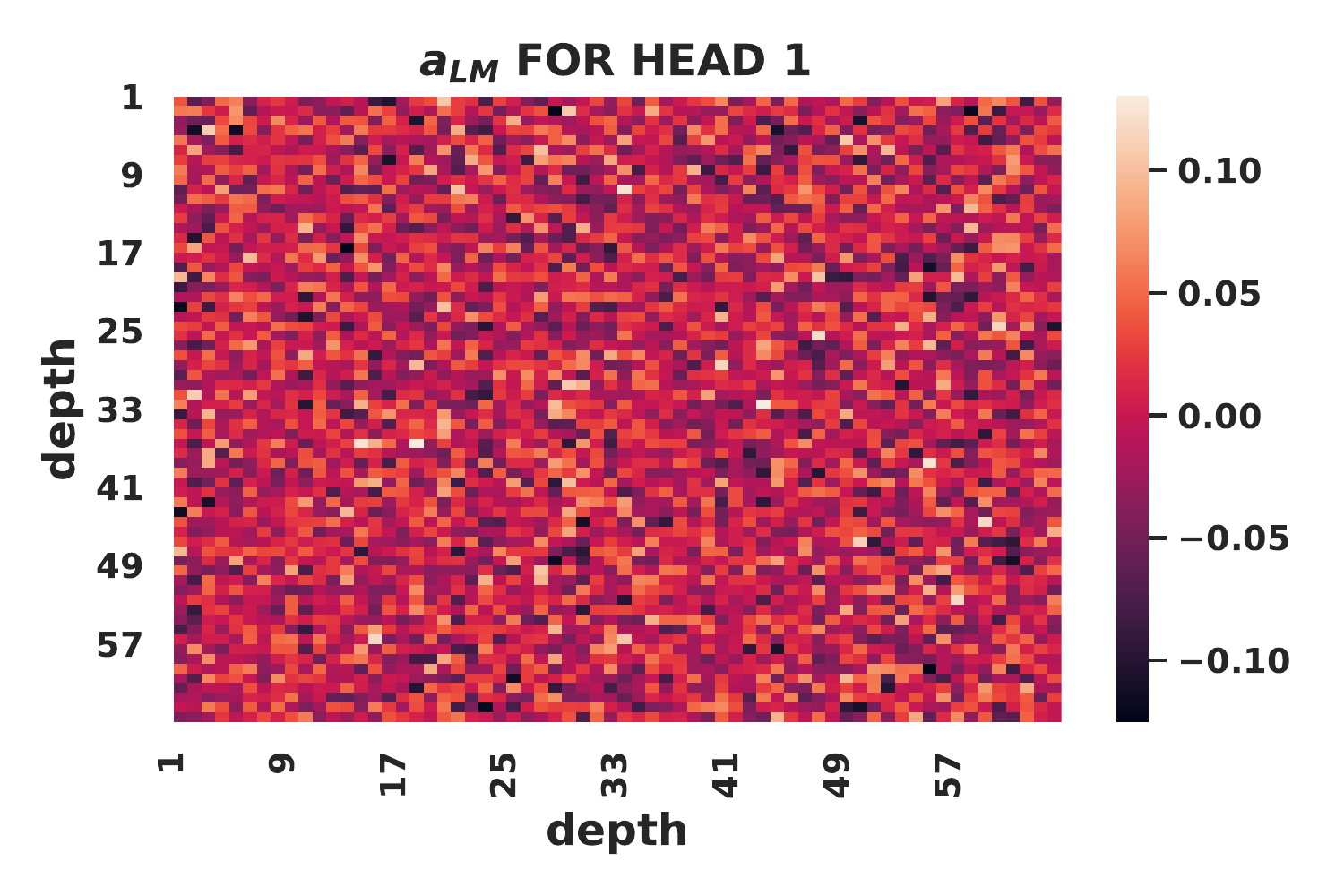}

\end{subfigure}
\hfill
\begin{subfigure}[b]{0.6\textwidth}
	\centering
	\includegraphics[width=1.1\textwidth]{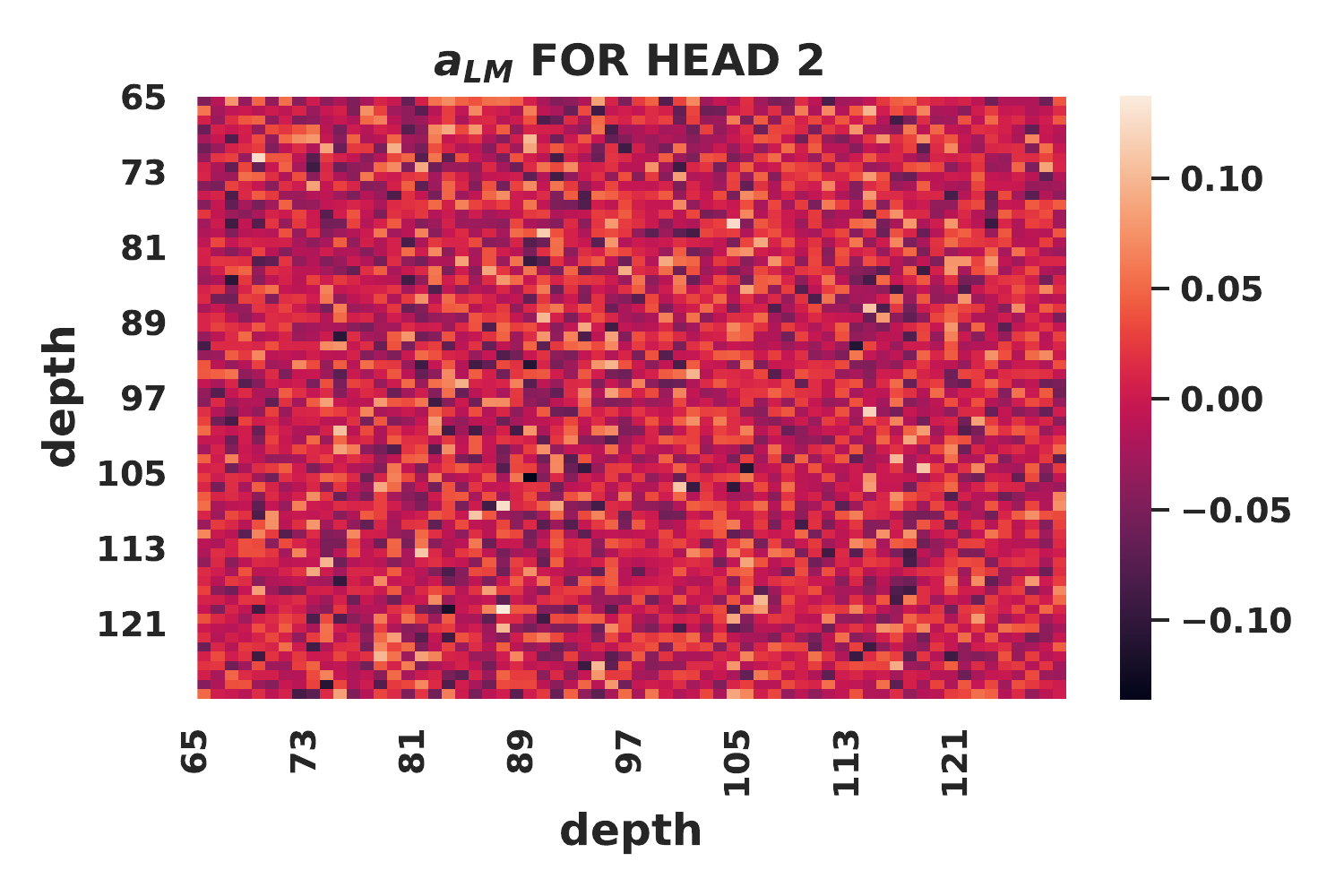}

\end{subfigure}
\hfill
\begin{subfigure}[b]{0.6\textwidth}
	\centering
	\includegraphics[width=1.1\textwidth]{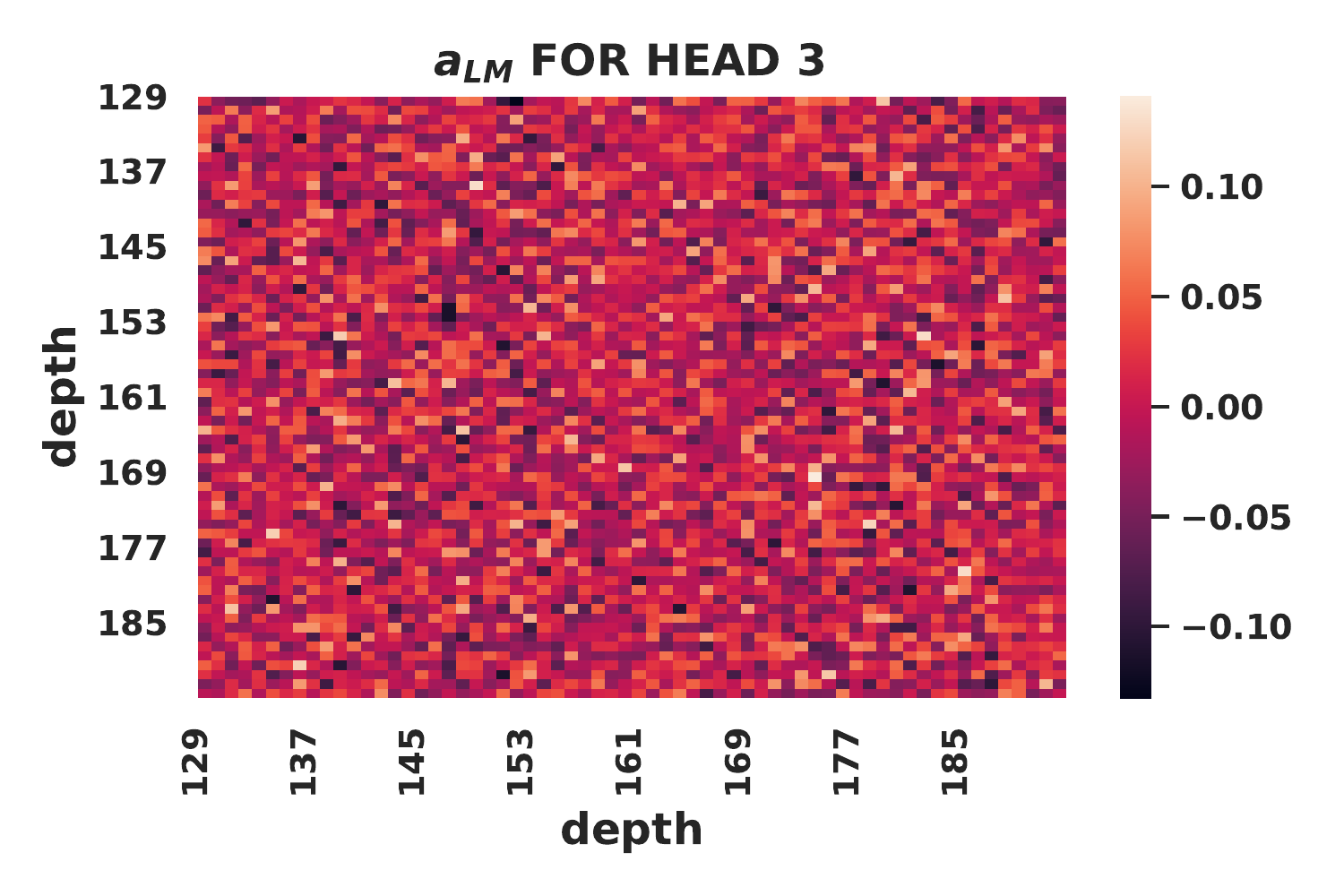}

\end{subfigure}
\hfill
\begin{subfigure}[b]{0.6\textwidth}
	\centering
	\includegraphics[width=1.1\textwidth]{Xheatmapmodel2avech4}

\end{subfigure}
\centering
\begin{subfigure}[b]{0.6\textwidth}
	\centering
	\includegraphics[width=1.1\textwidth]{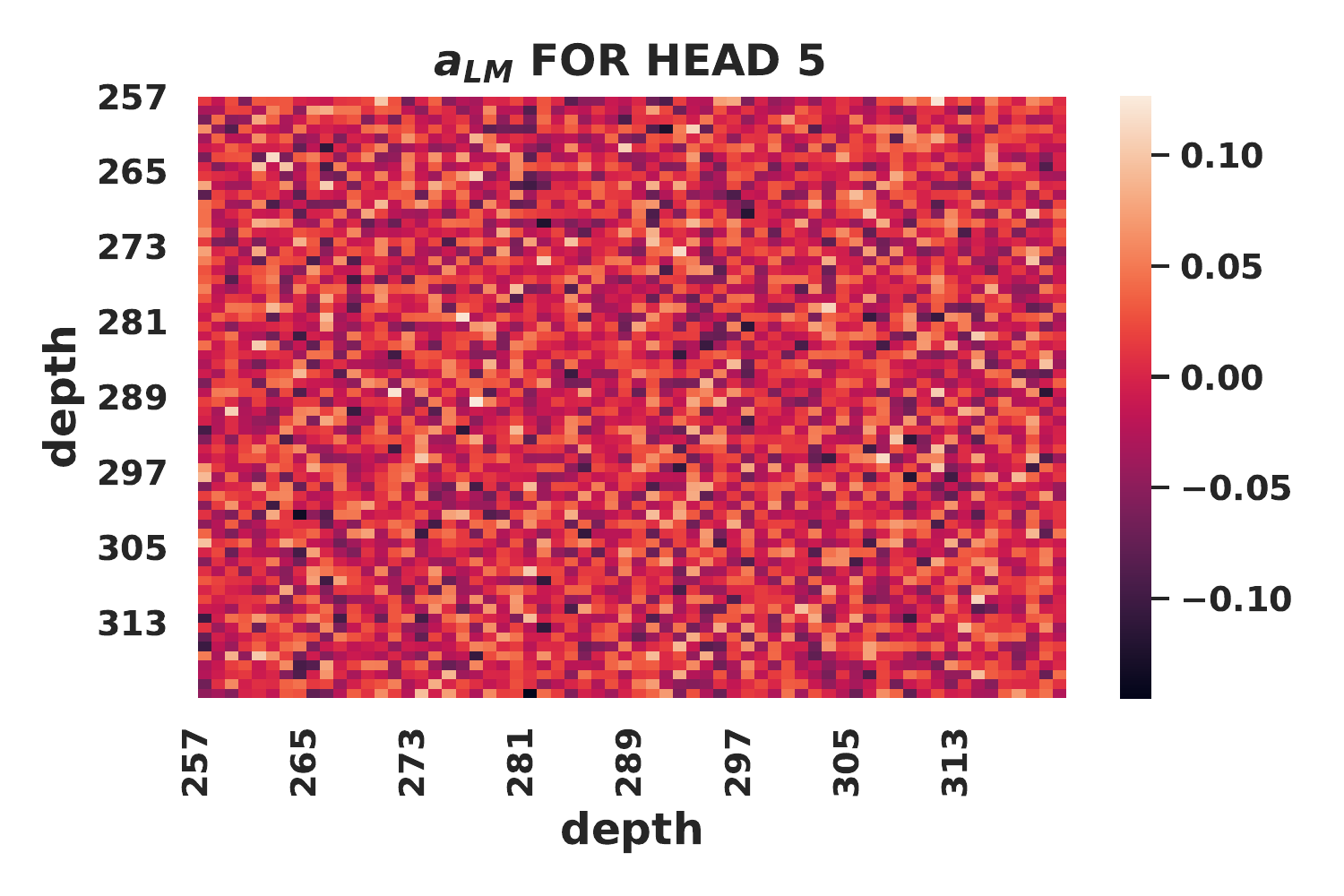}

\end{subfigure}
\hfill
\begin{subfigure}[b]{0.6\textwidth}
	\centering
	\includegraphics[width=1.1\textwidth]{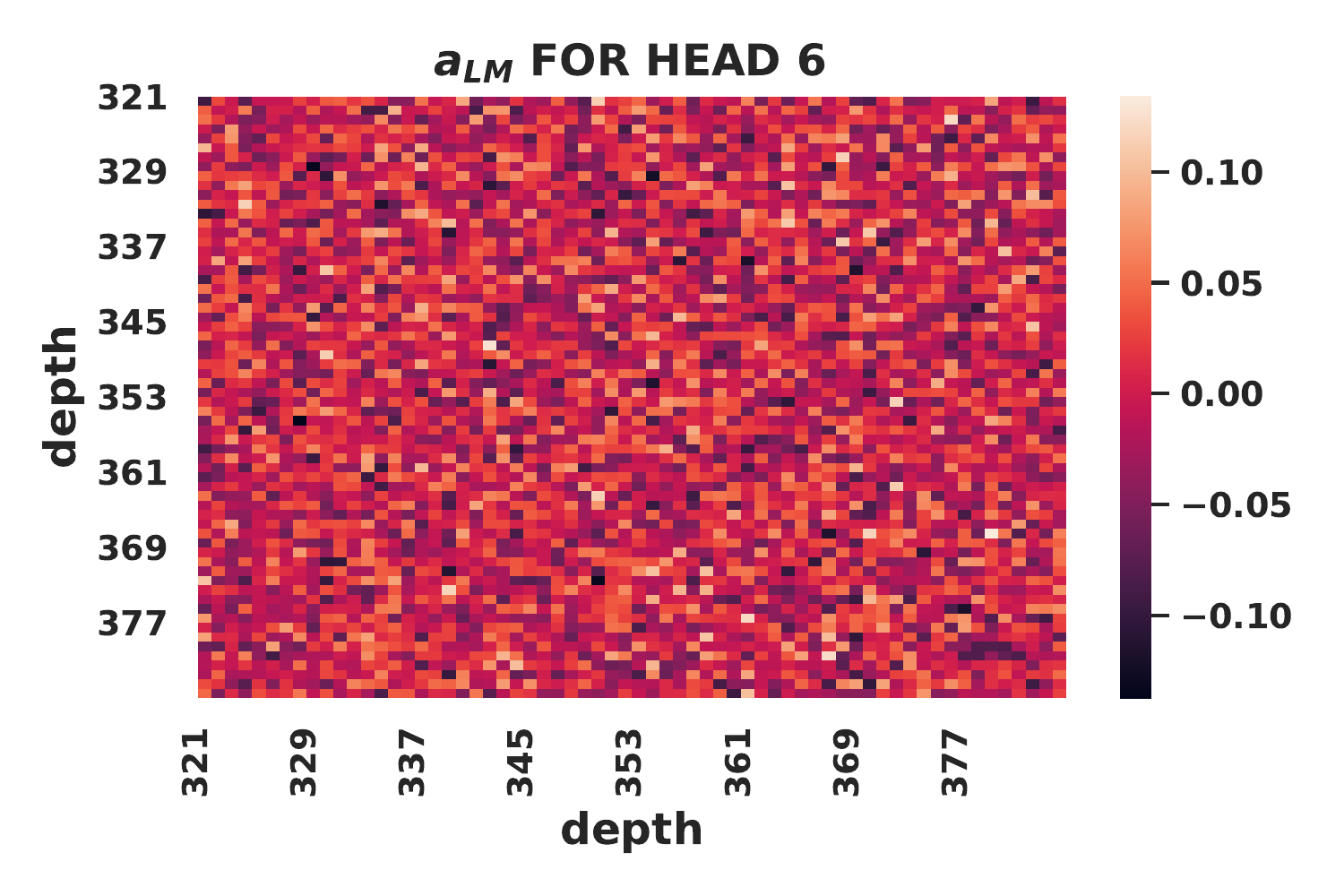}

\end{subfigure}
\hfill
\begin{subfigure}[b]{0.6\textwidth}
	\centering
	\includegraphics[width=1.1\textwidth]{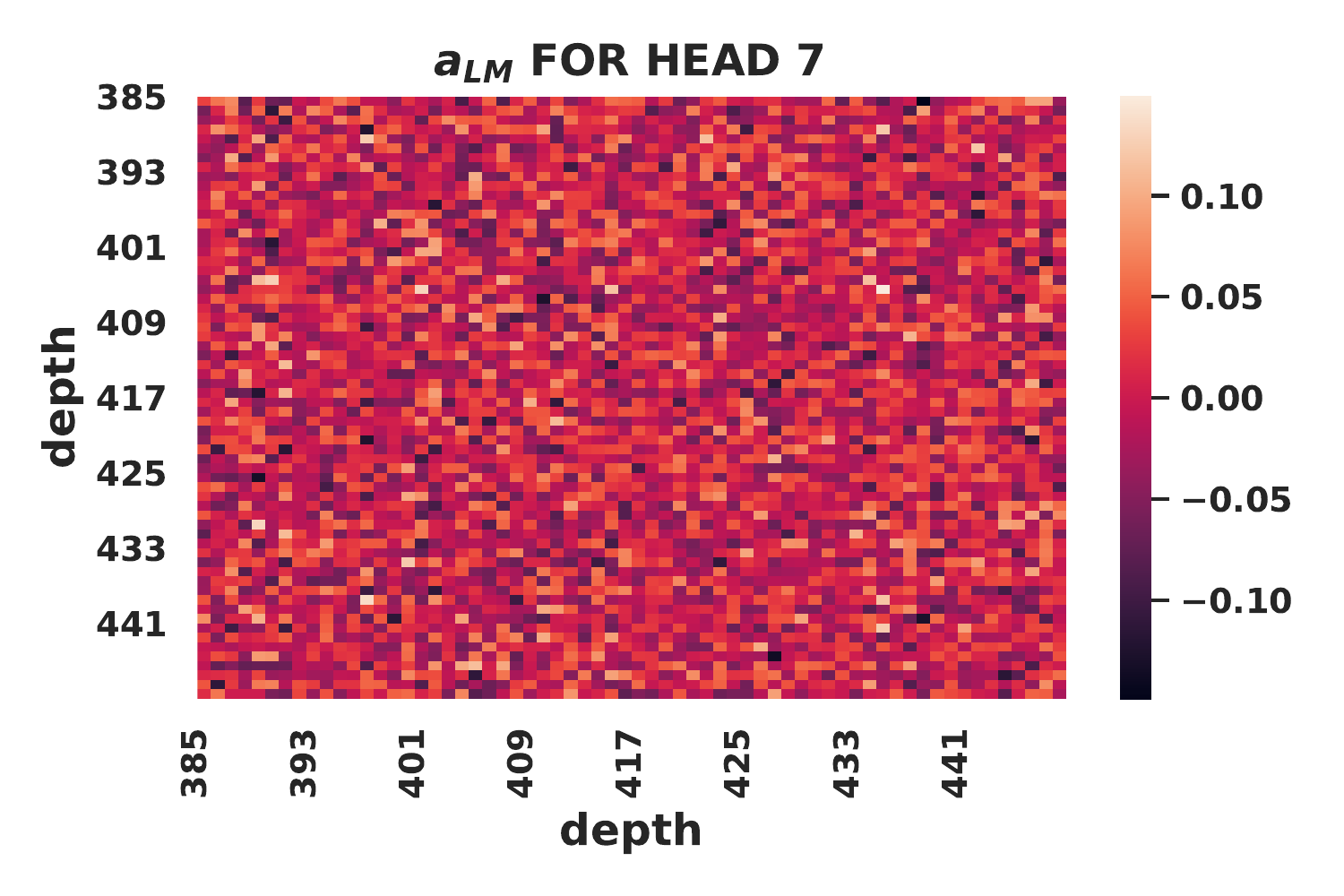}

\end{subfigure}
\hfill
\begin{subfigure}[b]{0.6\textwidth}
	\centering
	\includegraphics[width=1.1\textwidth]{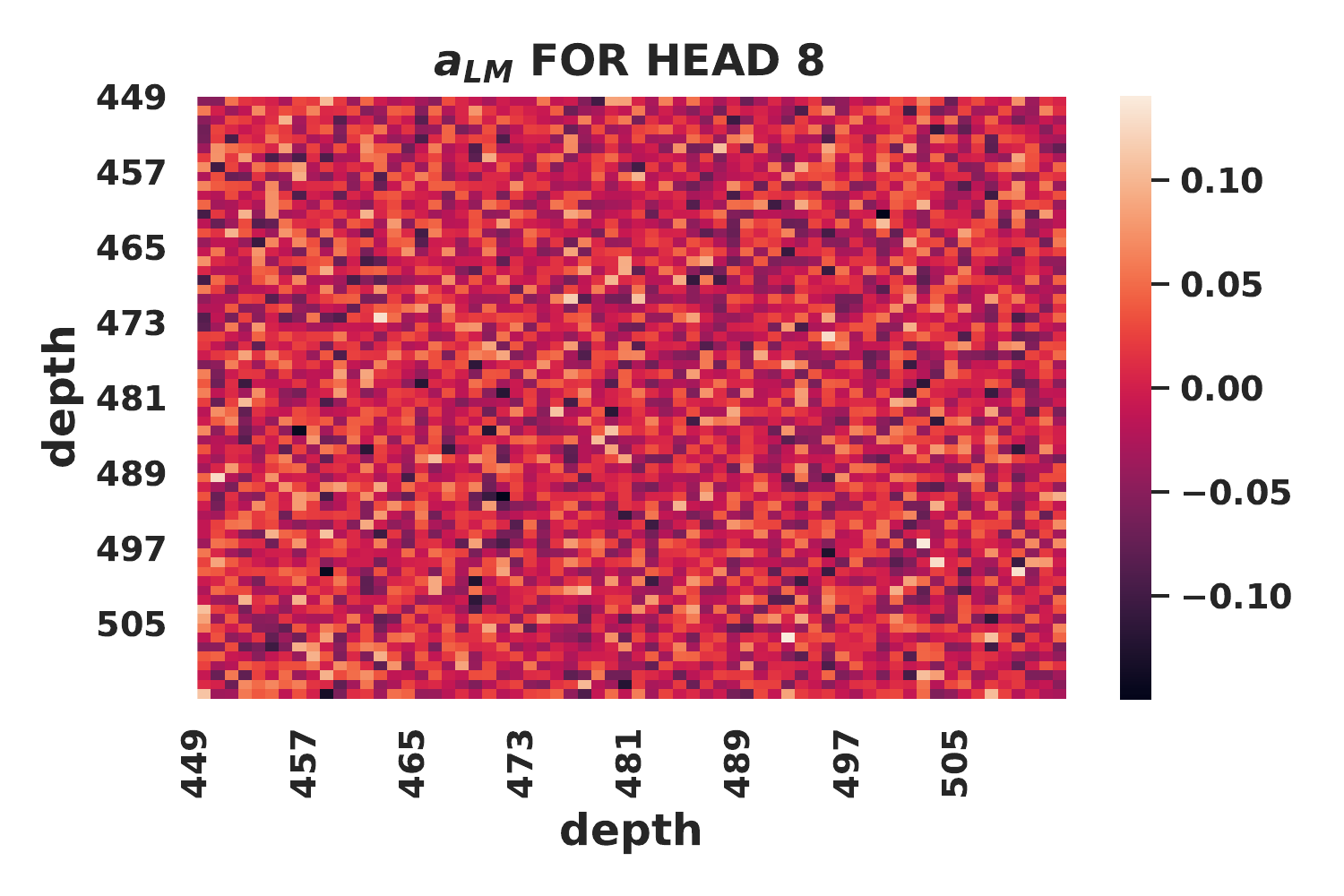}

\end{subfigure}
\caption{$\va_{LM}$ heatmap plots for all heads from XLM attention stage from Graph transformer model \#2.}
\label{fig4apx}
\end{adjustwidth}
\end{figure}    

\clearpage
\thispagestyle{headings}
\begin{figure}
\begin{adjustwidth}{-5em}{-5em}
\centering
\begin{subfigure}[b]{0.6\textwidth}
	\centering
	\includegraphics[width=1.1\textwidth]{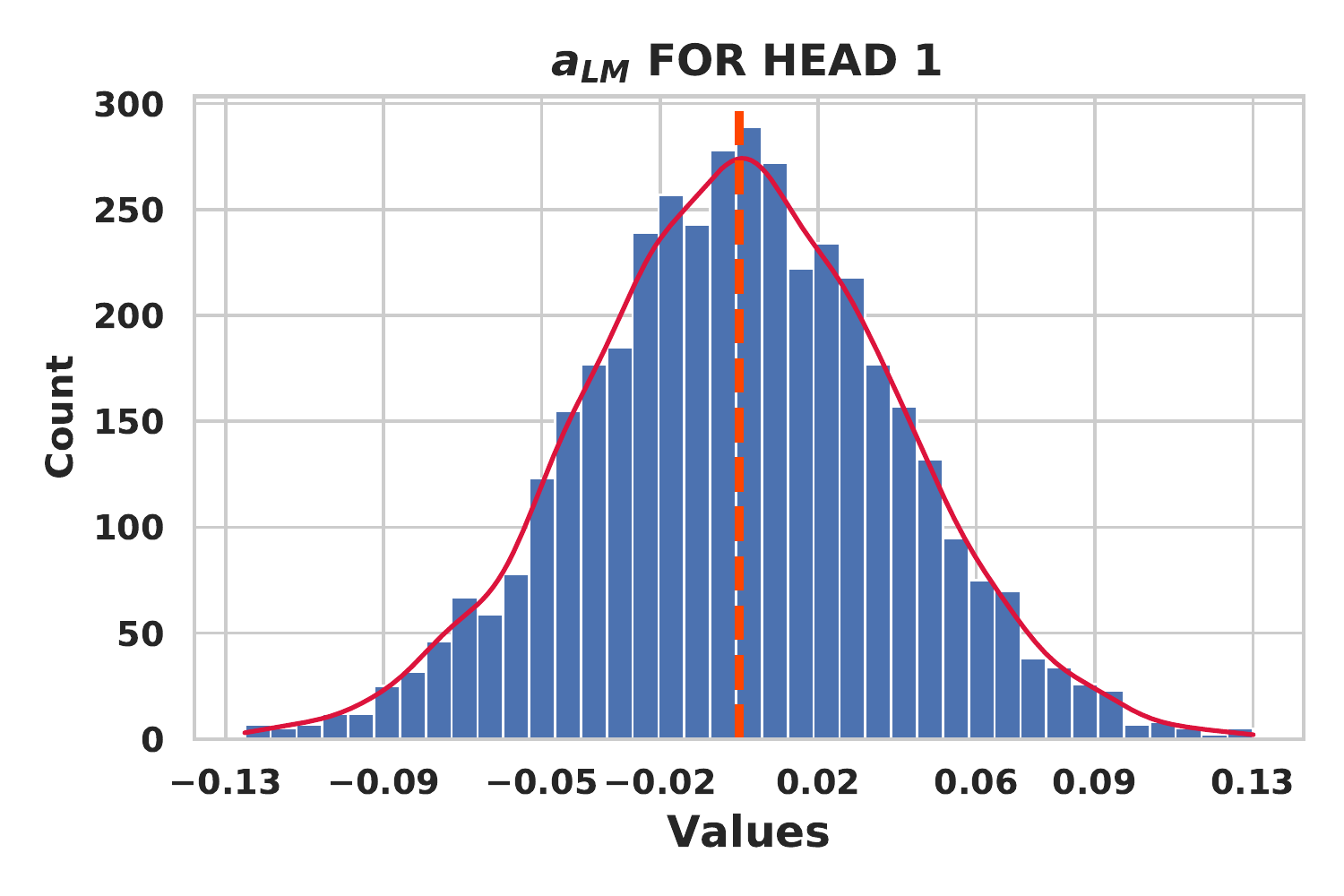}

\end{subfigure}
\hfill
\begin{subfigure}[b]{0.6\textwidth}
	\centering
	\includegraphics[width=1.1\textwidth]{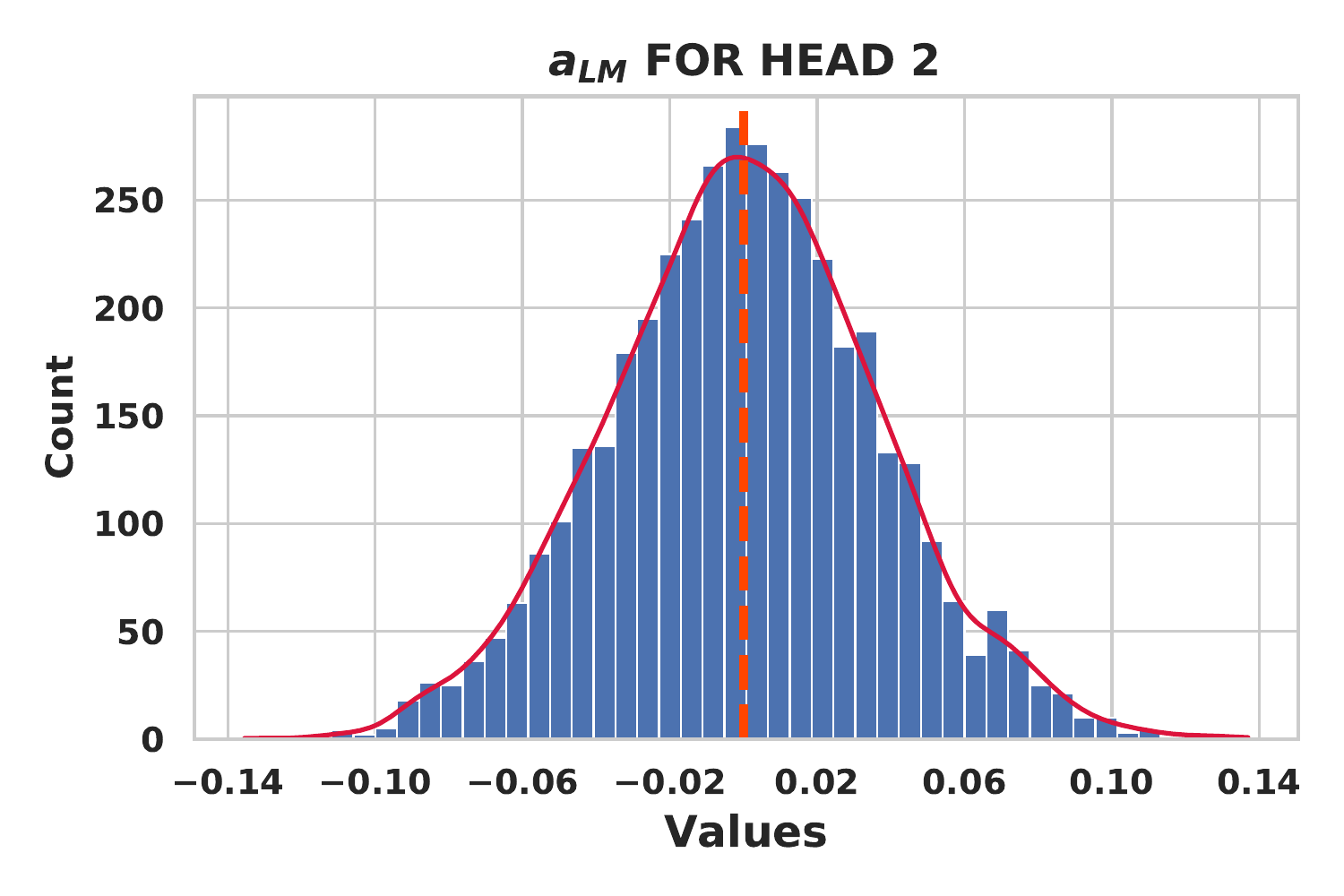}

\end{subfigure}
\hfill
\begin{subfigure}[b]{0.6\textwidth}
	\centering
	\includegraphics[width=1.1\textwidth]{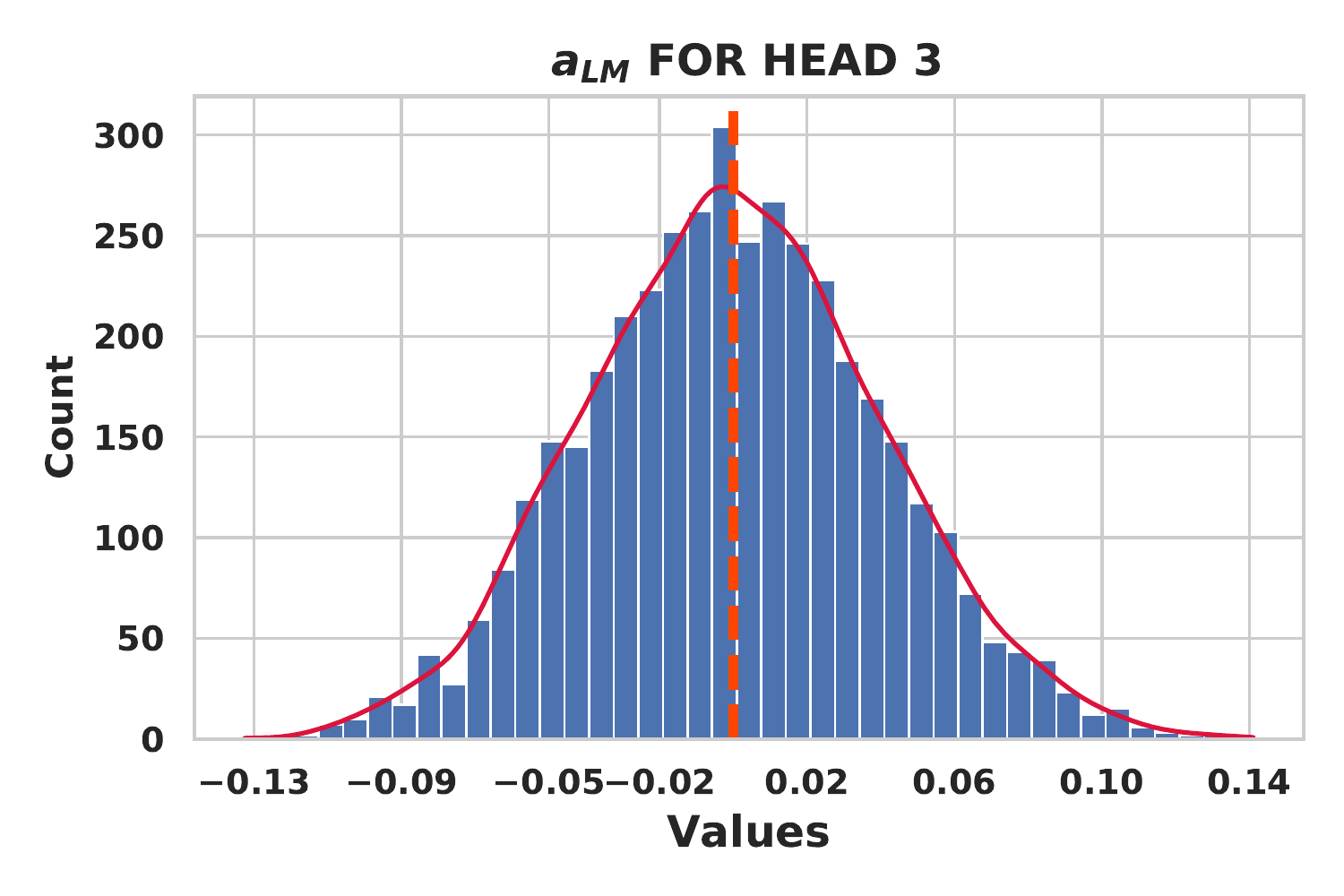}

\end{subfigure}
\hfill
\begin{subfigure}[b]{0.6\textwidth}
	\centering
	\includegraphics[width=1.1\textwidth]{Xhistmodel2avech4}

\end{subfigure}
\centering
\begin{subfigure}[b]{0.6\textwidth}
	\centering
	\includegraphics[width=1.1\textwidth]{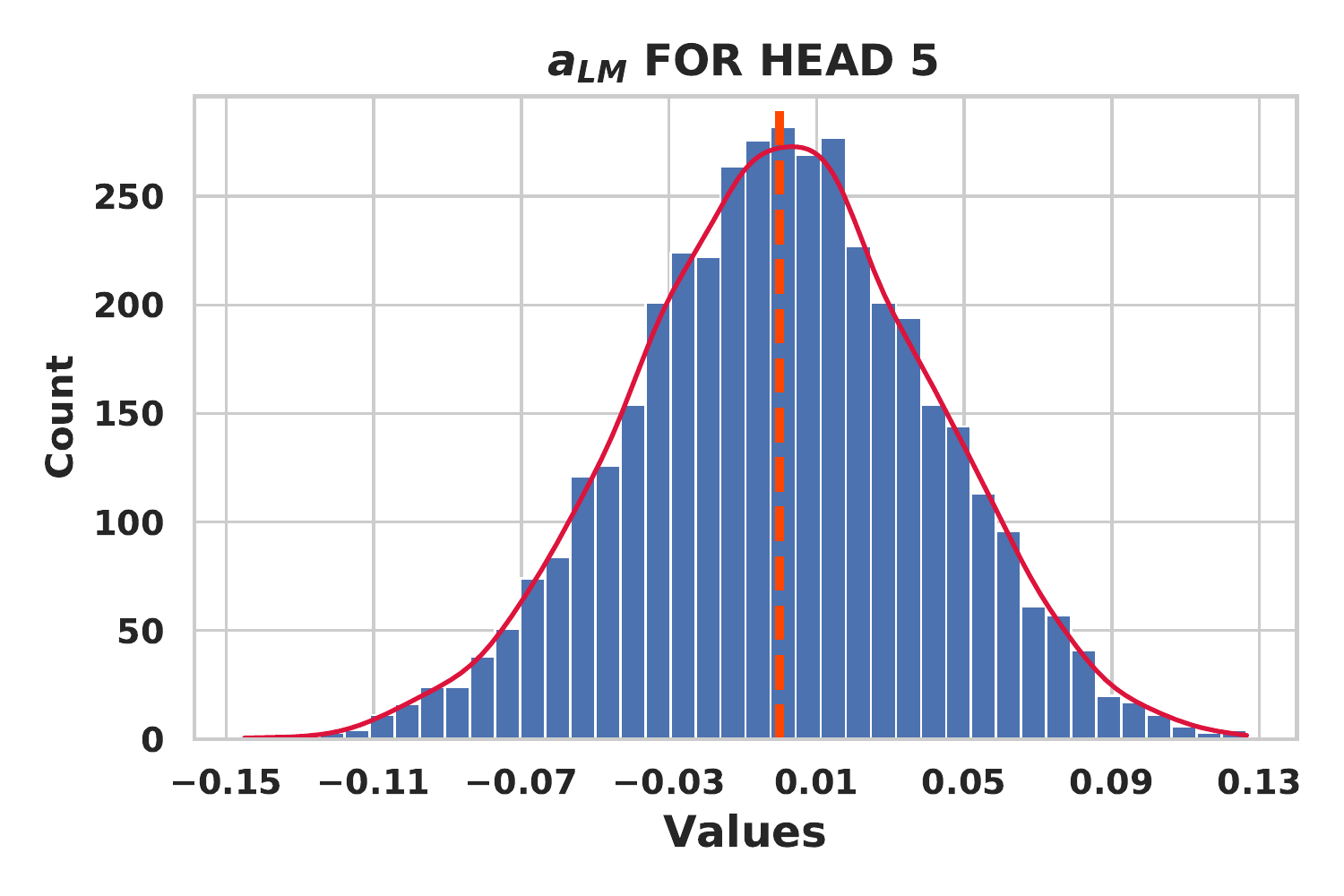}

\end{subfigure}
\hfill
\begin{subfigure}[b]{0.6\textwidth}
	\centering
	\includegraphics[width=1.1\textwidth]{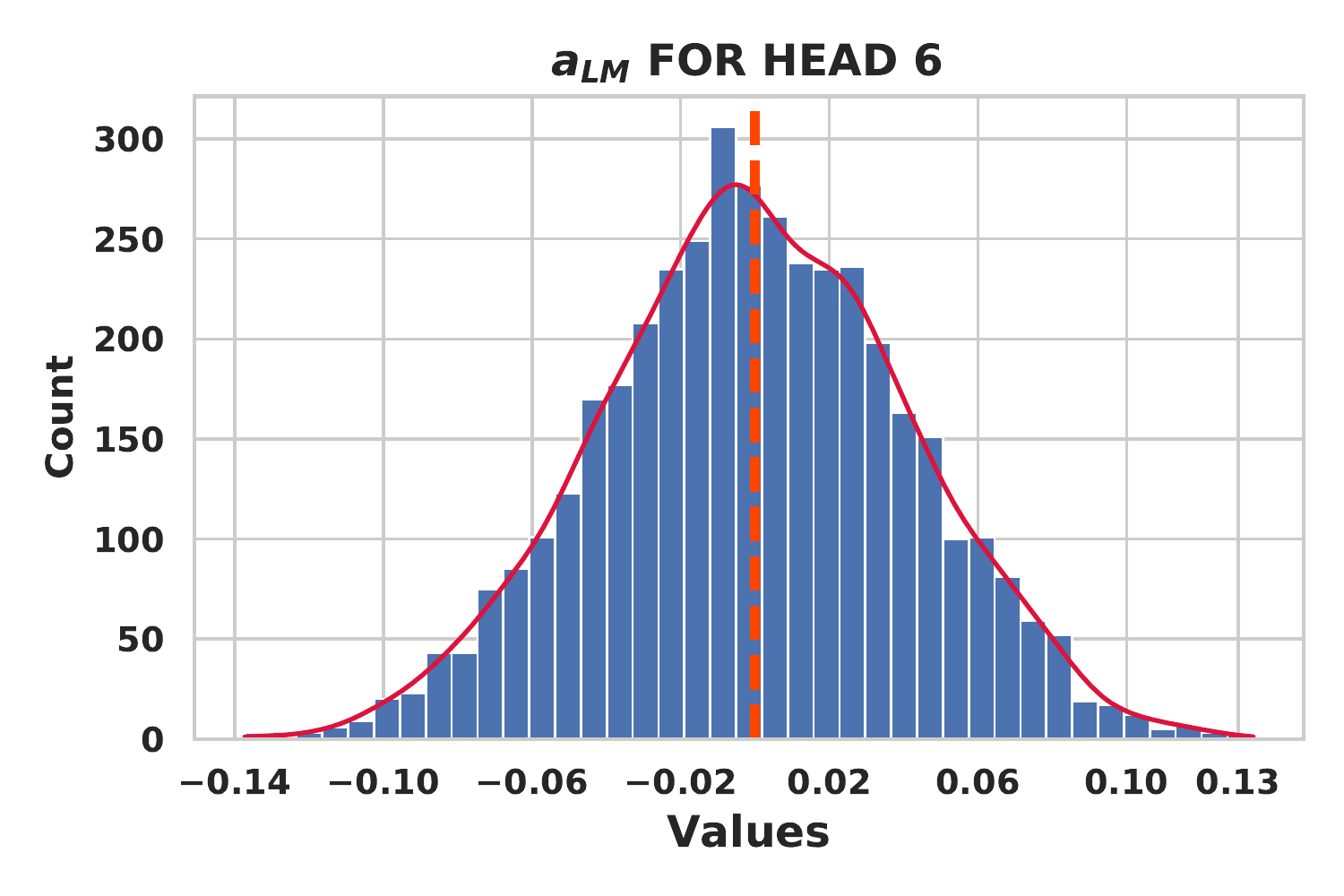}

\end{subfigure}
\hfill
\begin{subfigure}[b]{0.6\textwidth}
	\centering
	\includegraphics[width=1.1\textwidth]{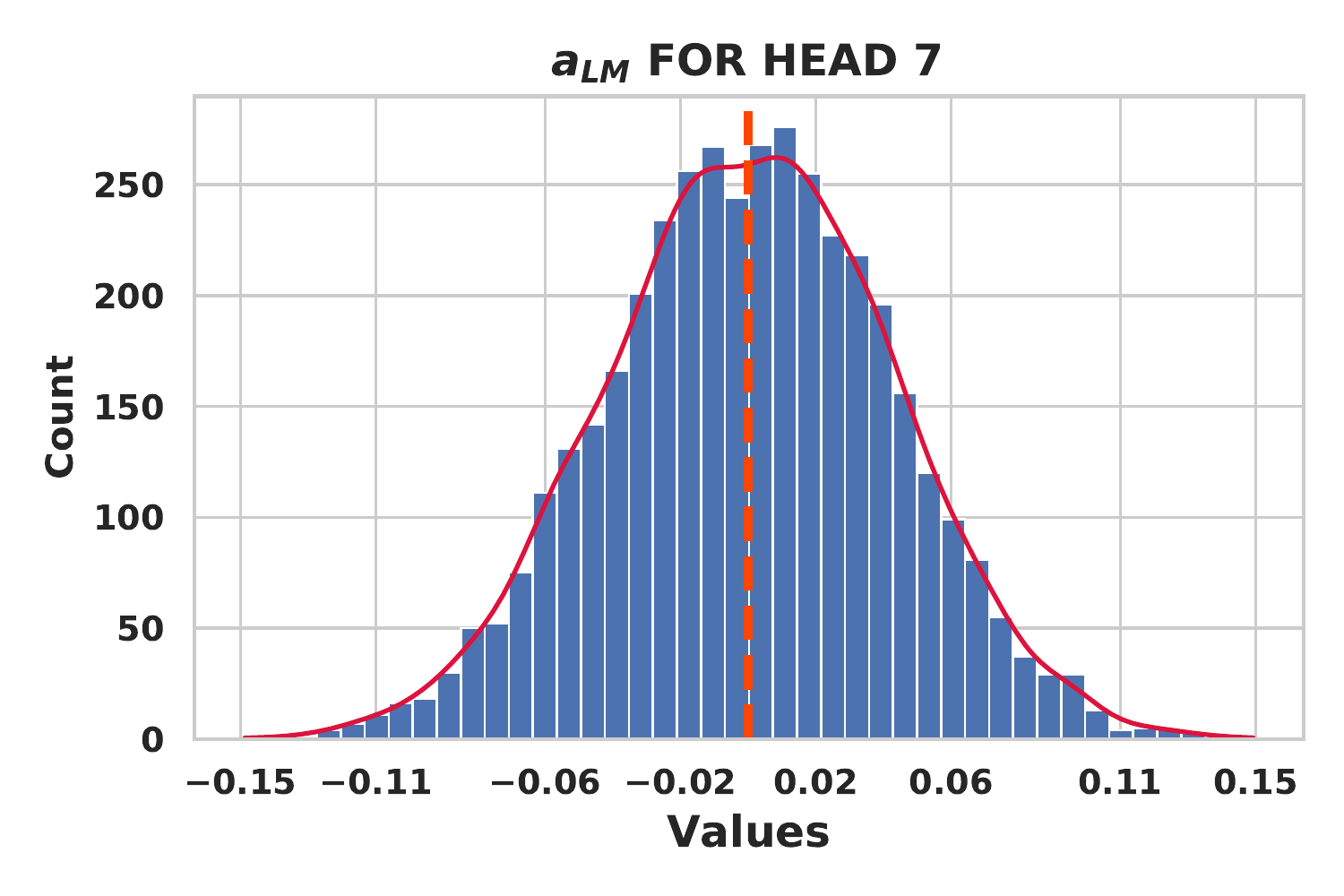}

\end{subfigure}
\hfill
\begin{subfigure}[b]{0.6\textwidth}
	\centering
	\includegraphics[width=1.1\textwidth]{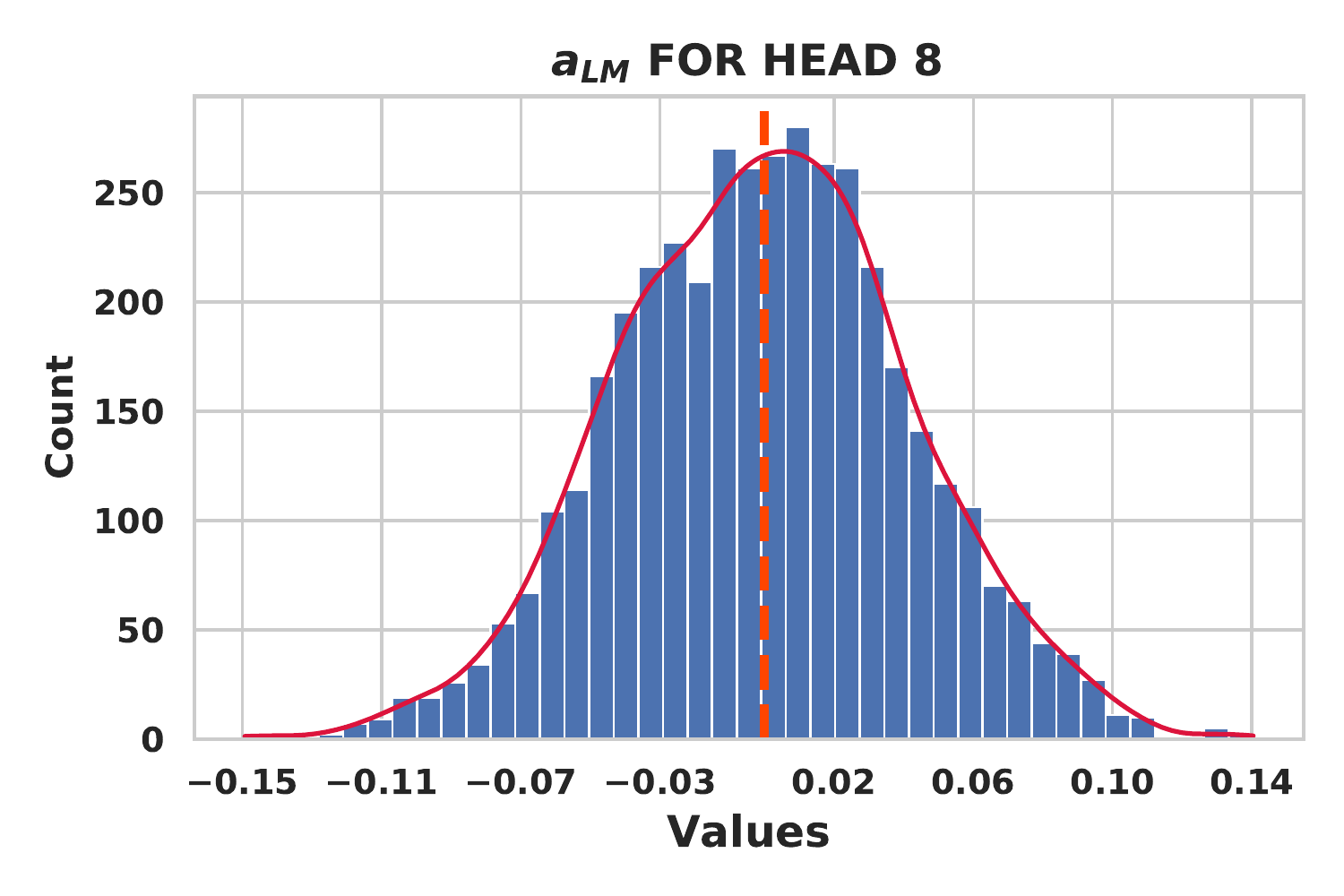}

\end{subfigure}
\caption{$\va_{LM}$ histogram plots for all heads from XLM attention stage from Graph transformer model \#2. Dashed line in orange marks zero value.}
\label{fig5apx}
\end{adjustwidth}
\end{figure}  

\clearpage
\thispagestyle{headings}
\begin{figure}
\begin{adjustwidth}{-5em}{-5em}
\centering
\begin{subfigure}[b]{0.6\textwidth}
	\centering
	\includegraphics[width=1.1\textwidth]{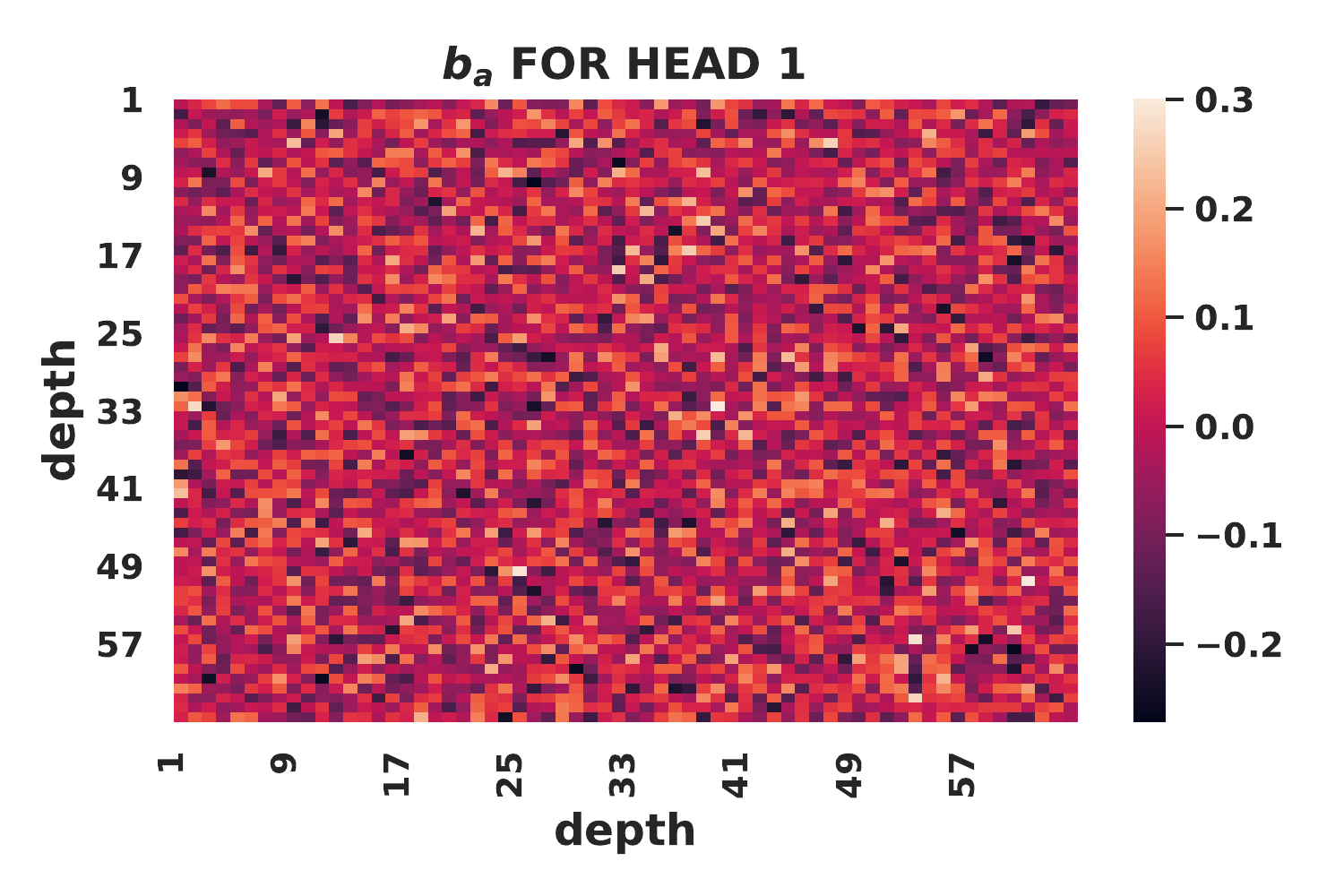}

\end{subfigure}
\hfill
\begin{subfigure}[b]{0.6\textwidth}
	\centering
	\includegraphics[width=1.1\textwidth]{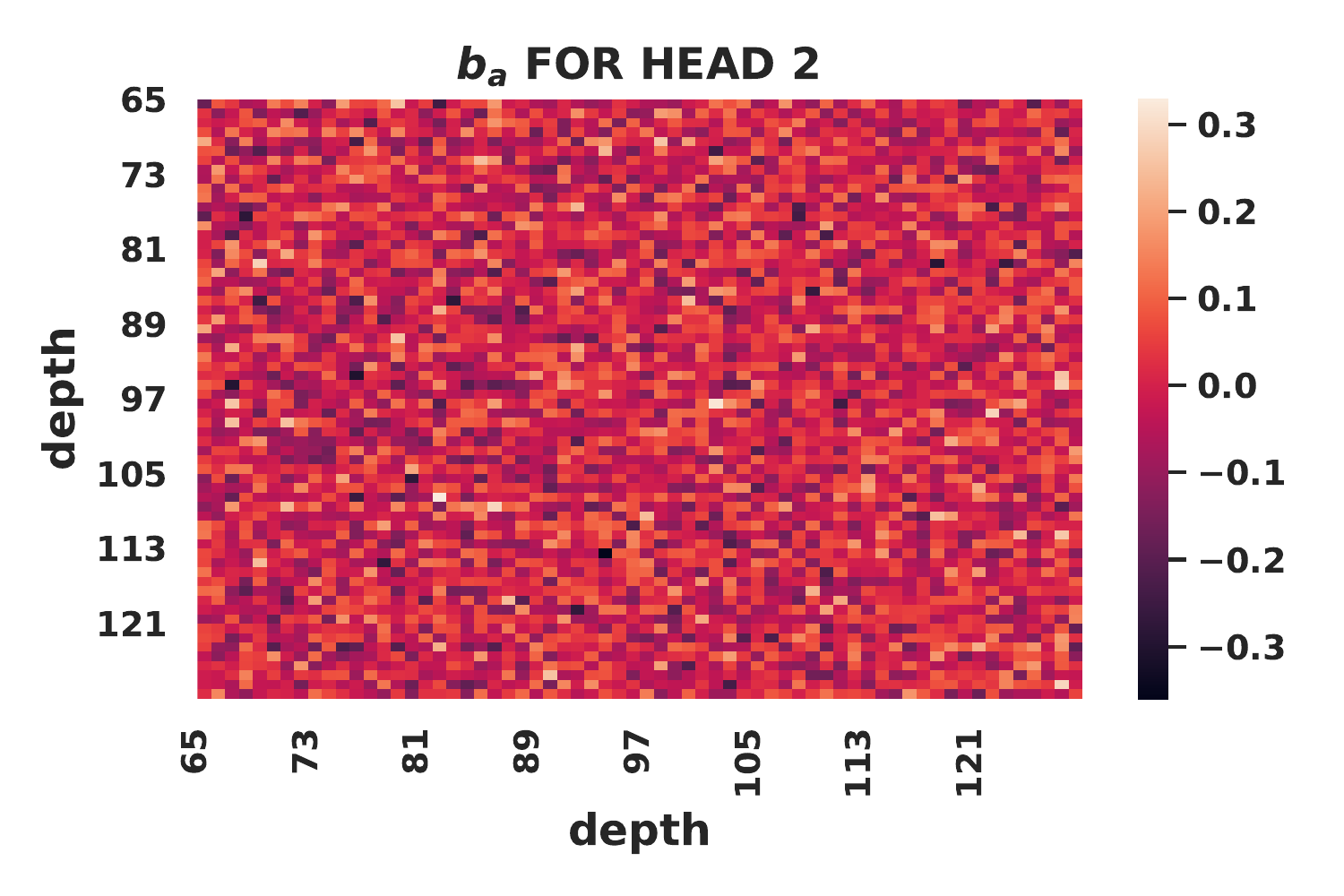}

\end{subfigure}
\hfill
\begin{subfigure}[b]{0.6\textwidth}
	\centering
	\includegraphics[width=1.1\textwidth]{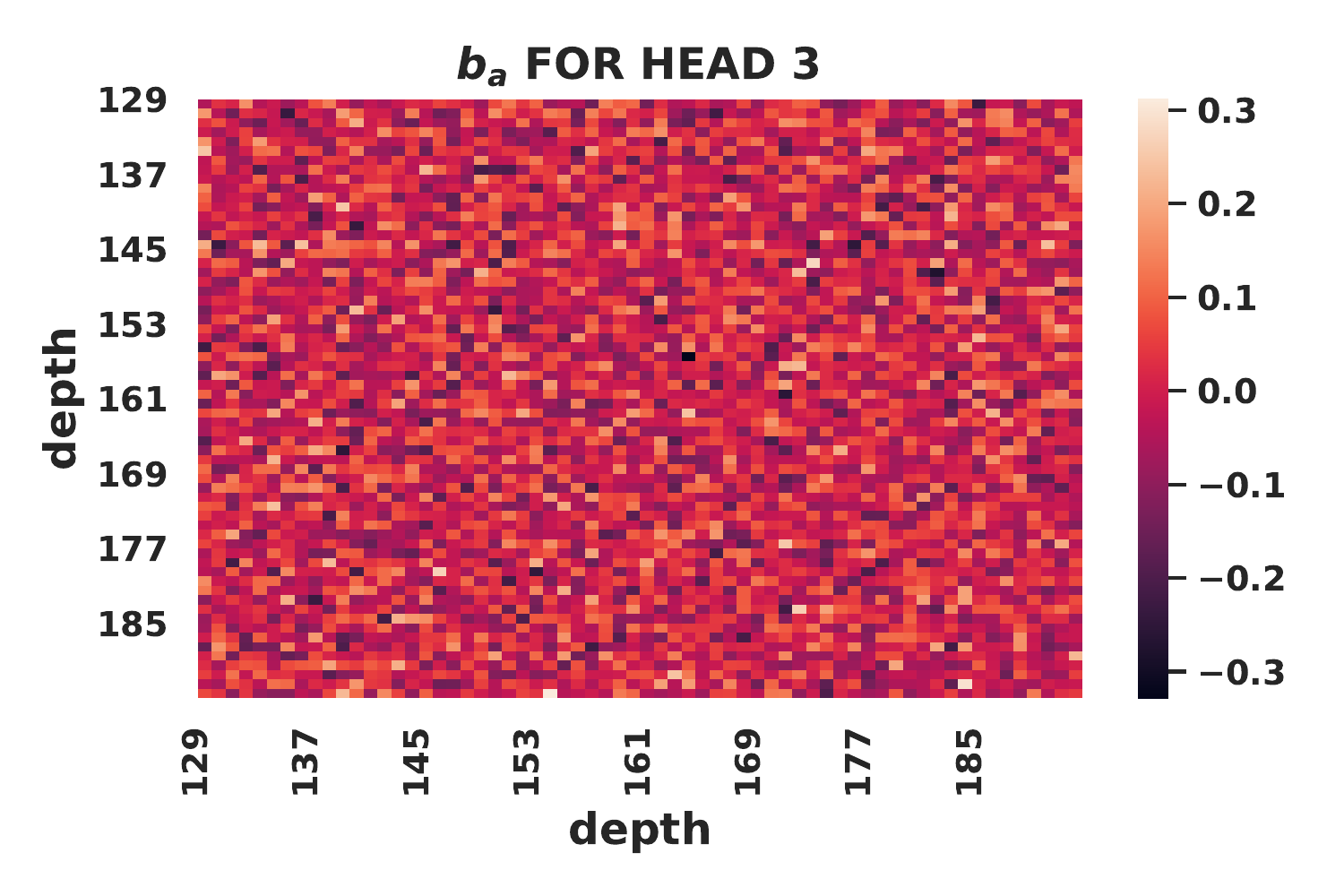}

\end{subfigure}
\hfill
\begin{subfigure}[b]{0.6\textwidth}
	\centering
	\includegraphics[width=1.1\textwidth]{Xheatmapmodel2bah4}

\end{subfigure}
\centering
\begin{subfigure}[b]{0.6\textwidth}
	\centering
	\includegraphics[width=1.1\textwidth]{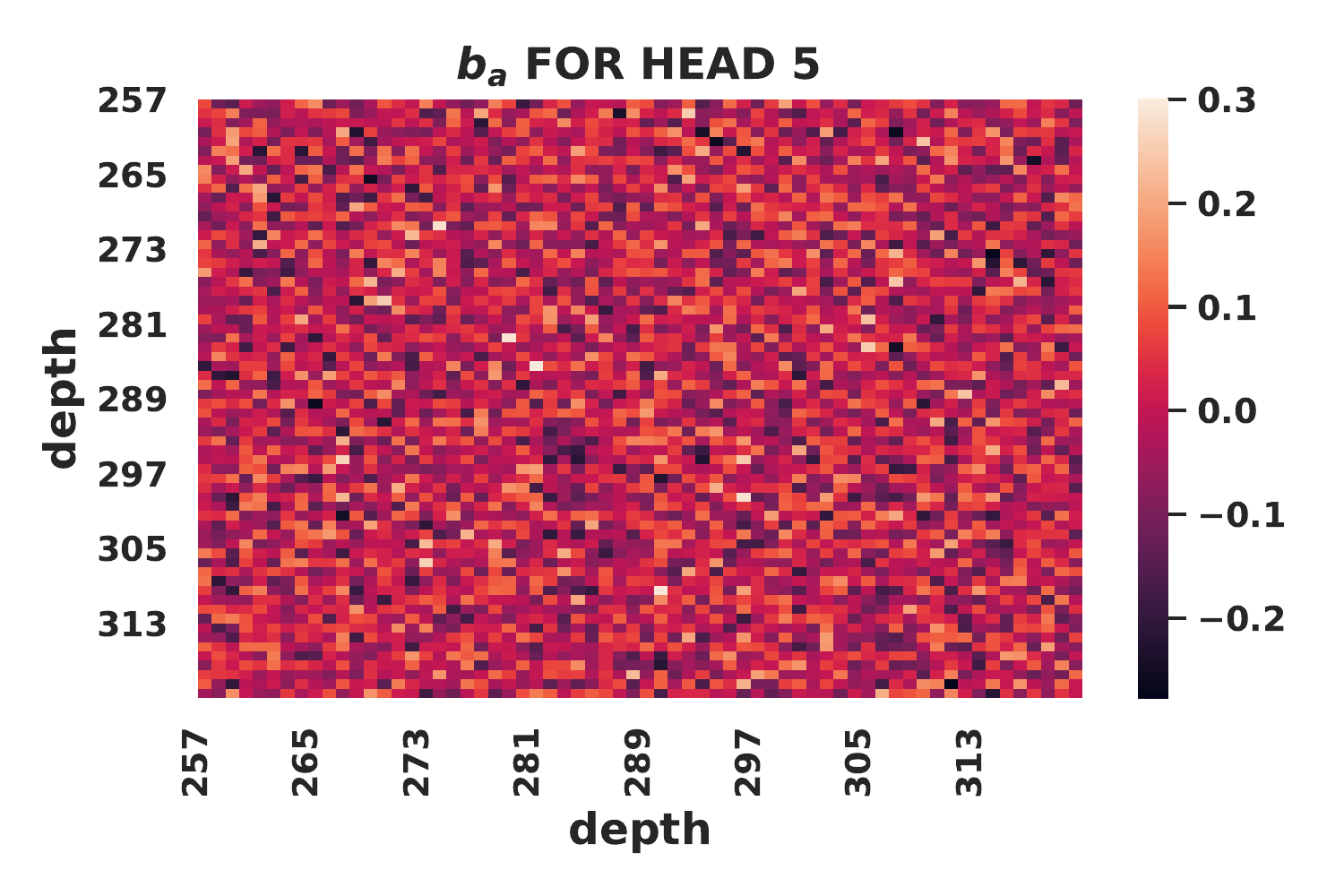}

\end{subfigure}
\hfill
\begin{subfigure}[b]{0.6\textwidth}
	\centering
	\includegraphics[width=1.1\textwidth]{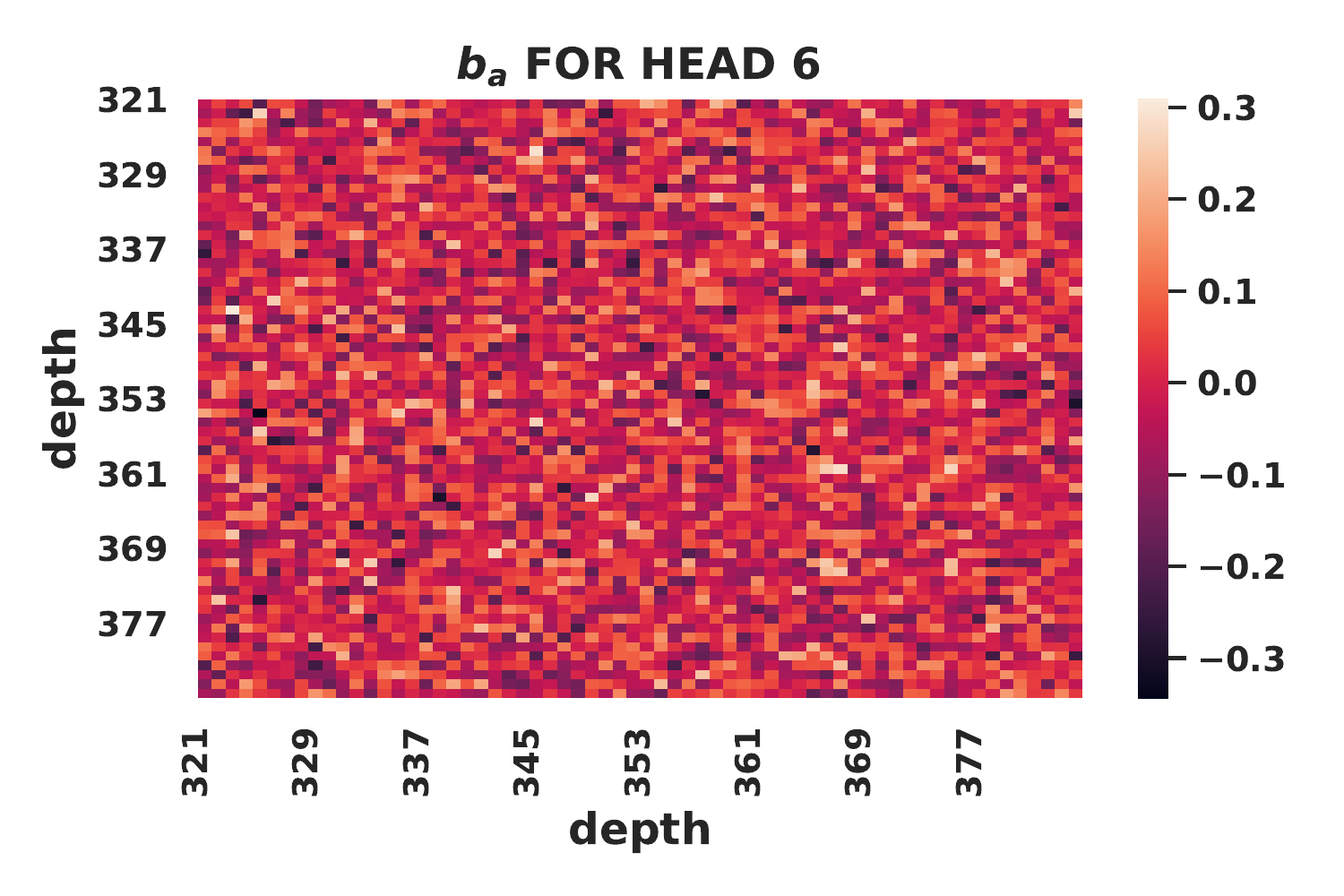}

\end{subfigure}
\hfill
\begin{subfigure}[b]{0.6\textwidth}
	\centering
	\includegraphics[width=1.1\textwidth]{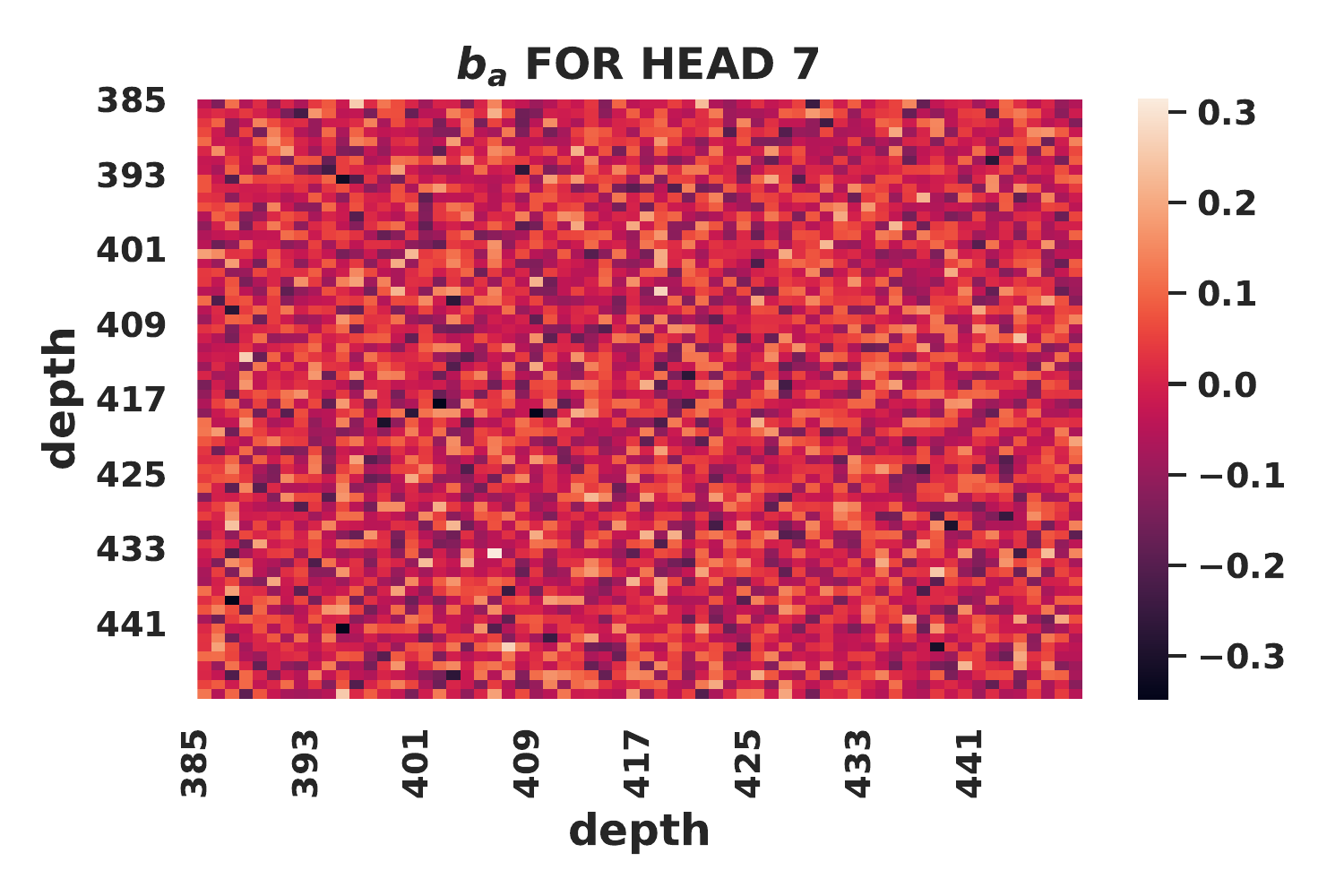}

\end{subfigure}
\hfill
\begin{subfigure}[b]{0.6\textwidth}
	\centering
	\includegraphics[width=1.1\textwidth]{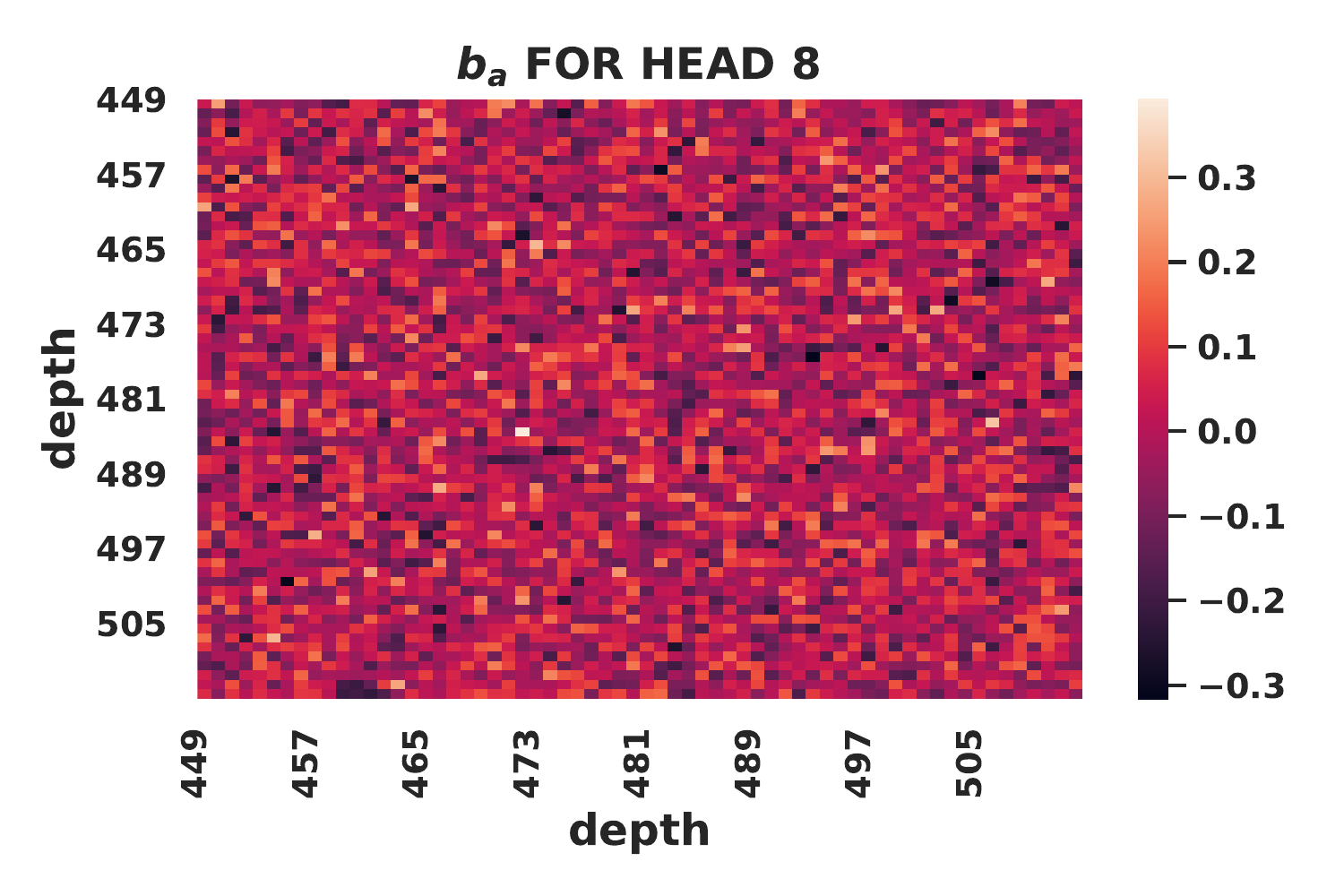}

\end{subfigure}
\caption{$\vb_{a}$ heatmap plots for all heads from XLM attention stage from graph transformer model \#2.}
\label{fig6apx}
\end{adjustwidth}
\end{figure}

\clearpage
\thispagestyle{headings}
\begin{figure}
\begin{adjustwidth}{-5em}{-5em}
\centering
\begin{subfigure}[b]{0.6\textwidth}
	\centering
	\includegraphics[width=1.1\textwidth]{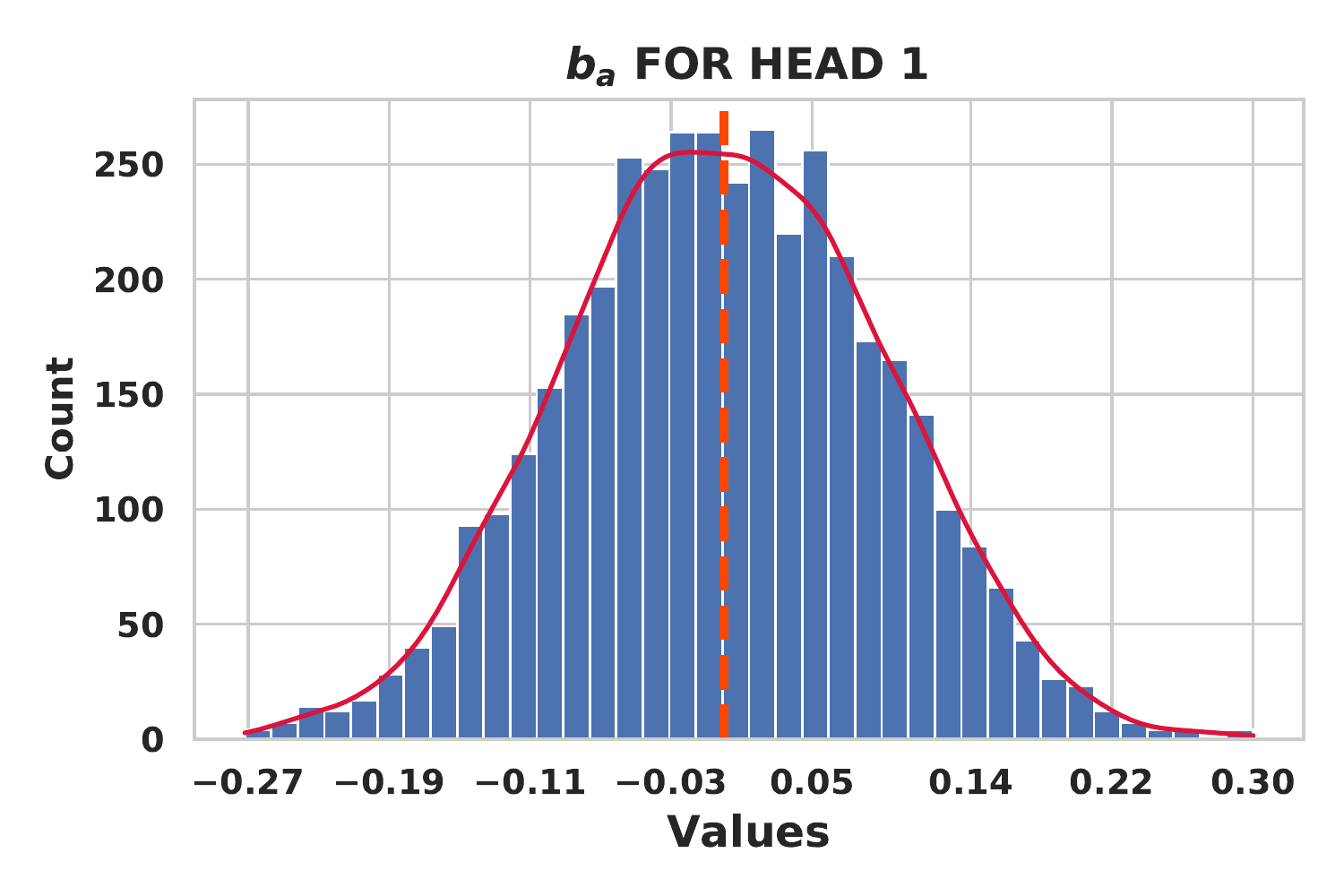}

\end{subfigure}
\hfill
\begin{subfigure}[b]{0.6\textwidth}
	\centering
	\includegraphics[width=1.1\textwidth]{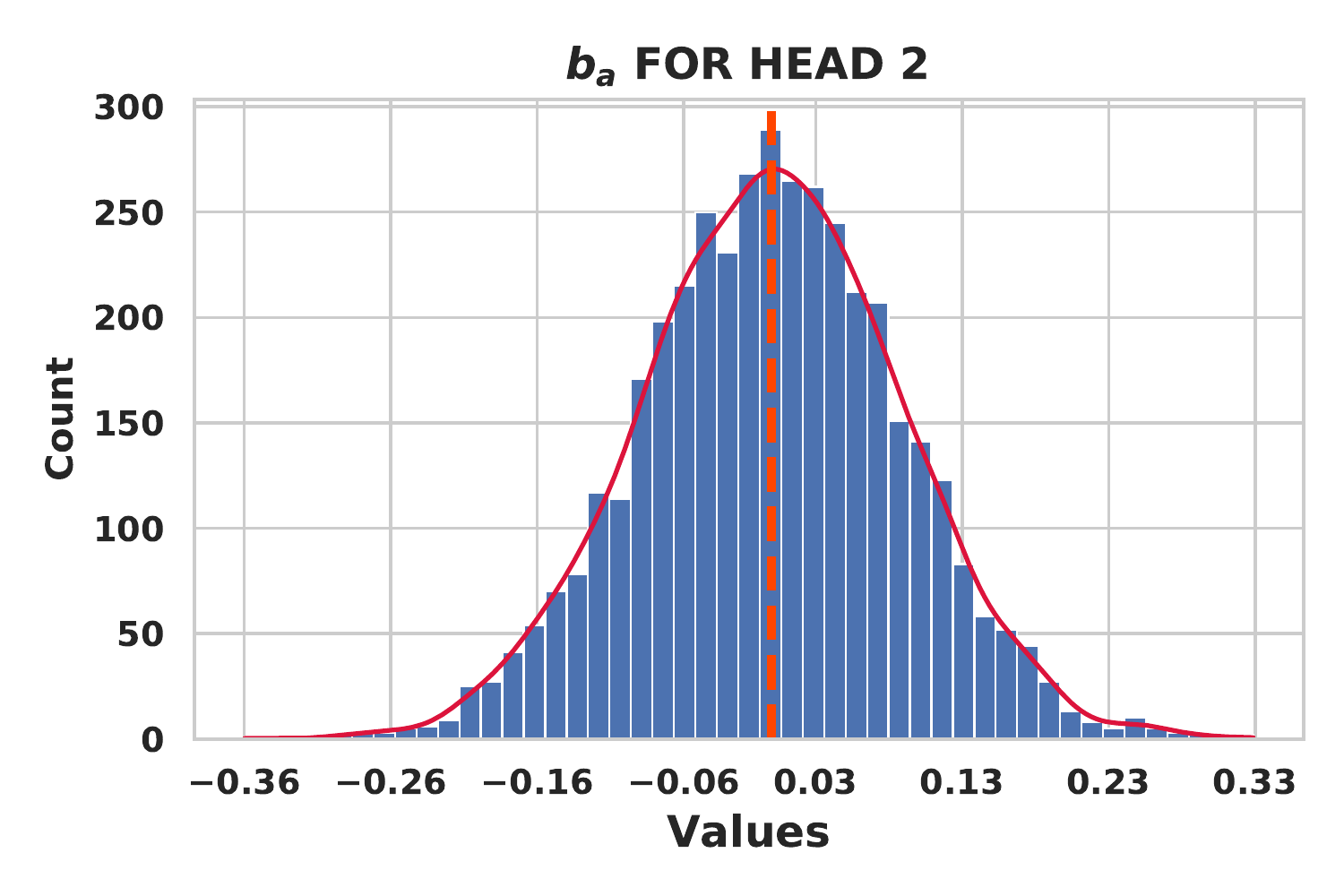}

\end{subfigure}
\hfill
\begin{subfigure}[b]{0.6\textwidth}
	\centering
	\includegraphics[width=1.1\textwidth]{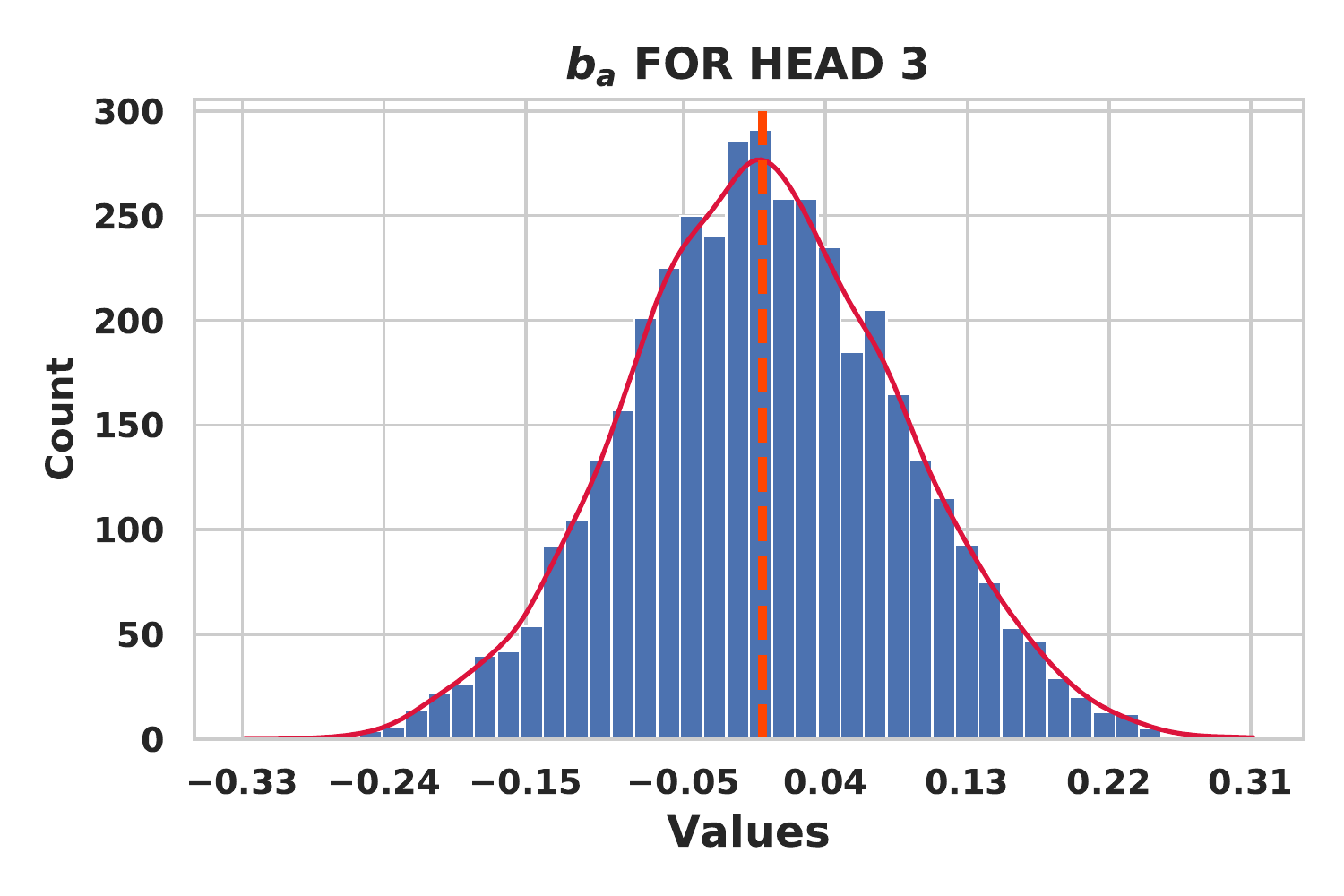}

\end{subfigure}
\hfill
\begin{subfigure}[b]{0.6\textwidth}
	\centering
	\includegraphics[width=1.1\textwidth]{Xhistmodel2bah4}

\end{subfigure}
\centering
\begin{subfigure}[b]{0.6\textwidth}
	\centering
	\includegraphics[width=1.1\textwidth]{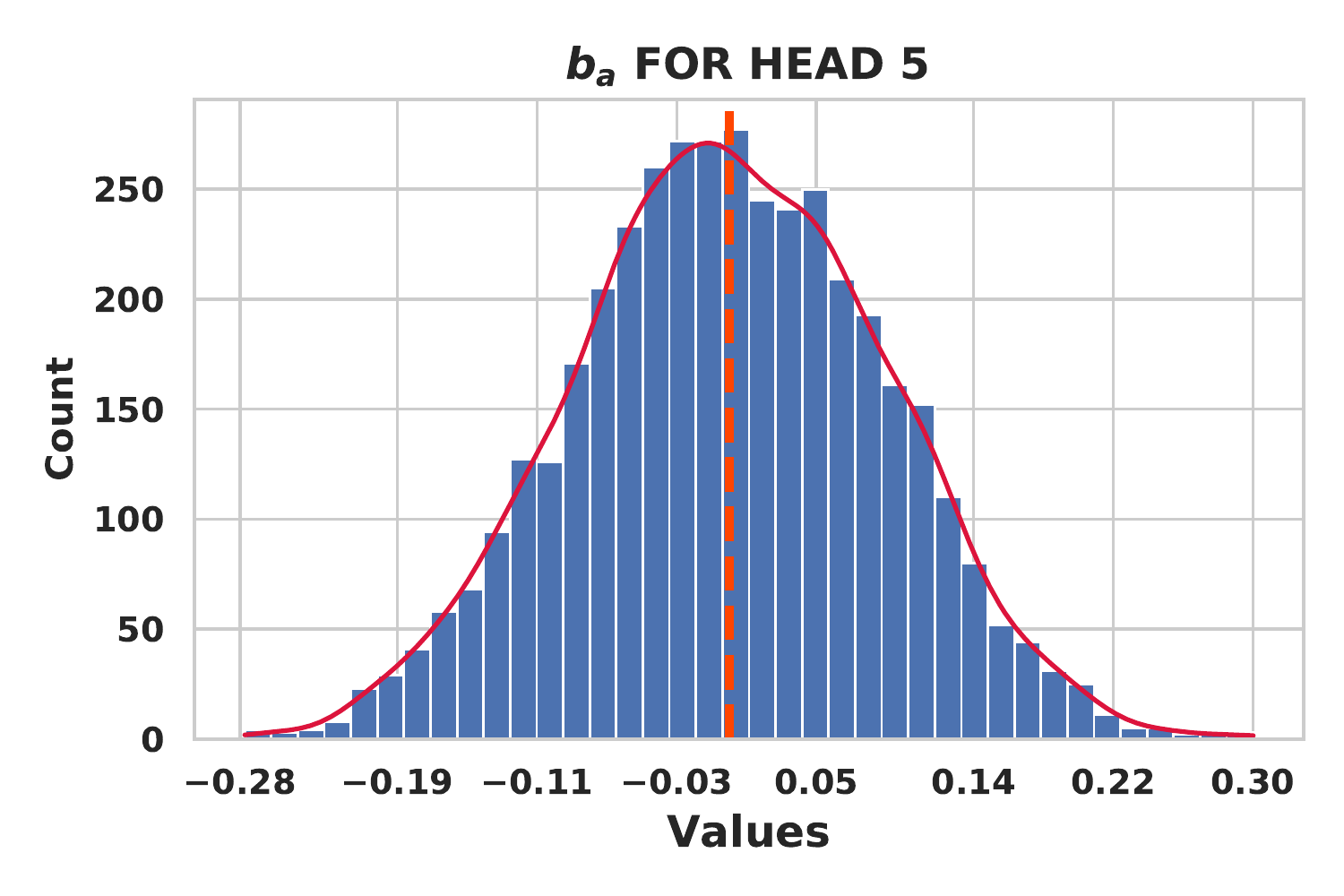}

\end{subfigure}
\hfill
\begin{subfigure}[b]{0.6\textwidth}
	\centering
	\includegraphics[width=1.1\textwidth]{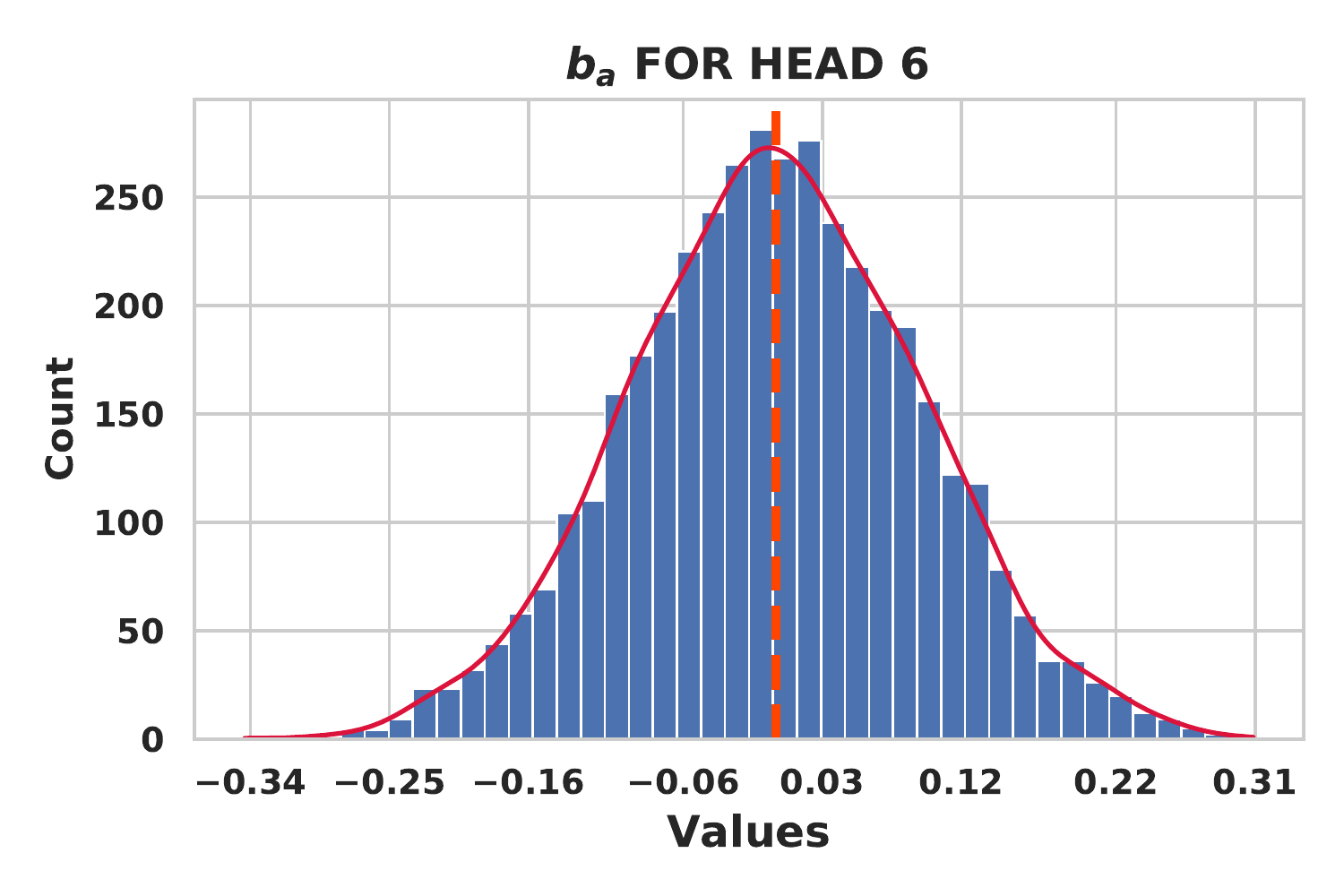}

\end{subfigure}
\hfill
\begin{subfigure}[b]{0.6\textwidth}
	\centering
	\includegraphics[width=1.1\textwidth]{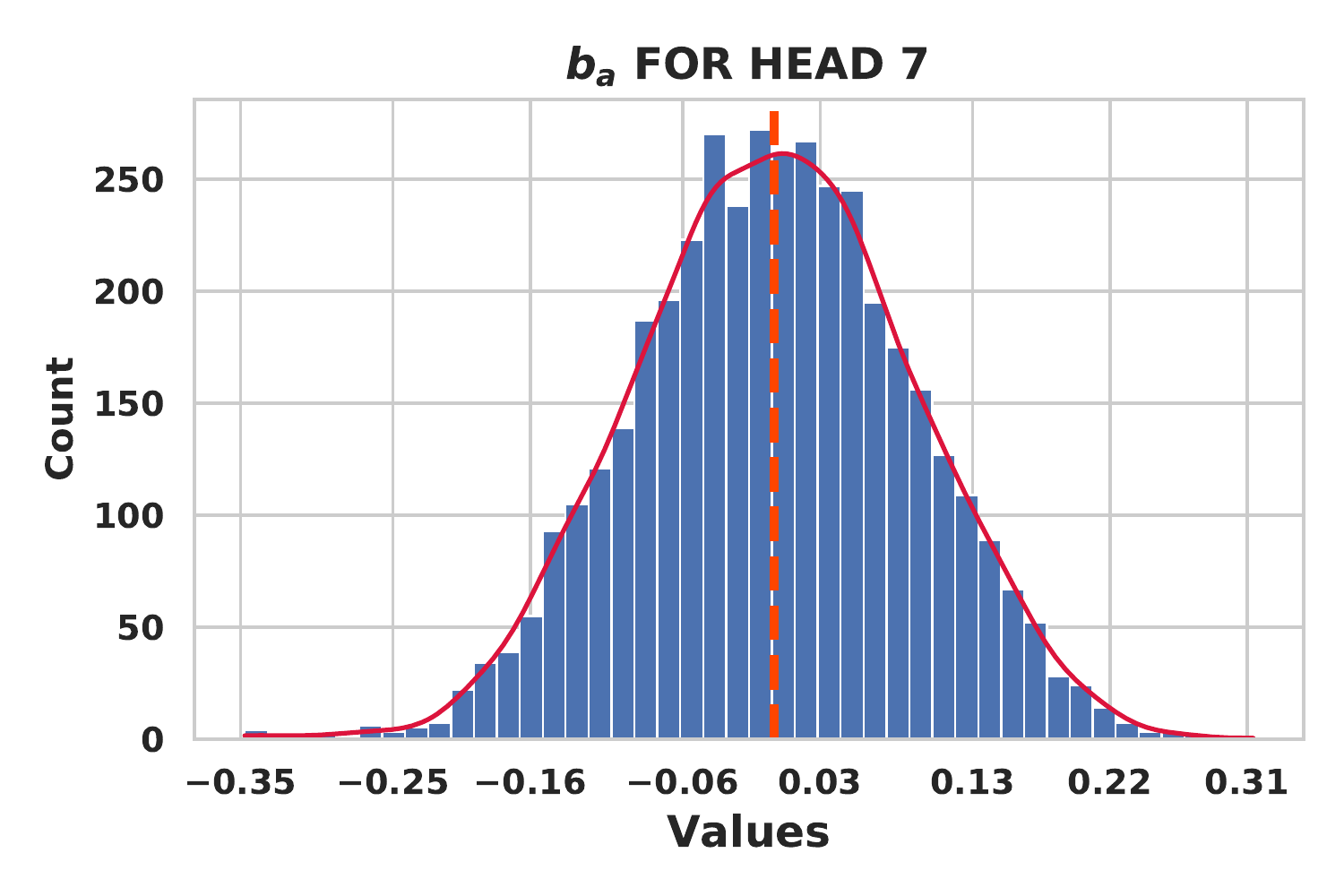}

\end{subfigure}
\hfill
\begin{subfigure}[b]{0.6\textwidth}
	\centering
	\includegraphics[width=1.1\textwidth]{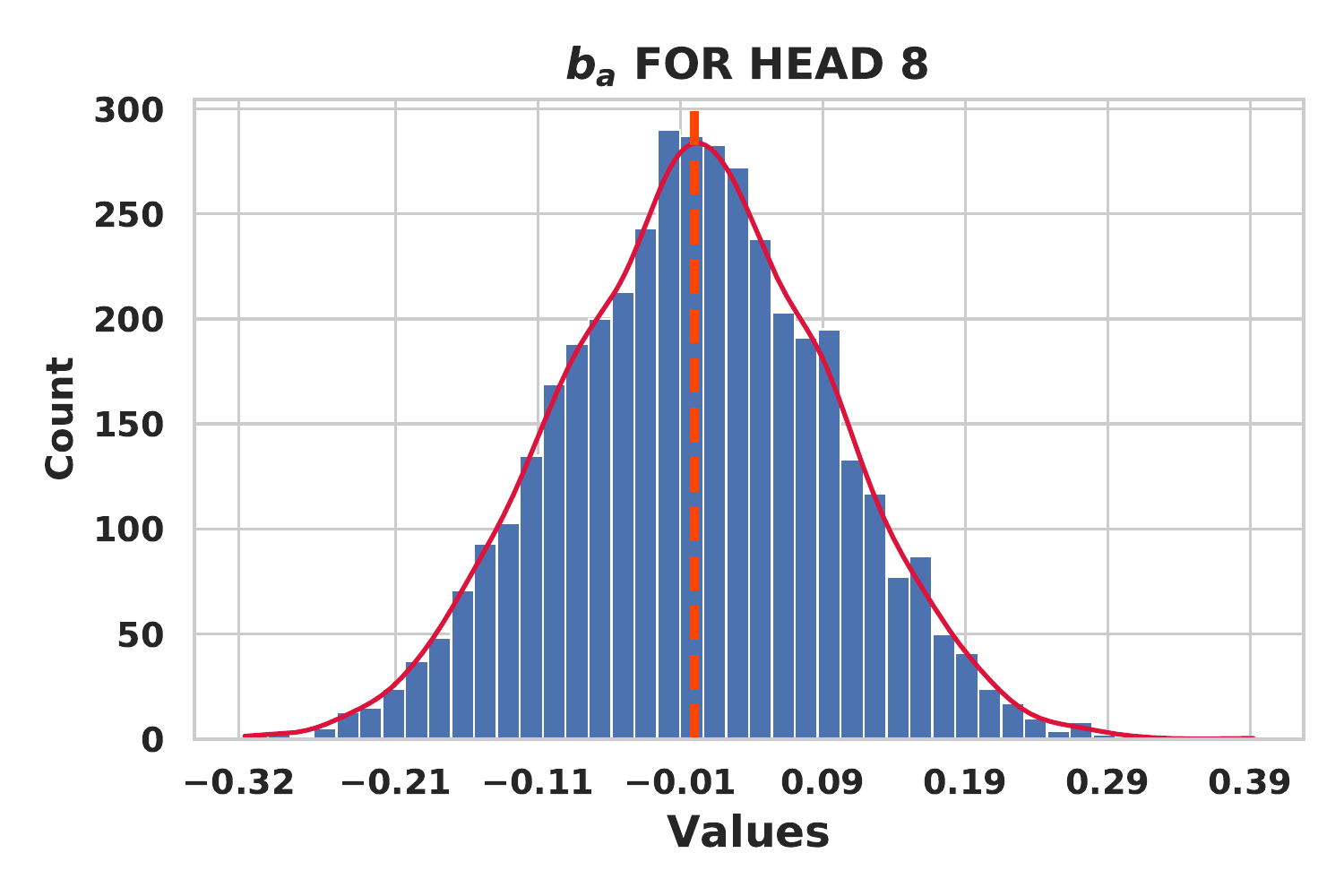}

\end{subfigure}
\caption{$\vb_{a}$ histogram plots for all heads from XLM attention stage from graph transformer model \#2. Dashed line in orange marks zero value.}
\label{fig7apx}
\end{adjustwidth}
\end{figure} 

\clearpage
\thispagestyle{headings}
\begin{figure}
\begin{adjustwidth}{-5em}{-5em}
\centering
\begin{subfigure}[b]{0.6\textwidth}
	\centering
	\includegraphics[width=1.1\textwidth]{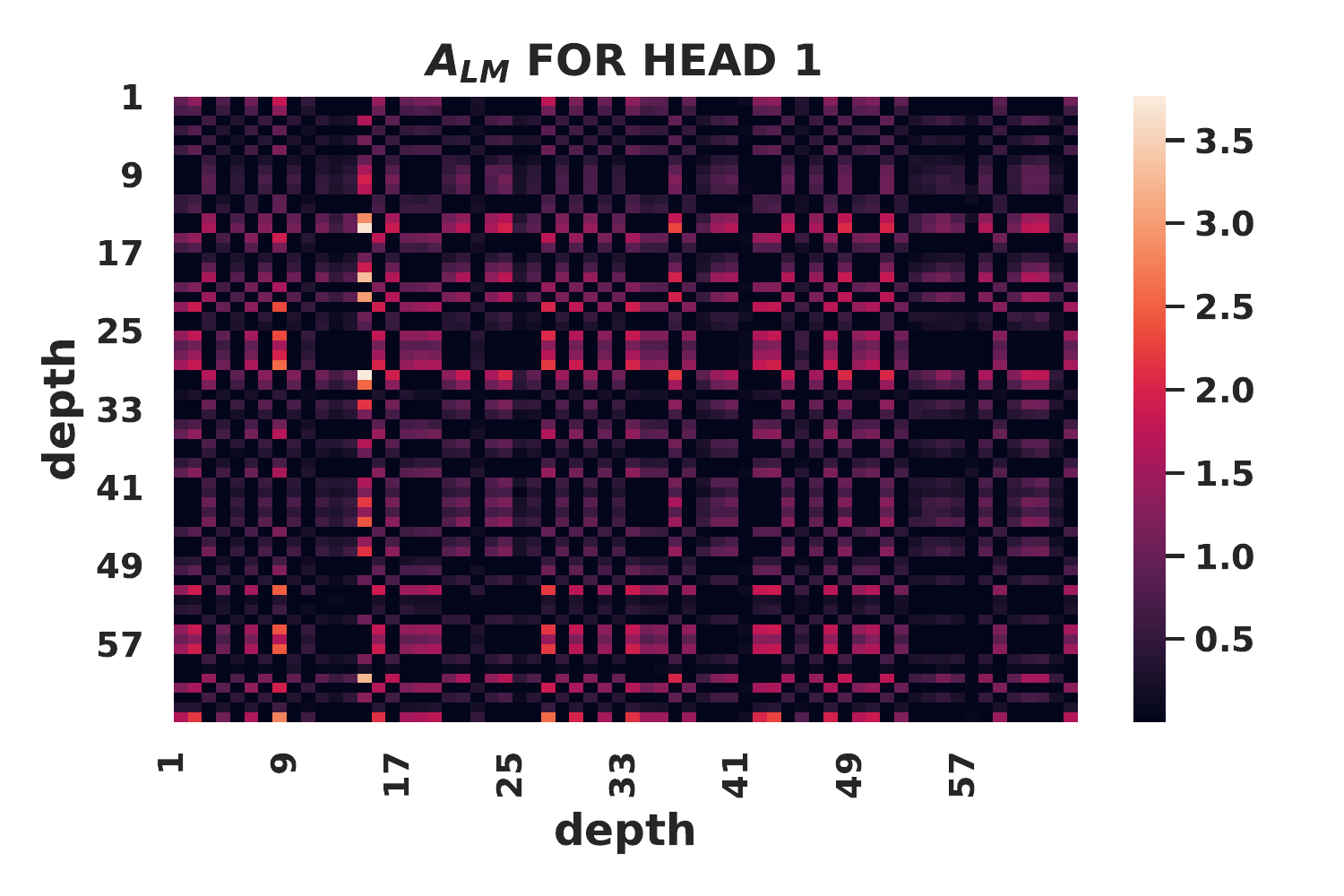}

\end{subfigure}
\hfill
\begin{subfigure}[b]{0.6\textwidth}
	\centering
	\includegraphics[width=1.1\textwidth]{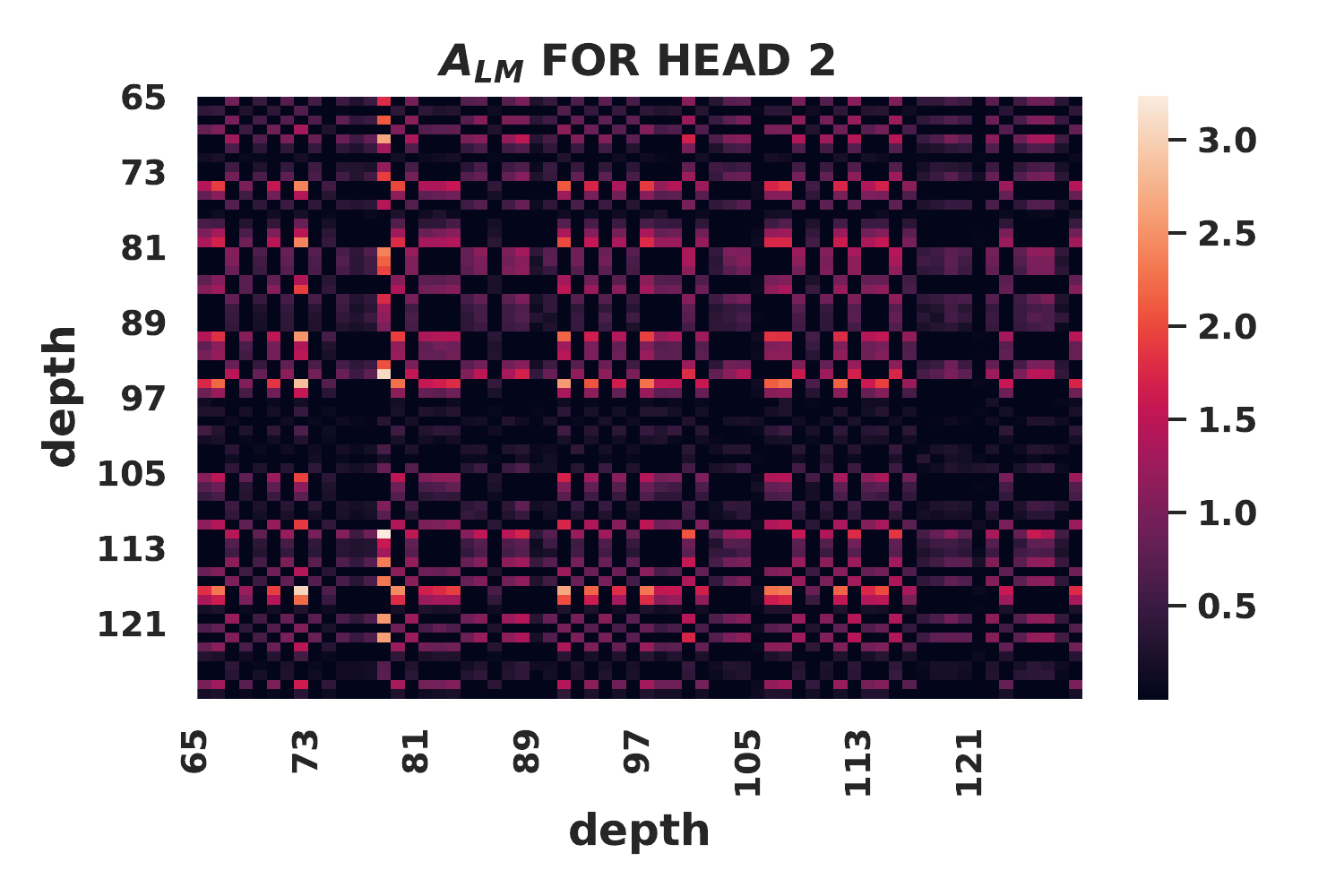}

\end{subfigure}
\hfill
\begin{subfigure}[b]{0.6\textwidth}
	\centering
	\includegraphics[width=1.1\textwidth]{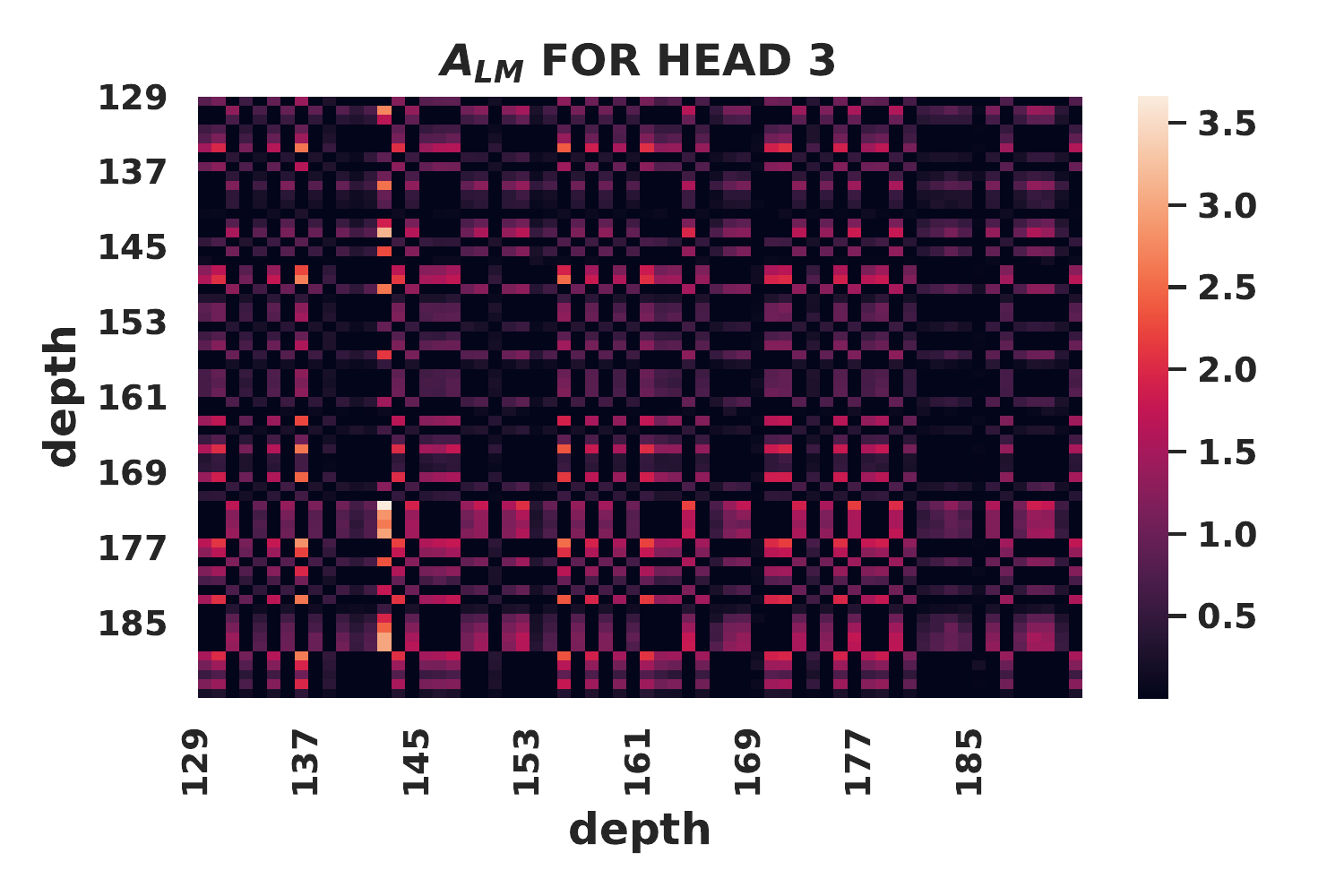}

\end{subfigure}
\hfill
\begin{subfigure}[b]{0.6\textwidth}
	\centering
	\includegraphics[width=1.1\textwidth]{Xheatmapmodel2almh4}

\end{subfigure}
\centering
\begin{subfigure}[b]{0.6\textwidth}
	\centering
	\includegraphics[width=1.1\textwidth]{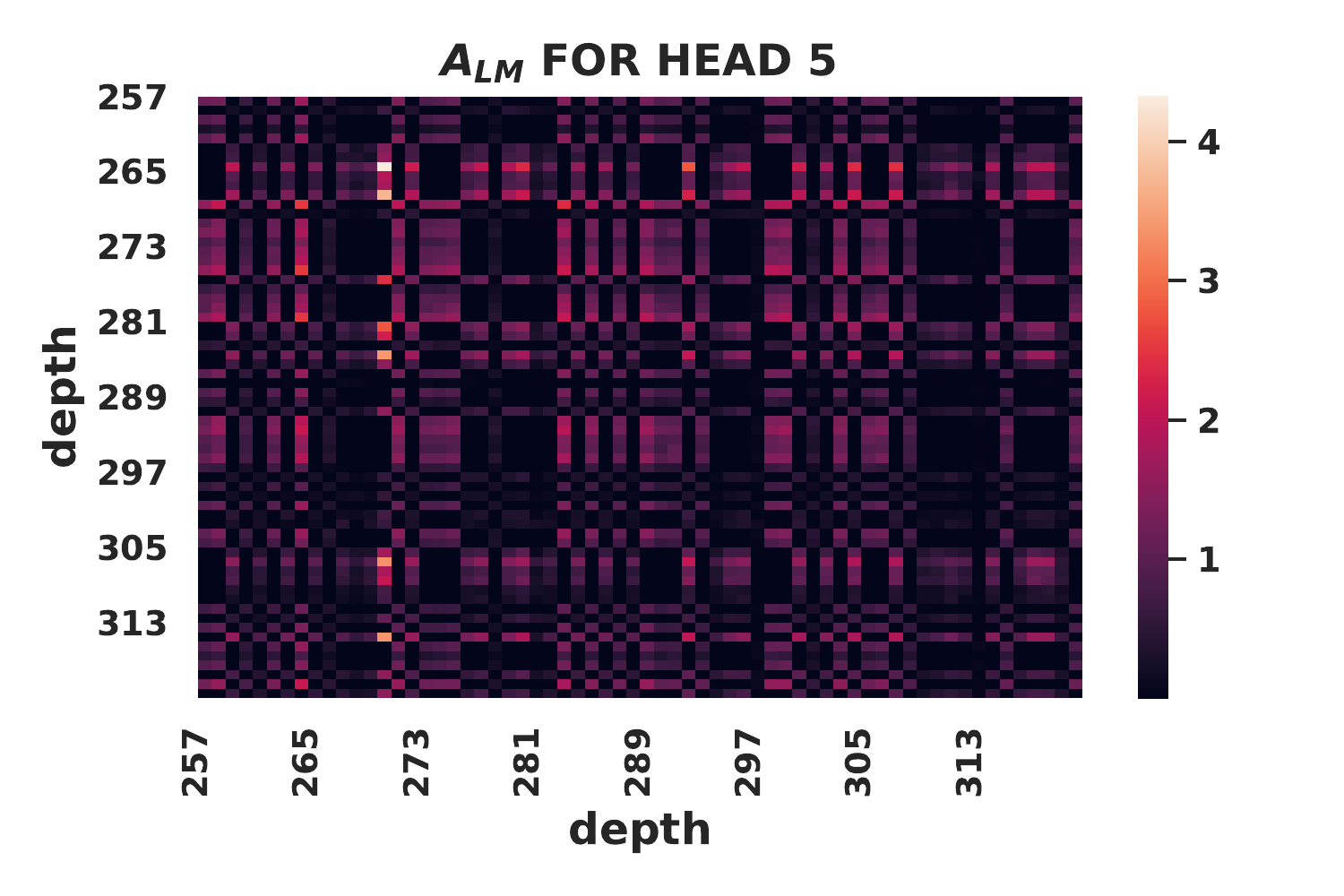}

\end{subfigure}
\hfill
\begin{subfigure}[b]{0.6\textwidth}
	\centering
	\includegraphics[width=1.1\textwidth]{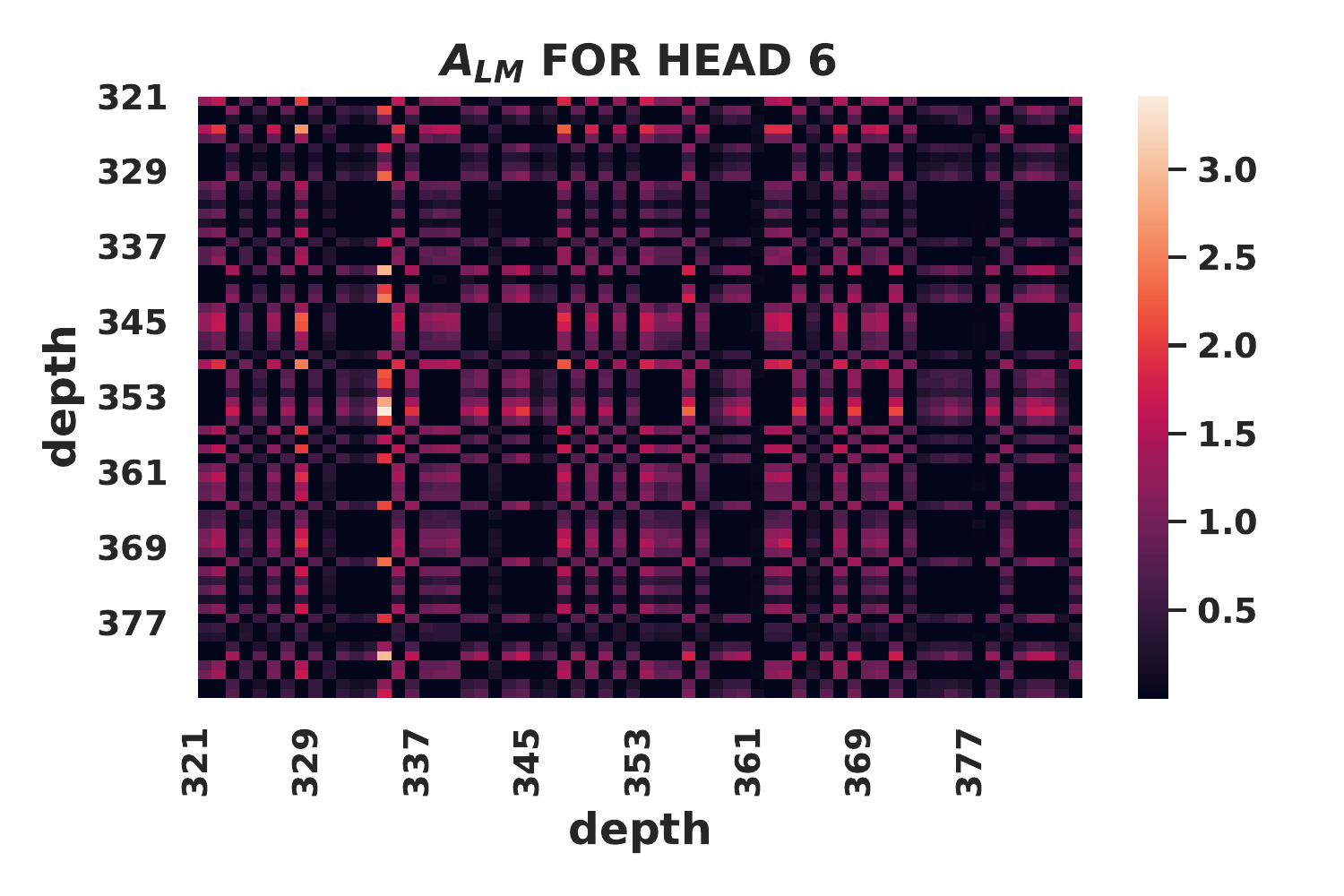}

\end{subfigure}
\hfill
\begin{subfigure}[b]{0.6\textwidth}
	\centering
	\includegraphics[width=1.1\textwidth]{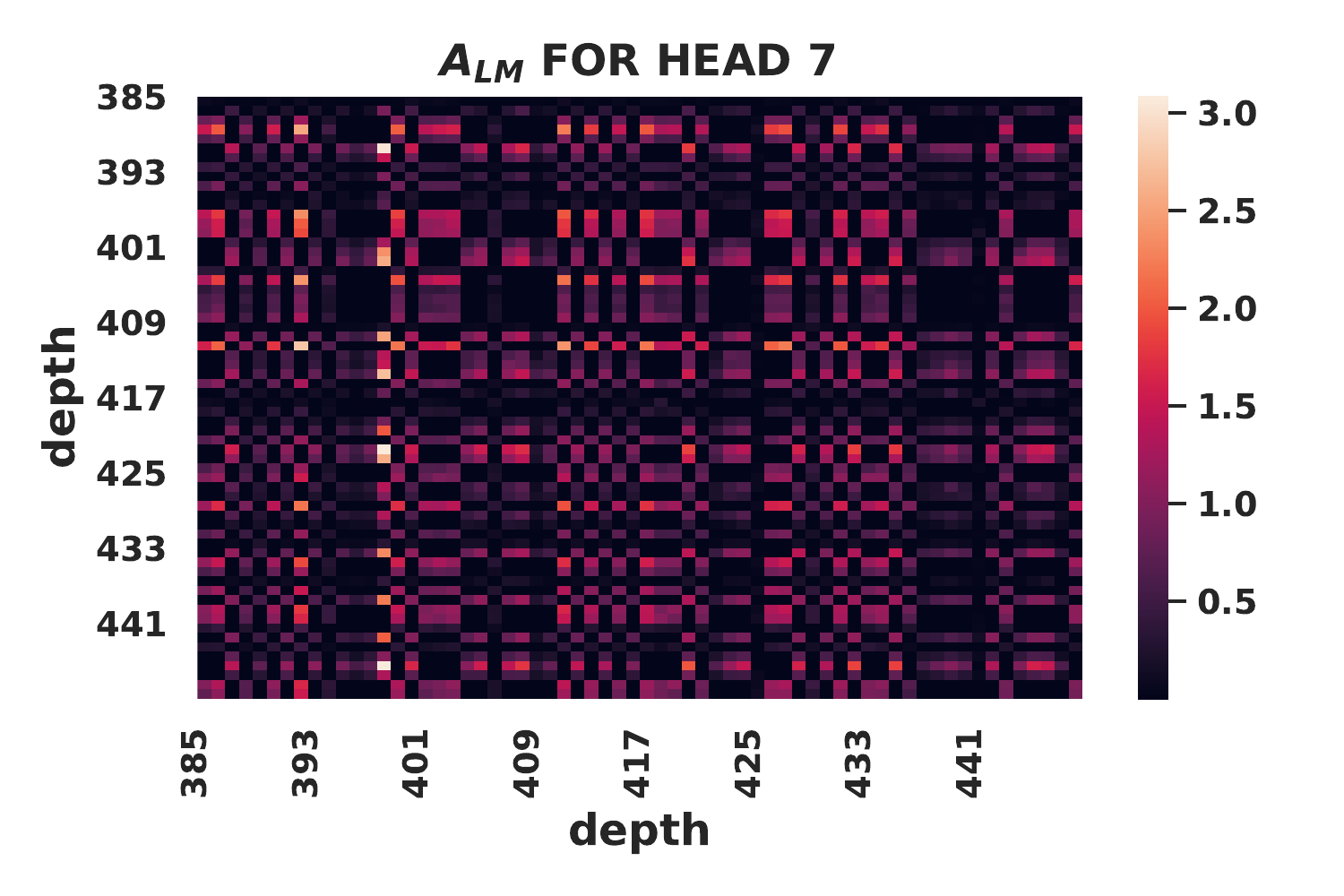}

\end{subfigure}
\hfill
\begin{subfigure}[b]{0.6\textwidth}
	\centering
	\includegraphics[width=1.1\textwidth]{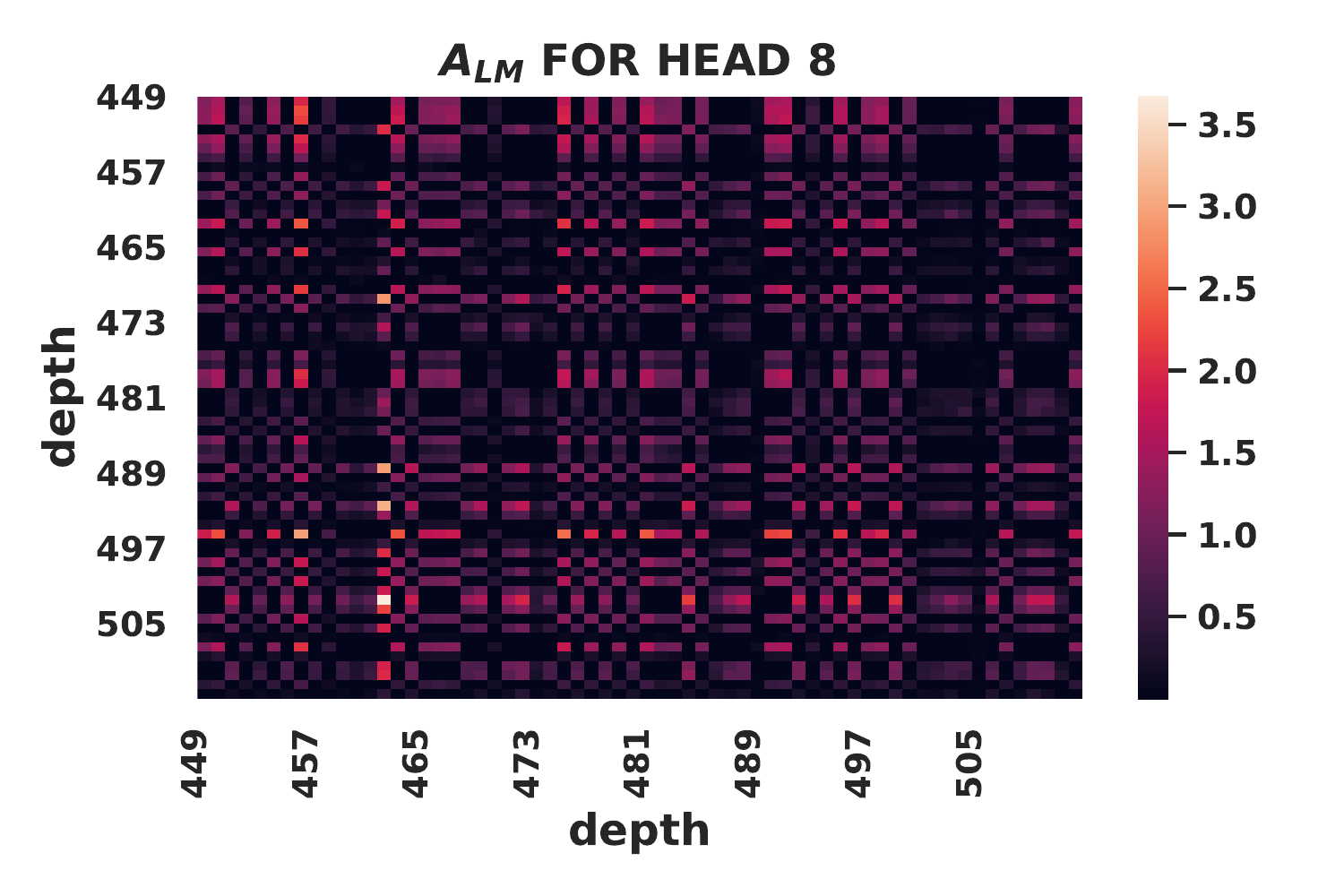}

\end{subfigure}
\caption{$\mA_{LM}$ heatmap plots for all heads from XLM attention stage from graph transformer model \#2.}
\label{fig8apx}
\end{adjustwidth}
\end{figure} 

\clearpage
\thispagestyle{headings}
\begin{figure}
\begin{adjustwidth}{-5em}{-5em}
\centering
\begin{subfigure}[b]{0.6\textwidth}
	\centering
	\includegraphics[width=1.1\textwidth]{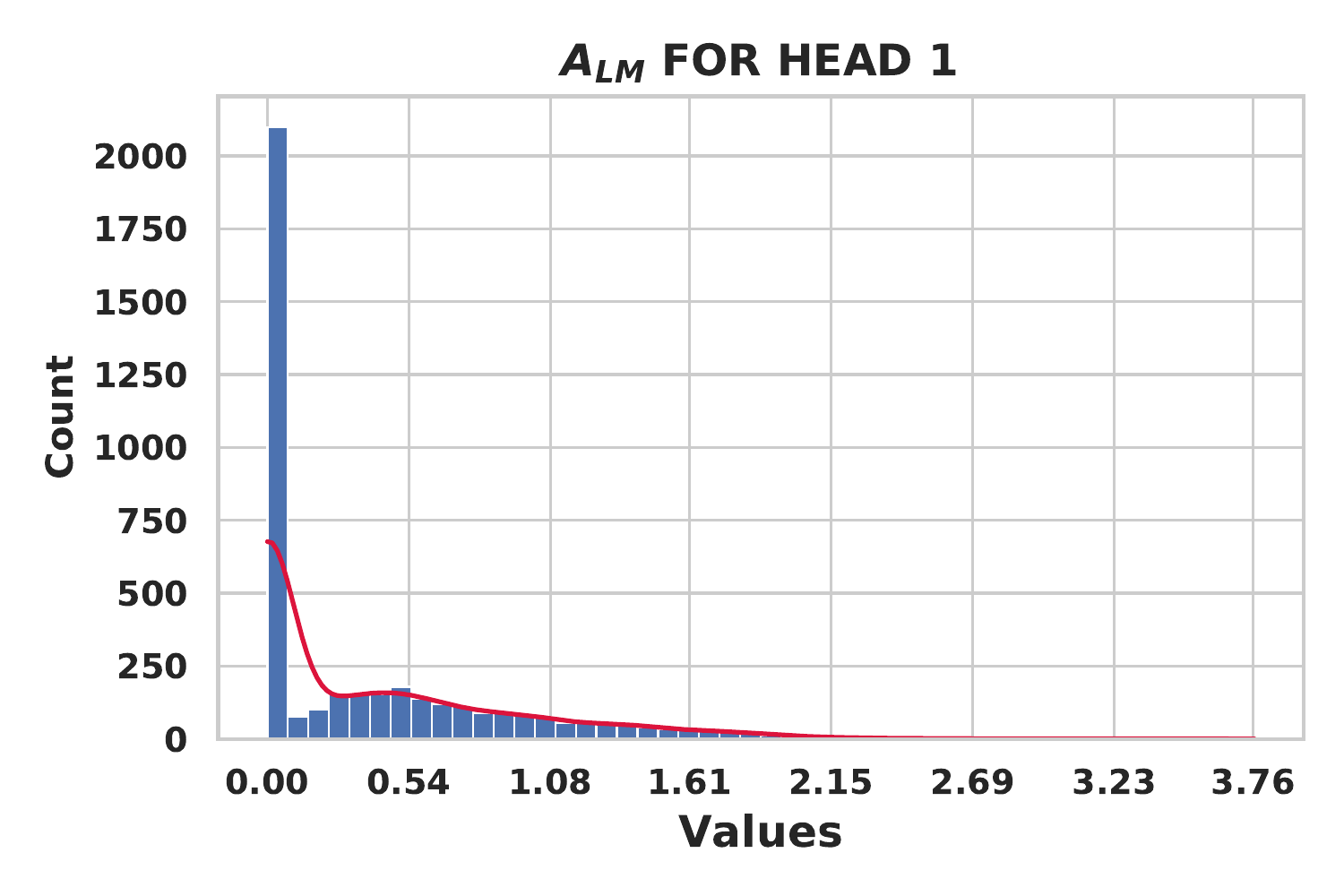}

\end{subfigure}
\hfill
\begin{subfigure}[b]{0.6\textwidth}
	\centering
	\includegraphics[width=1.1\textwidth]{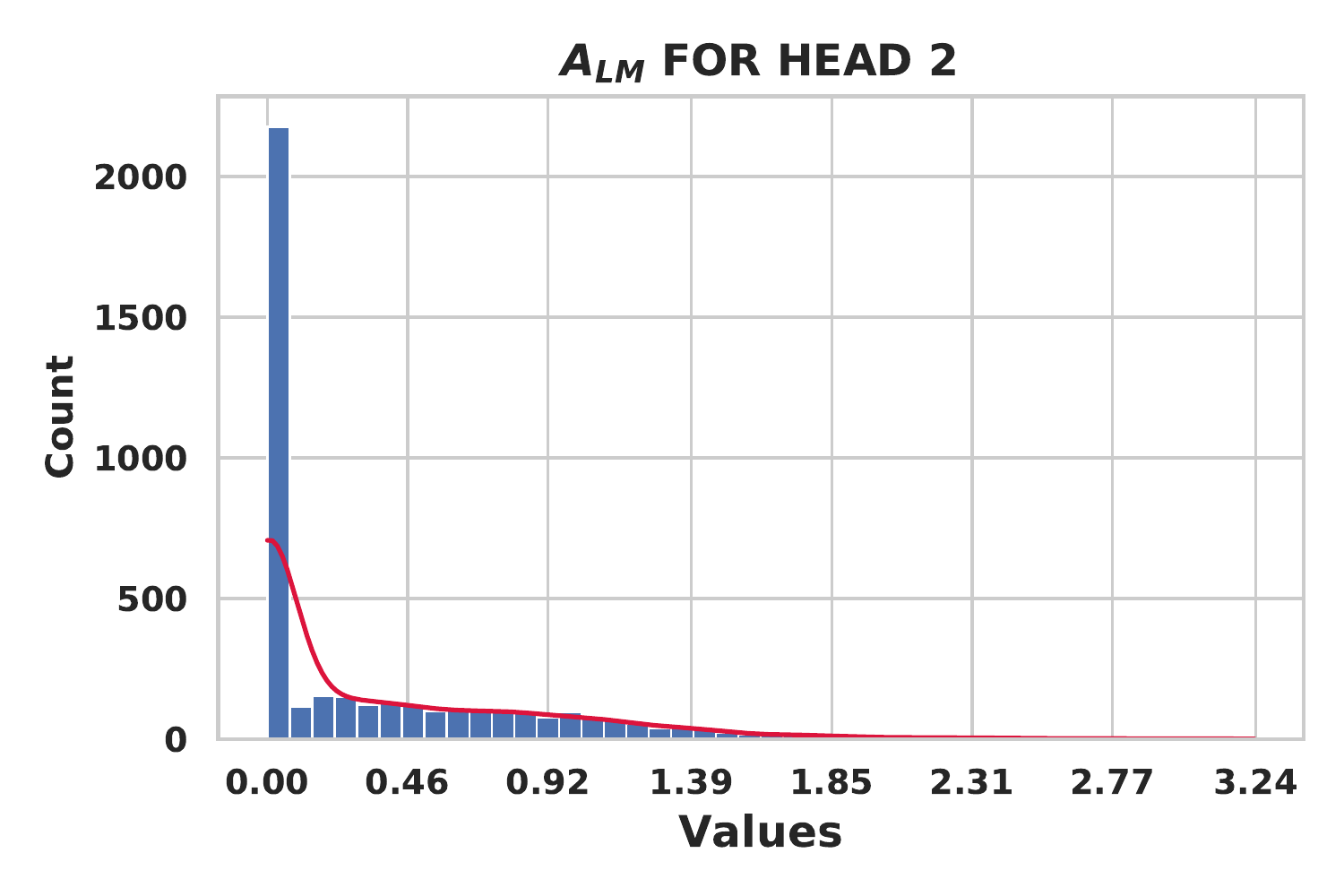}

\end{subfigure}
\hfill
\begin{subfigure}[b]{0.6\textwidth}
	\centering
	\includegraphics[width=1.1\textwidth]{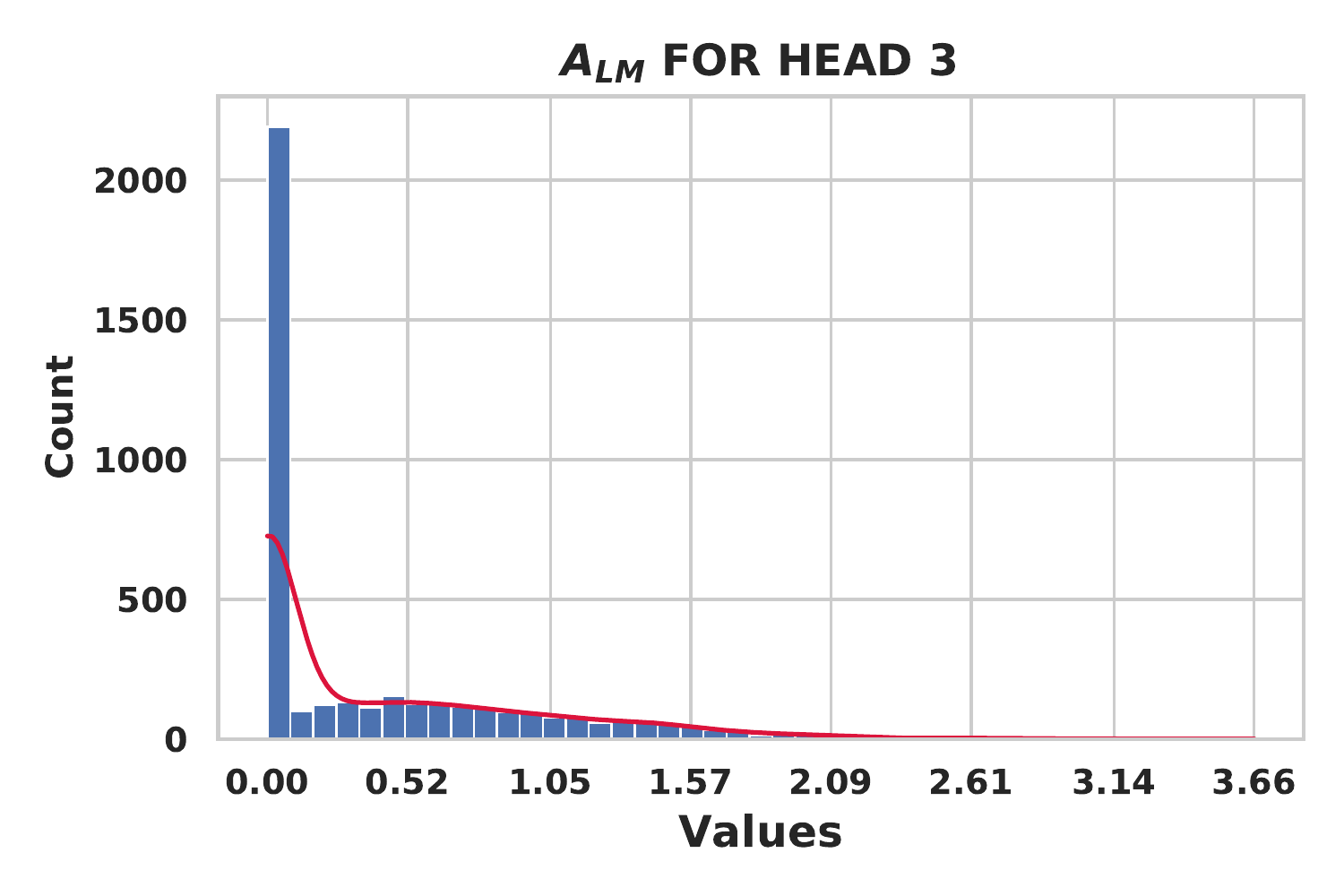}

\end{subfigure}
\hfill
\begin{subfigure}[b]{0.6\textwidth}
	\centering
	\includegraphics[width=1.1\textwidth]{Xhistmodel2almh4}

\end{subfigure}
\centering
\begin{subfigure}[b]{0.6\textwidth}
	\centering
	\includegraphics[width=1.1\textwidth]{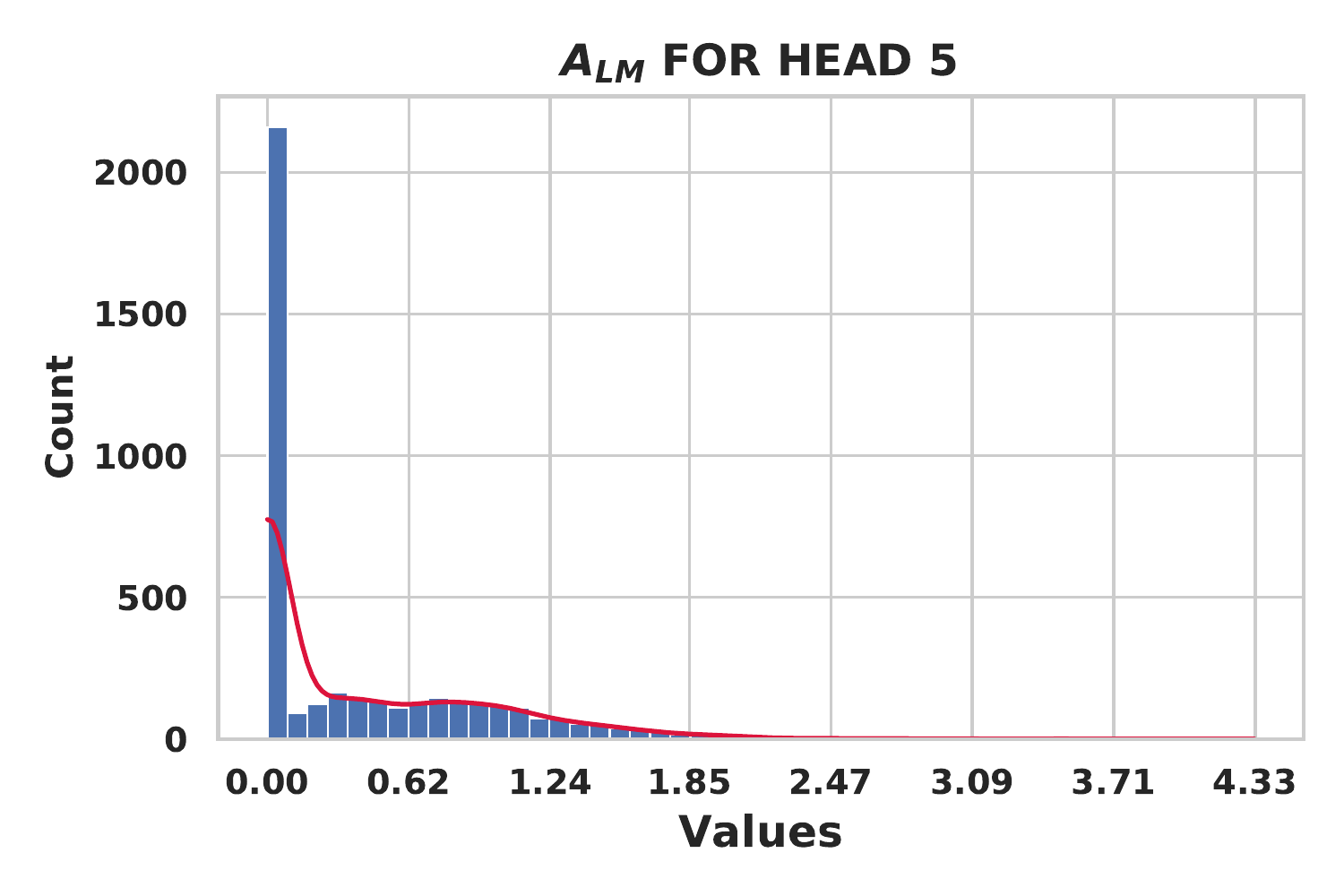}

\end{subfigure}
\hfill
\begin{subfigure}[b]{0.6\textwidth}
	\centering
	\includegraphics[width=1.1\textwidth]{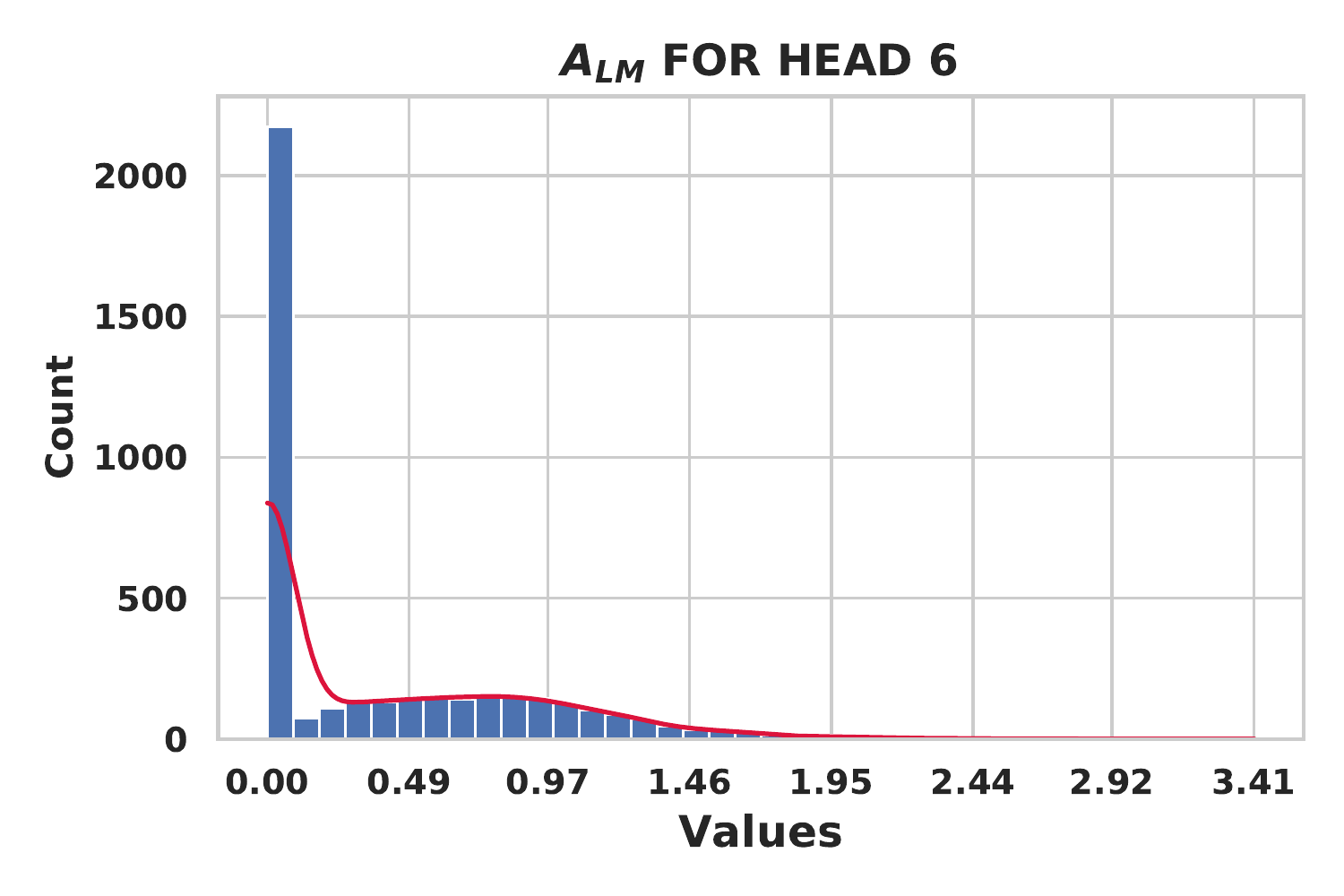}

\end{subfigure}
\hfill
\begin{subfigure}[b]{0.6\textwidth}
	\centering
	\includegraphics[width=1.1\textwidth]{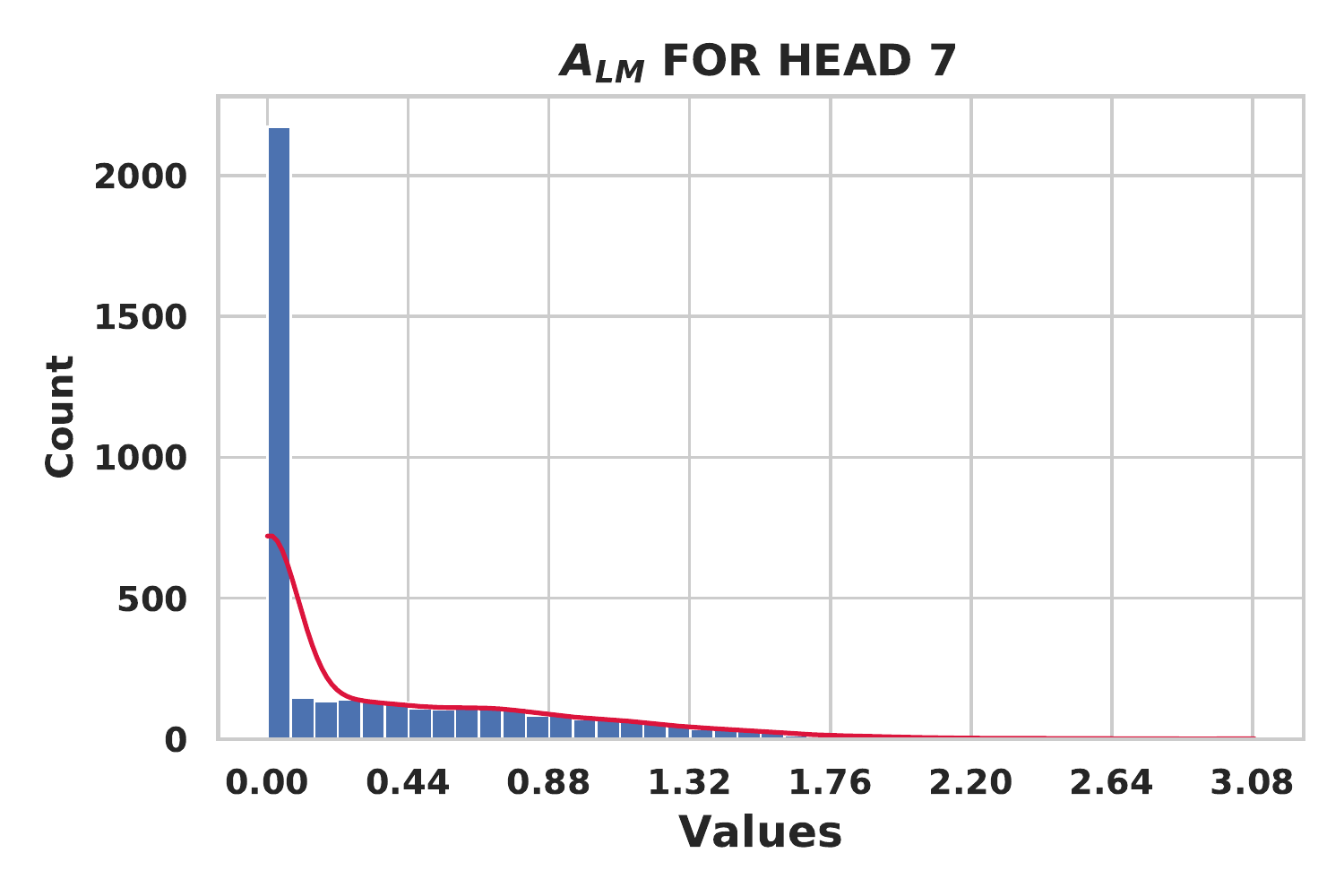}

\end{subfigure}
\hfill
\begin{subfigure}[b]{0.6\textwidth}
	\centering
	\includegraphics[width=1.1\textwidth]{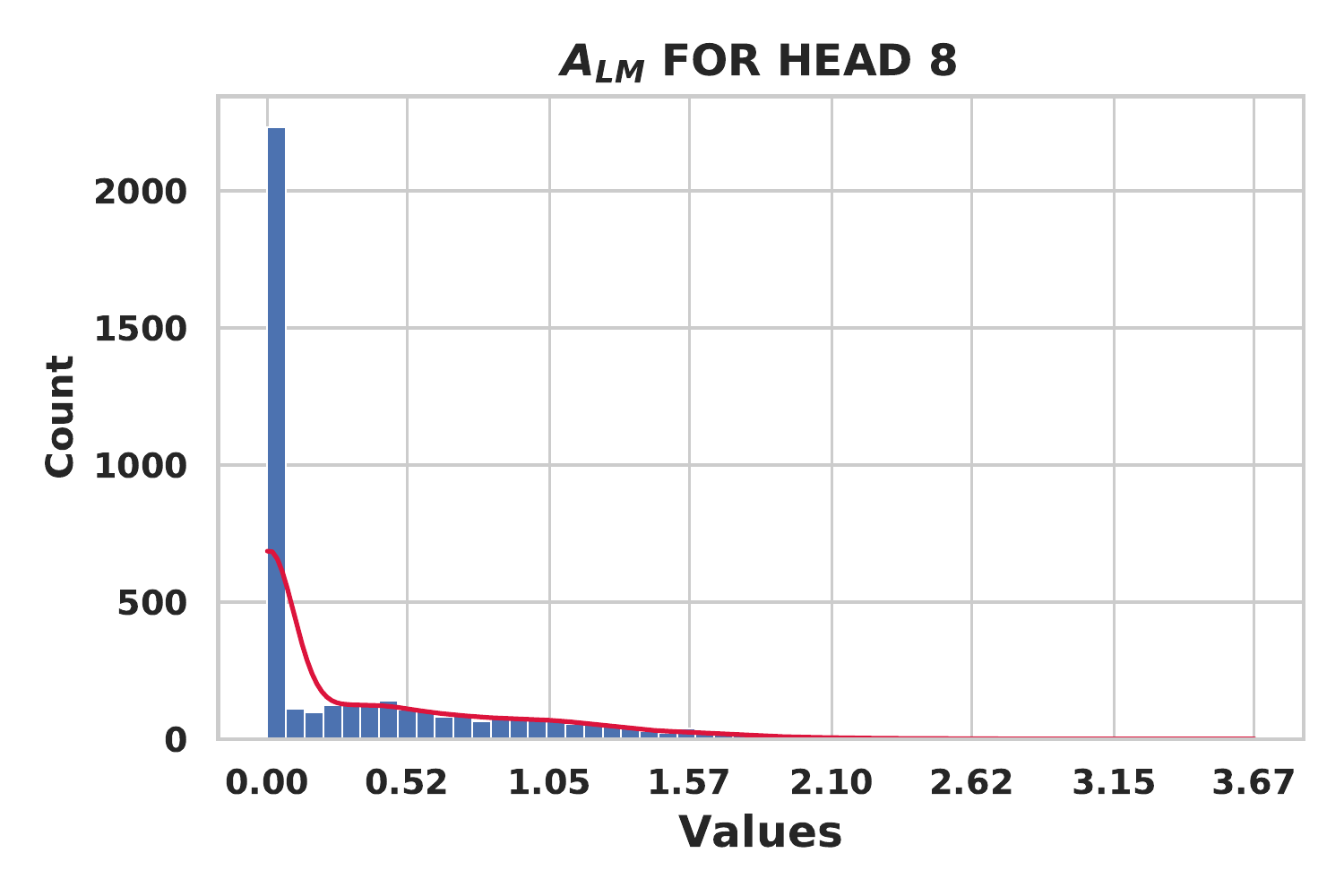}

\end{subfigure}
\caption{$\mA_{LM}$ histogram plots for all heads from XLM attention stage from graph transformer model \#2.}
\label{fig9apx}
\end{adjustwidth}
\end{figure}
\clearpage
\thispagestyle{headings}

\begin{figure}
\begin{adjustwidth}{-5em}{-5em}
\centering
\begin{subfigure}[b]{0.6\textwidth}
	\centering
	\includegraphics[width=1.1\textwidth]{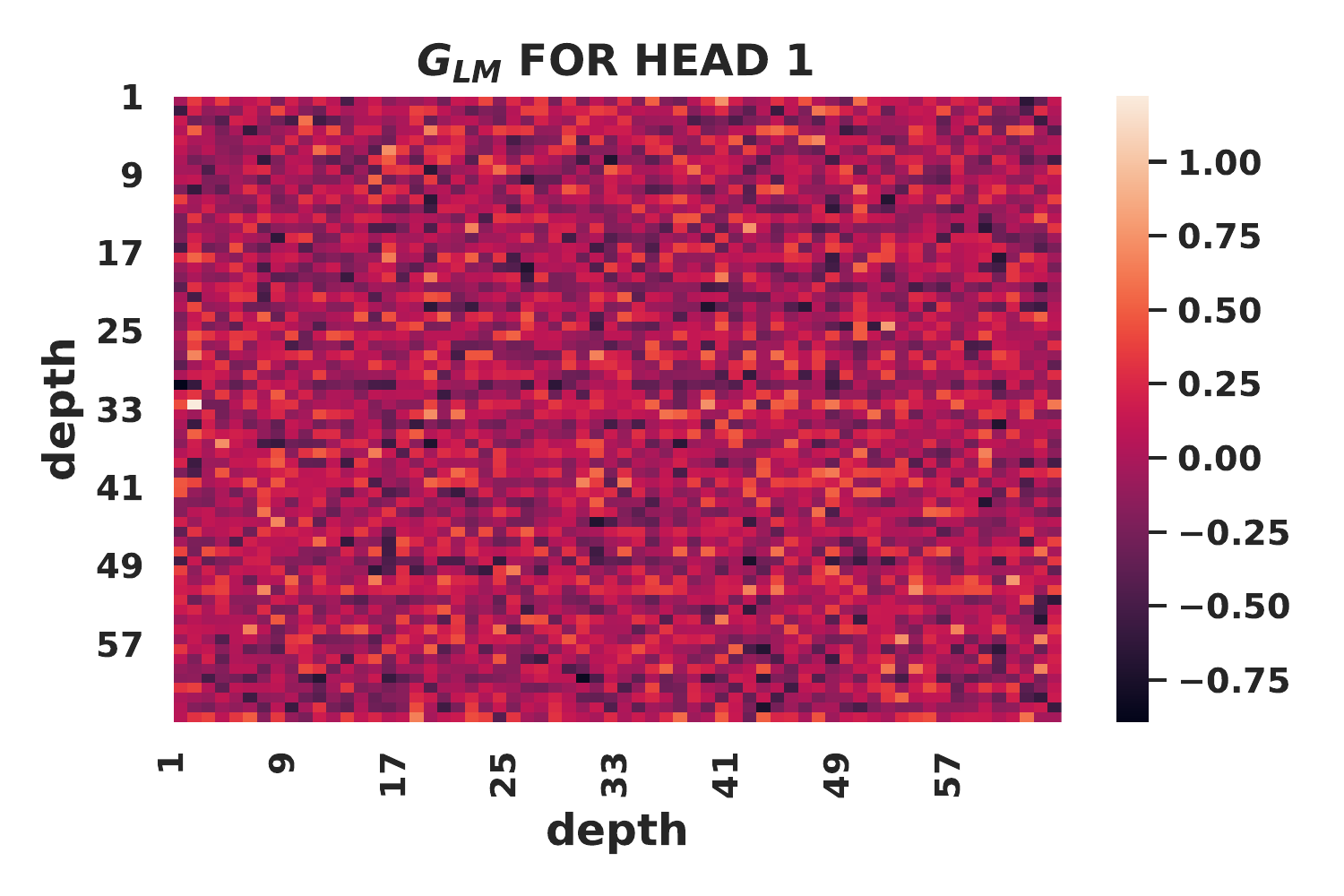}

\end{subfigure}
\hfill
\begin{subfigure}[b]{0.6\textwidth}
	\centering
	\includegraphics[width=1.1\textwidth]{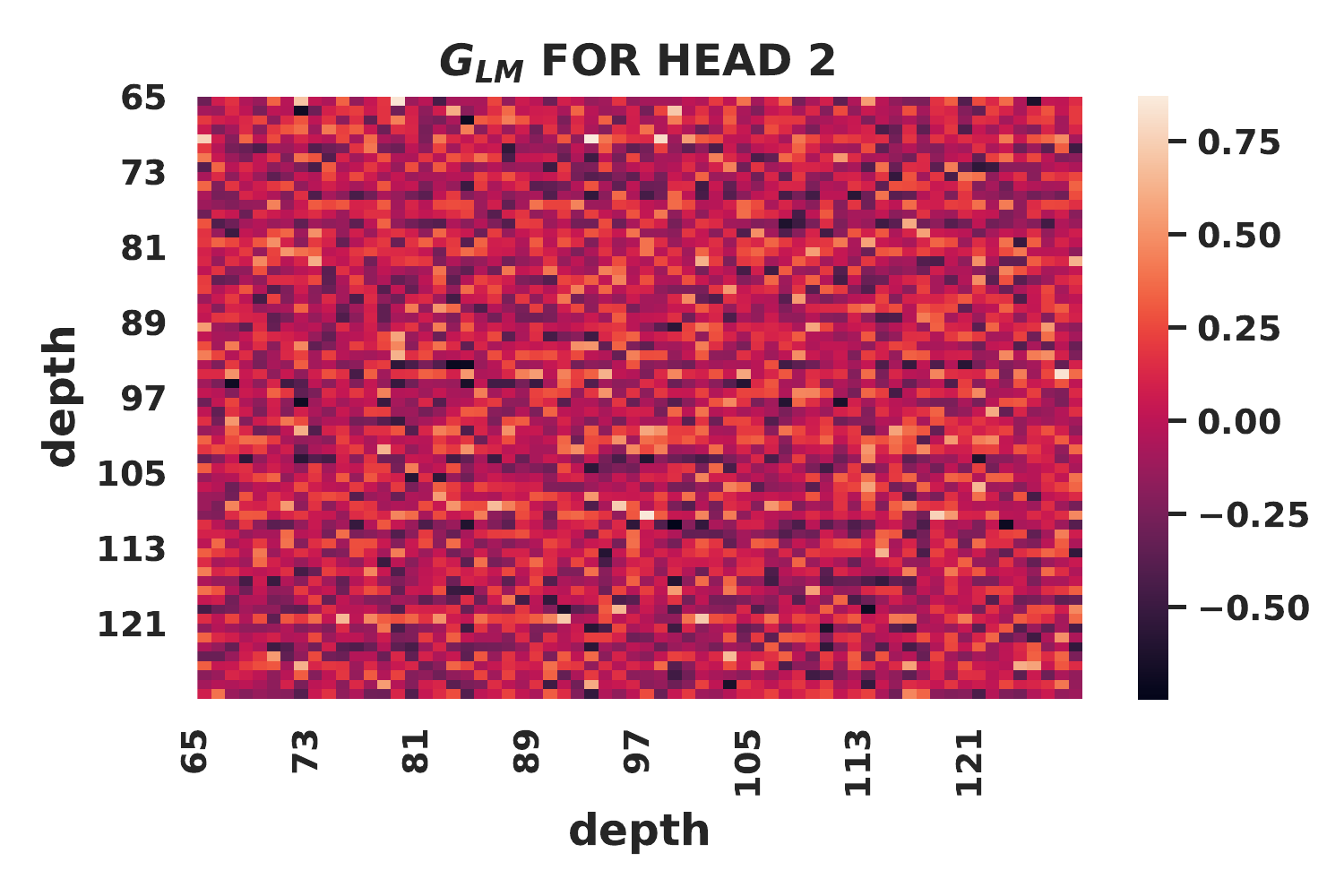}

\end{subfigure}
\hfill
\begin{subfigure}[b]{0.6\textwidth}
	\centering
	\includegraphics[width=1.1\textwidth]{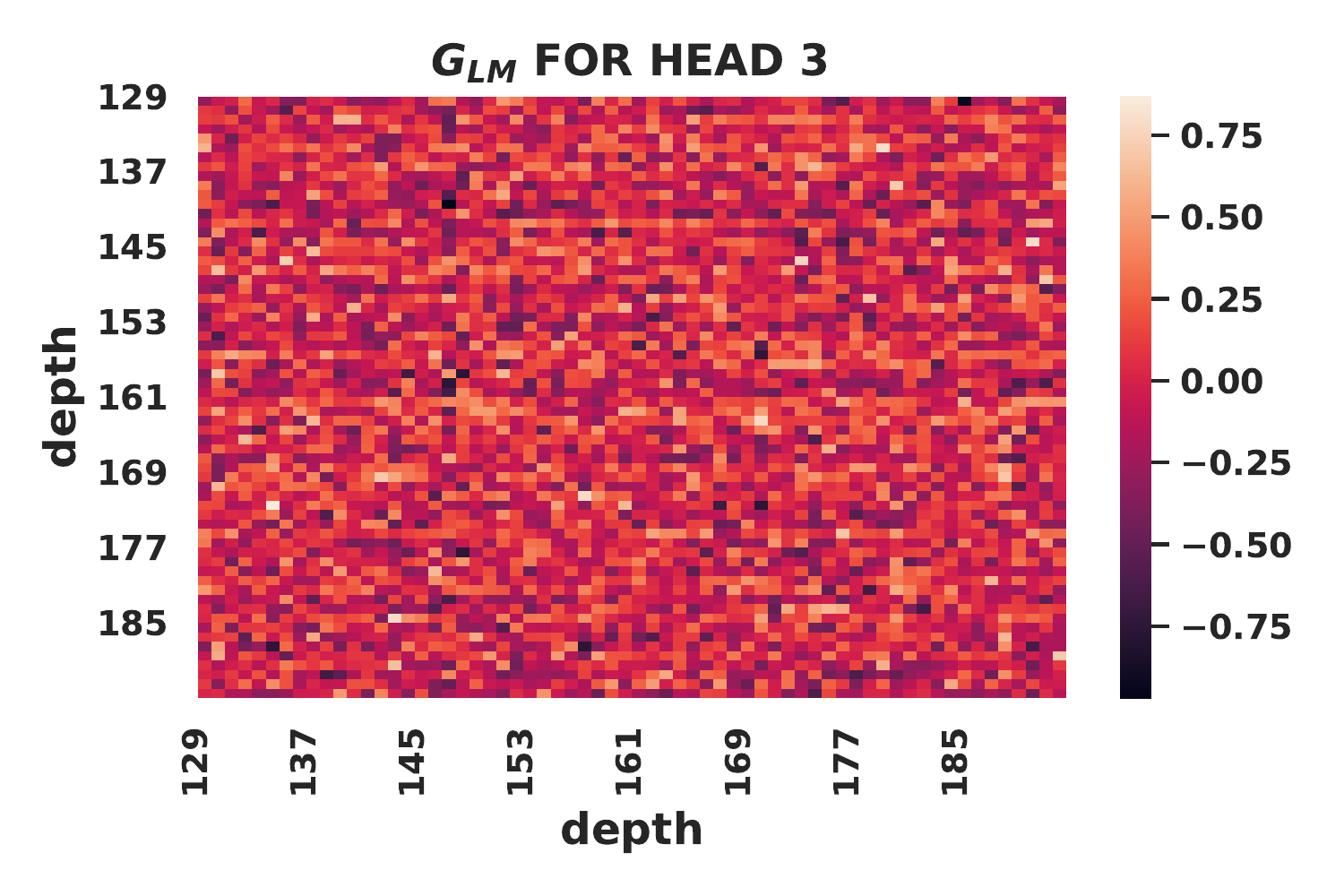}

\end{subfigure}
\hfill
\begin{subfigure}[b]{0.6\textwidth}
	\centering
	\includegraphics[width=1.1\textwidth]{Xheatmapmodel2glmh4}

\end{subfigure}
\centering
\begin{subfigure}[b]{0.6\textwidth}
	\centering
	\includegraphics[width=1.1\textwidth]{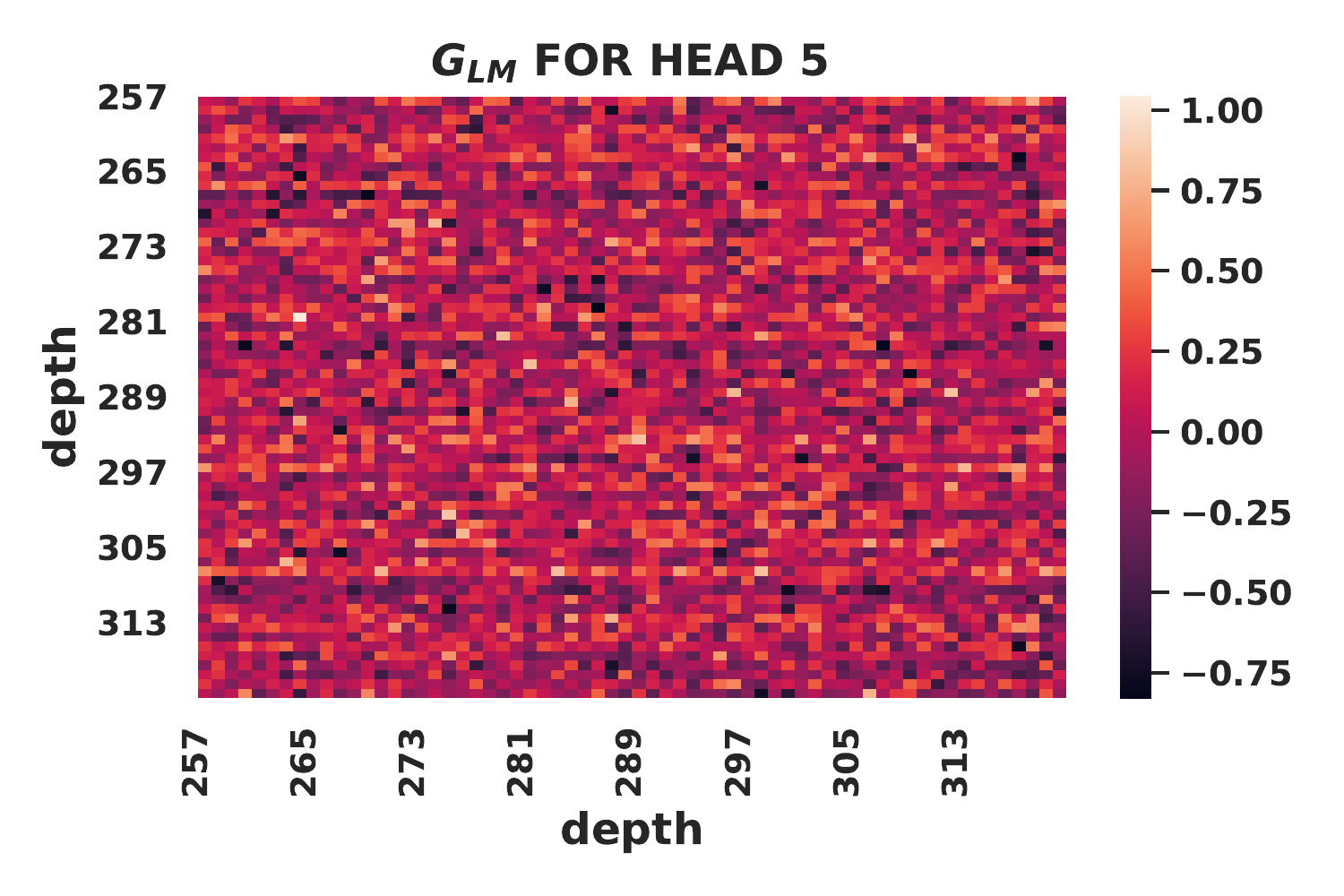}

\end{subfigure}
\hfill
\begin{subfigure}[b]{0.6\textwidth}
	\centering
	\includegraphics[width=1.1\textwidth]{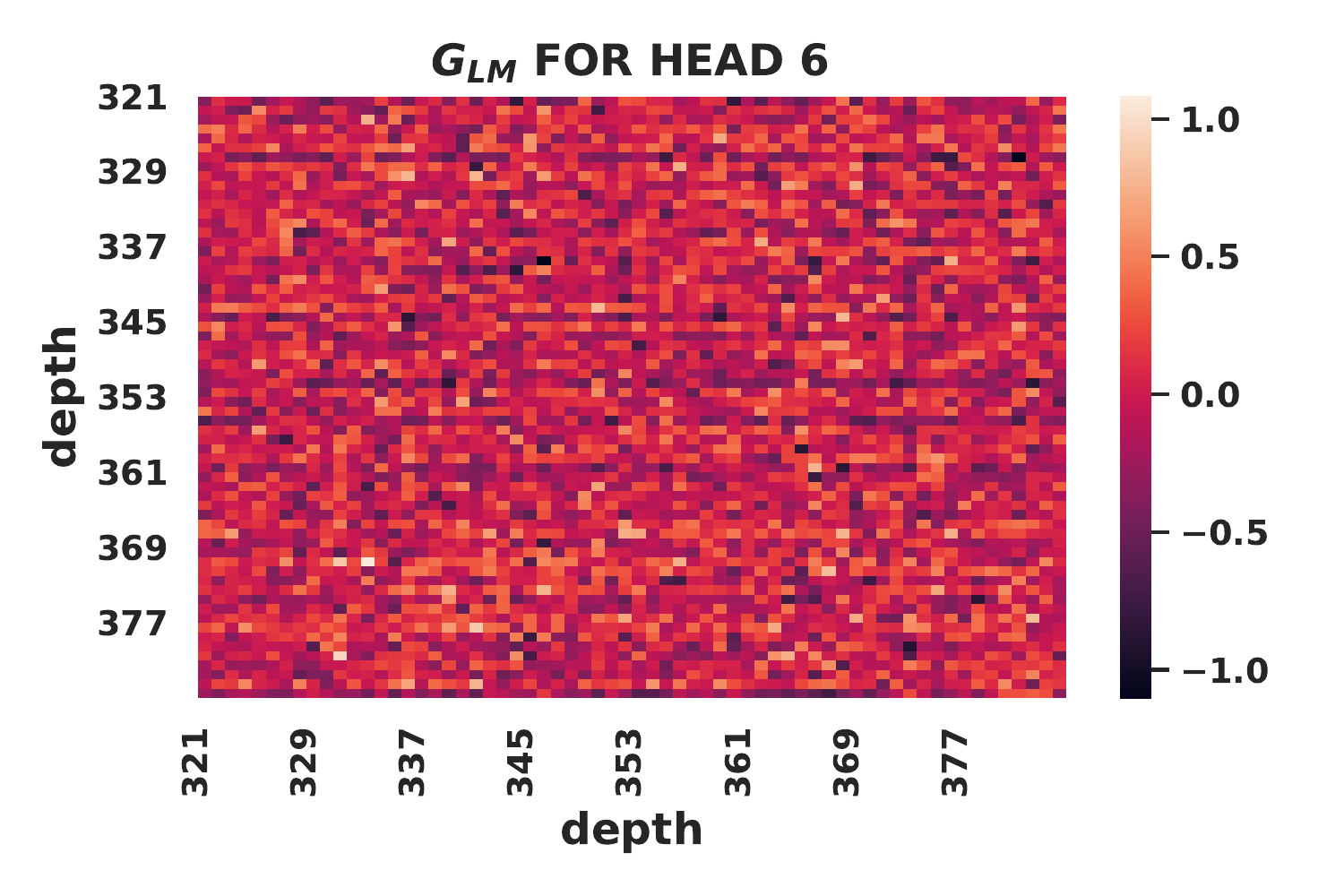}

\end{subfigure}
\hfill
\begin{subfigure}[b]{0.6\textwidth}
	\centering
	\includegraphics[width=1.1\textwidth]{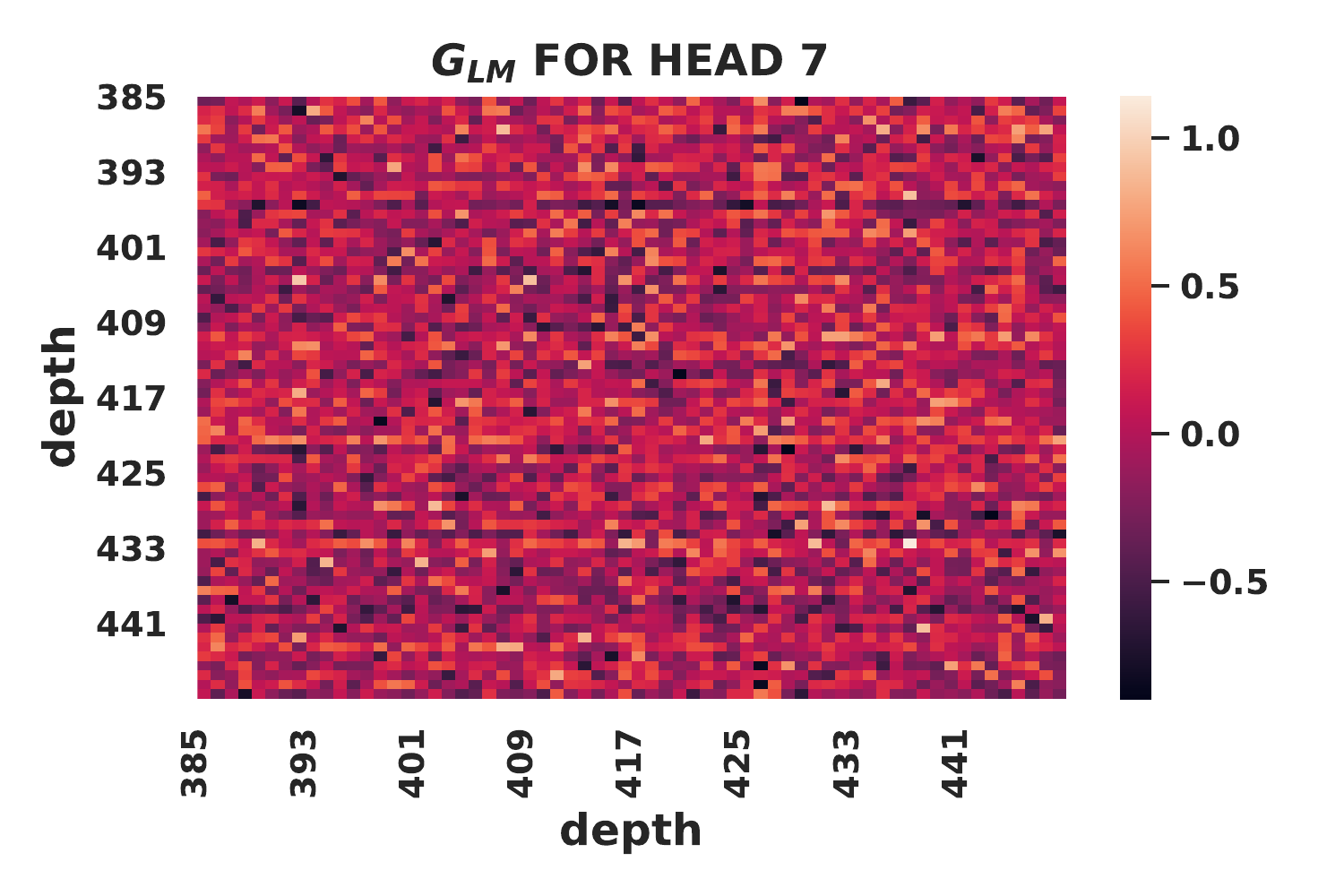}

\end{subfigure}
\hfill
\begin{subfigure}[b]{0.6\textwidth}
	\centering
	\includegraphics[width=1.1\textwidth]{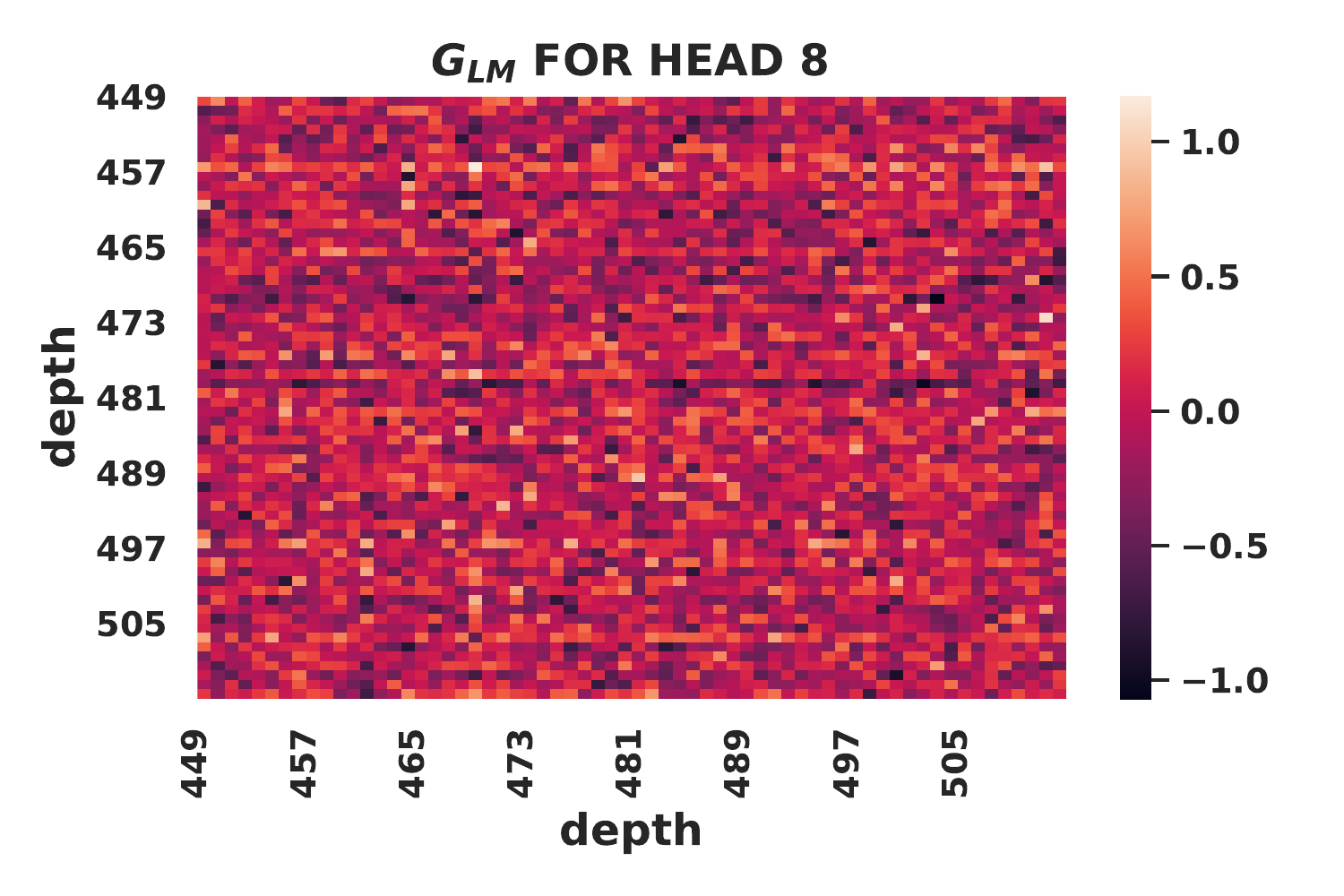}

\end{subfigure}
\caption{$\mG_{LM}$ heatmap plots for all heads from XLM attention stage from graph transformer model \#2.}
\label{fig10apx}
\end{adjustwidth}
\end{figure}

\clearpage
\thispagestyle{headings}

\begin{figure}
\begin{adjustwidth}{-5em}{-5em}
\centering
\begin{subfigure}[b]{0.6\textwidth}
	\centering
	\includegraphics[width=1.1\textwidth]{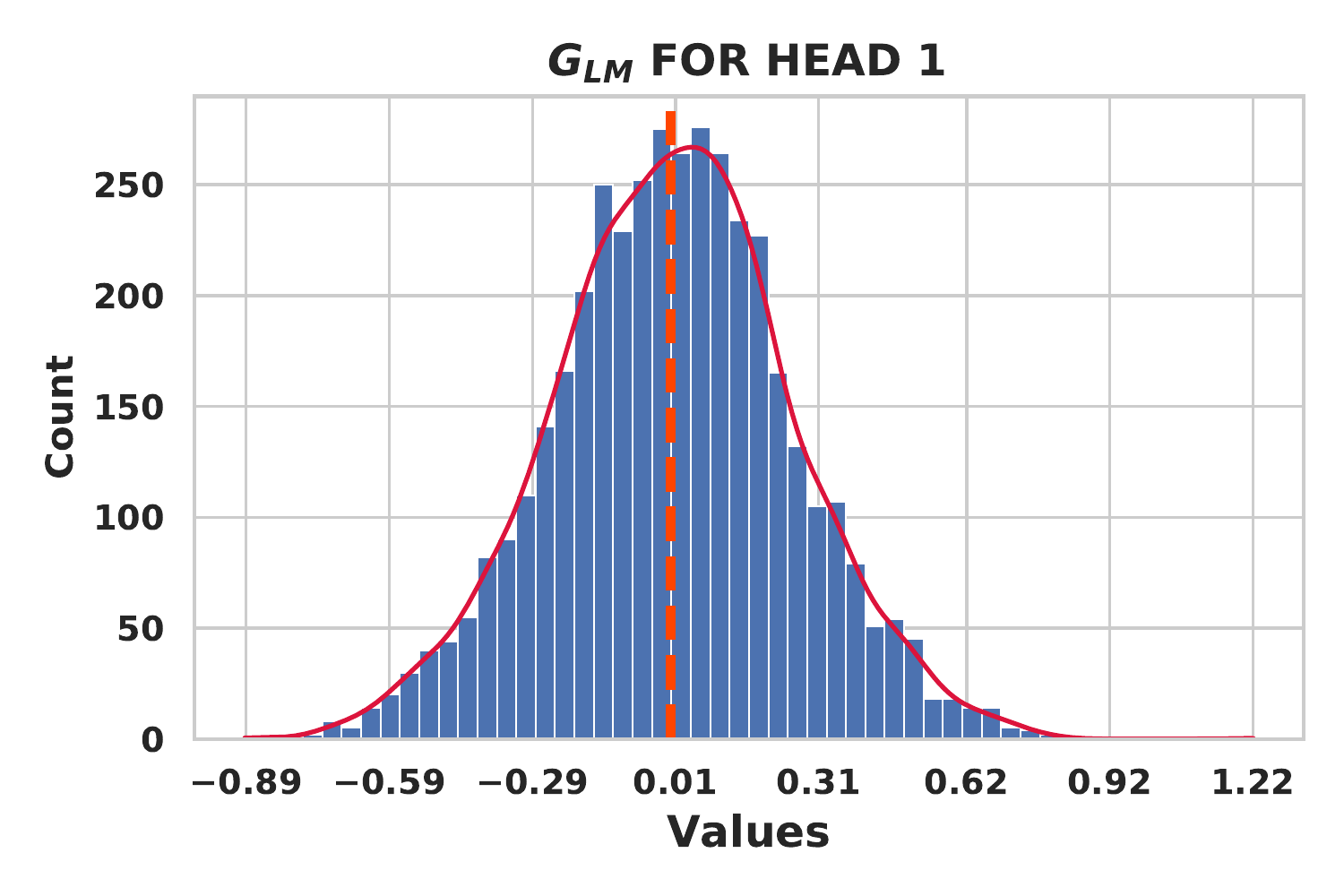}

\end{subfigure}
\hfill
\begin{subfigure}[b]{0.6\textwidth}
	\centering
	\includegraphics[width=1.1\textwidth]{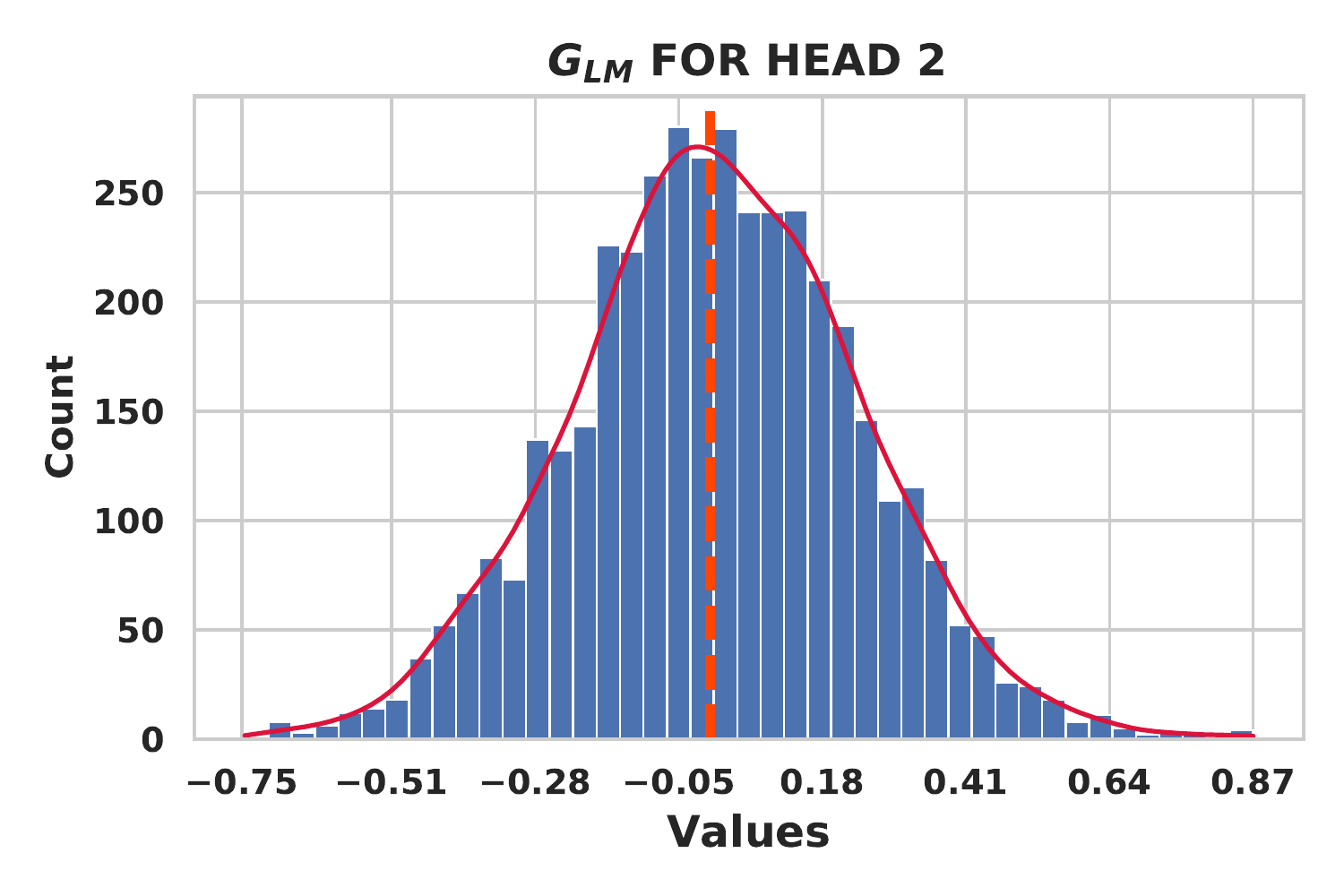}

\end{subfigure}
\hfill
\begin{subfigure}[b]{0.6\textwidth}
	\centering
	\includegraphics[width=1.1\textwidth]{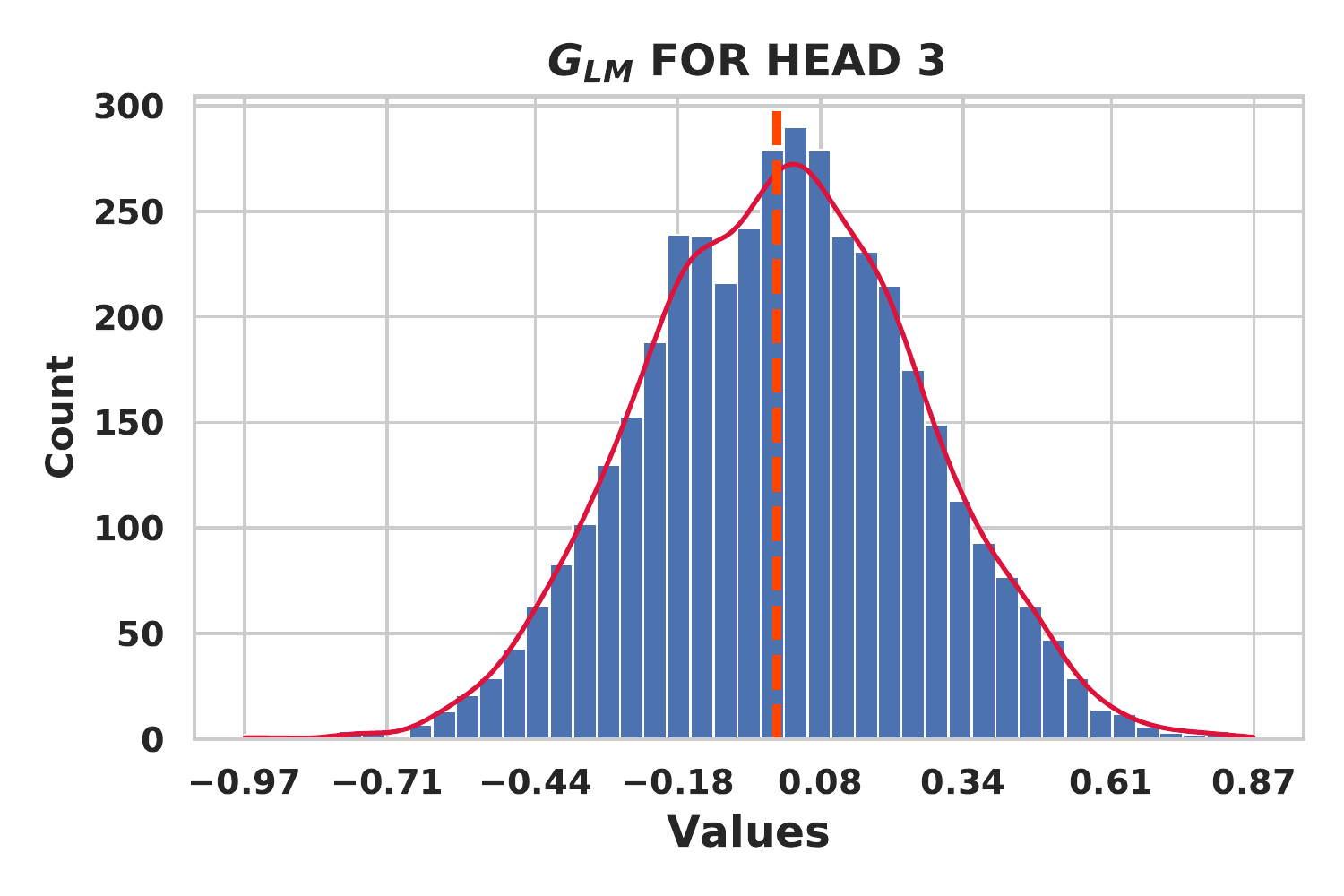}

\end{subfigure}
\hfill
\begin{subfigure}[b]{0.6\textwidth}
	\centering
	\includegraphics[width=1.1\textwidth]{Xhistmodel2glmh4}

\end{subfigure}
\centering
\begin{subfigure}[b]{0.6\textwidth}
	\centering
	\includegraphics[width=1.1\textwidth]{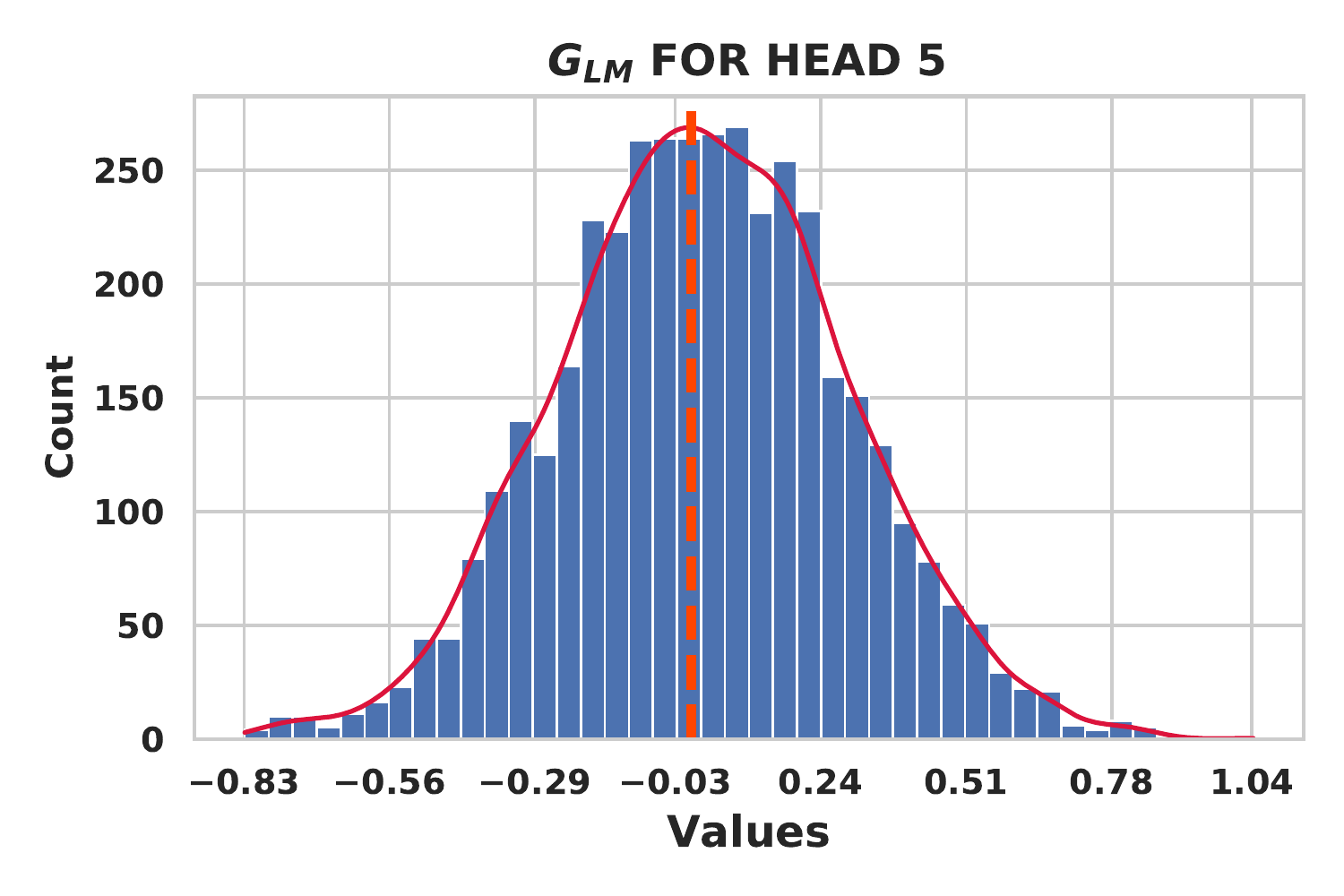}

\end{subfigure}
\hfill
\begin{subfigure}[b]{0.6\textwidth}
	\centering
	\includegraphics[width=1.1\textwidth]{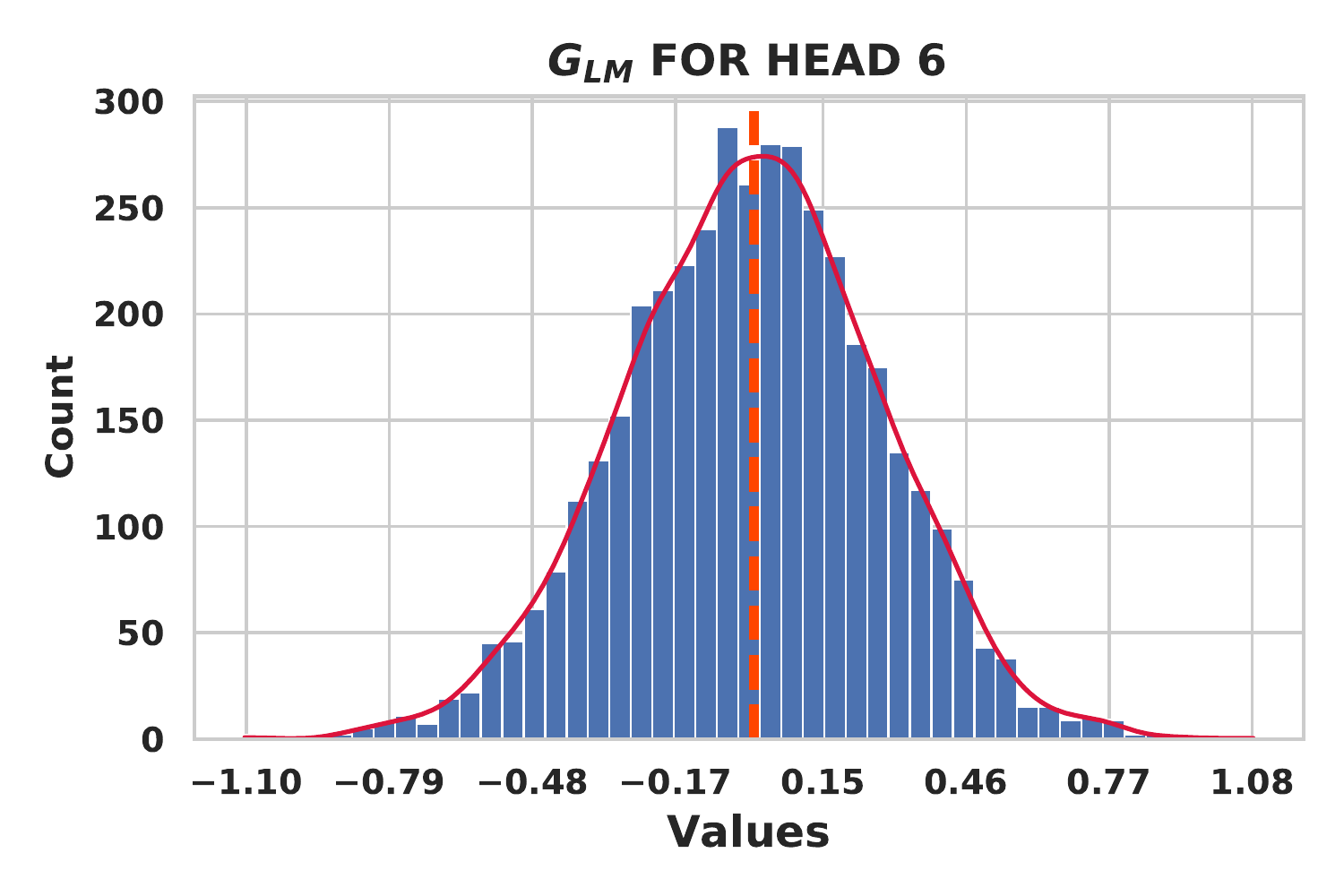}

\end{subfigure}
\hfill
\begin{subfigure}[b]{0.6\textwidth}
	\centering
	\includegraphics[width=1.1\textwidth]{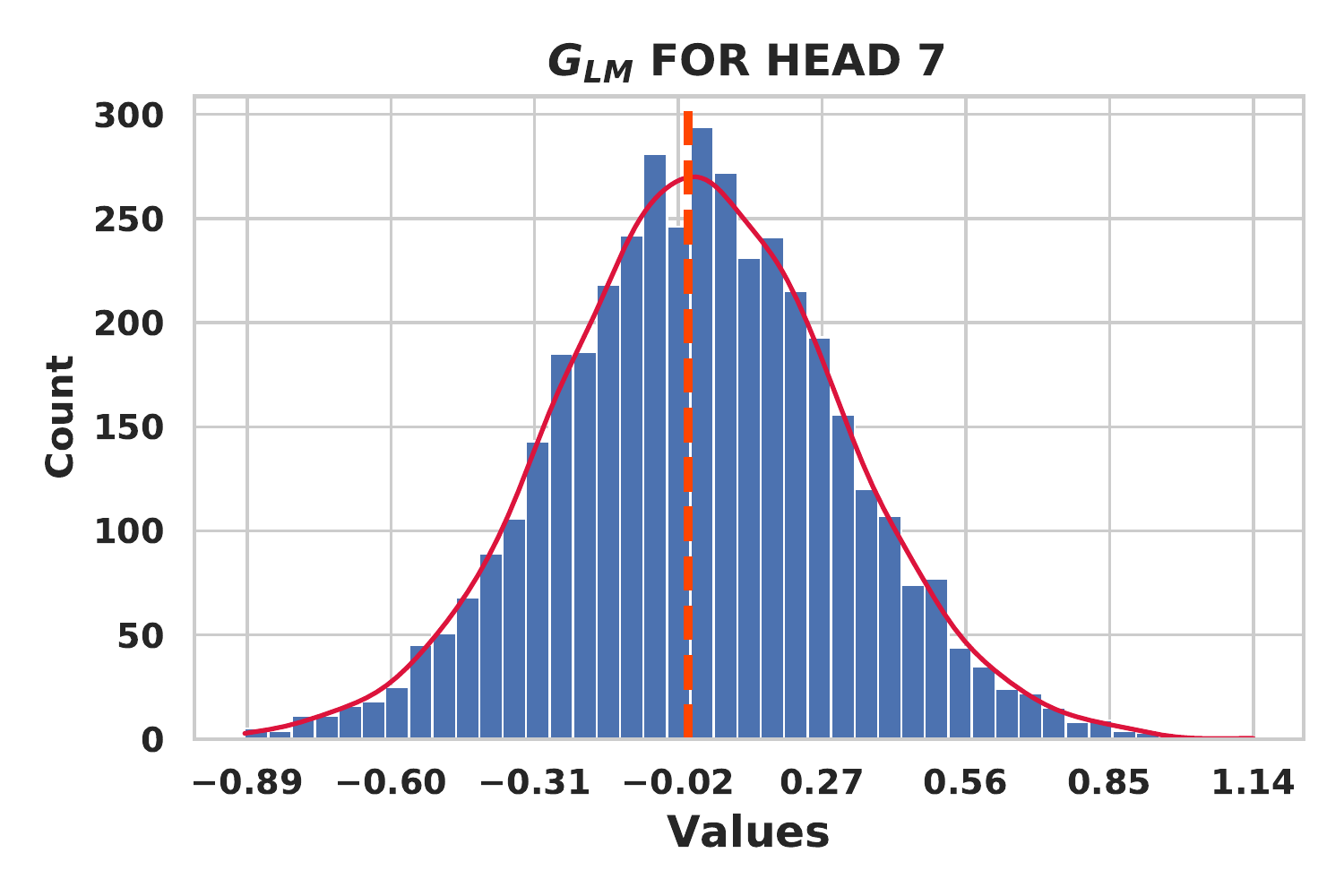}

\end{subfigure}
\hfill
\begin{subfigure}[b]{0.6\textwidth}
	\centering
	\includegraphics[width=1.1\textwidth]{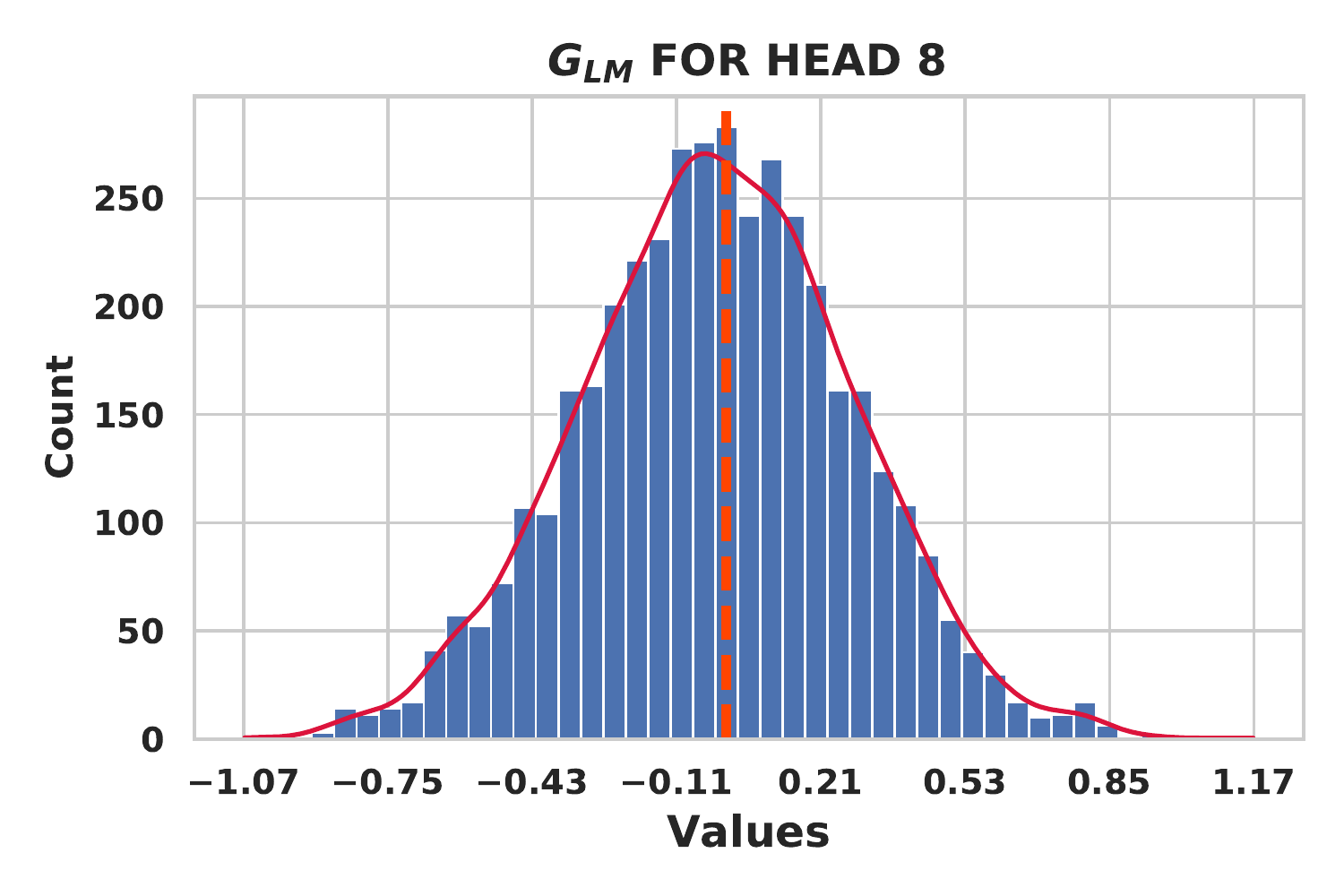}

\end{subfigure}
\caption{$\mG_{LM}$ histogram plots for all heads from XLM attention stage from graph transformer model \#2. Dashed line in orange marks zero value.}
\label{fig11apx}
\end{adjustwidth}
\end{figure}


\thispagestyle{headings}
\begin{figure}
\begin{adjustwidth}{-5em}{-5em}
\centering
\begin{subfigure}[b]{0.6\textwidth}
	\centering
	\includegraphics[width=1.1\textwidth]{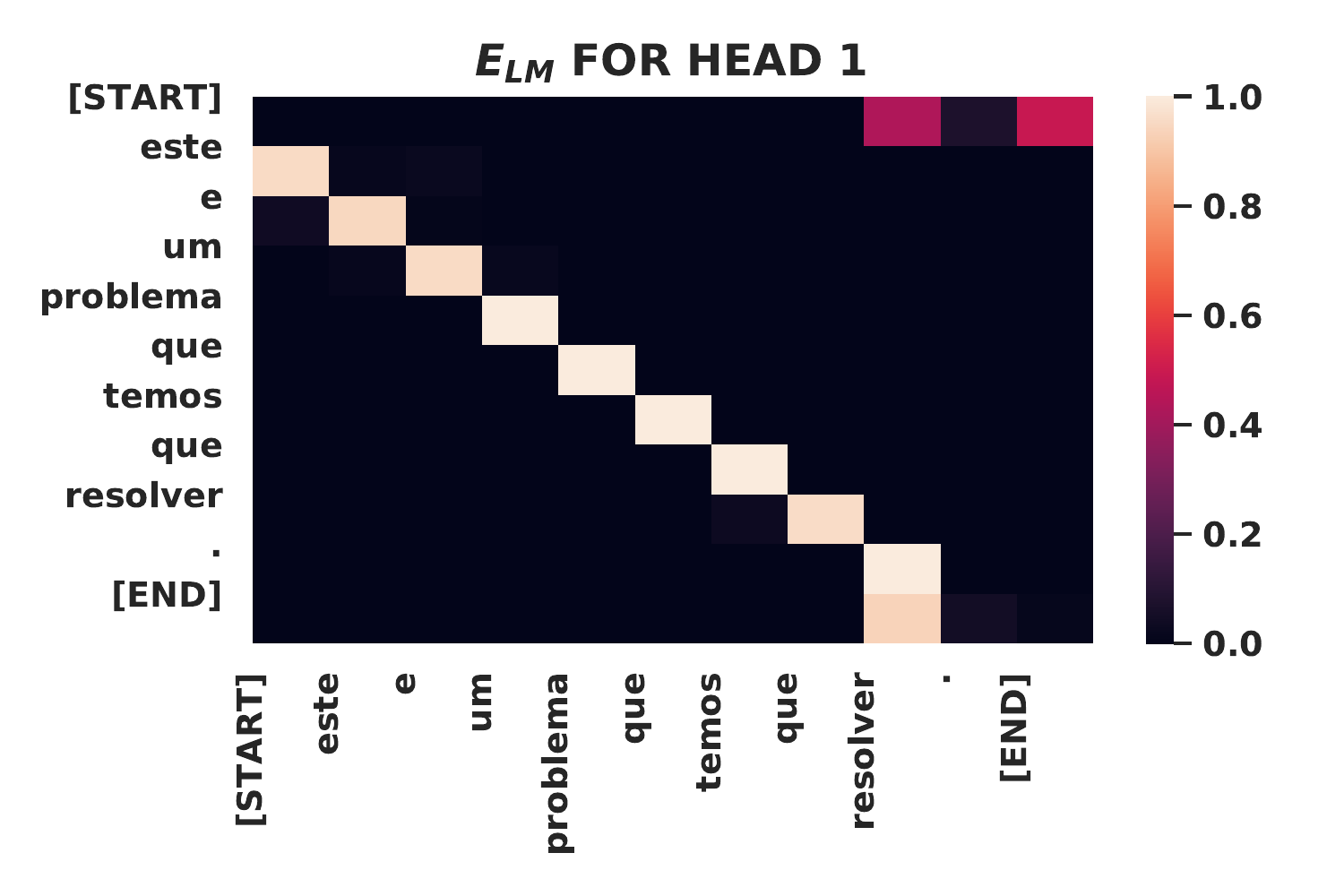}

\end{subfigure}
\hfill
\begin{subfigure}[b]{0.6\textwidth}
	\centering
	\includegraphics[width=1.1\textwidth]{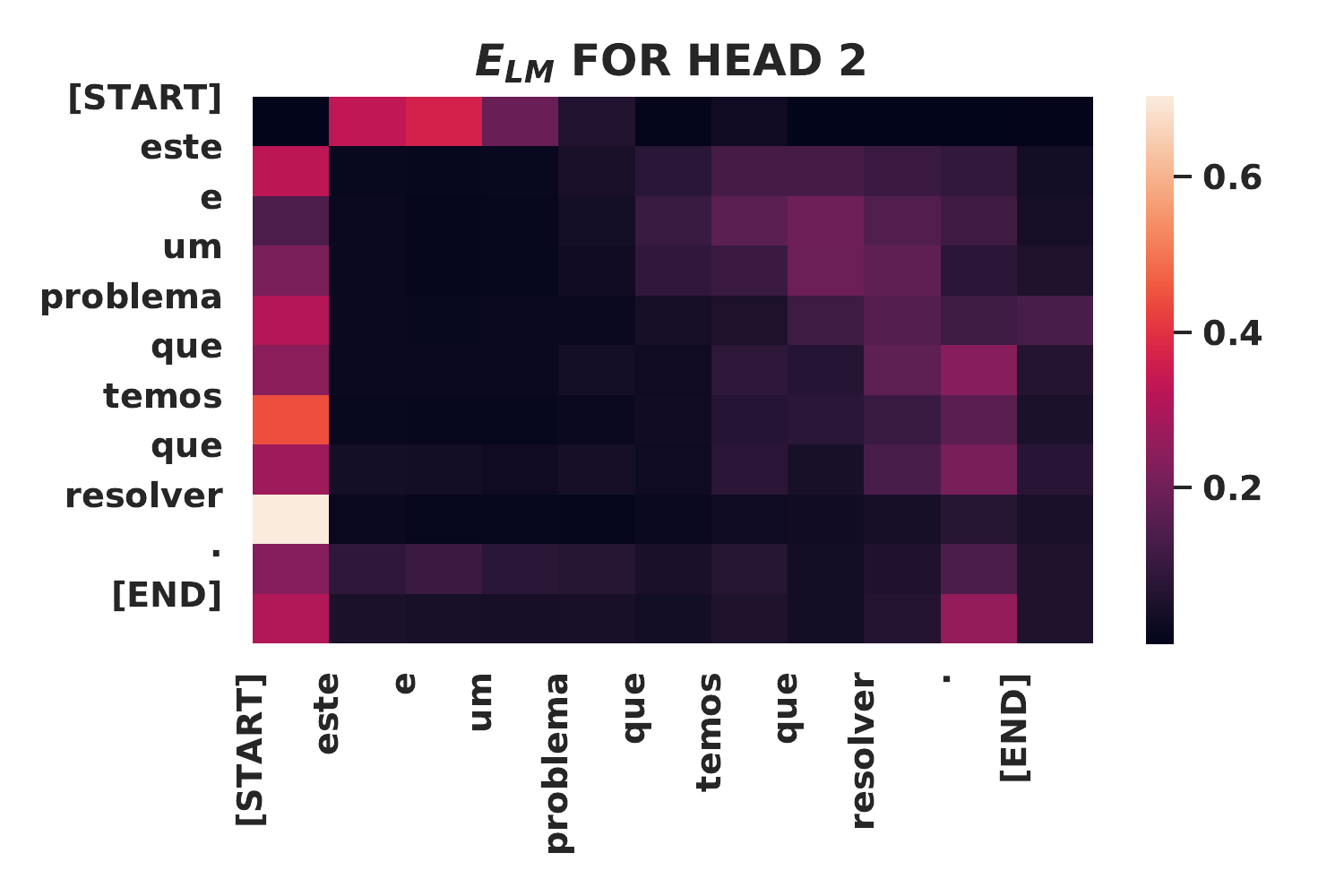}

\end{subfigure}
\hfill
\begin{subfigure}[b]{0.6\textwidth}
	\centering
	\includegraphics[width=1.1\textwidth]{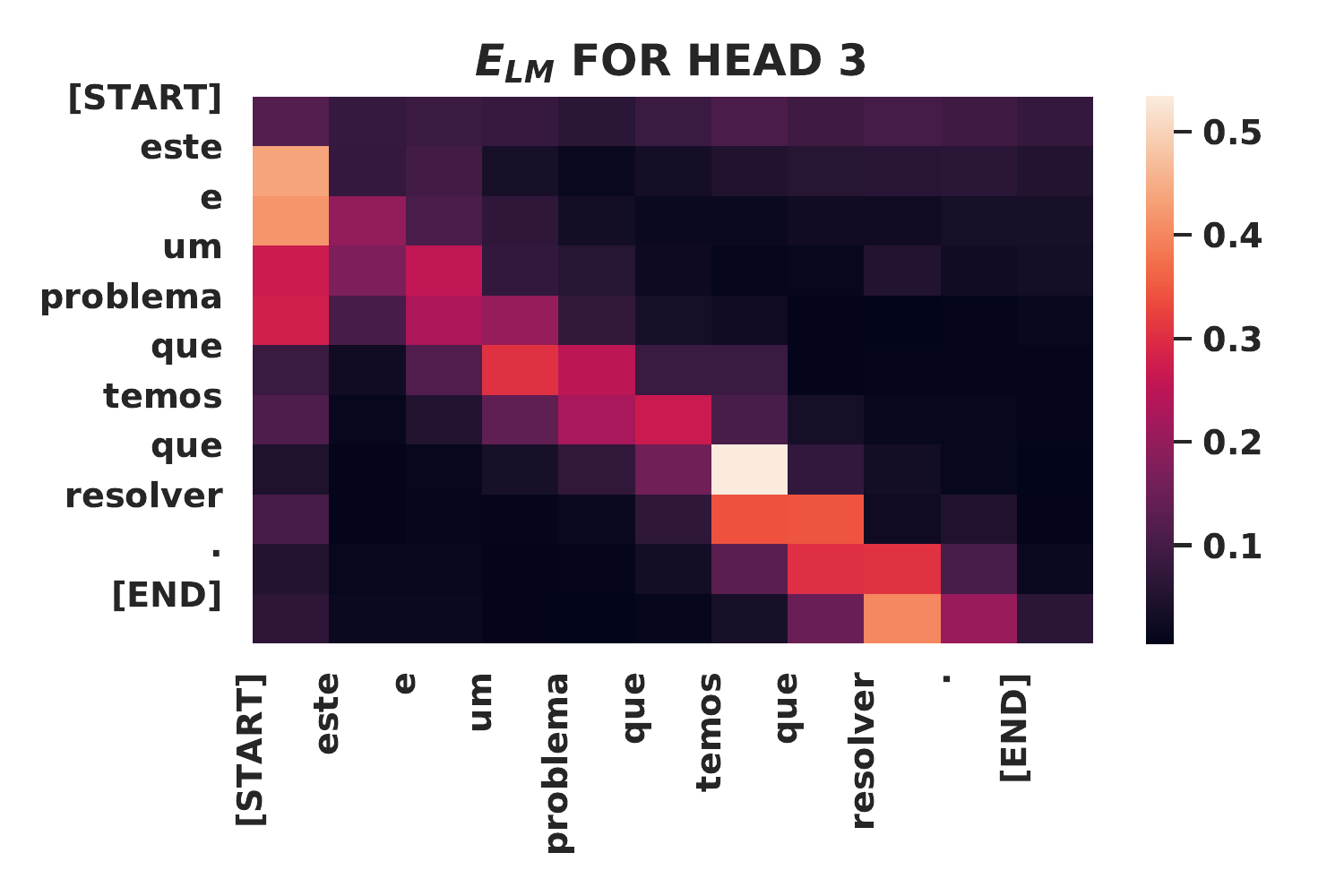}

\end{subfigure}
\hfill
\begin{subfigure}[b]{0.6\textwidth}
	\centering
	\includegraphics[width=1.1\textwidth]{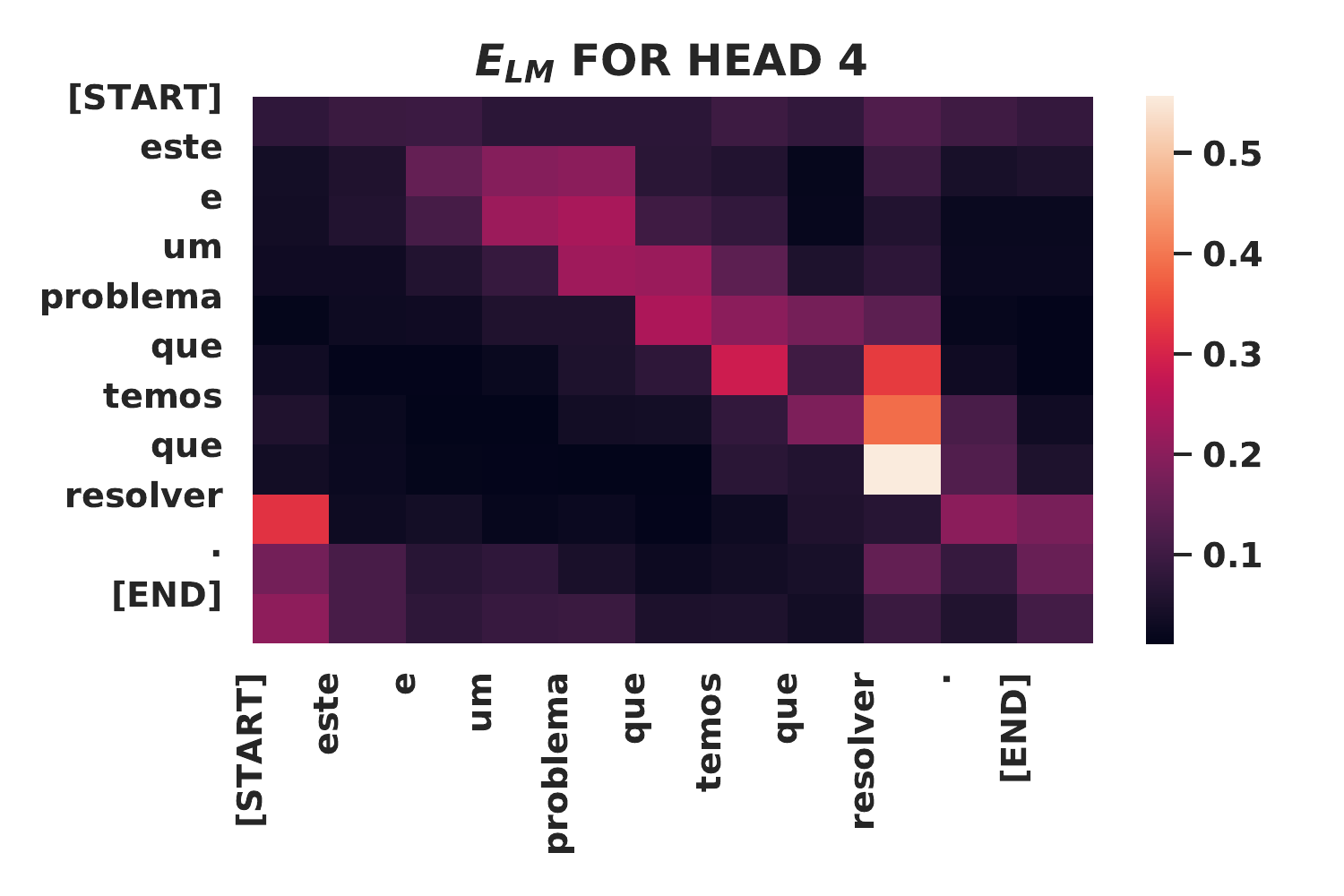}

\end{subfigure}
\centering
\begin{subfigure}[b]{0.6\textwidth}
	\centering
	\includegraphics[width=1.1\textwidth]{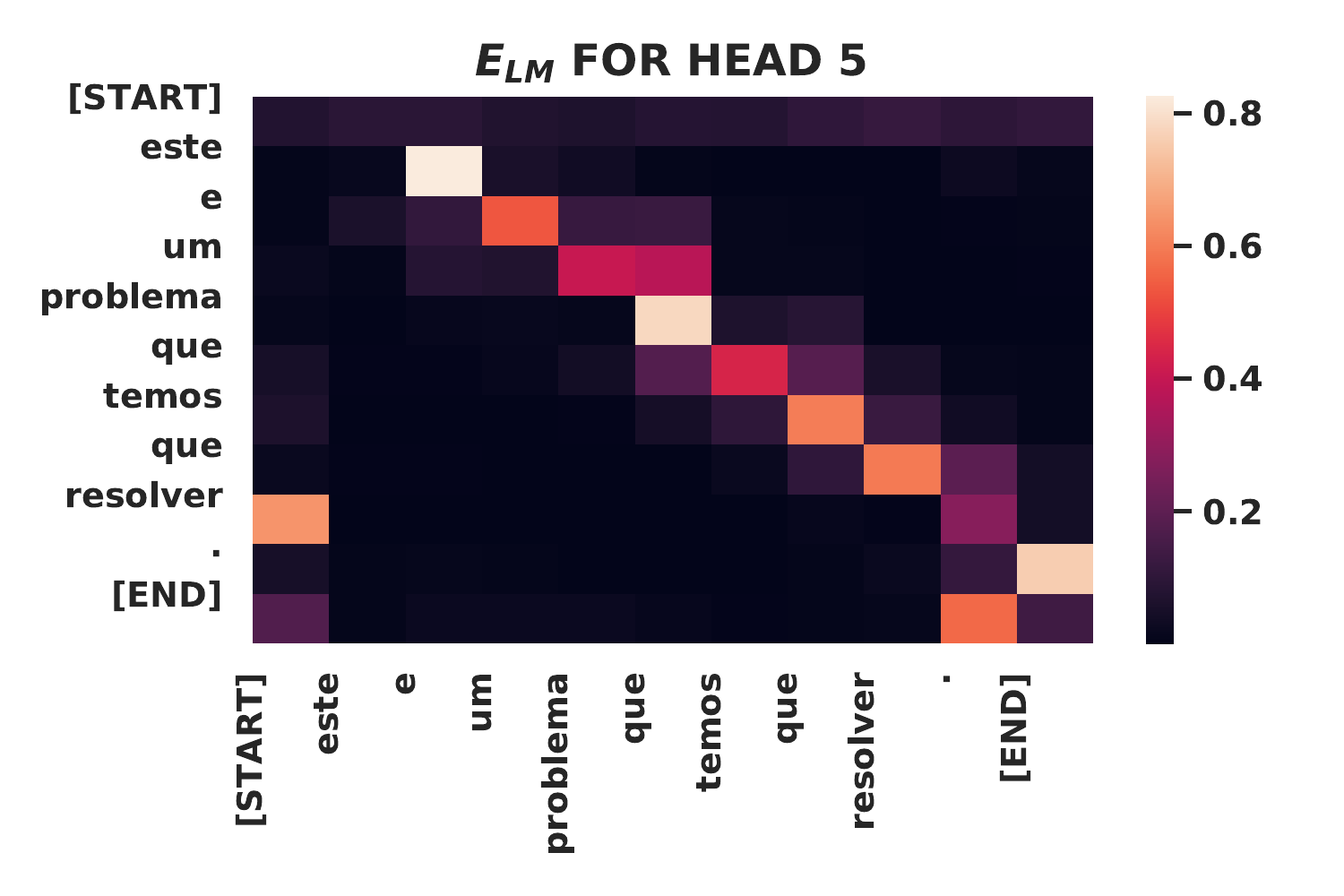}

\end{subfigure}
\hfill
\begin{subfigure}[b]{0.6\textwidth}
	\centering
	\includegraphics[width=1.1\textwidth]{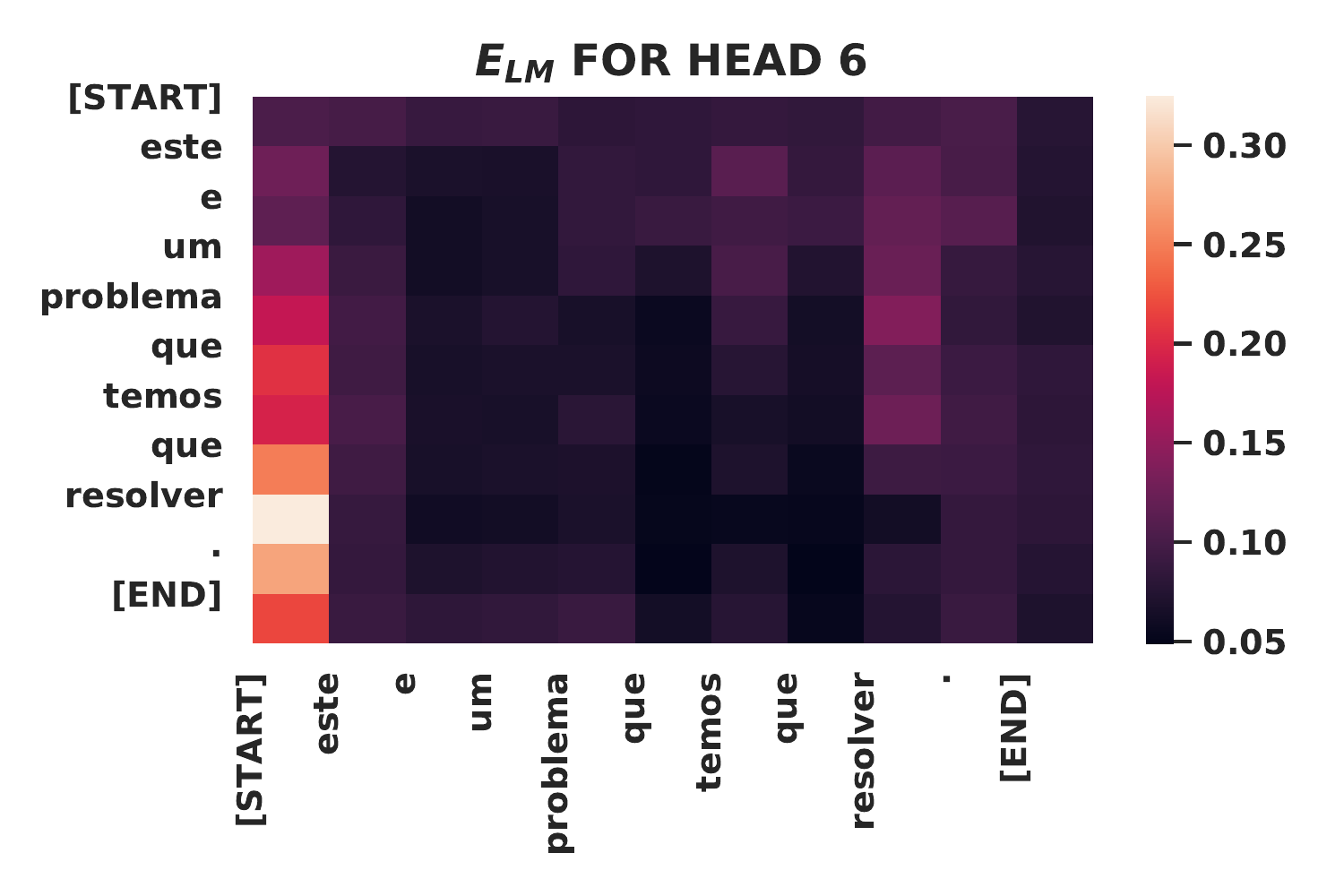}

\end{subfigure}
\hfill
\begin{subfigure}[b]{0.6\textwidth}
	\centering
	\includegraphics[width=1.1\textwidth]{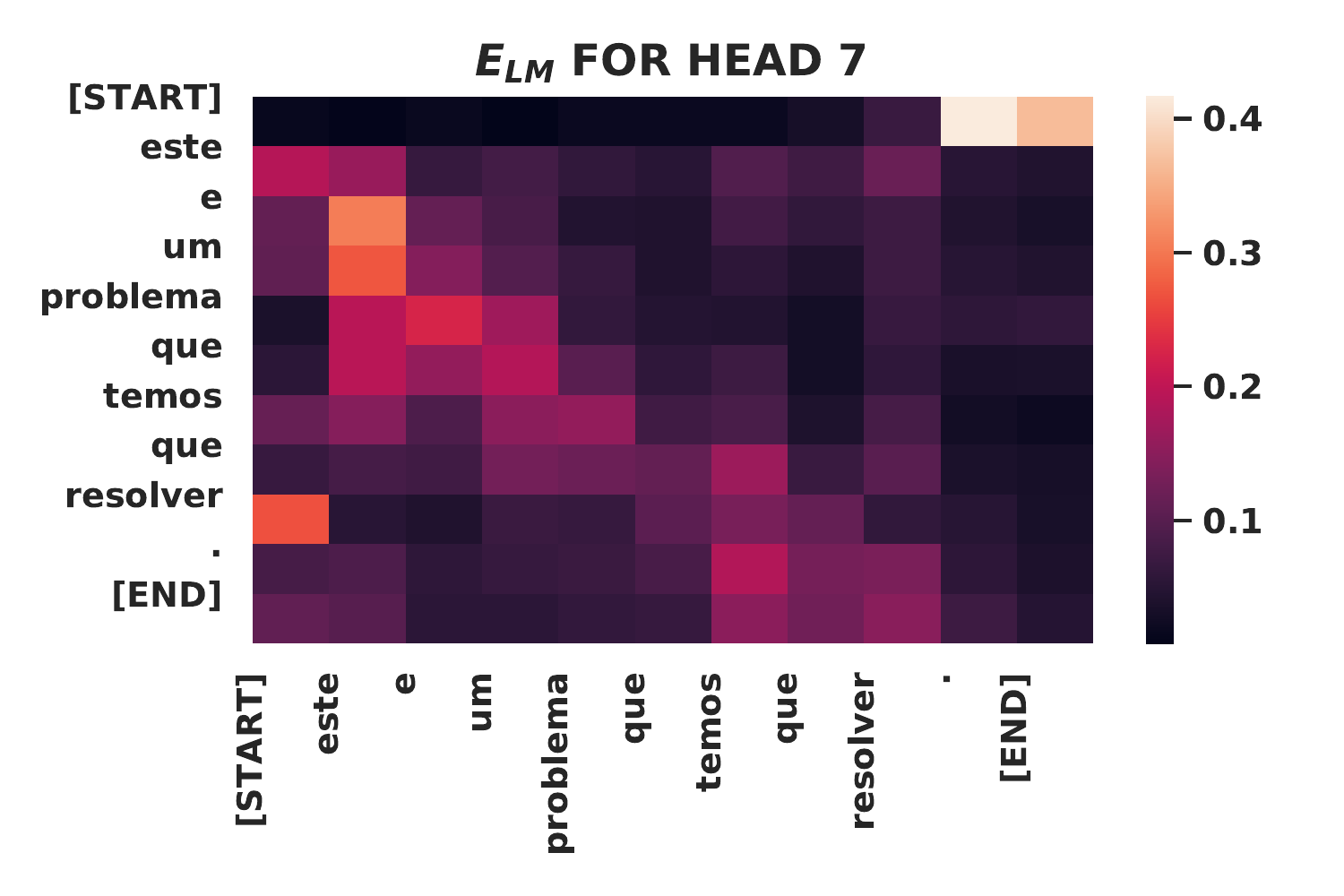}

\end{subfigure}
\hfill
\begin{subfigure}[b]{0.6\textwidth}
	\centering
	\includegraphics[width=1.1\textwidth]{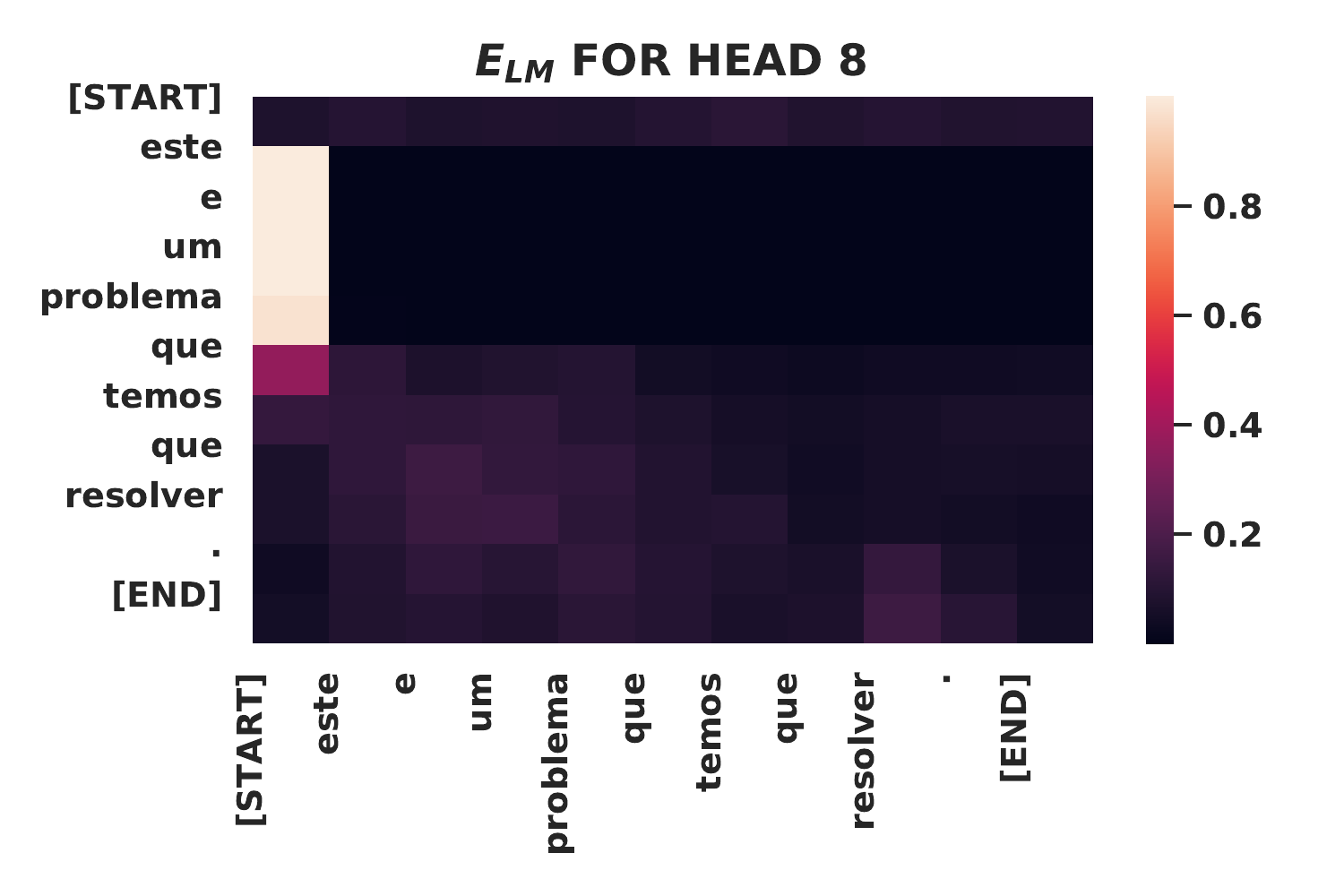}

\end{subfigure}
\caption{$\mE_{LM}$ heatmap plots for all heads from SLM attention stage from graph transformer model \#2 for PT-EN translation task.}
\label{fig12apx}
\end{adjustwidth}
\end{figure}  

\clearpage
\thispagestyle{headings}
\begin{figure}
\begin{adjustwidth}{-5em}{-5em}
\centering
\begin{subfigure}[b]{0.6\textwidth}
	\centering
	\includegraphics[width=1.1\textwidth]{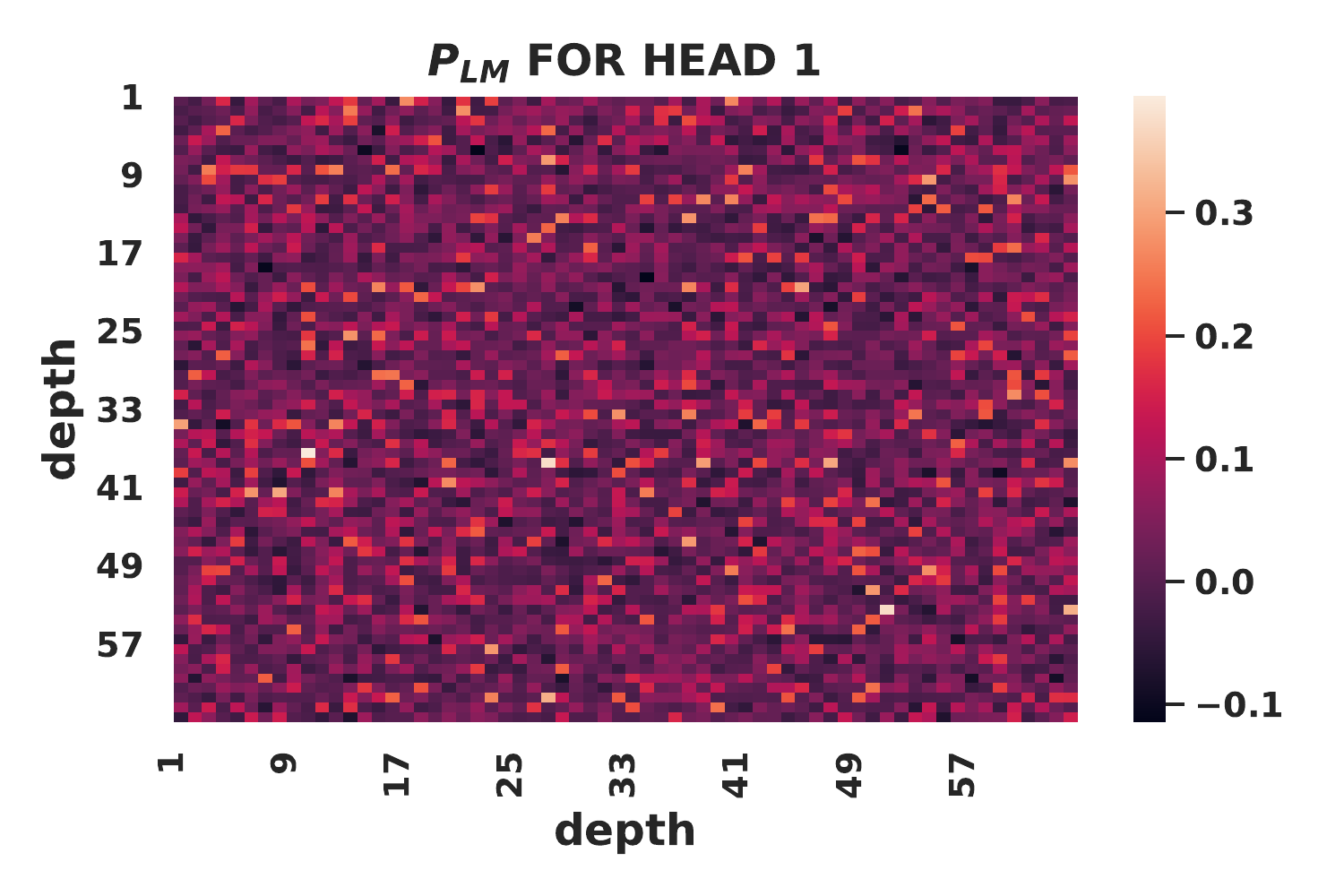}

\end{subfigure}
\hfill
\begin{subfigure}[b]{0.6\textwidth}
	\centering
	\includegraphics[width=1.1\textwidth]{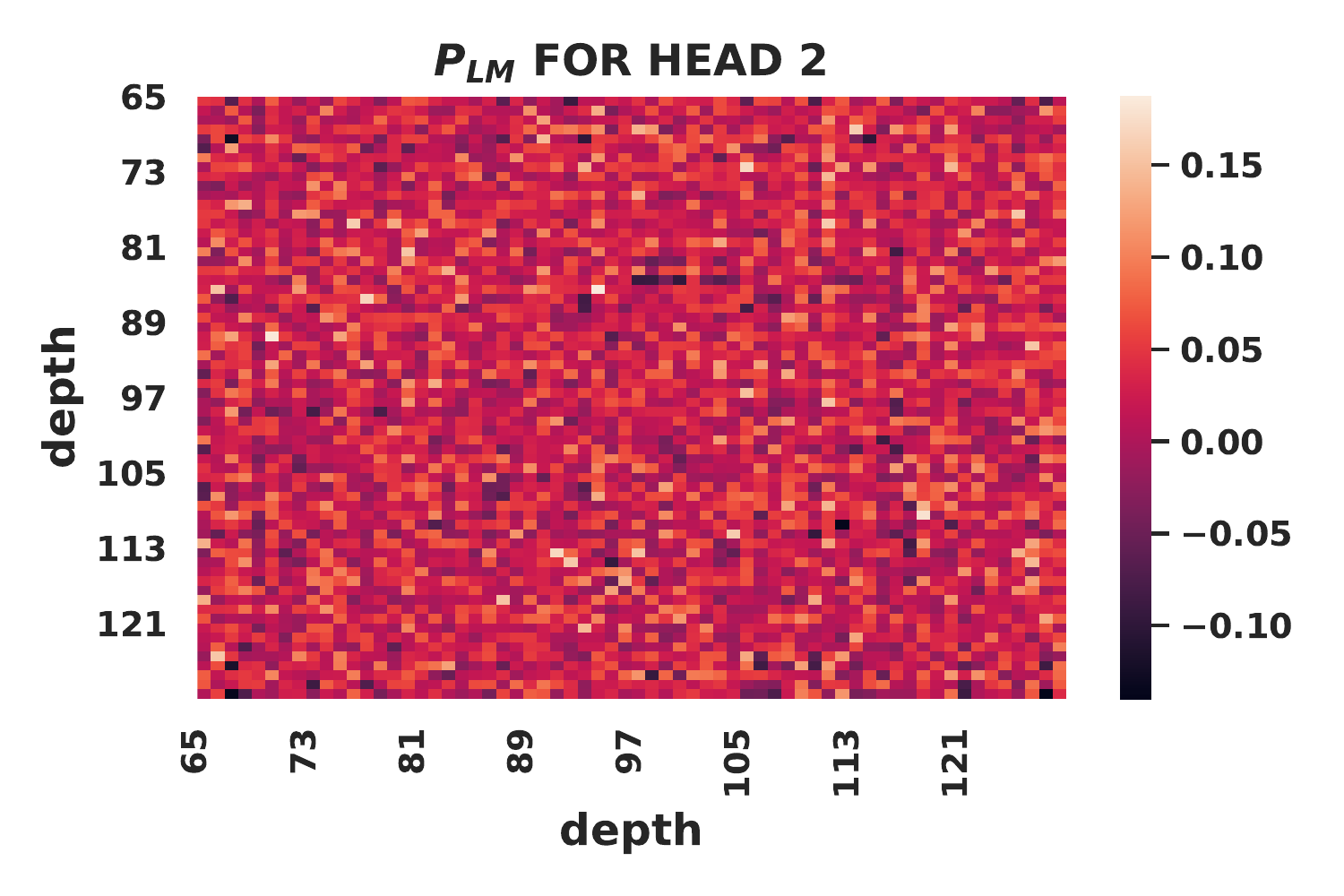}

\end{subfigure}
\hfill
\begin{subfigure}[b]{0.6\textwidth}
	\centering
	\includegraphics[width=1.1\textwidth]{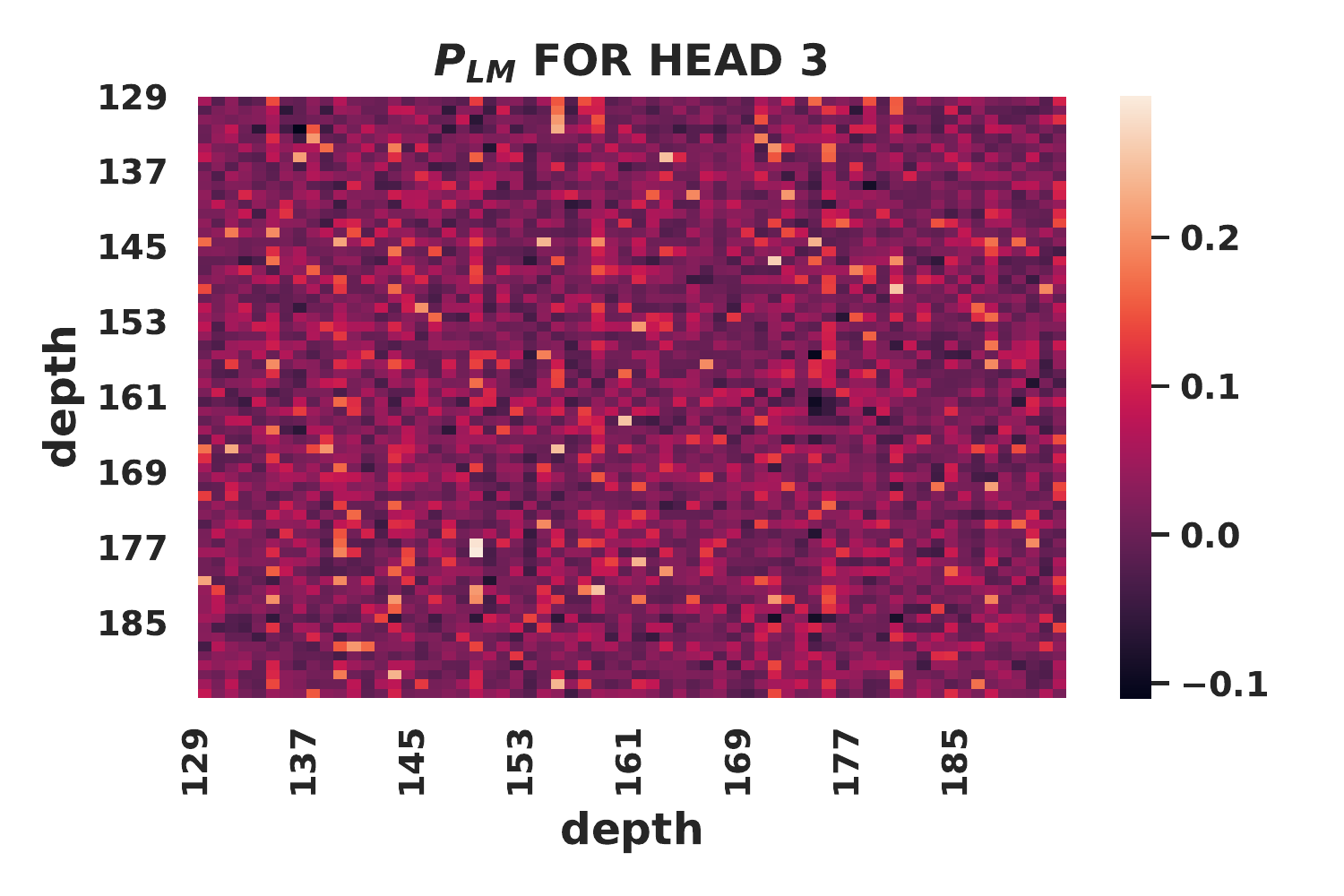}

\end{subfigure}
\hfill
\begin{subfigure}[b]{0.6\textwidth}
	\centering
	\includegraphics[width=1.1\textwidth]{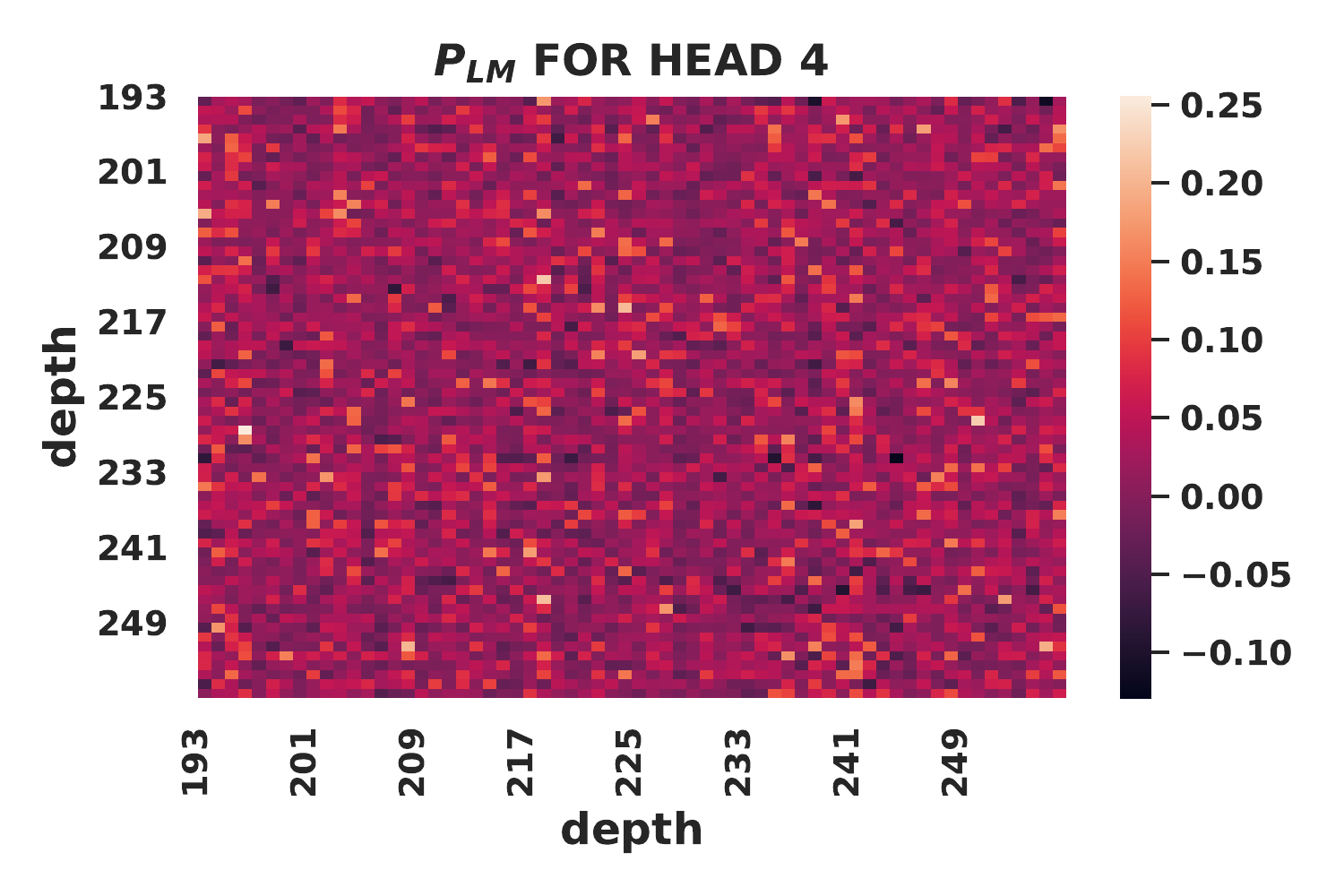}

\end{subfigure}
\centering
\begin{subfigure}[b]{0.6\textwidth}
	\centering
	\includegraphics[width=1.1\textwidth]{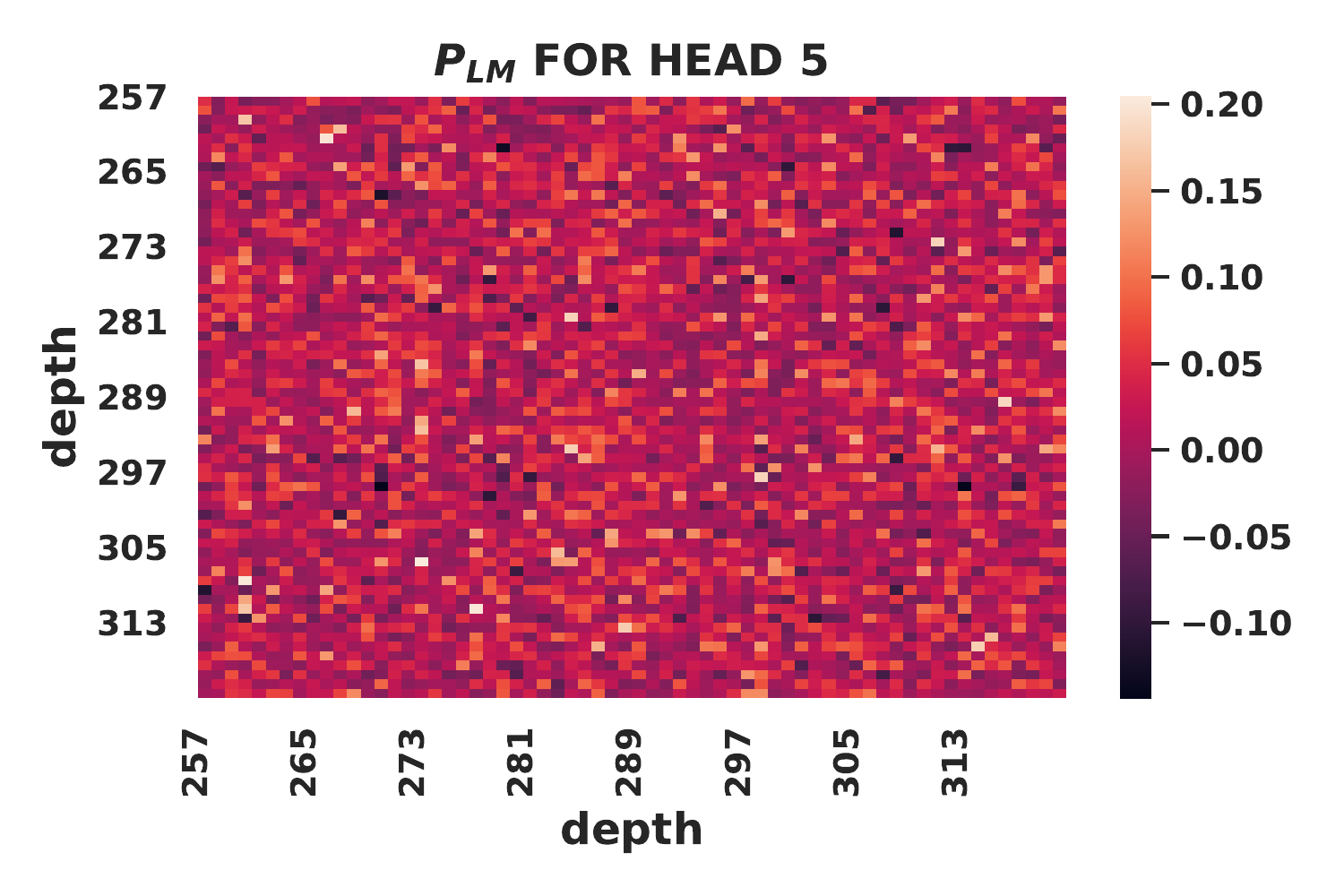}

\end{subfigure}
\hfill
\begin{subfigure}[b]{0.6\textwidth}
	\centering
	\includegraphics[width=1.1\textwidth]{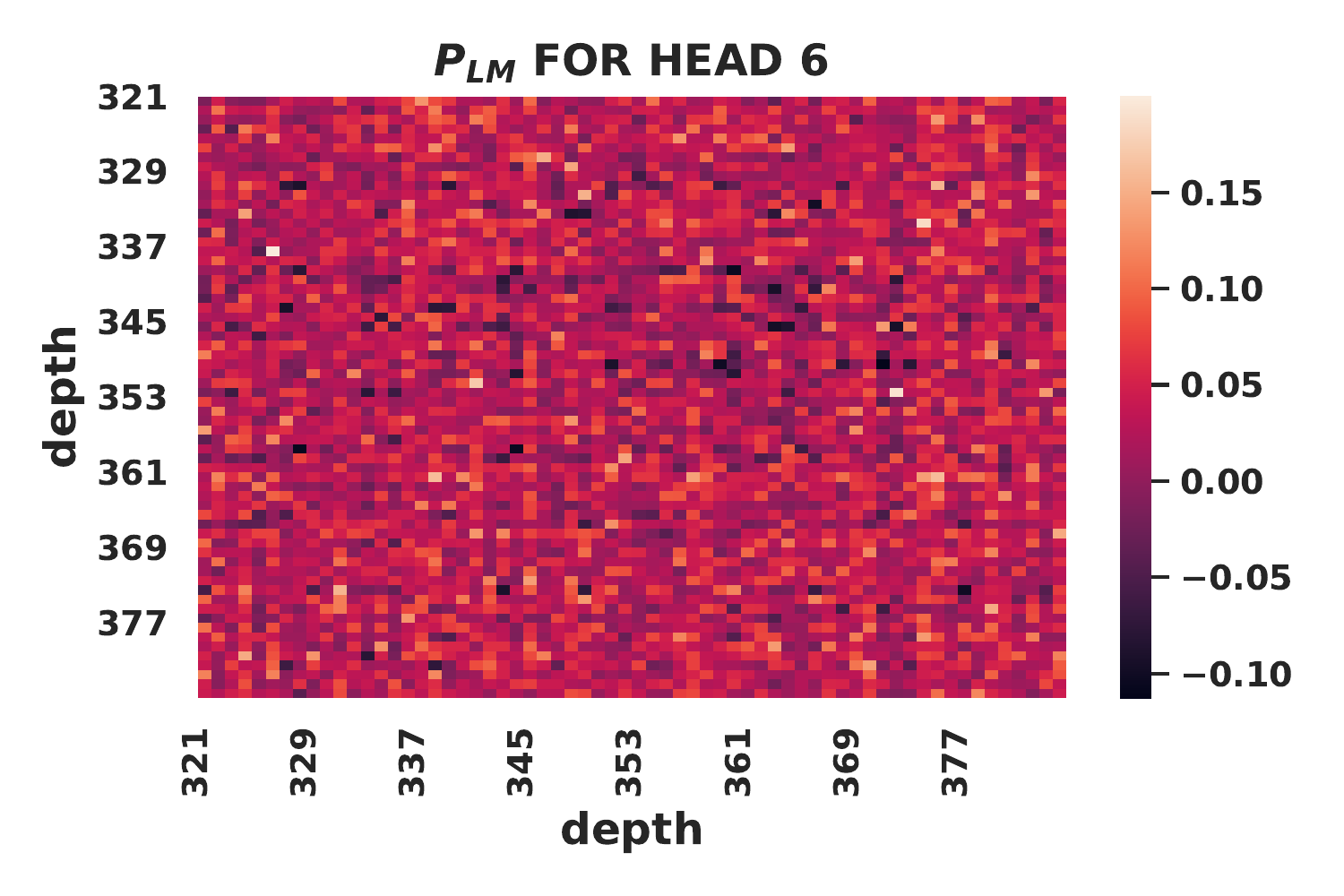}

\end{subfigure}
\hfill
\begin{subfigure}[b]{0.6\textwidth}
	\centering
	\includegraphics[width=1.1\textwidth]{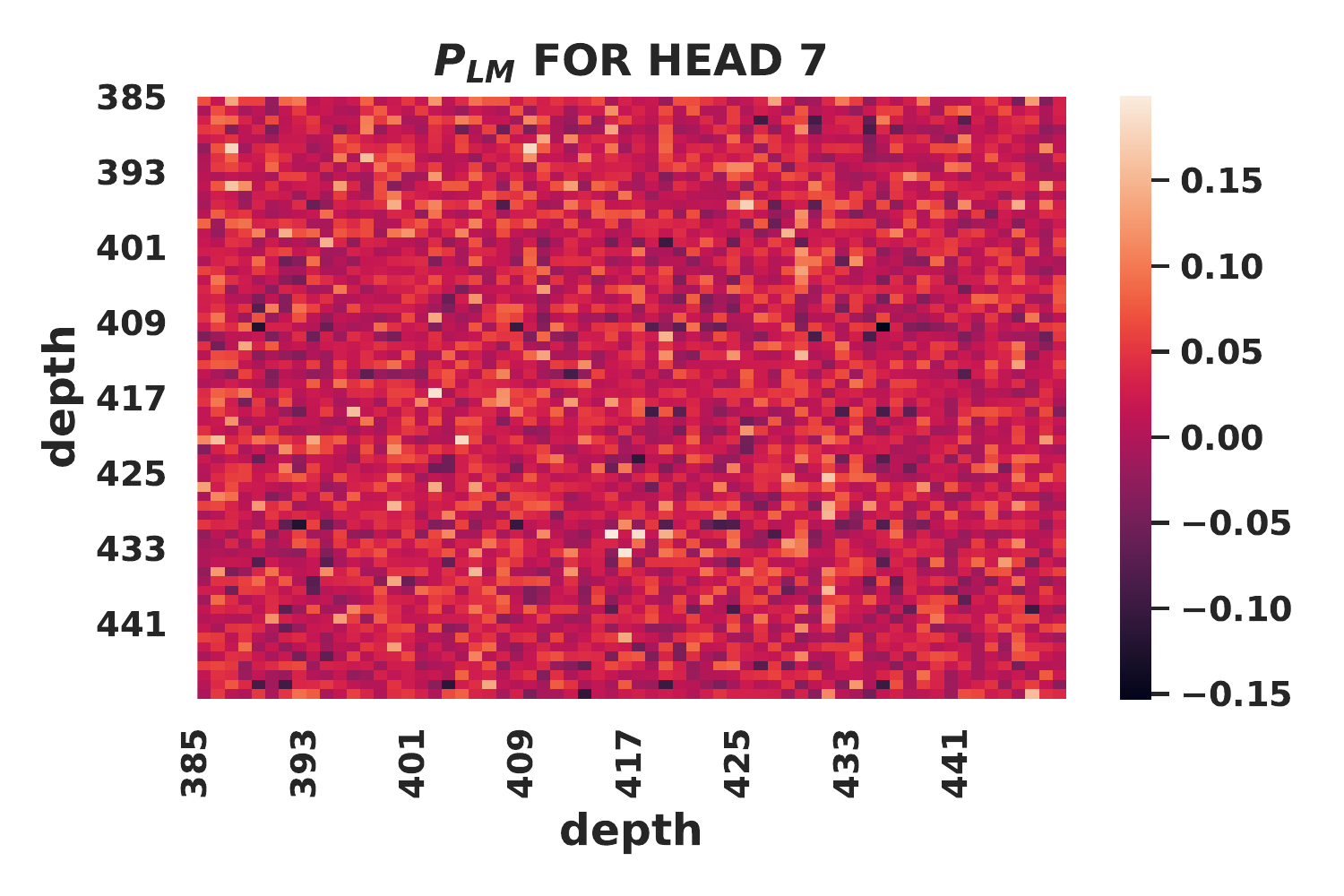}

\end{subfigure}
\hfill
\begin{subfigure}[b]{0.6\textwidth}
	\centering
	\includegraphics[width=1.1\textwidth]{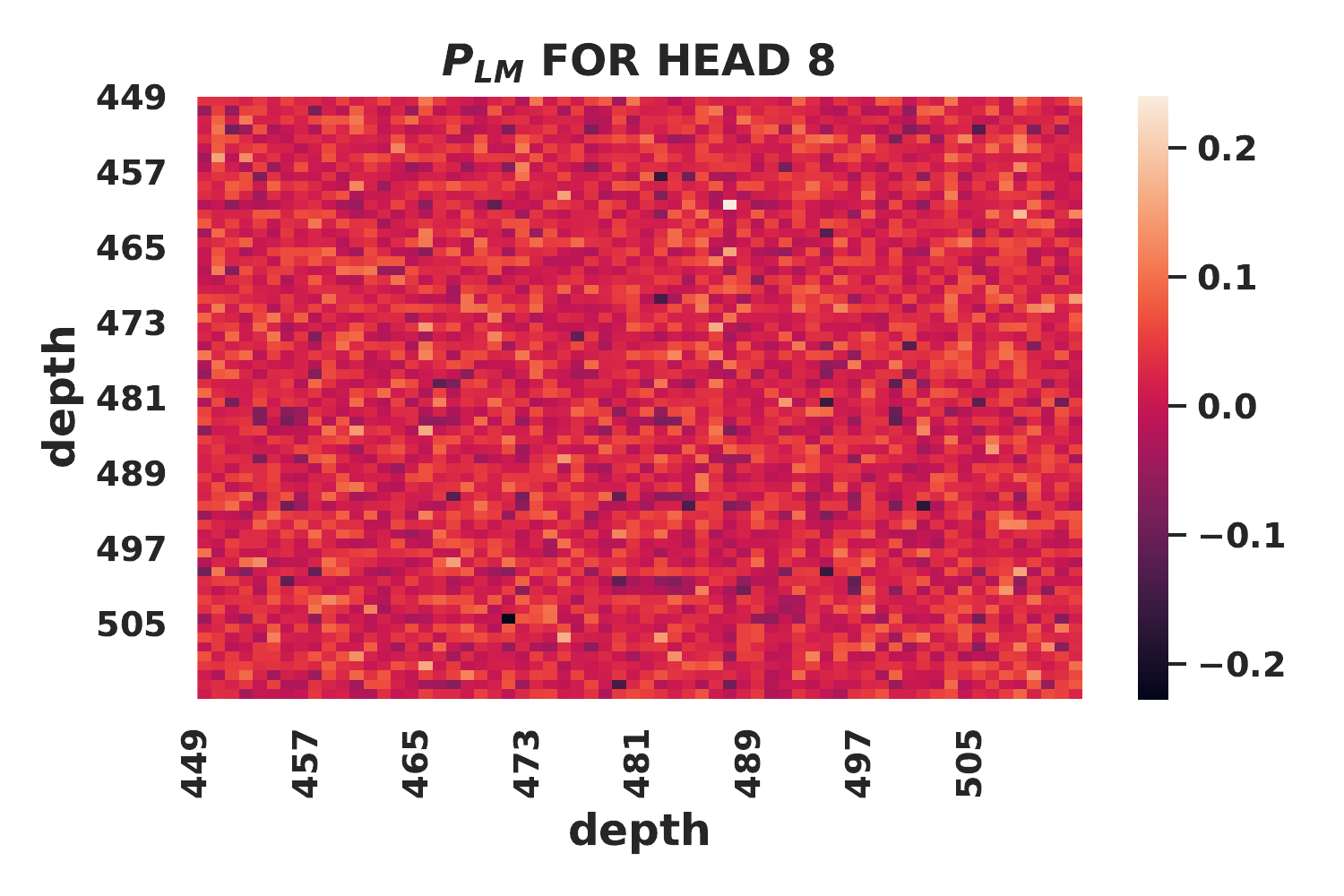}

\end{subfigure}
\caption{$\mP_{LM}$ heatmap plots for all heads from SLM attention stage from graph transformer model \#2 for PT-EN translation task. }
\label{fig13apx}
\end{adjustwidth}
\end{figure}

\clearpage
\thispagestyle{headings}

\begin{figure}
\begin{adjustwidth}{-5em}{-5em}
\centering
\begin{subfigure}[b]{0.6\textwidth}
	\centering
	\includegraphics[width=1.1\textwidth]{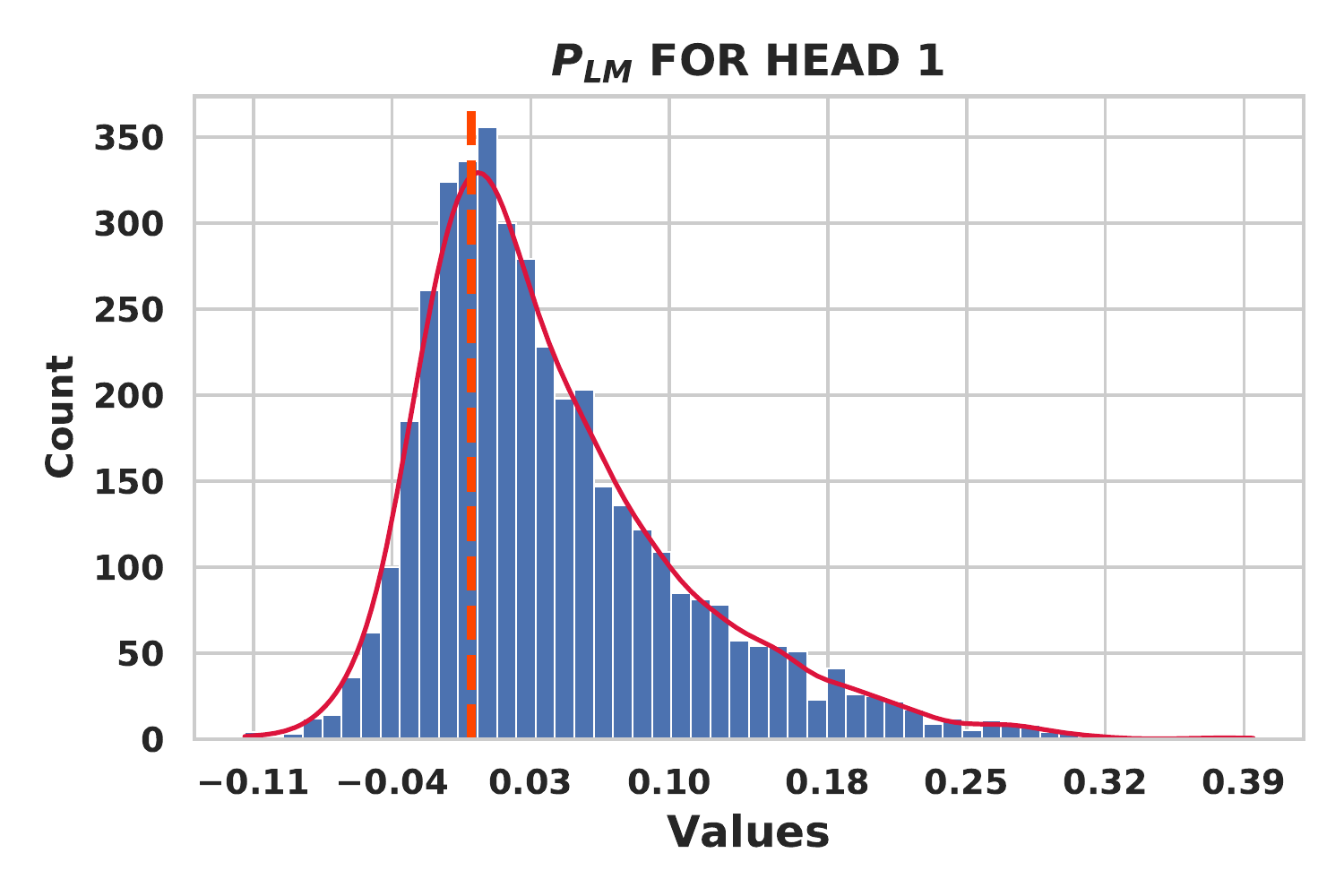}

\end{subfigure}
\hfill
\begin{subfigure}[b]{0.6\textwidth}
	\centering
	\includegraphics[width=1.1\textwidth]{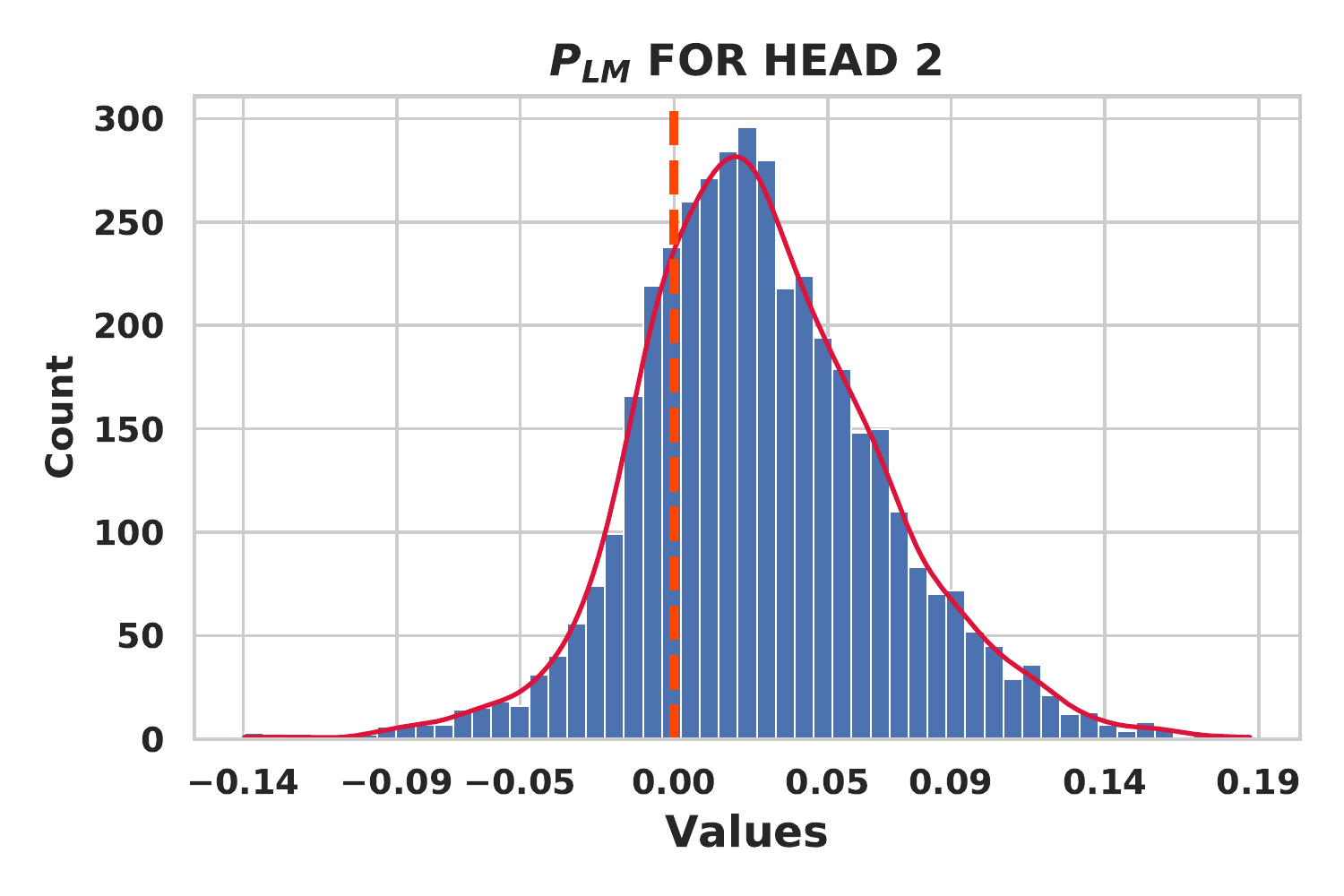}

\end{subfigure}
\hfill
\begin{subfigure}[b]{0.6\textwidth}
	\centering
	\includegraphics[width=1.1\textwidth]{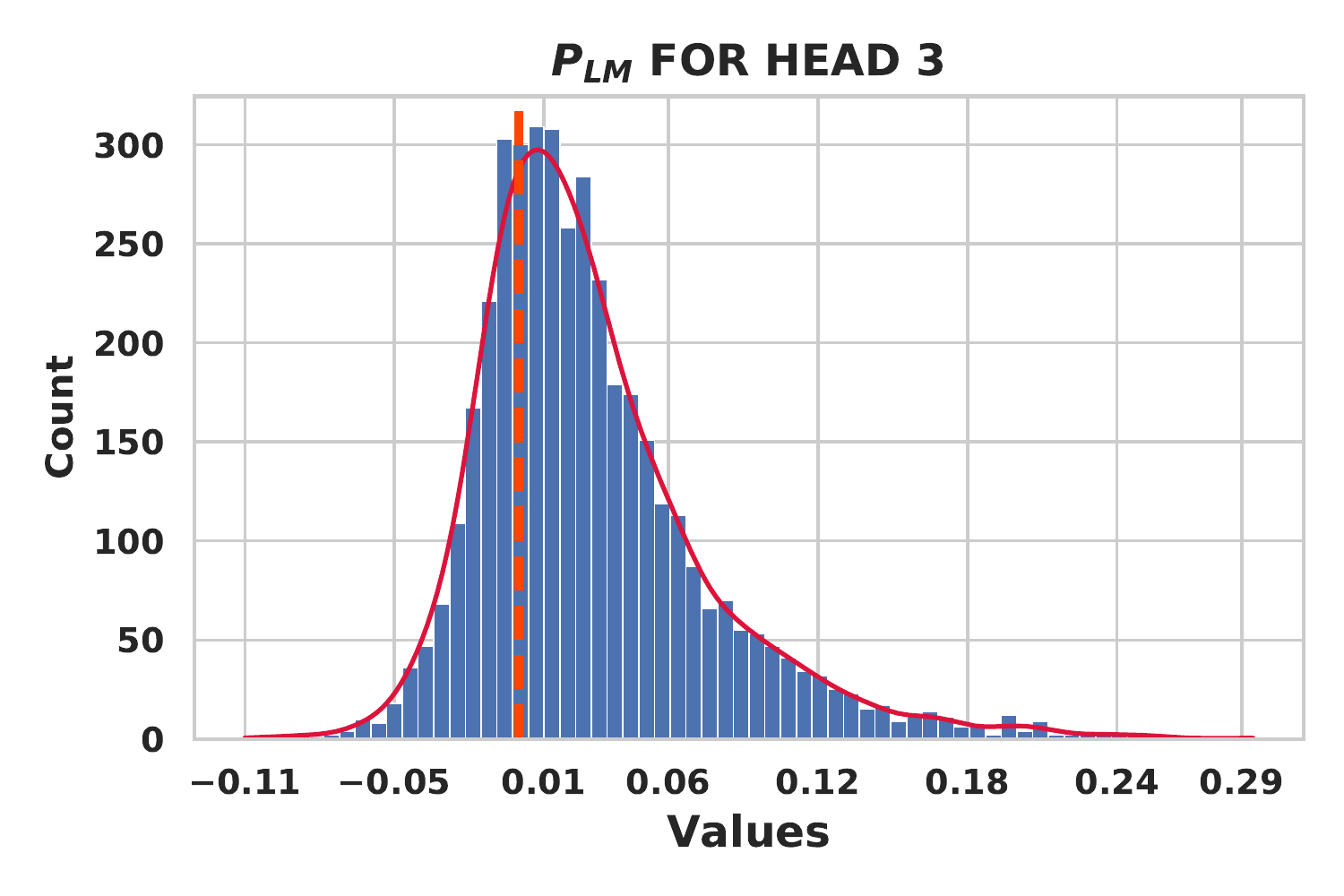}

\end{subfigure}
\hfill
\begin{subfigure}[b]{0.6\textwidth}
	\centering
	\includegraphics[width=1.1\textwidth]{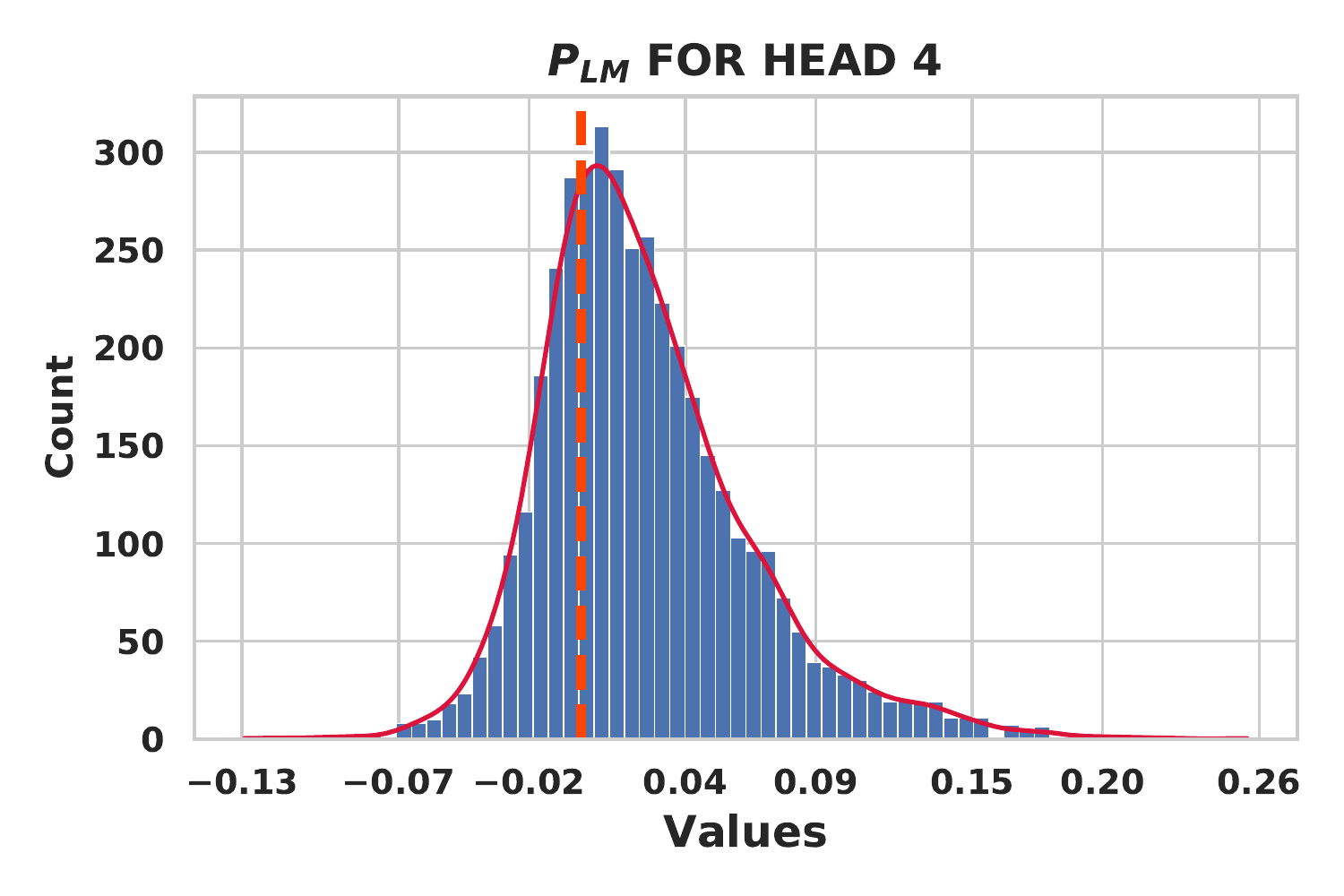}

\end{subfigure}
\centering
\begin{subfigure}[b]{0.6\textwidth}
	\centering
	\includegraphics[width=1.1\textwidth]{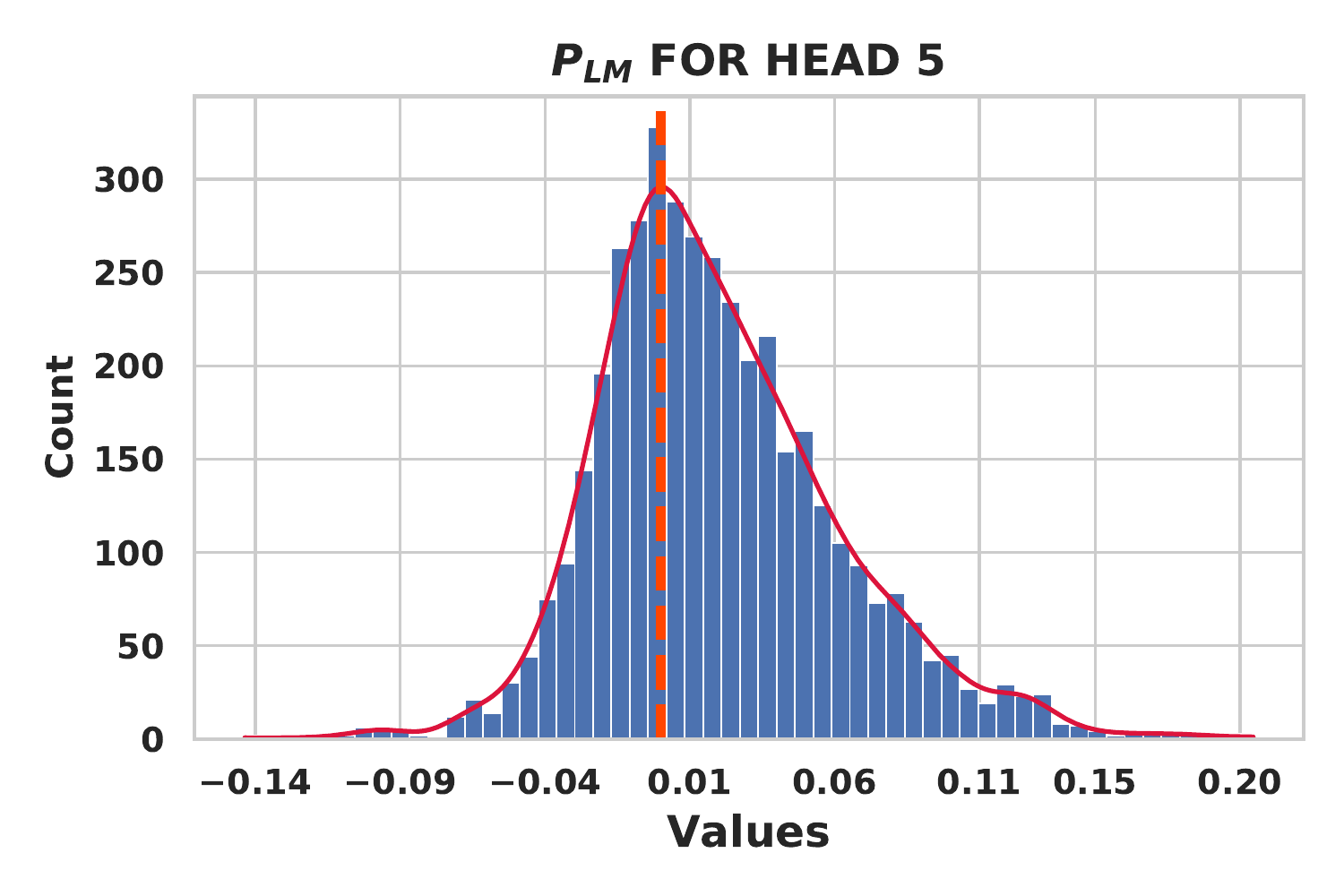}

\end{subfigure}
\hfill
\begin{subfigure}[b]{0.6\textwidth}
	\centering
	\includegraphics[width=1.1\textwidth]{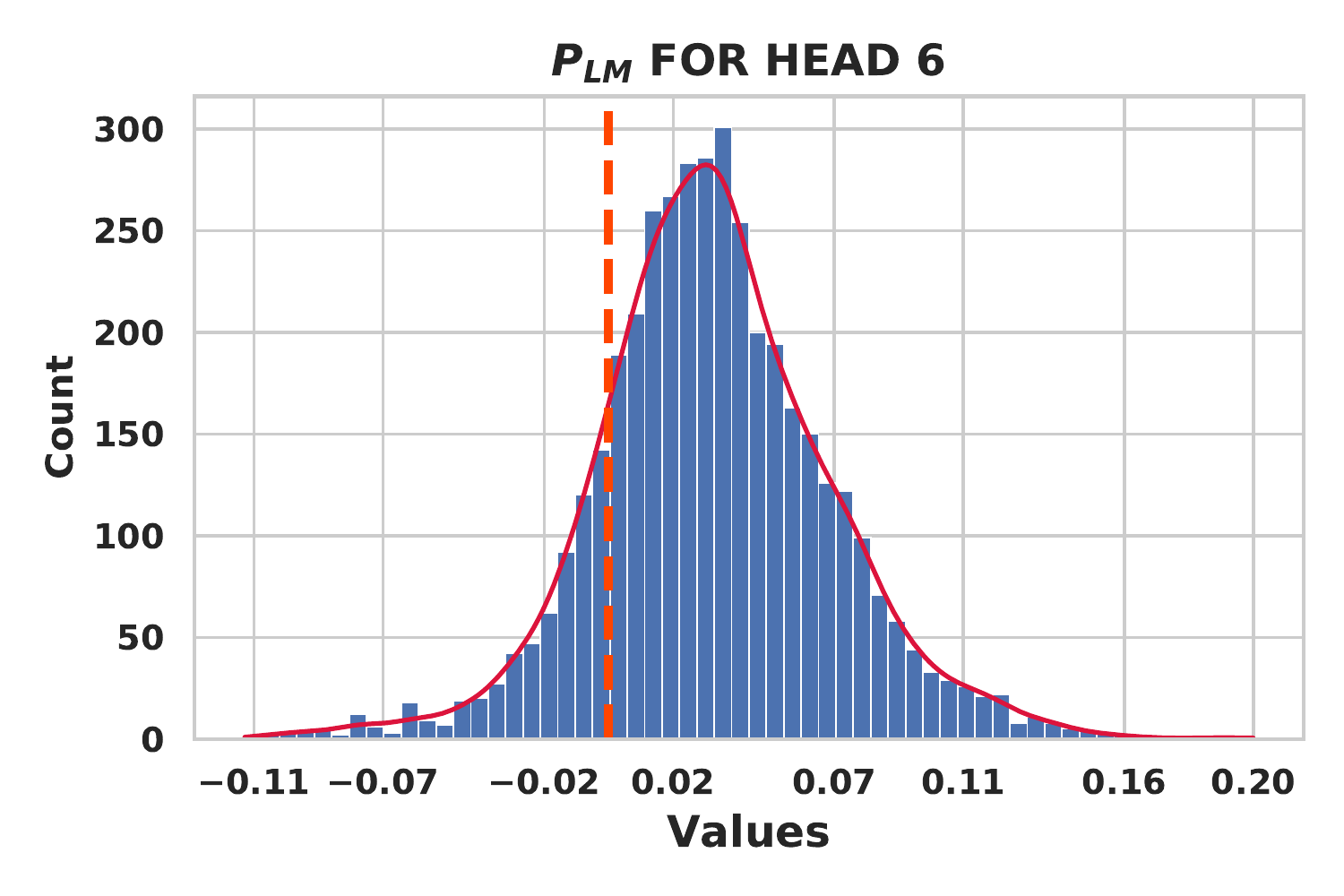}

\end{subfigure}
\hfill
\begin{subfigure}[b]{0.6\textwidth}
	\centering
	\includegraphics[width=1.1\textwidth]{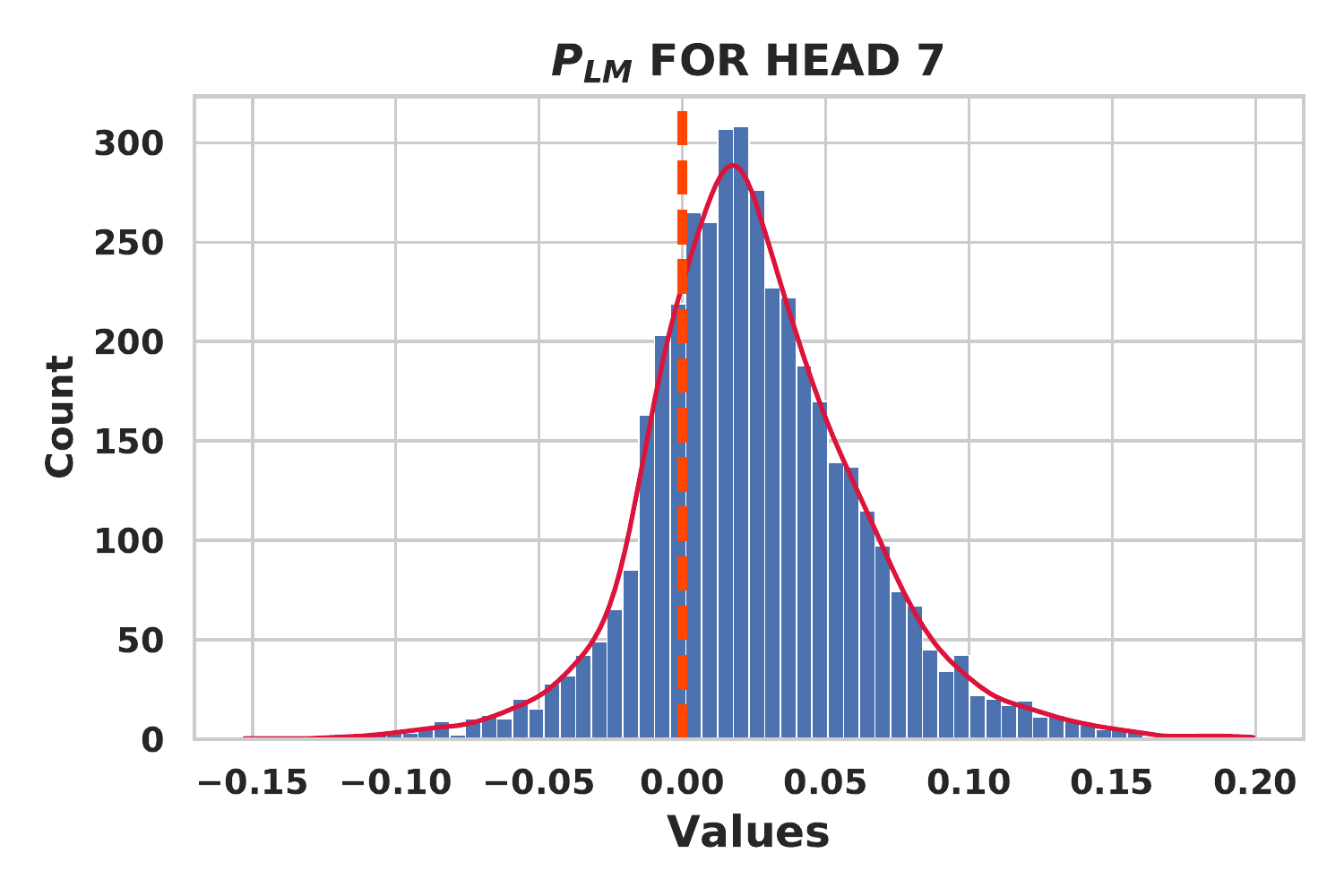}

\end{subfigure}
\hfill
\begin{subfigure}[b]{0.6\textwidth}
	\centering
	\includegraphics[width=1.1\textwidth]{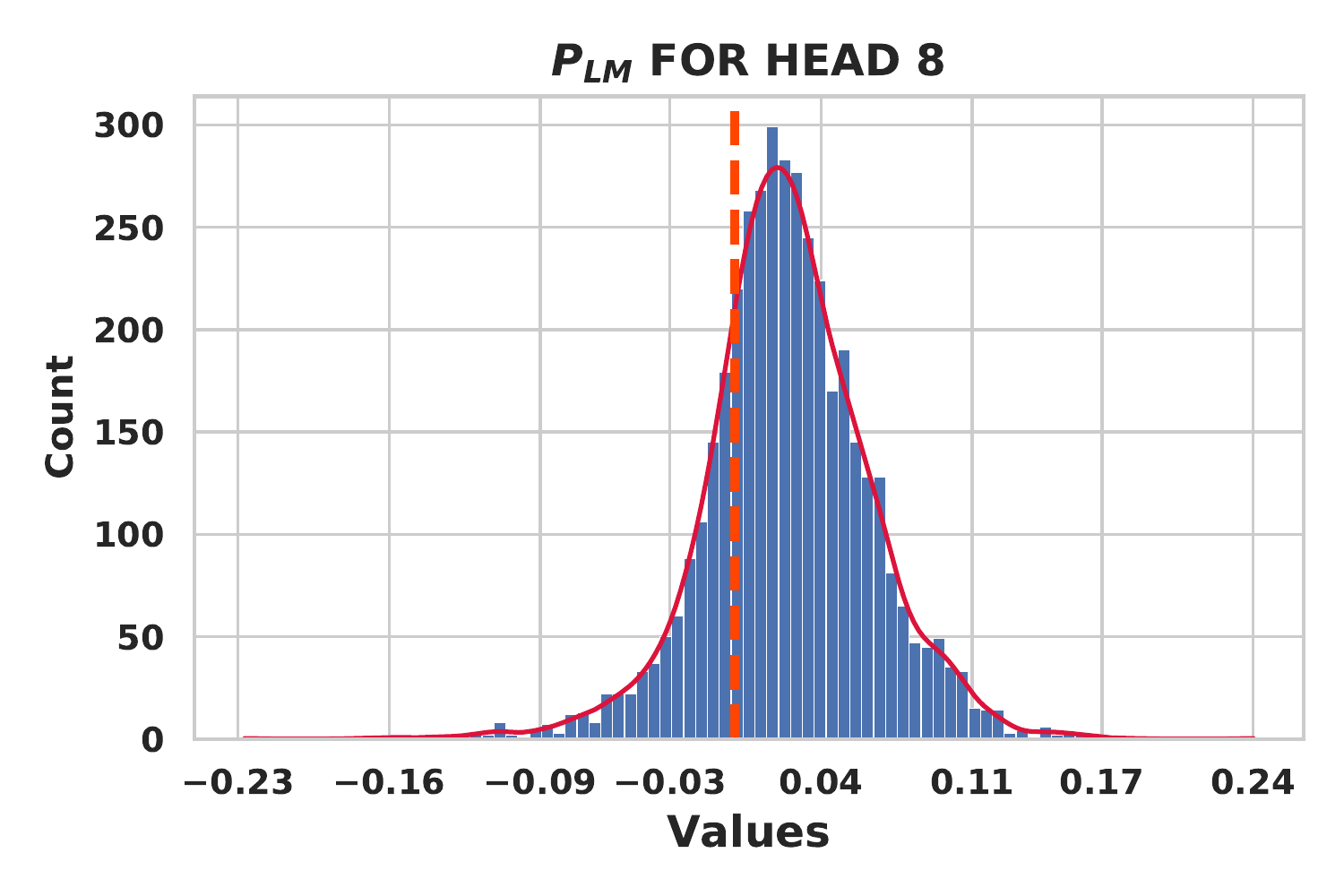}

\end{subfigure}
\caption{$\mP_{LM}$ histogram plots for all heads from SLM attention stage from graph transformer model \#2 for PT-EN translation task. Dashed line in orange marks zero value.}
\label{fig14apx}
\end{adjustwidth}
\end{figure}    

\clearpage
\thispagestyle{headings}

\begin{figure}
\begin{adjustwidth}{-5em}{-5em}
\centering
\begin{subfigure}[b]{0.6\textwidth}
	\centering
	\includegraphics[width=1.1\textwidth]{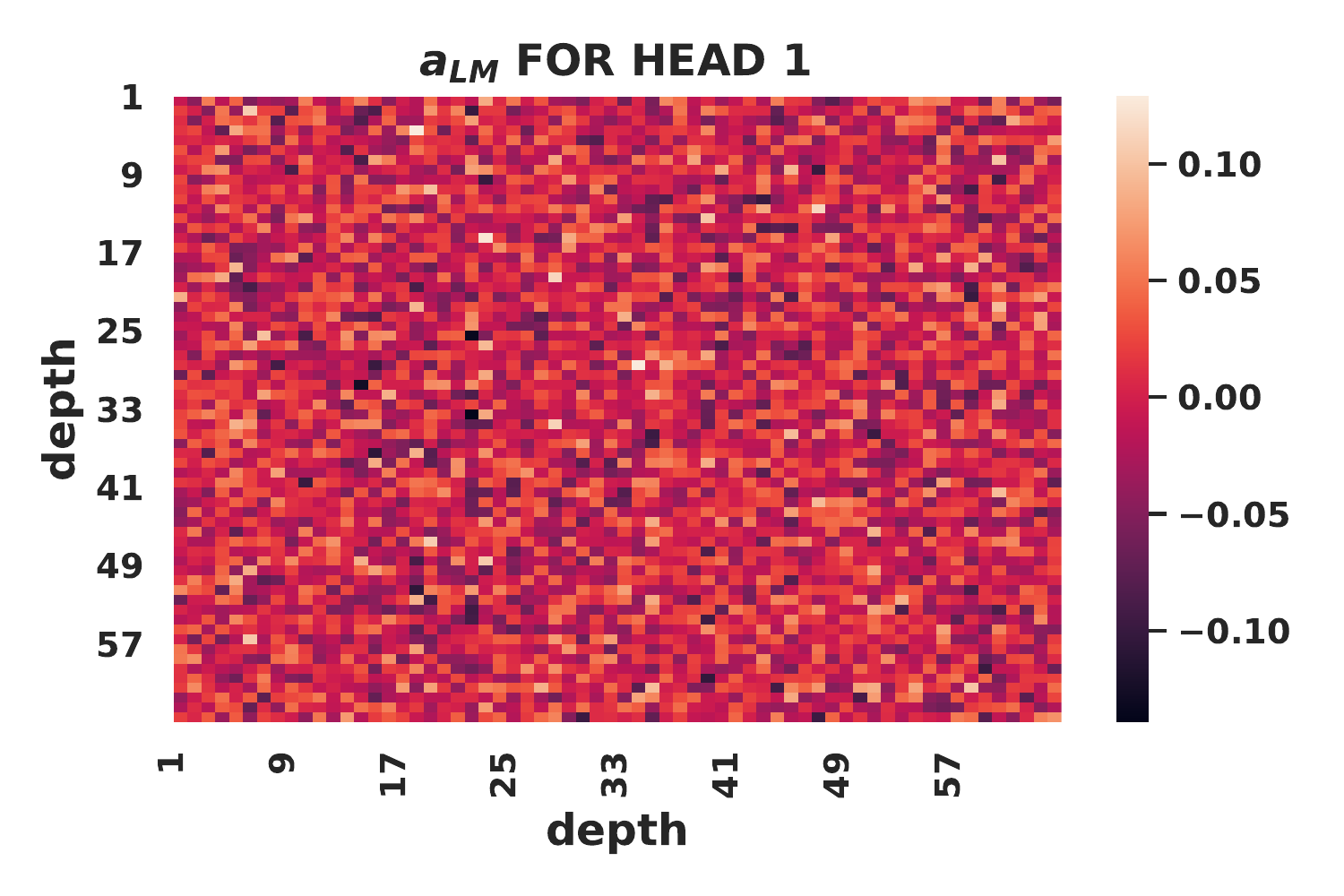}

\end{subfigure}
\hfill
\begin{subfigure}[b]{0.6\textwidth}
	\centering
	\includegraphics[width=1.1\textwidth]{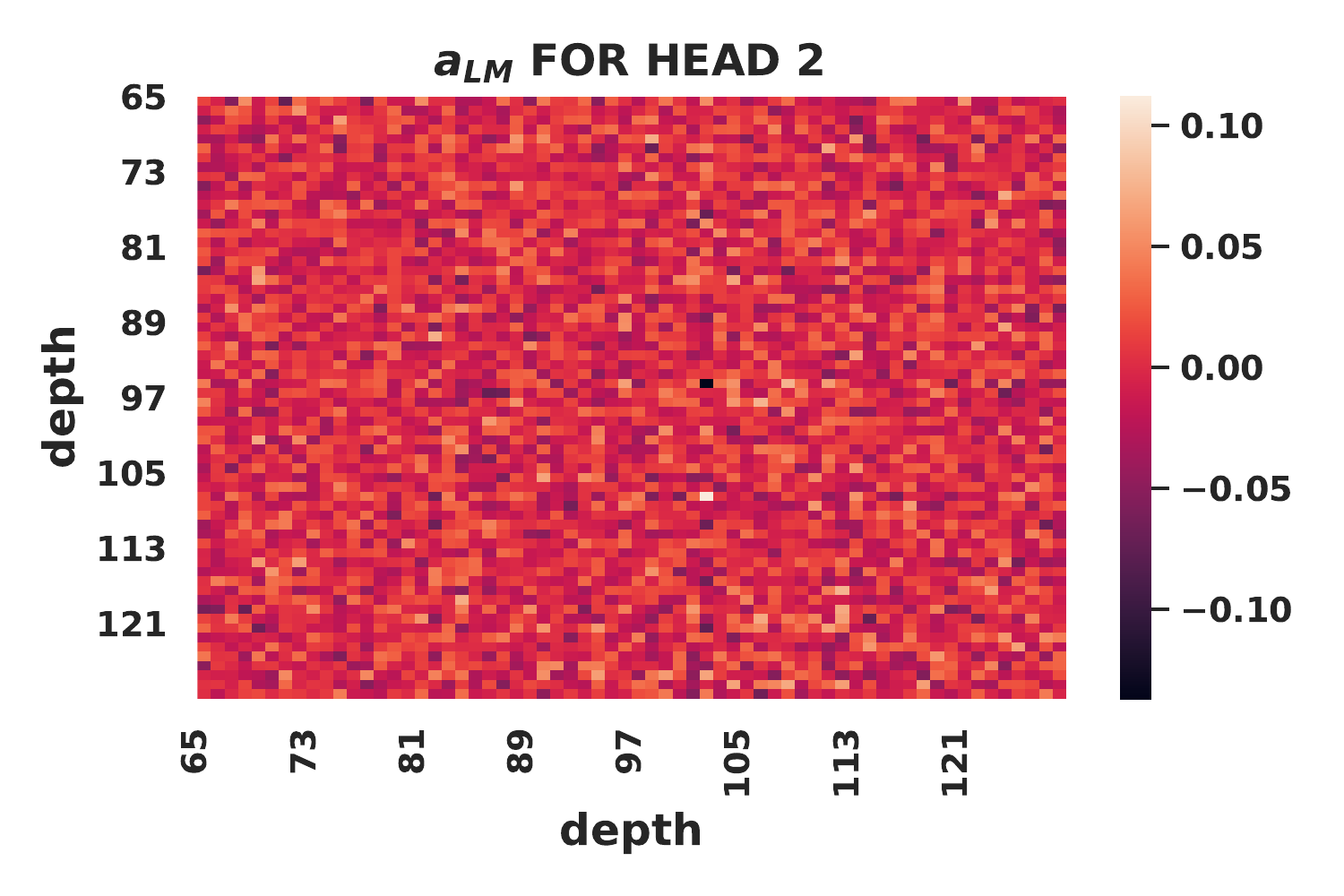}

\end{subfigure}
\hfill
\begin{subfigure}[b]{0.6\textwidth}
	\centering
	\includegraphics[width=1.1\textwidth]{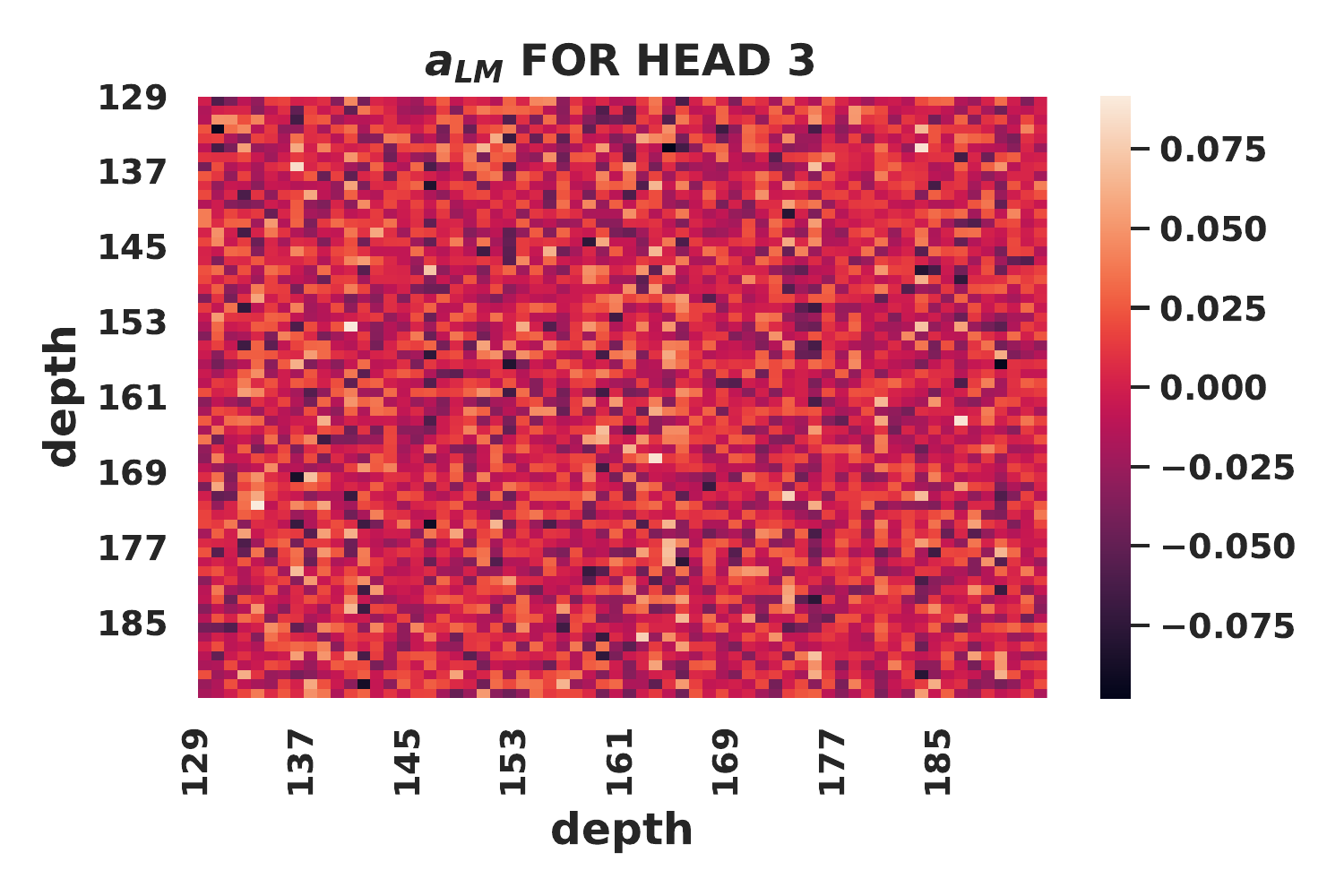}

\end{subfigure}
\hfill
\begin{subfigure}[b]{0.6\textwidth}
	\centering
	\includegraphics[width=1.1\textwidth]{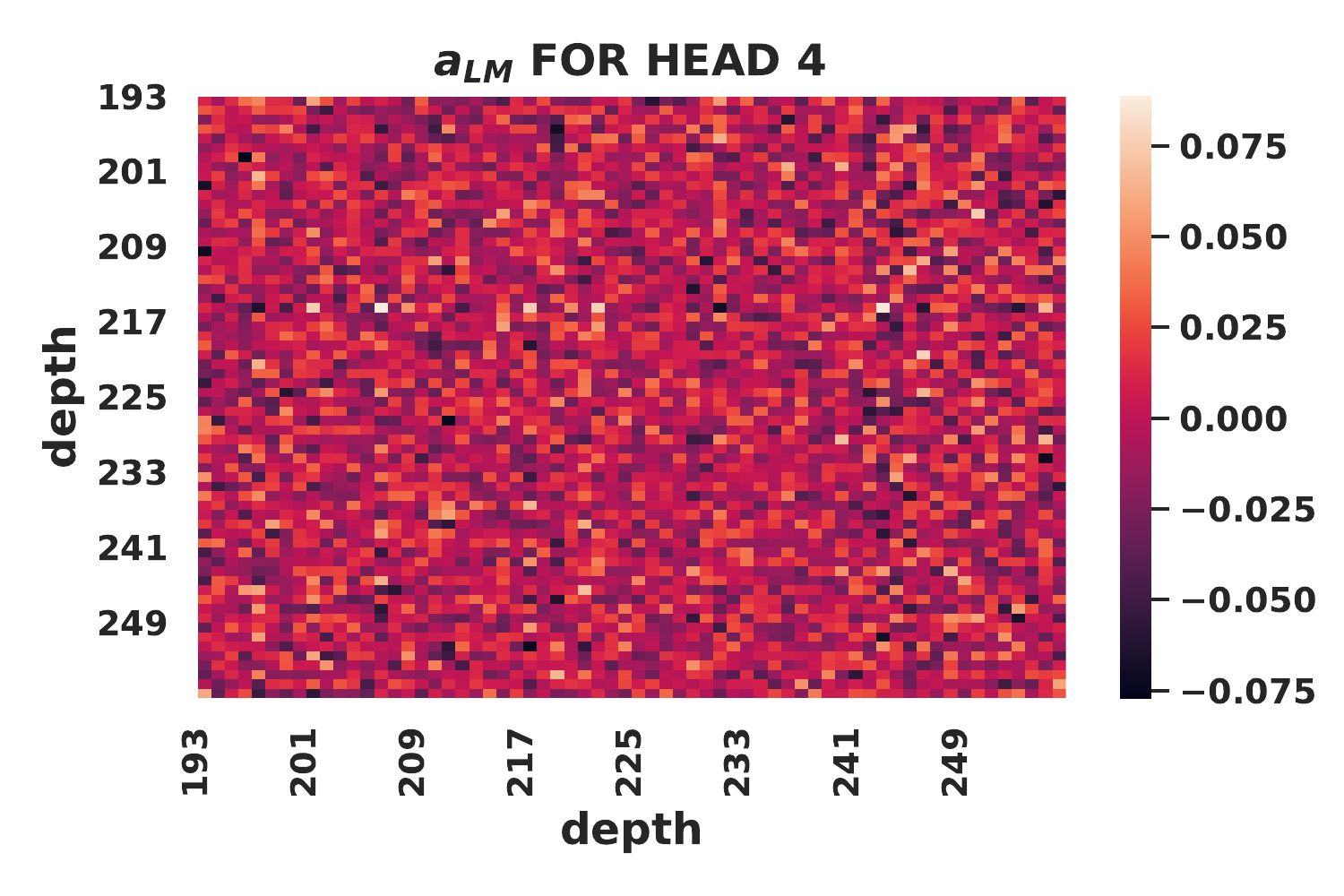}

\end{subfigure}
\centering
\begin{subfigure}[b]{0.6\textwidth}
	\centering
	\includegraphics[width=1.1\textwidth]{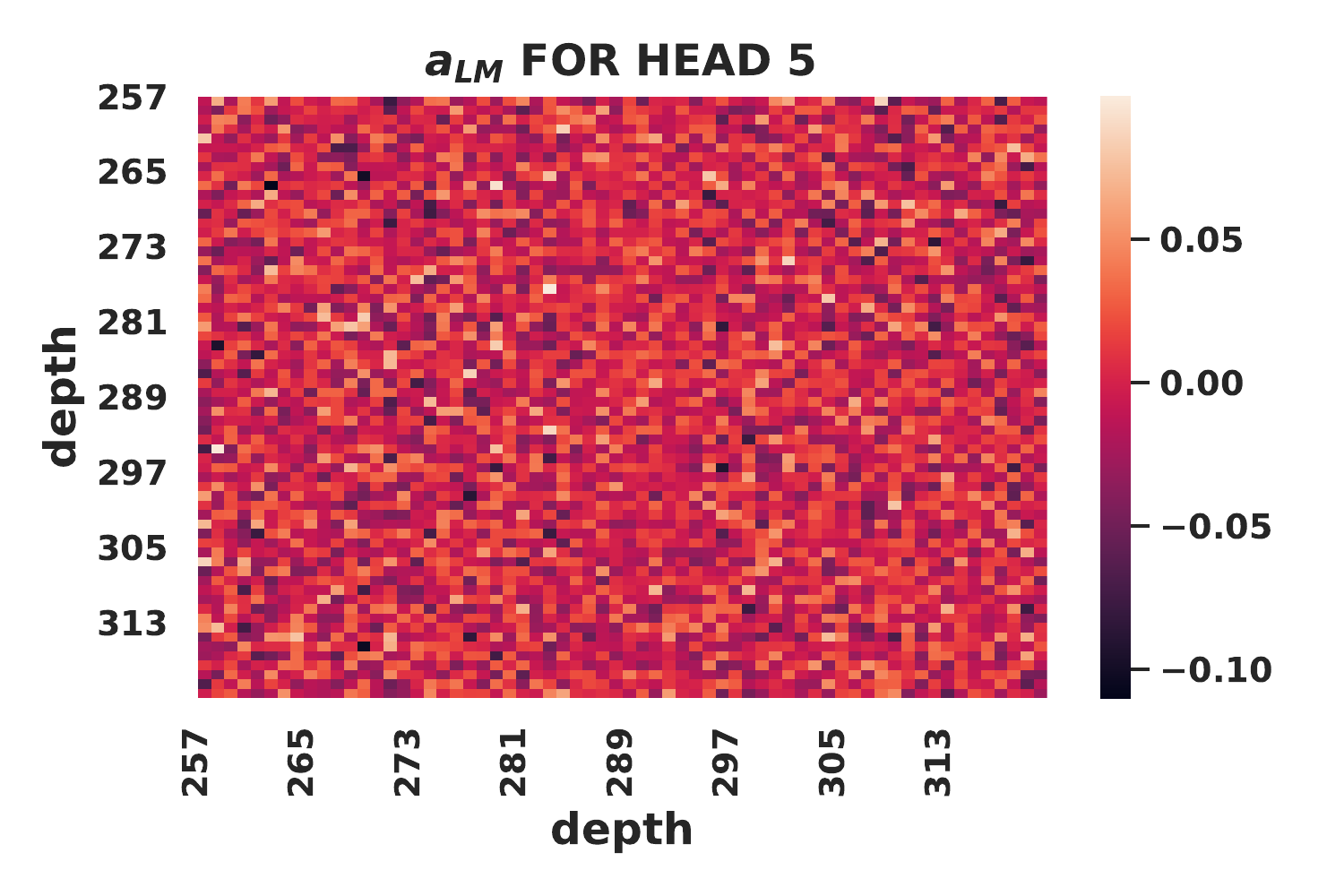}

\end{subfigure}
\hfill
\begin{subfigure}[b]{0.6\textwidth}
	\centering
	\includegraphics[width=1.1\textwidth]{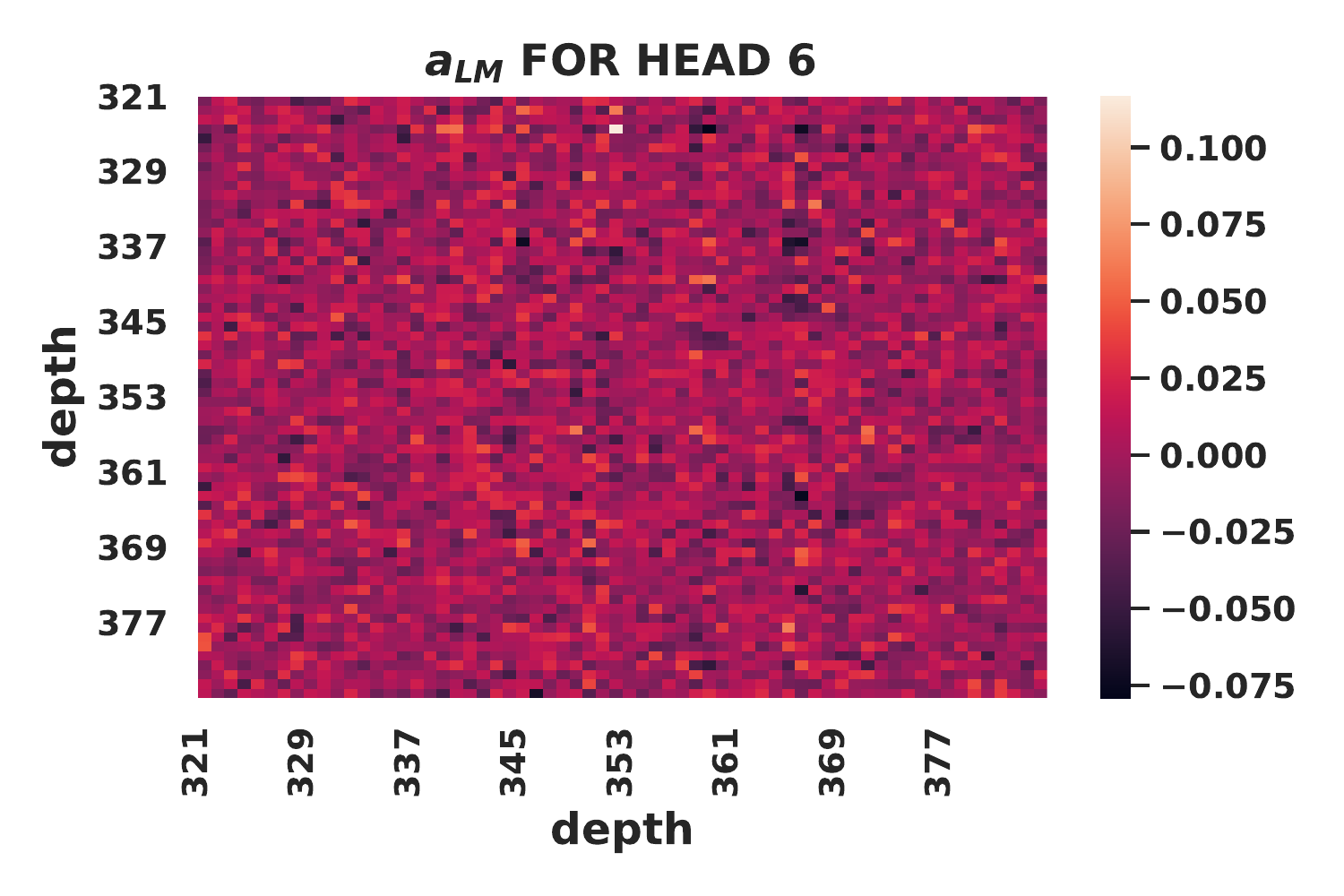}

\end{subfigure}
\hfill
\begin{subfigure}[b]{0.6\textwidth}
	\centering
	\includegraphics[width=1.1\textwidth]{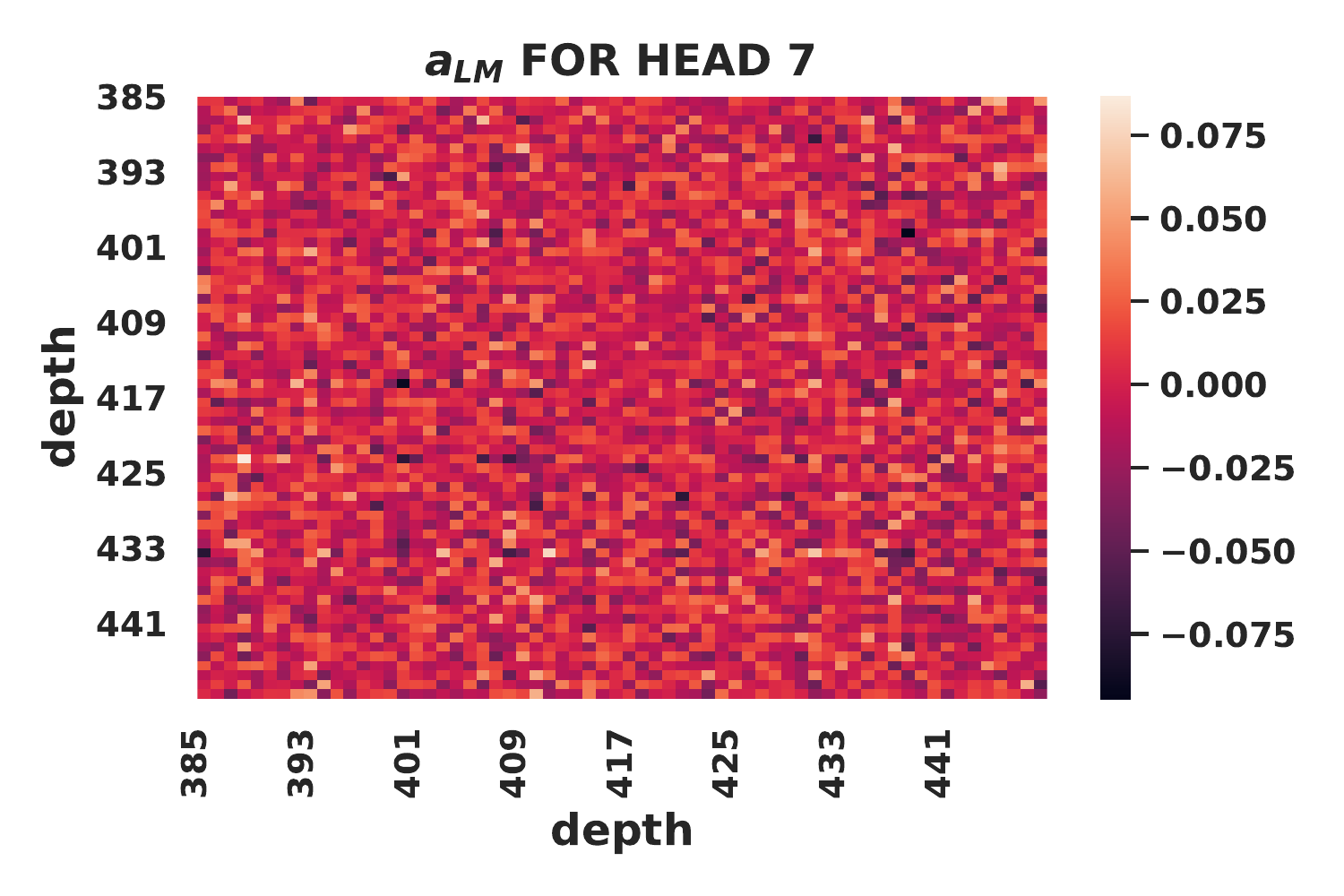}

\end{subfigure}
\hfill
\begin{subfigure}[b]{0.6\textwidth}
	\centering
	\includegraphics[width=1.1\textwidth]{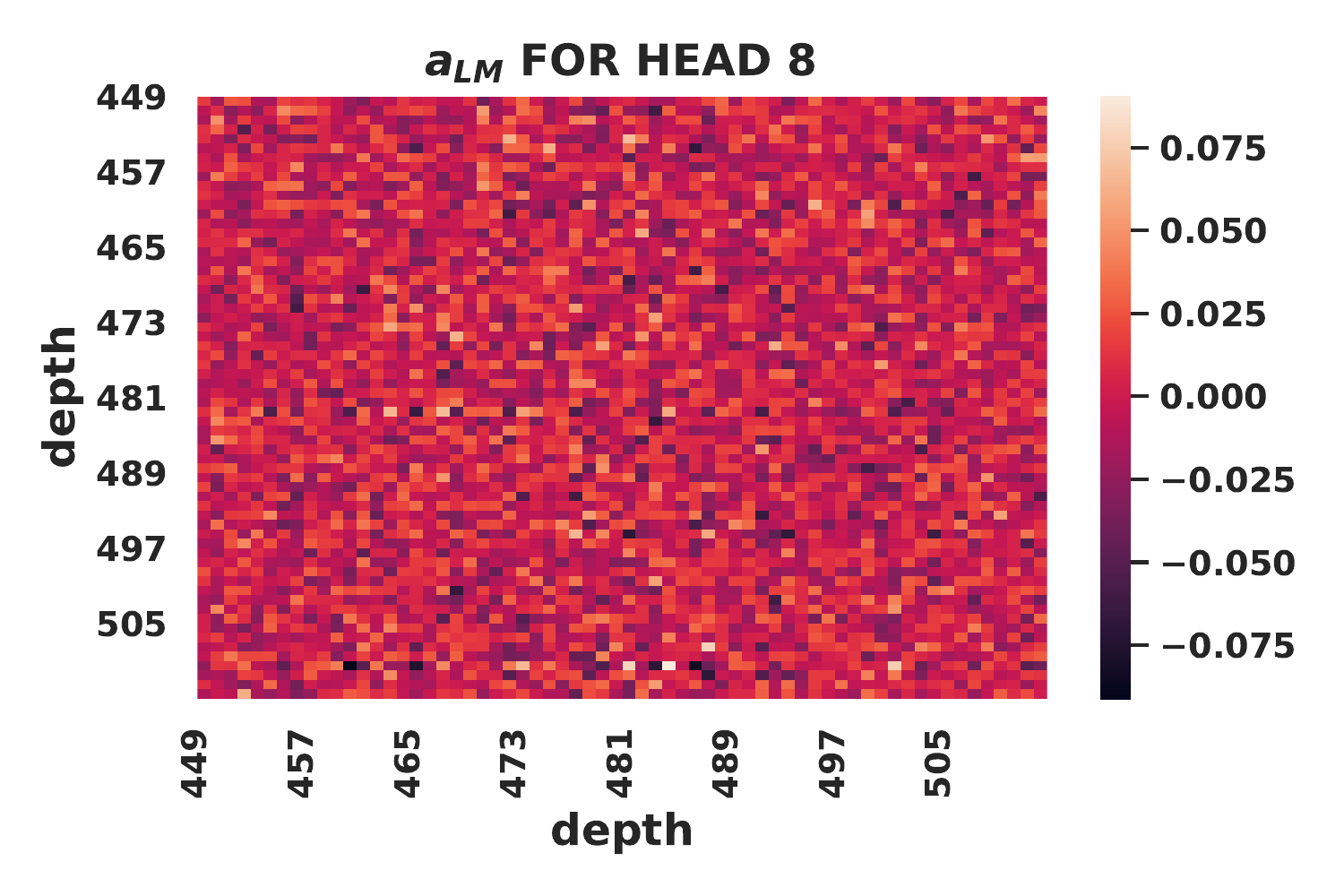}

\end{subfigure}
\caption{$\va_{LM}$ heatmap plots for all heads from SLM attention stage from Graph transformer model \#2 for PT-EN translation task.}
\label{fig15apx}
\end{adjustwidth}
\end{figure}    

\clearpage
\thispagestyle{headings}
\begin{figure}
\begin{adjustwidth}{-5em}{-5em}
\centering
\begin{subfigure}[b]{0.6\textwidth}
	\centering
	\includegraphics[width=1.1\textwidth]{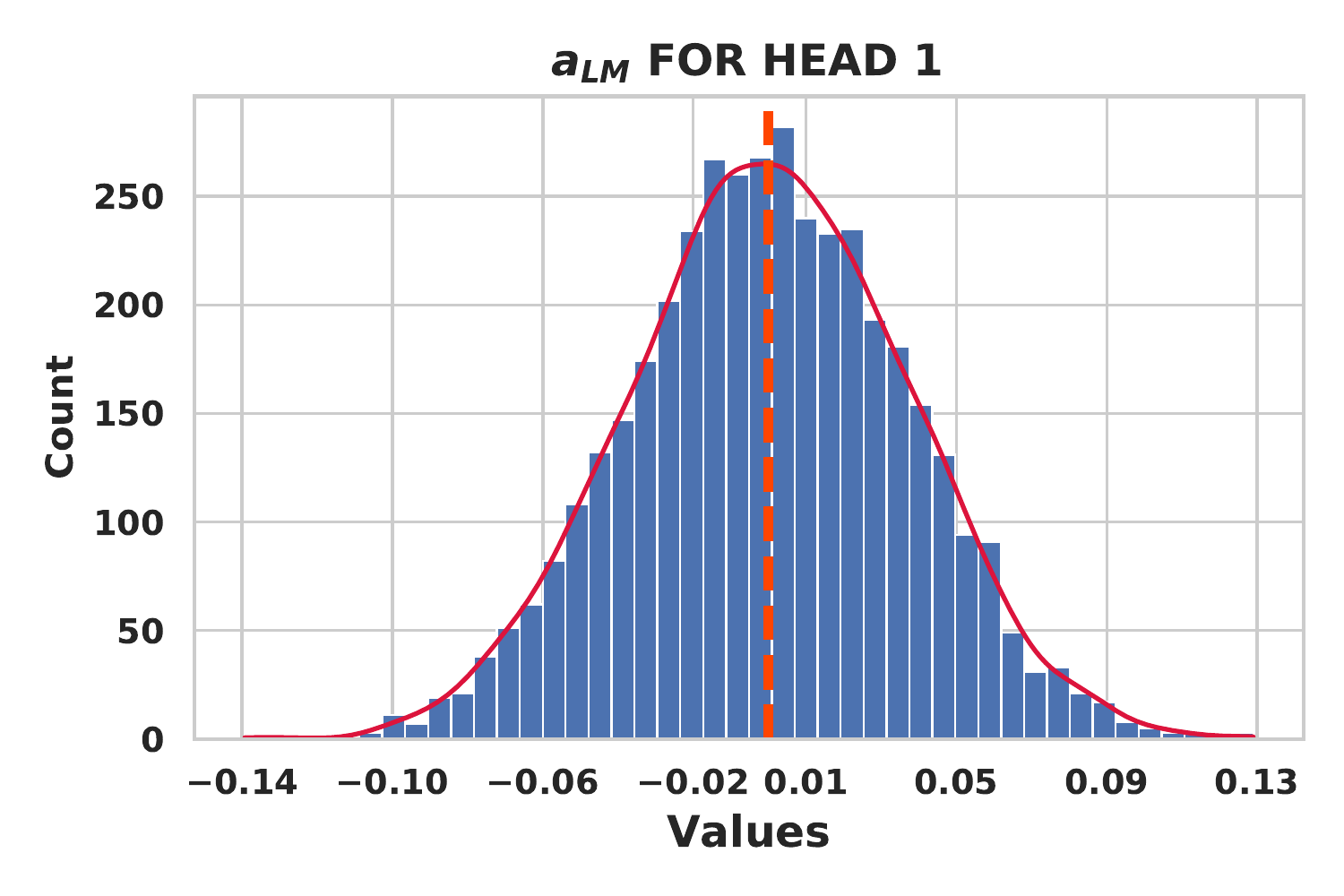}

\end{subfigure}
\hfill
\begin{subfigure}[b]{0.6\textwidth}
	\centering
	\includegraphics[width=1.1\textwidth]{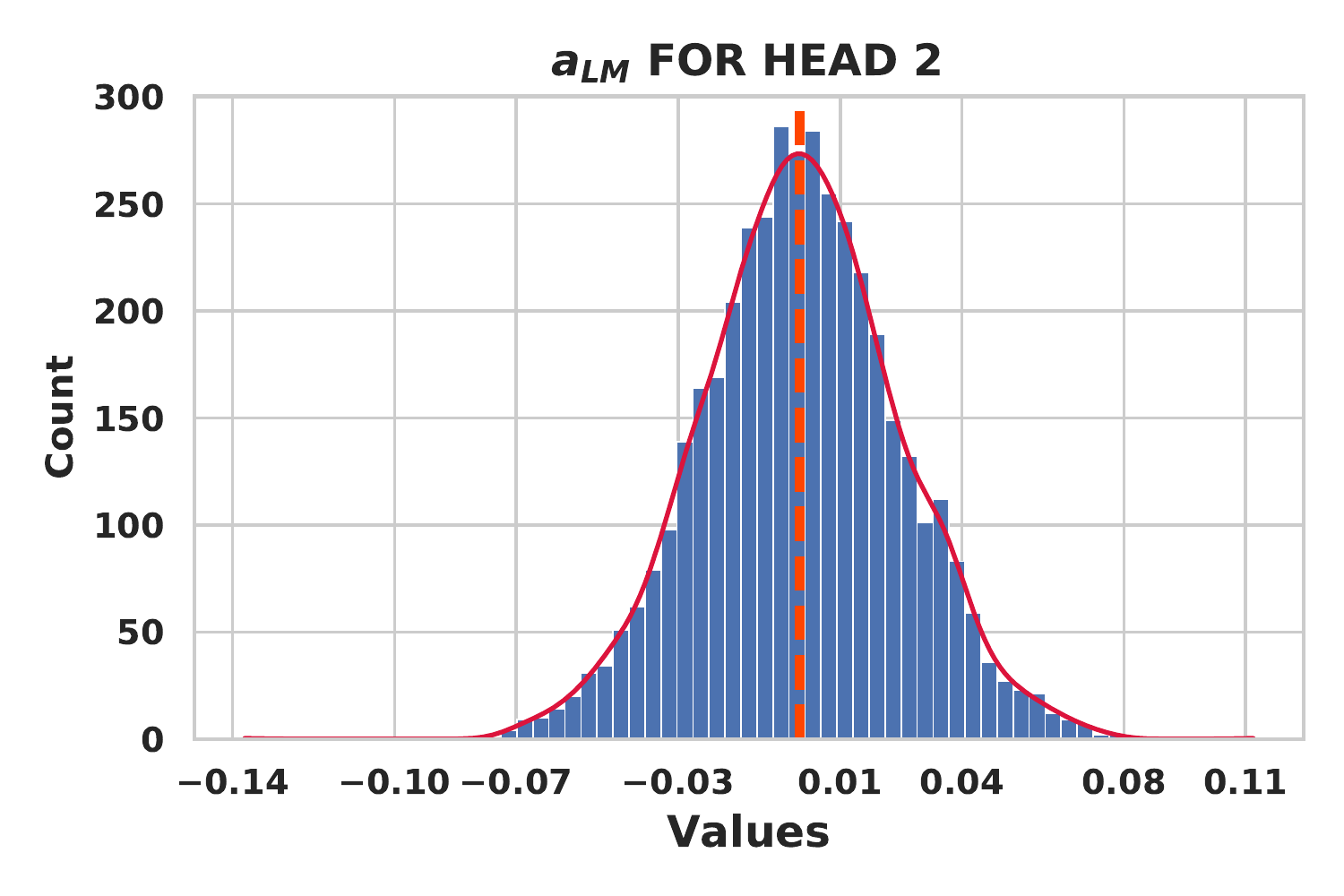}

\end{subfigure}
\hfill
\begin{subfigure}[b]{0.6\textwidth}
	\centering
	\includegraphics[width=1.1\textwidth]{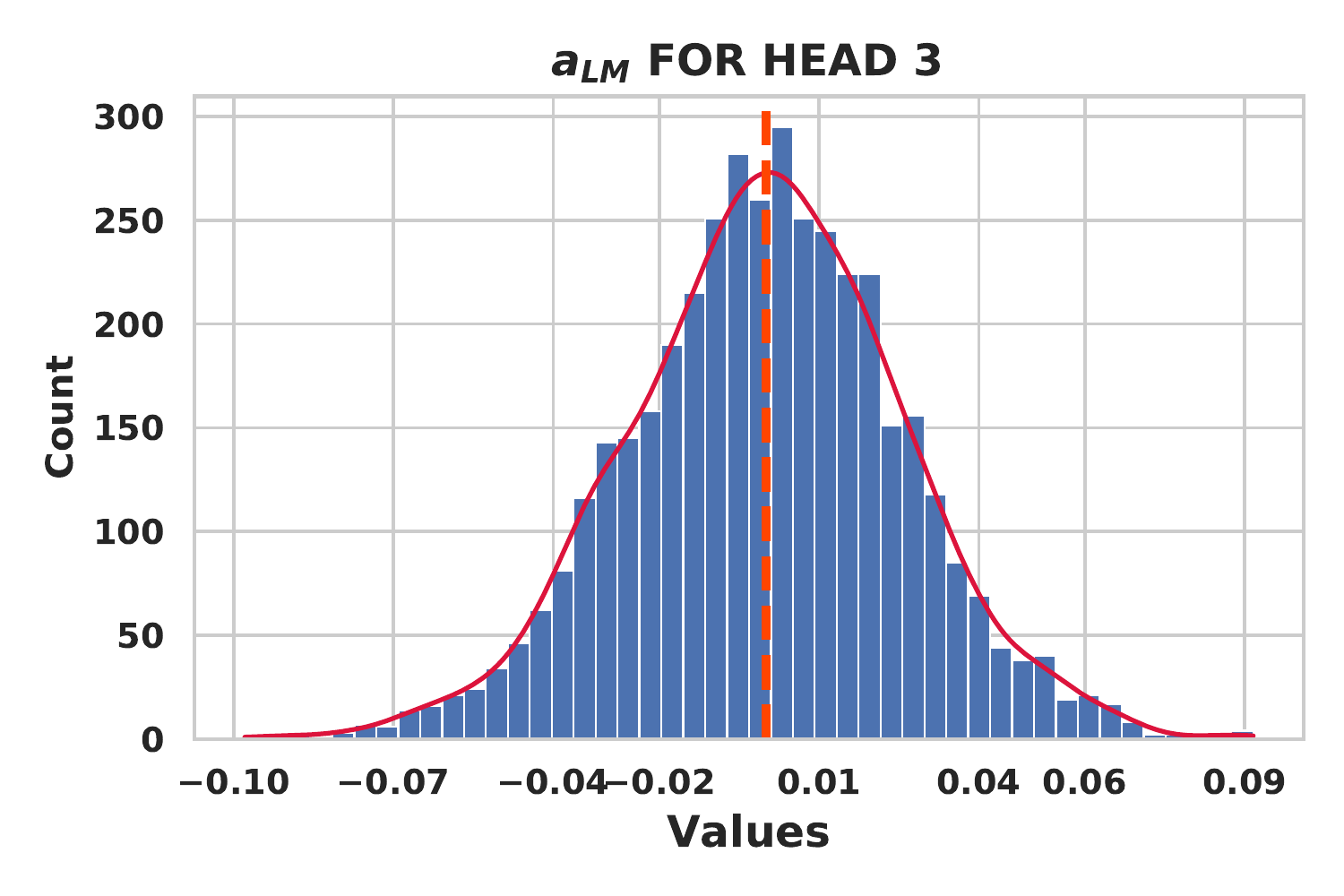}

\end{subfigure}
\hfill
\begin{subfigure}[b]{0.6\textwidth}
	\centering
	\includegraphics[width=1.1\textwidth]{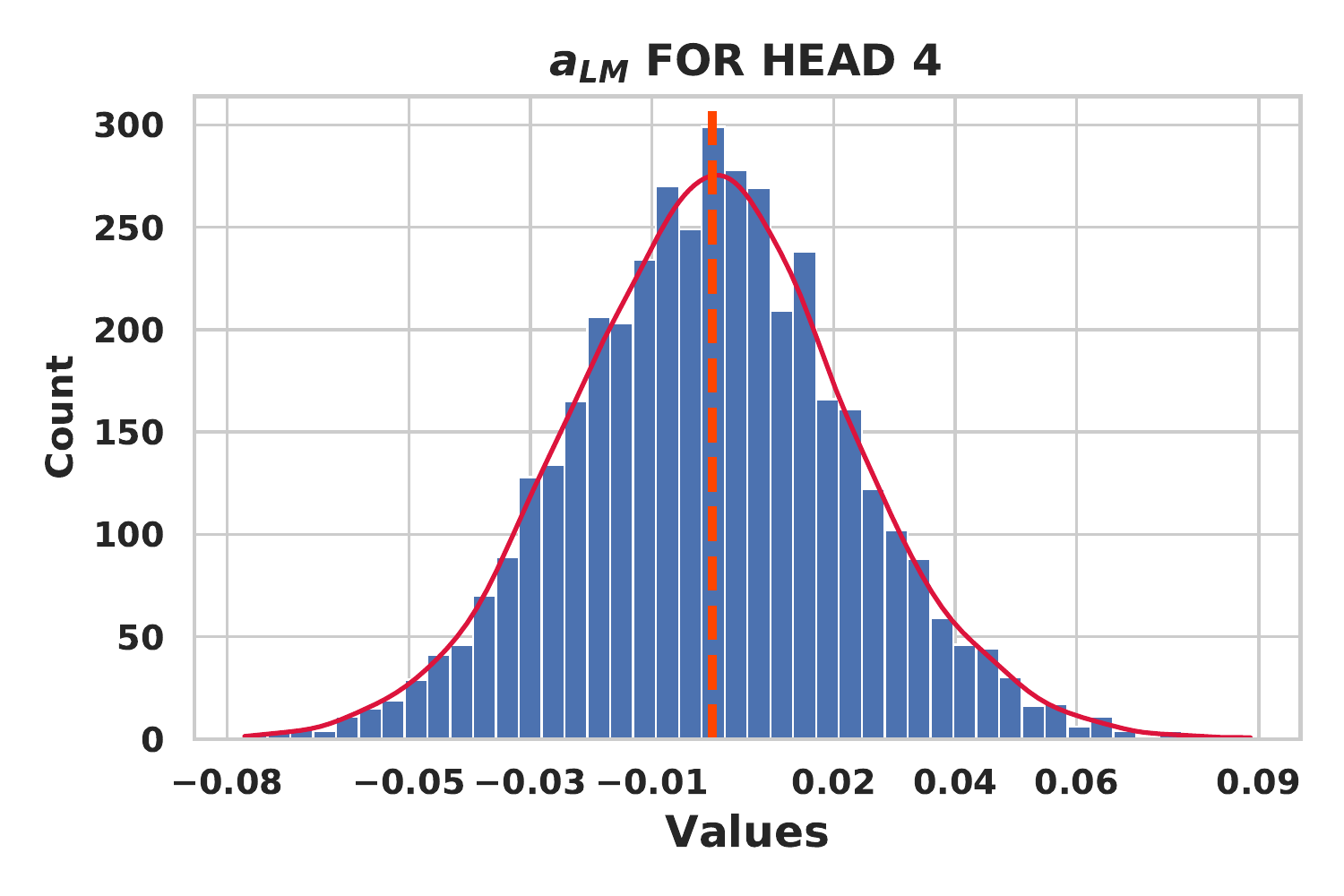}

\end{subfigure}
\centering
\begin{subfigure}[b]{0.6\textwidth}
	\centering
	\includegraphics[width=1.1\textwidth]{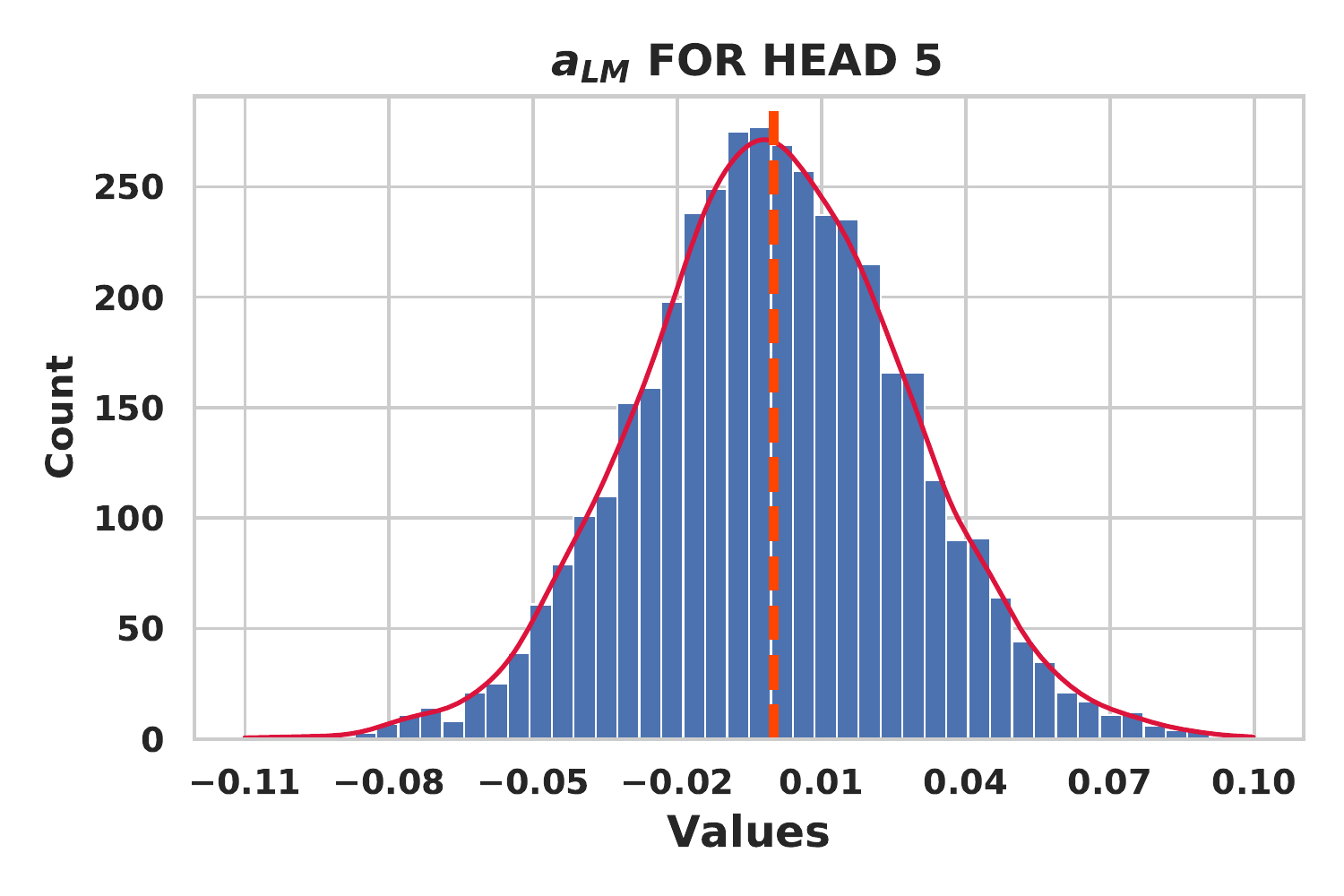}

\end{subfigure}
\hfill
\begin{subfigure}[b]{0.6\textwidth}
	\centering
	\includegraphics[width=1.1\textwidth]{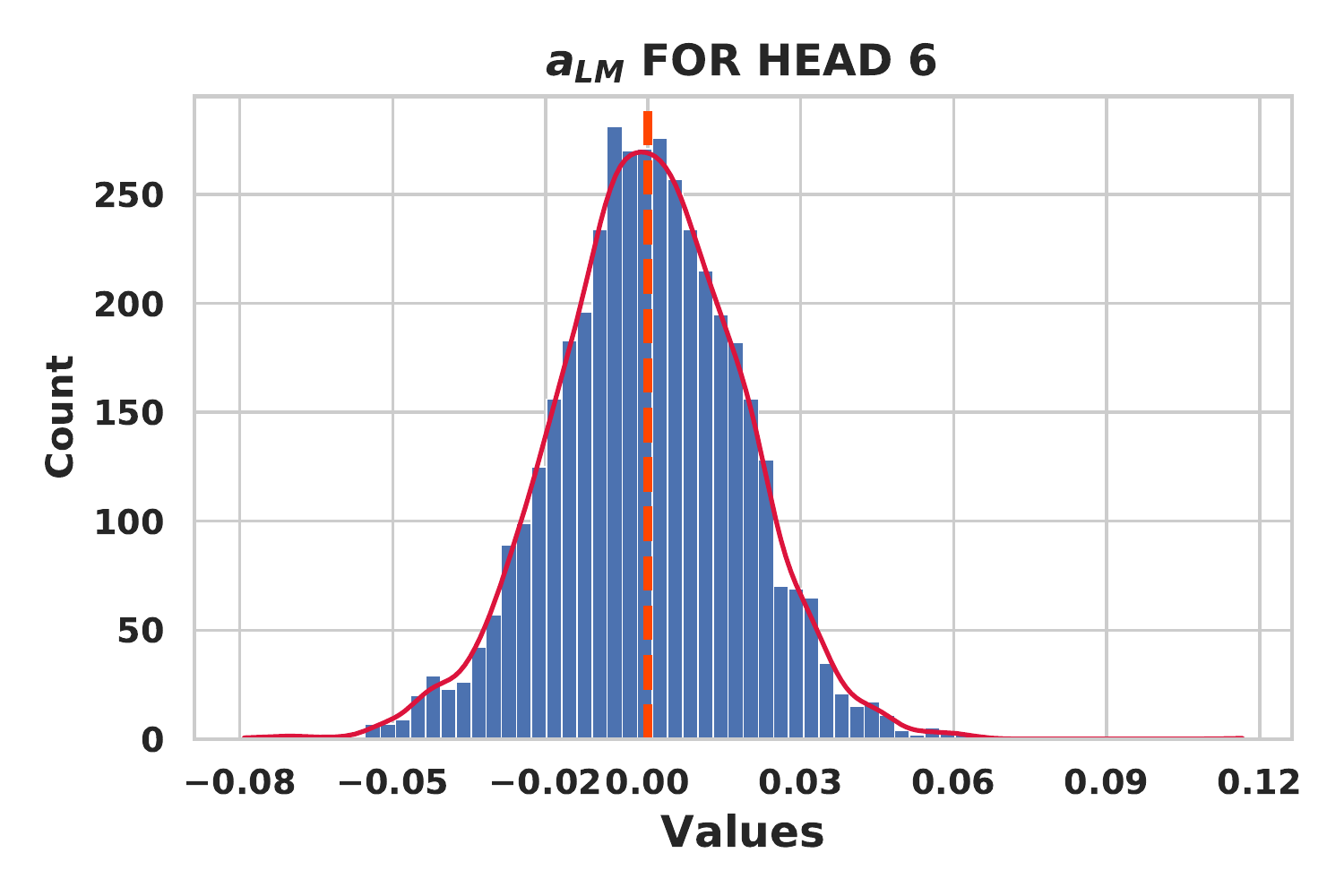}

\end{subfigure}
\hfill
\begin{subfigure}[b]{0.6\textwidth}
	\centering
	\includegraphics[width=1.1\textwidth]{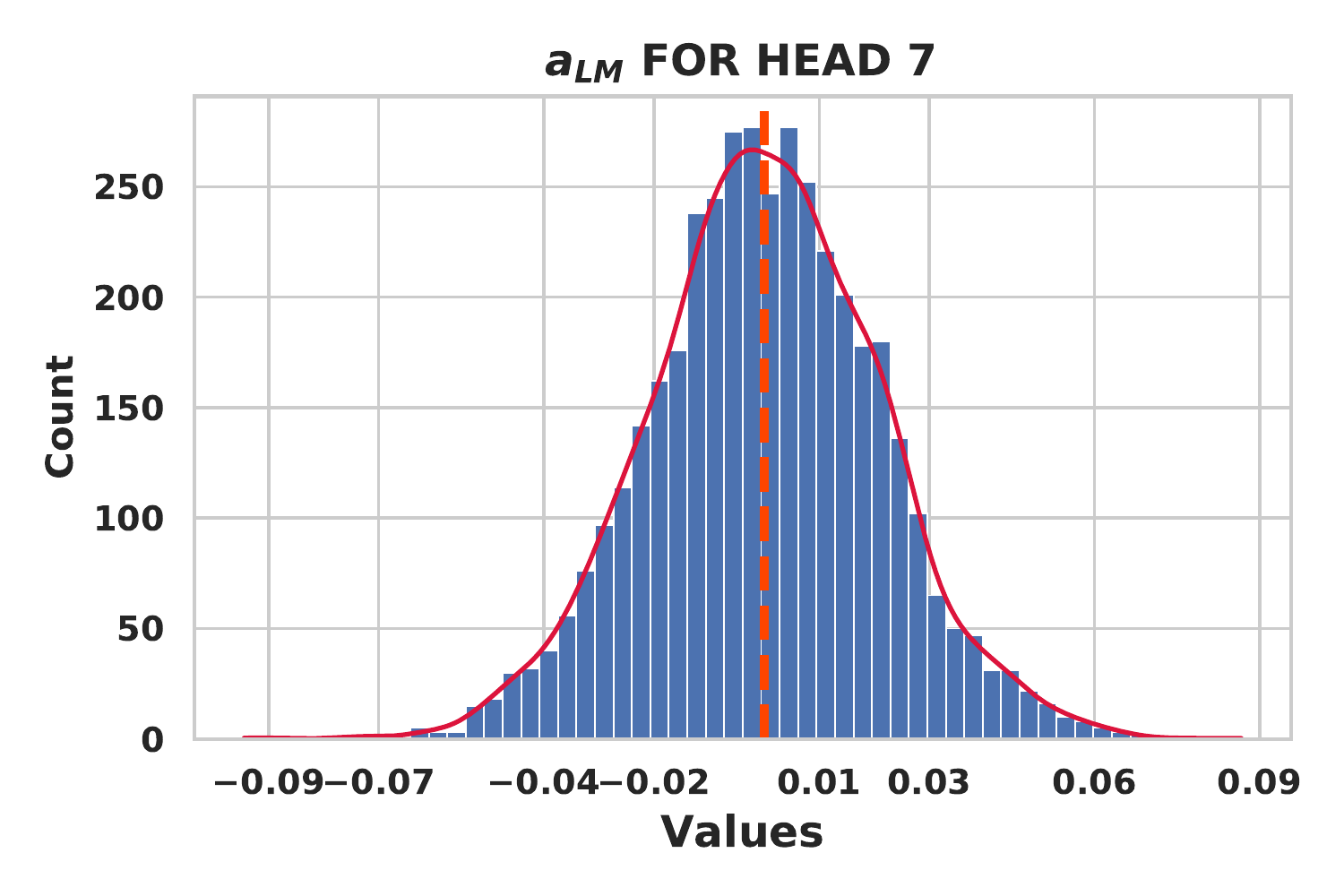}

\end{subfigure}
\hfill
\begin{subfigure}[b]{0.6\textwidth}
	\centering
	\includegraphics[width=1.1\textwidth]{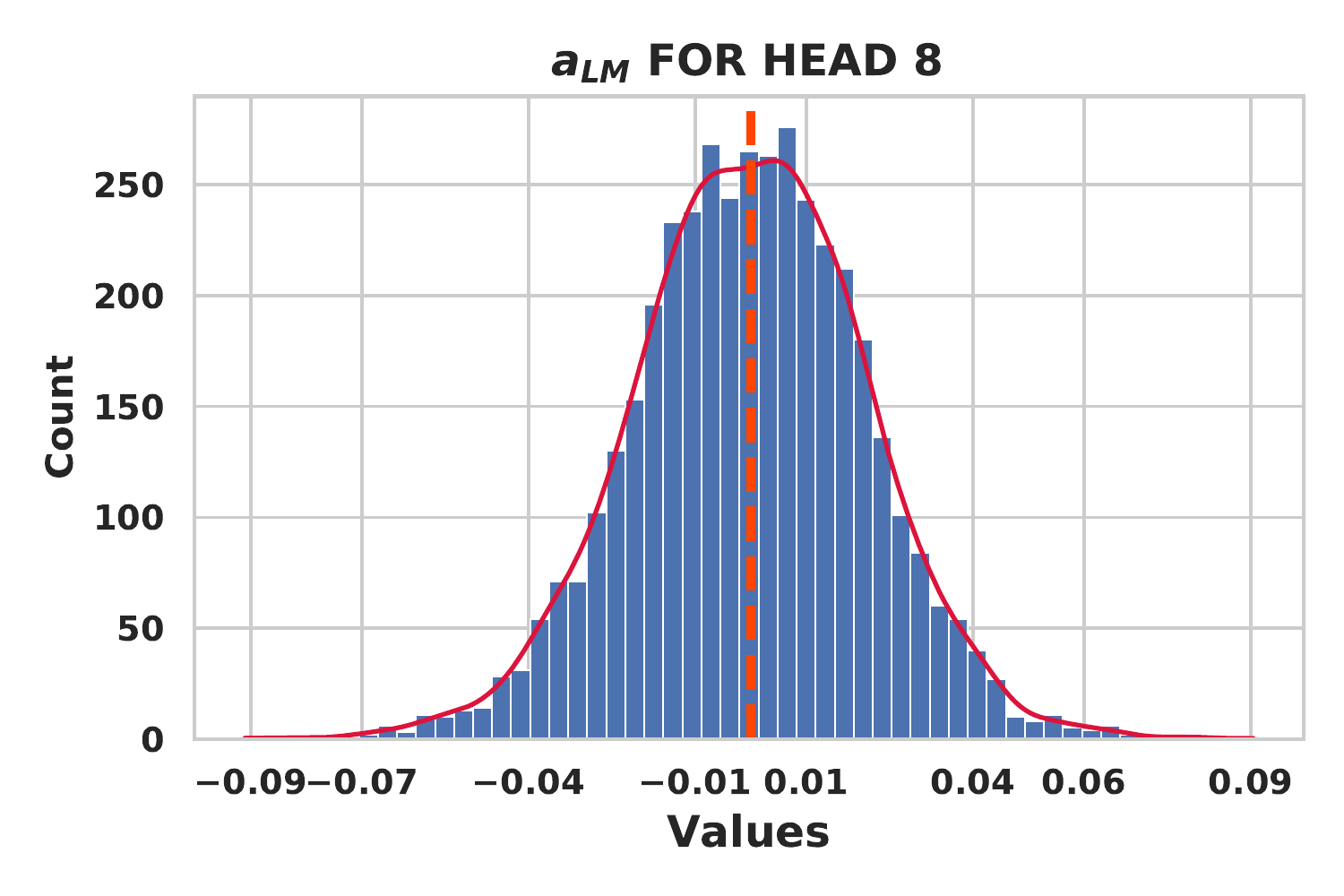}

\end{subfigure}
\caption{$\va_{LM}$ histogram plots for all heads from SLM attention stage from Graph transformer model \#2 for PT-EN translation task. Dashed line in orange marks zero value.}
\label{fig16apx}
\end{adjustwidth}
\end{figure}  

\clearpage
\thispagestyle{headings}
\begin{figure}
\begin{adjustwidth}{-5em}{-5em}
\centering
\begin{subfigure}[b]{0.6\textwidth}
	\centering
	\includegraphics[width=1.1\textwidth]{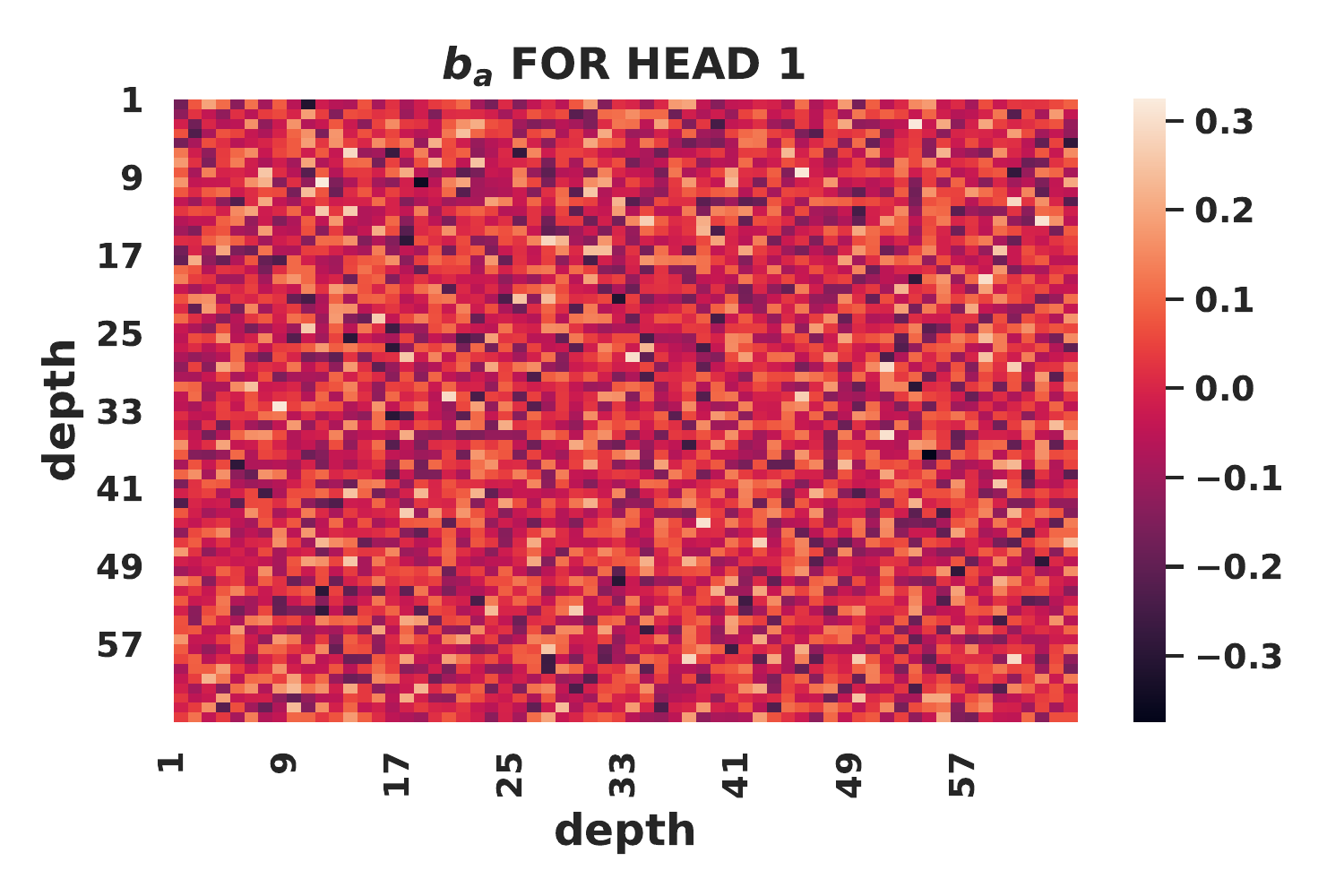}

\end{subfigure}
\hfill
\begin{subfigure}[b]{0.6\textwidth}
	\centering
	\includegraphics[width=1.1\textwidth]{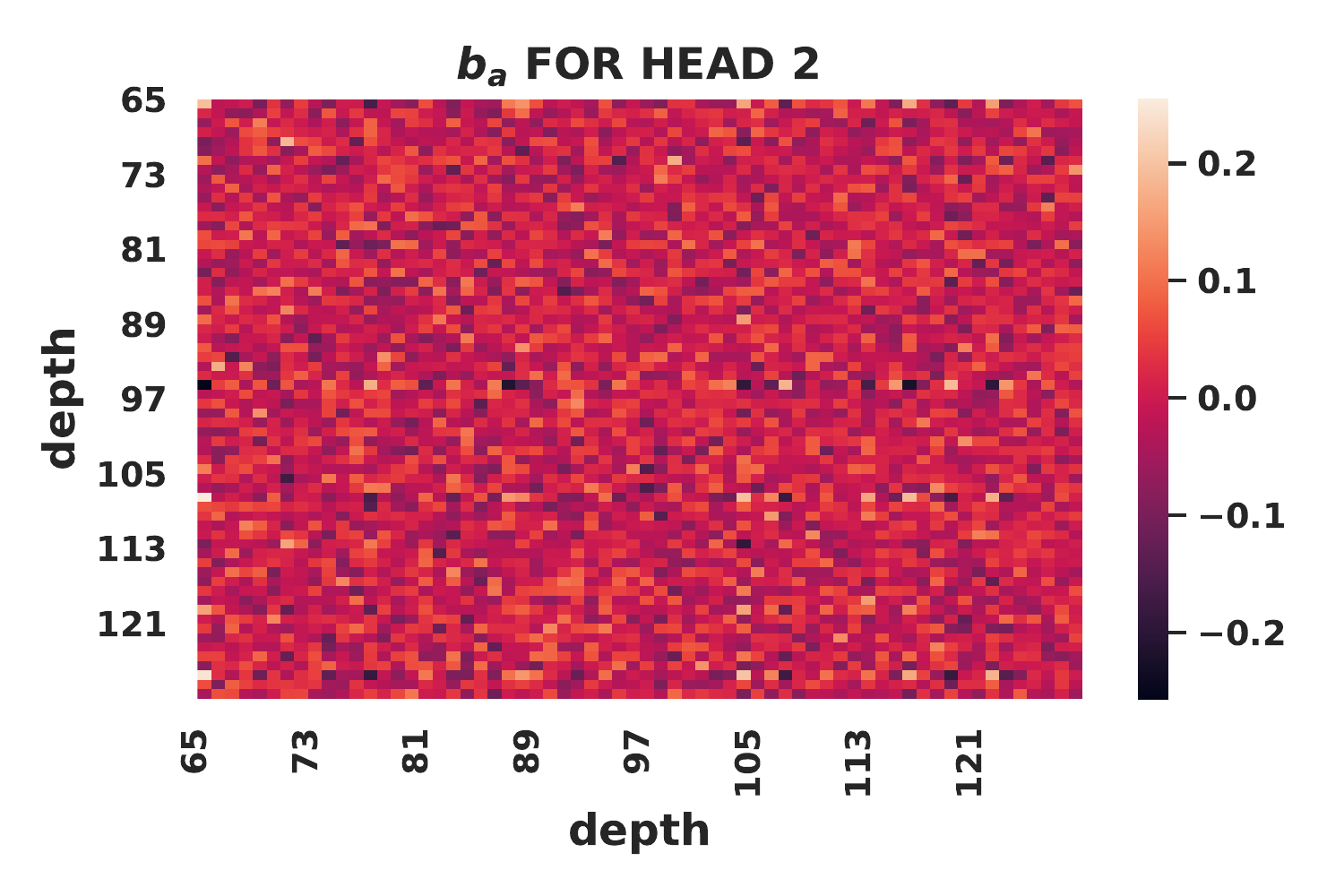}

\end{subfigure}
\hfill
\begin{subfigure}[b]{0.6\textwidth}
	\centering
	\includegraphics[width=1.1\textwidth]{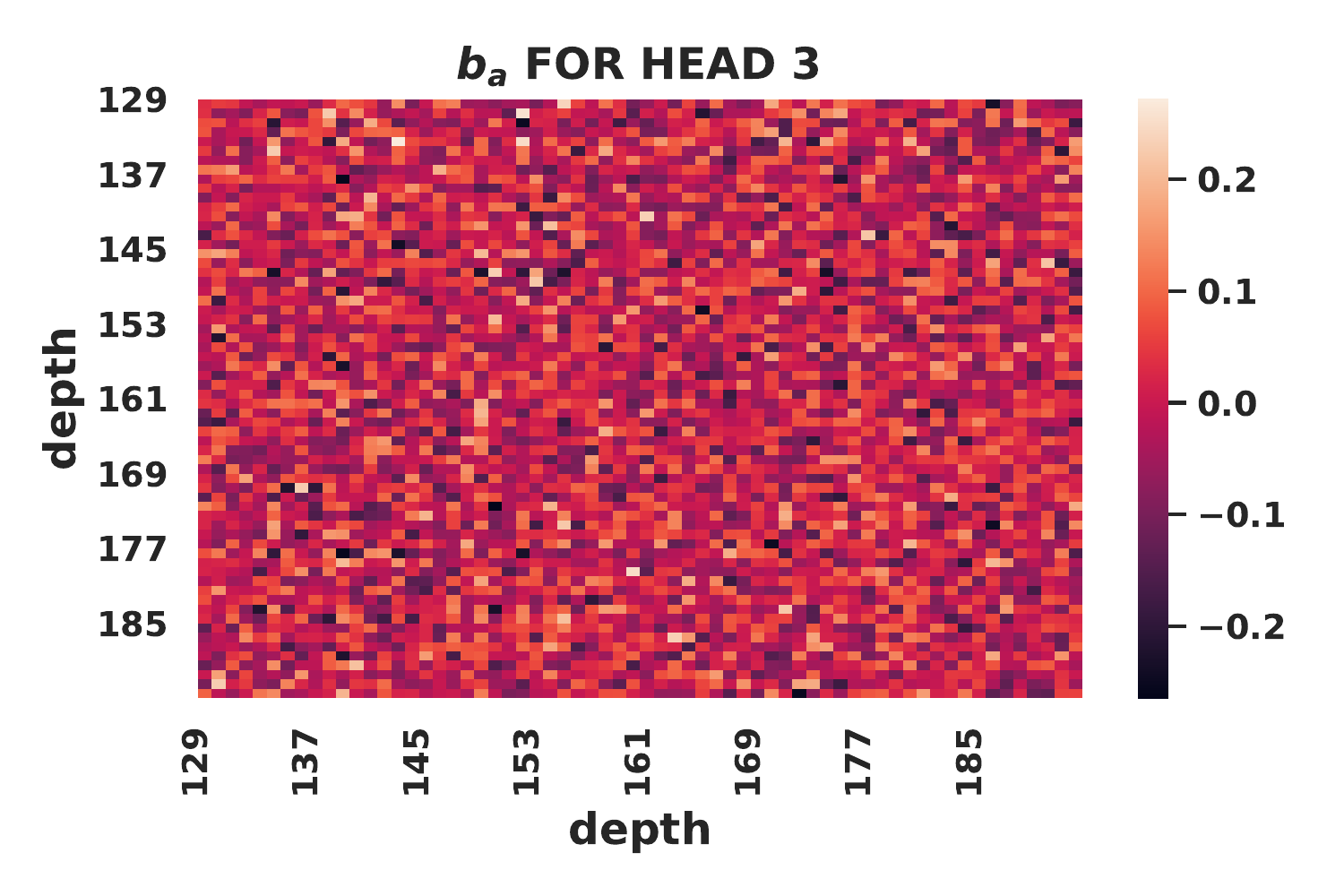}

\end{subfigure}
\hfill
\begin{subfigure}[b]{0.6\textwidth}
	\centering
	\includegraphics[width=1.1\textwidth]{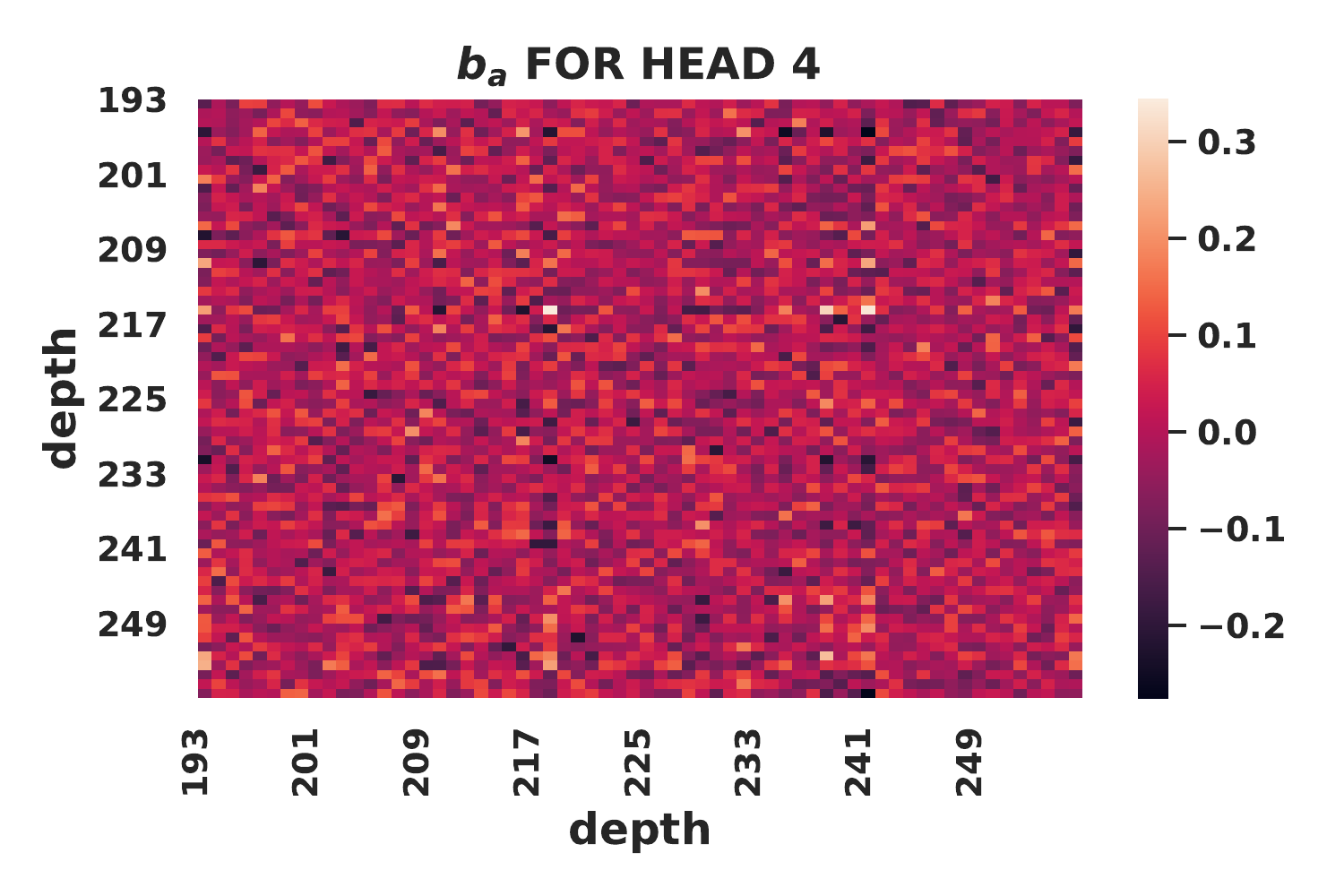}

\end{subfigure}
\centering
\begin{subfigure}[b]{0.6\textwidth}
	\centering
	\includegraphics[width=1.1\textwidth]{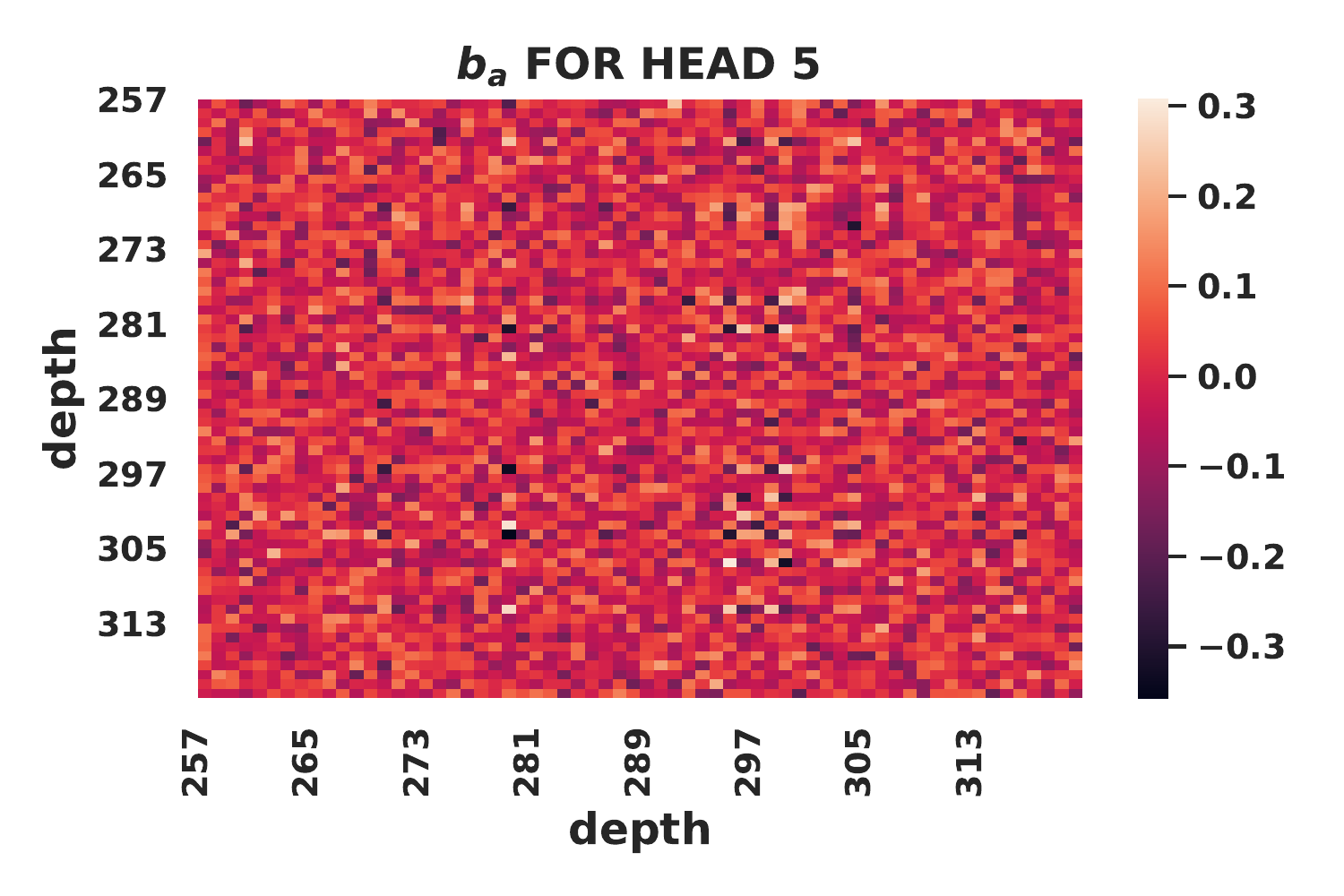}

\end{subfigure}
\hfill
\begin{subfigure}[b]{0.6\textwidth}
	\centering
	\includegraphics[width=1.1\textwidth]{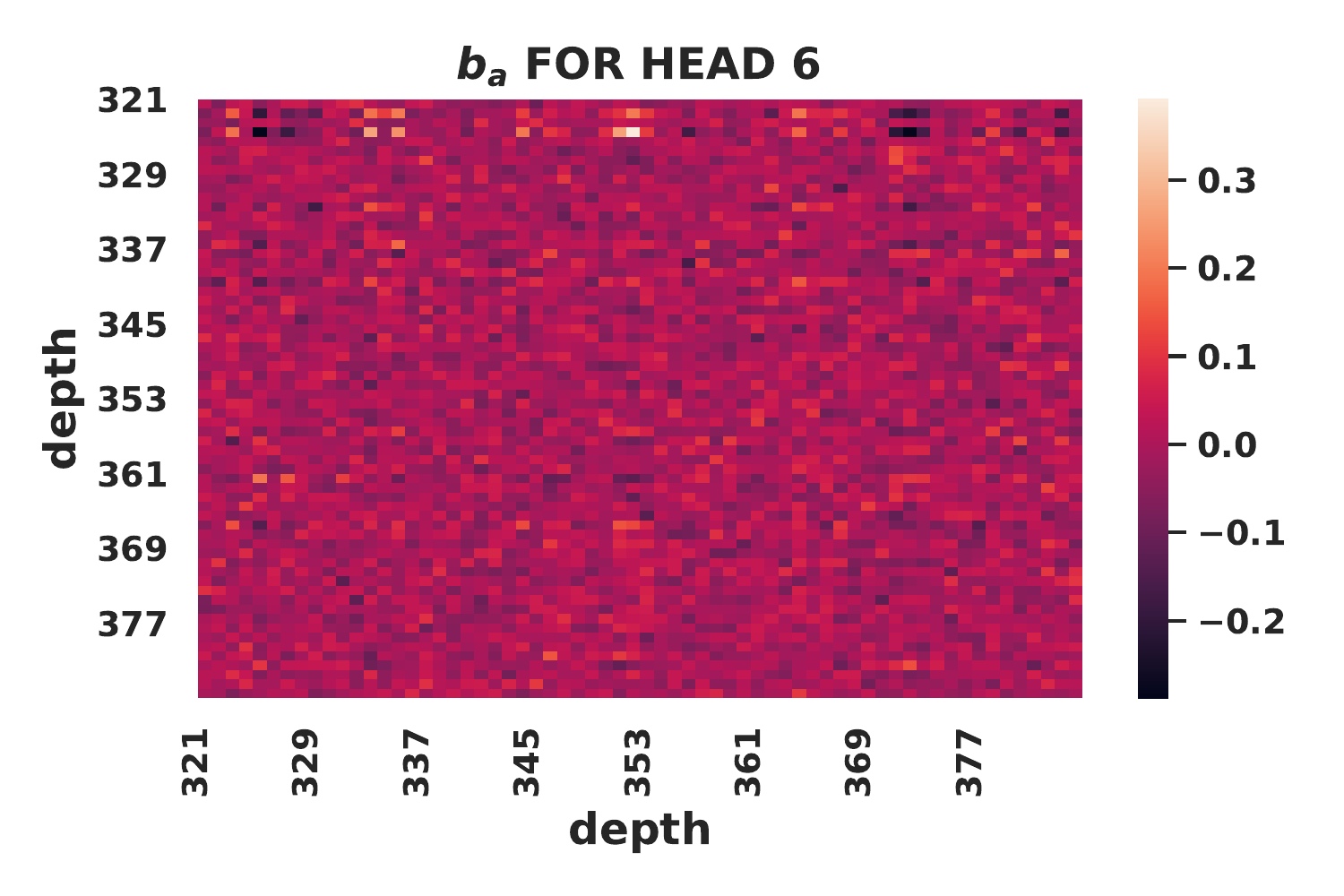}

\end{subfigure}
\hfill
\begin{subfigure}[b]{0.6\textwidth}
	\centering
	\includegraphics[width=1.1\textwidth]{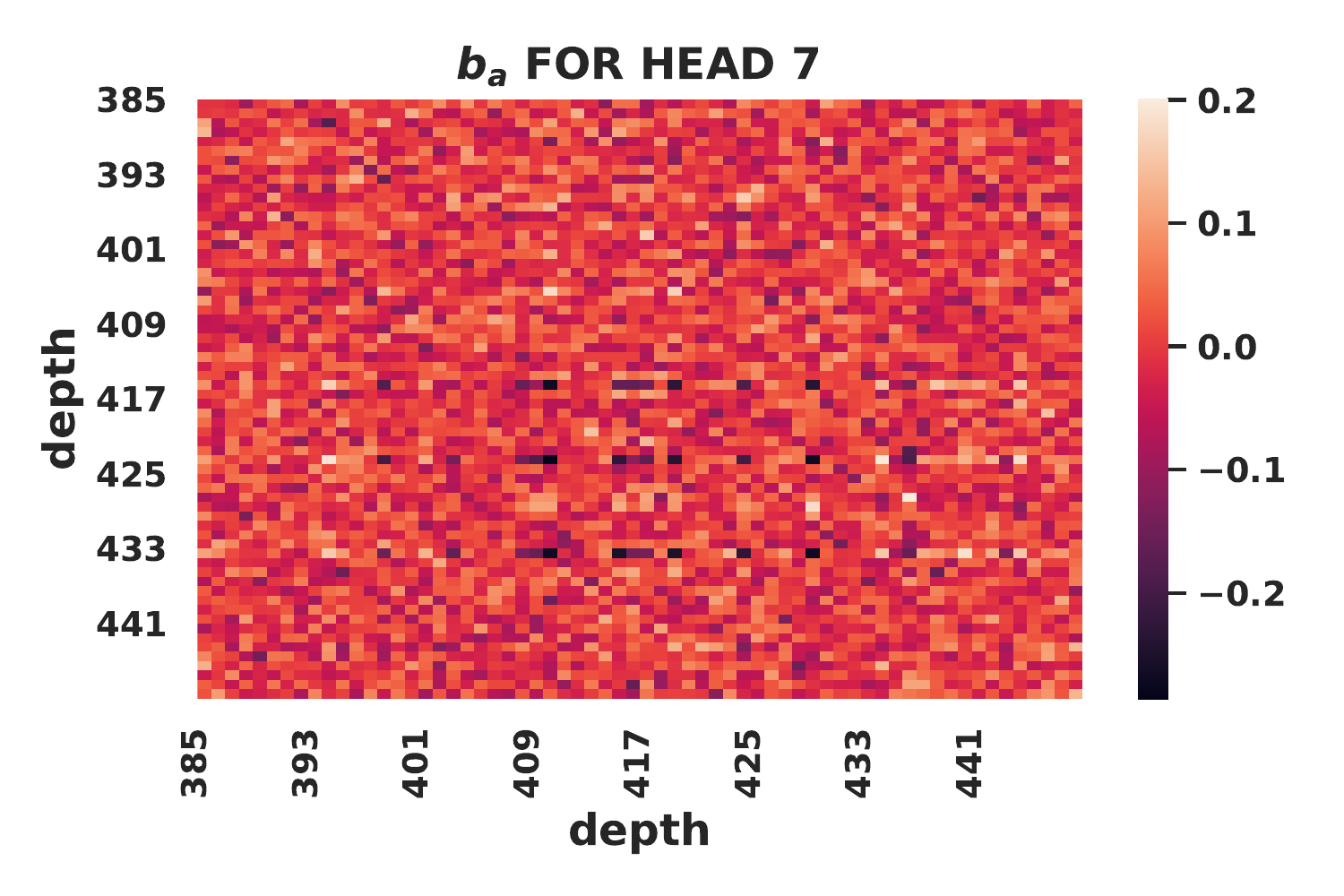}

\end{subfigure}
\hfill
\begin{subfigure}[b]{0.6\textwidth}
	\centering
	\includegraphics[width=1.1\textwidth]{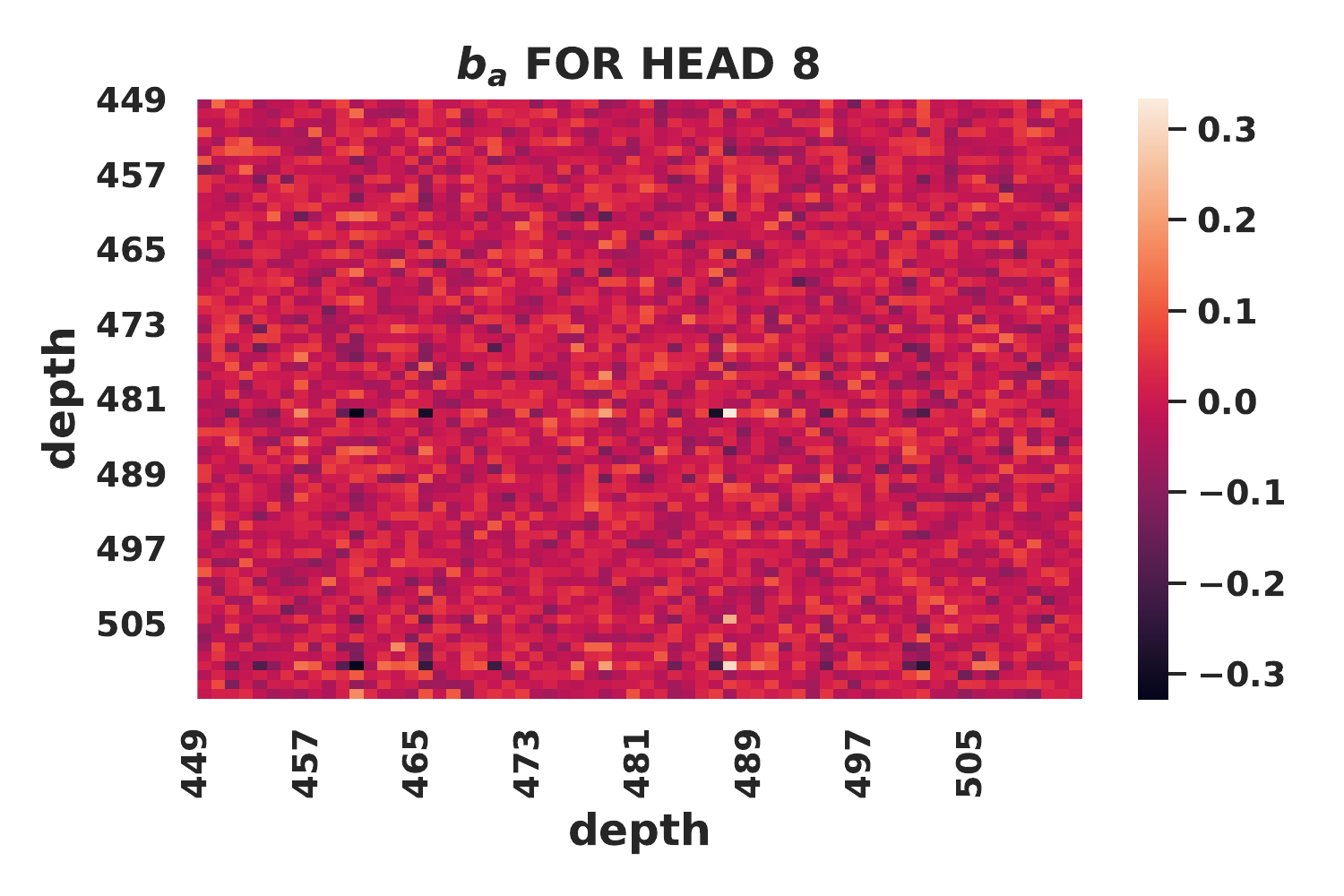}

\end{subfigure}
\caption{$\vb_{a}$ heatmap plots for all heads from SLM attention stage from graph transformer model \#2 for PT-EN translation task.}
\label{fig17apx}
\end{adjustwidth}
\end{figure}

\clearpage
\thispagestyle{headings}
\begin{figure}
\begin{adjustwidth}{-5em}{-5em}
\centering
\begin{subfigure}[b]{0.6\textwidth}
	\centering
	\includegraphics[width=1.1\textwidth]{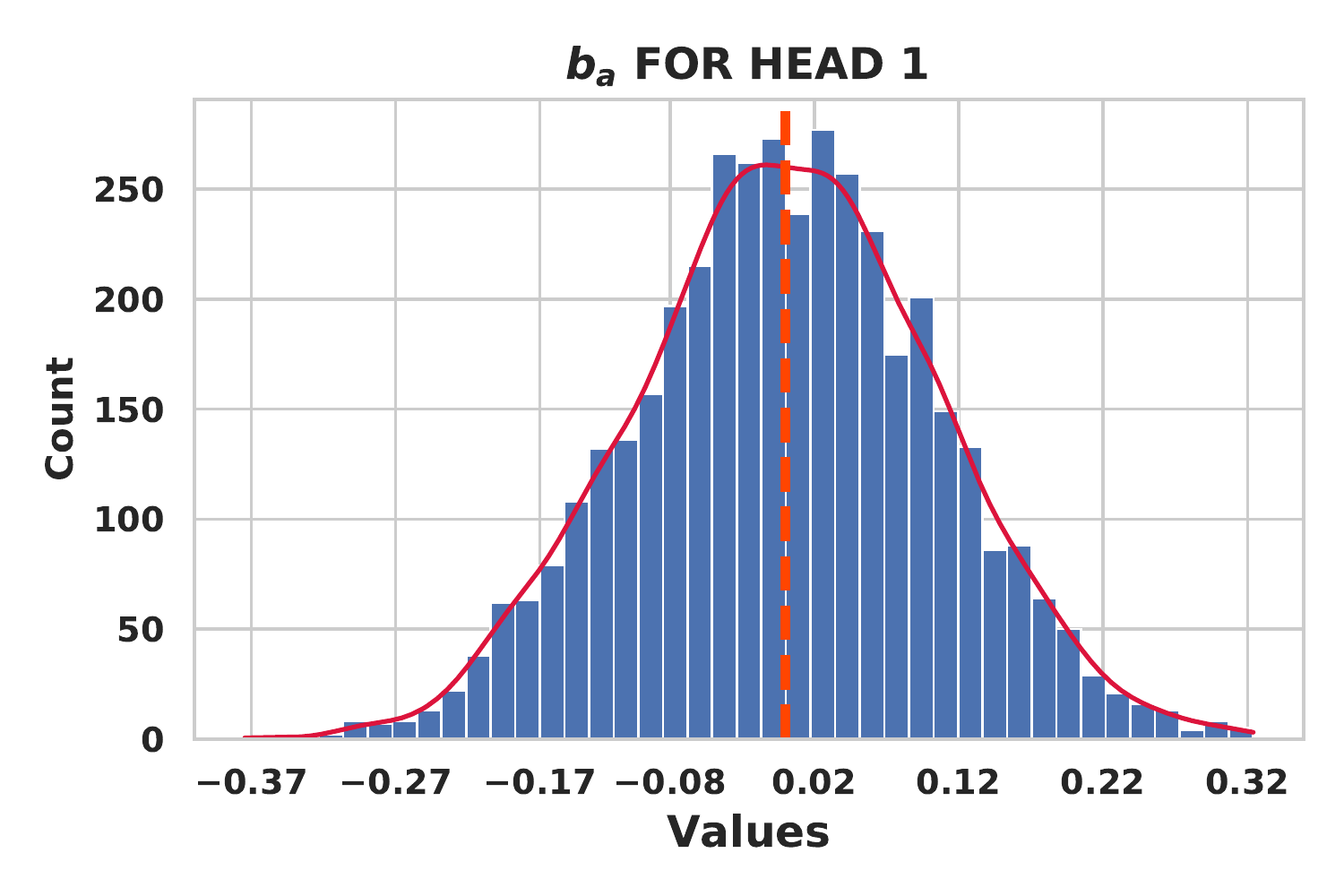}

\end{subfigure}
\hfill
\begin{subfigure}[b]{0.6\textwidth}
	\centering
	\includegraphics[width=1.1\textwidth]{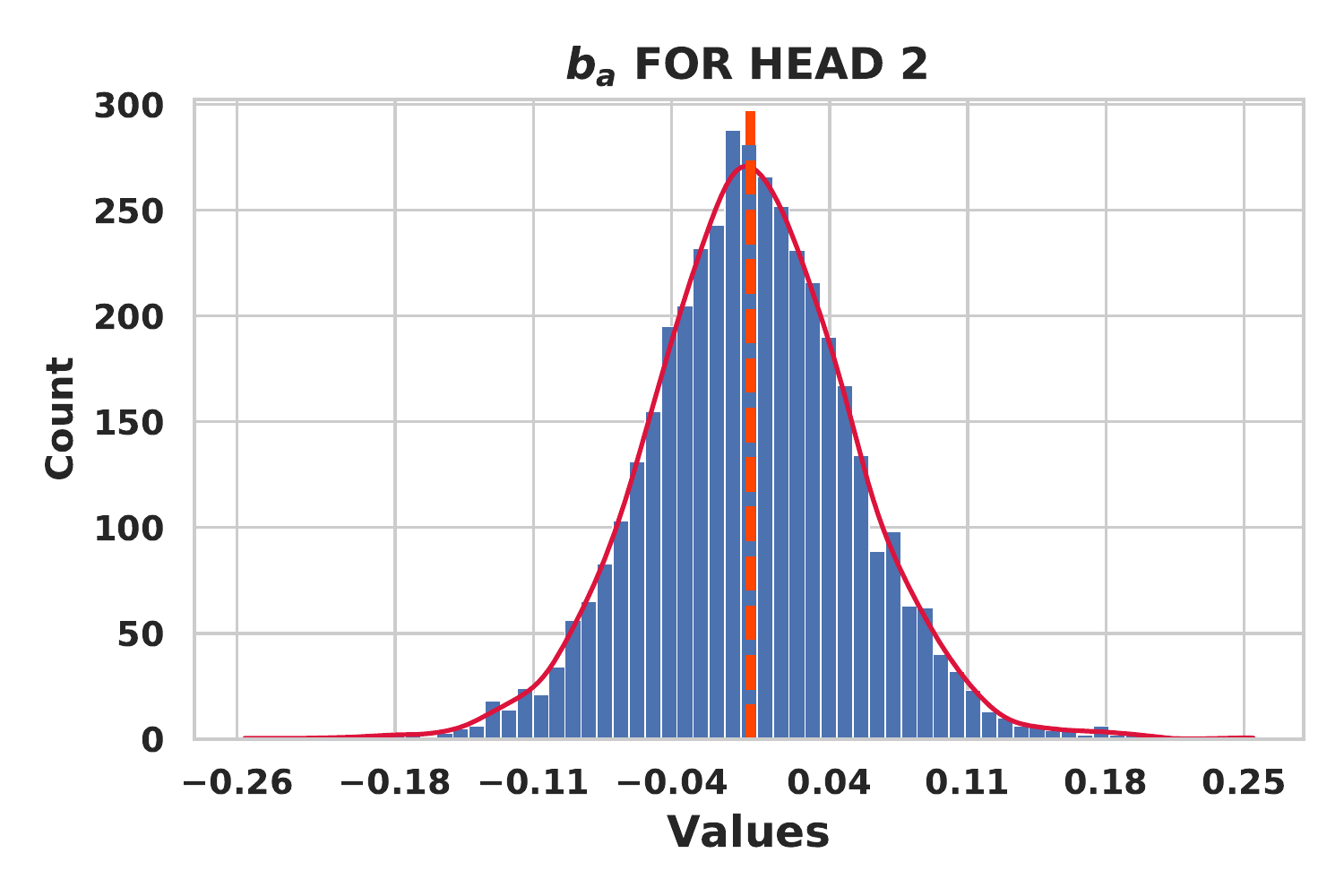}

\end{subfigure}
\hfill
\begin{subfigure}[b]{0.6\textwidth}
	\centering
	\includegraphics[width=1.1\textwidth]{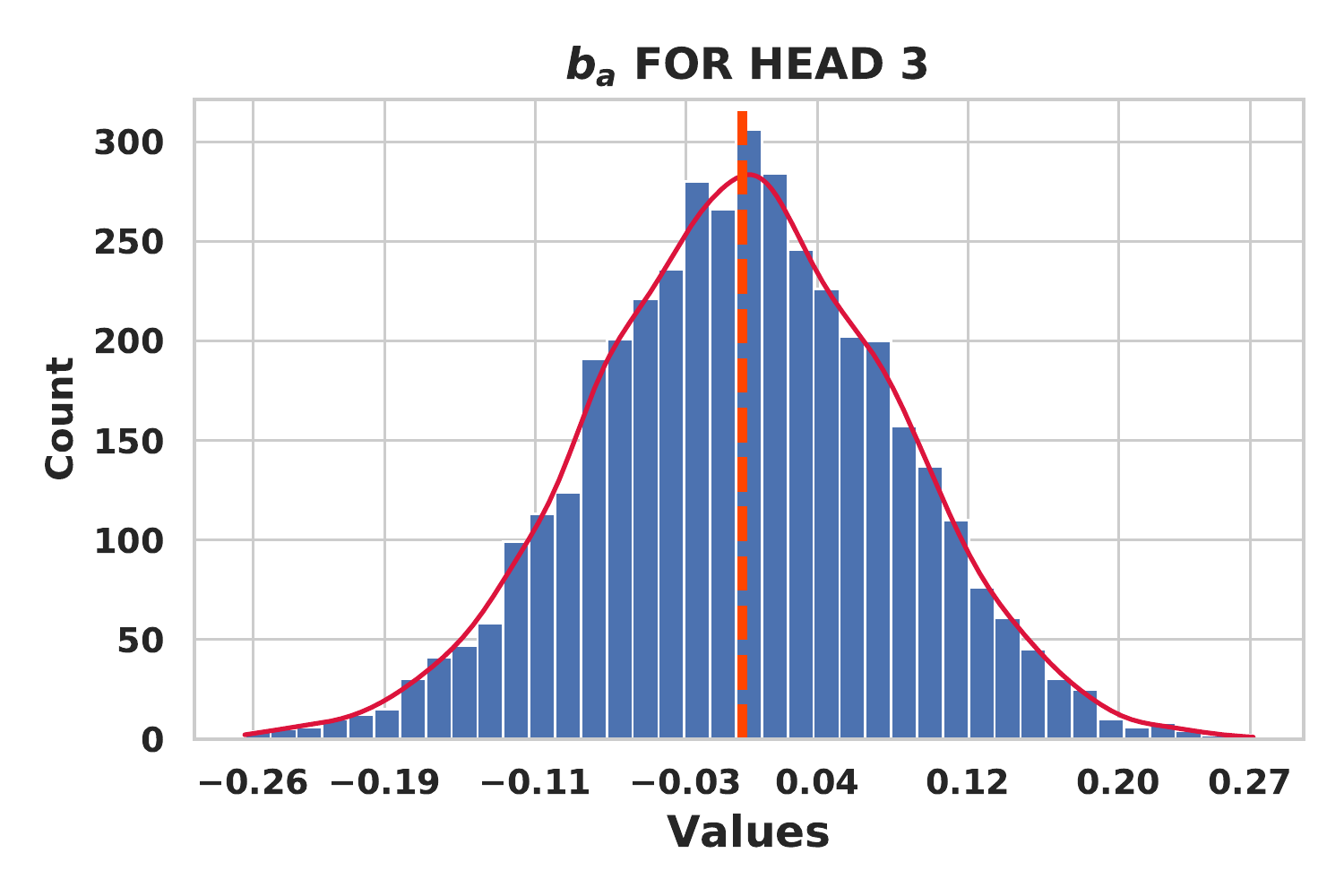}

\end{subfigure}
\hfill
\begin{subfigure}[b]{0.6\textwidth}
	\centering
	\includegraphics[width=1.1\textwidth]{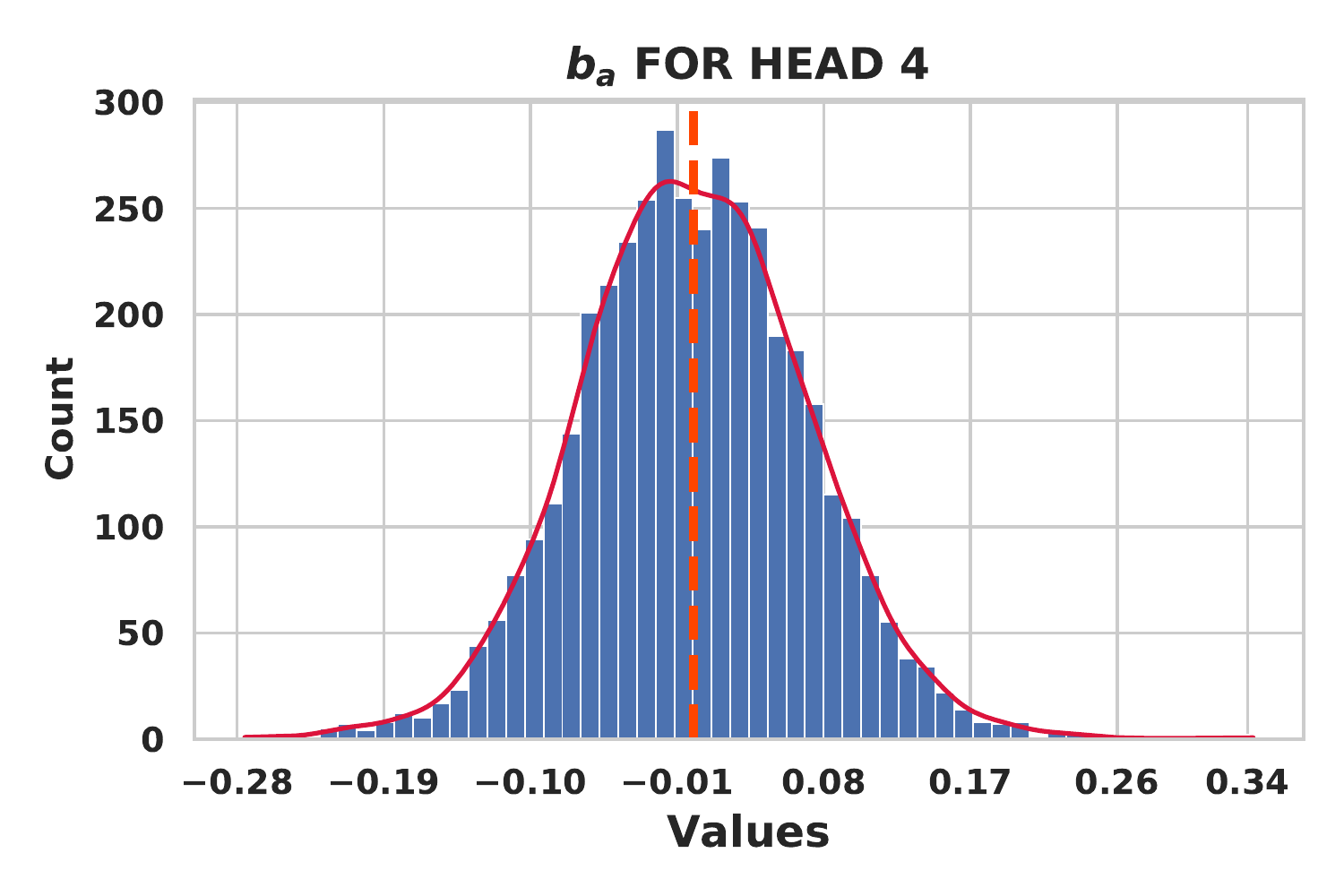}

\end{subfigure}
\centering
\begin{subfigure}[b]{0.6\textwidth}
	\centering
	\includegraphics[width=1.1\textwidth]{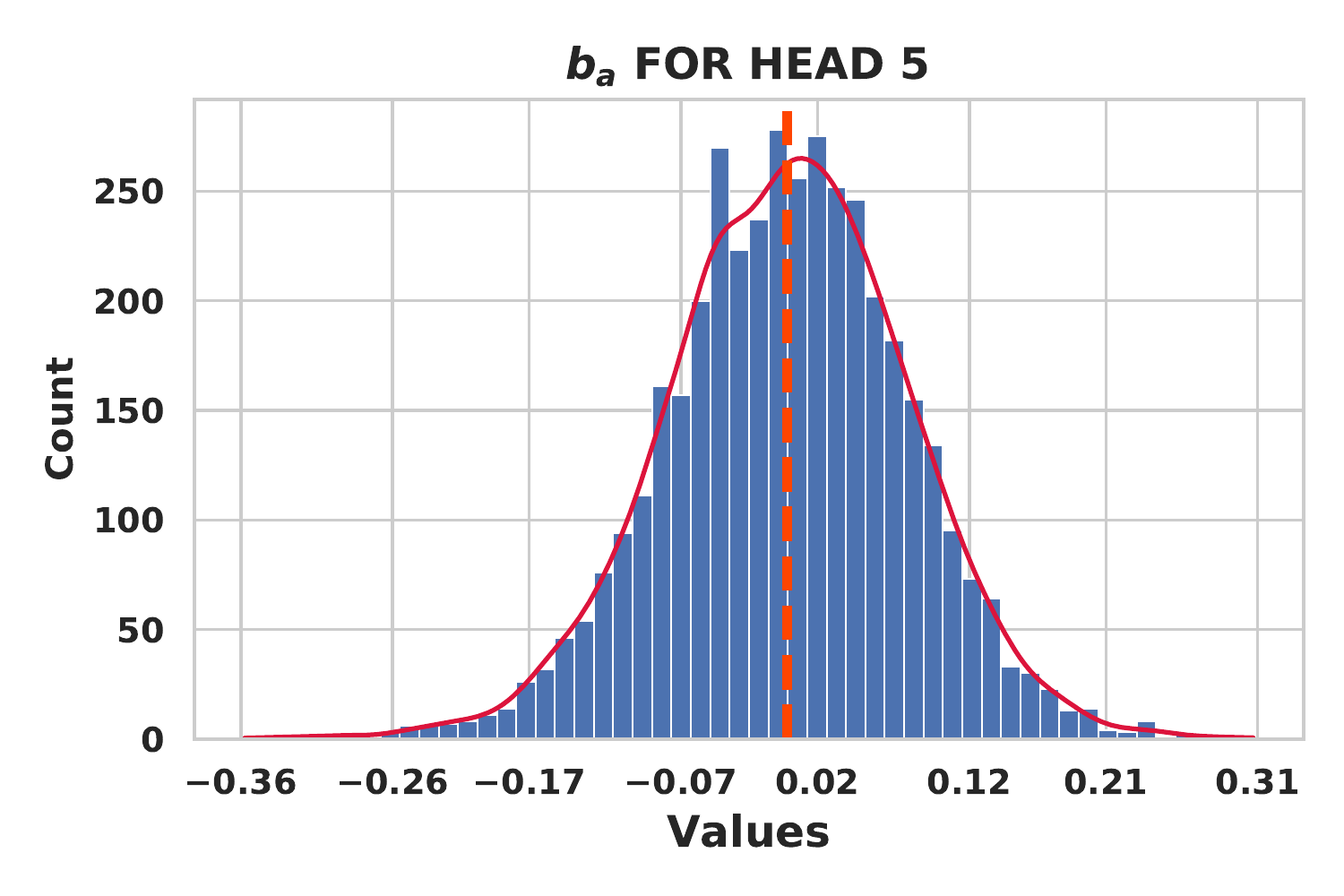}

\end{subfigure}
\hfill
\begin{subfigure}[b]{0.6\textwidth}
	\centering
	\includegraphics[width=1.1\textwidth]{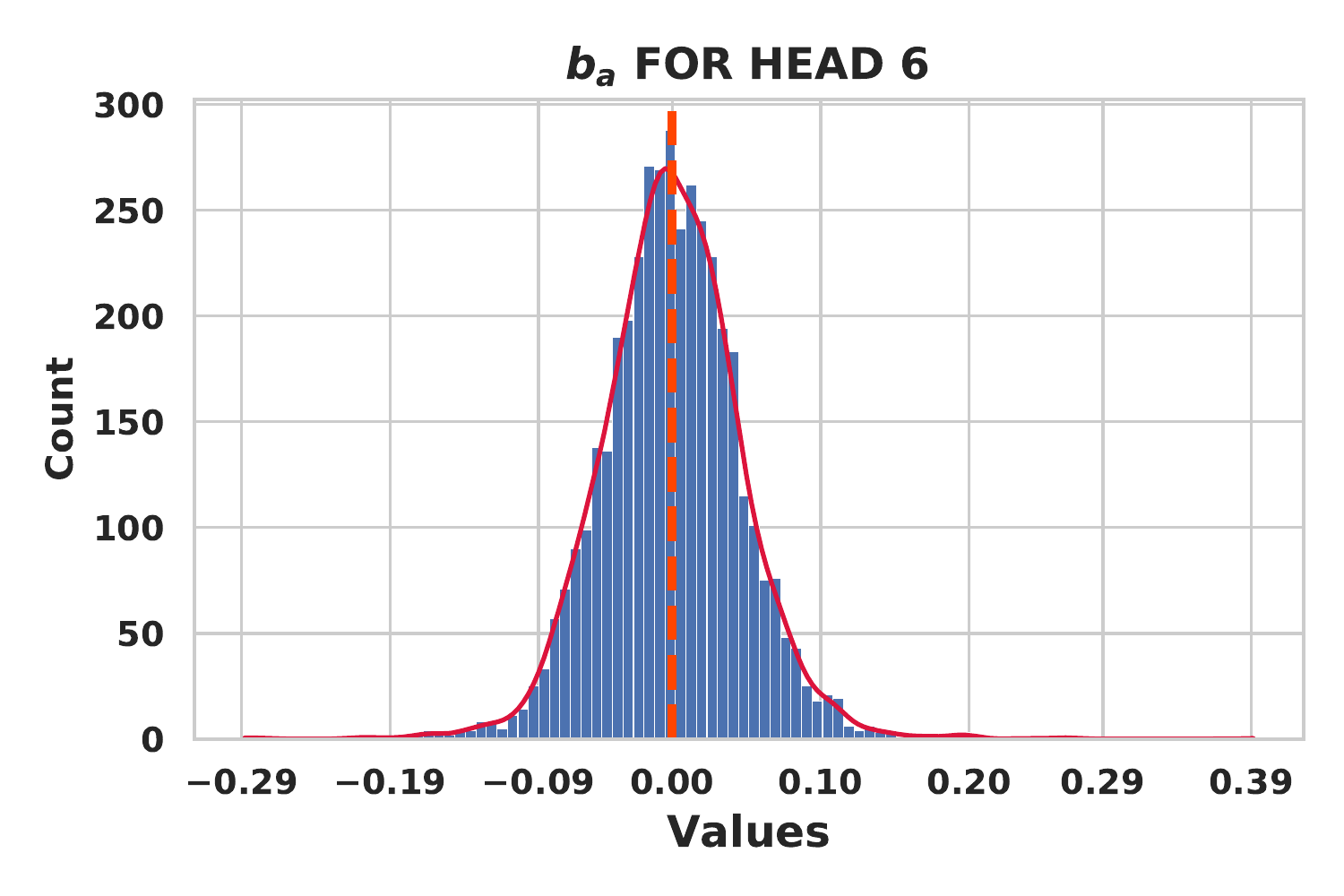}

\end{subfigure}
\hfill
\begin{subfigure}[b]{0.6\textwidth}
	\centering
	\includegraphics[width=1.1\textwidth]{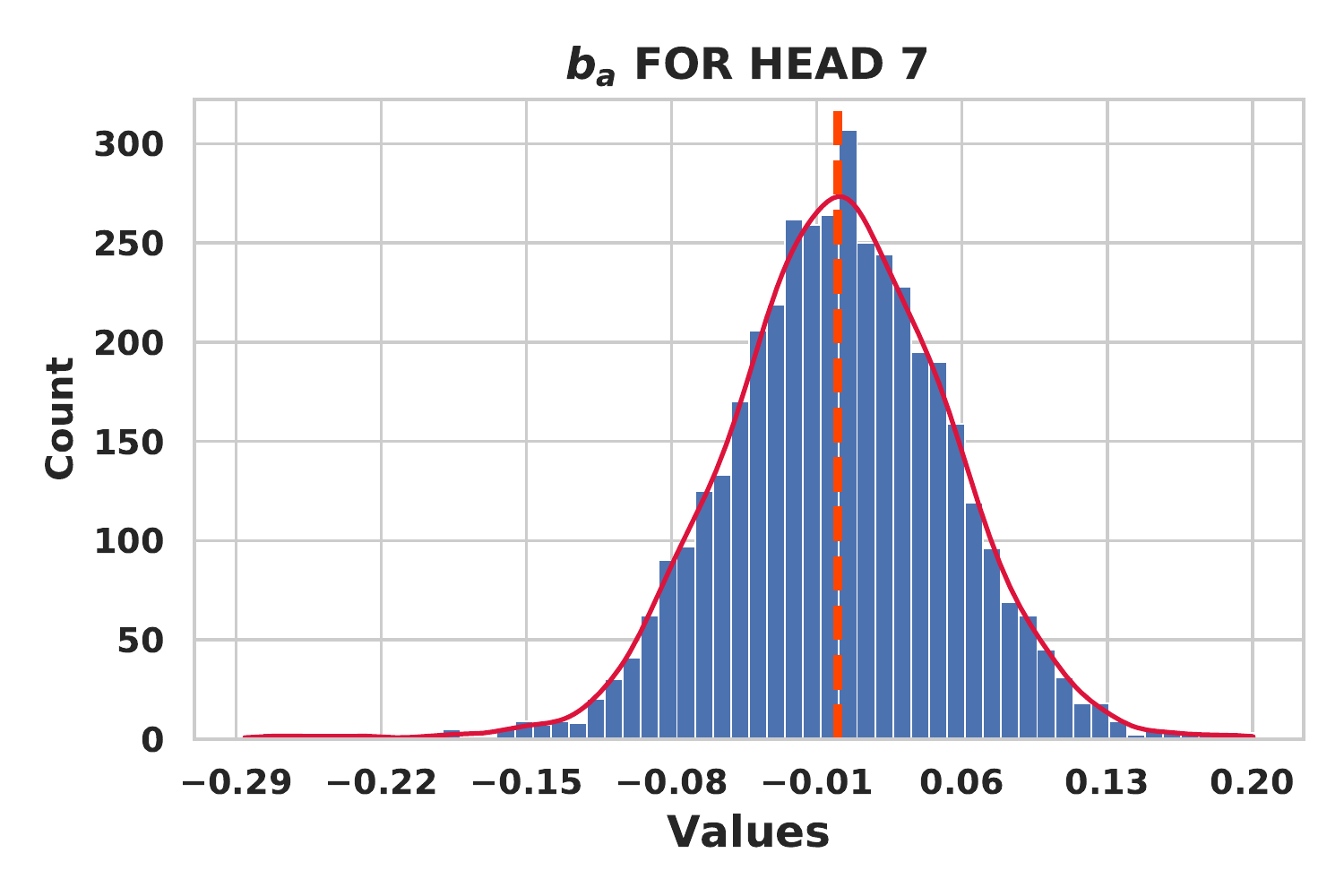}

\end{subfigure}
\hfill
\begin{subfigure}[b]{0.6\textwidth}
	\centering
	\includegraphics[width=1.1\textwidth]{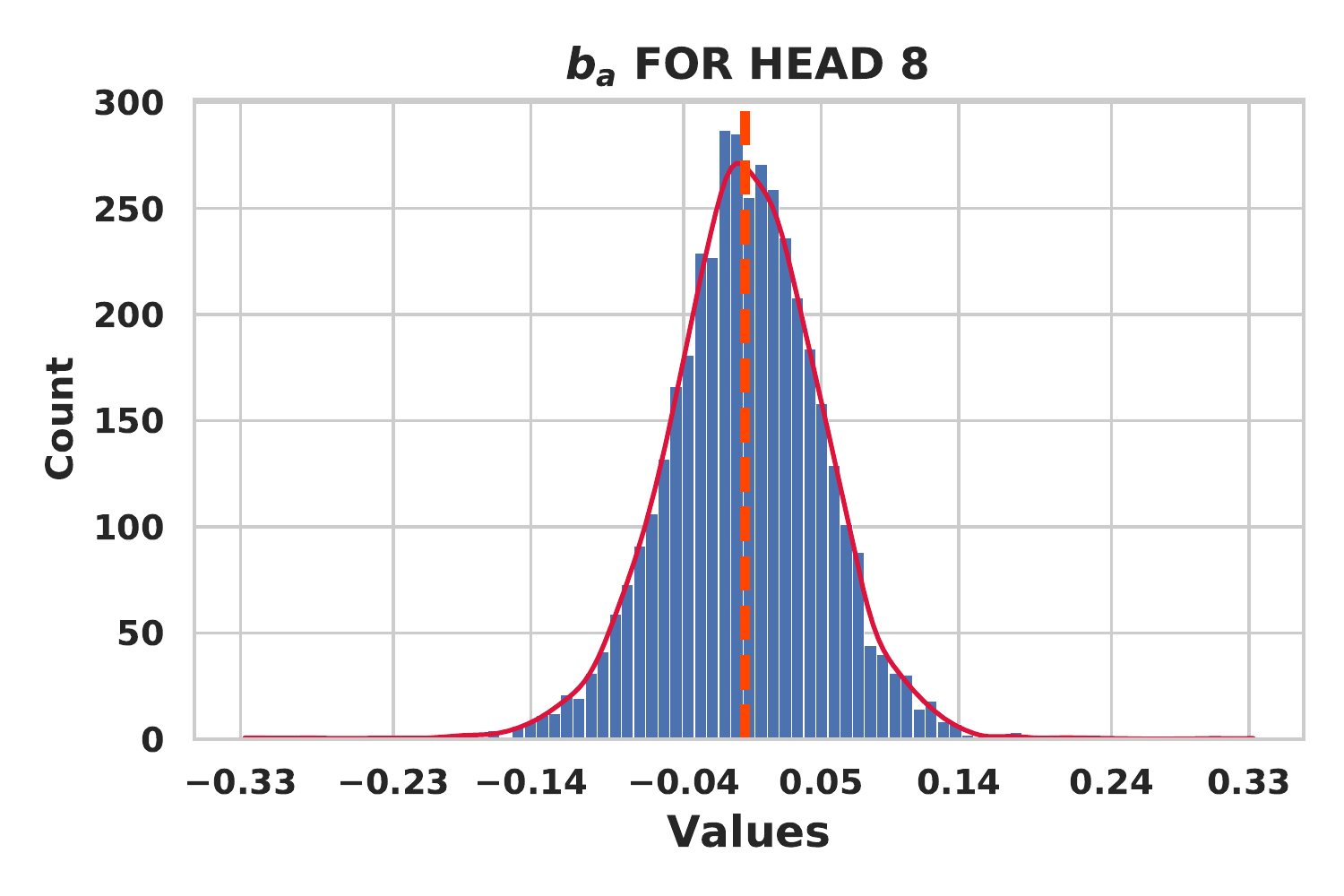}

\end{subfigure}
\caption{$\vb_{a}$ histogram plots for all heads from SLM attention stage from graph transformer model \#2 for PT-EN translation task. Dashed line in orange marks zero value.}
\label{fig18apx}
\end{adjustwidth}
\end{figure} 

\clearpage
\thispagestyle{headings}
\begin{figure}
\begin{adjustwidth}{-5em}{-5em}
\centering
\begin{subfigure}[b]{0.6\textwidth}
	\centering
	\includegraphics[width=1.1\textwidth]{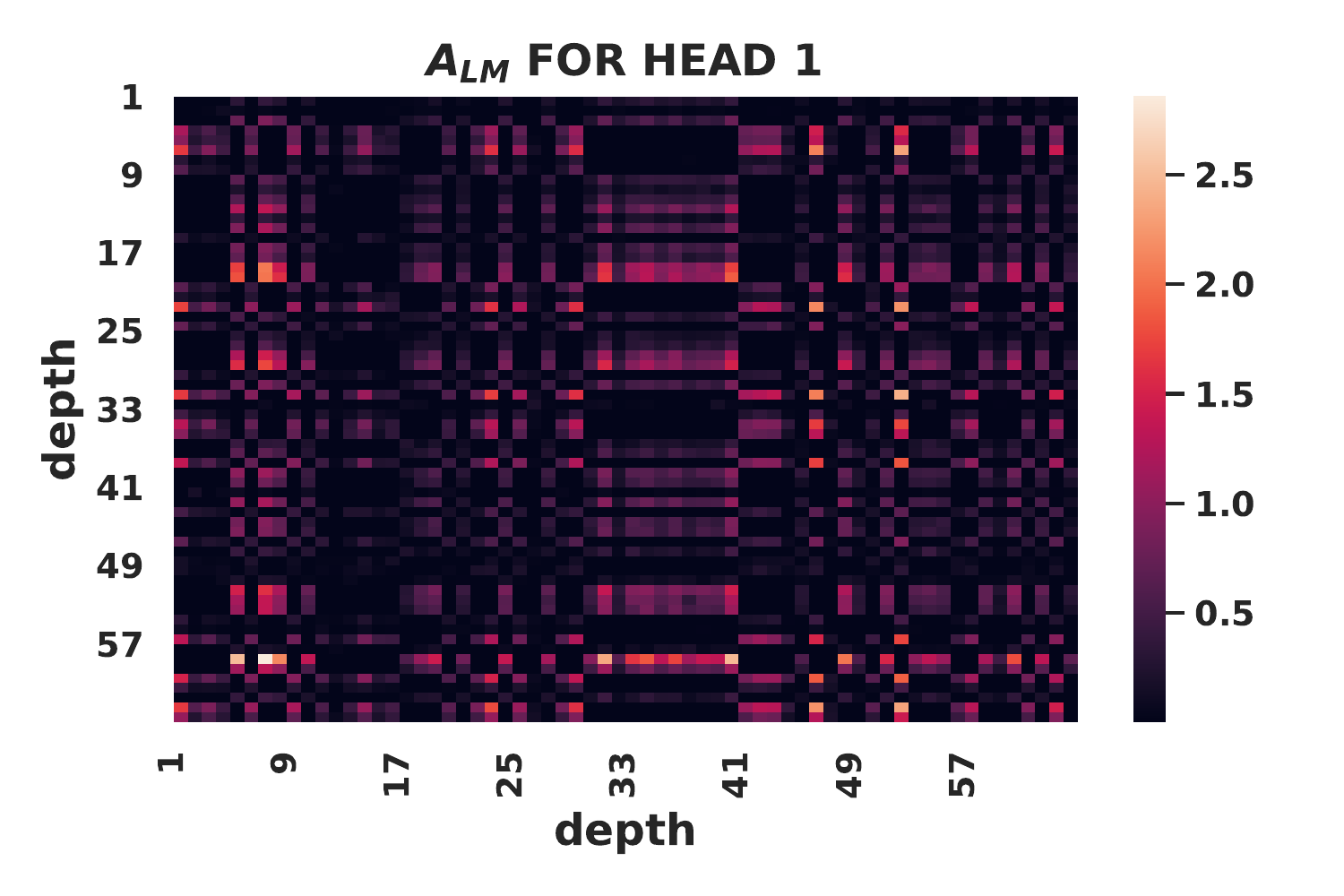}

\end{subfigure}
\hfill
\begin{subfigure}[b]{0.6\textwidth}
	\centering
	\includegraphics[width=1.1\textwidth]{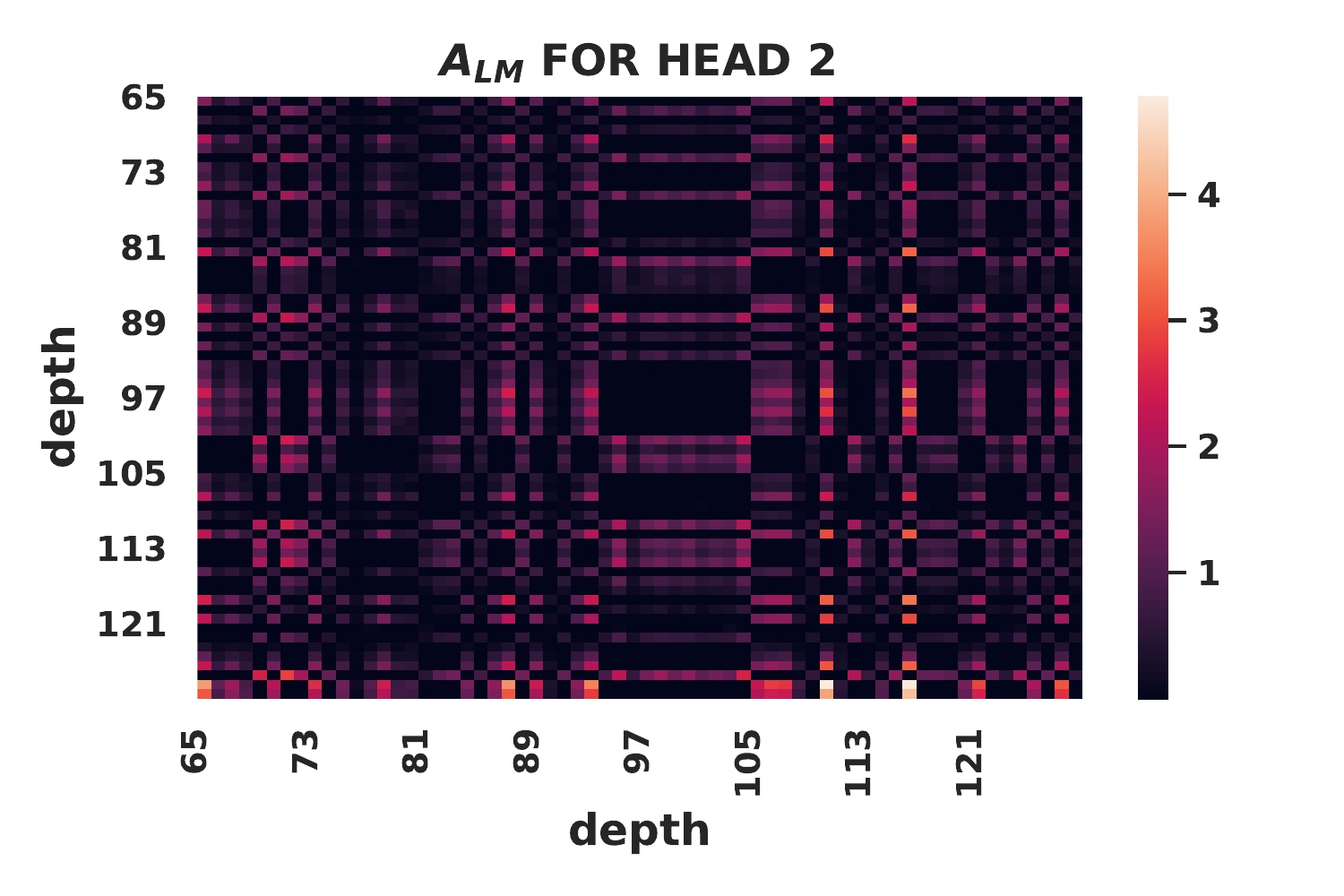}

\end{subfigure}
\hfill
\begin{subfigure}[b]{0.6\textwidth}
	\centering
	\includegraphics[width=1.1\textwidth]{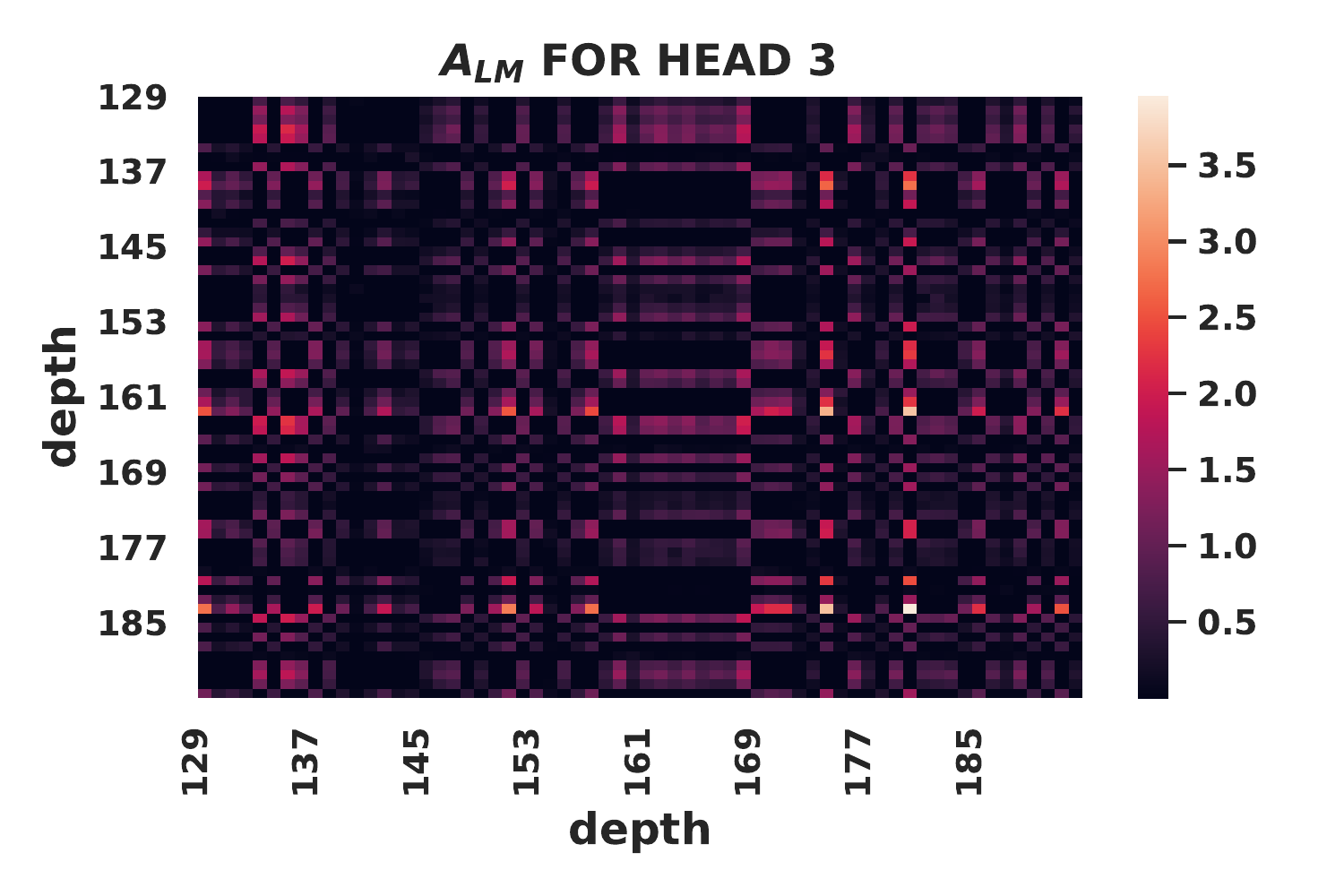}

\end{subfigure}
\hfill
\begin{subfigure}[b]{0.6\textwidth}
	\centering
	\includegraphics[width=1.1\textwidth]{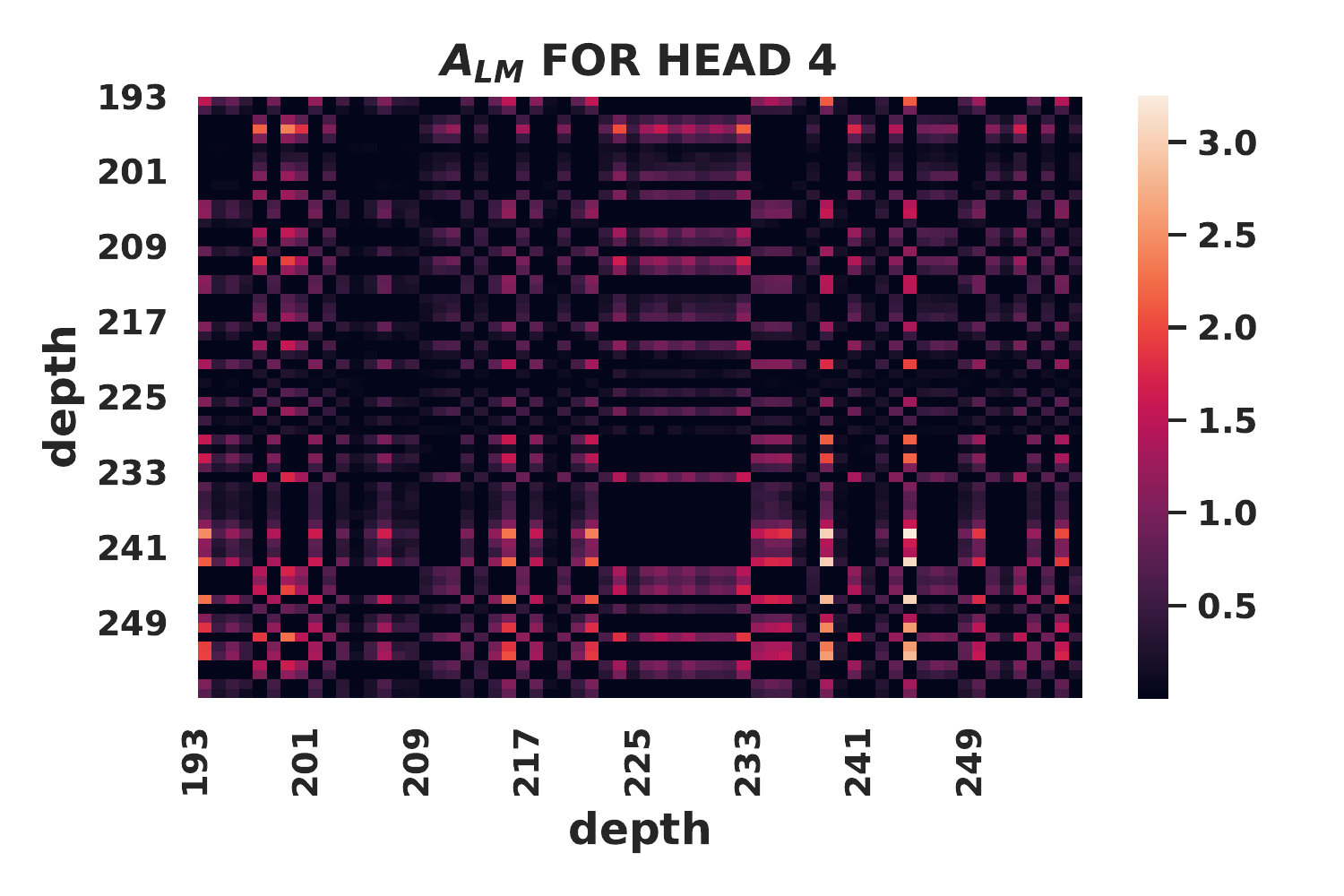}

\end{subfigure}
\centering
\begin{subfigure}[b]{0.6\textwidth}
	\centering
	\includegraphics[width=1.1\textwidth]{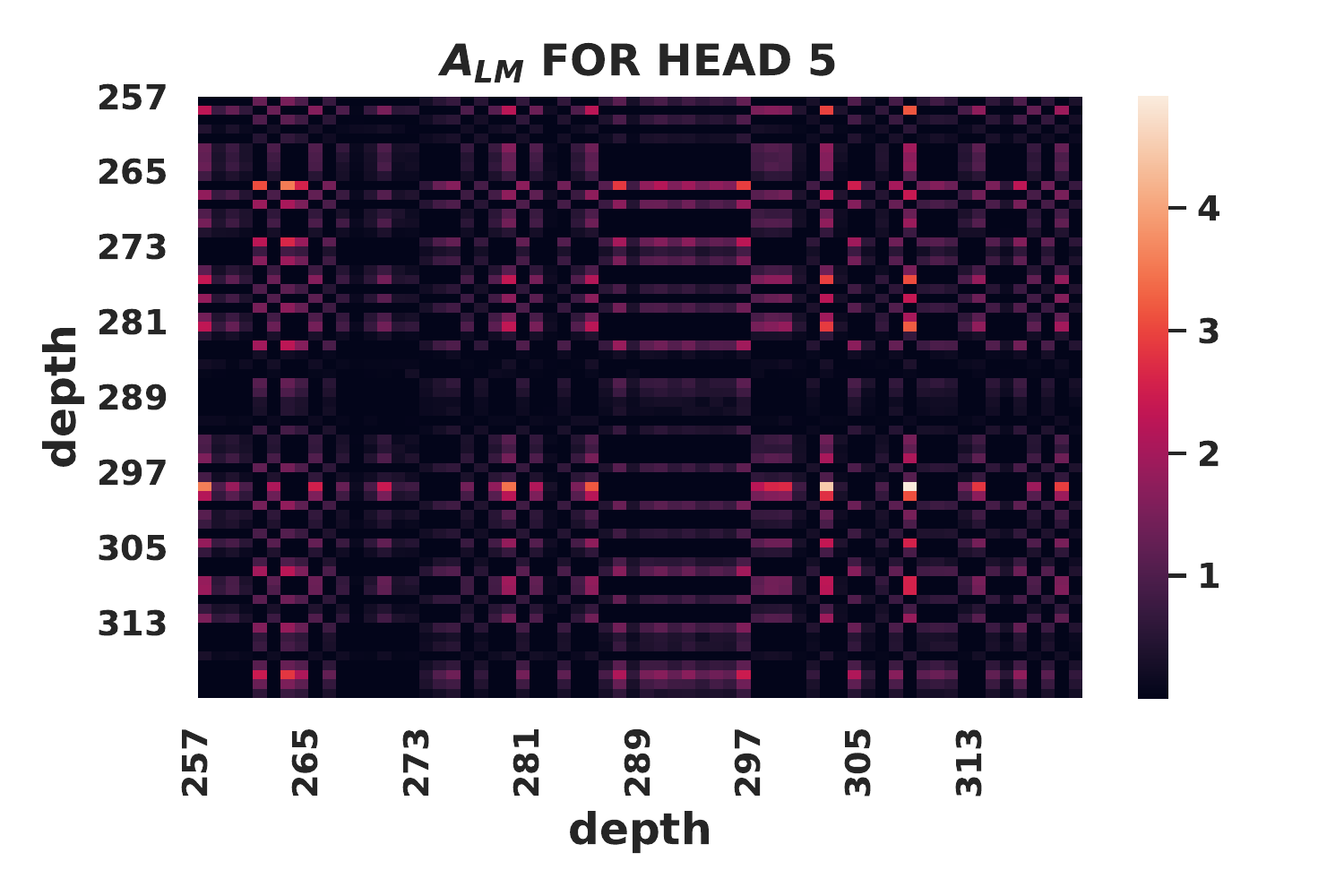}

\end{subfigure}
\hfill
\begin{subfigure}[b]{0.6\textwidth}
	\centering
	\includegraphics[width=1.1\textwidth]{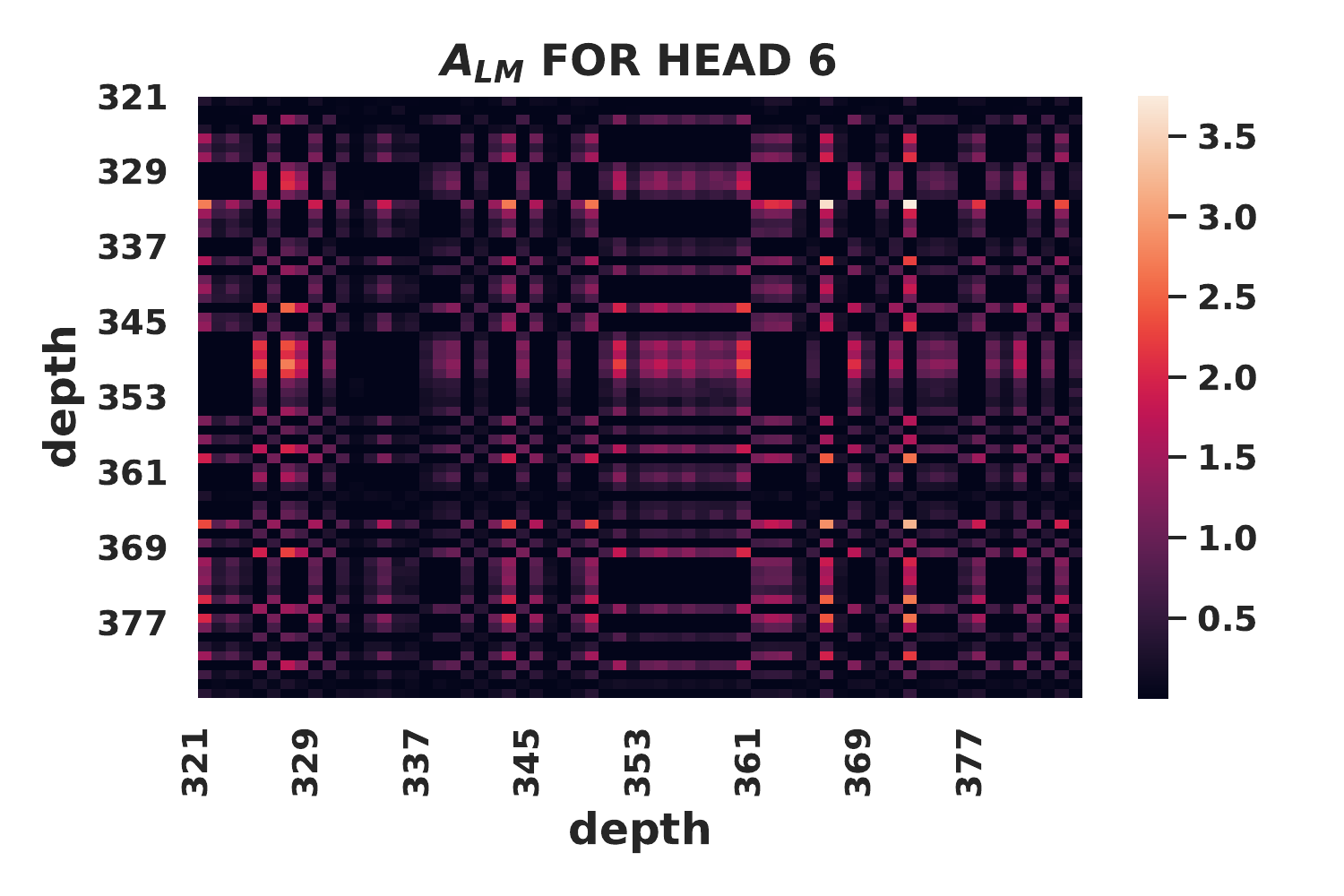}

\end{subfigure}
\hfill
\begin{subfigure}[b]{0.6\textwidth}
	\centering
	\includegraphics[width=1.1\textwidth]{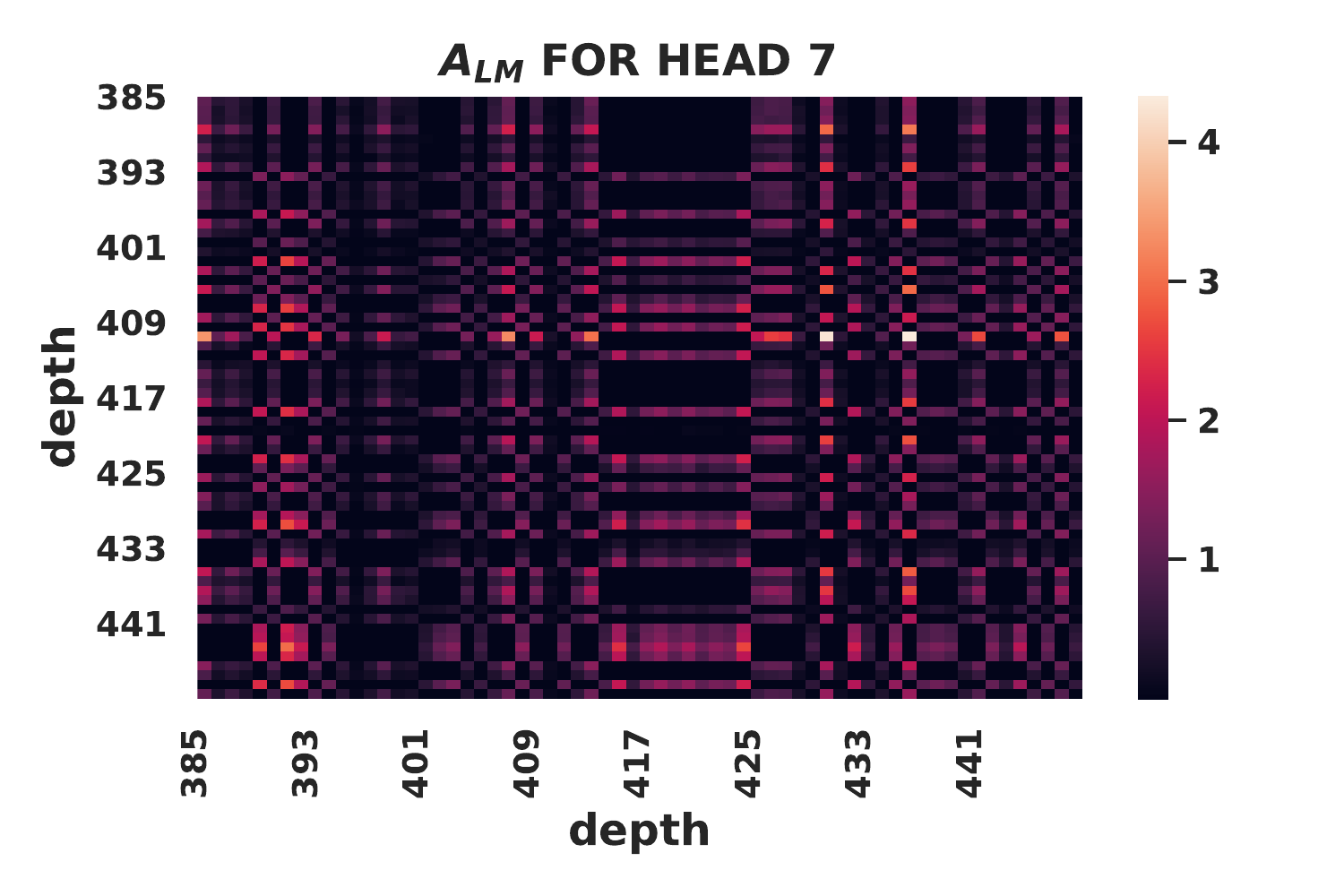}

\end{subfigure}
\hfill
\begin{subfigure}[b]{0.6\textwidth}
	\centering
	\includegraphics[width=1.1\textwidth]{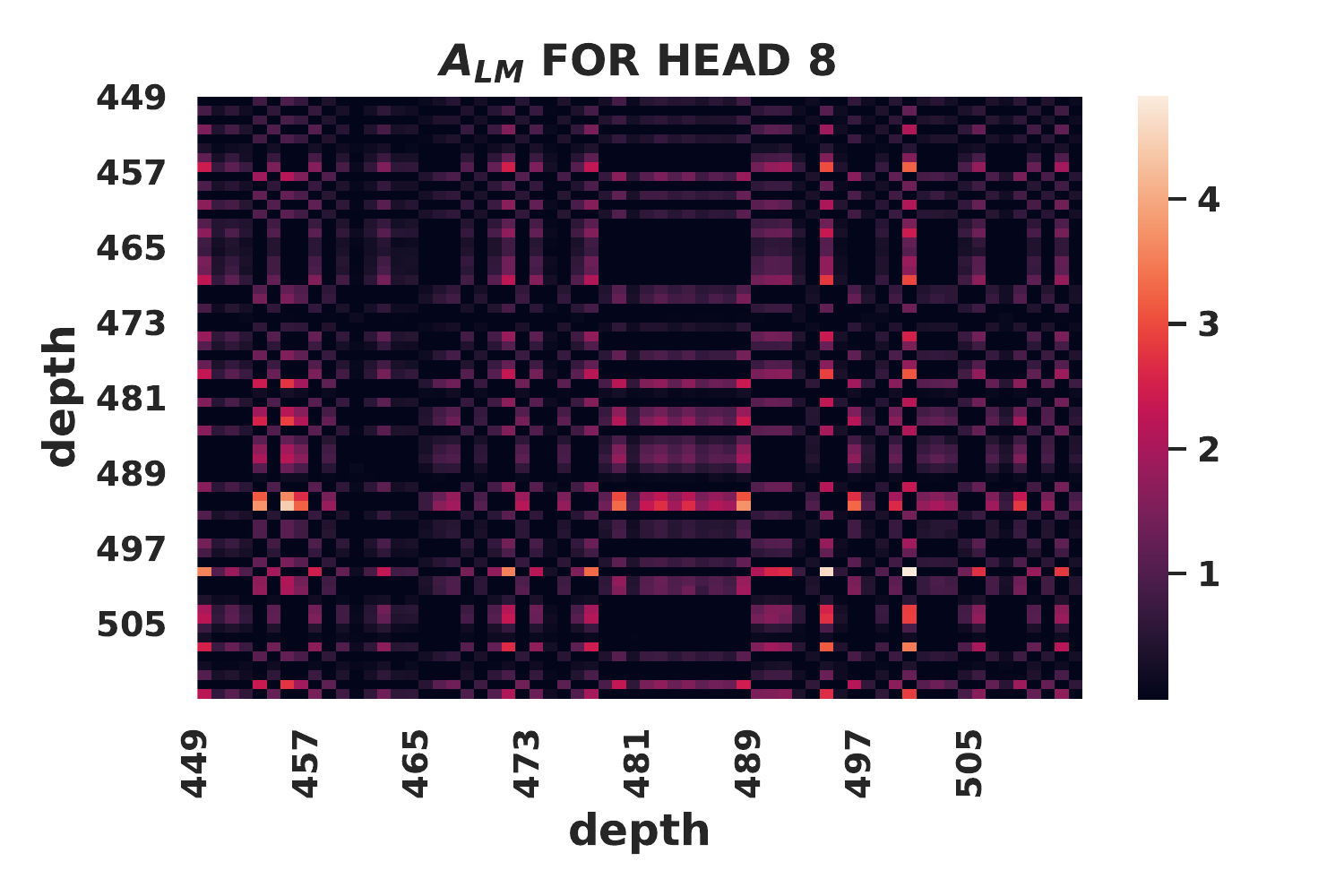}

\end{subfigure}
\caption{$\mA_{LM}$ heatmap plots for all heads from SLM attention stage from graph transformer model \#2 for PT-EN translation task.}
\label{fig19apx}
\end{adjustwidth}
\end{figure} 

\clearpage
\thispagestyle{headings}
\begin{figure}
\begin{adjustwidth}{-5em}{-5em}
\centering
\begin{subfigure}[b]{0.6\textwidth}
	\centering
	\includegraphics[width=1.1\textwidth]{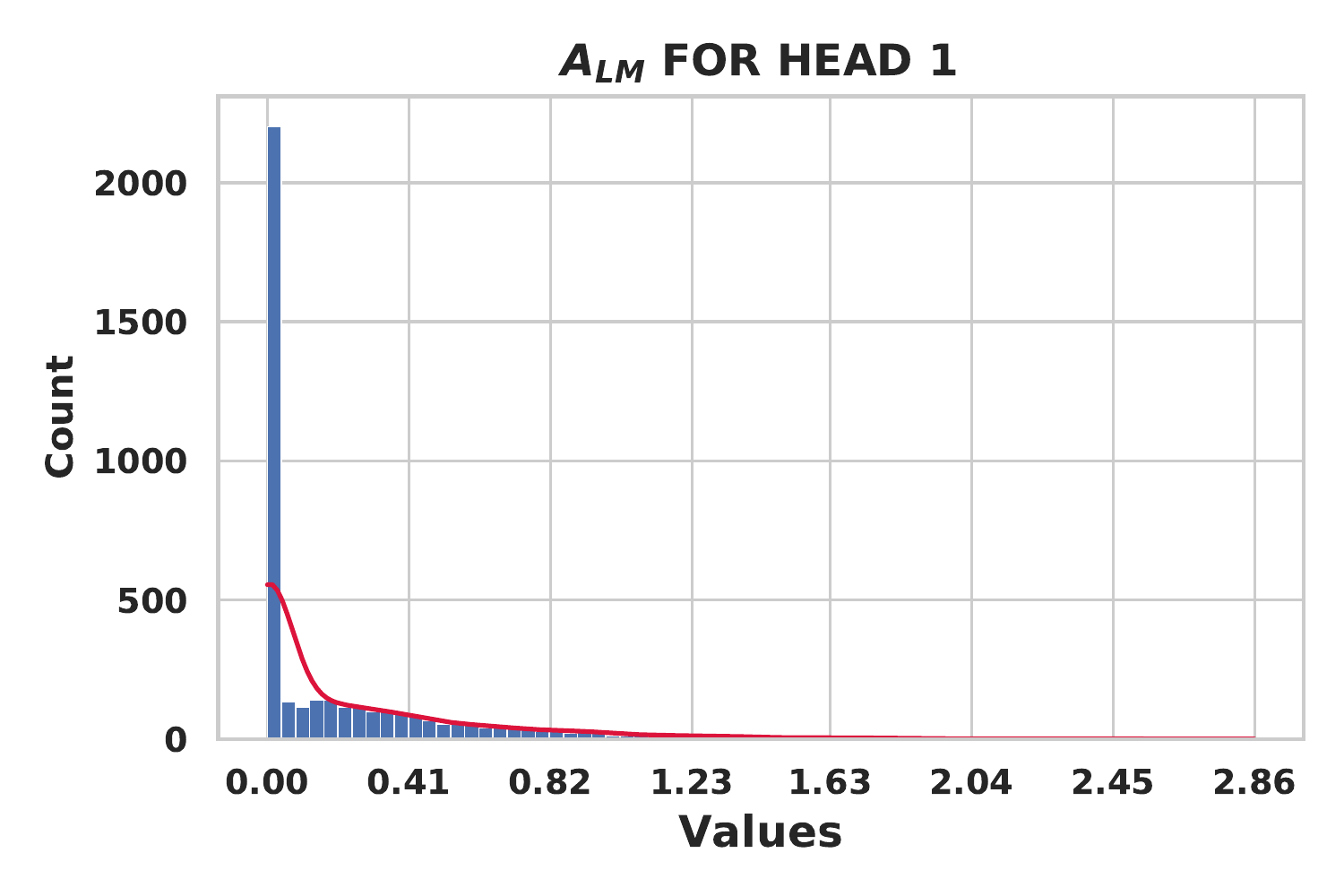}

\end{subfigure}
\hfill
\begin{subfigure}[b]{0.6\textwidth}
	\centering
	\includegraphics[width=1.1\textwidth]{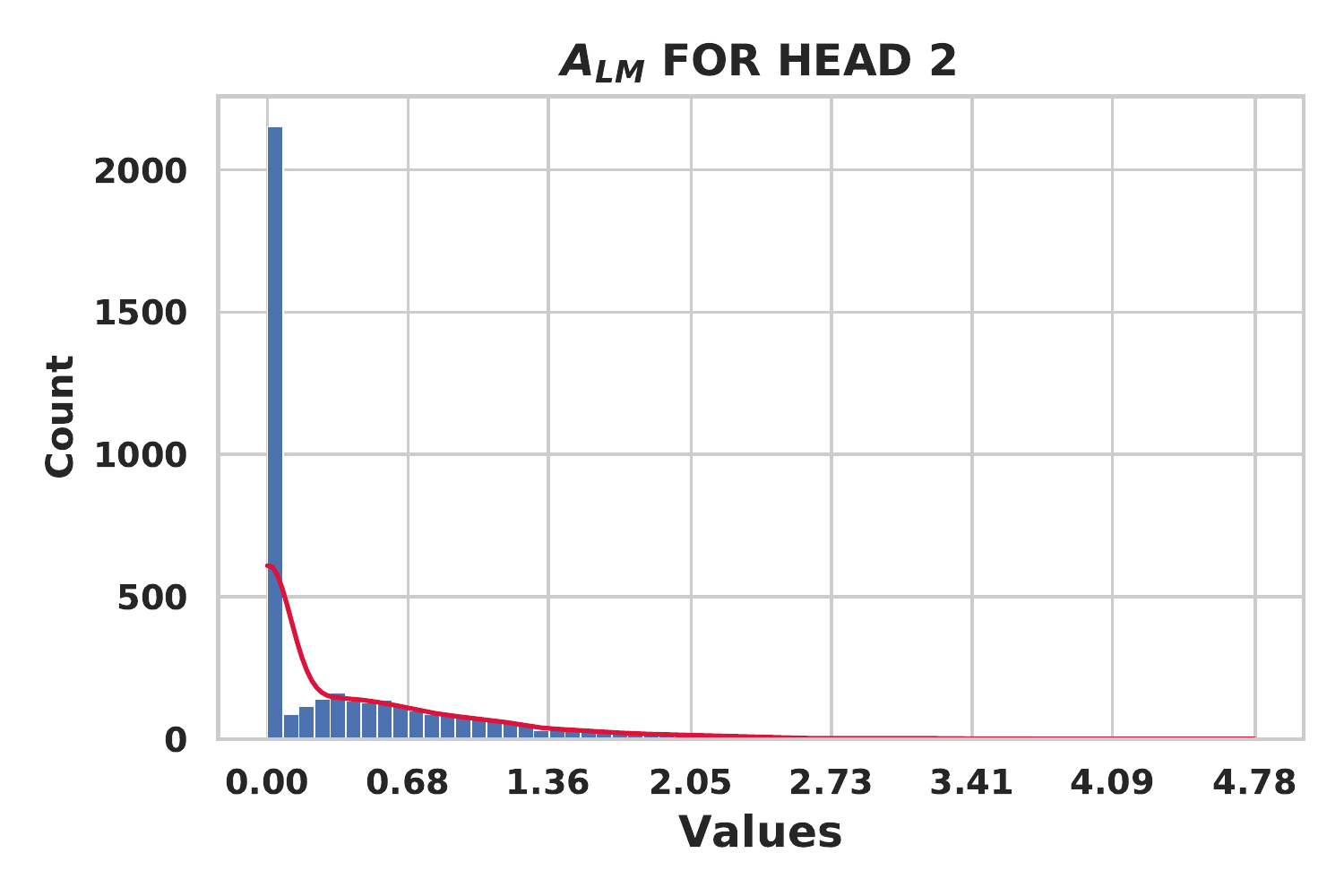}

\end{subfigure}
\hfill
\begin{subfigure}[b]{0.6\textwidth}
	\centering
	\includegraphics[width=1.1\textwidth]{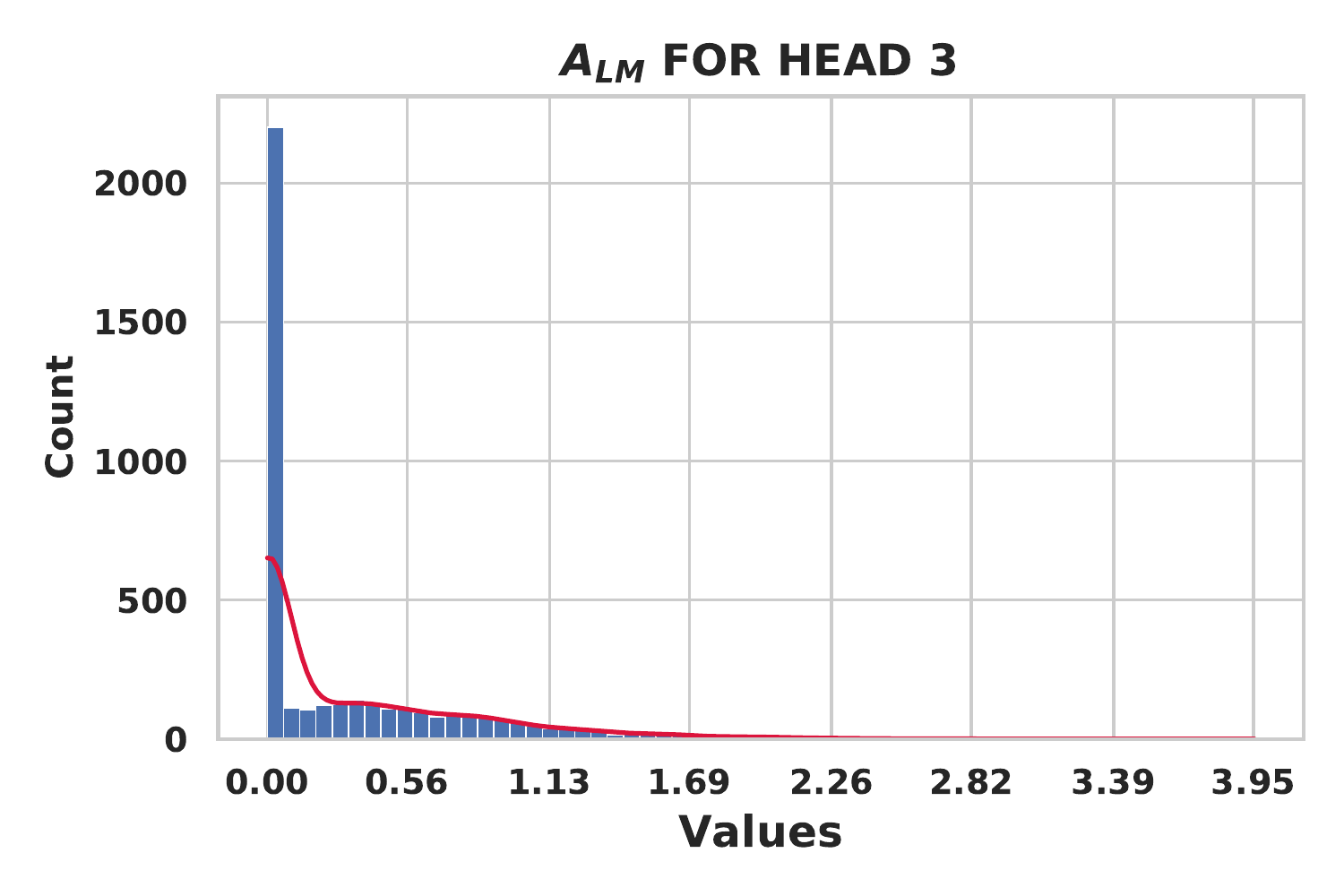}

\end{subfigure}
\hfill
\begin{subfigure}[b]{0.6\textwidth}
	\centering
	\includegraphics[width=1.1\textwidth]{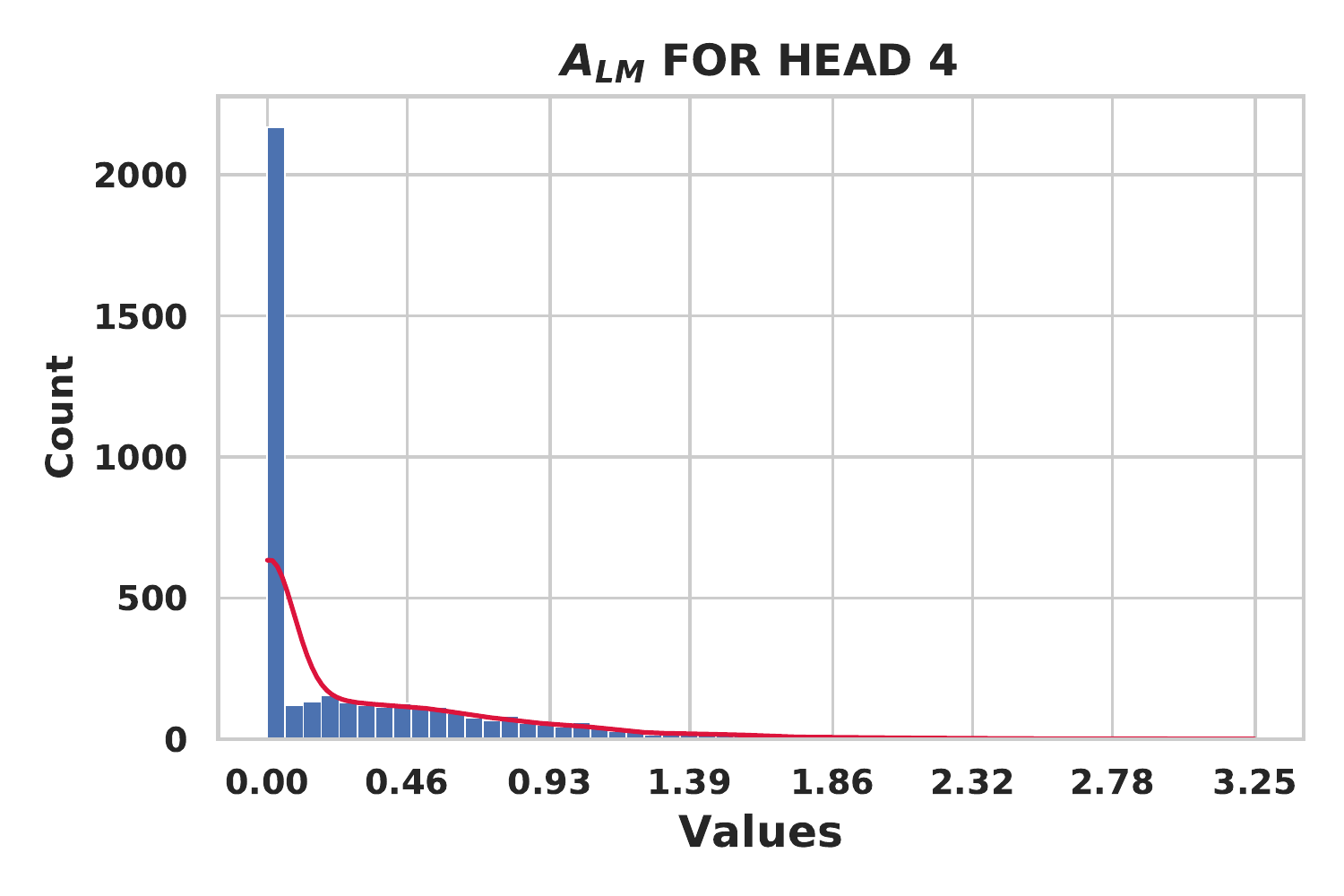}

\end{subfigure}
\centering
\begin{subfigure}[b]{0.6\textwidth}
	\centering
	\includegraphics[width=1.1\textwidth]{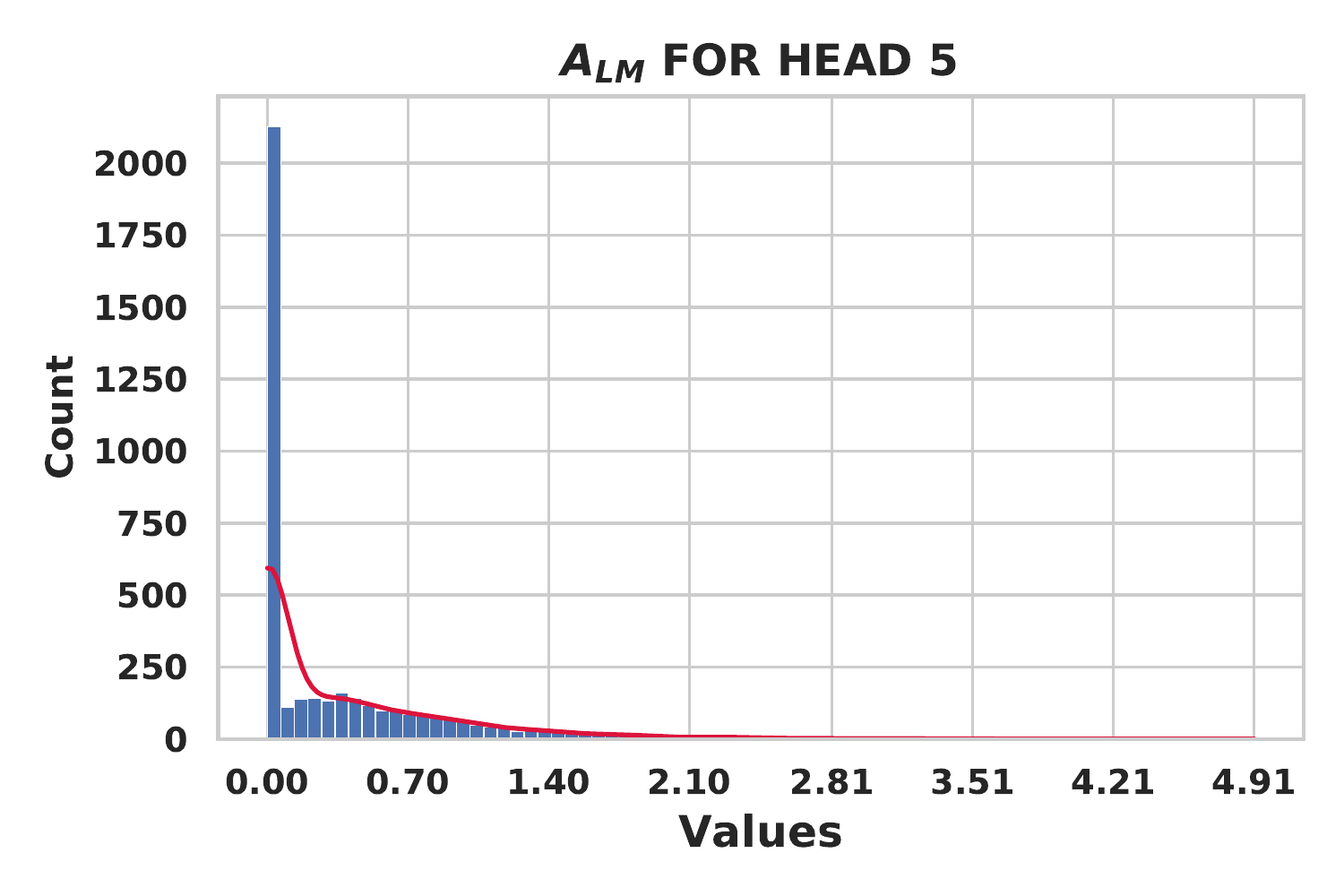}

\end{subfigure}
\hfill
\begin{subfigure}[b]{0.6\textwidth}
	\centering
	\includegraphics[width=1.1\textwidth]{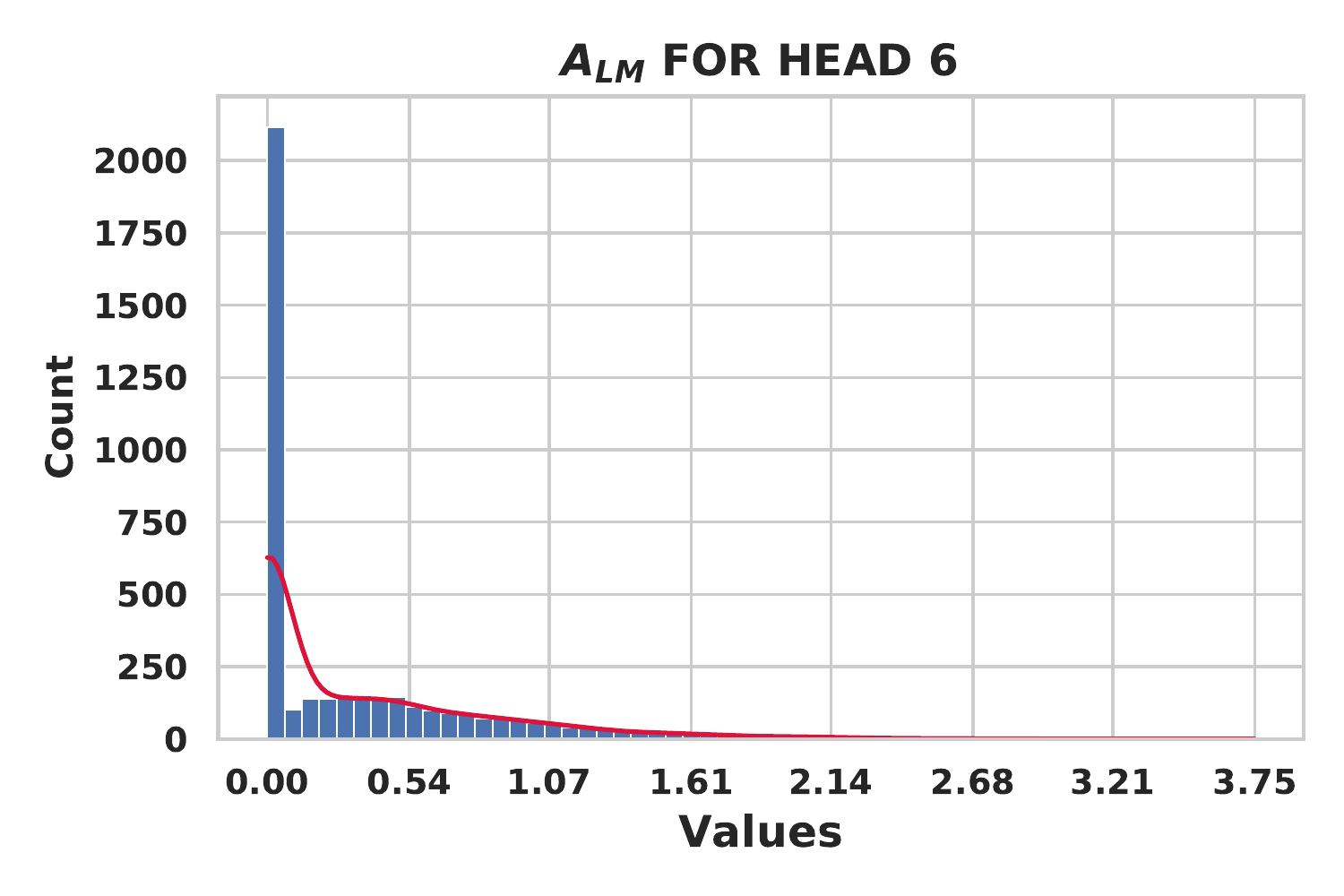}

\end{subfigure}
\hfill
\begin{subfigure}[b]{0.6\textwidth}
	\centering
	\includegraphics[width=1.1\textwidth]{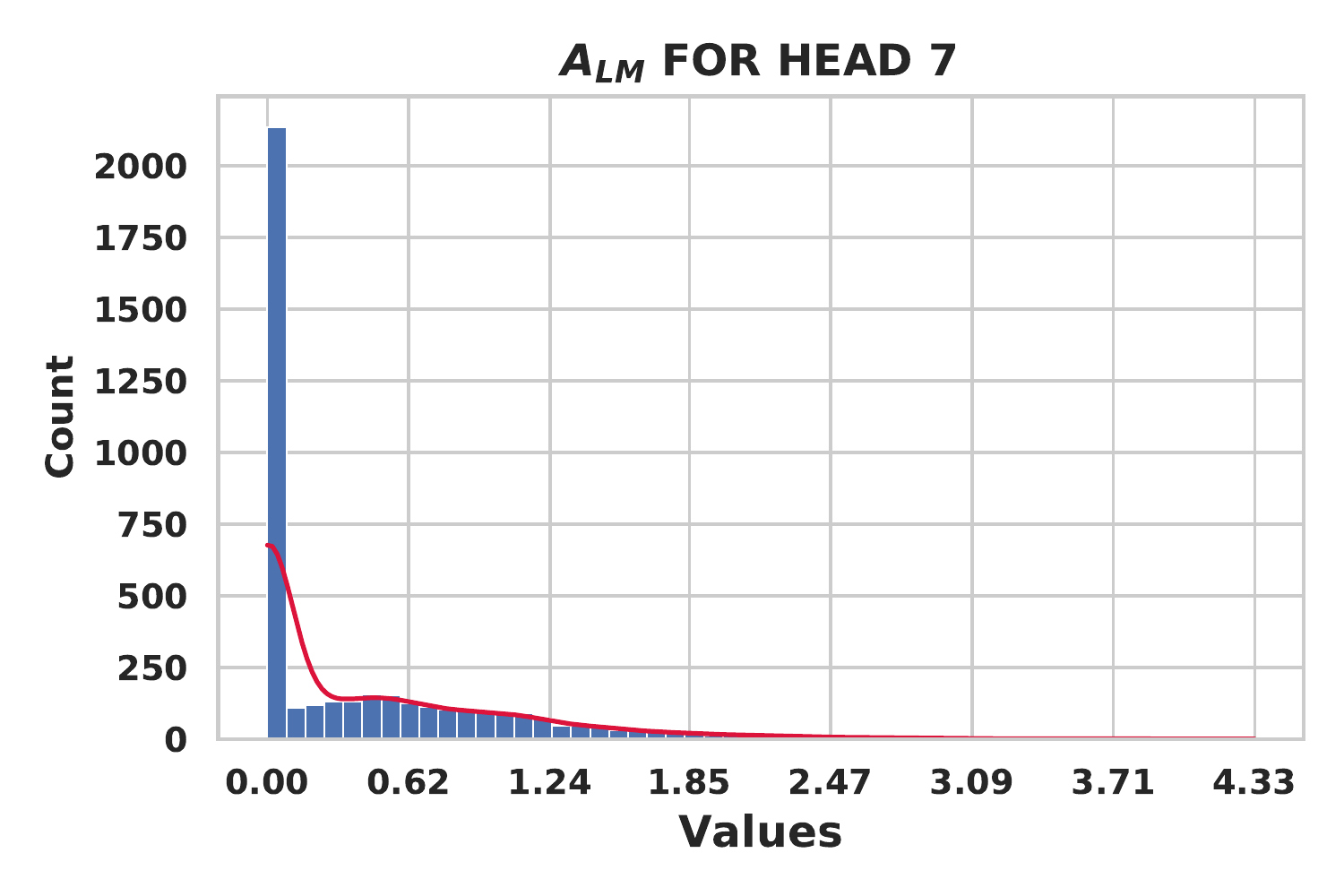}

\end{subfigure}
\hfill
\begin{subfigure}[b]{0.6\textwidth}
	\centering
	\includegraphics[width=1.1\textwidth]{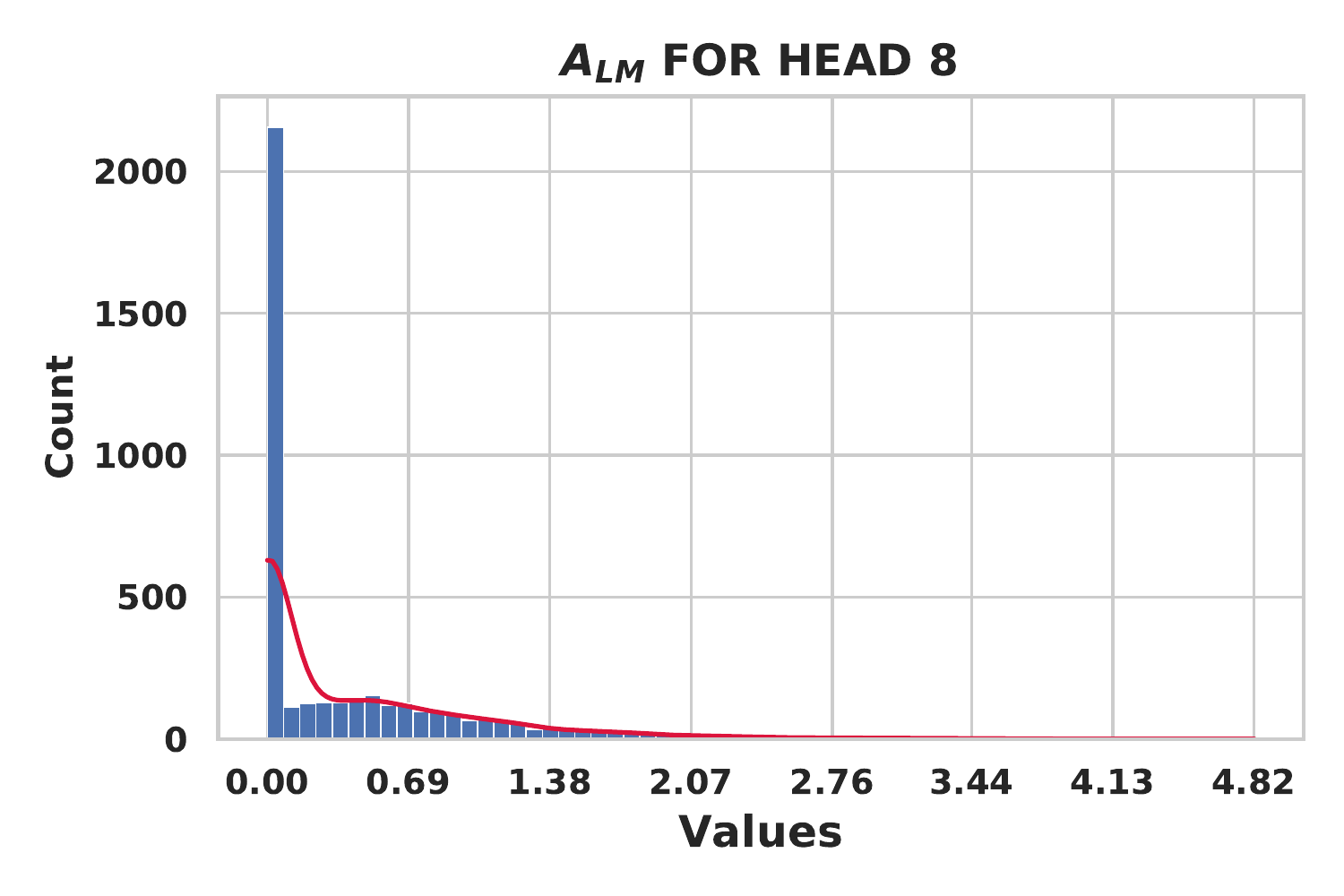}

\end{subfigure}
\caption{$\mA_{LM}$ histogram plots for all heads from SLM attention stage from graph transformer model \#2 for PT-EN translation task.}
\label{fig20apx}
\end{adjustwidth}
\end{figure}
\clearpage
\thispagestyle{headings}

\begin{figure}
\begin{adjustwidth}{-5em}{-5em}
\centering
\begin{subfigure}[b]{0.6\textwidth}
	\centering
	\includegraphics[width=1.1\textwidth]{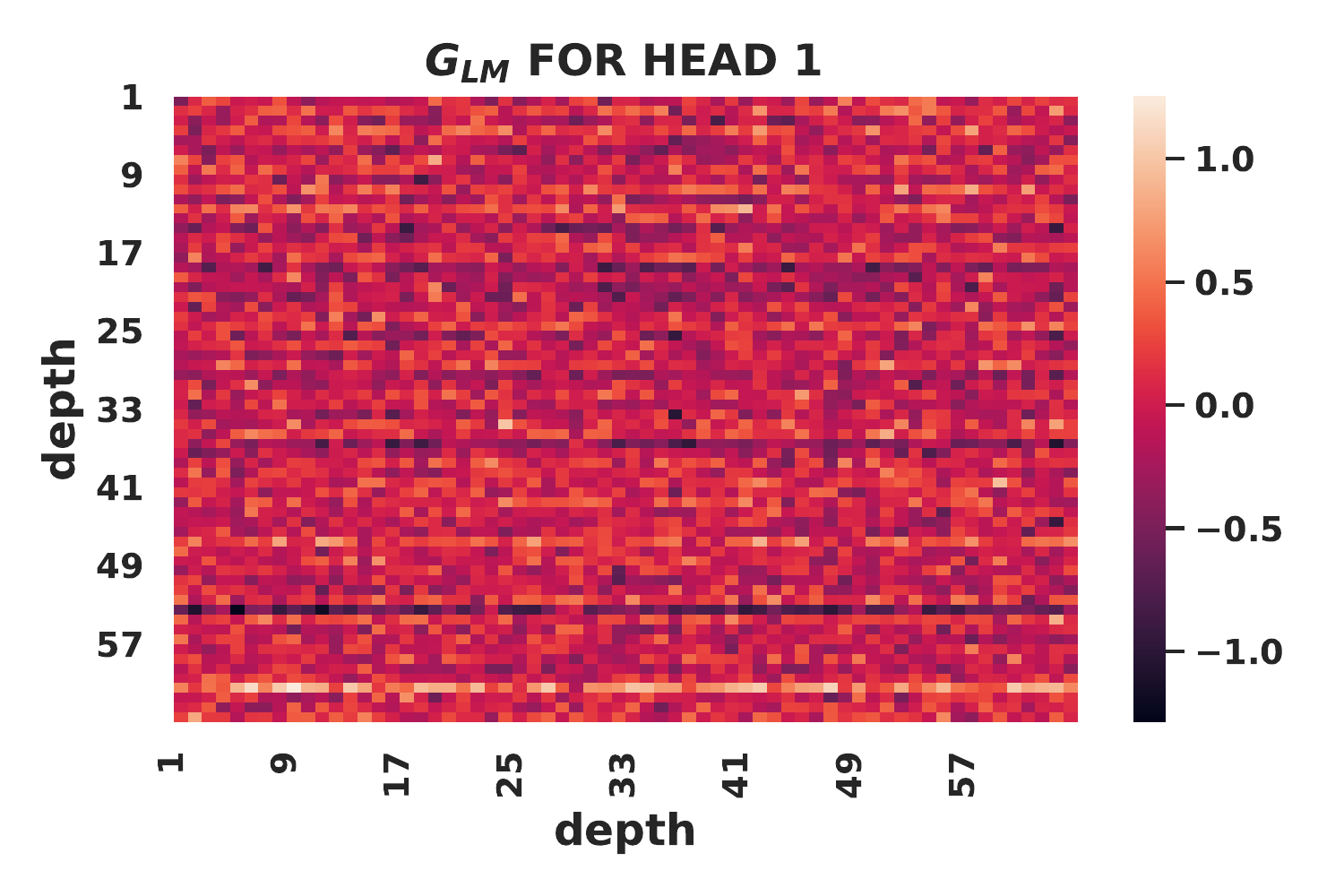}

\end{subfigure}
\hfill
\begin{subfigure}[b]{0.6\textwidth}
	\centering
	\includegraphics[width=1.1\textwidth]{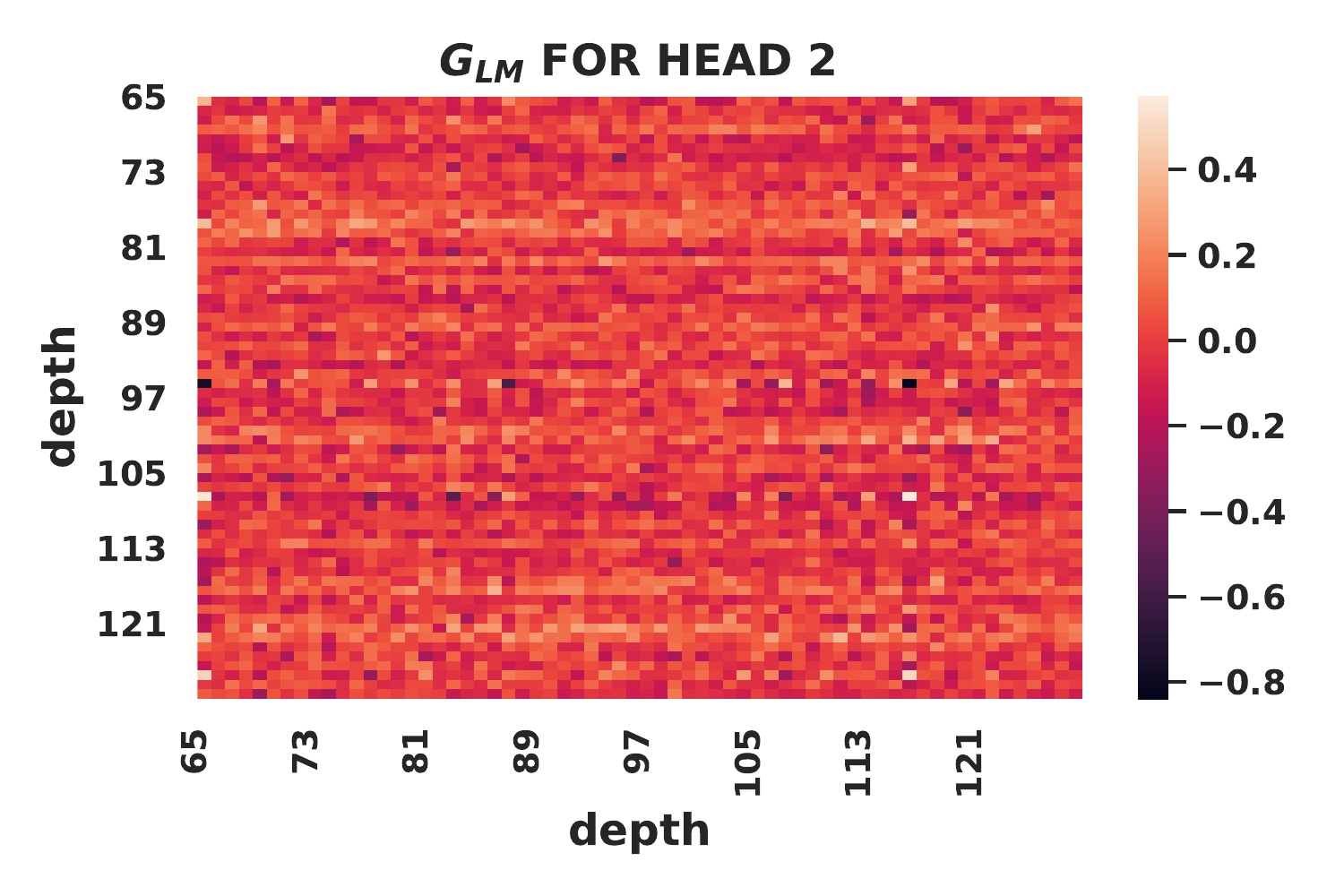}

\end{subfigure}
\hfill
\begin{subfigure}[b]{0.6\textwidth}
	\centering
	\includegraphics[width=1.1\textwidth]{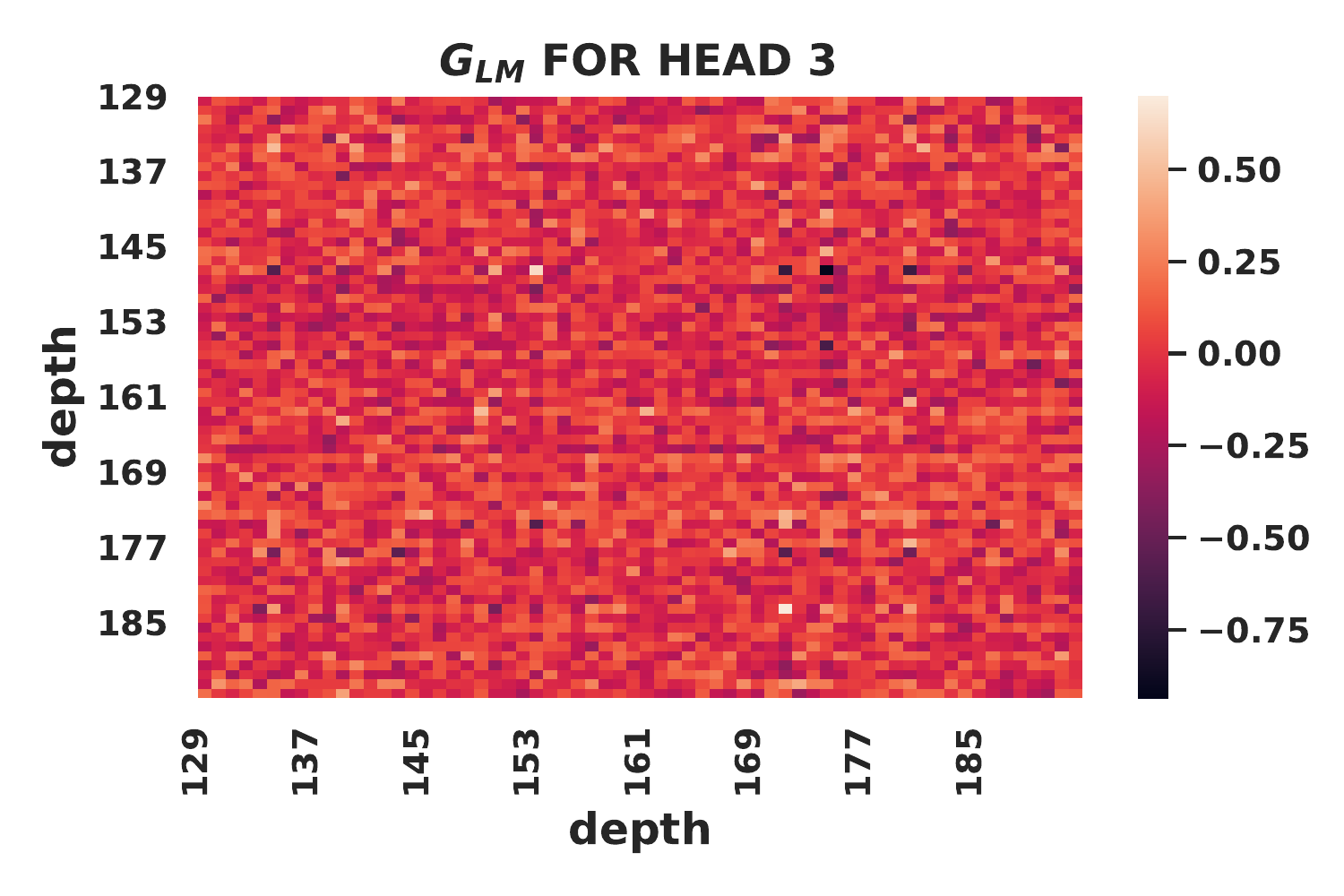}

\end{subfigure}
\hfill
\begin{subfigure}[b]{0.6\textwidth}
	\centering
	\includegraphics[width=1.1\textwidth]{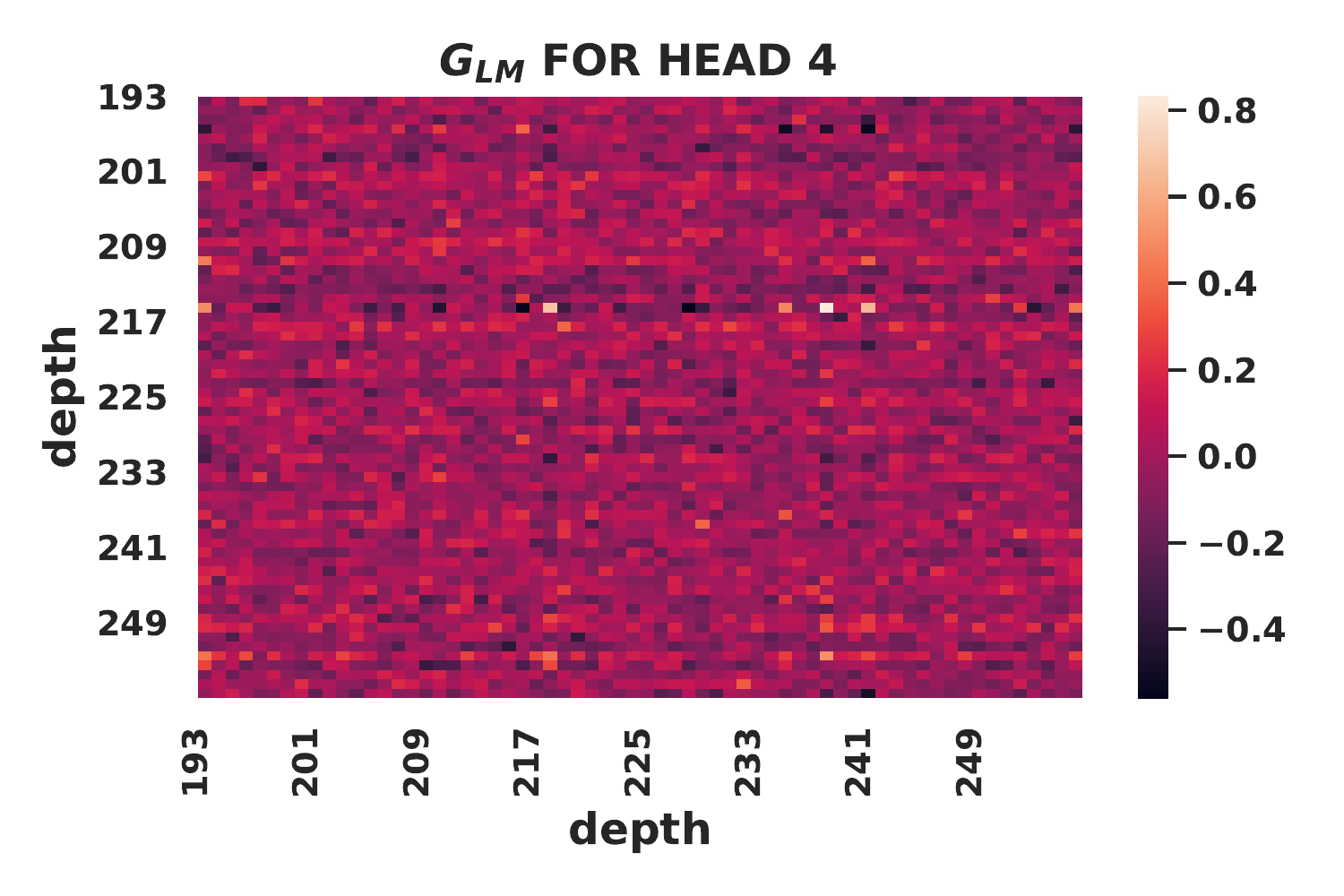}

\end{subfigure}
\centering
\begin{subfigure}[b]{0.6\textwidth}
	\centering
	\includegraphics[width=1.1\textwidth]{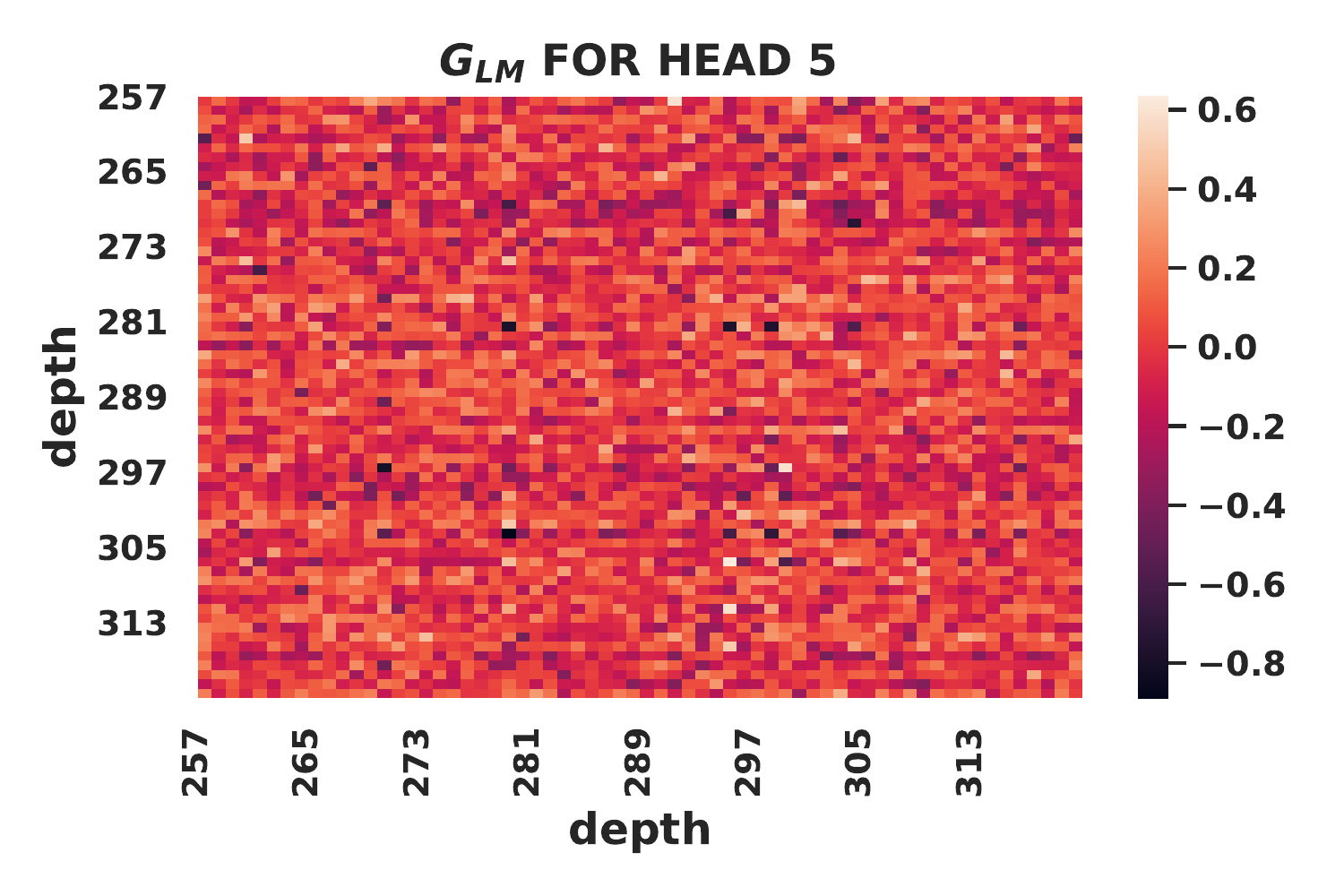}

\end{subfigure}
\hfill
\begin{subfigure}[b]{0.6\textwidth}
	\centering
	\includegraphics[width=1.1\textwidth]{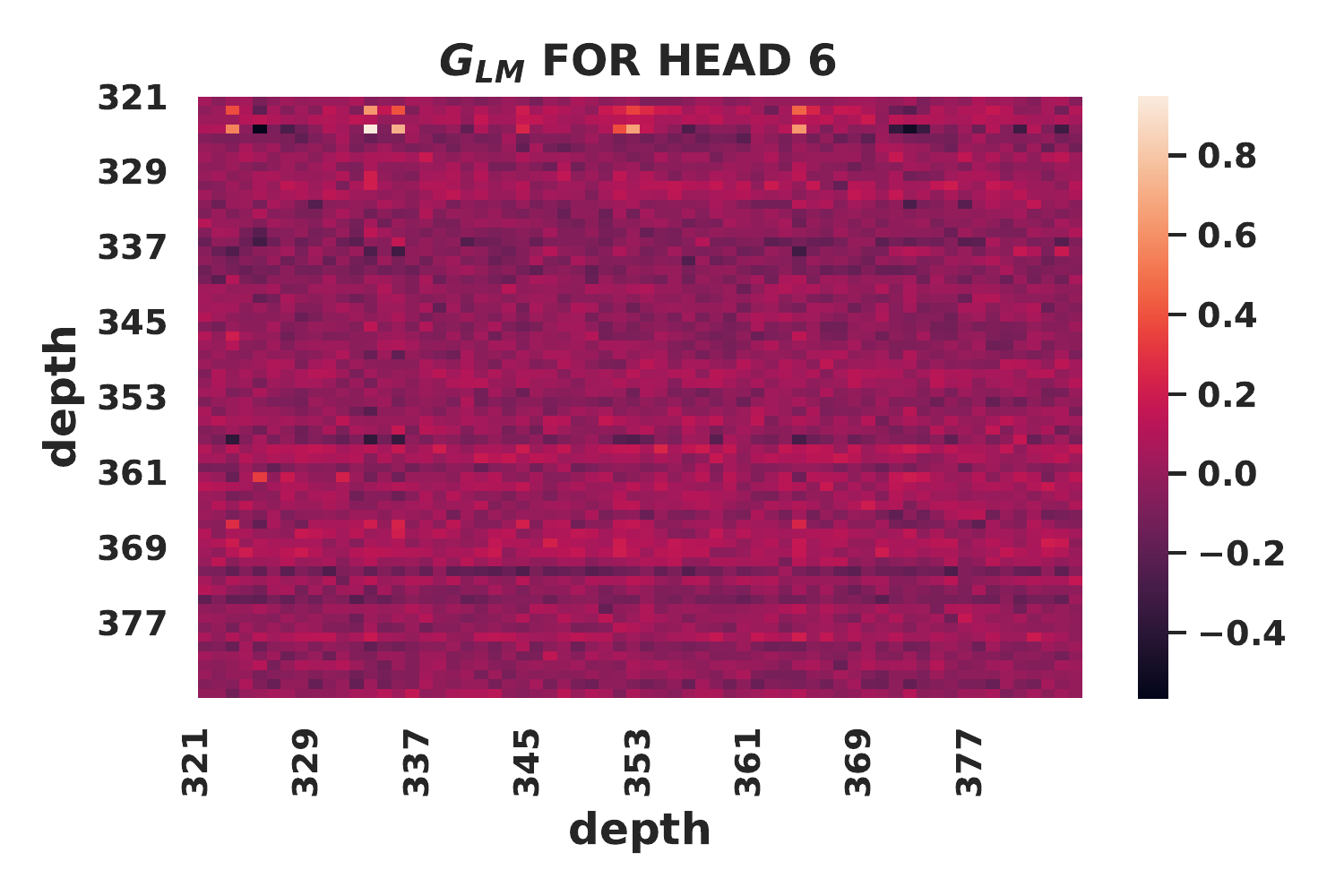}

\end{subfigure}
\hfill
\begin{subfigure}[b]{0.6\textwidth}
	\centering
	\includegraphics[width=1.1\textwidth]{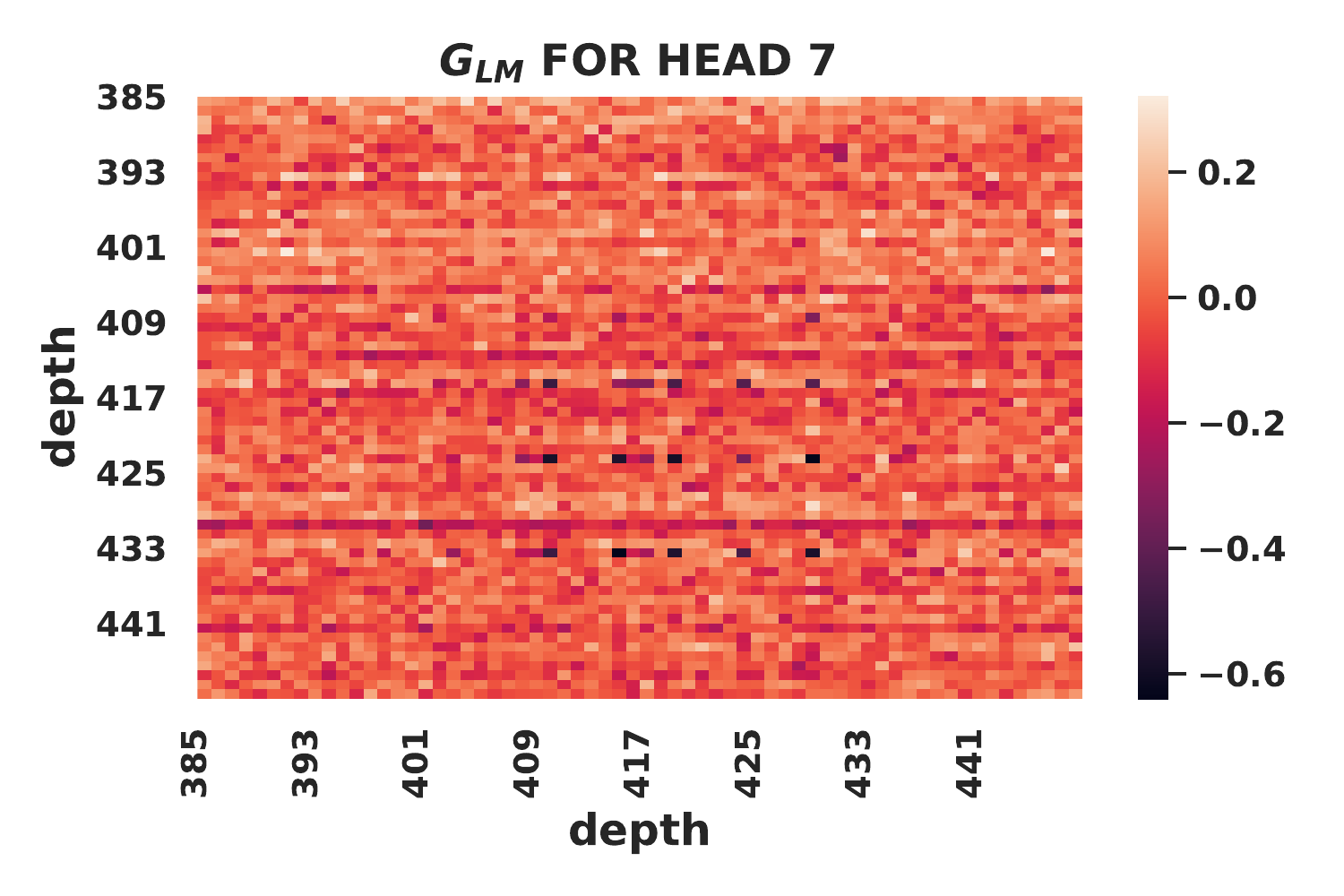}

\end{subfigure}
\hfill
\begin{subfigure}[b]{0.6\textwidth}
	\centering
	\includegraphics[width=1.1\textwidth]{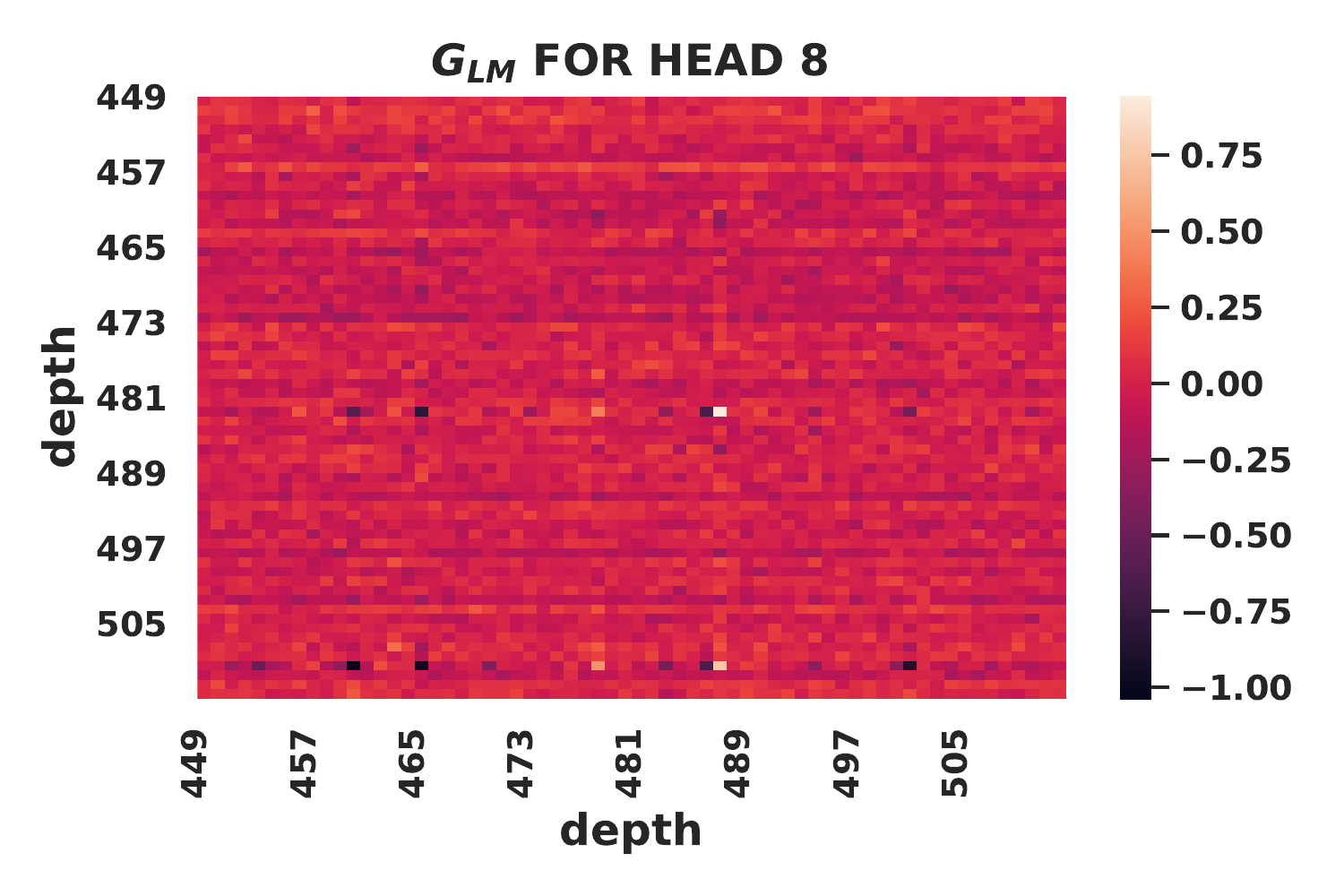}

\end{subfigure}
\caption{$\mG_{LM}$ heatmap plots for all heads from SLM attention stage from graph transformer model \#2 for PT-EN translation task.}
\label{fig21apx}
\end{adjustwidth}
\end{figure}

\clearpage

\thispagestyle{headings}
\begin{figure}
\begin{adjustwidth}{-5em}{-5em}
\centering
\begin{subfigure}[b]{0.6\textwidth}
	\centering
	\includegraphics[width=1.1\textwidth]{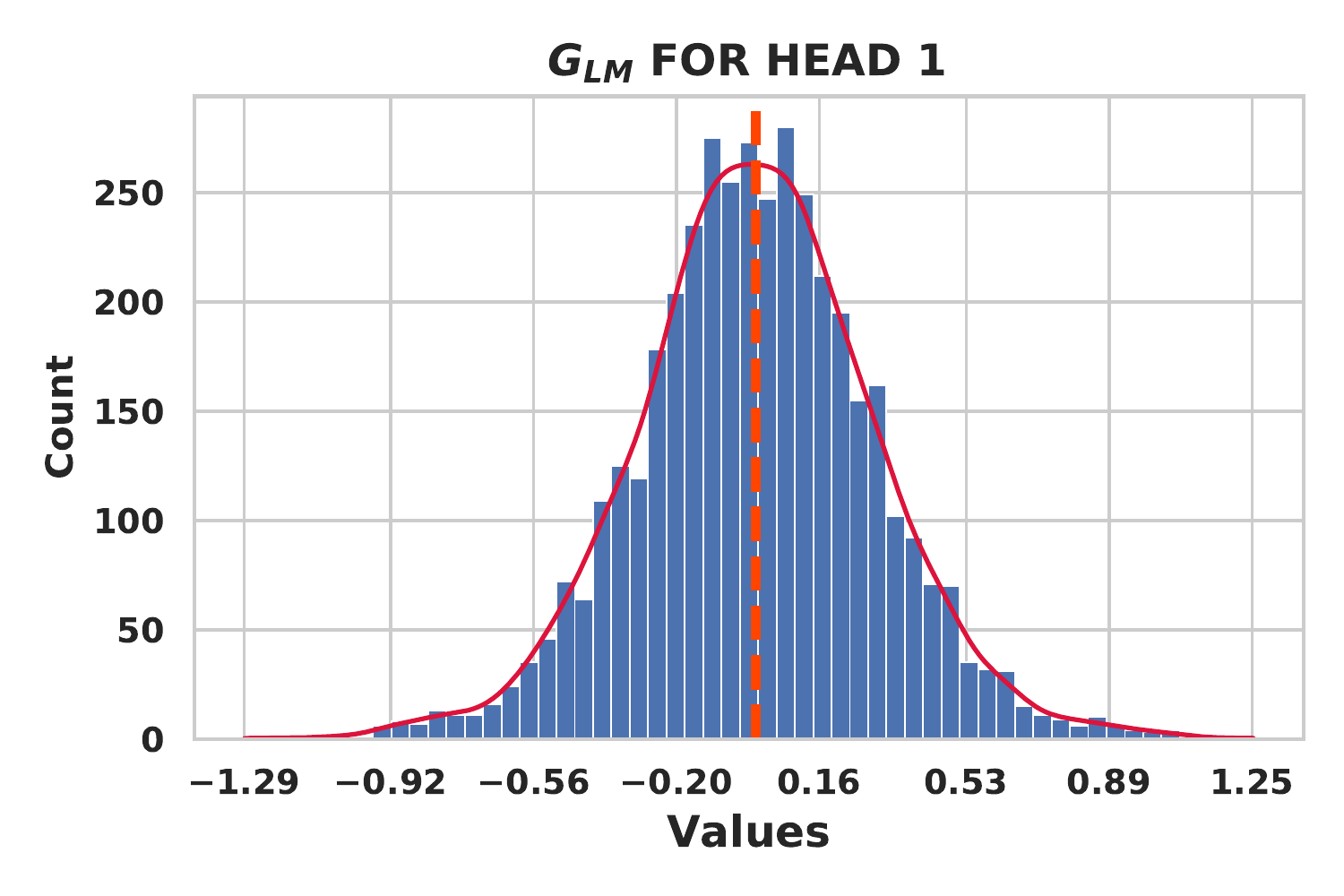}

\end{subfigure}
\hfill
\begin{subfigure}[b]{0.6\textwidth}
	\centering
	\includegraphics[width=1.1\textwidth]{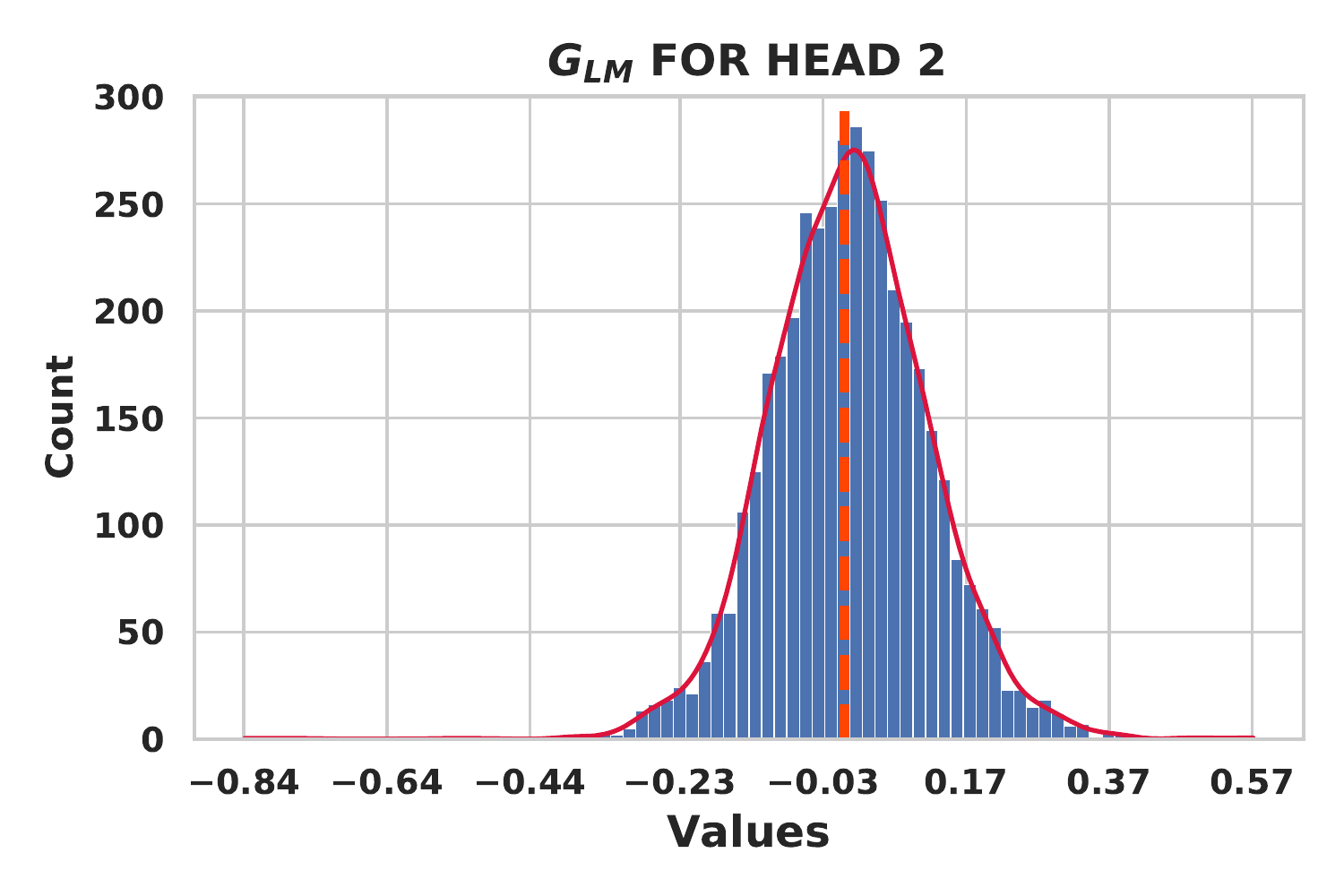}

\end{subfigure}
\hfill
\begin{subfigure}[b]{0.6\textwidth}
	\centering
	\includegraphics[width=1.1\textwidth]{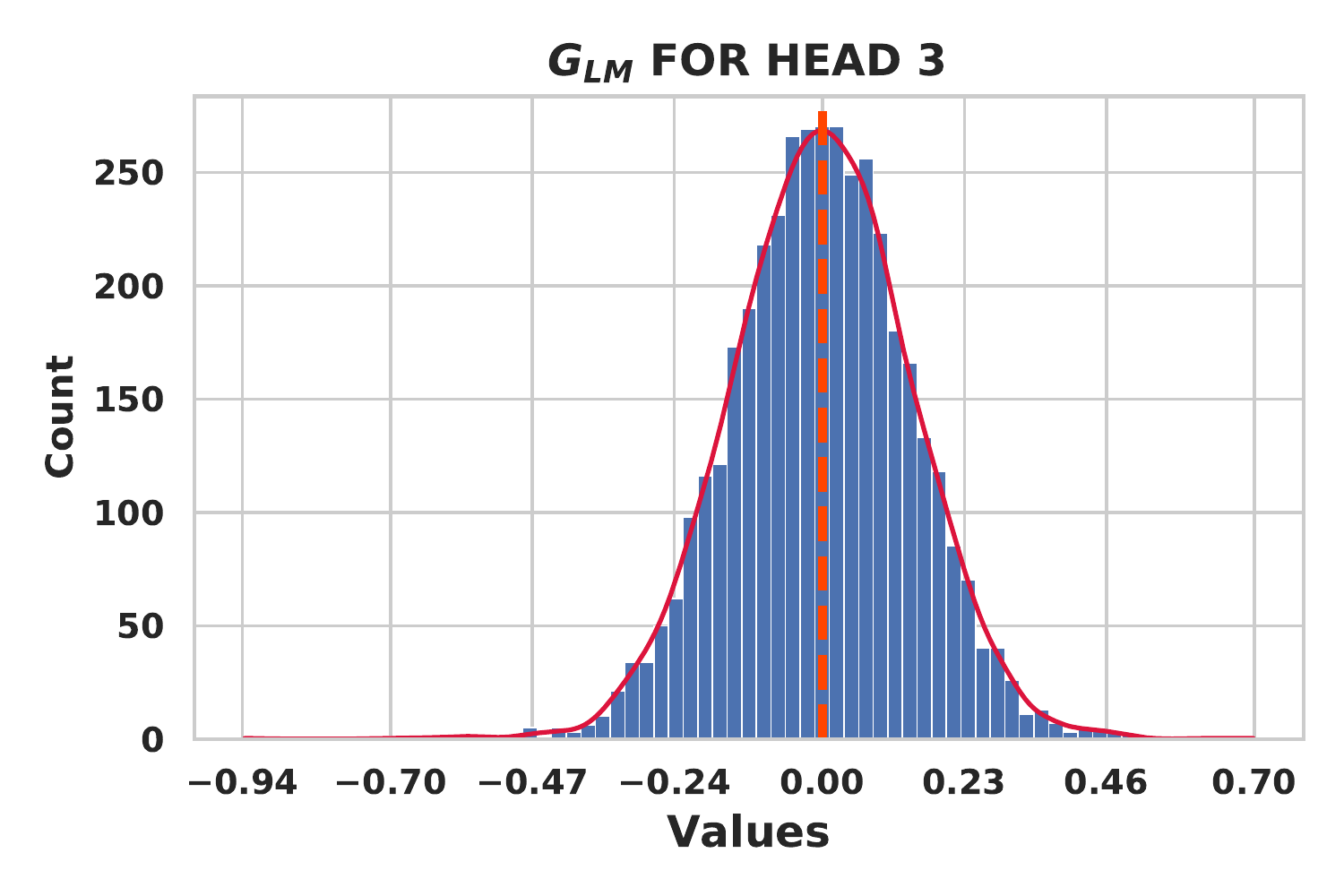}

\end{subfigure}
\hfill
\begin{subfigure}[b]{0.6\textwidth}
	\centering
	\includegraphics[width=1.1\textwidth]{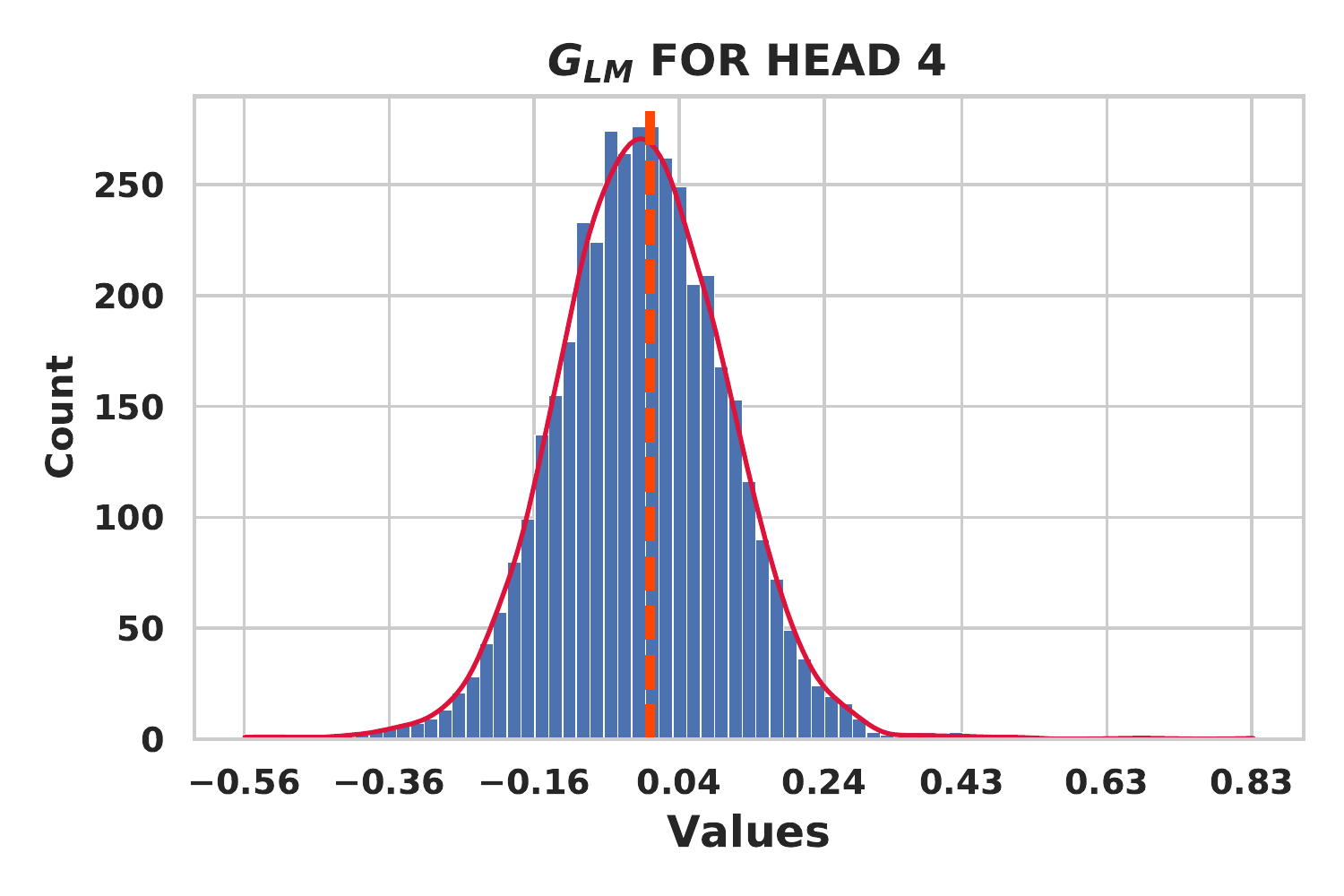}

\end{subfigure}
\centering
\begin{subfigure}[b]{0.6\textwidth}
	\centering
	\includegraphics[width=1.1\textwidth]{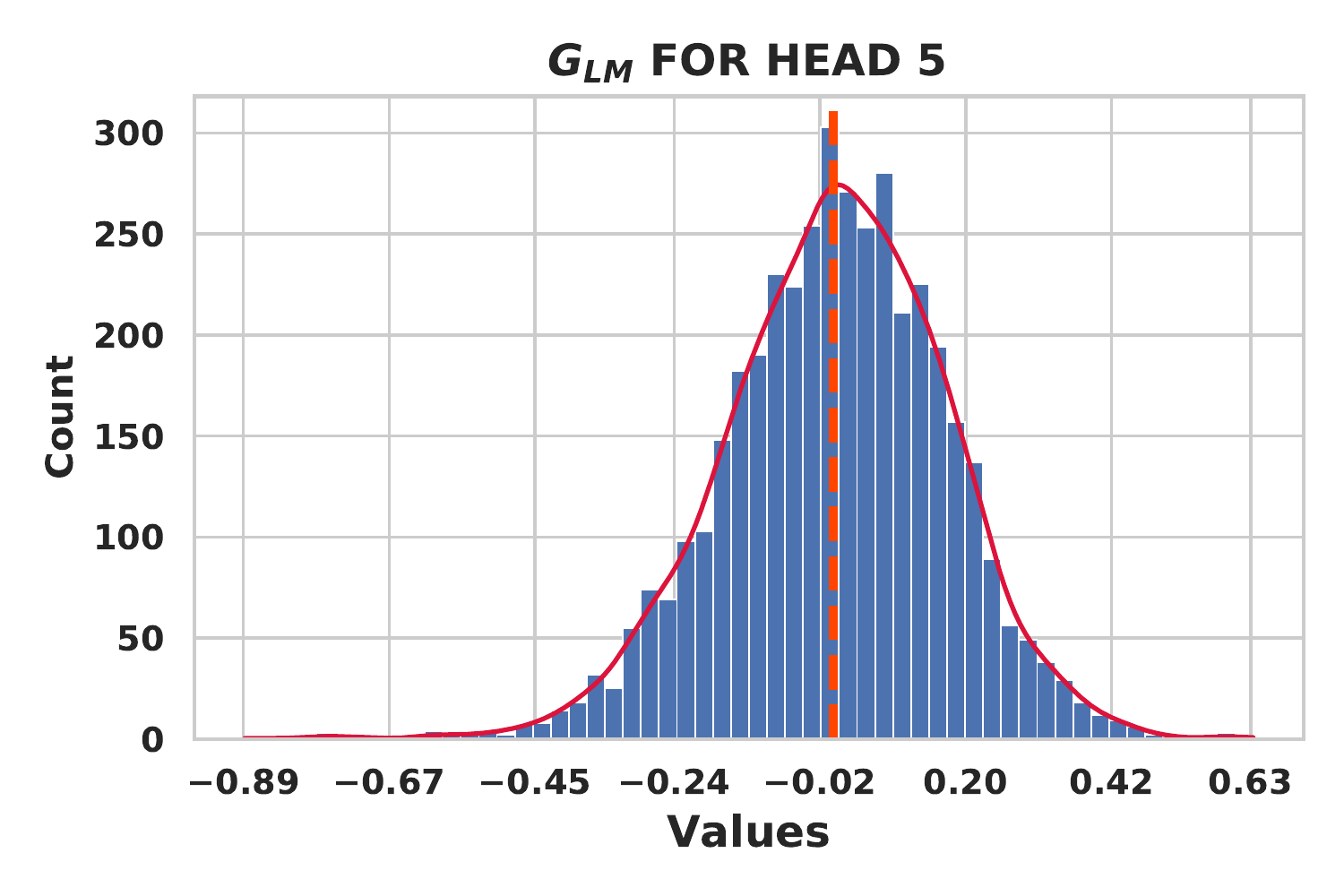}

\end{subfigure}
\hfill
\begin{subfigure}[b]{0.6\textwidth}
	\centering
	\includegraphics[width=1.1\textwidth]{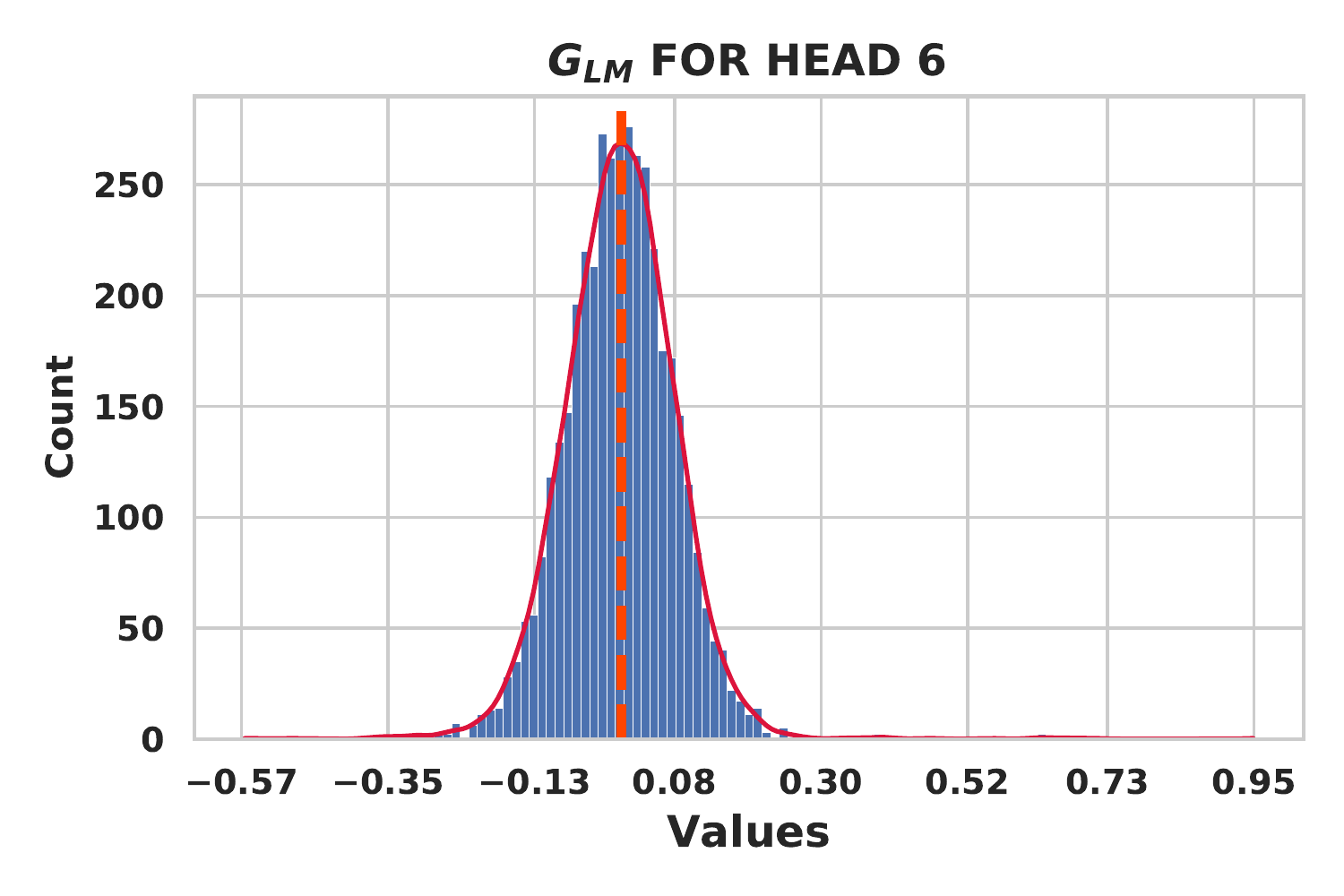}

\end{subfigure}
\hfill
\begin{subfigure}[b]{0.6\textwidth}
	\centering
	\includegraphics[width=1.1\textwidth]{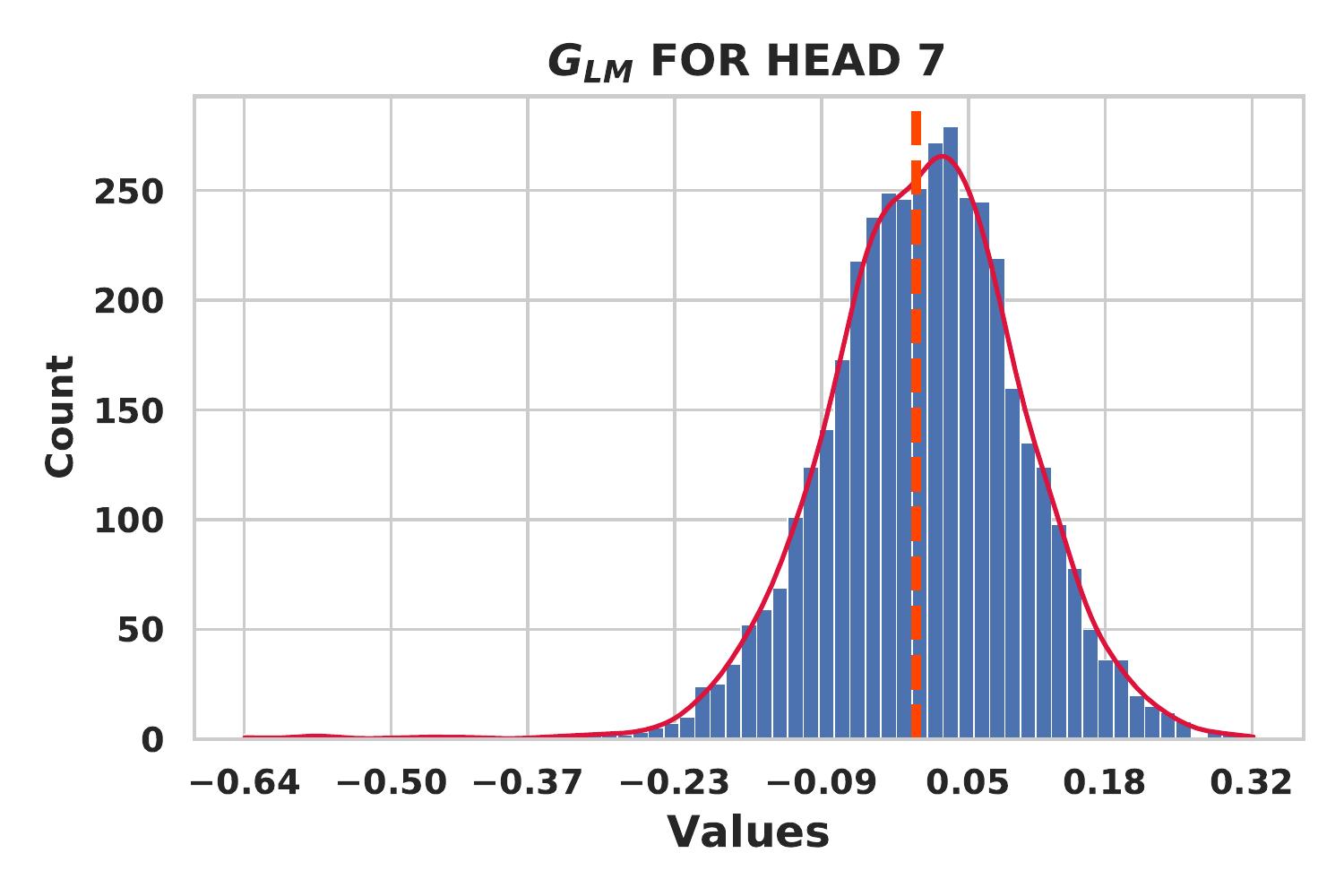}

\end{subfigure}
\hfill
\begin{subfigure}[b]{0.6\textwidth}
	\centering
	\includegraphics[width=1.1\textwidth]{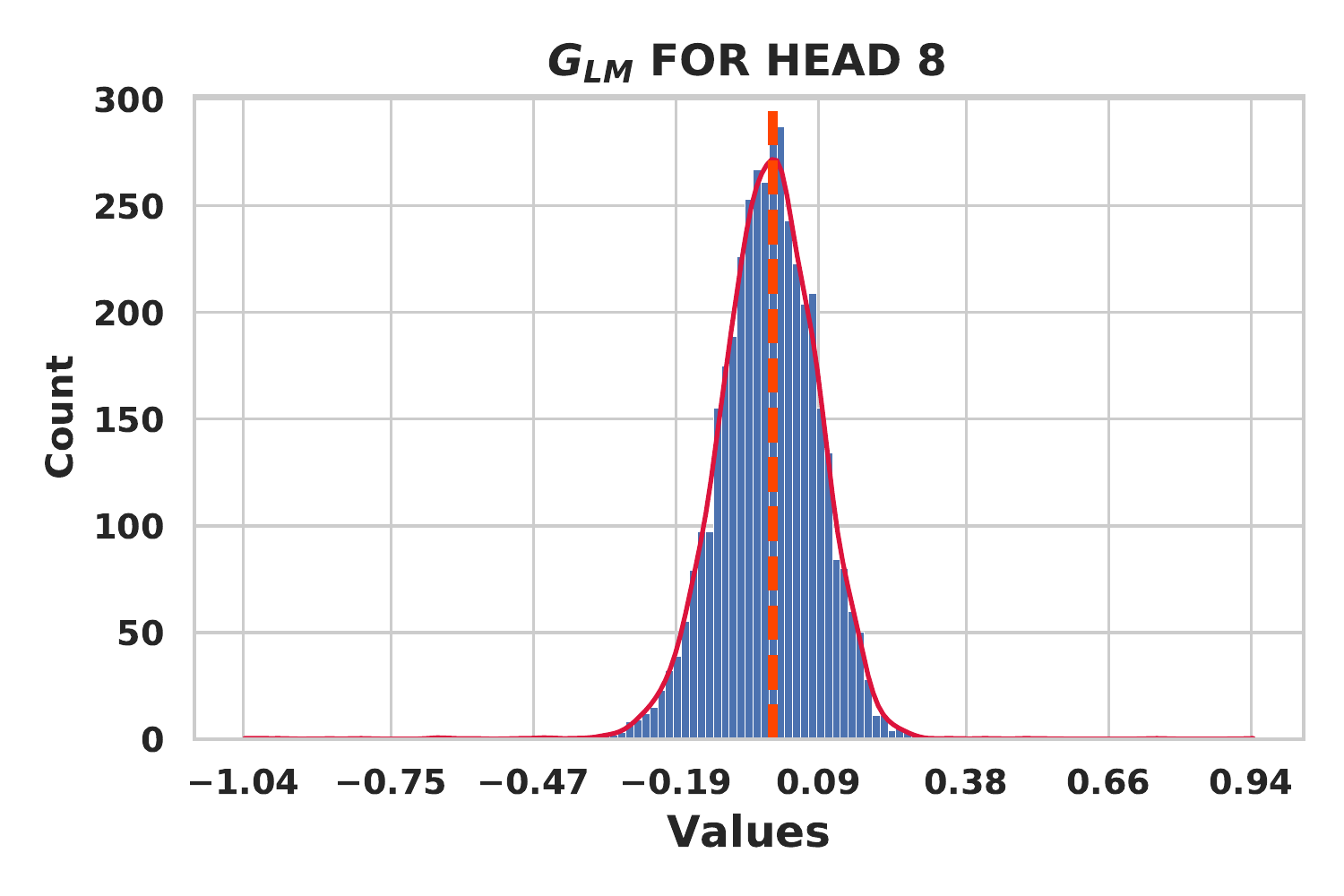}

\end{subfigure}
\caption{$\mG_{LM}$ histogram plots for all heads from SLM attention stage from graph transformer model \#2 for PT-EN translation task. Dashed line in orange marks zero value.}
\label{fig22apx}
\end{adjustwidth}
\end{figure}


\thispagestyle{headings}
\begin{figure}
\begin{adjustwidth}{-5em}{-5em}
\centering
\begin{subfigure}[b]{0.6\textwidth}
	\centering
	\includegraphics[width=1.1\textwidth]{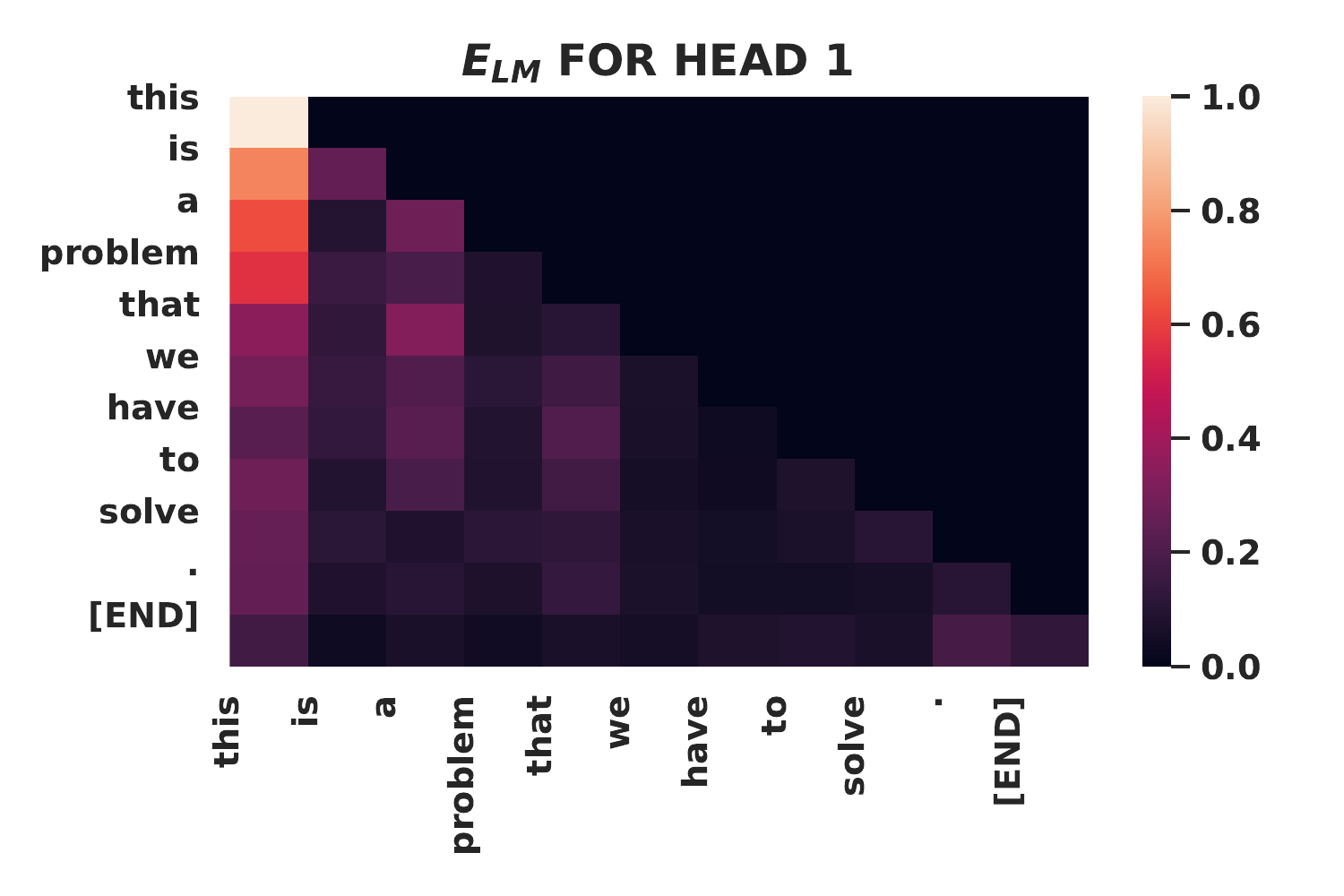}

\end{subfigure}
\hfill
\begin{subfigure}[b]{0.6\textwidth}
	\centering
	\includegraphics[width=1.1\textwidth]{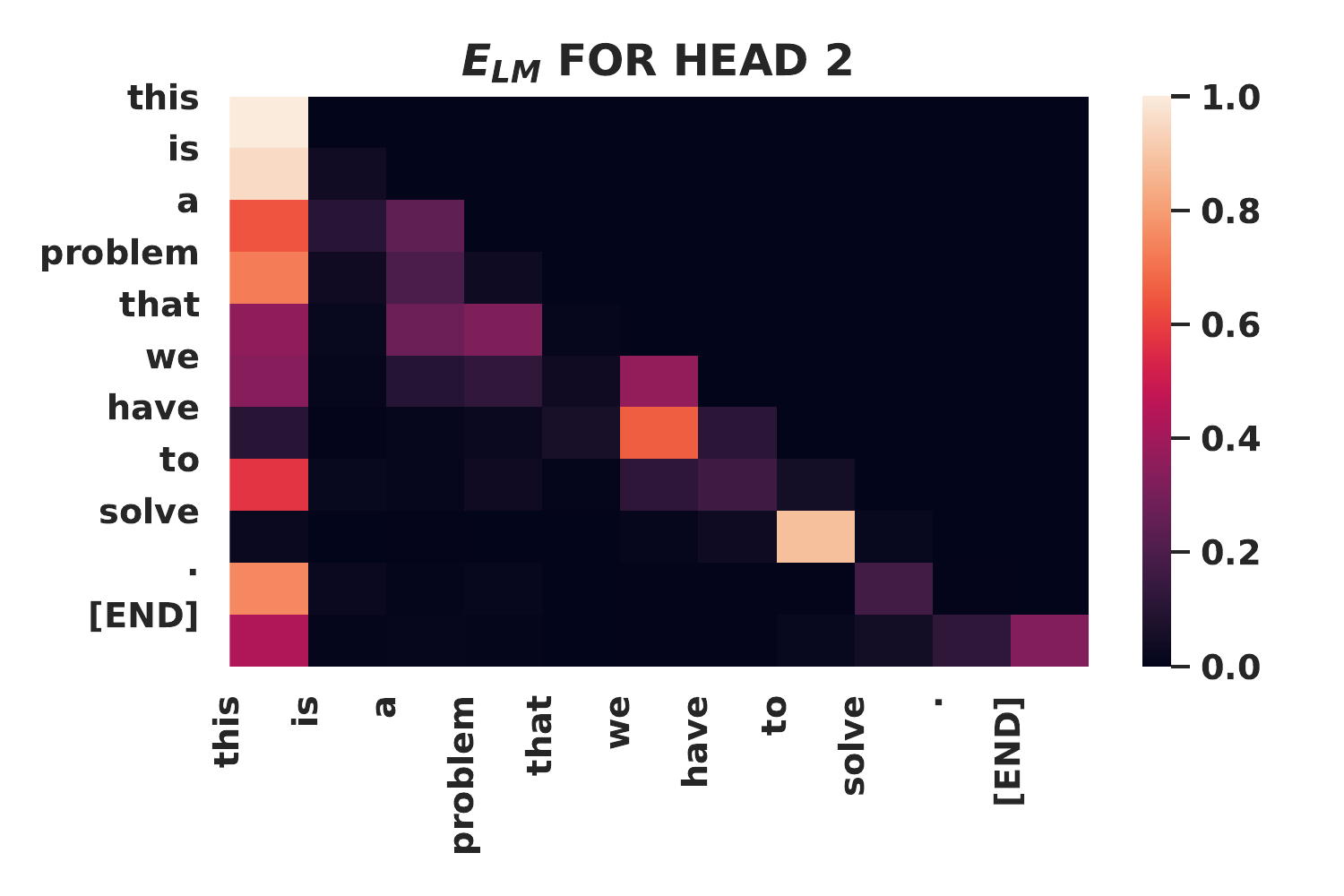}

\end{subfigure}
\hfill
\begin{subfigure}[b]{0.6\textwidth}
	\centering
	\includegraphics[width=1.1\textwidth]{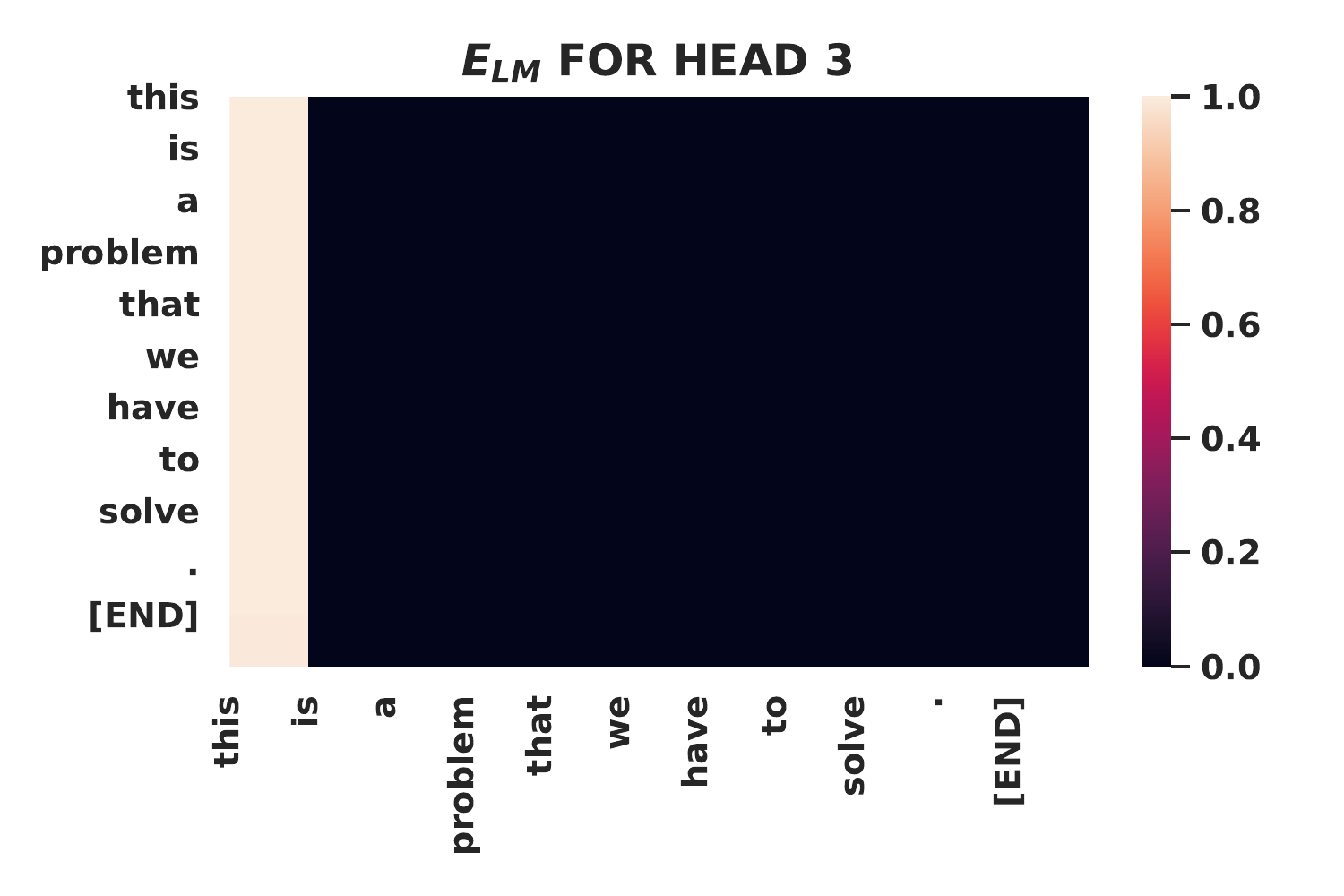}

\end{subfigure}
\hfill
\begin{subfigure}[b]{0.6\textwidth}
	\centering
	\includegraphics[width=1.1\textwidth]{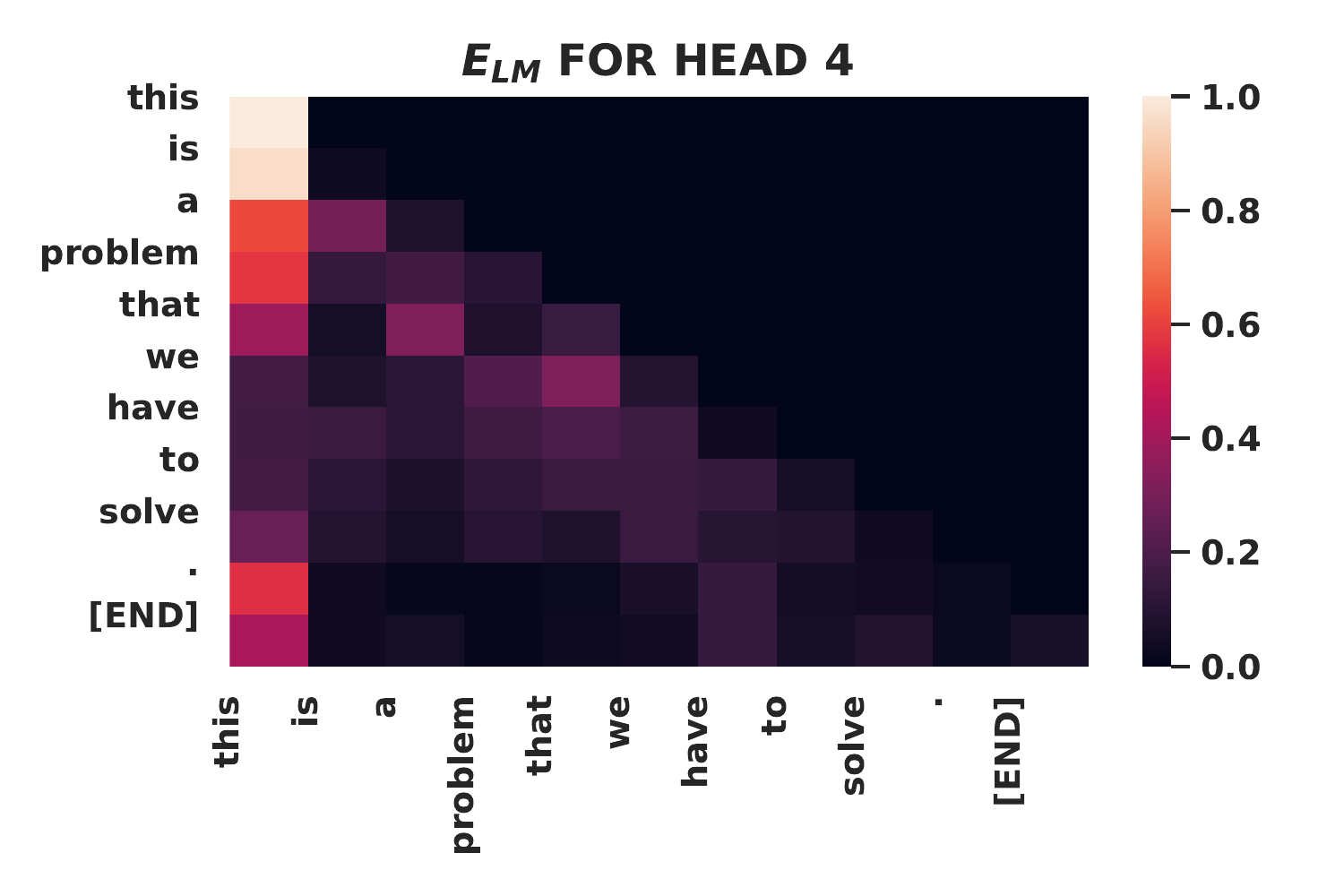}

\end{subfigure}
\centering
\begin{subfigure}[b]{0.6\textwidth}
	\centering
	\includegraphics[width=1.1\textwidth]{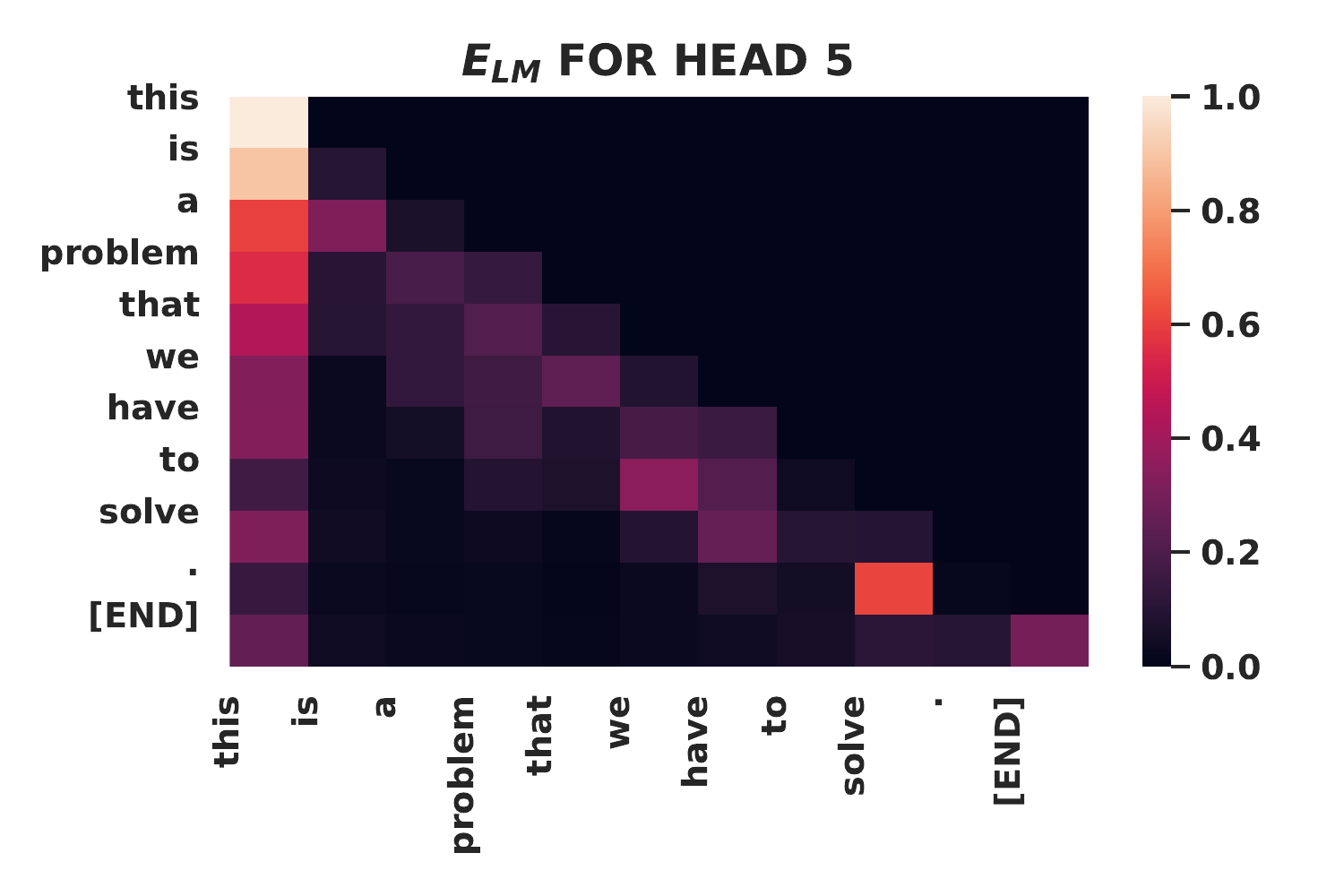}

\end{subfigure}
\hfill
\begin{subfigure}[b]{0.6\textwidth}
	\centering
	\includegraphics[width=1.1\textwidth]{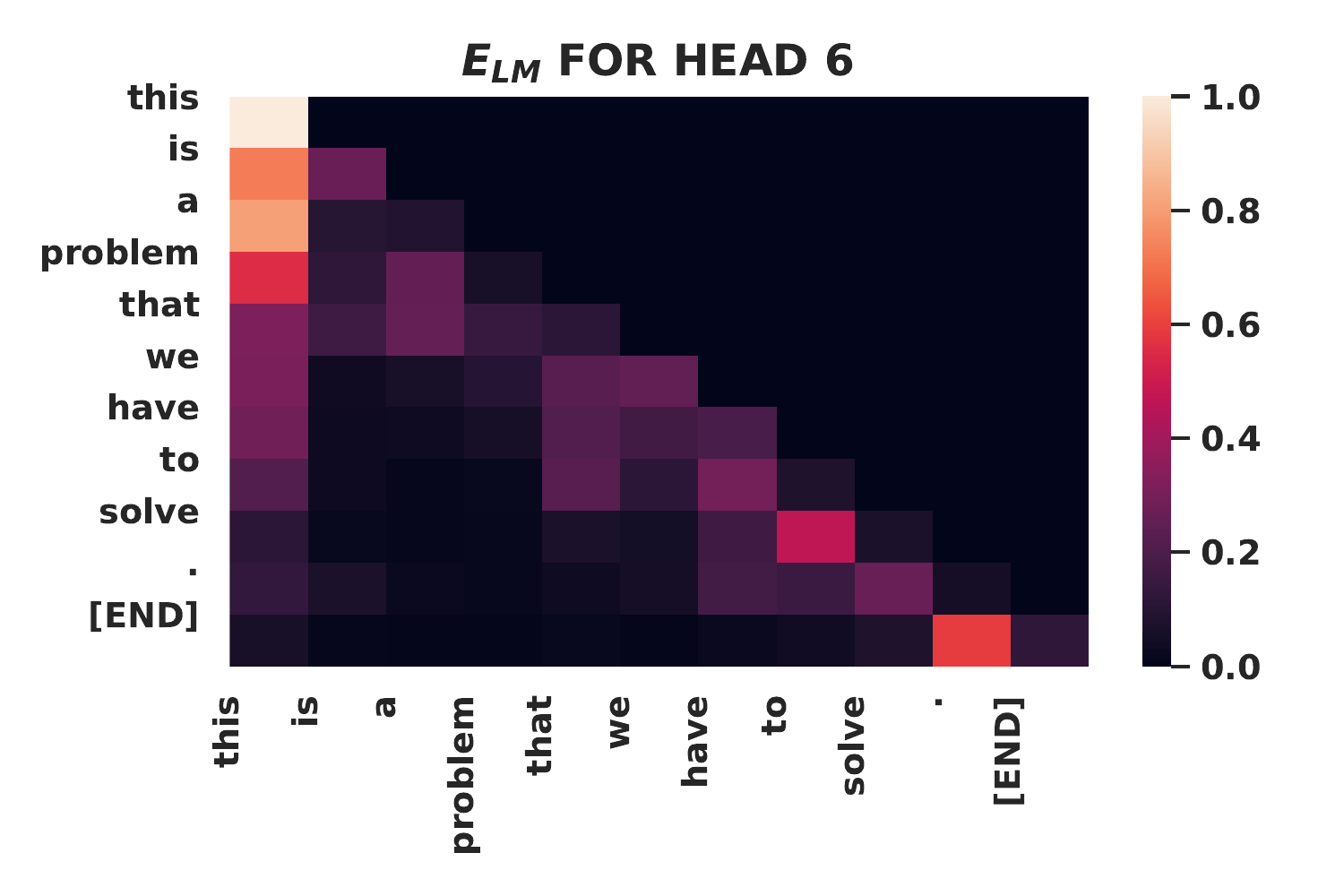}

\end{subfigure}
\hfill
\begin{subfigure}[b]{0.6\textwidth}
	\centering
	\includegraphics[width=1.1\textwidth]{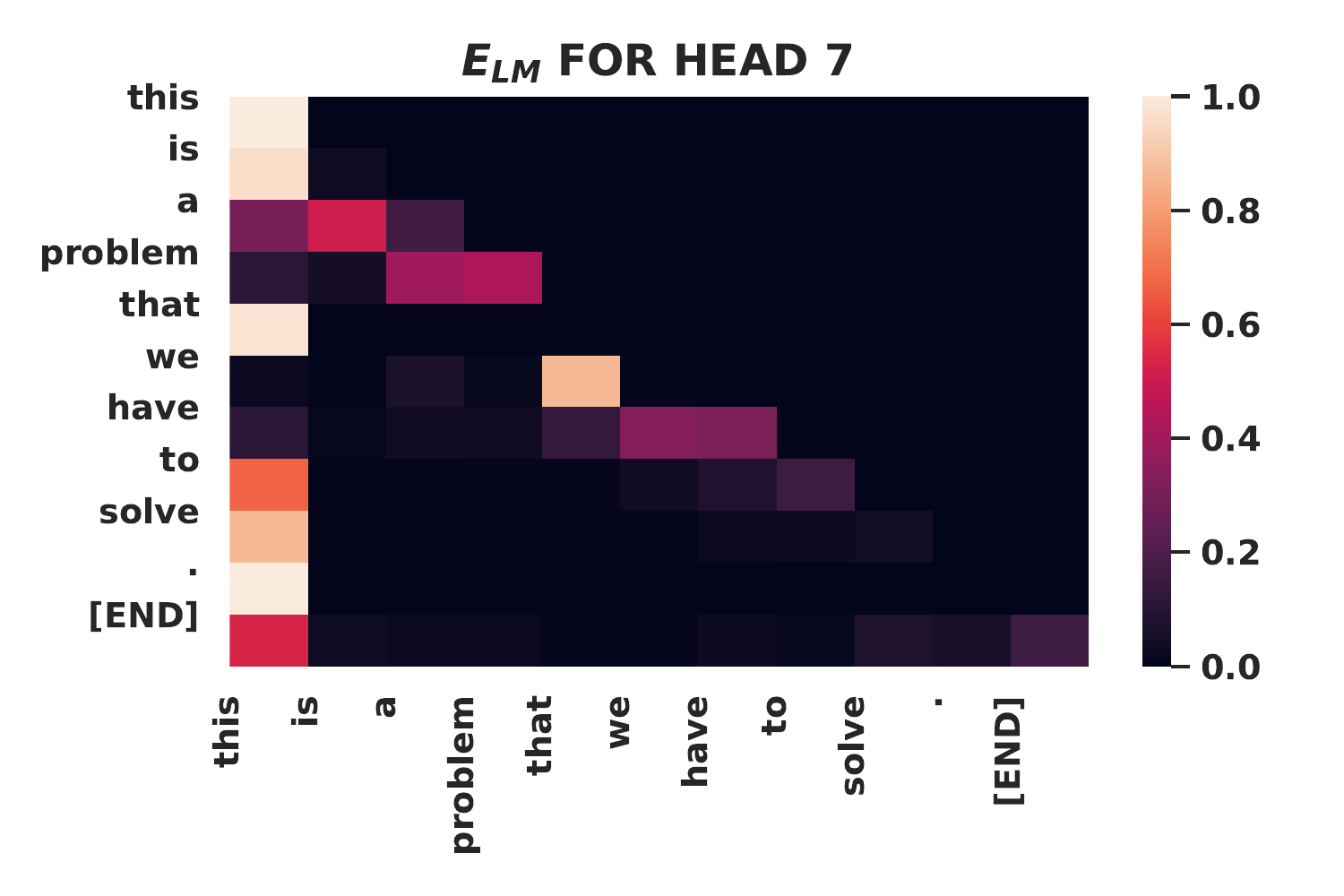}

\end{subfigure}
\hfill
\begin{subfigure}[b]{0.6\textwidth}
	\centering
	\includegraphics[width=1.1\textwidth]{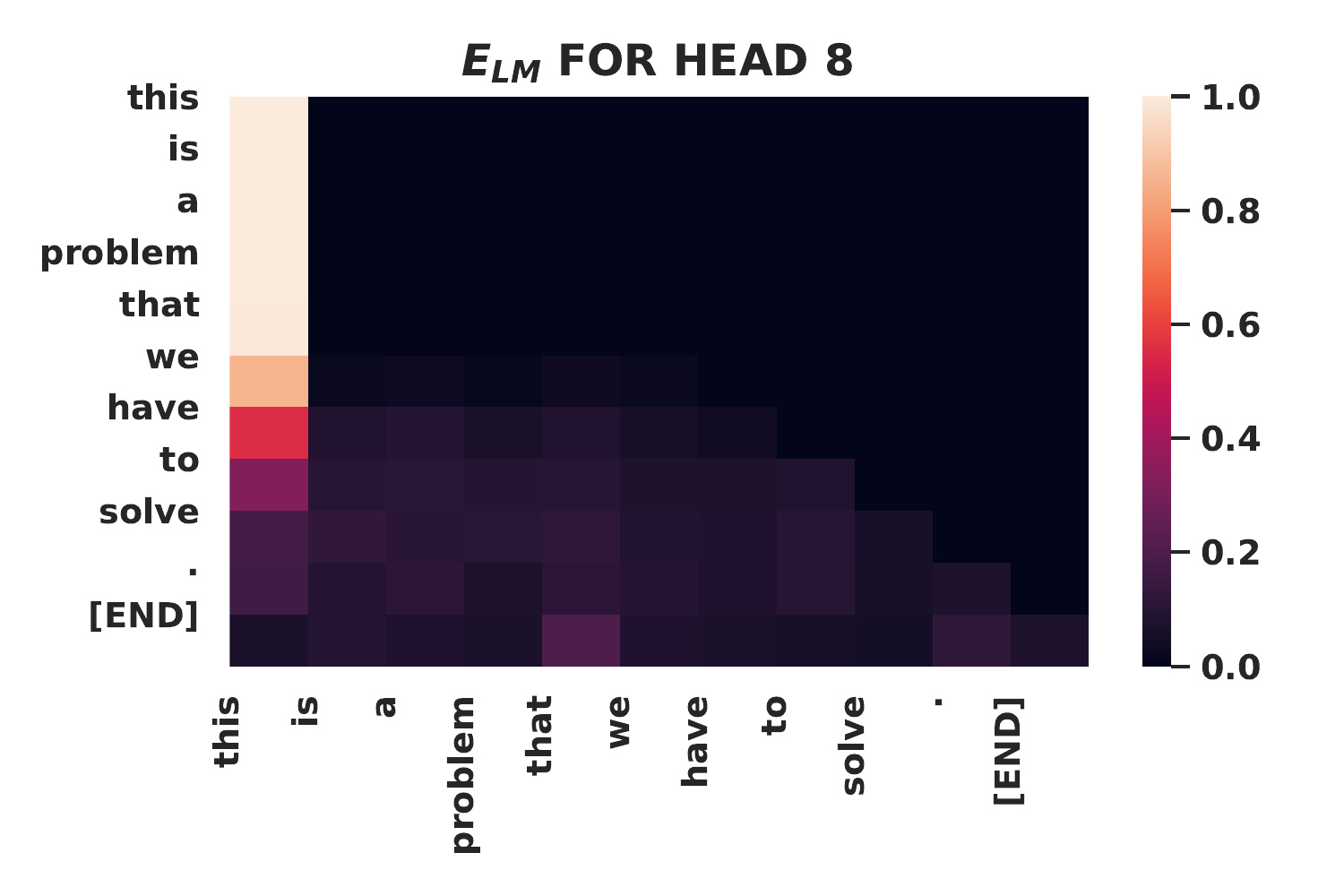}

\end{subfigure}
\caption{$\mE_{LM}$ heatmap plots for all heads from TLM attention stage from graph transformer model \#2 for PT-EN translation task.}
\label{fig23apx}
\end{adjustwidth}
\end{figure}  

\clearpage
\thispagestyle{headings}
\begin{figure}
\begin{adjustwidth}{-5em}{-5em}
\centering
\begin{subfigure}[b]{0.6\textwidth}
	\centering
	\includegraphics[width=1.1\textwidth]{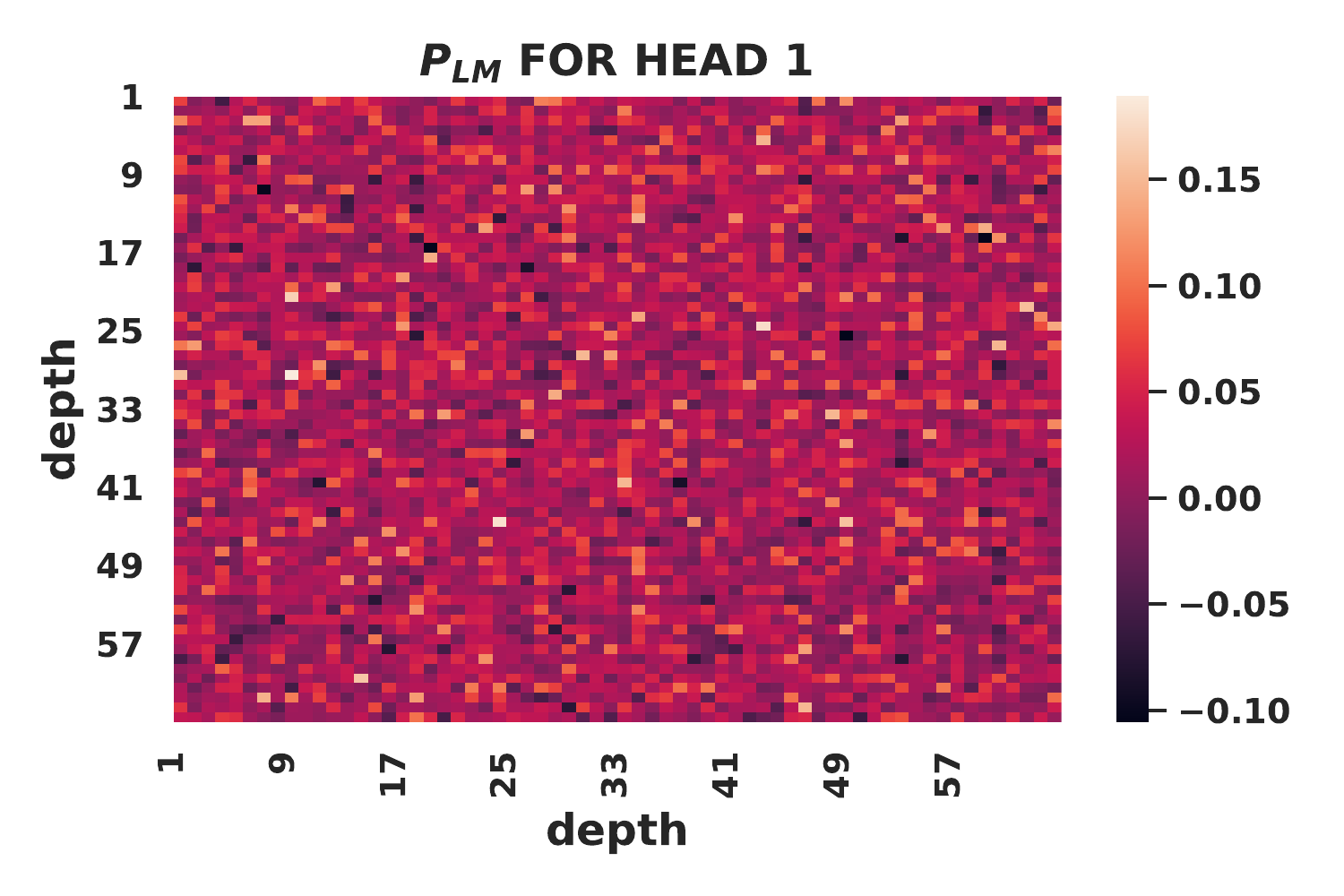}

\end{subfigure}
\hfill
\begin{subfigure}[b]{0.6\textwidth}
	\centering
	\includegraphics[width=1.1\textwidth]{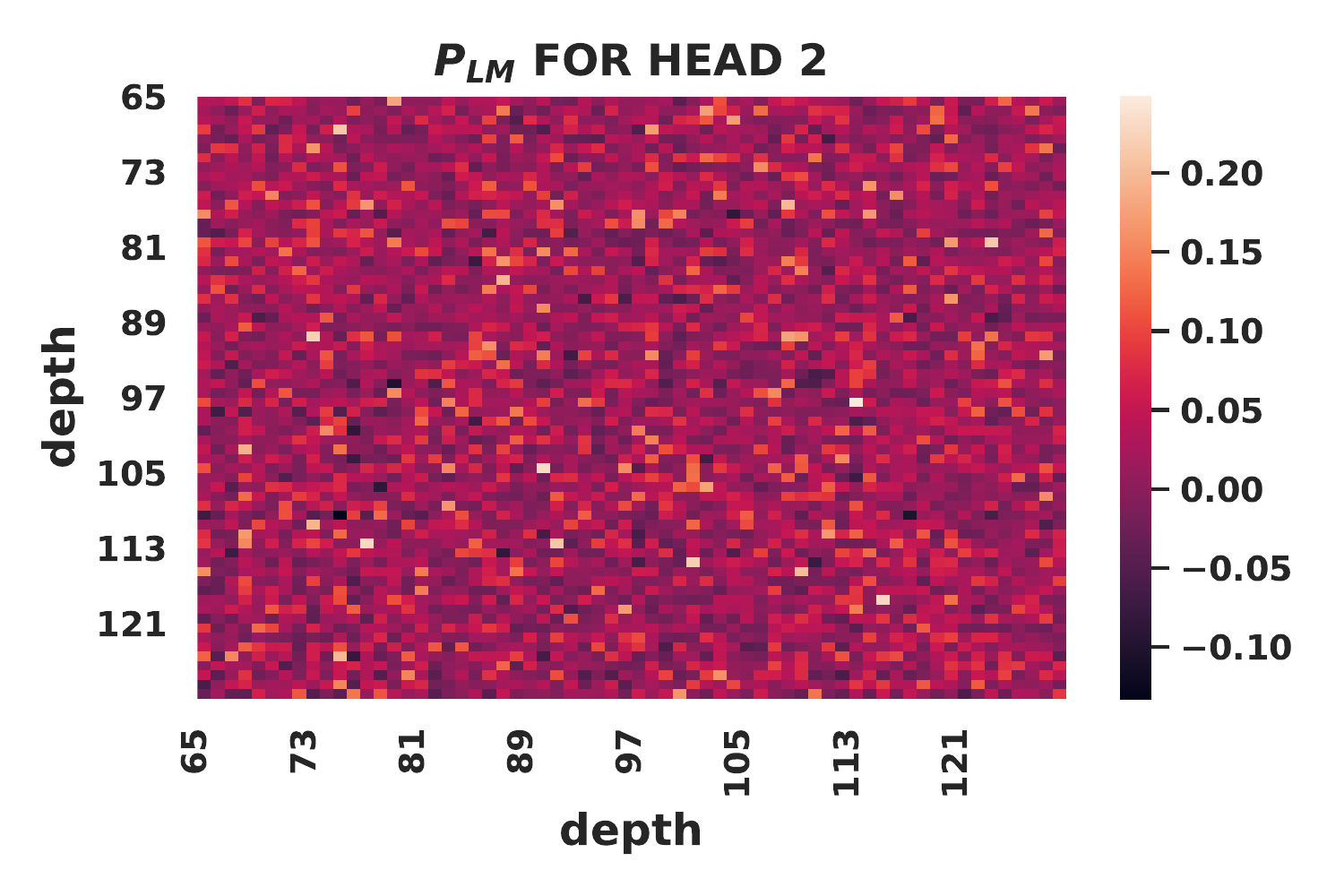}

\end{subfigure}
\hfill
\begin{subfigure}[b]{0.6\textwidth}
	\centering
	\includegraphics[width=1.1\textwidth]{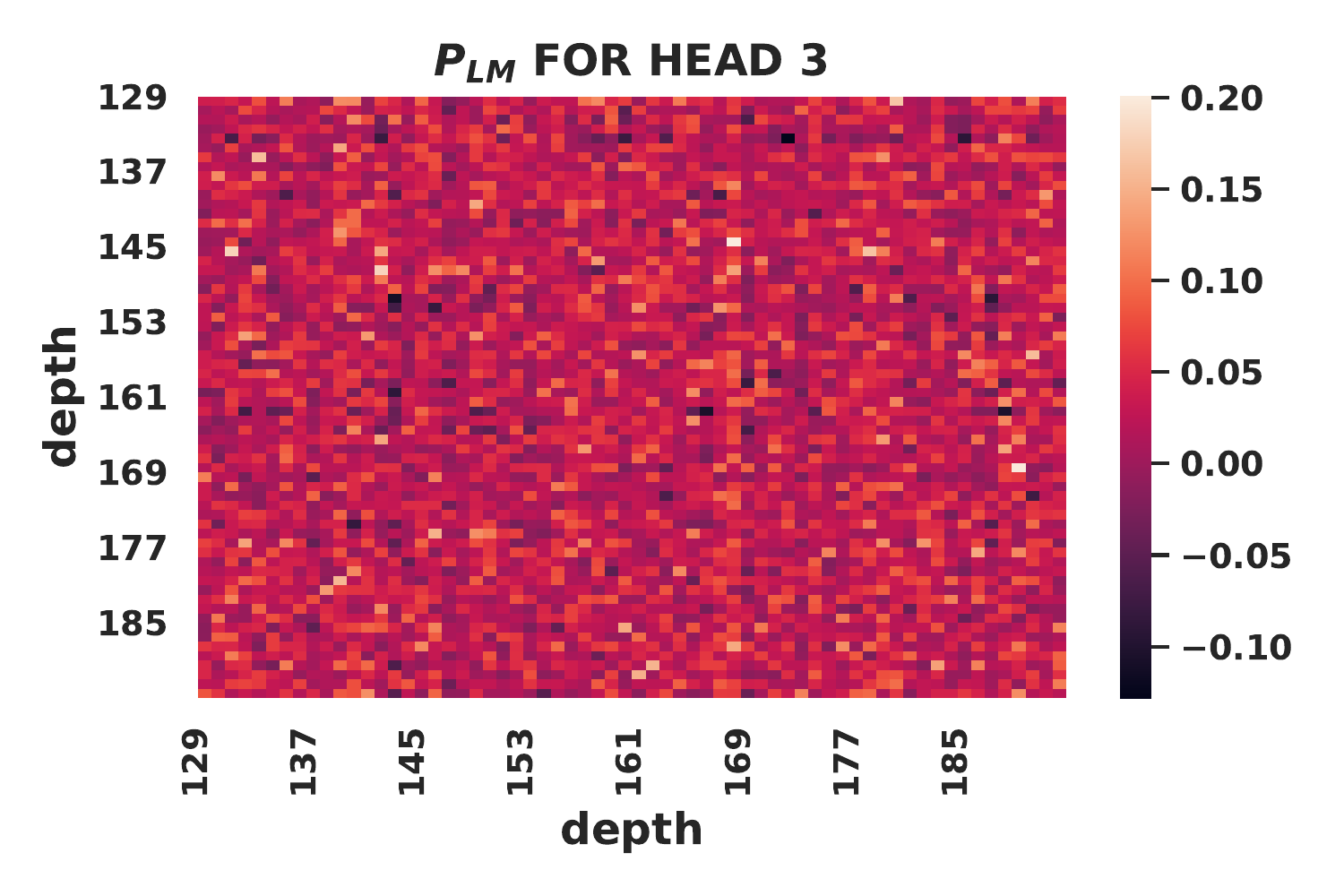}

\end{subfigure}
\hfill
\begin{subfigure}[b]{0.6\textwidth}
	\centering
	\includegraphics[width=1.1\textwidth]{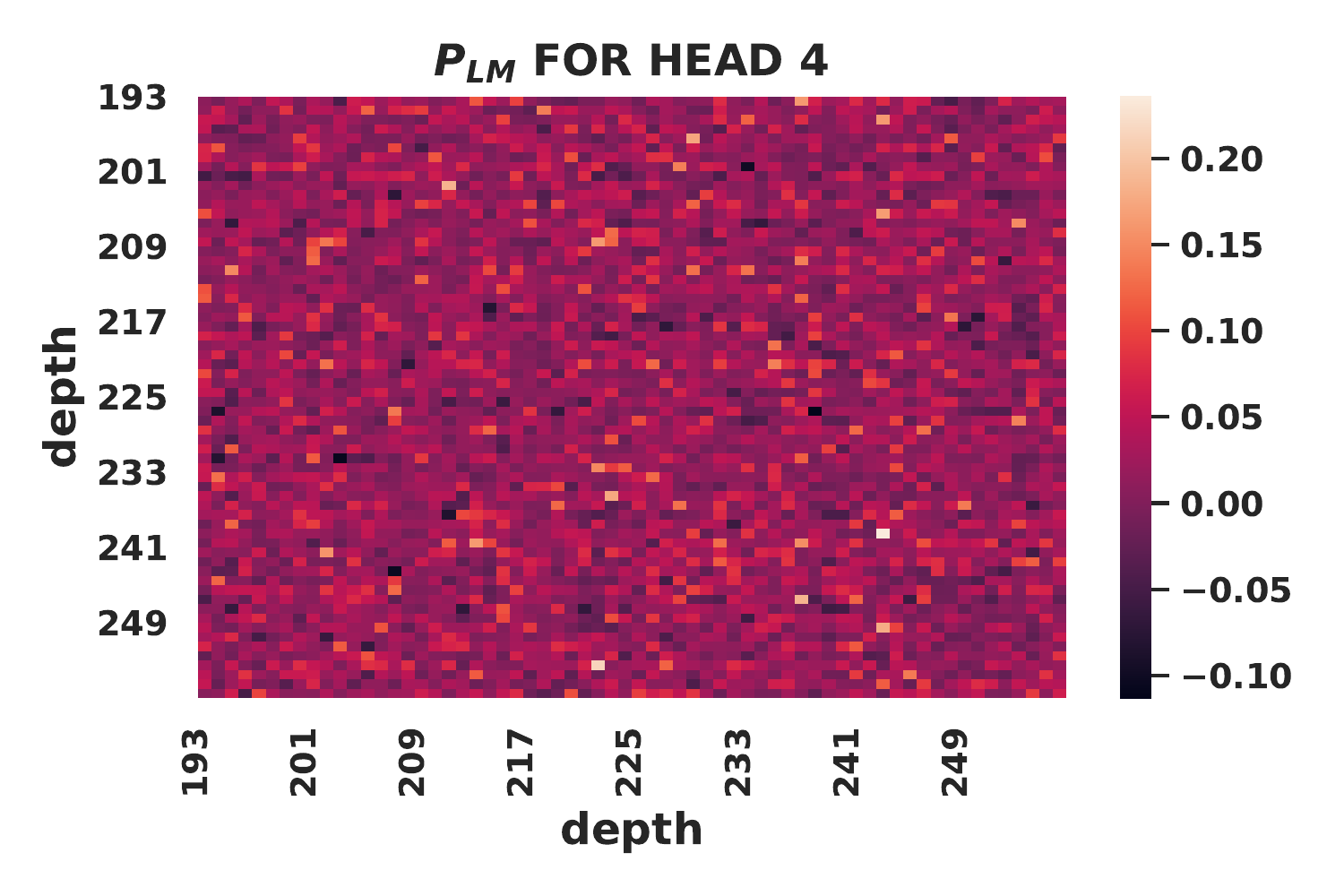}

\end{subfigure}
\centering
\begin{subfigure}[b]{0.6\textwidth}
	\centering
	\includegraphics[width=1.1\textwidth]{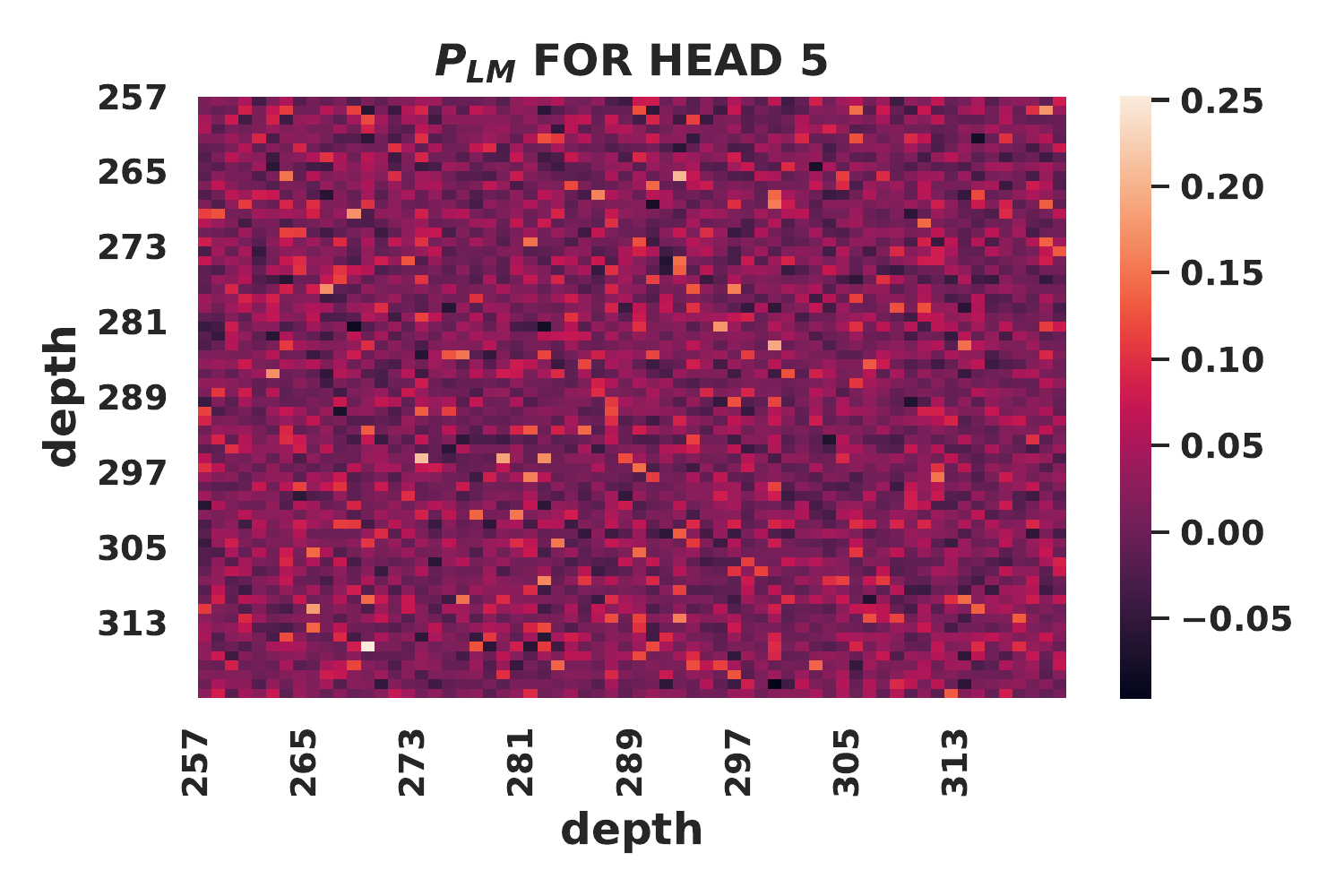}

\end{subfigure}
\hfill
\begin{subfigure}[b]{0.6\textwidth}
	\centering
	\includegraphics[width=1.1\textwidth]{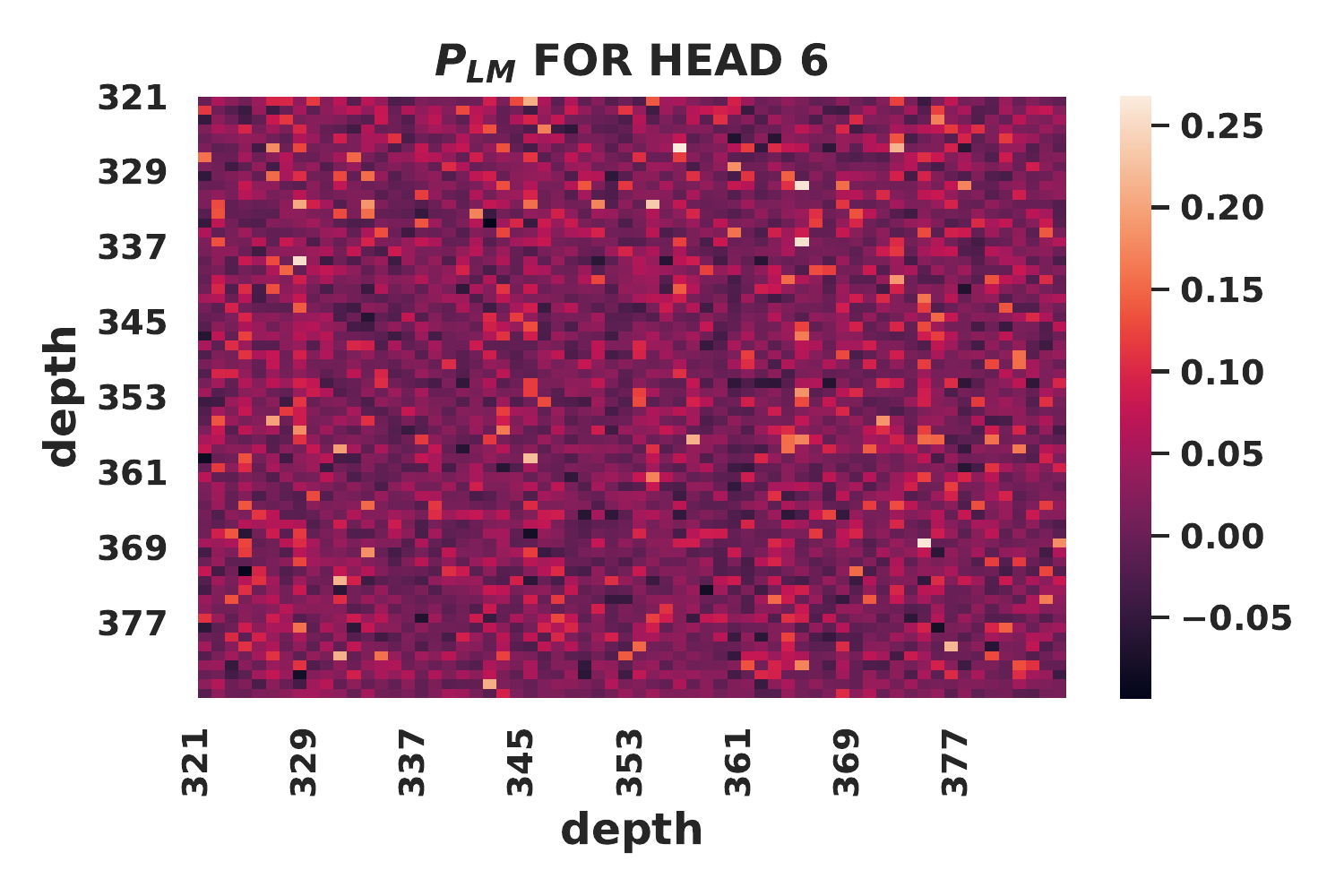}

\end{subfigure}
\hfill
\begin{subfigure}[b]{0.6\textwidth}
	\centering
	\includegraphics[width=1.1\textwidth]{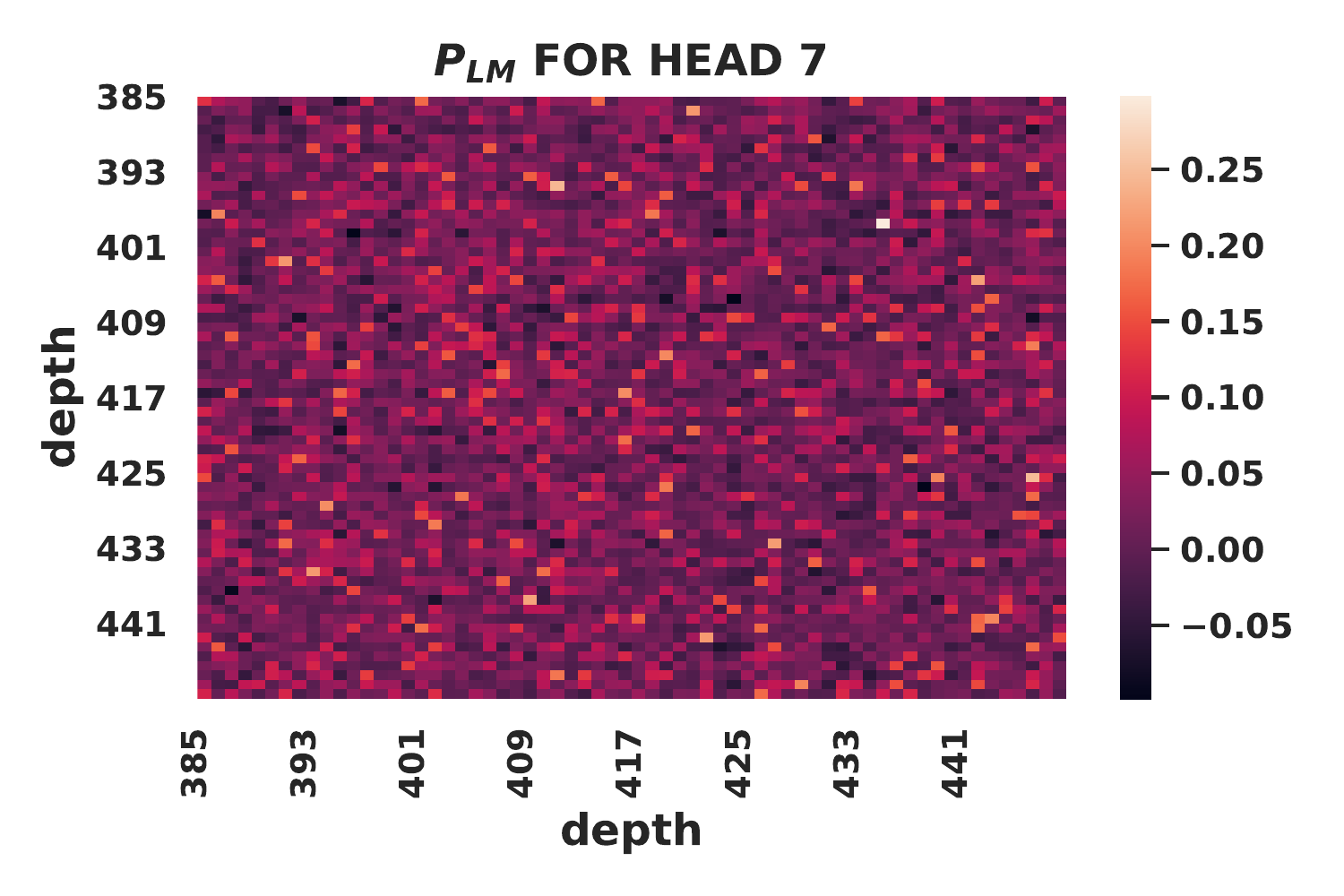}

\end{subfigure}
\hfill
\begin{subfigure}[b]{0.6\textwidth}
	\centering
	\includegraphics[width=1.1\textwidth]{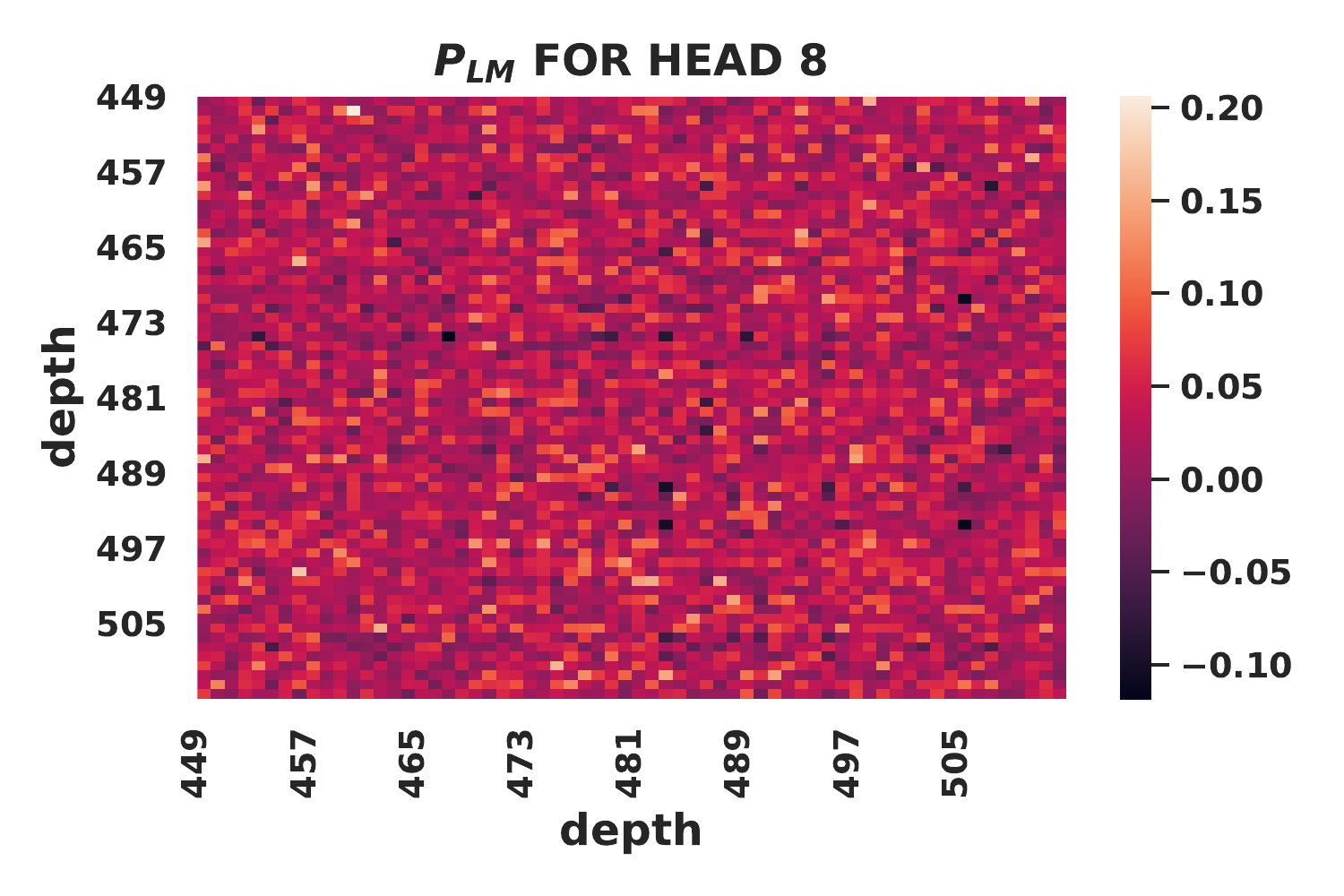}

\end{subfigure}
\caption{$\mP_{LM}$ heatmap plots for all heads from TLM attention stage from graph transformer model \#2 for PT-EN translation task.}
\label{fig24apx}
\end{adjustwidth}
\end{figure}

\clearpage
\thispagestyle{headings}

\begin{figure}
\begin{adjustwidth}{-5em}{-5em}
\centering
\begin{subfigure}[b]{0.6\textwidth}
	\centering
	\includegraphics[width=1.1\textwidth]{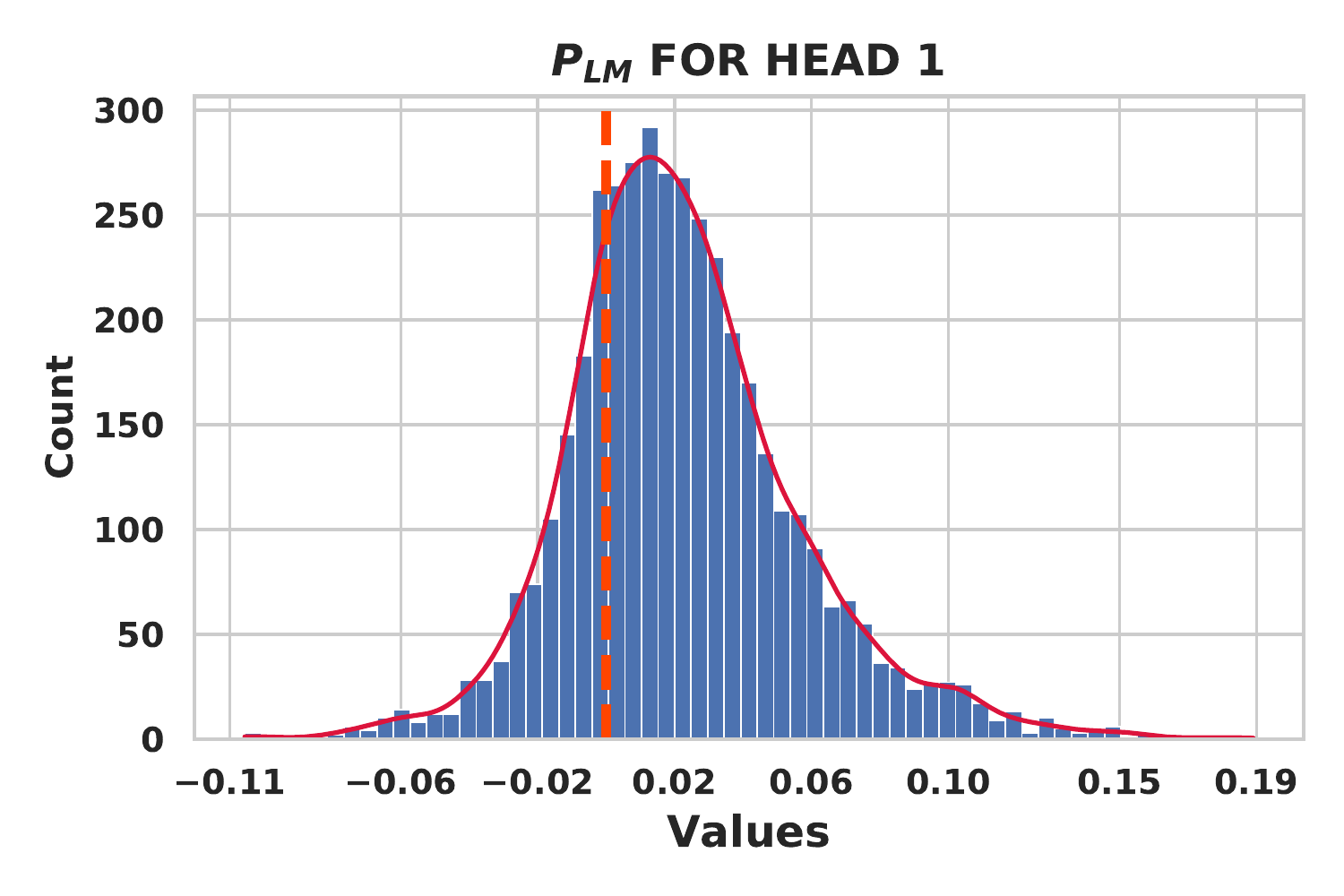}

\end{subfigure}
\hfill
\begin{subfigure}[b]{0.6\textwidth}
	\centering
	\includegraphics[width=1.1\textwidth]{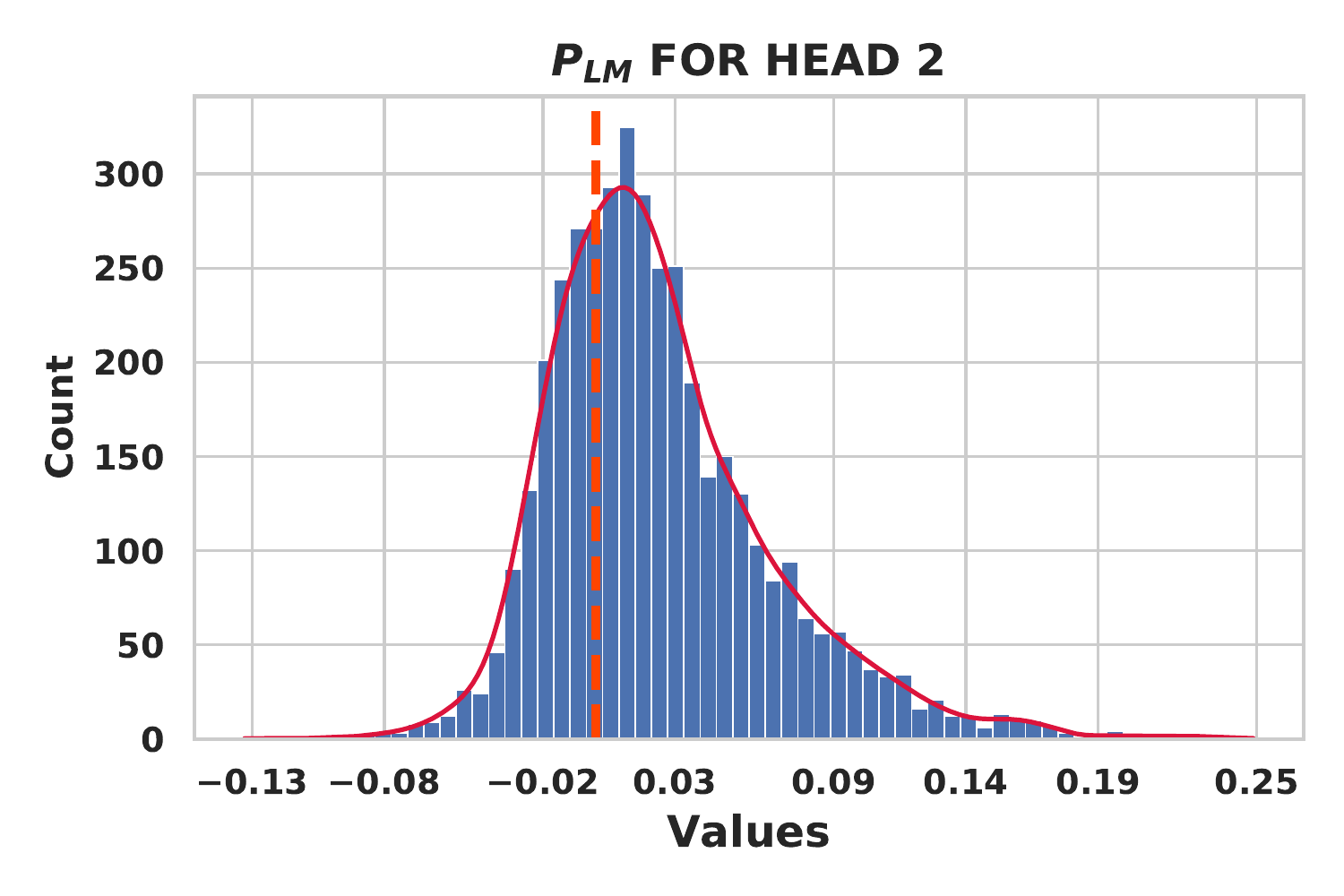}

\end{subfigure}
\hfill
\begin{subfigure}[b]{0.6\textwidth}
	\centering
	\includegraphics[width=1.1\textwidth]{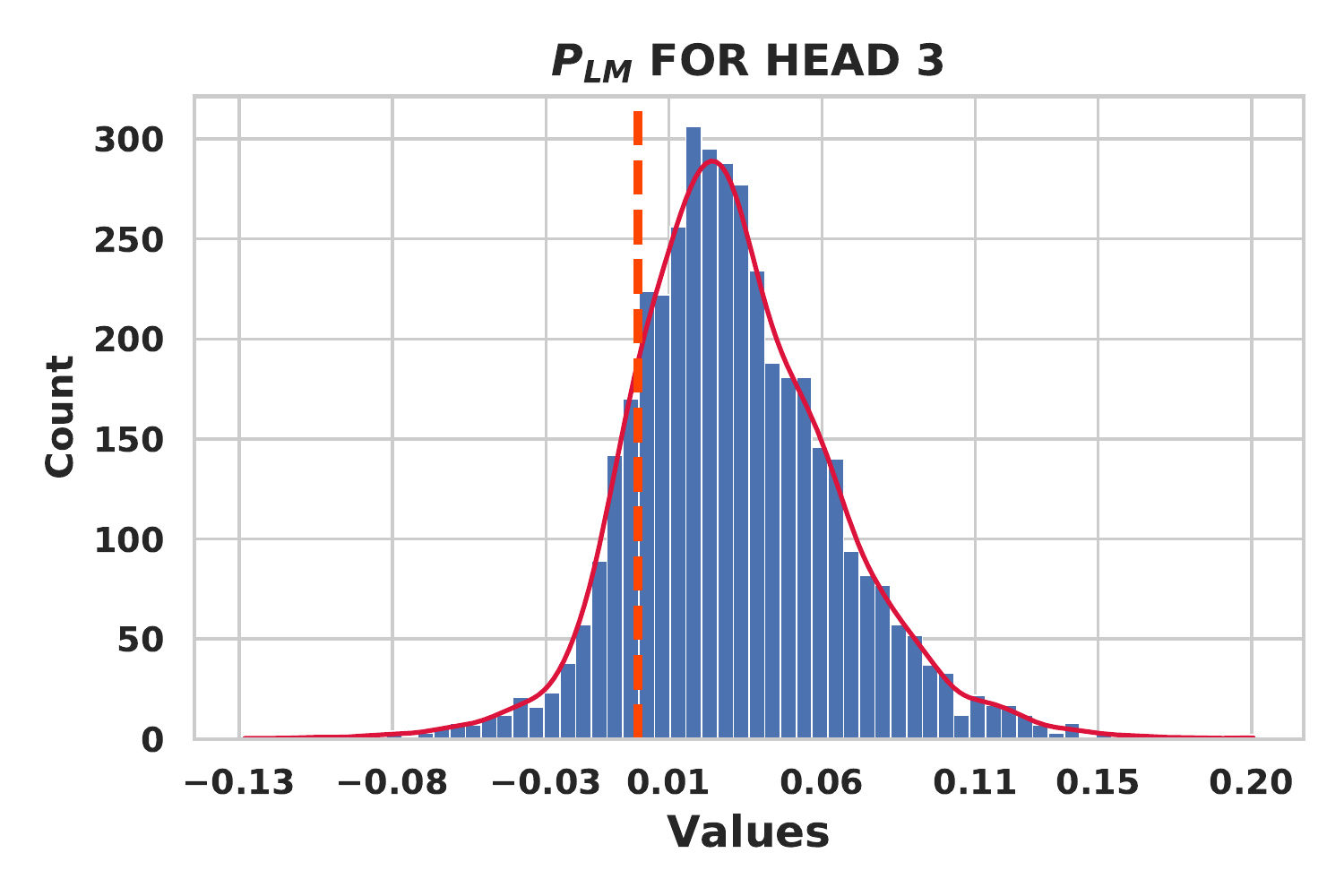}

\end{subfigure}
\hfill
\begin{subfigure}[b]{0.6\textwidth}
	\centering
	\includegraphics[width=1.1\textwidth]{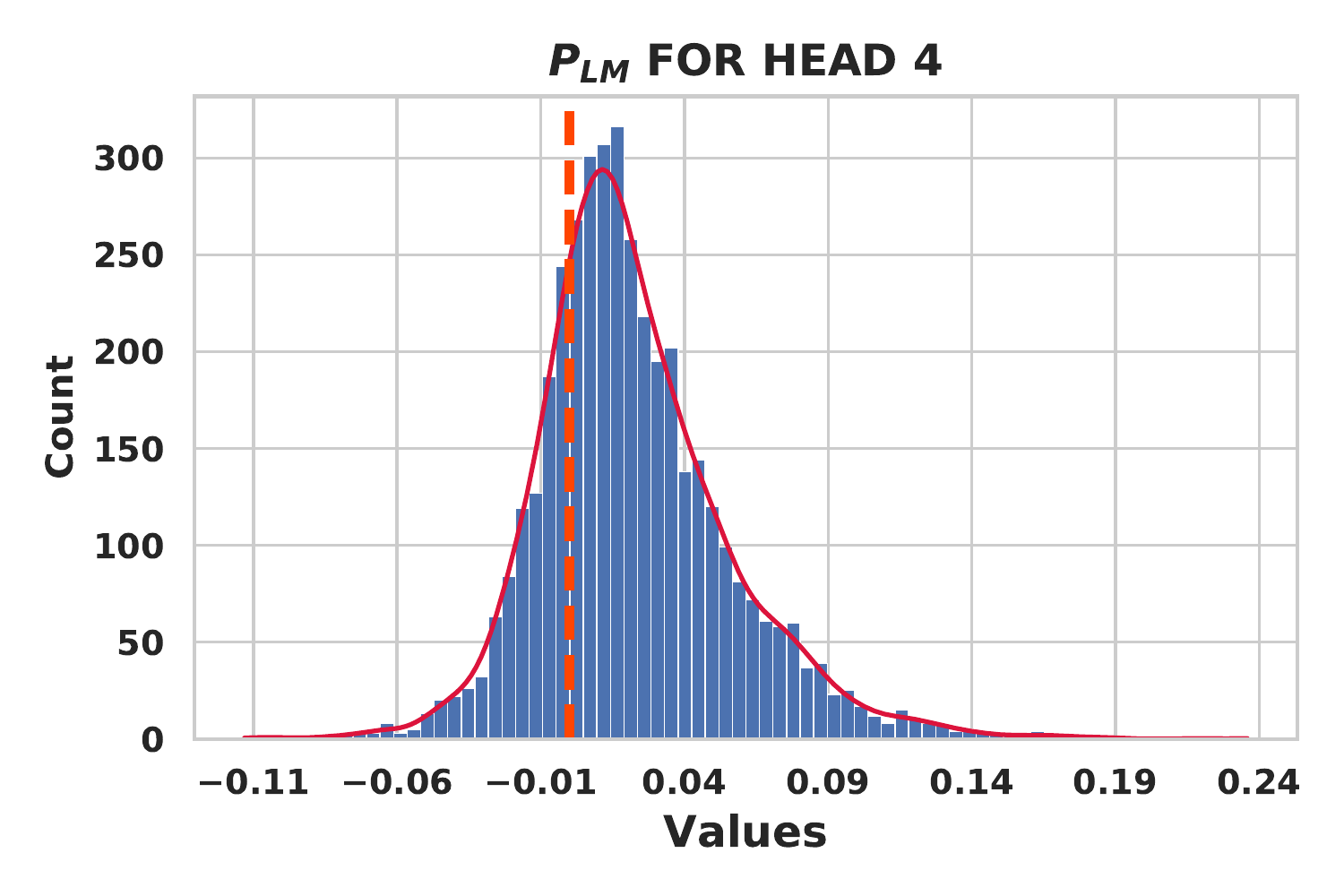}

\end{subfigure}
\centering
\begin{subfigure}[b]{0.6\textwidth}
	\centering
	\includegraphics[width=1.1\textwidth]{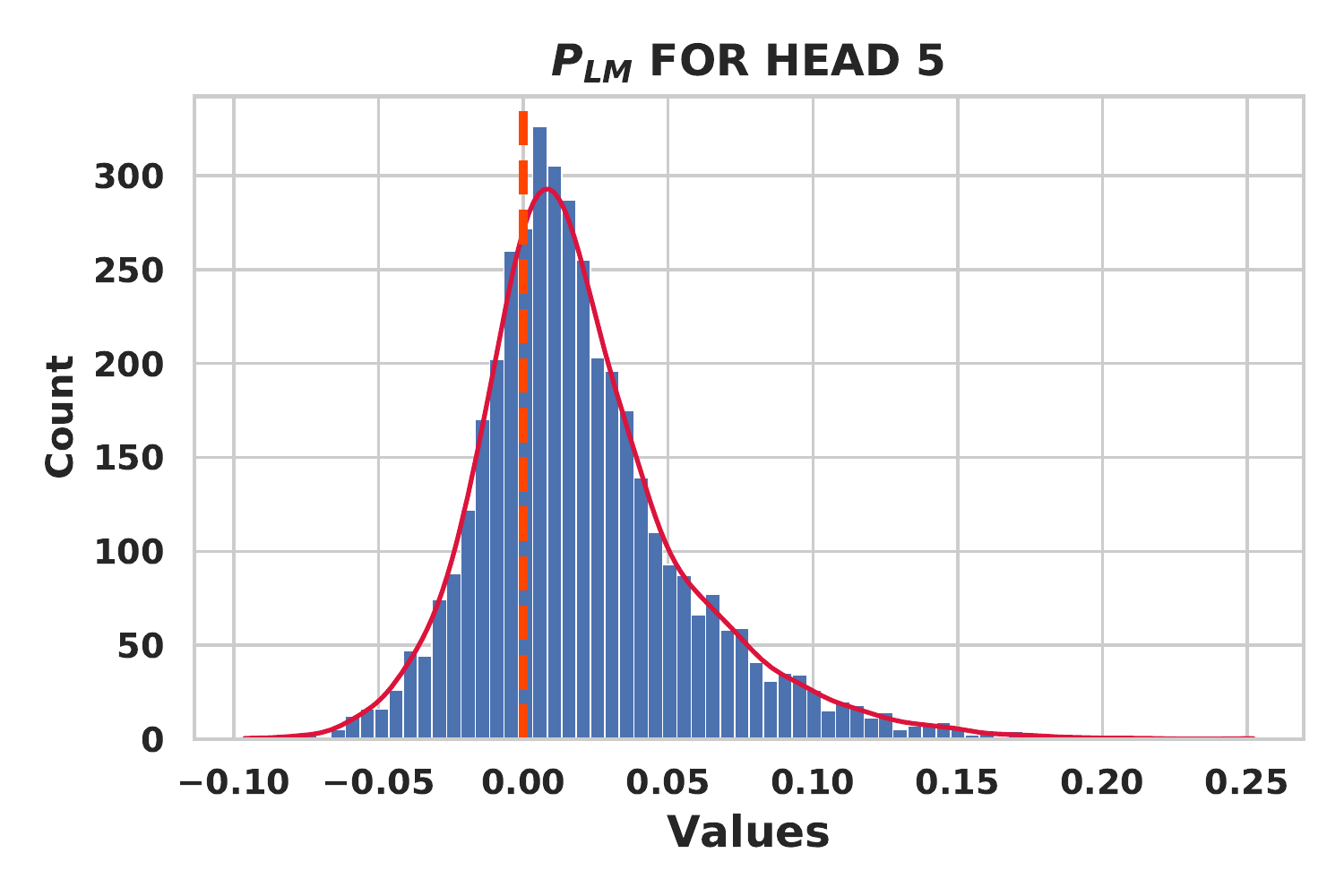}

\end{subfigure}
\hfill
\begin{subfigure}[b]{0.6\textwidth}
	\centering
	\includegraphics[width=1.1\textwidth]{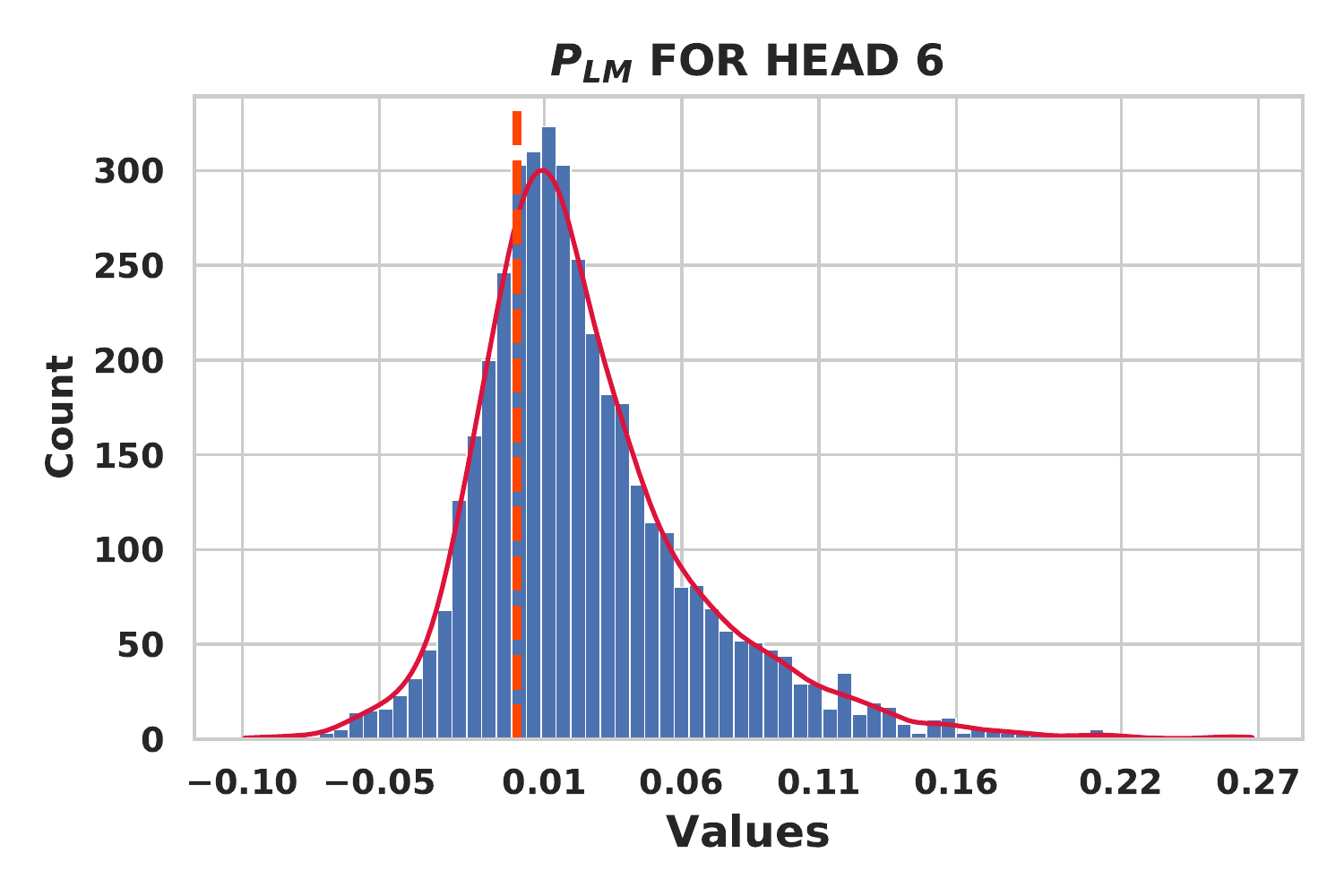}

\end{subfigure}
\hfill
\begin{subfigure}[b]{0.6\textwidth}
	\centering
	\includegraphics[width=1.1\textwidth]{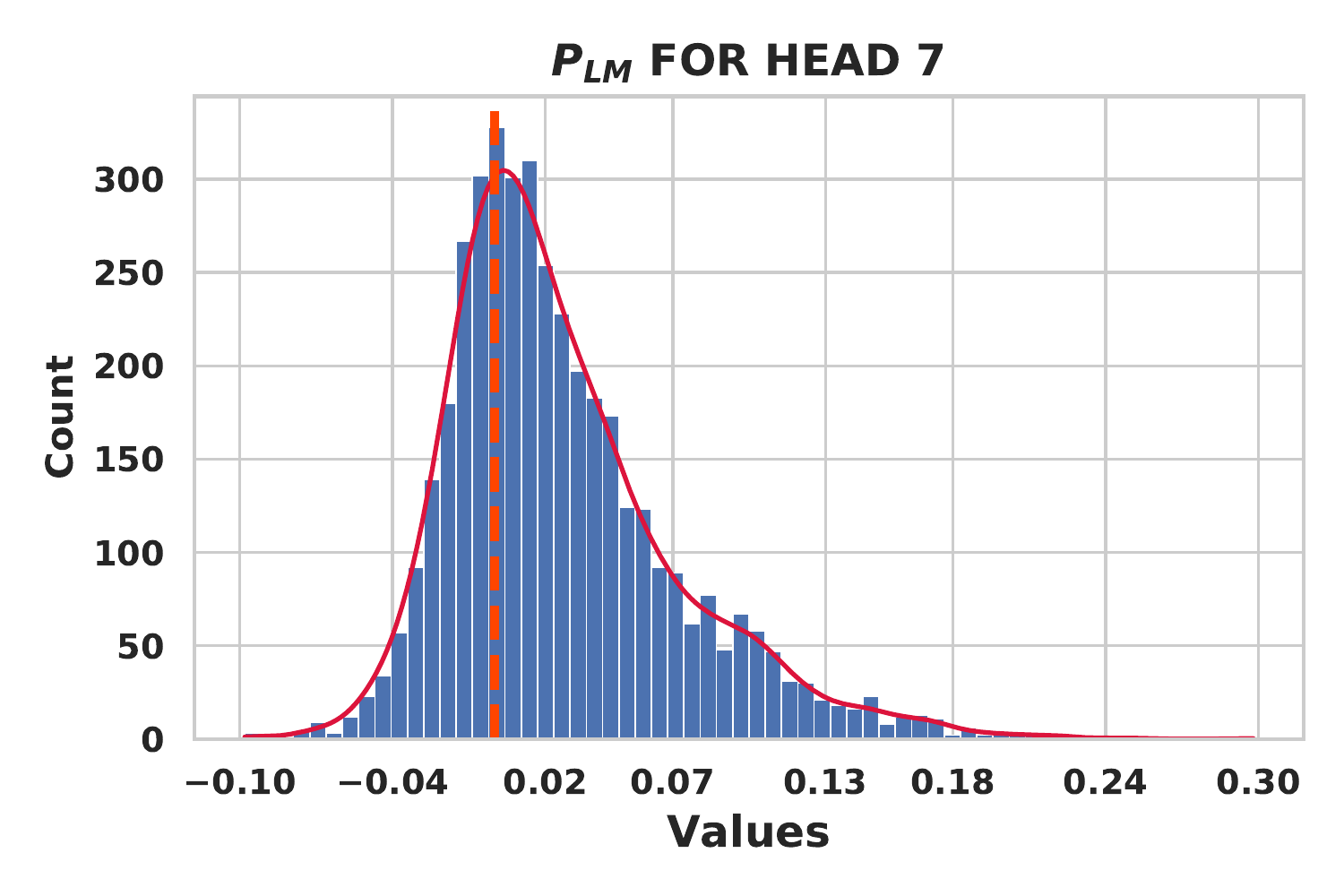}

\end{subfigure}
\hfill
\begin{subfigure}[b]{0.6\textwidth}
	\centering
	\includegraphics[width=1.1\textwidth]{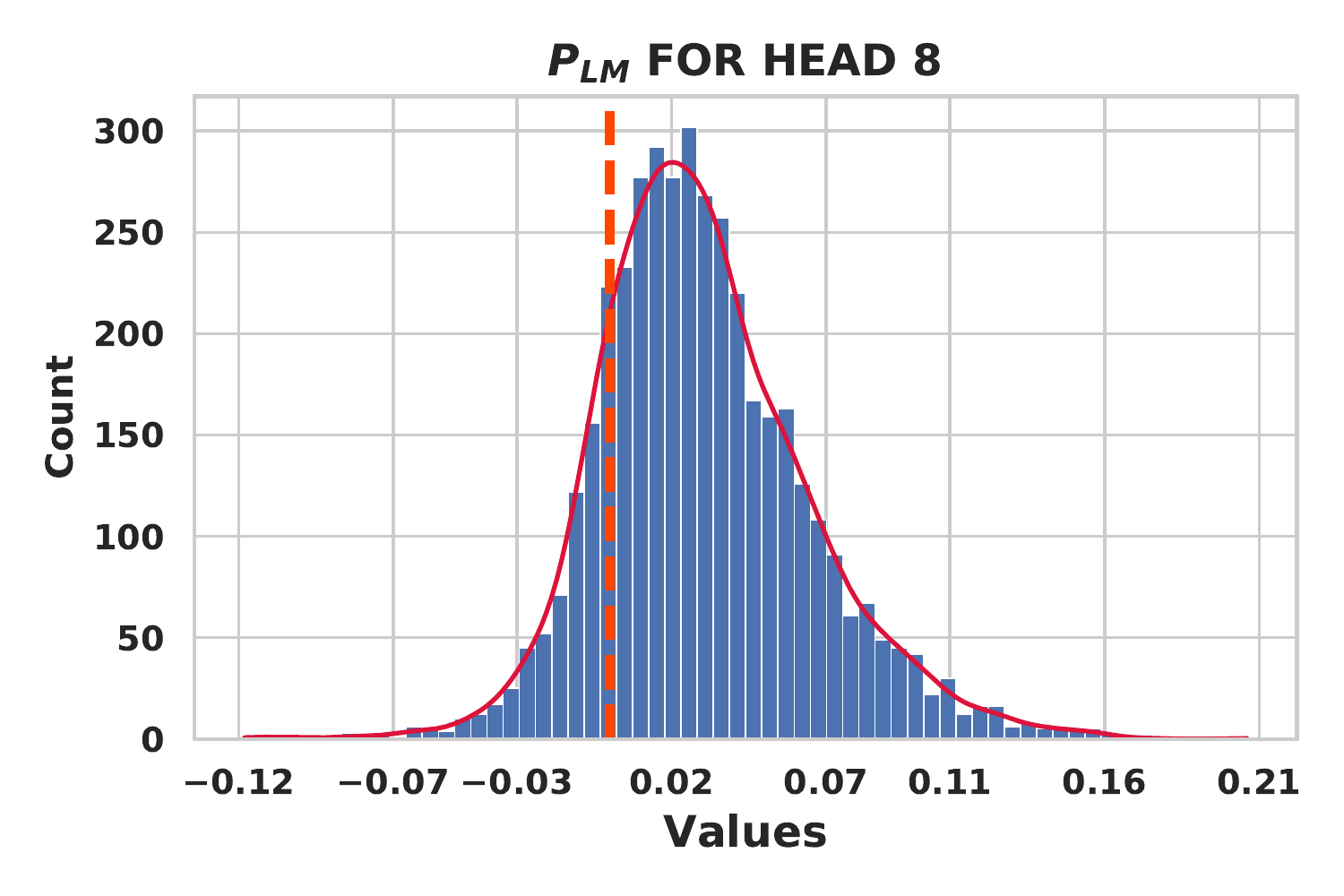}

\end{subfigure}
\caption{$\mP_{LM}$ histogram plots for all heads from TLM attention stage from graph transformer model \#2 for PT-EN translation task. Dashed line in orange marks zero value.}
\label{fig25apx}
\end{adjustwidth}
\end{figure}    

\clearpage
\thispagestyle{headings}

\begin{figure}
\begin{adjustwidth}{-5em}{-5em}
\centering
\begin{subfigure}[b]{0.6\textwidth}
	\centering
	\includegraphics[width=1.1\textwidth]{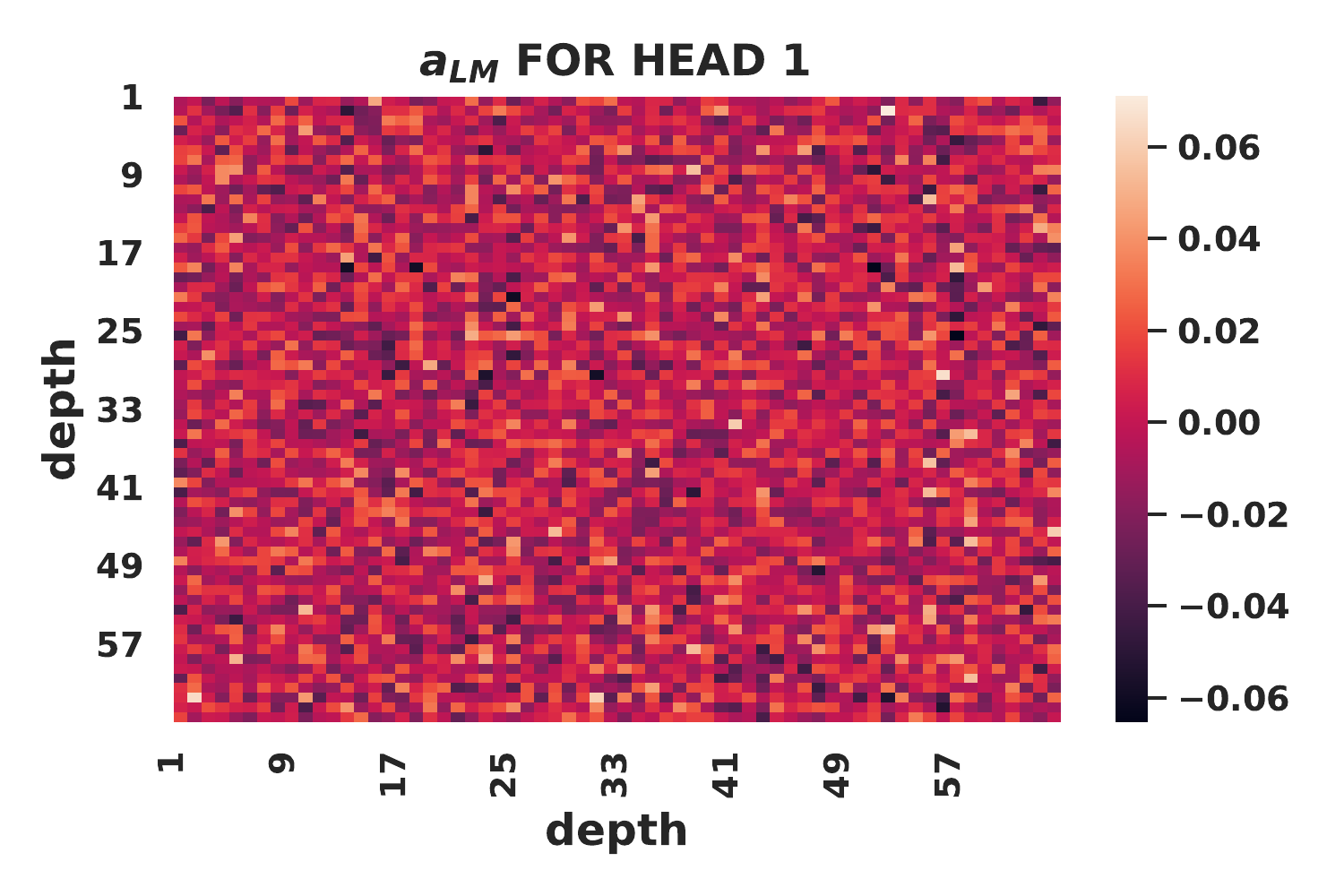}

\end{subfigure}
\hfill
\begin{subfigure}[b]{0.6\textwidth}
	\centering
	\includegraphics[width=1.1\textwidth]{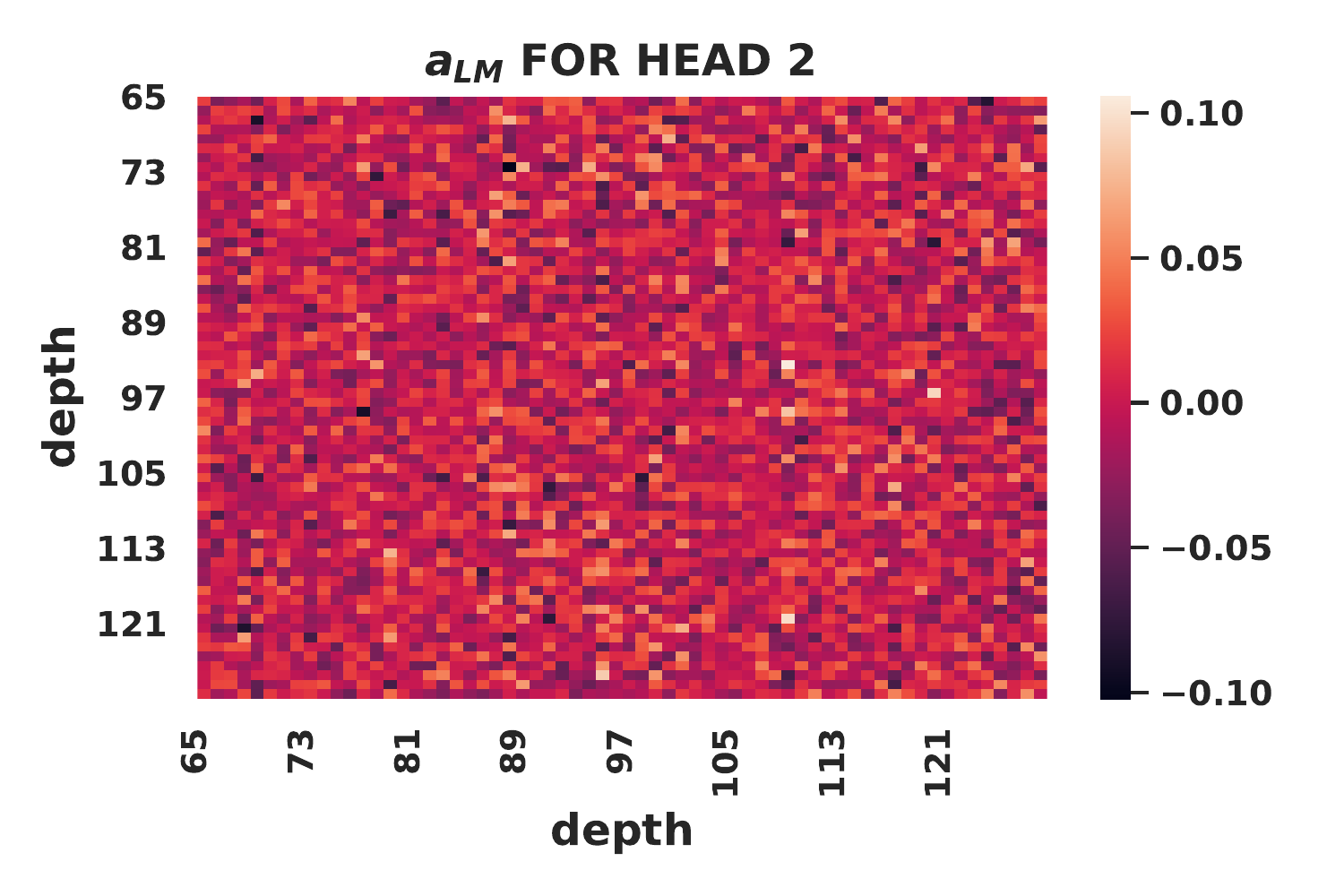}

\end{subfigure}
\hfill
\begin{subfigure}[b]{0.6\textwidth}
	\centering
	\includegraphics[width=1.1\textwidth]{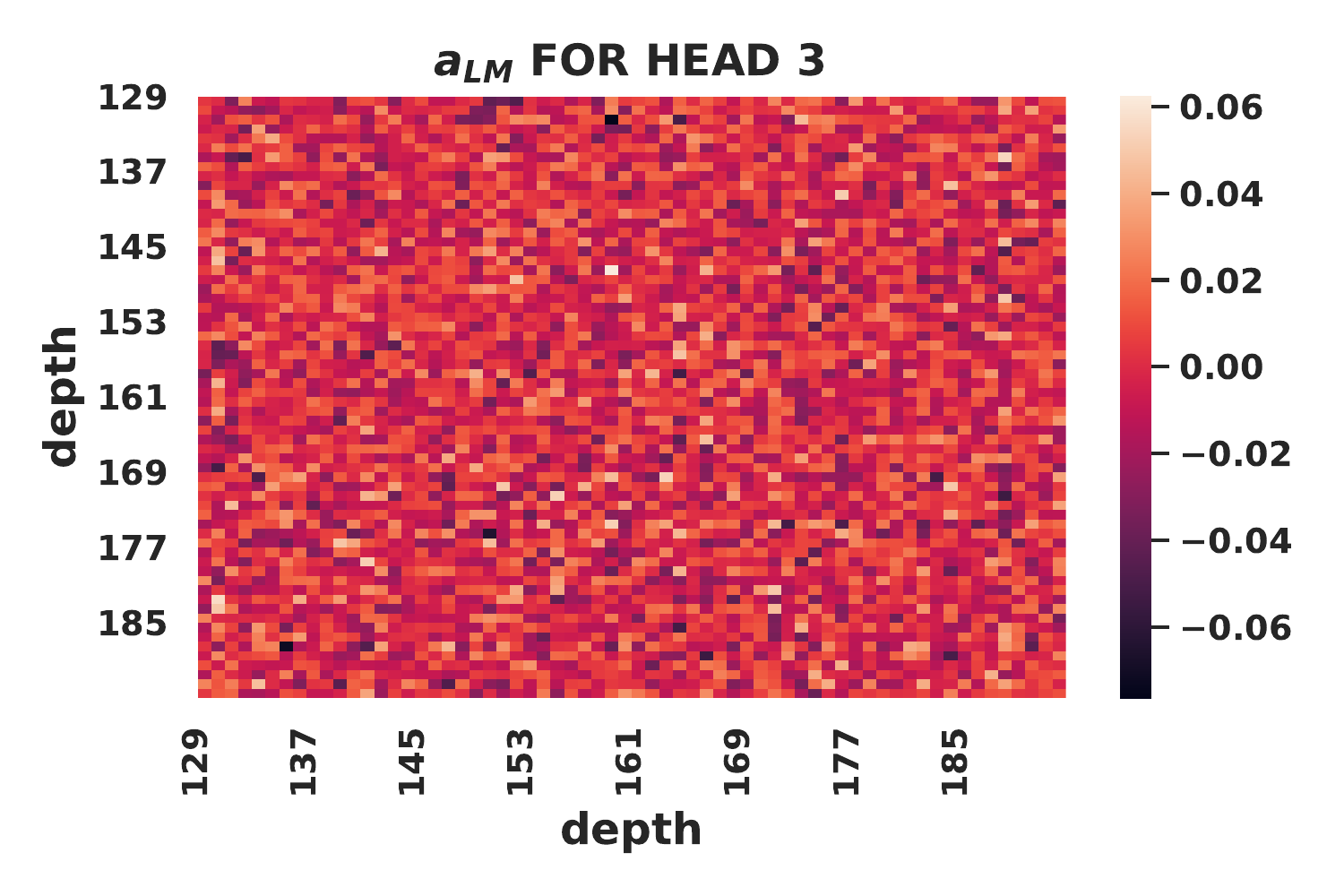}

\end{subfigure}
\hfill
\begin{subfigure}[b]{0.6\textwidth}
	\centering
	\includegraphics[width=1.1\textwidth]{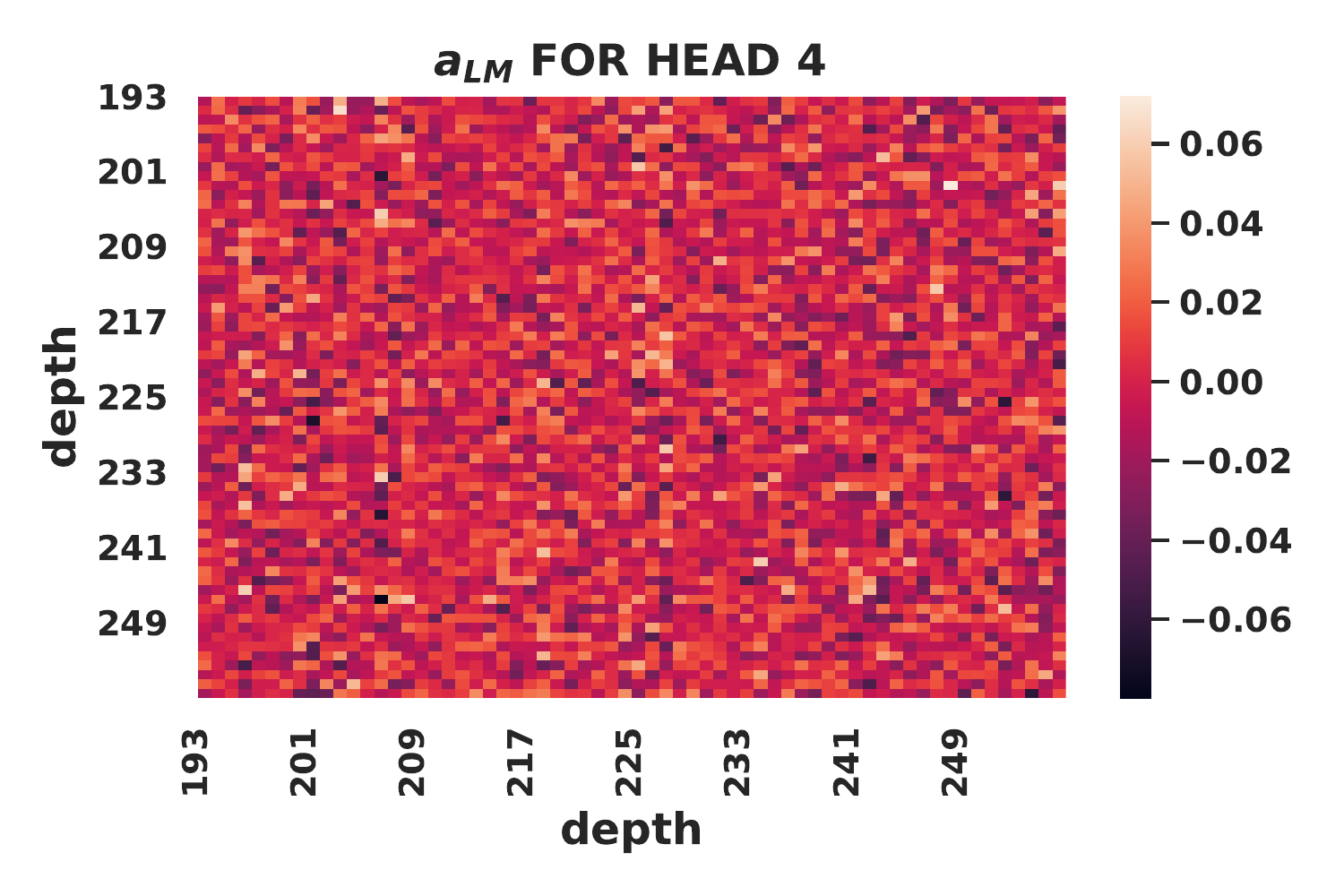}

\end{subfigure}
\centering
\begin{subfigure}[b]{0.6\textwidth}
	\centering
	\includegraphics[width=1.1\textwidth]{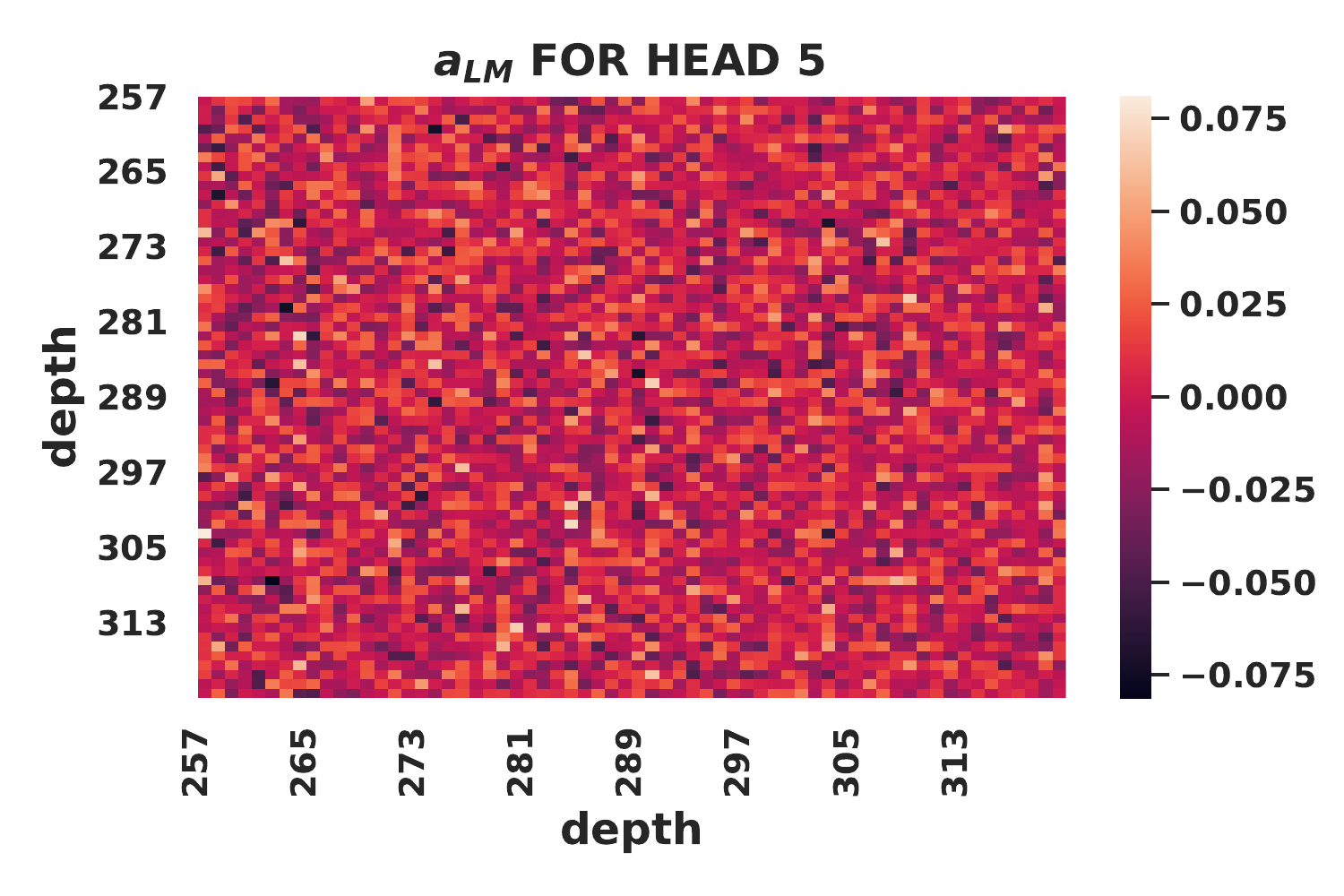}

\end{subfigure}
\hfill
\begin{subfigure}[b]{0.6\textwidth}
	\centering
	\includegraphics[width=1.1\textwidth]{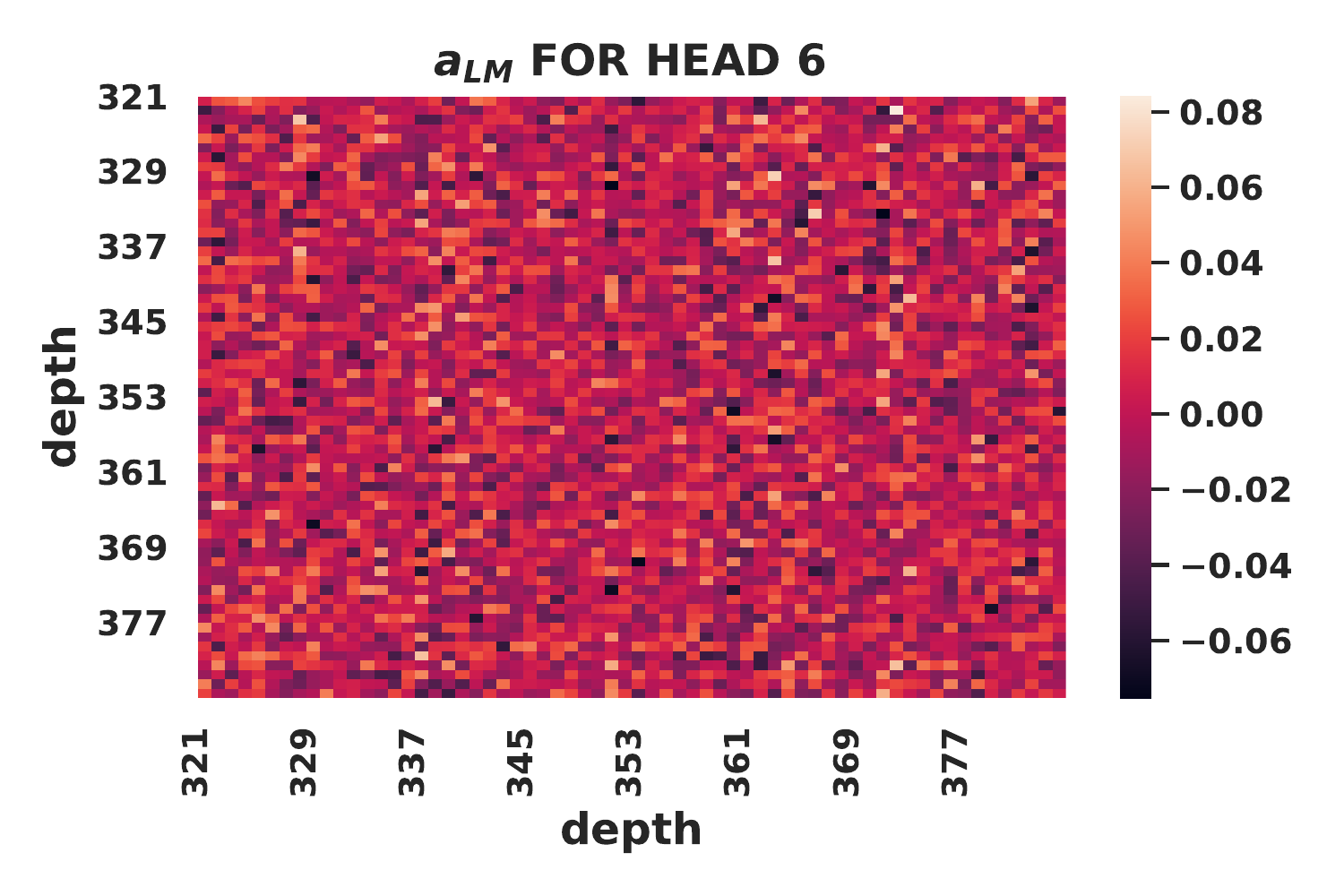}

\end{subfigure}
\hfill
\begin{subfigure}[b]{0.6\textwidth}
	\centering
	\includegraphics[width=1.1\textwidth]{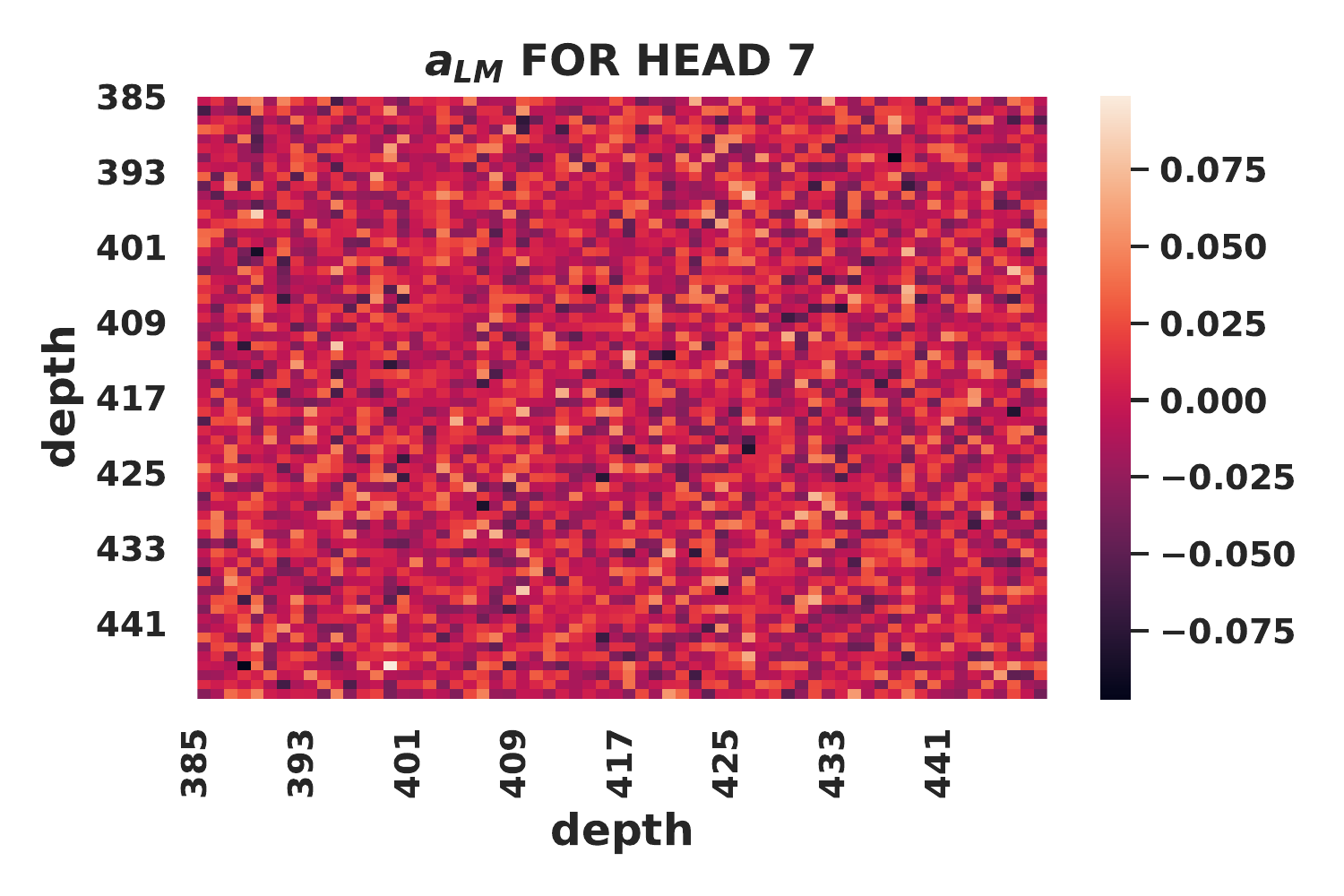}

\end{subfigure}
\hfill
\begin{subfigure}[b]{0.6\textwidth}
	\centering
	\includegraphics[width=1.1\textwidth]{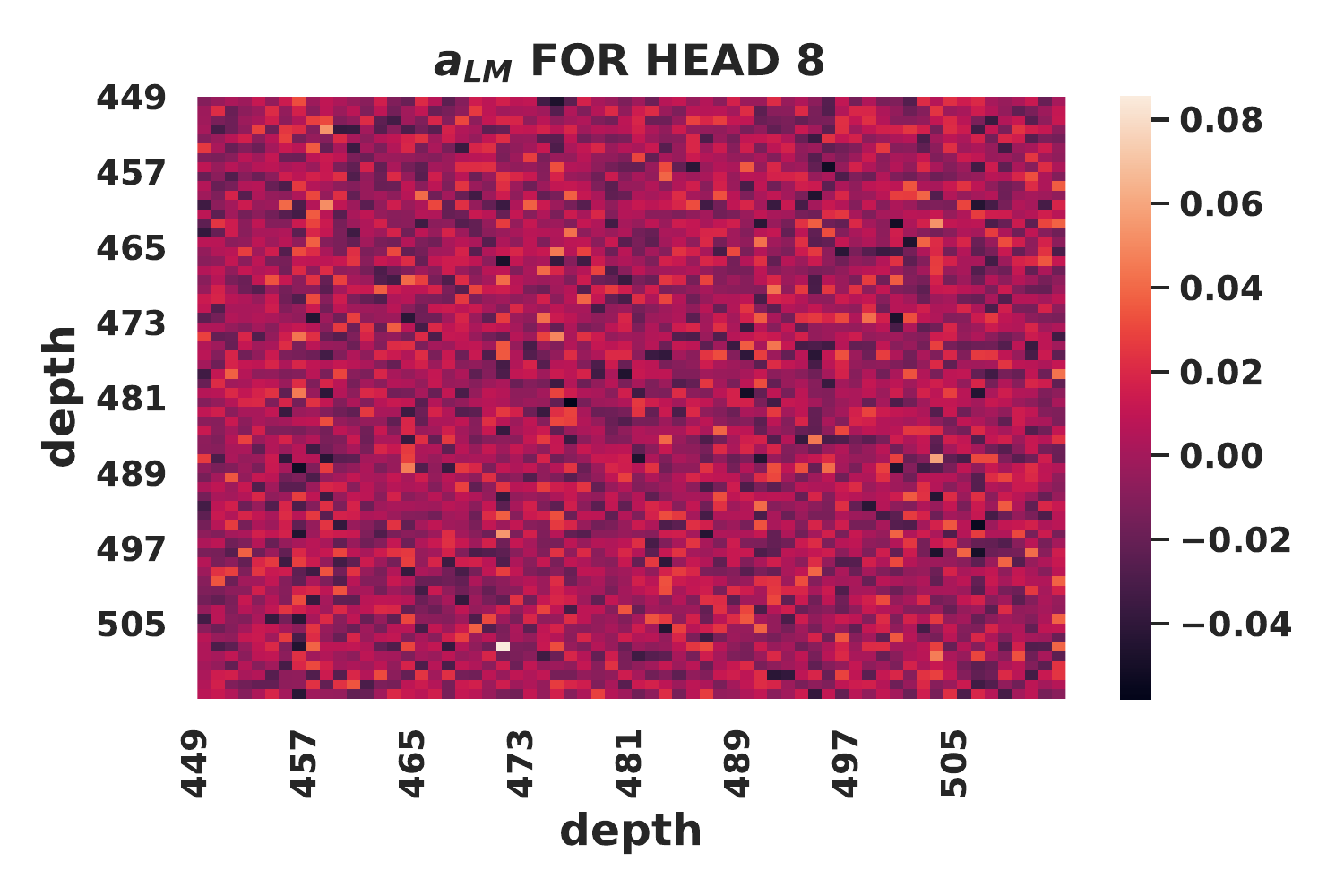}

\end{subfigure}
\caption{$\va_{LM}$ heatmap plots for all heads from TLM attention stage from Graph transformer model \#2 for PT-EN translation task.}
\label{fig26apx}
\end{adjustwidth}
\end{figure}    

\clearpage
\thispagestyle{headings}
\begin{figure}
\begin{adjustwidth}{-5em}{-5em}
\centering
\begin{subfigure}[b]{0.6\textwidth}
	\centering
	\includegraphics[width=1.1\textwidth]{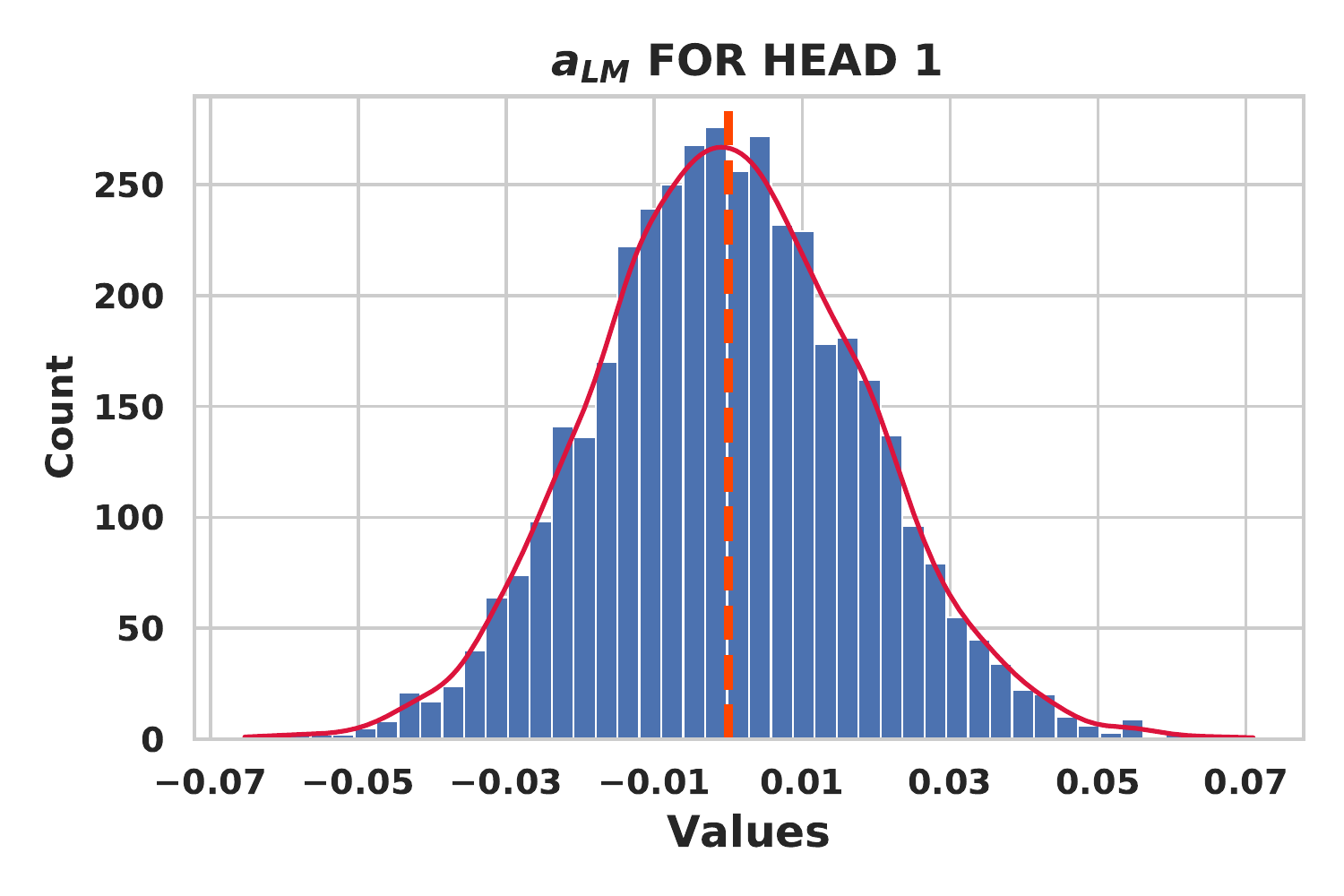}

\end{subfigure}
\hfill
\begin{subfigure}[b]{0.6\textwidth}
	\centering
	\includegraphics[width=1.1\textwidth]{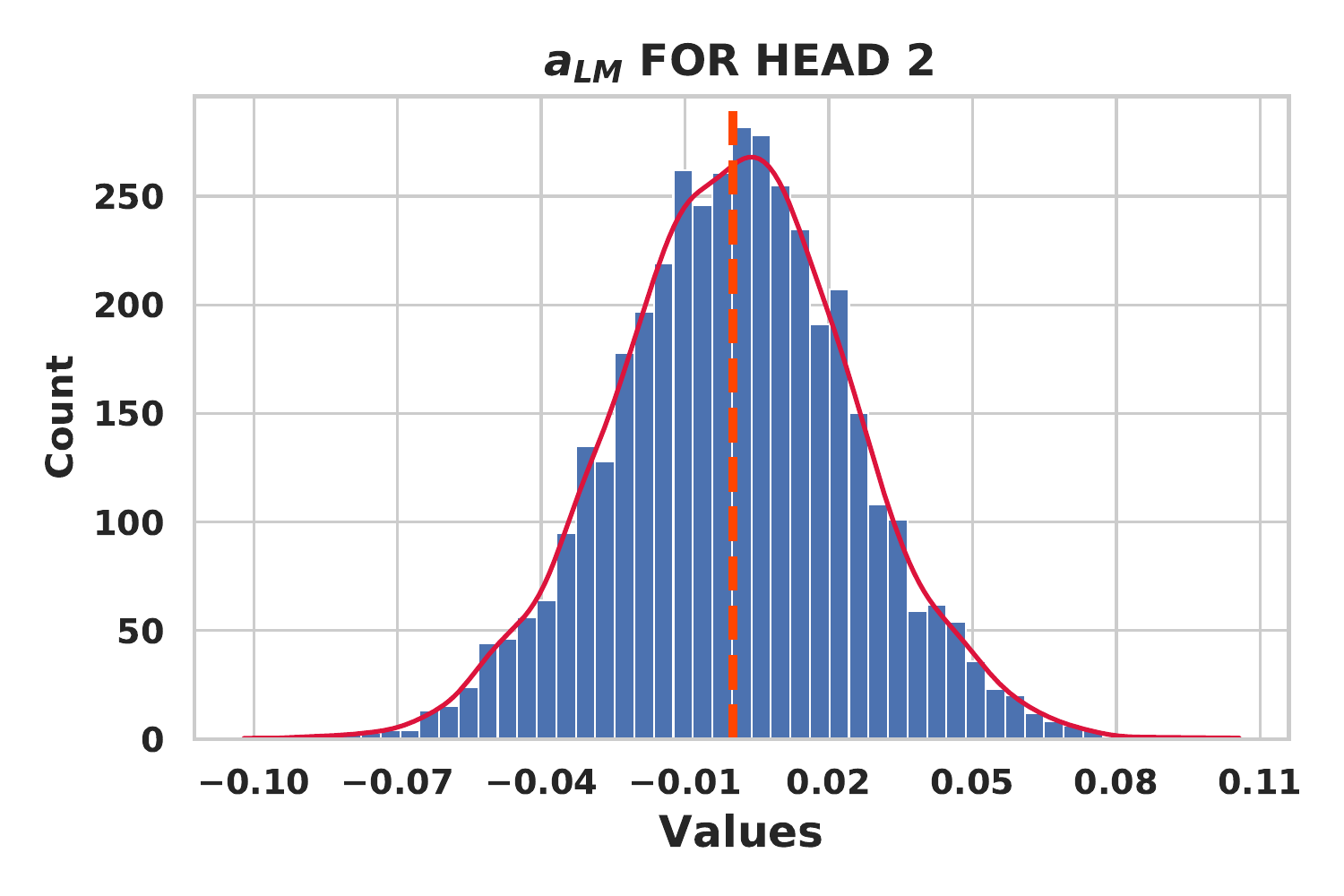}

\end{subfigure}
\hfill
\begin{subfigure}[b]{0.6\textwidth}
	\centering
	\includegraphics[width=1.1\textwidth]{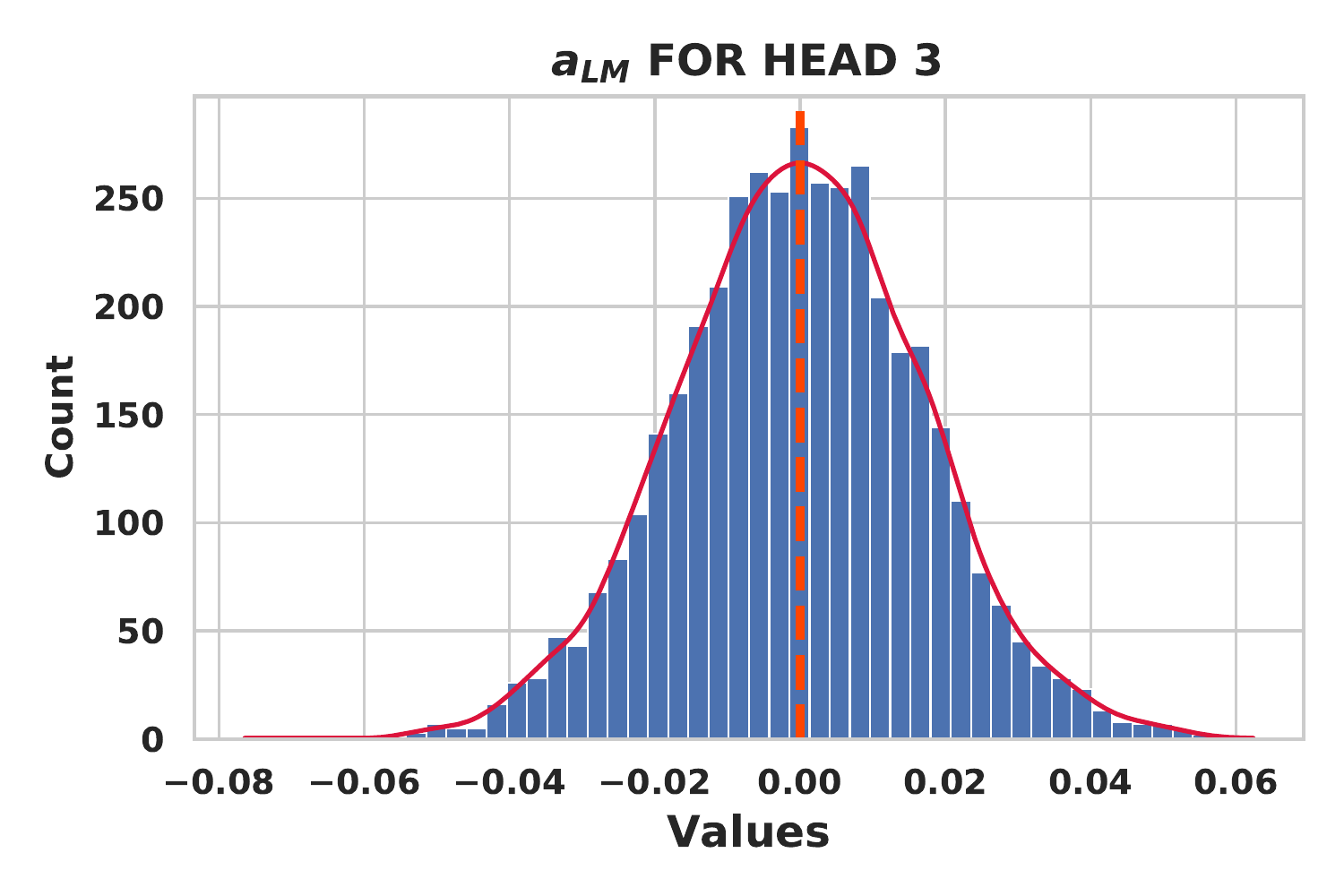}

\end{subfigure}
\hfill
\begin{subfigure}[b]{0.6\textwidth}
	\centering
	\includegraphics[width=1.1\textwidth]{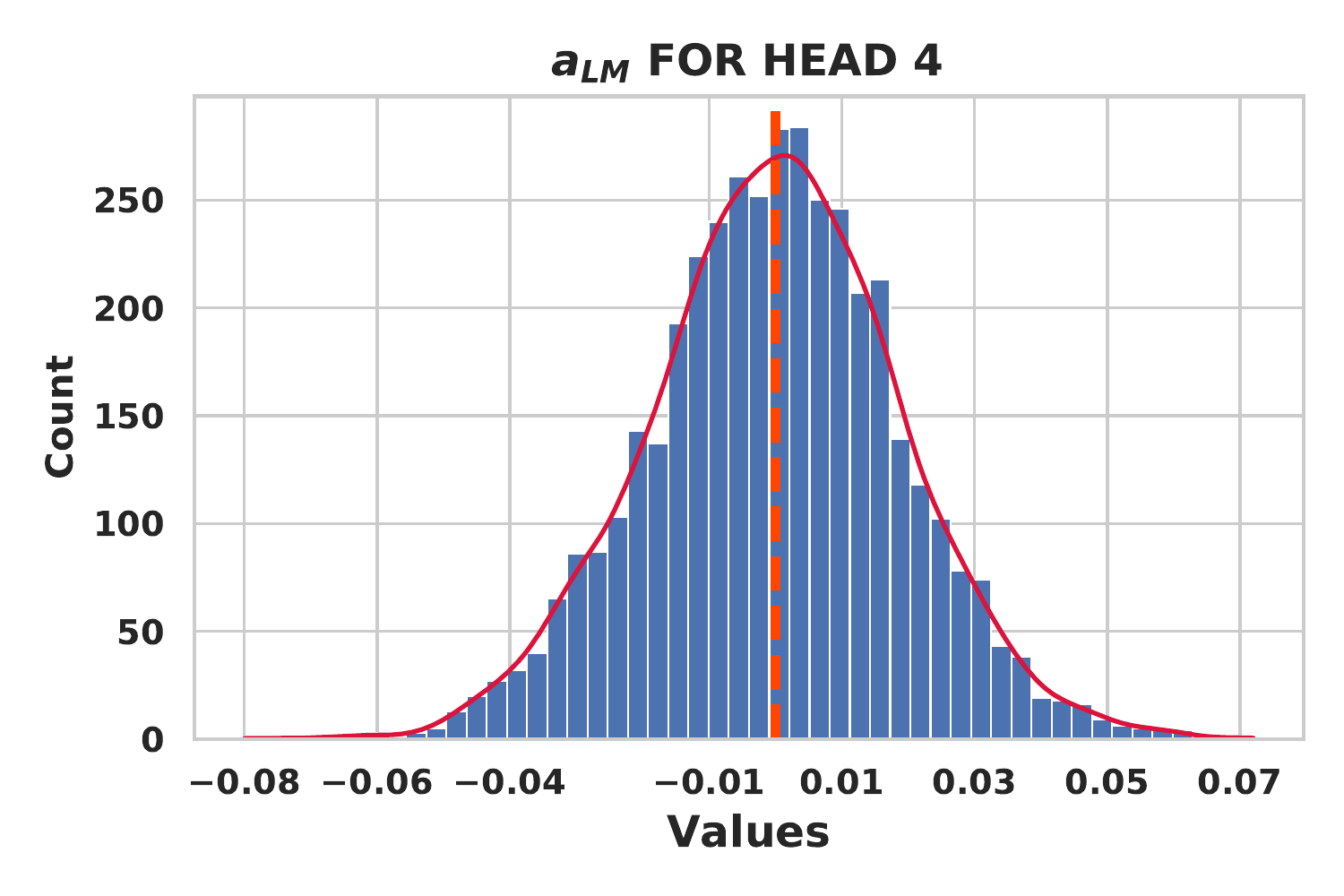}

\end{subfigure}
\centering
\begin{subfigure}[b]{0.6\textwidth}
	\centering
	\includegraphics[width=1.1\textwidth]{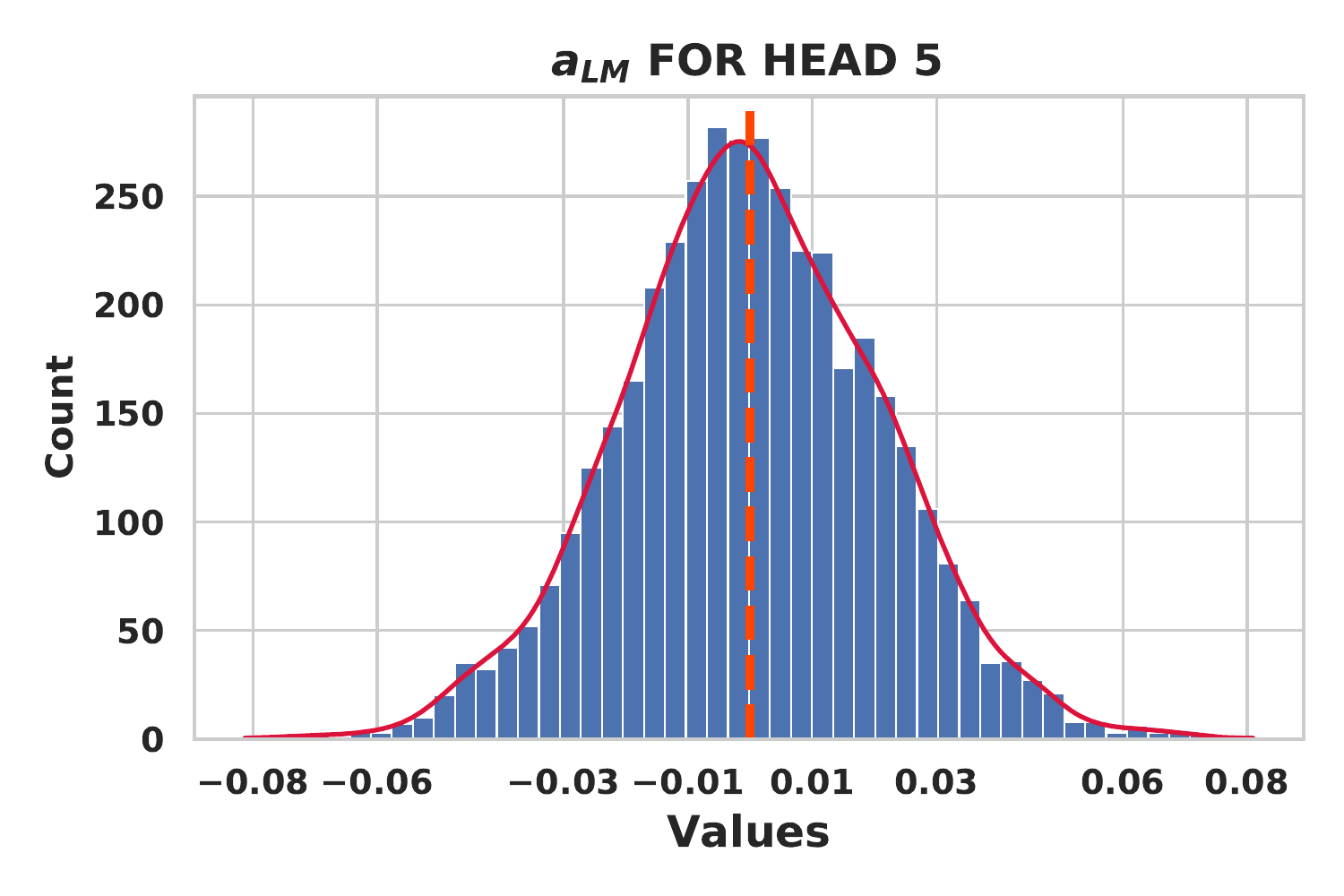}

\end{subfigure}
\hfill
\begin{subfigure}[b]{0.6\textwidth}
	\centering
	\includegraphics[width=1.1\textwidth]{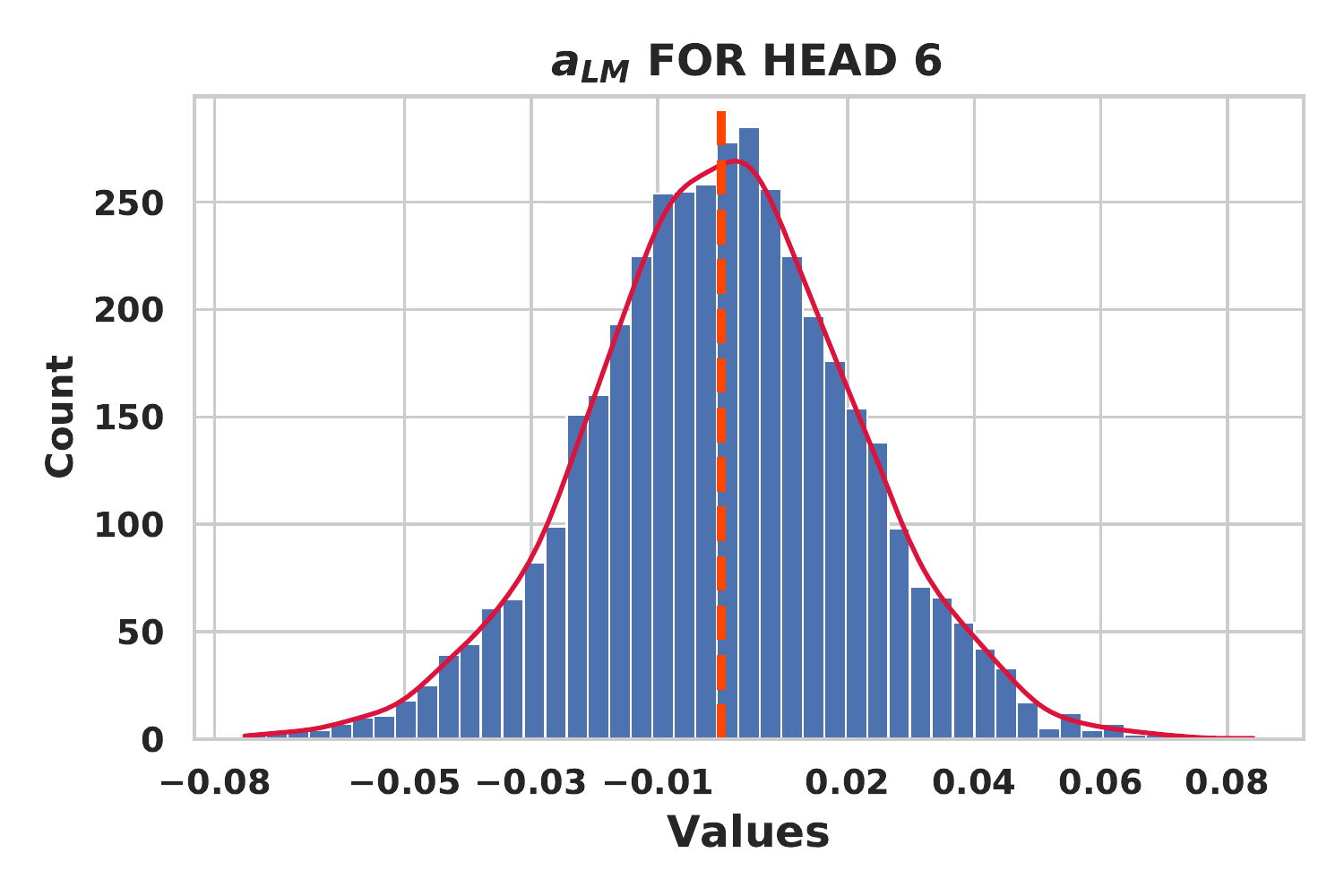}

\end{subfigure}
\hfill
\begin{subfigure}[b]{0.6\textwidth}
	\centering
	\includegraphics[width=1.1\textwidth]{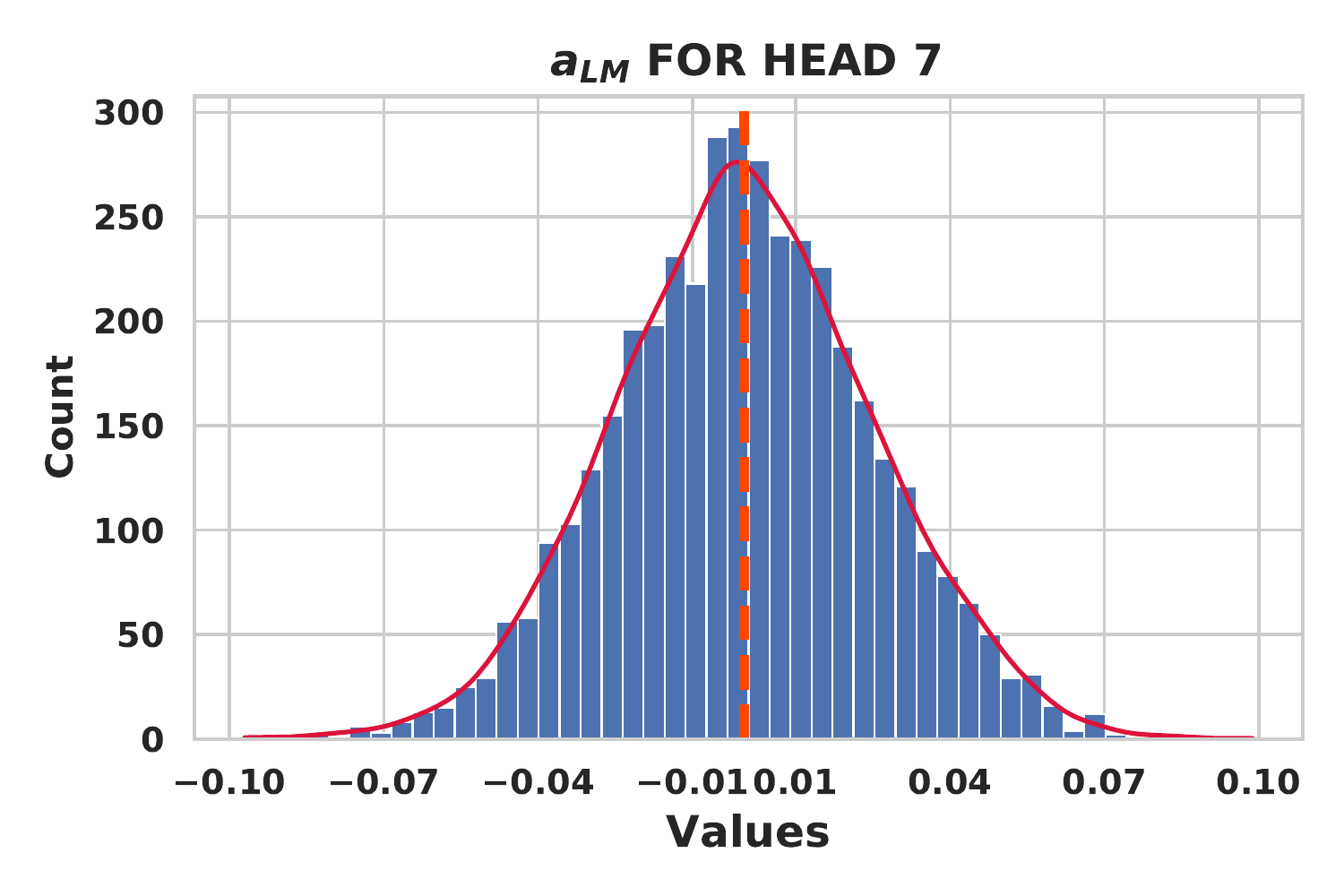}

\end{subfigure}
\hfill
\begin{subfigure}[b]{0.6\textwidth}
	\centering
	\includegraphics[width=1.1\textwidth]{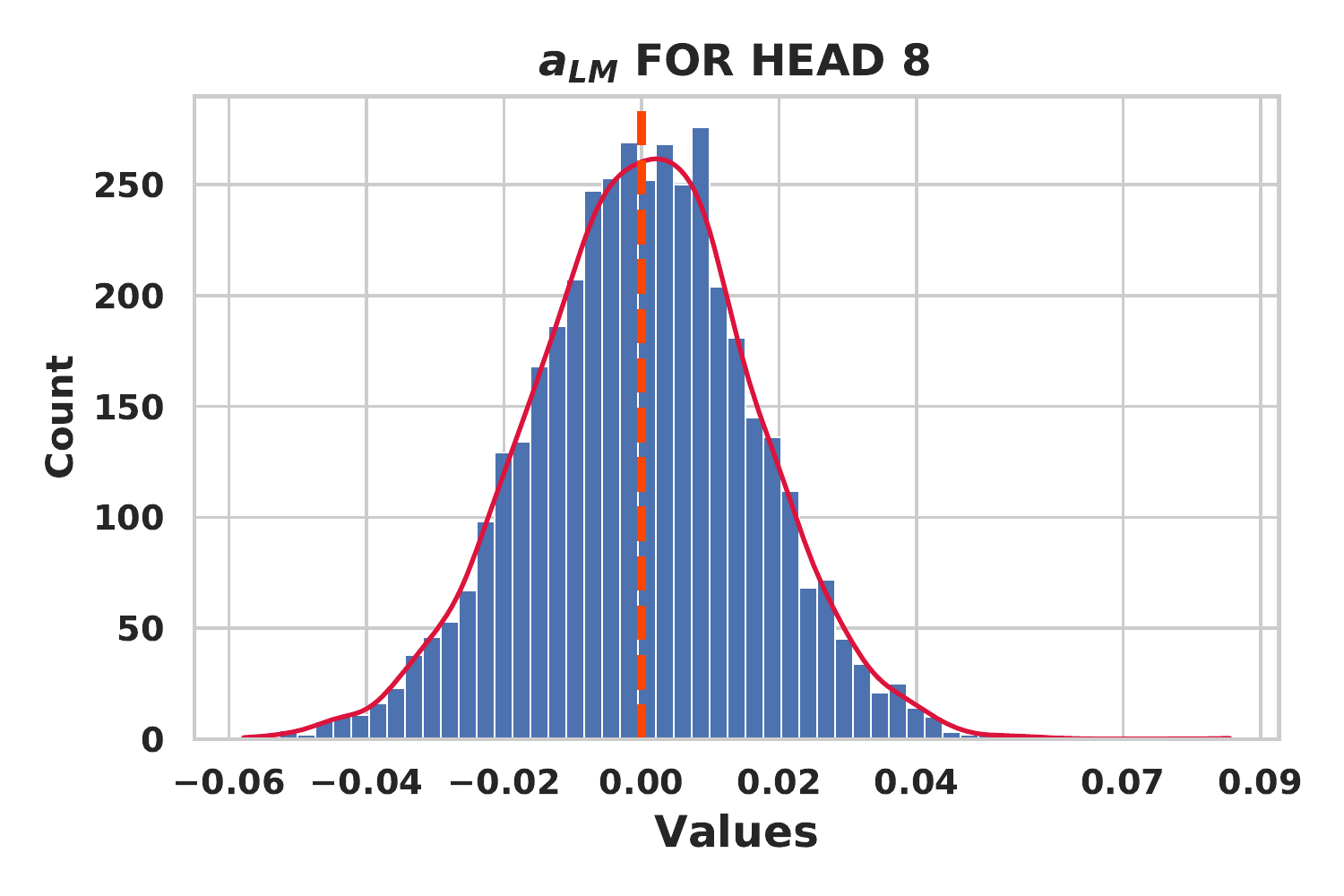}

\end{subfigure}
\caption{$\va_{LM}$ histogram plots for all heads from TLM attention stage from Graph transformer model \#2 for PT-EN translation task. Dashed line in orange marks zero value.}
\label{fig27apx}
\end{adjustwidth}
\end{figure}  

\clearpage
\thispagestyle{headings}
\begin{figure}
\begin{adjustwidth}{-5em}{-5em}
\centering
\begin{subfigure}[b]{0.6\textwidth}
	\centering
	\includegraphics[width=1.1\textwidth]{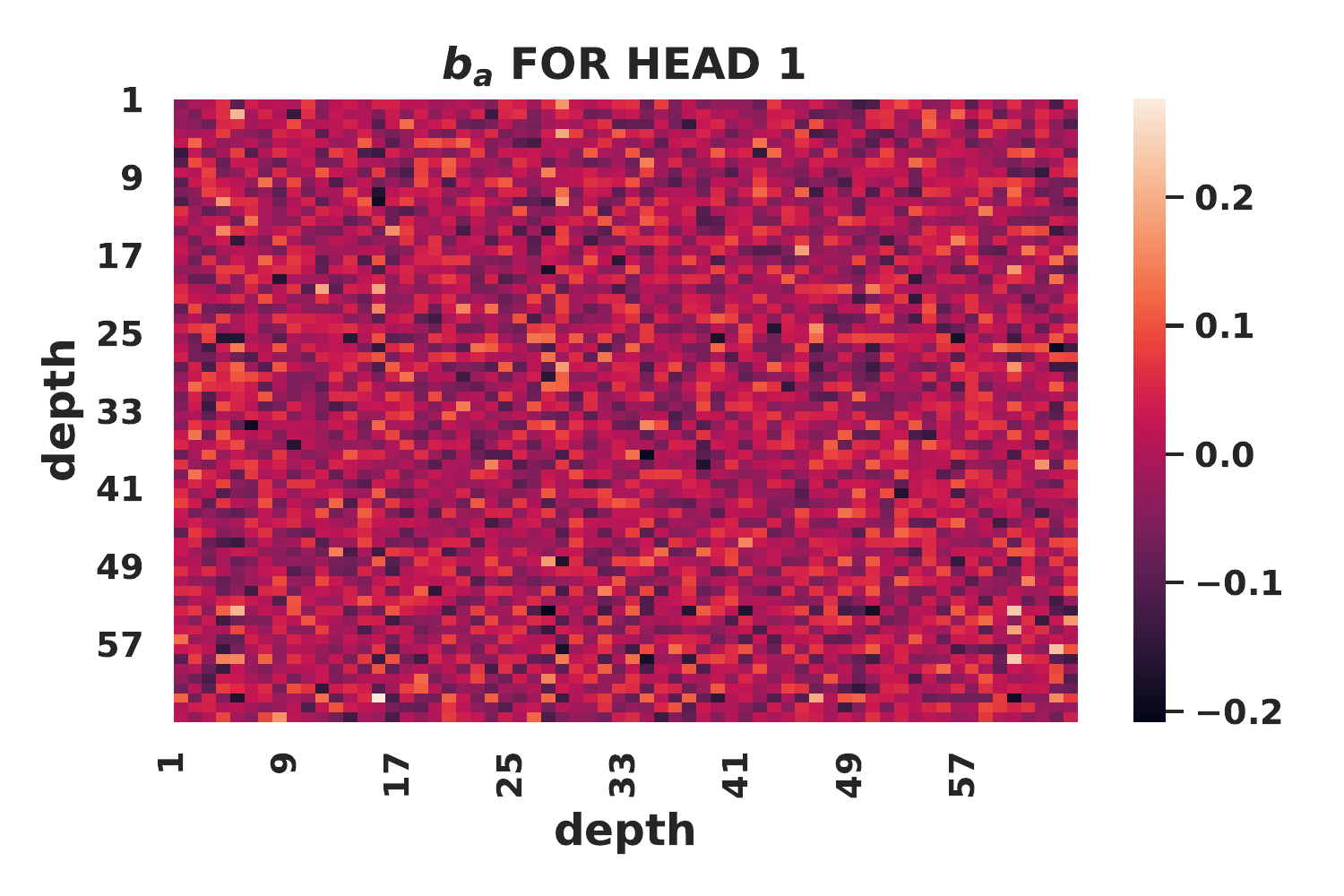}

\end{subfigure}
\hfill
\begin{subfigure}[b]{0.6\textwidth}
	\centering
	\includegraphics[width=1.1\textwidth]{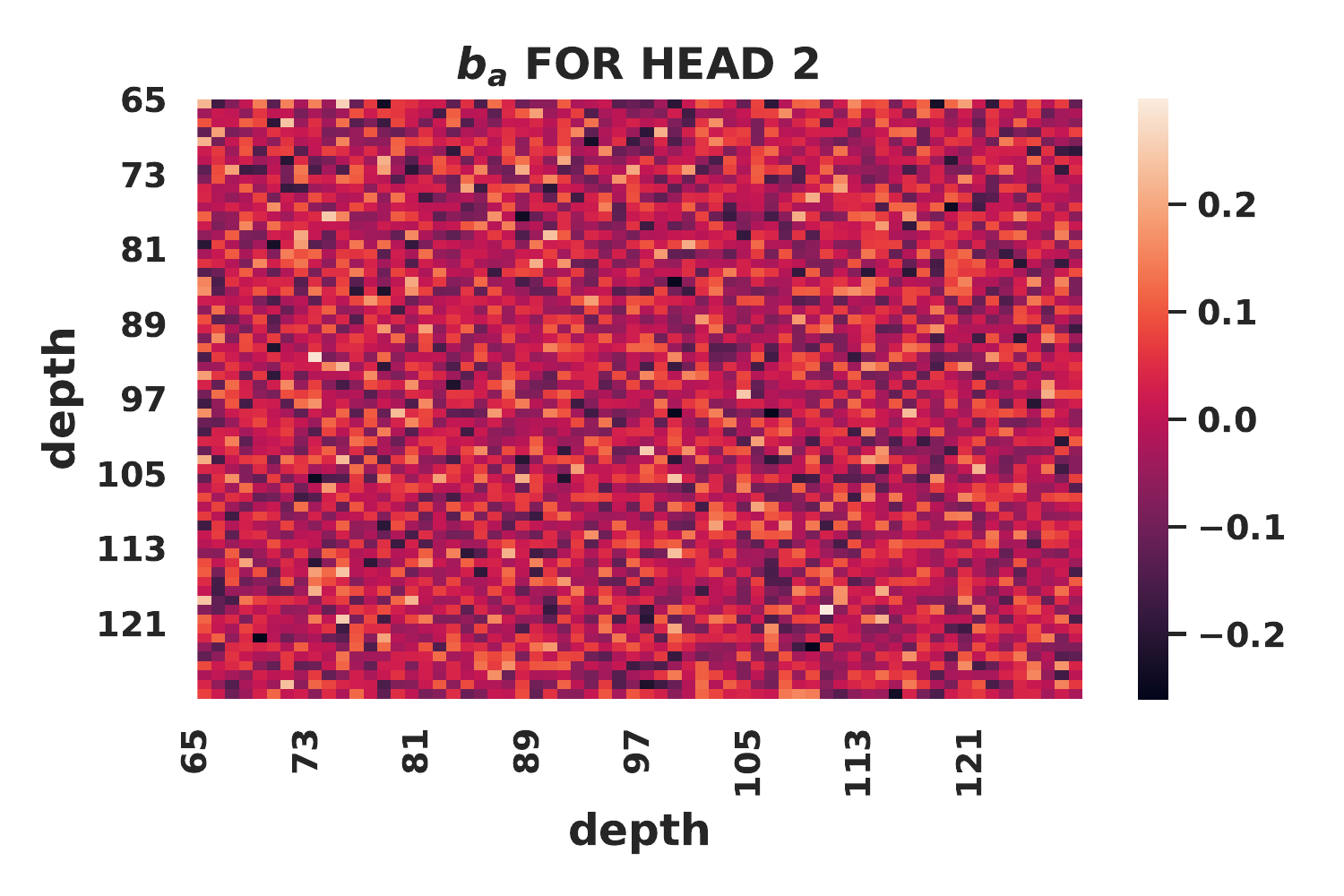}

\end{subfigure}
\hfill
\begin{subfigure}[b]{0.6\textwidth}
	\centering
	\includegraphics[width=1.1\textwidth]{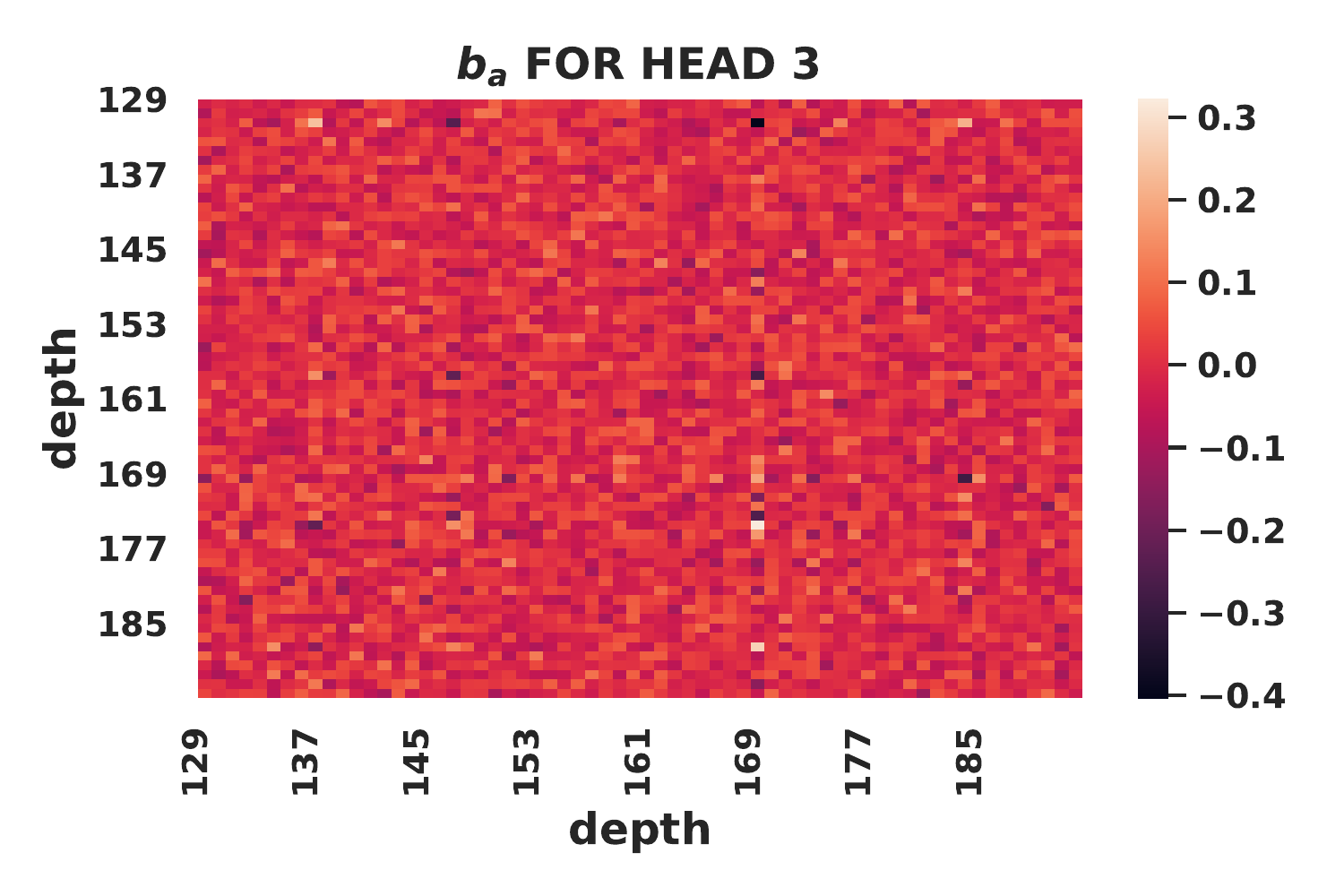}

\end{subfigure}
\hfill
\begin{subfigure}[b]{0.6\textwidth}
	\centering
	\includegraphics[width=1.1\textwidth]{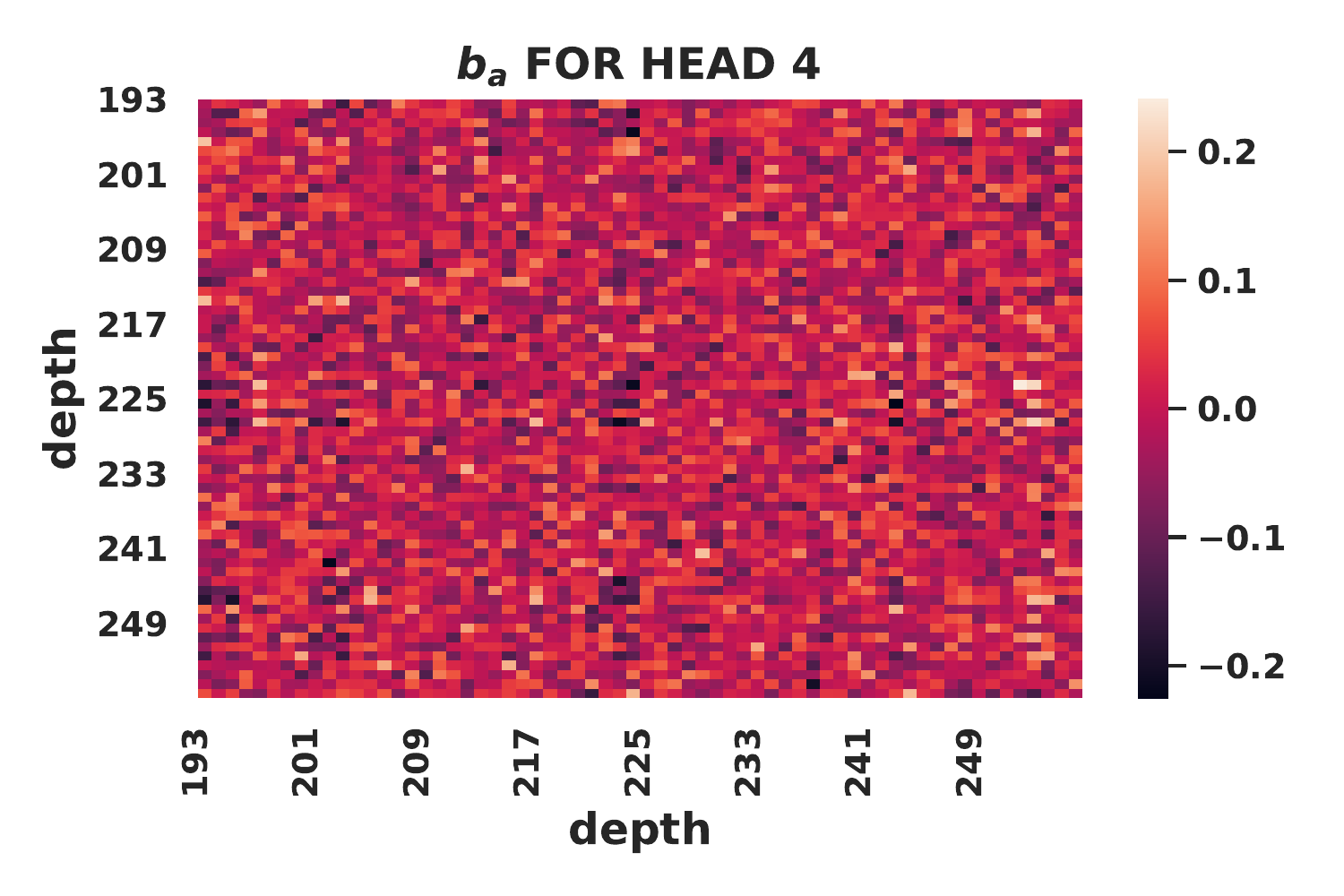}

\end{subfigure}
\centering
\begin{subfigure}[b]{0.6\textwidth}
	\centering
	\includegraphics[width=1.1\textwidth]{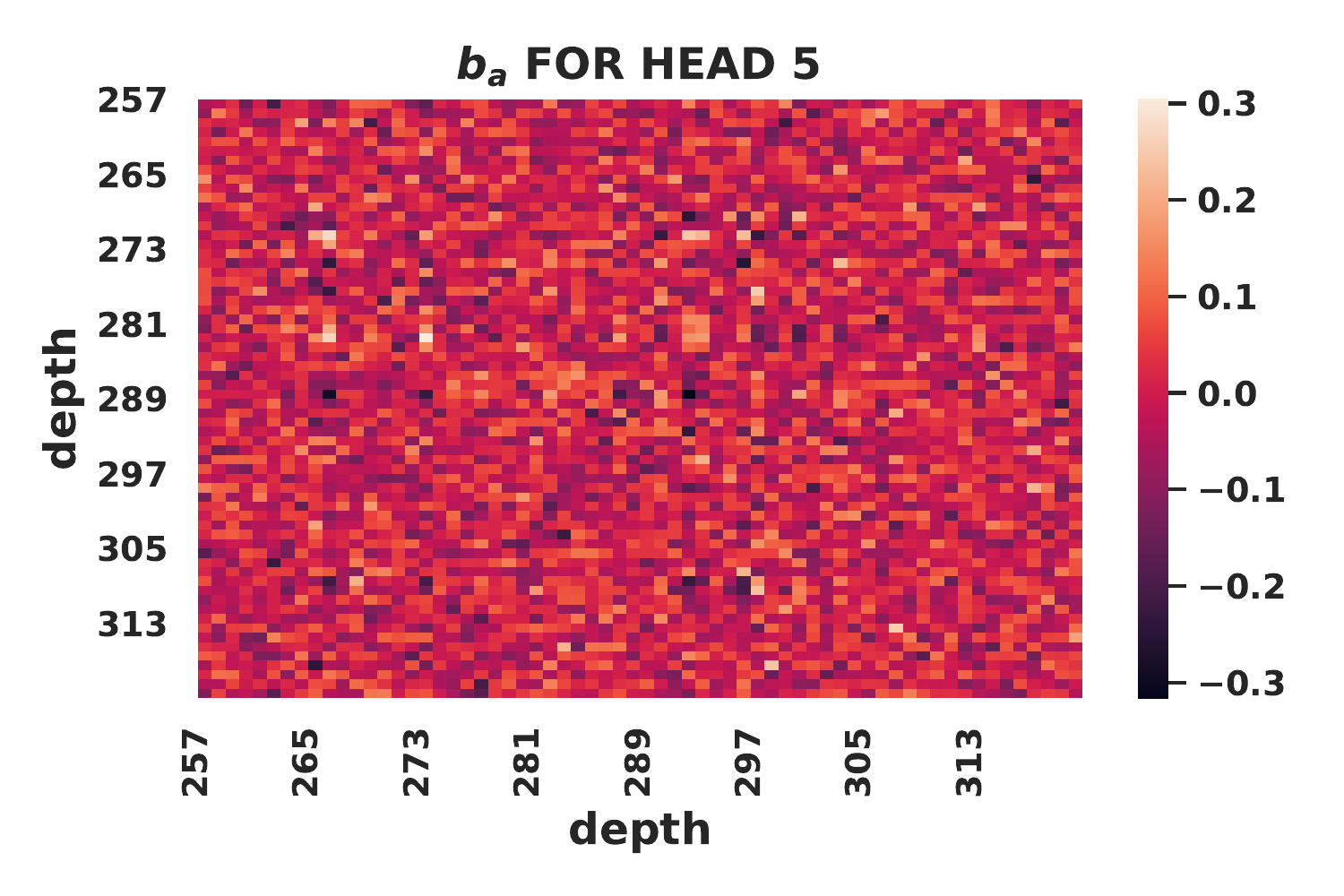}

\end{subfigure}
\hfill
\begin{subfigure}[b]{0.6\textwidth}
	\centering
	\includegraphics[width=1.1\textwidth]{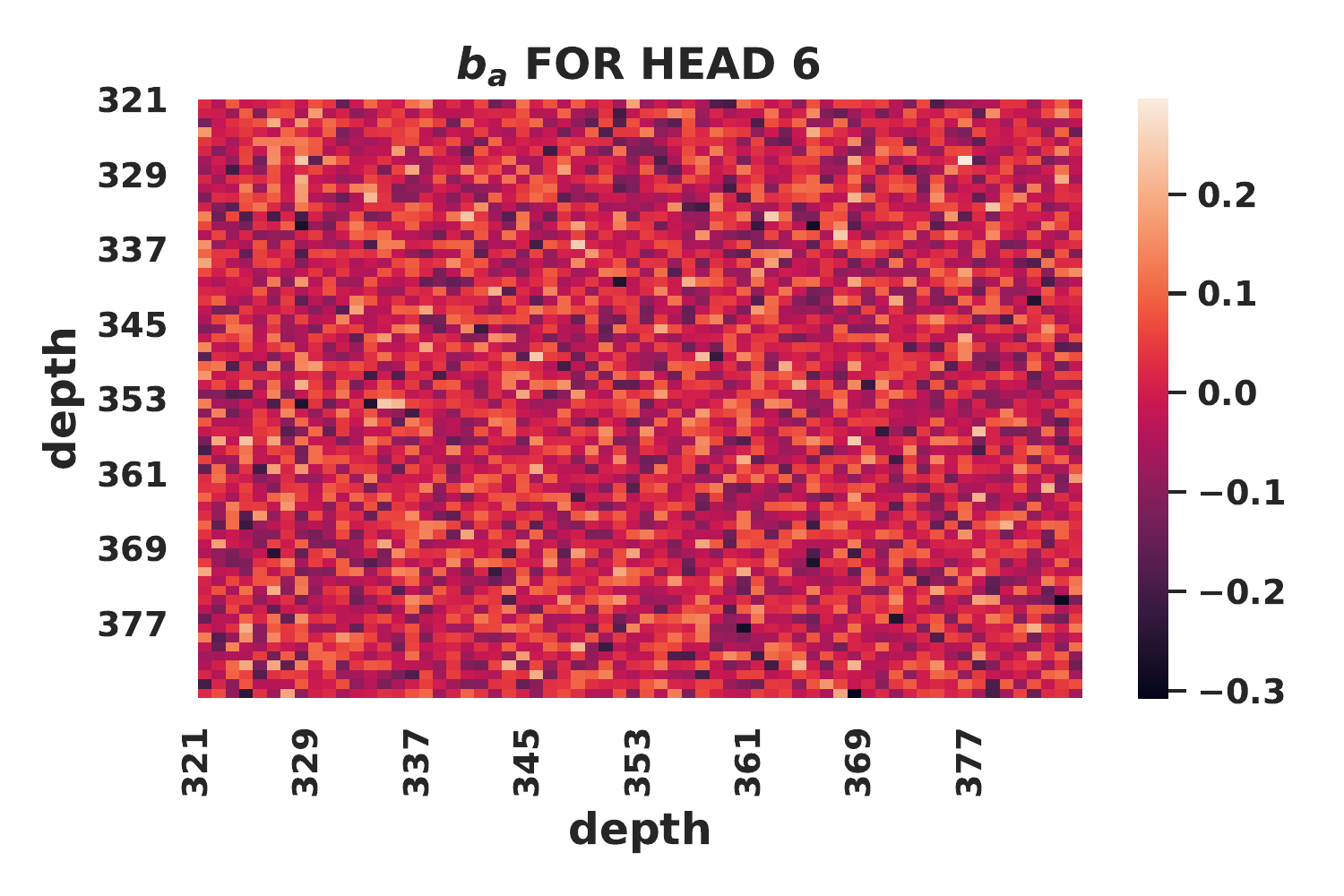}

\end{subfigure}
\hfill
\begin{subfigure}[b]{0.6\textwidth}
	\centering
	\includegraphics[width=1.1\textwidth]{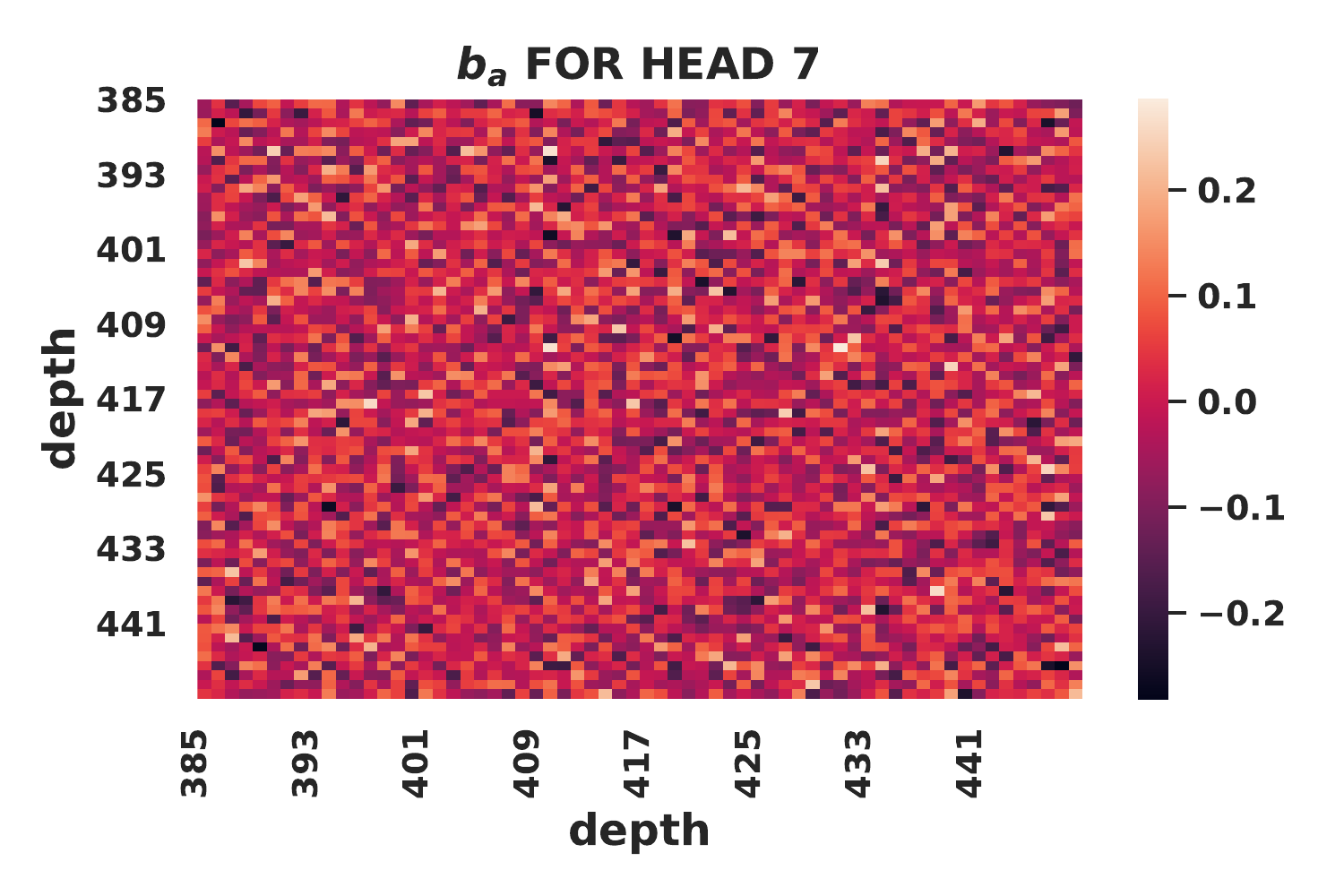}

\end{subfigure}
\hfill
\begin{subfigure}[b]{0.6\textwidth}
	\centering
	\includegraphics[width=1.1\textwidth]{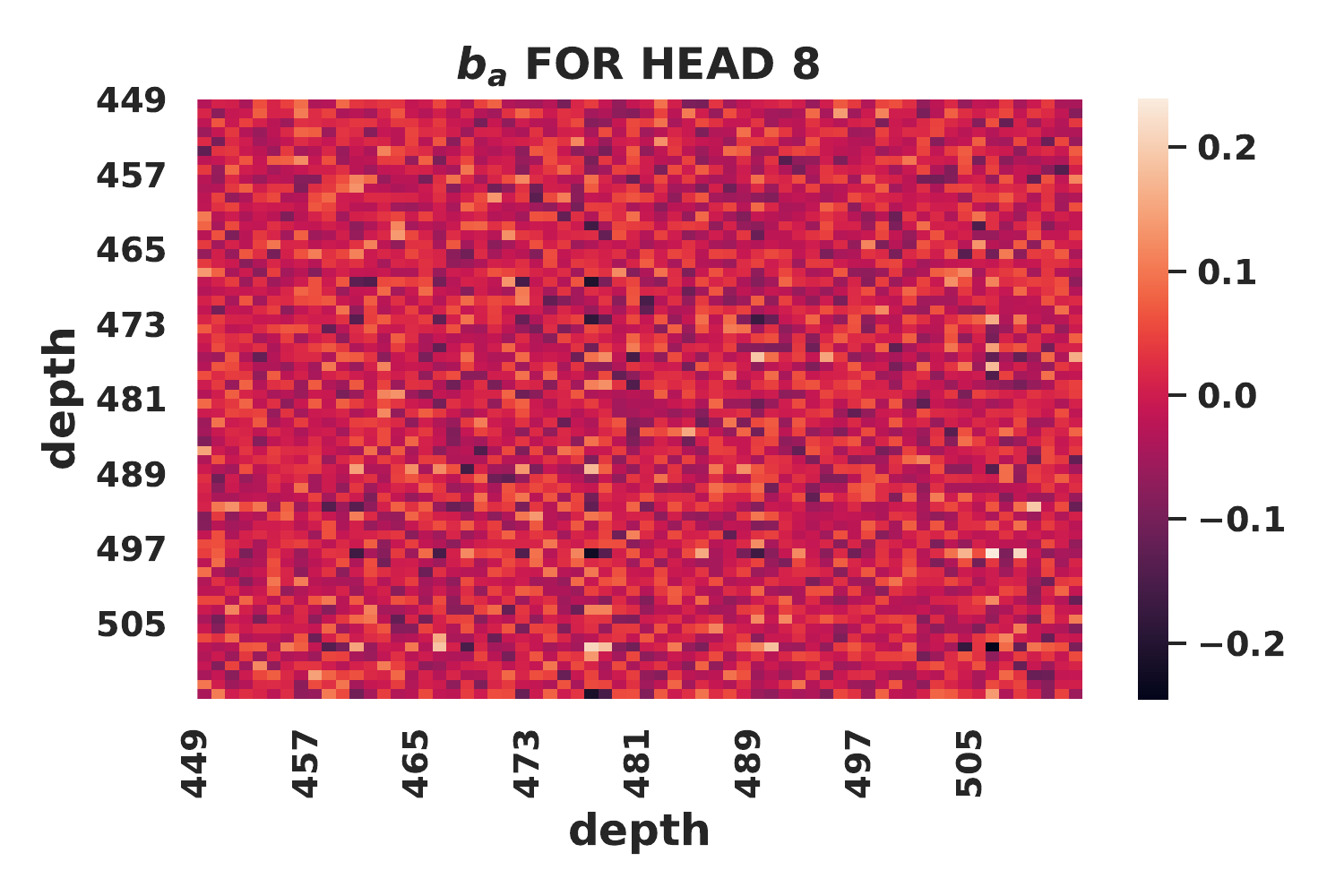}

\end{subfigure}
\caption{$\vb_{a}$ heatmap plots for all heads from TLM attention stage from graph transformer model \#2 for PT-EN translation task.}
\label{fig28apx}
\end{adjustwidth}
\end{figure}

\clearpage
\thispagestyle{headings}

\begin{figure}
\begin{adjustwidth}{-5em}{-5em}
\centering
\begin{subfigure}[b]{0.6\textwidth}
	\centering
	\includegraphics[width=1.1\textwidth]{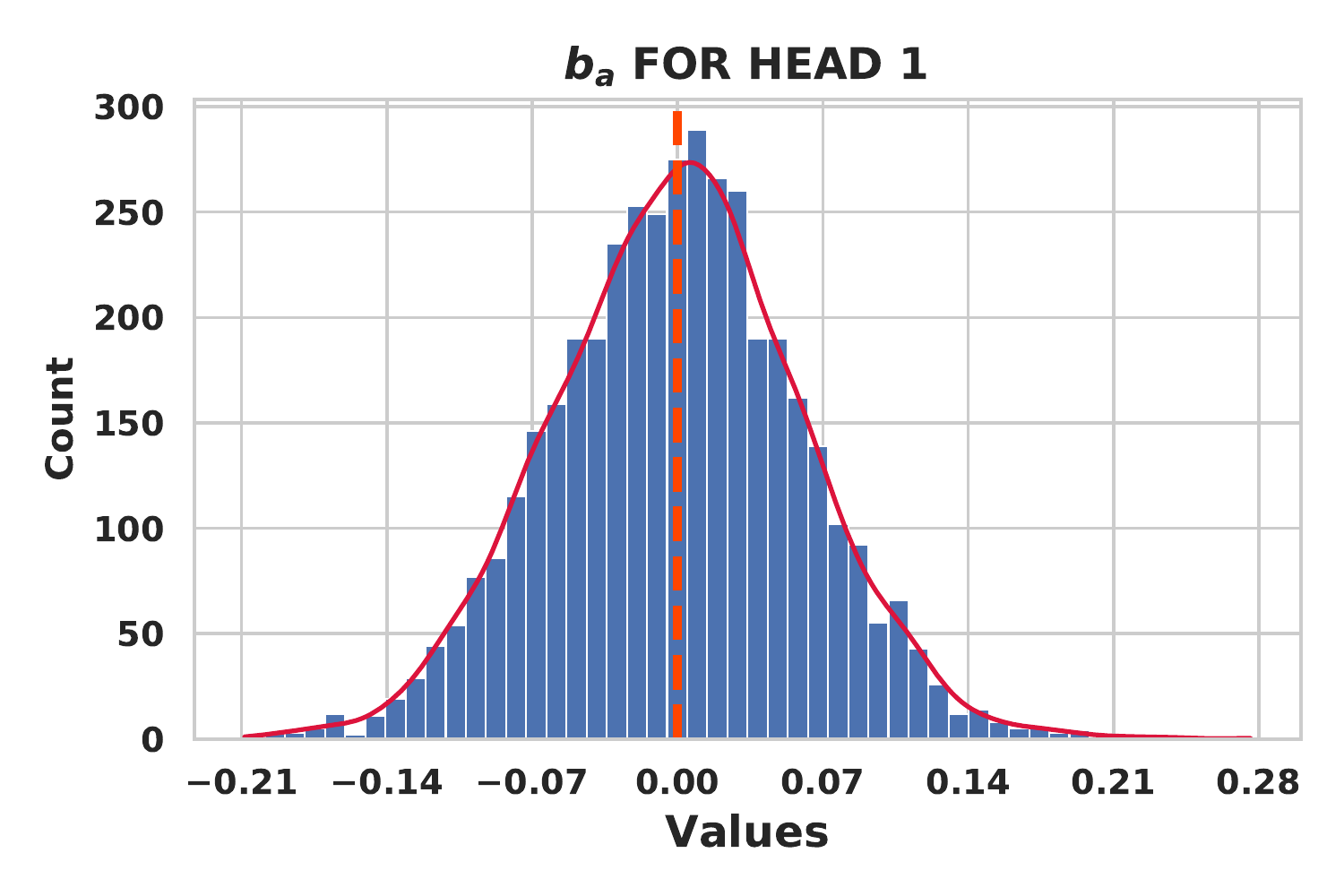}

\end{subfigure}
\hfill
\begin{subfigure}[b]{0.6\textwidth}
	\centering
	\includegraphics[width=1.1\textwidth]{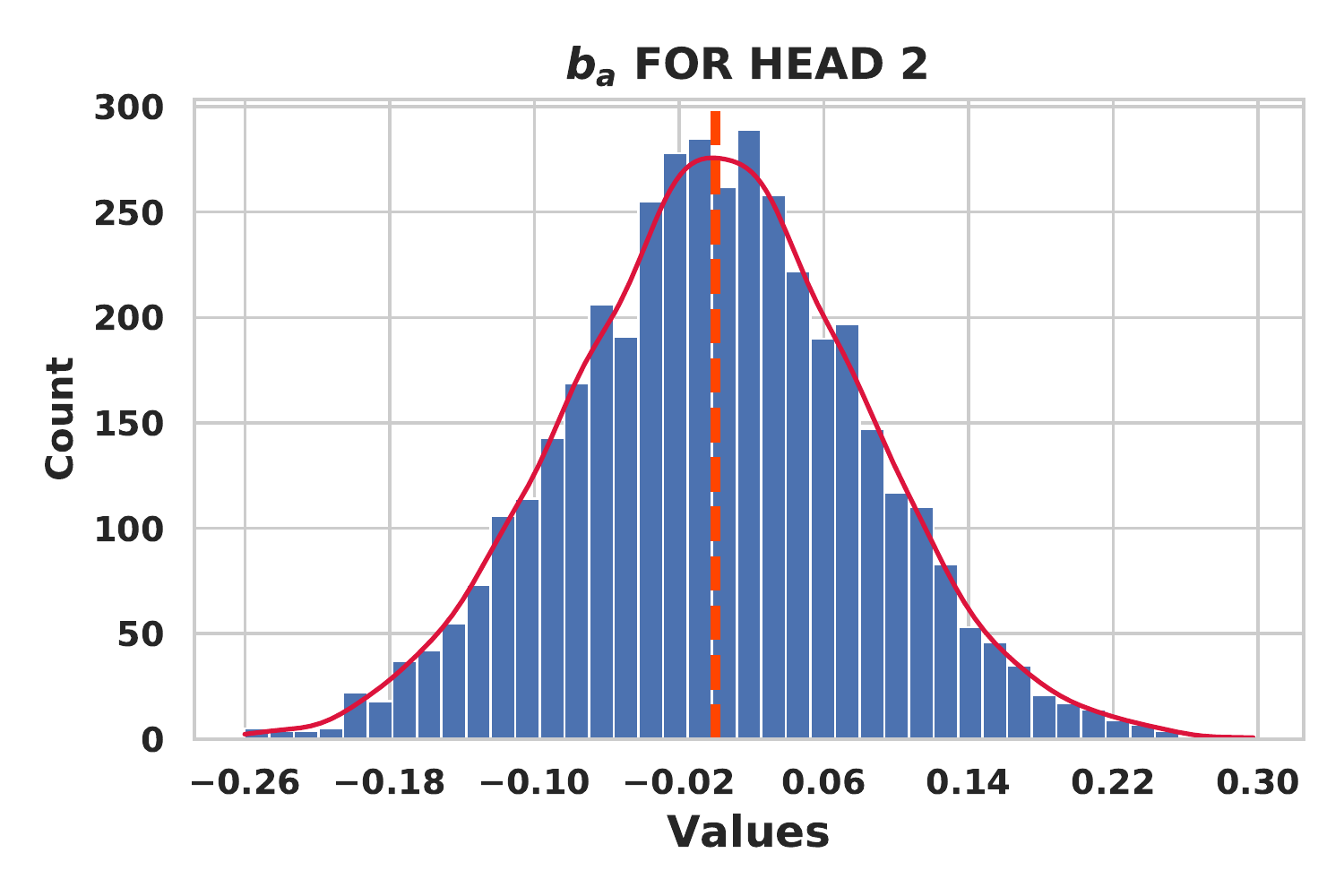}

\end{subfigure}
\hfill
\begin{subfigure}[b]{0.6\textwidth}
	\centering
	\includegraphics[width=1.1\textwidth]{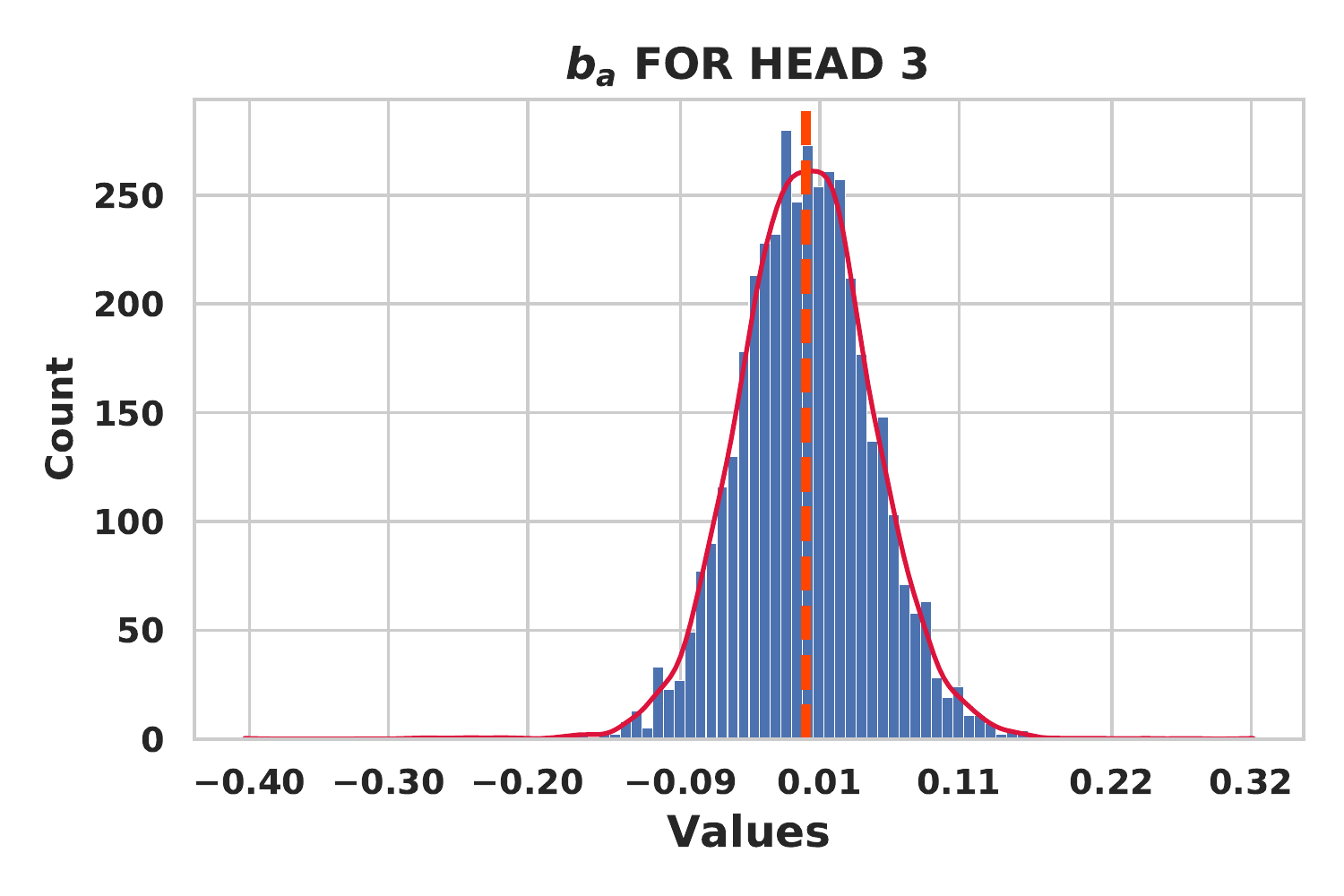}

\end{subfigure}
\hfill
\begin{subfigure}[b]{0.6\textwidth}
	\centering
	\includegraphics[width=1.1\textwidth]{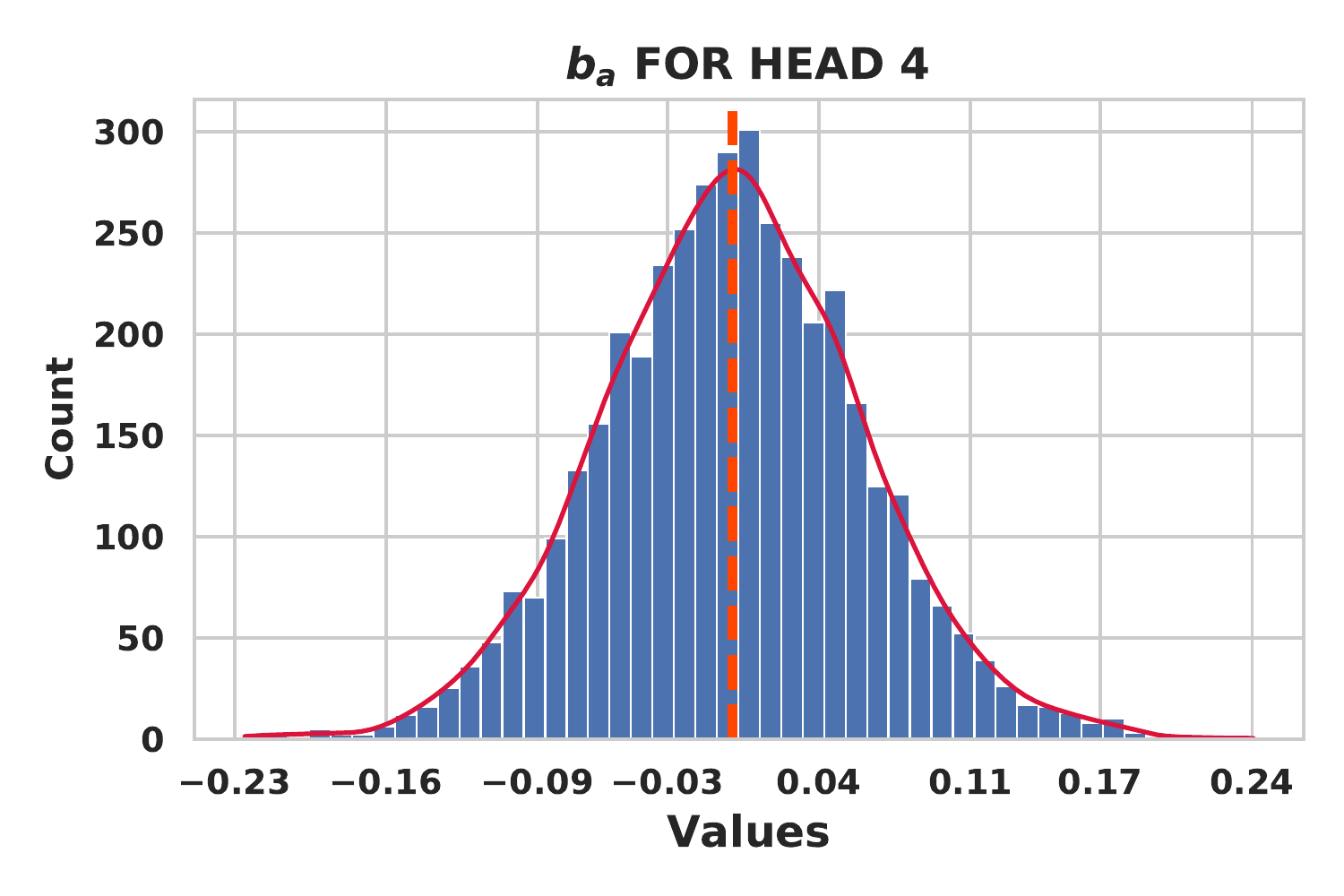}

\end{subfigure}
\centering
\begin{subfigure}[b]{0.6\textwidth}
	\centering
	\includegraphics[width=1.1\textwidth]{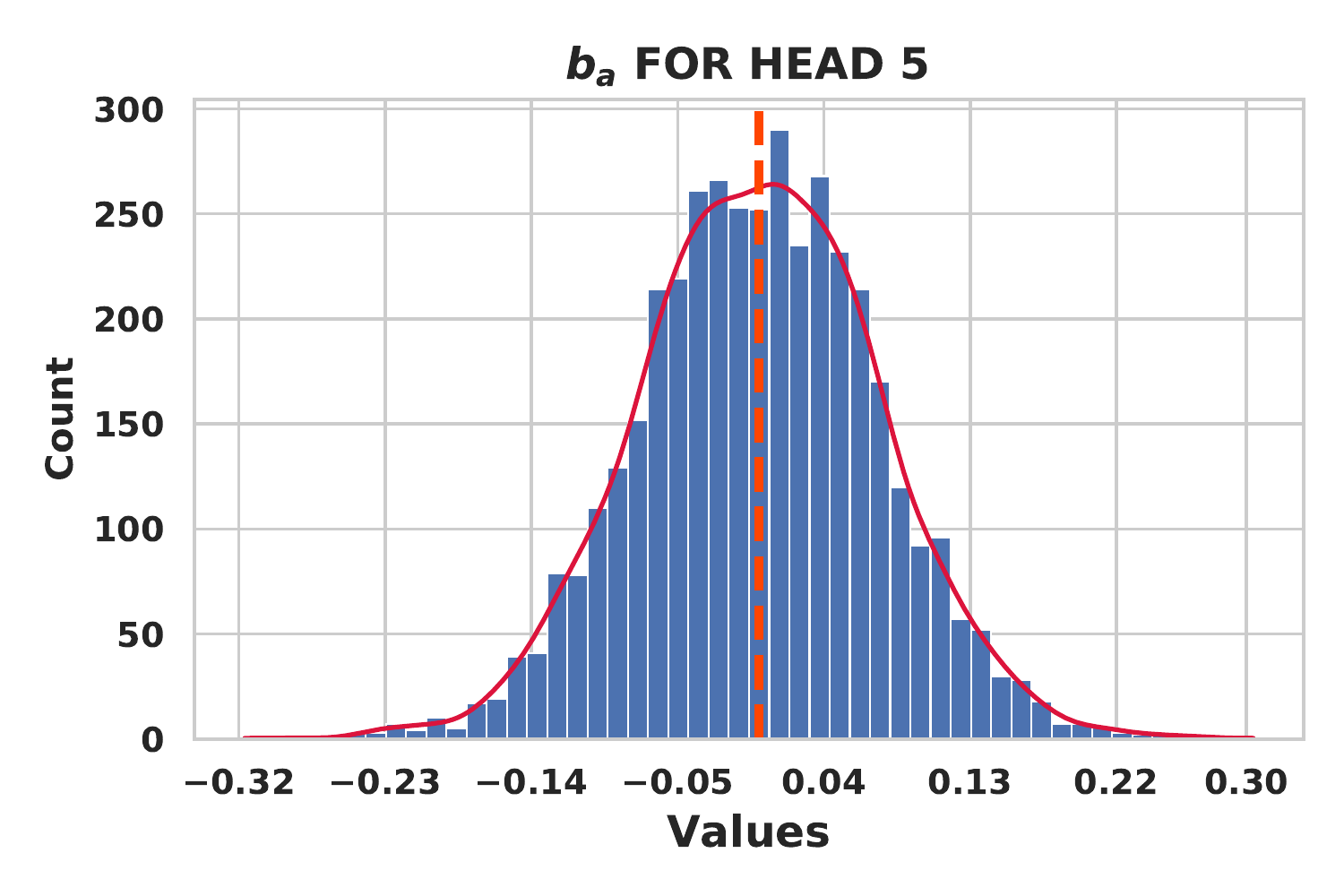}

\end{subfigure}
\hfill
\begin{subfigure}[b]{0.6\textwidth}
	\centering
	\includegraphics[width=1.1\textwidth]{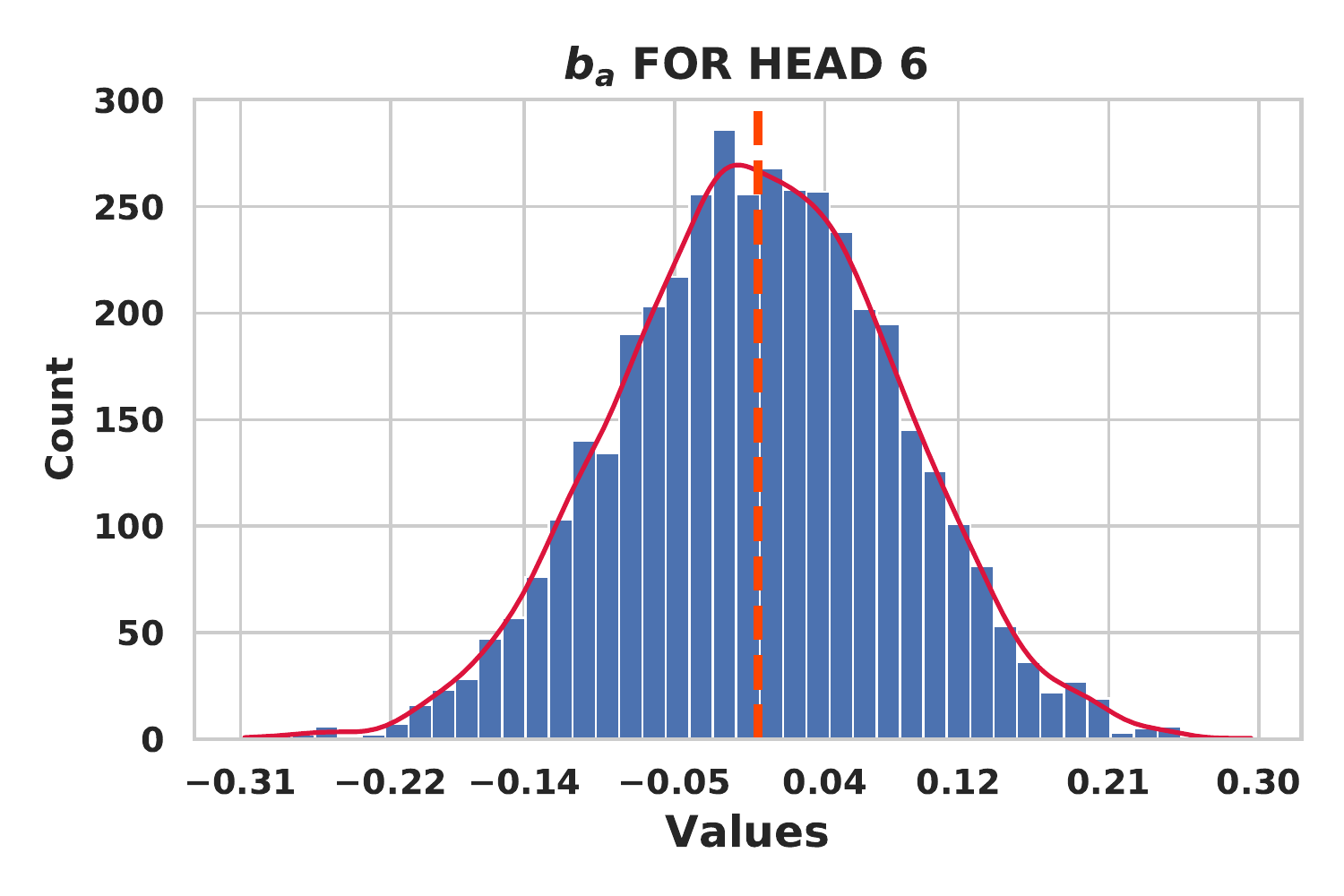}

\end{subfigure}
\hfill
\begin{subfigure}[b]{0.6\textwidth}
	\centering
	\includegraphics[width=1.1\textwidth]{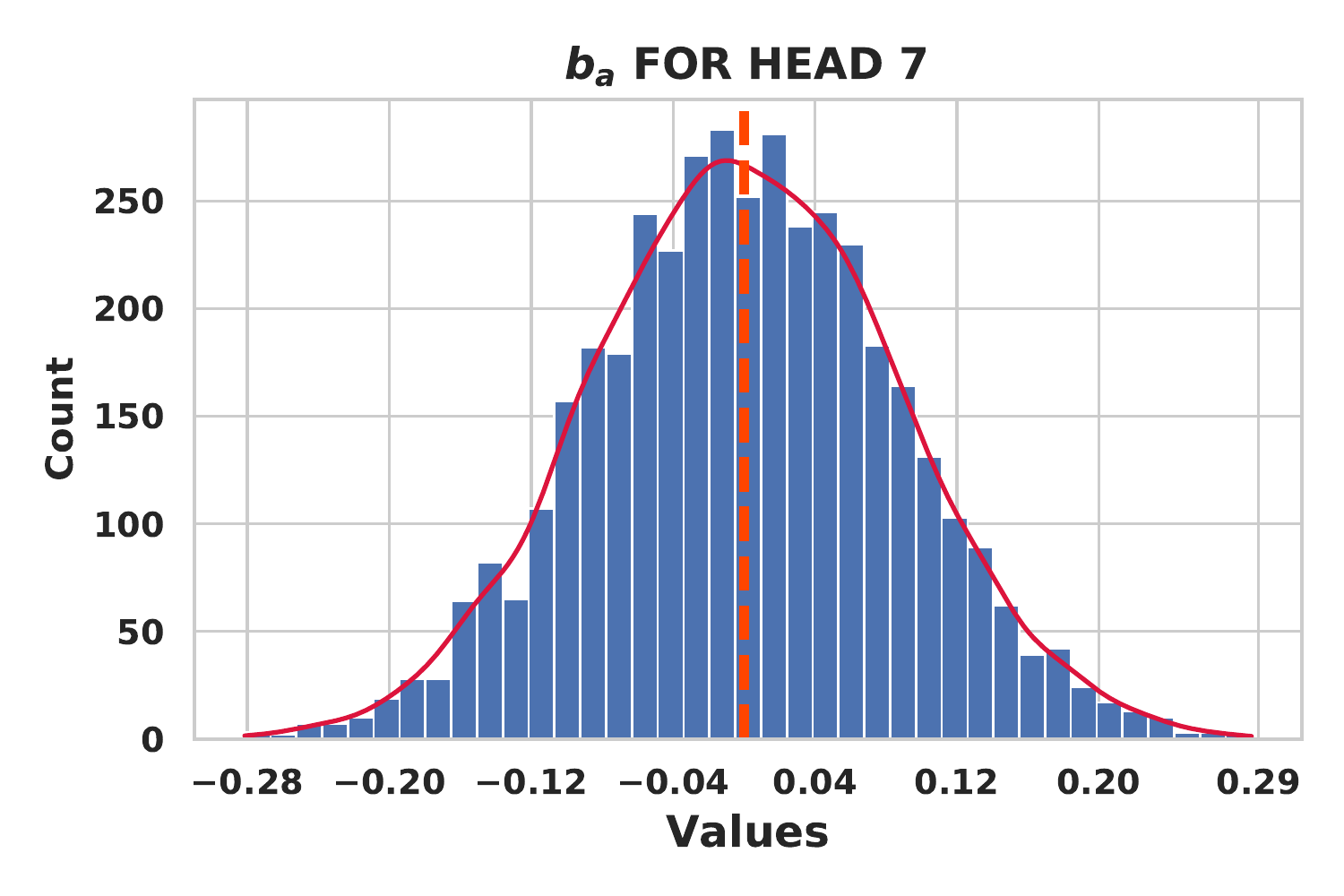}

\end{subfigure}
\hfill
\begin{subfigure}[b]{0.6\textwidth}
	\centering
	\includegraphics[width=1.1\textwidth]{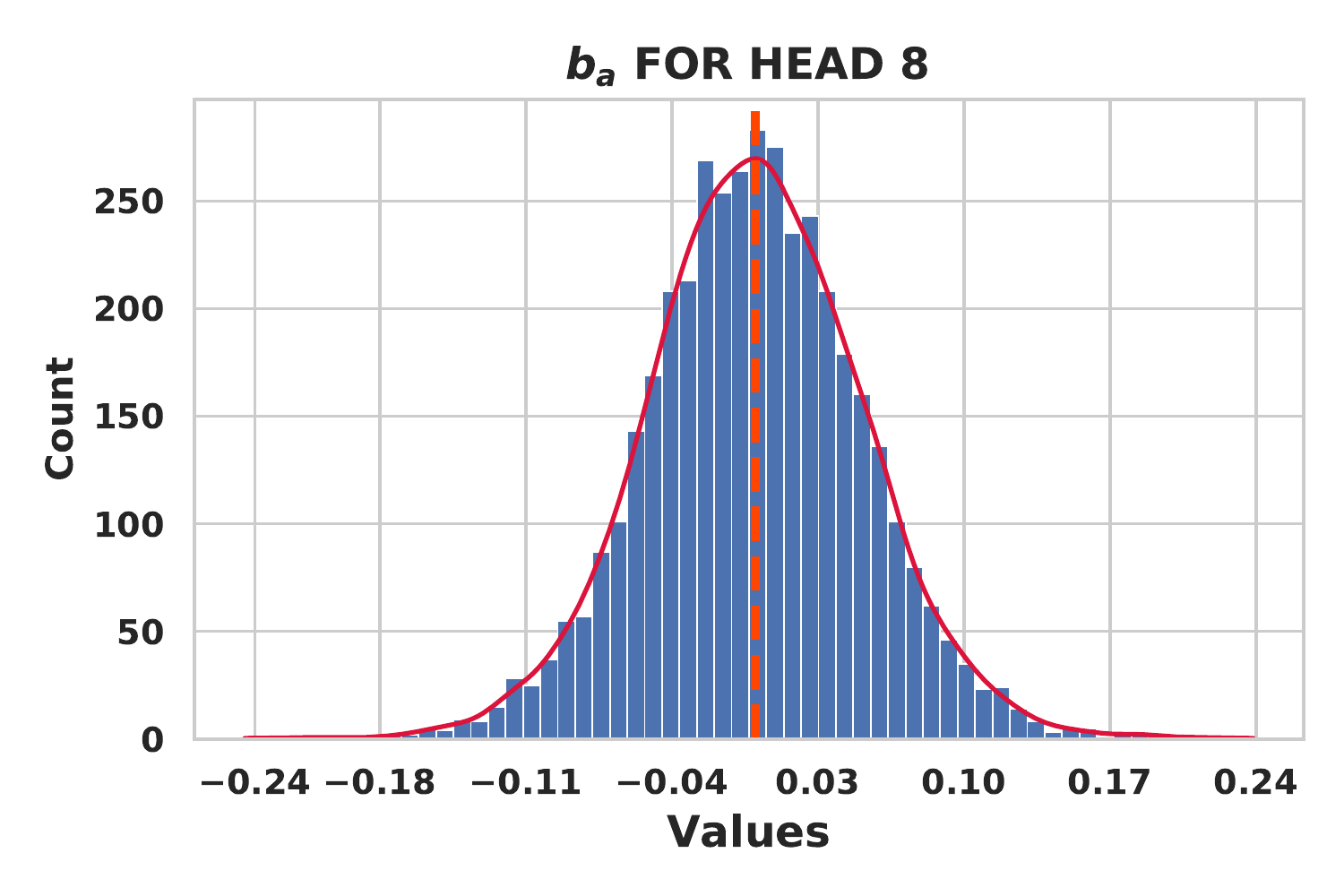}

\end{subfigure}
\caption{$\vb_{a}$ histogram plots for all heads from TLM attention stage from graph transformer model \#2 for PT-EN translation task. Dashed line in orange marks zero value.}
\label{fig29apx}
\end{adjustwidth}
\end{figure} 

\clearpage
\thispagestyle{headings}

\begin{figure}
\begin{adjustwidth}{-5em}{-5em}
\centering
\begin{subfigure}[b]{0.6\textwidth}
	\centering
	\includegraphics[width=1.1\textwidth]{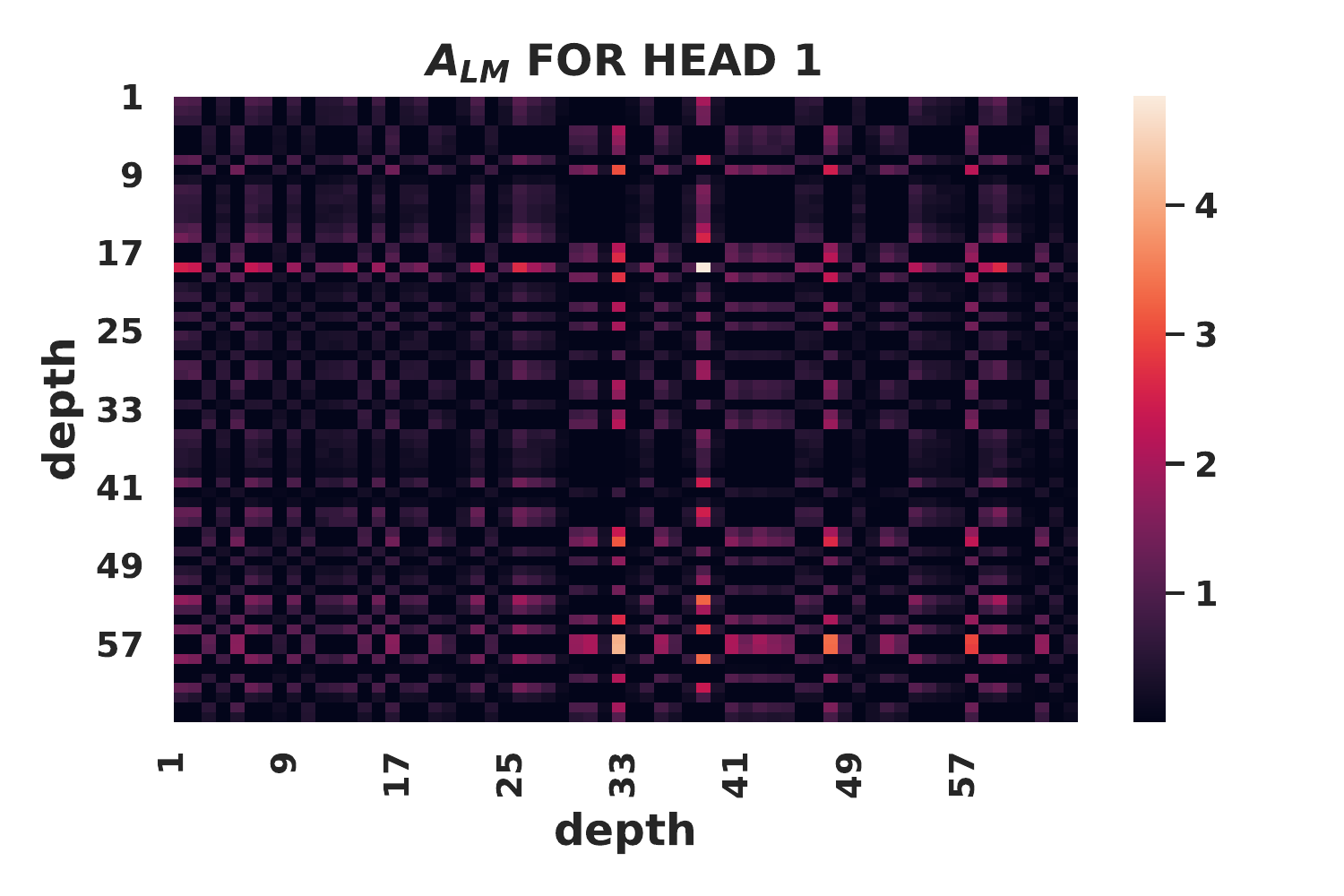}

\end{subfigure}
\hfill
\begin{subfigure}[b]{0.6\textwidth}
	\centering
	\includegraphics[width=1.1\textwidth]{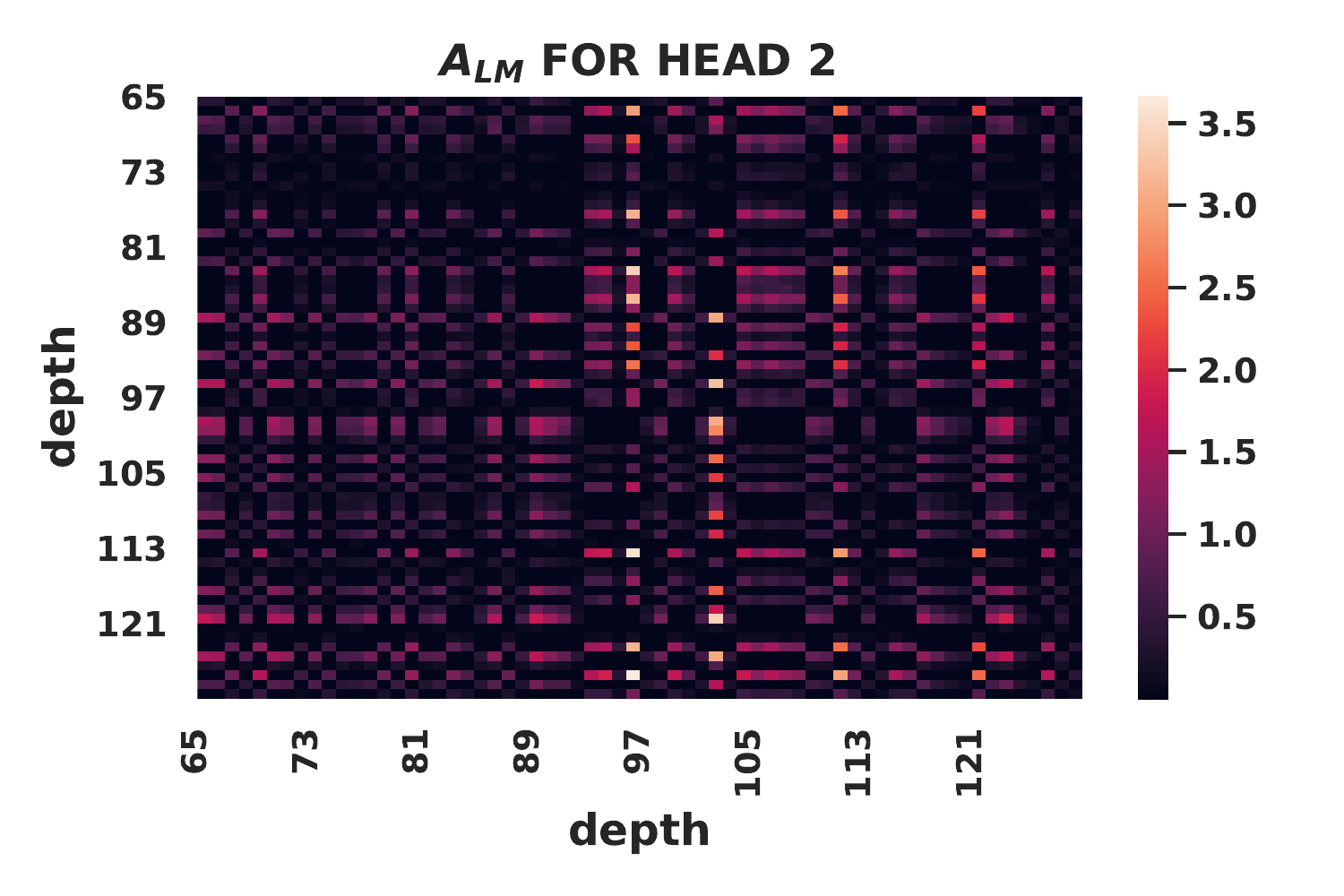}

\end{subfigure}
\hfill
\begin{subfigure}[b]{0.6\textwidth}
	\centering
	\includegraphics[width=1.1\textwidth]{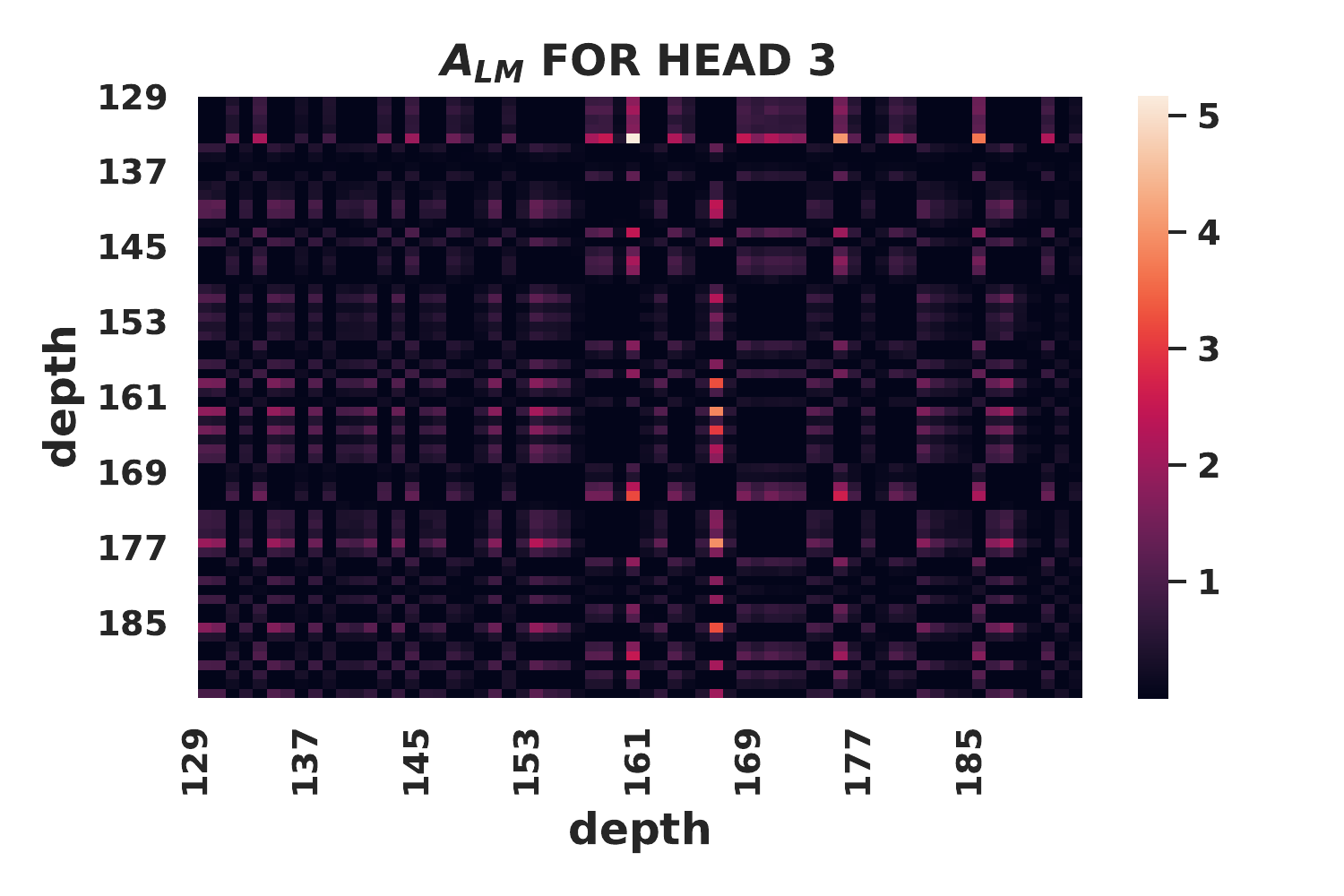}

\end{subfigure}
\hfill
\begin{subfigure}[b]{0.6\textwidth}
	\centering
	\includegraphics[width=1.1\textwidth]{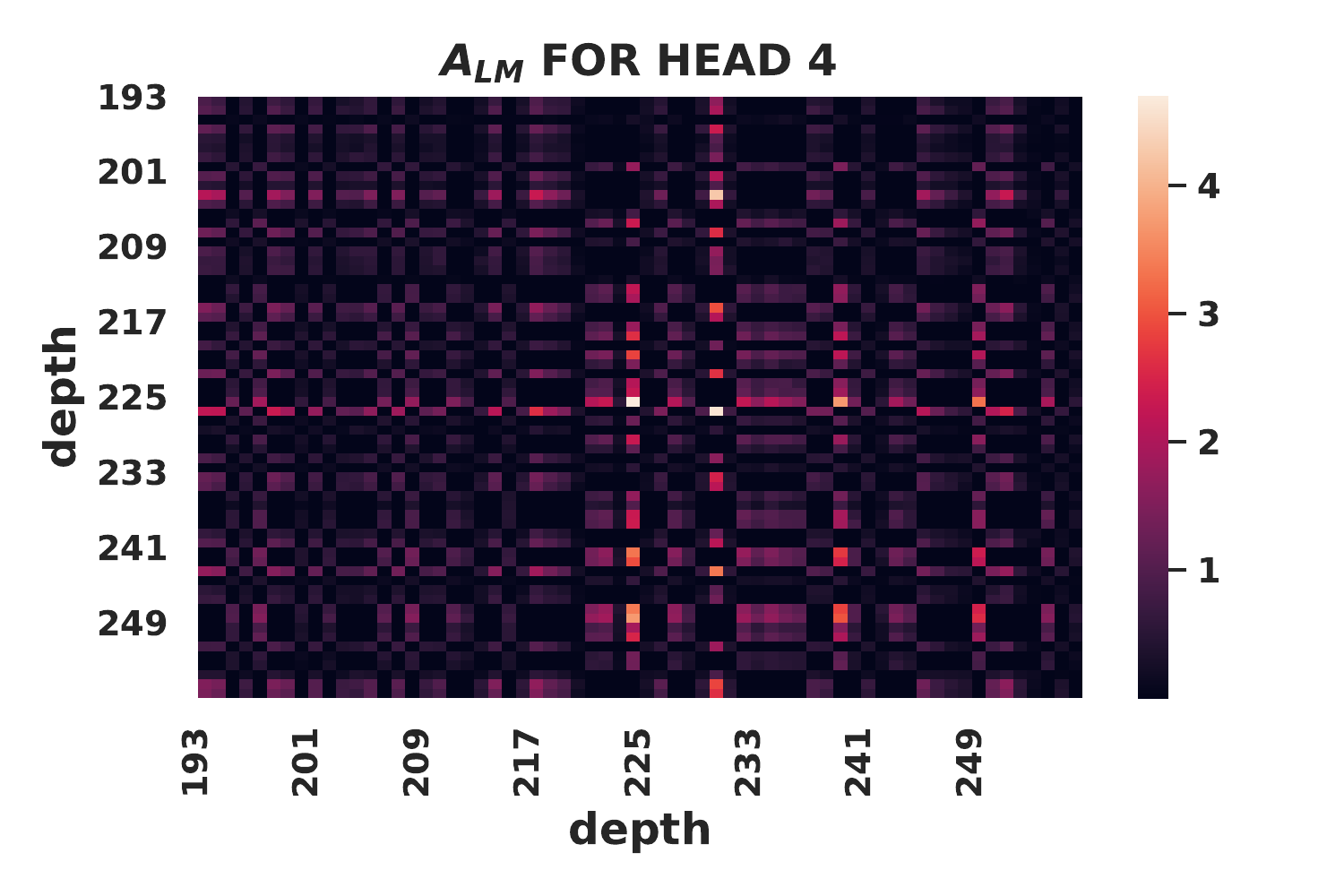}

\end{subfigure}
\centering
\begin{subfigure}[b]{0.6\textwidth}
	\centering
	\includegraphics[width=1.1\textwidth]{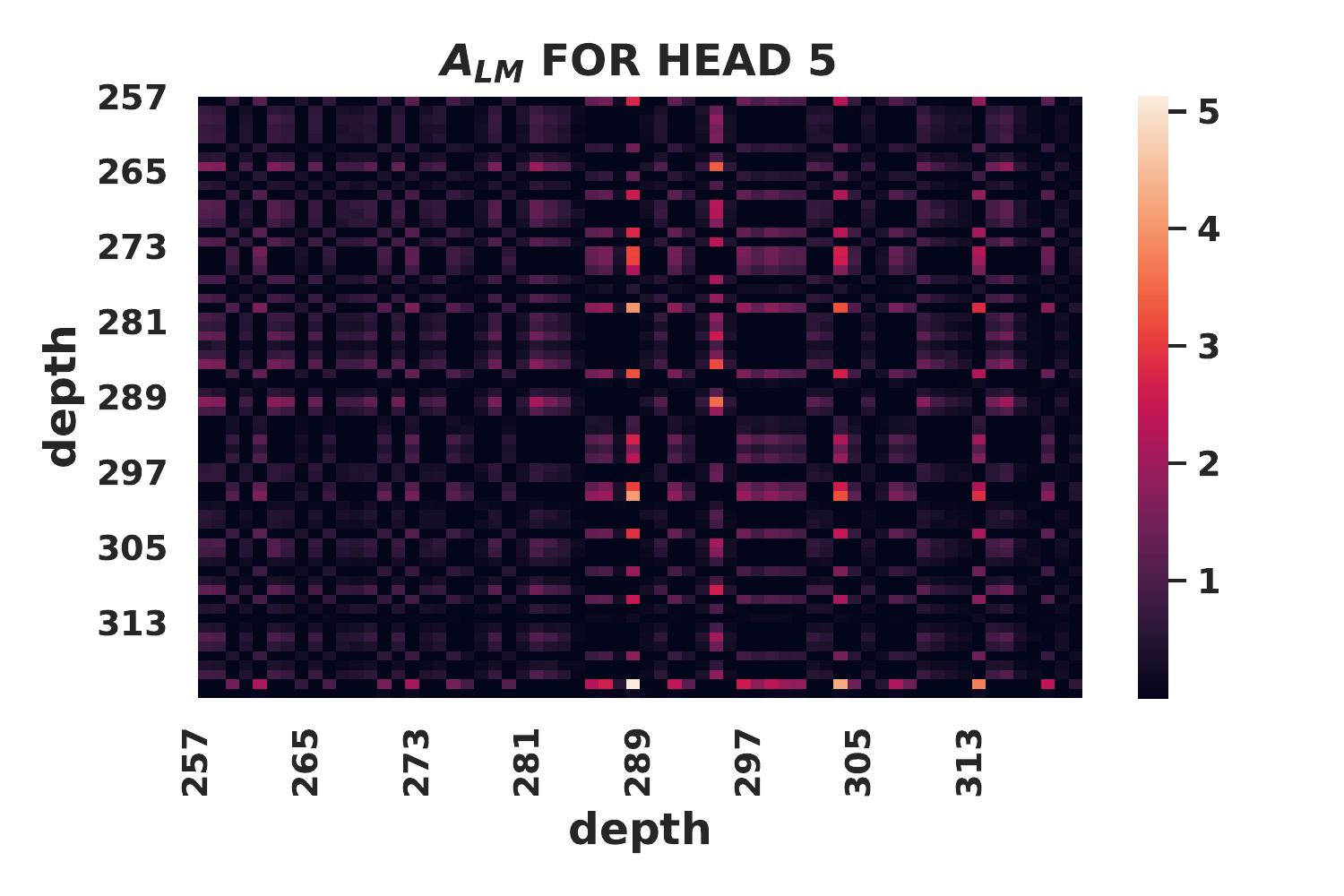}

\end{subfigure}
\hfill
\begin{subfigure}[b]{0.6\textwidth}
	\centering
	\includegraphics[width=1.1\textwidth]{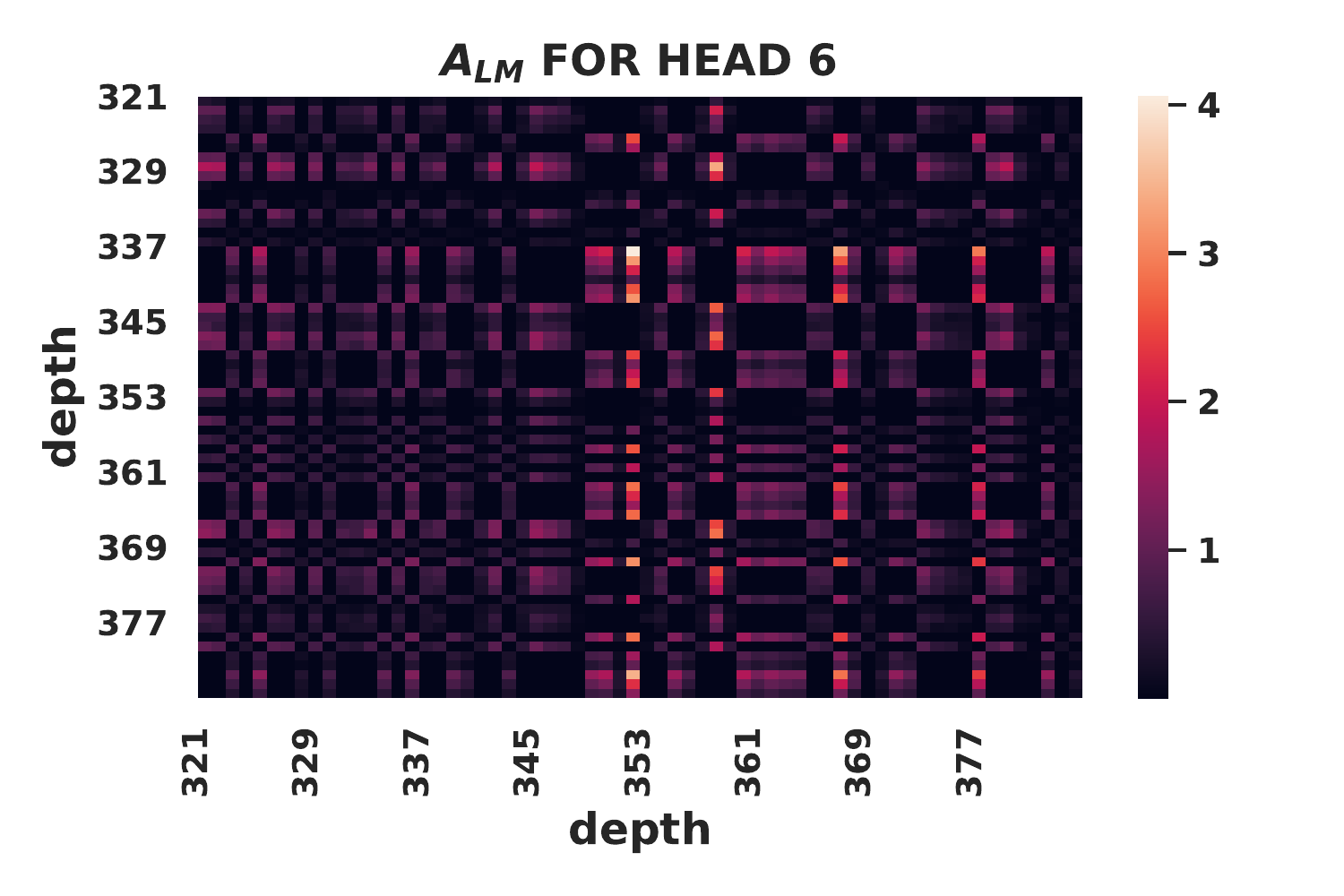}

\end{subfigure}
\hfill
\begin{subfigure}[b]{0.6\textwidth}
	\centering
	\includegraphics[width=1.1\textwidth]{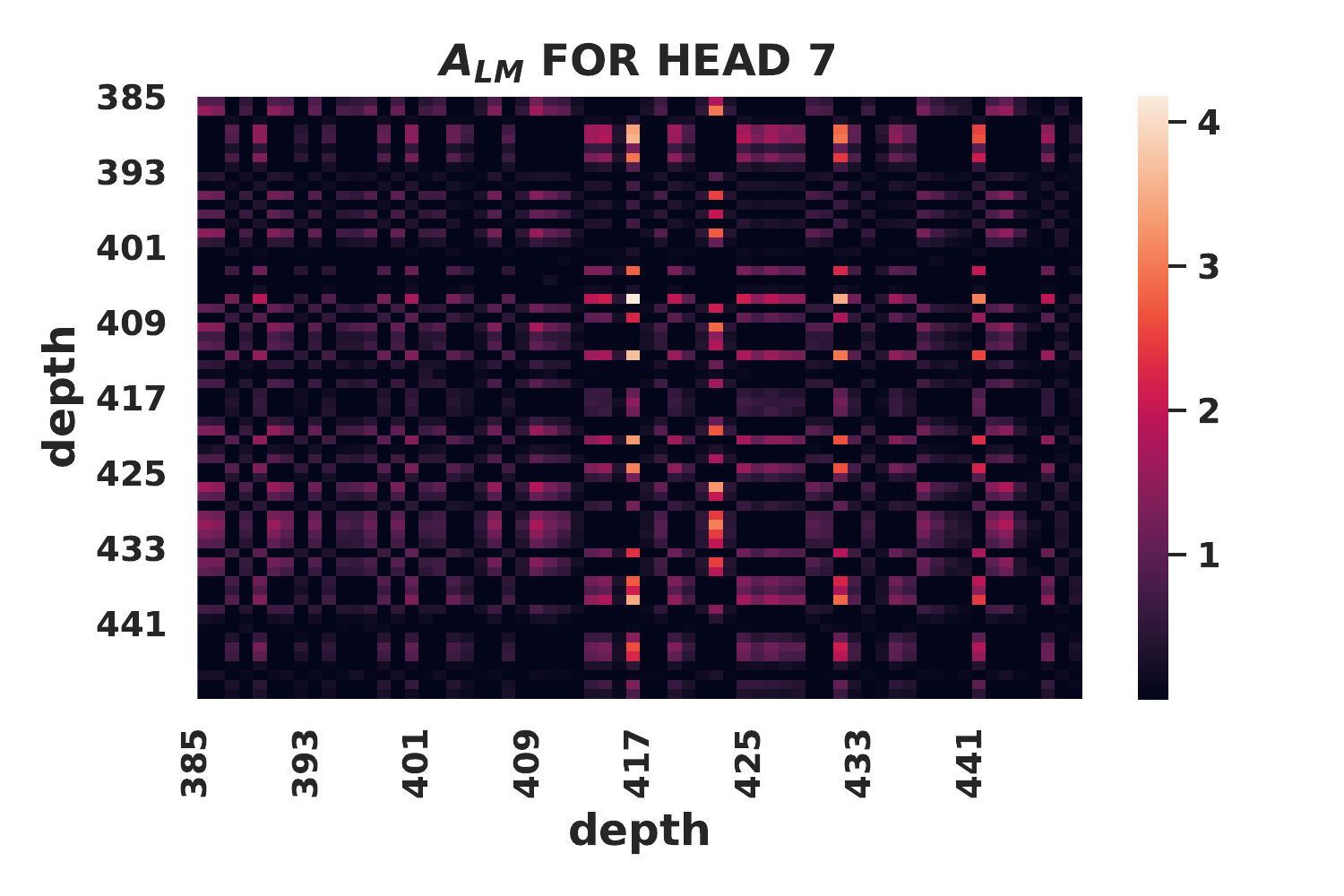}

\end{subfigure}
\hfill
\begin{subfigure}[b]{0.6\textwidth}
	\centering
	\includegraphics[width=1.1\textwidth]{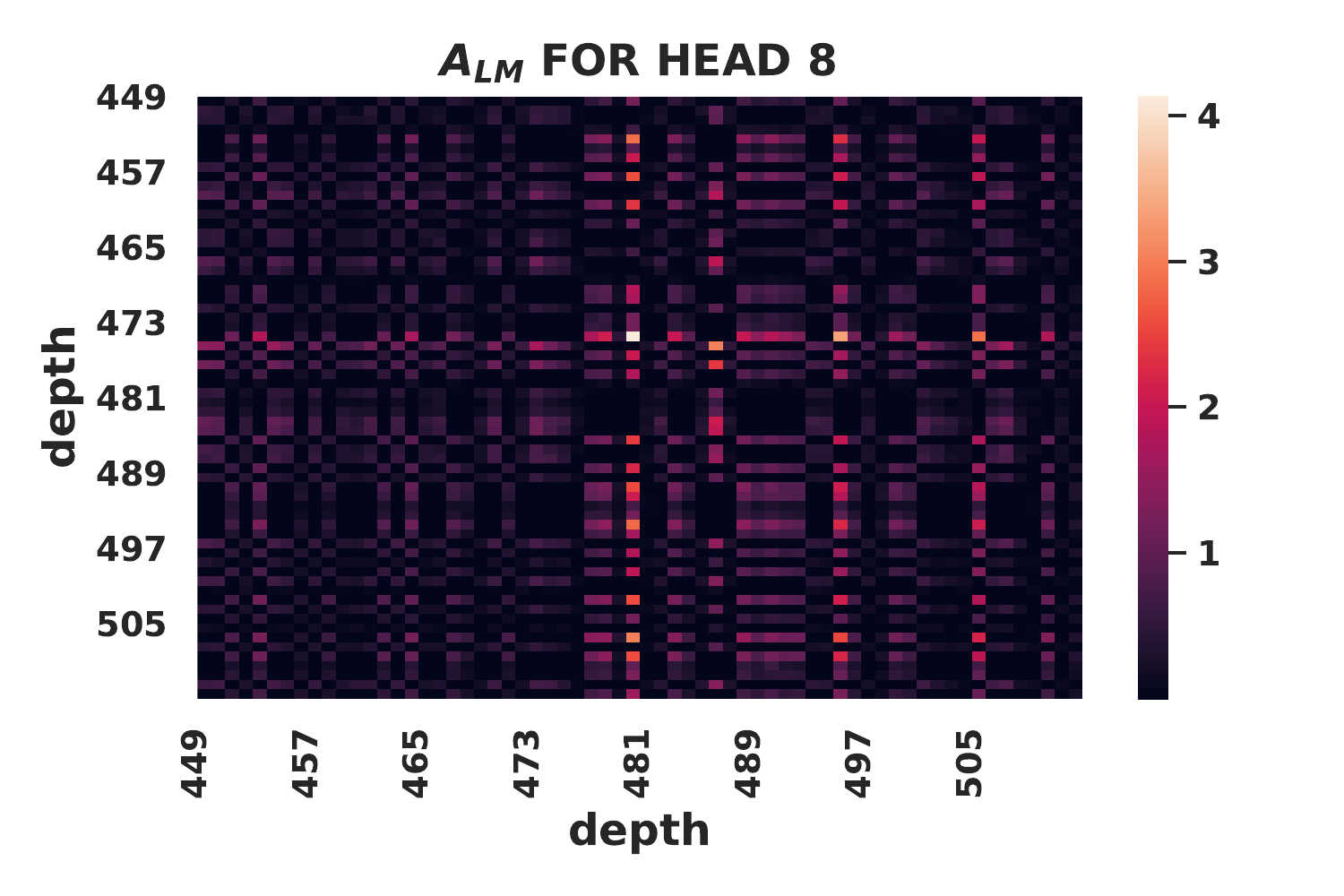}

\end{subfigure}
\caption{$\mA_{LM}$ heatmap plots for all heads from TLM attention stage from graph transformer model \#2 for PT-EN translation task.}
\label{fig30apx}
\end{adjustwidth}
\end{figure} 

\clearpage
\thispagestyle{headings}

\begin{figure}
\begin{adjustwidth}{-5em}{-5em}
\centering
\begin{subfigure}[b]{0.6\textwidth}
	\centering
	\includegraphics[width=1.1\textwidth]{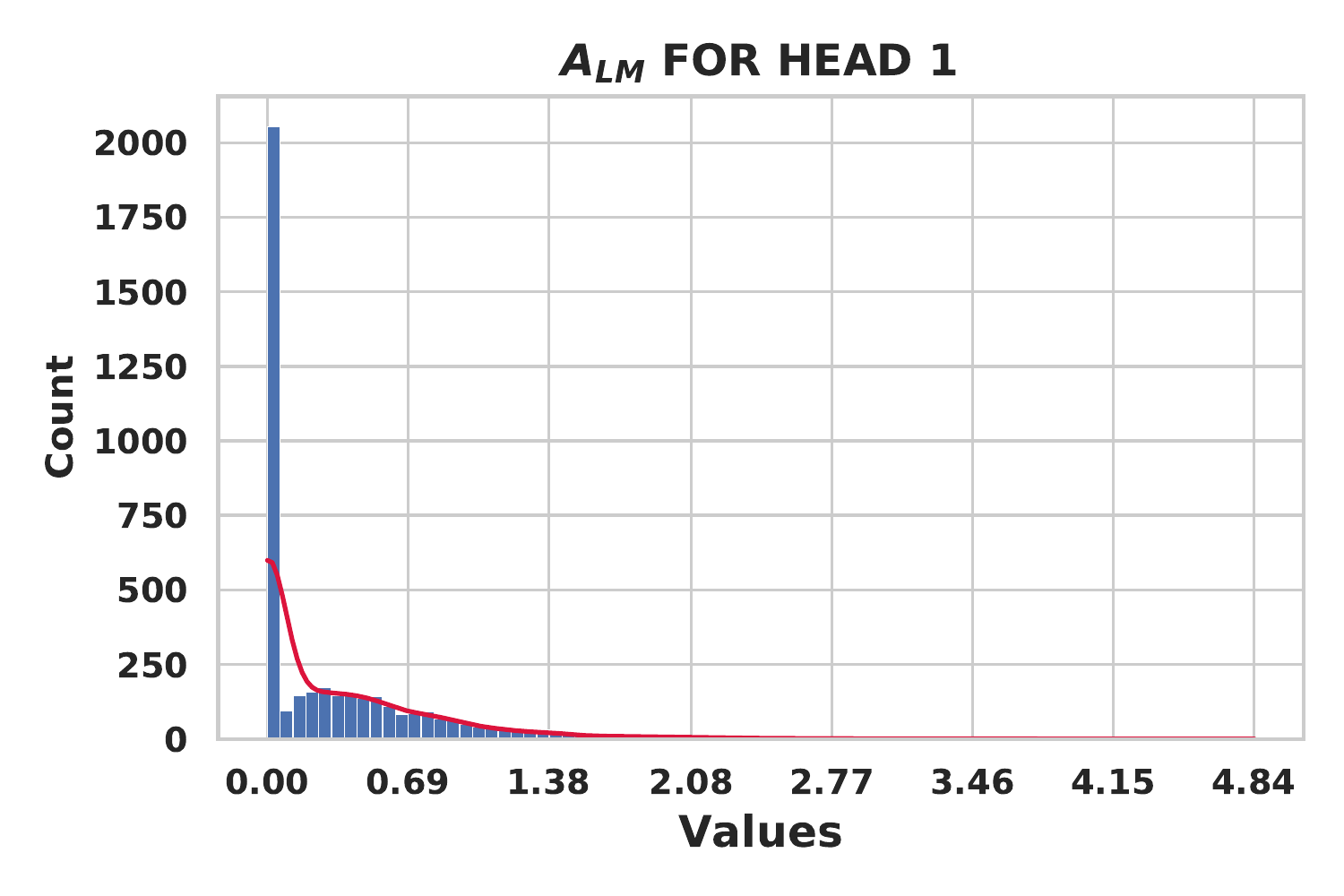}

\end{subfigure}
\hfill
\begin{subfigure}[b]{0.6\textwidth}
	\centering
	\includegraphics[width=1.1\textwidth]{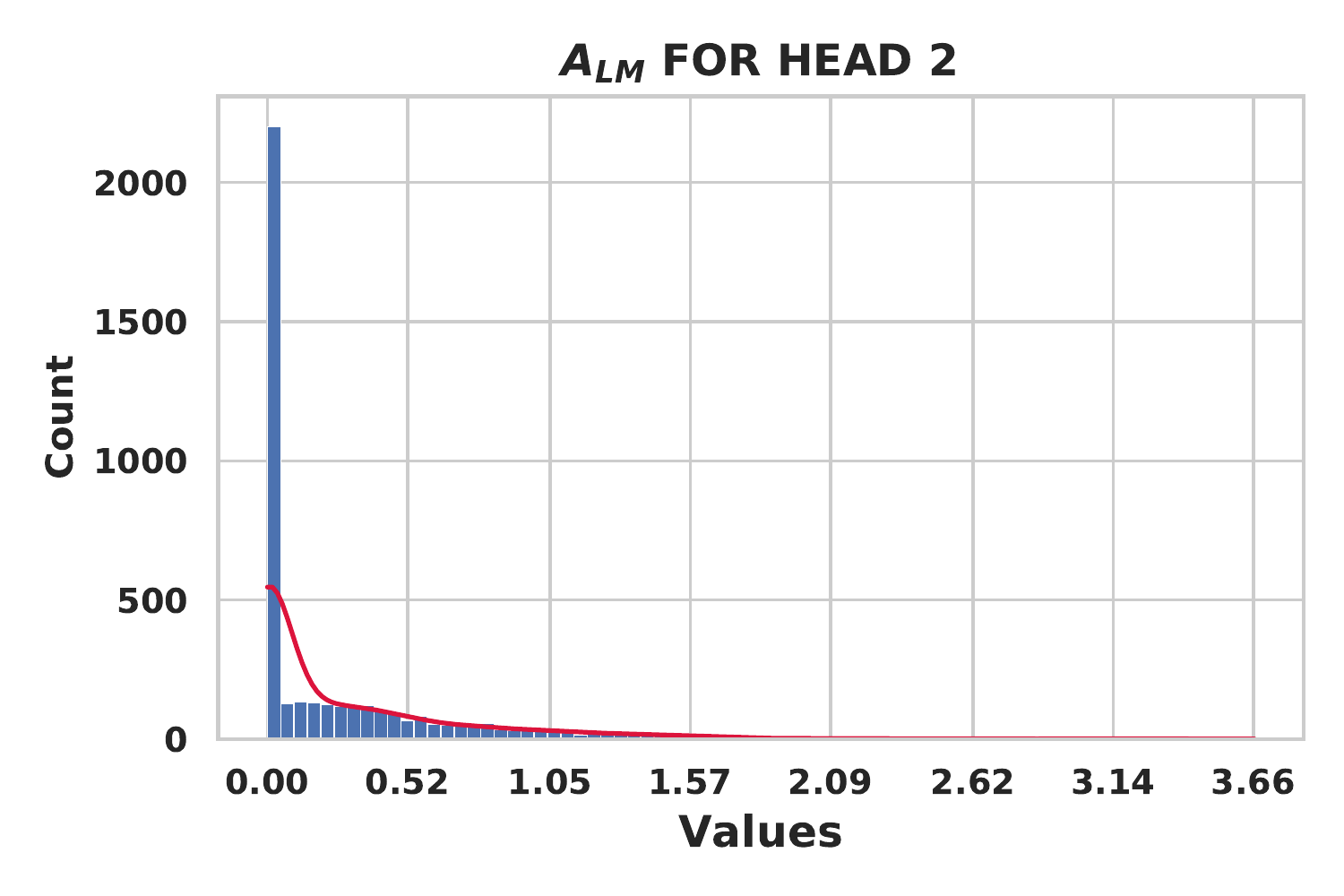}

\end{subfigure}
\hfill
\begin{subfigure}[b]{0.6\textwidth}
	\centering
	\includegraphics[width=1.1\textwidth]{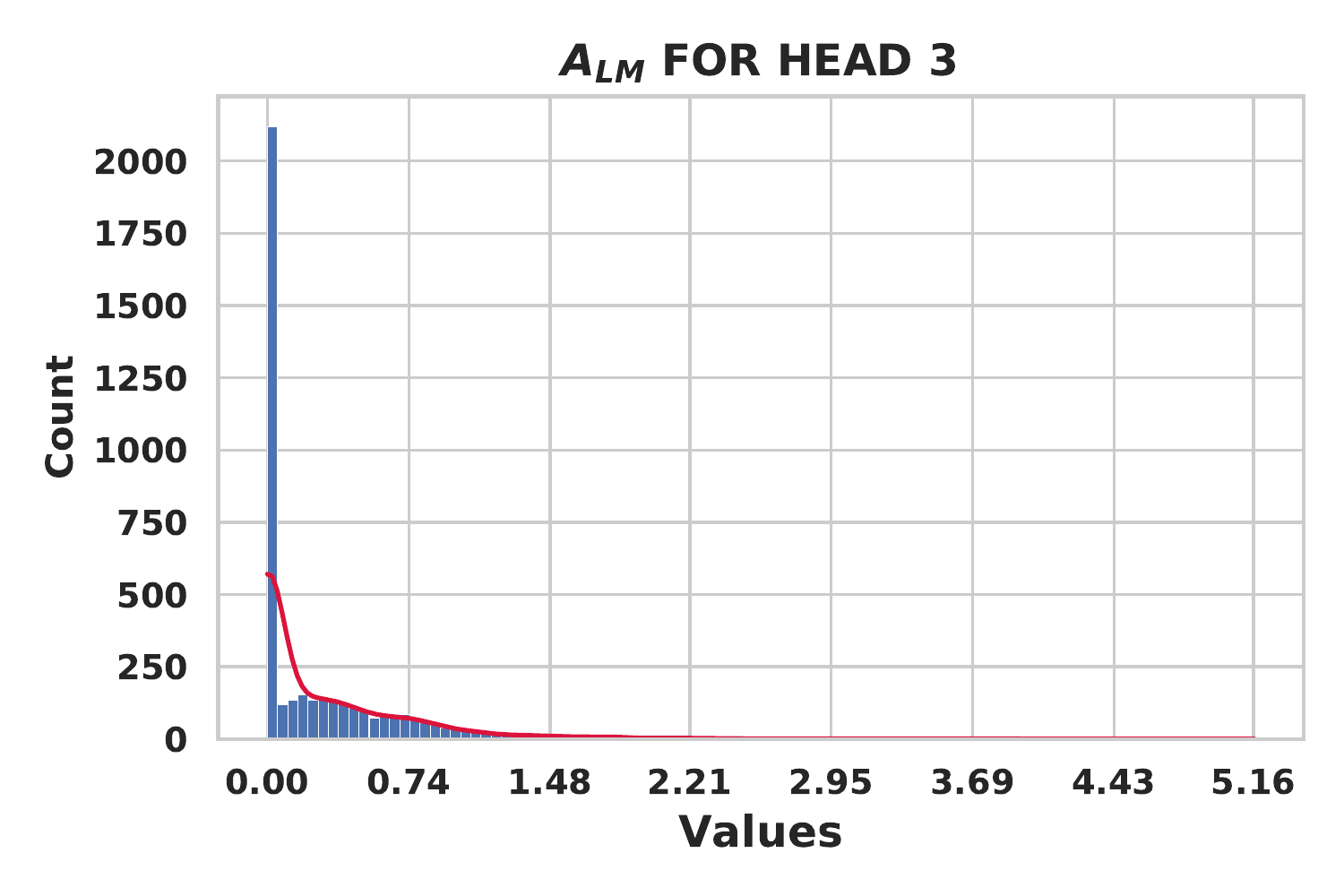}

\end{subfigure}
\hfill
\begin{subfigure}[b]{0.6\textwidth}
	\centering
	\includegraphics[width=1.1\textwidth]{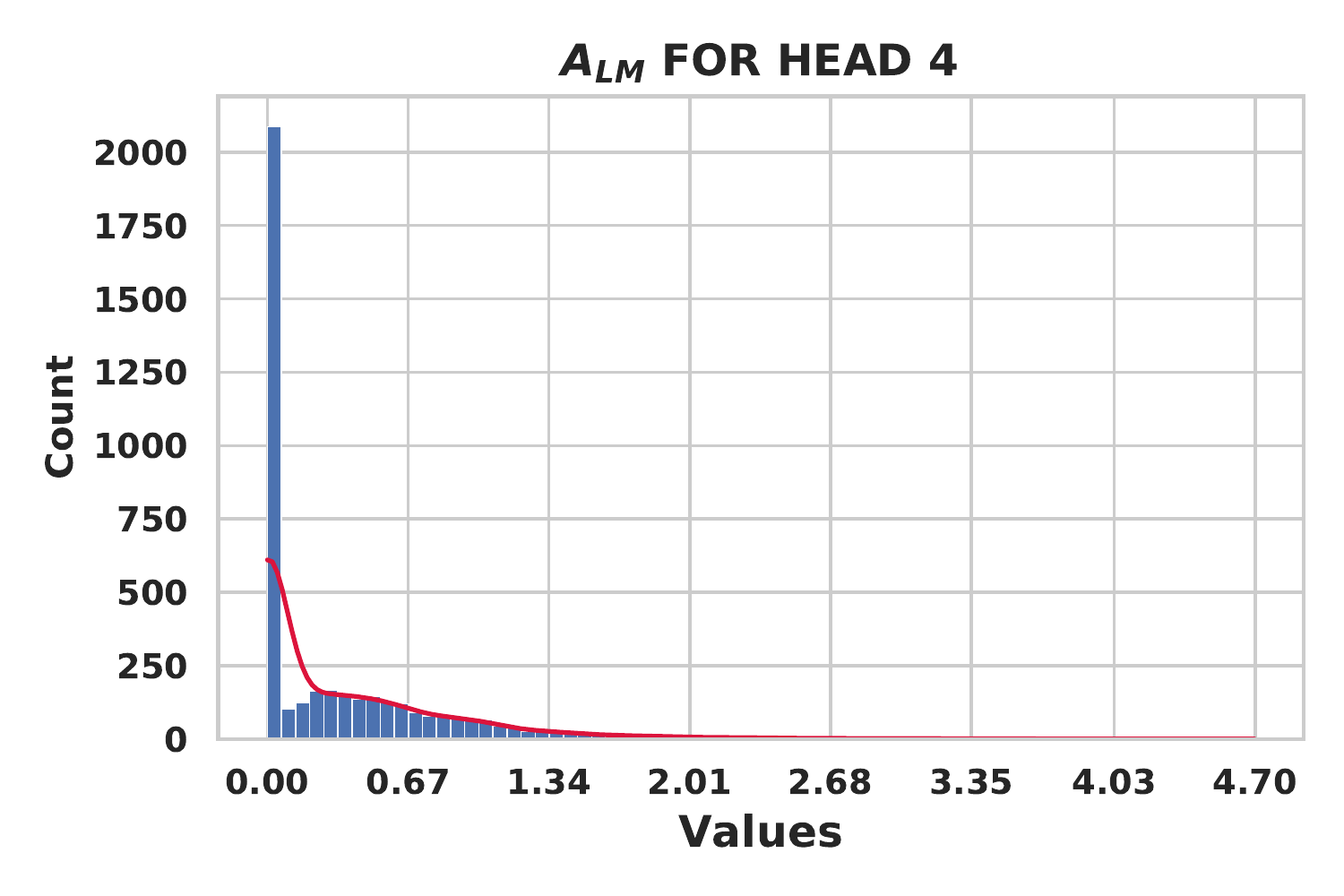}

\end{subfigure}
\centering
\begin{subfigure}[b]{0.6\textwidth}
	\centering
	\includegraphics[width=1.1\textwidth]{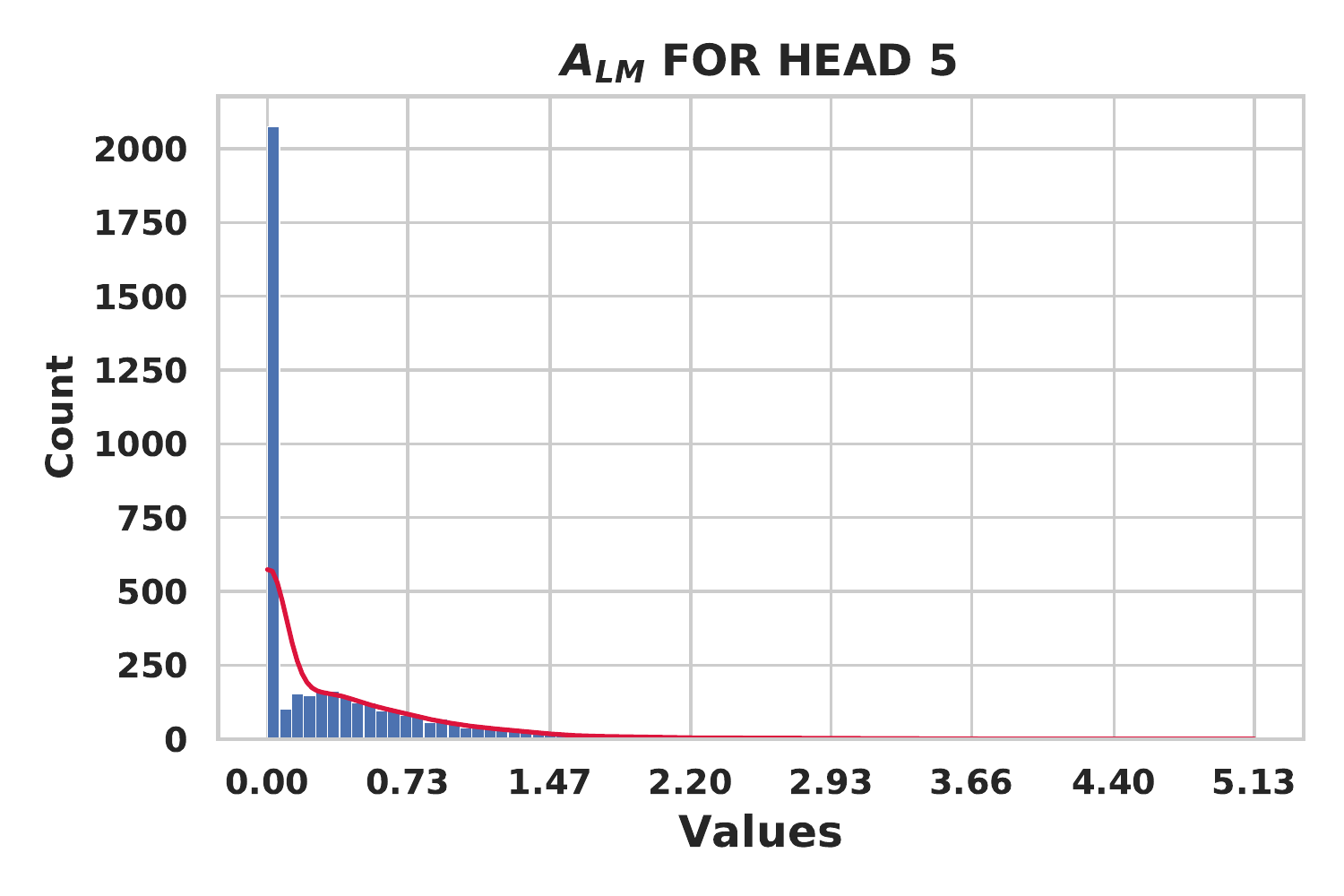}

\end{subfigure}
\hfill
\begin{subfigure}[b]{0.6\textwidth}
	\centering
	\includegraphics[width=1.1\textwidth]{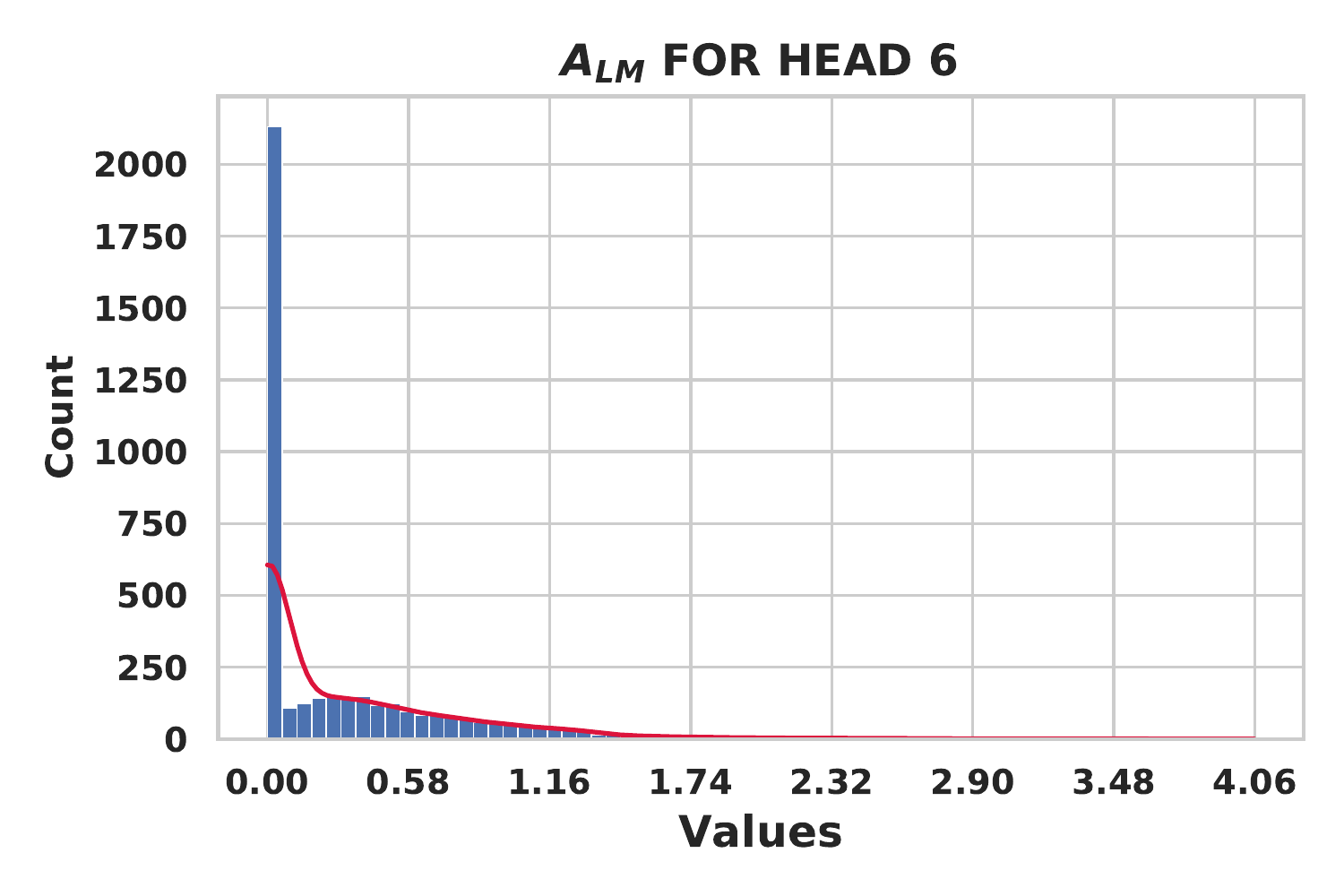}

\end{subfigure}
\hfill
\begin{subfigure}[b]{0.6\textwidth}
	\centering
	\includegraphics[width=1.1\textwidth]{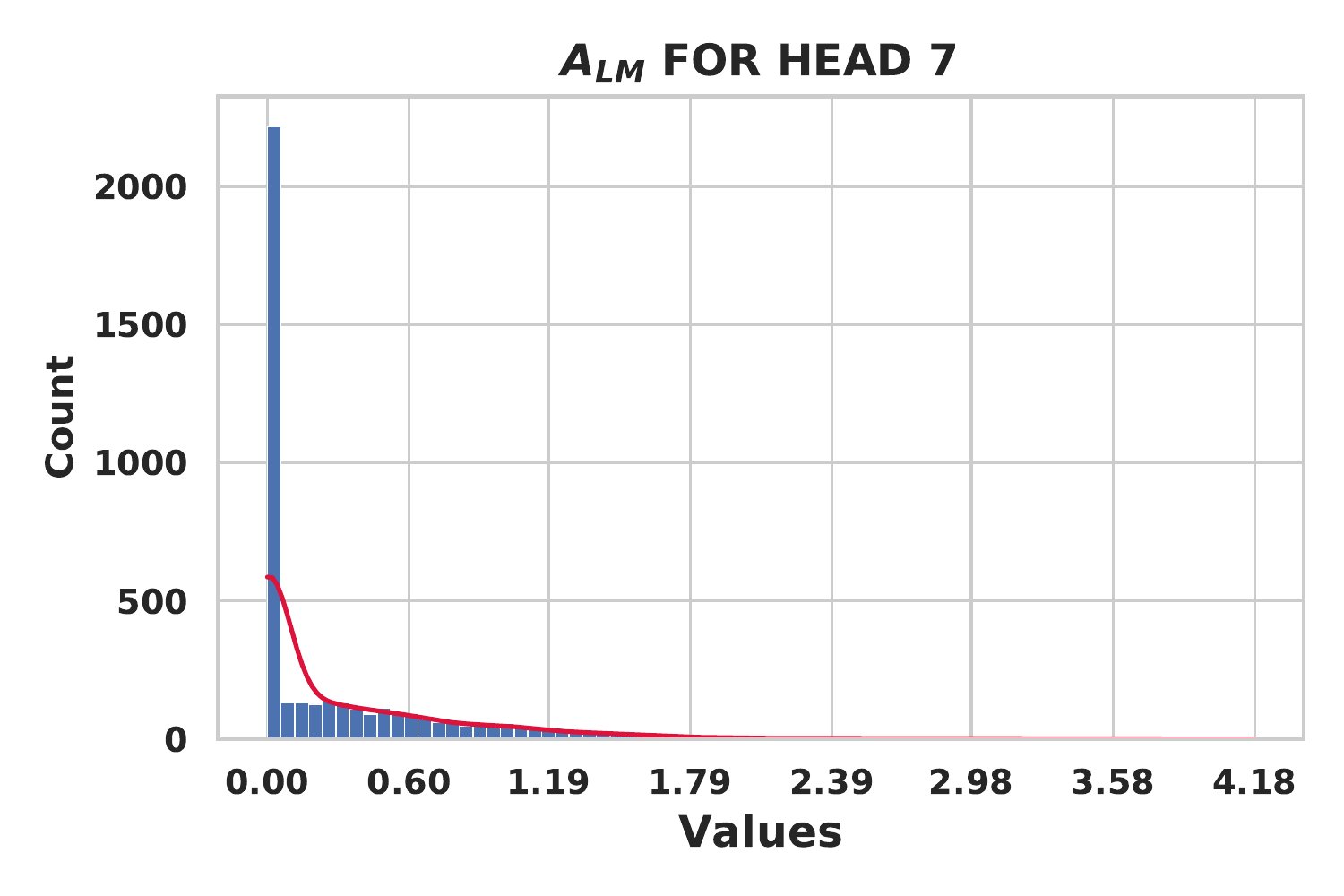}

\end{subfigure}
\hfill
\begin{subfigure}[b]{0.6\textwidth}
	\centering
	\includegraphics[width=1.1\textwidth]{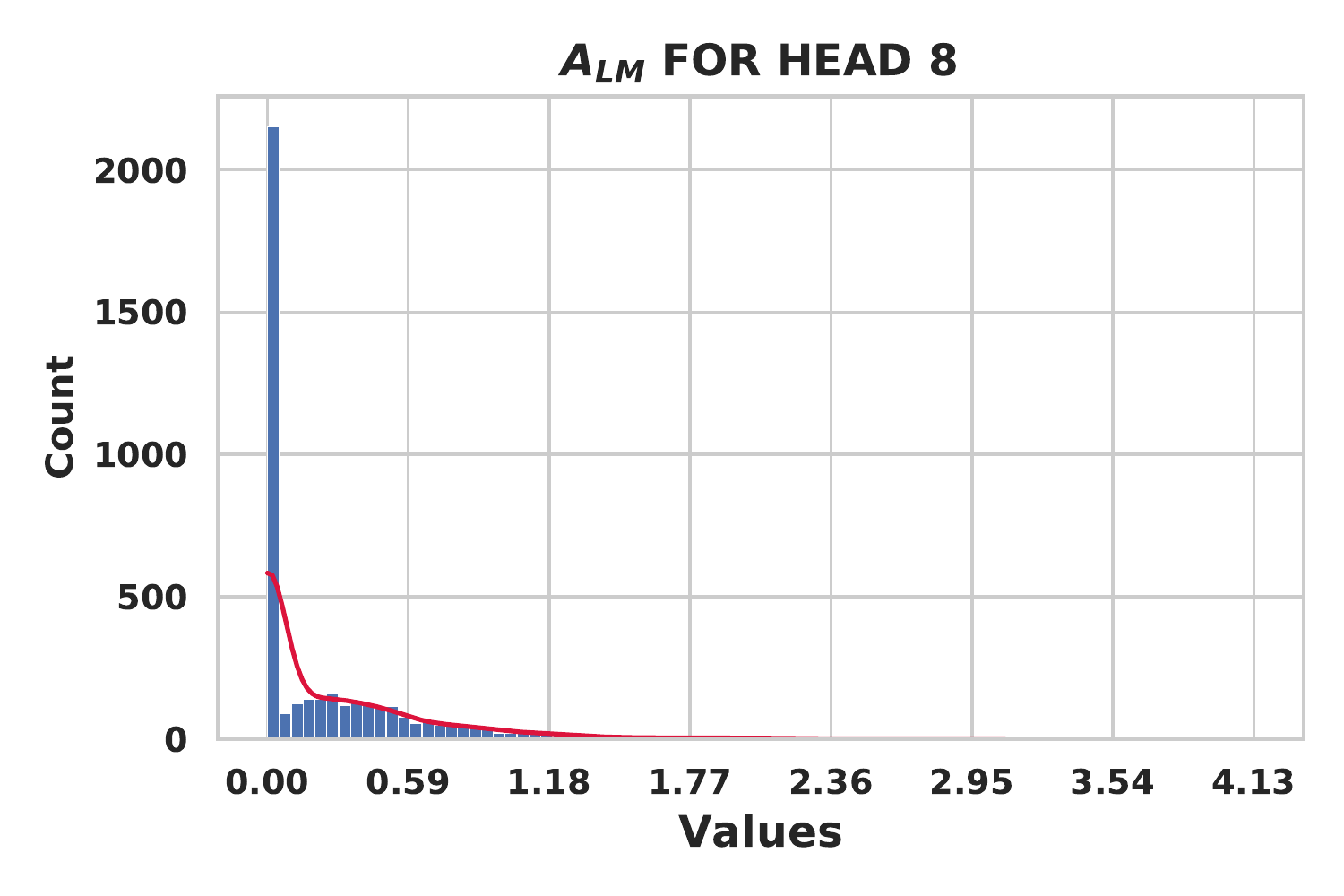}

\end{subfigure}
\caption{$\mA_{LM}$ histogram plots for all heads from TLM attention stage from graph transformer model \#2 for PT-EN translation task.}
\label{fig31apx}
\end{adjustwidth}
\end{figure}

\clearpage
\thispagestyle{headings}

\begin{figure}
\begin{adjustwidth}{-5em}{-5em}
\centering
\begin{subfigure}[b]{0.6\textwidth}
	\centering
	\includegraphics[width=1.1\textwidth]{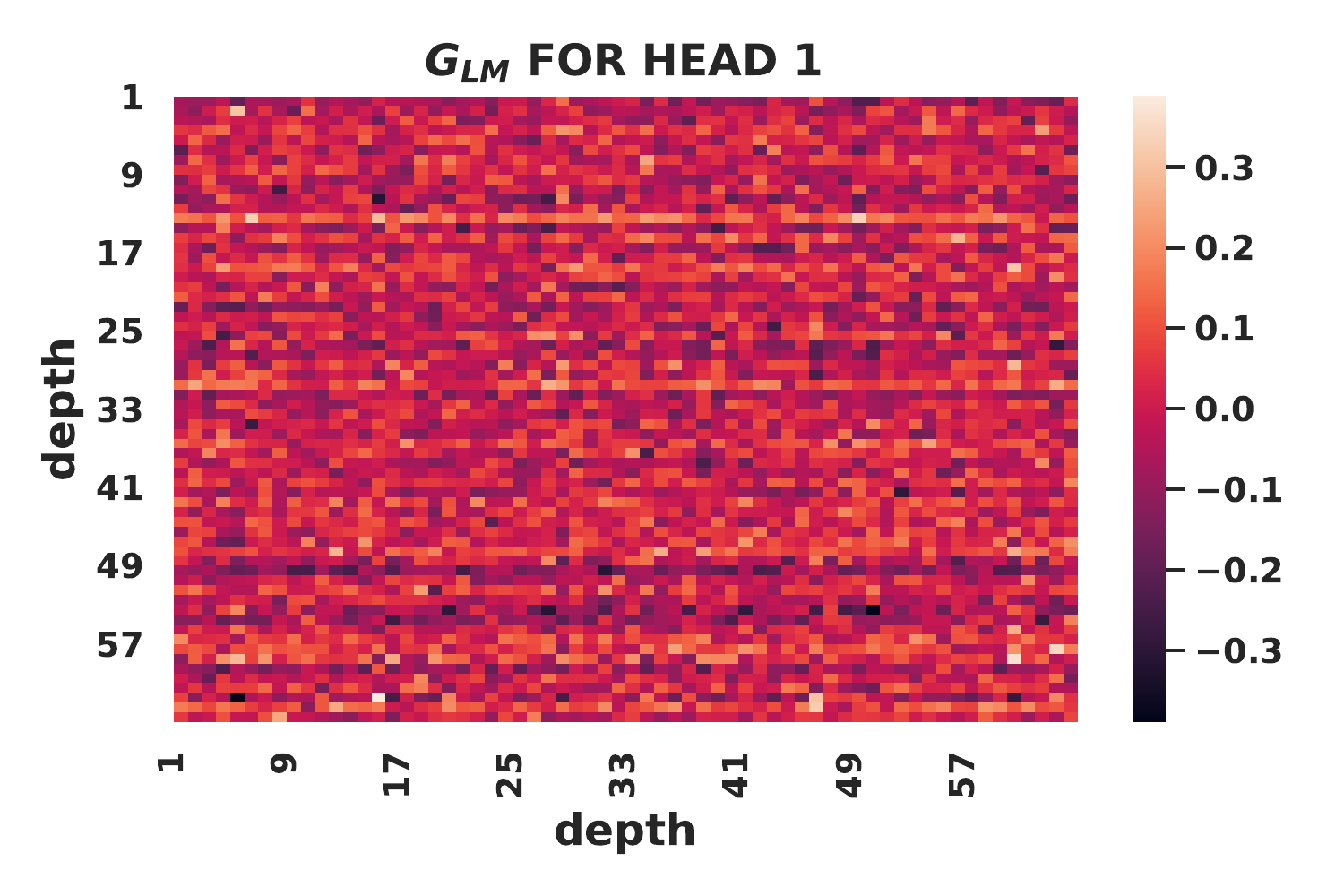}

\end{subfigure}
\hfill
\begin{subfigure}[b]{0.6\textwidth}
	\centering
	\includegraphics[width=1.1\textwidth]{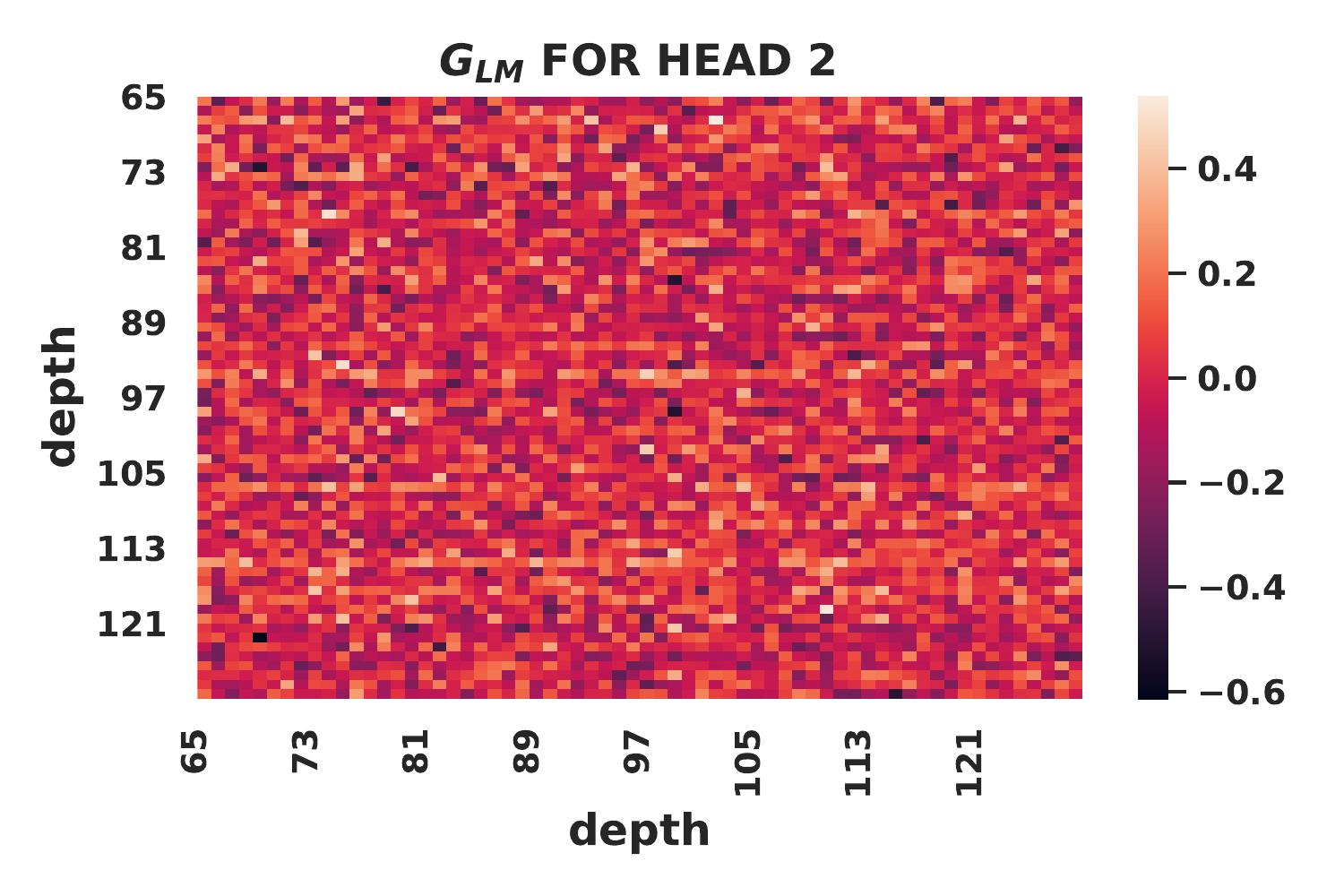}

\end{subfigure}
\hfill
\begin{subfigure}[b]{0.6\textwidth}
	\centering
	\includegraphics[width=1.1\textwidth]{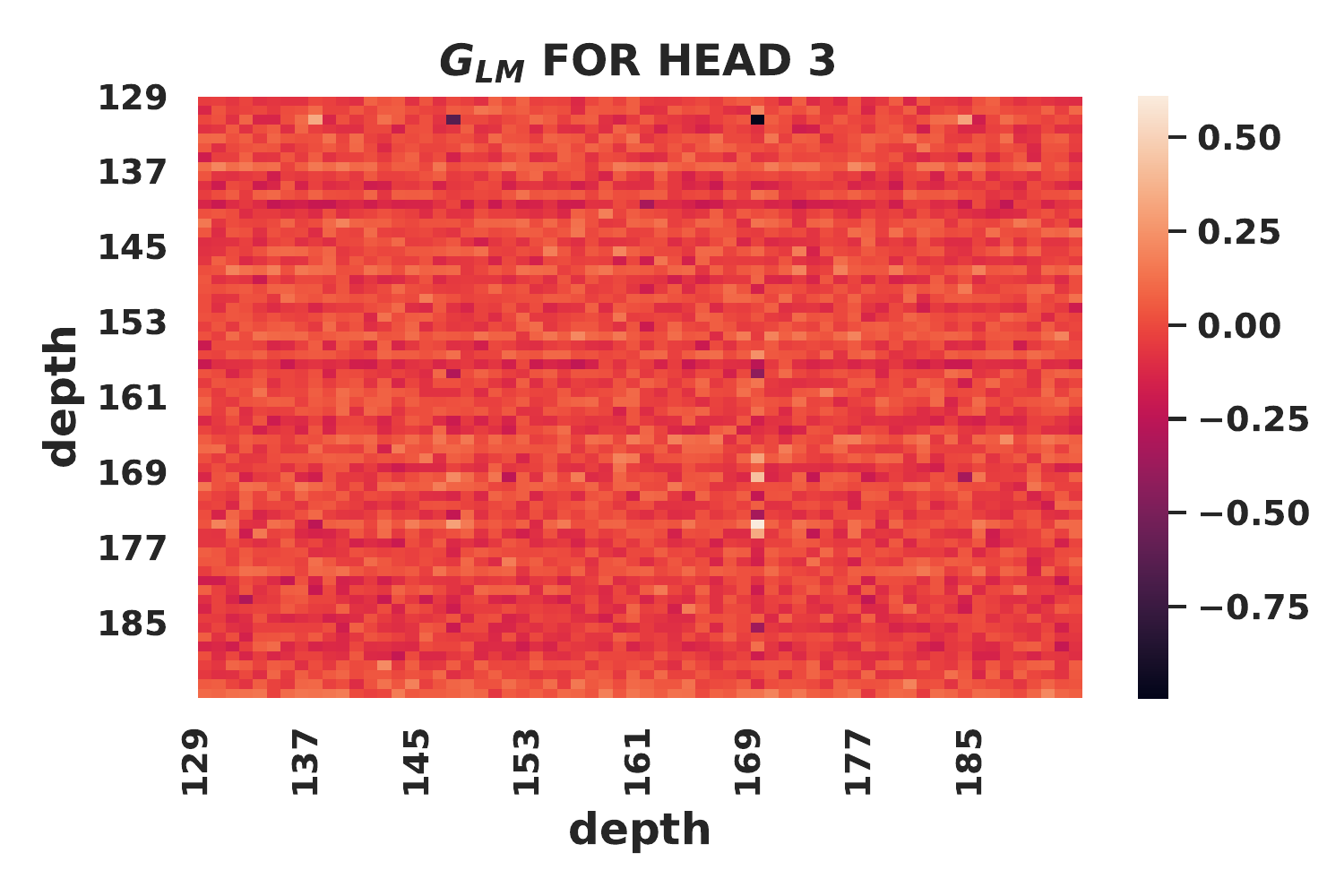}

\end{subfigure}
\hfill
\begin{subfigure}[b]{0.6\textwidth}
	\centering
	\includegraphics[width=1.1\textwidth]{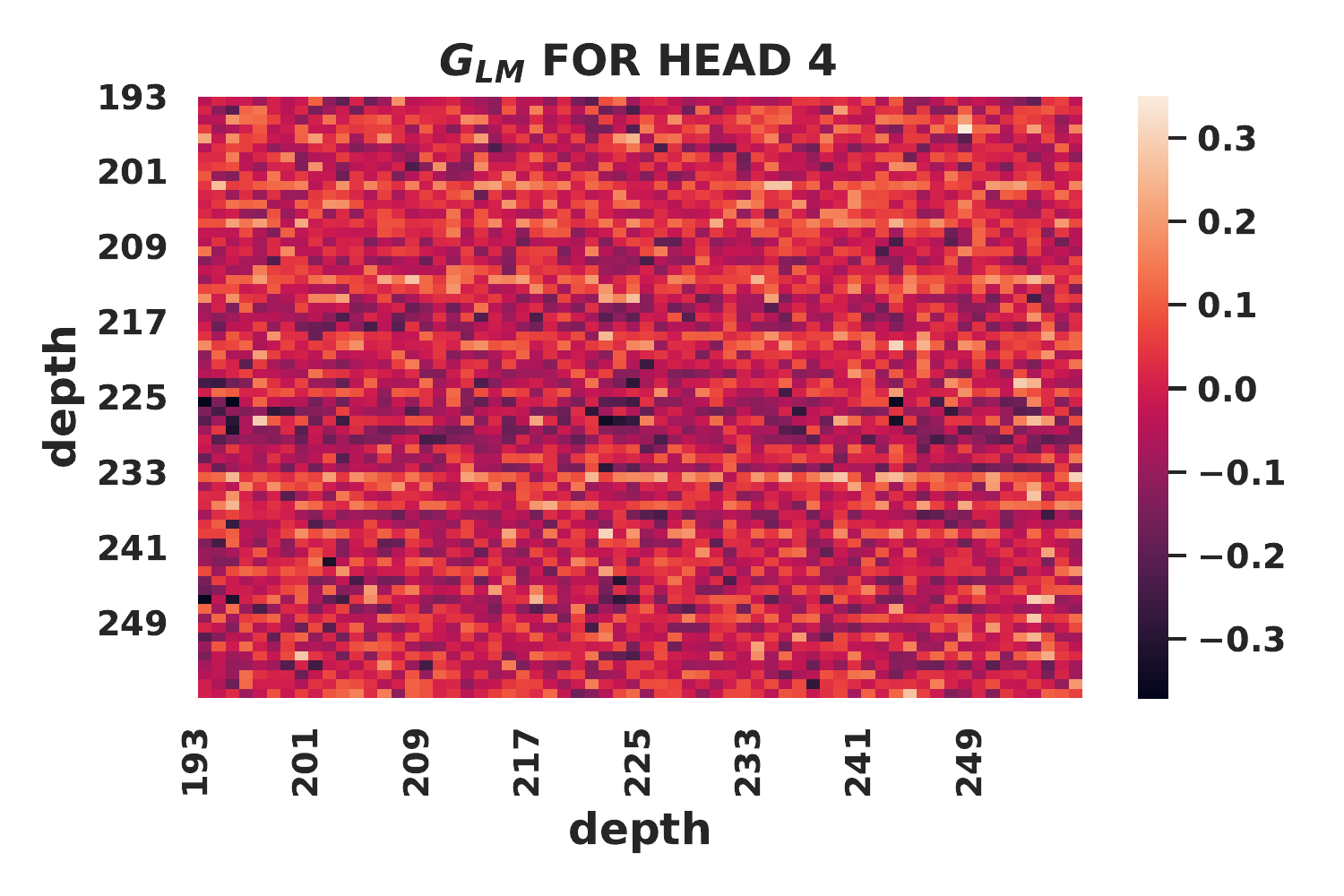}

\end{subfigure}
\centering
\begin{subfigure}[b]{0.6\textwidth}
	\centering
	\includegraphics[width=1.1\textwidth]{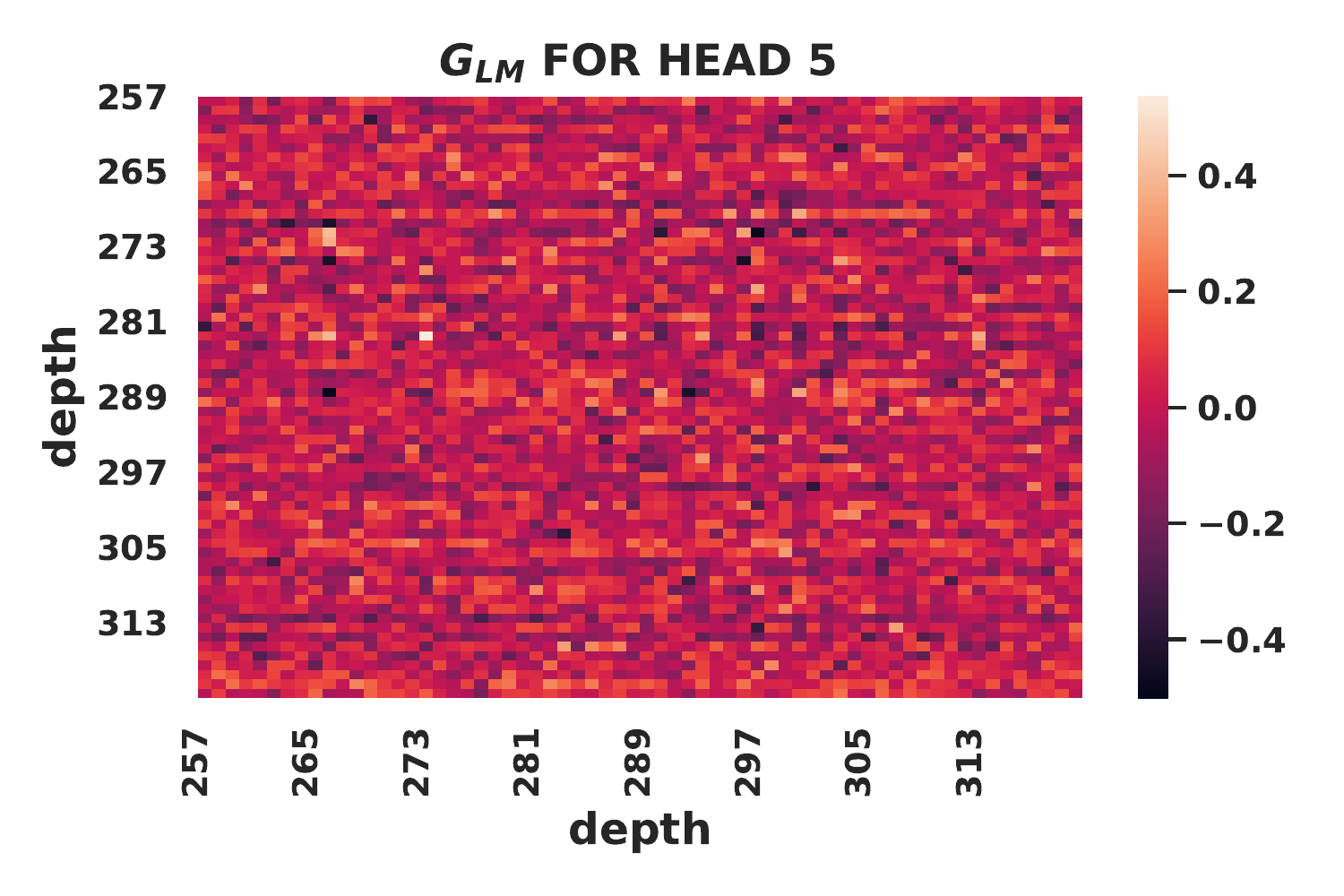}

\end{subfigure}
\hfill
\begin{subfigure}[b]{0.6\textwidth}
	\centering
	\includegraphics[width=1.1\textwidth]{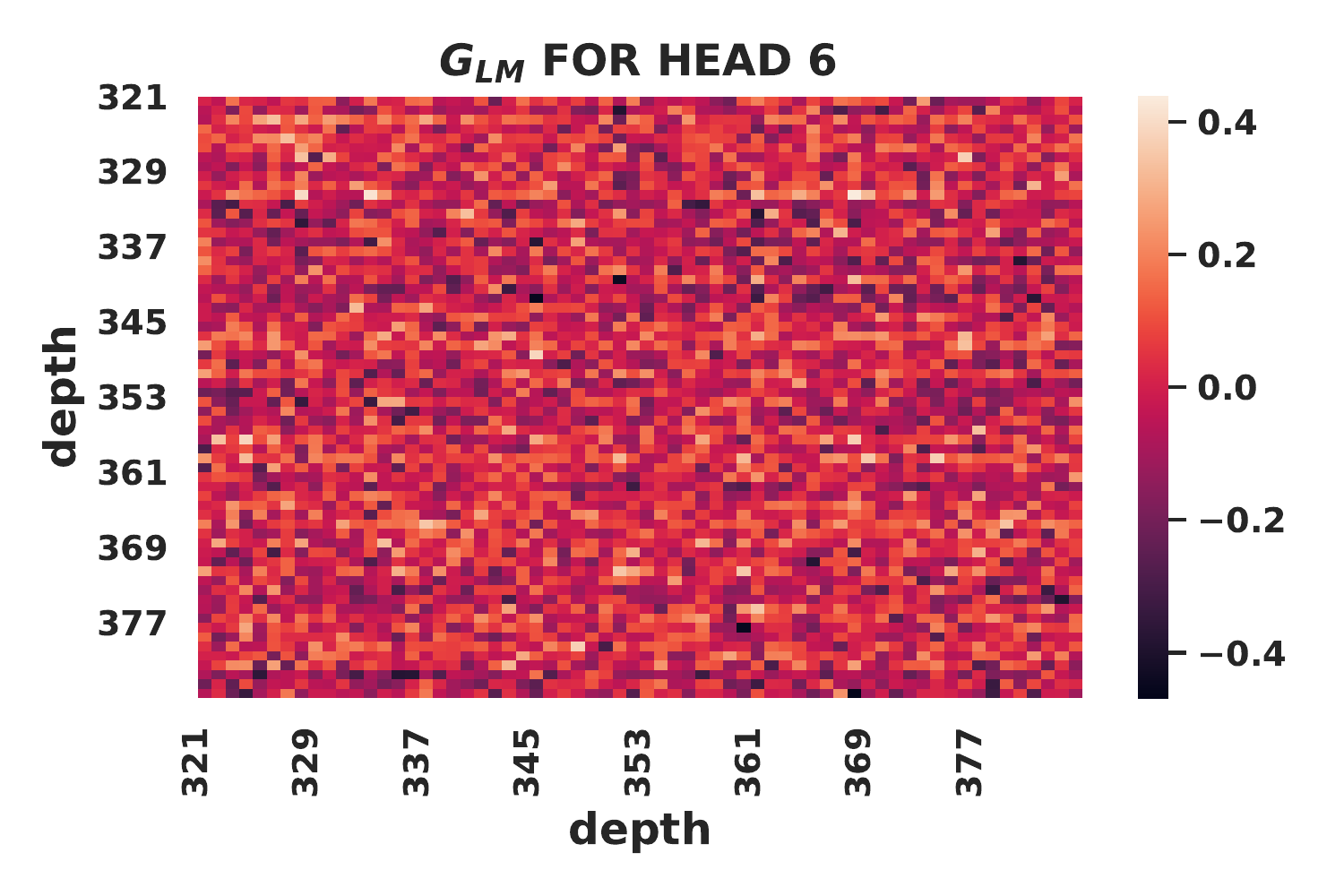}

\end{subfigure}
\hfill
\begin{subfigure}[b]{0.6\textwidth}
	\centering
	\includegraphics[width=1.1\textwidth]{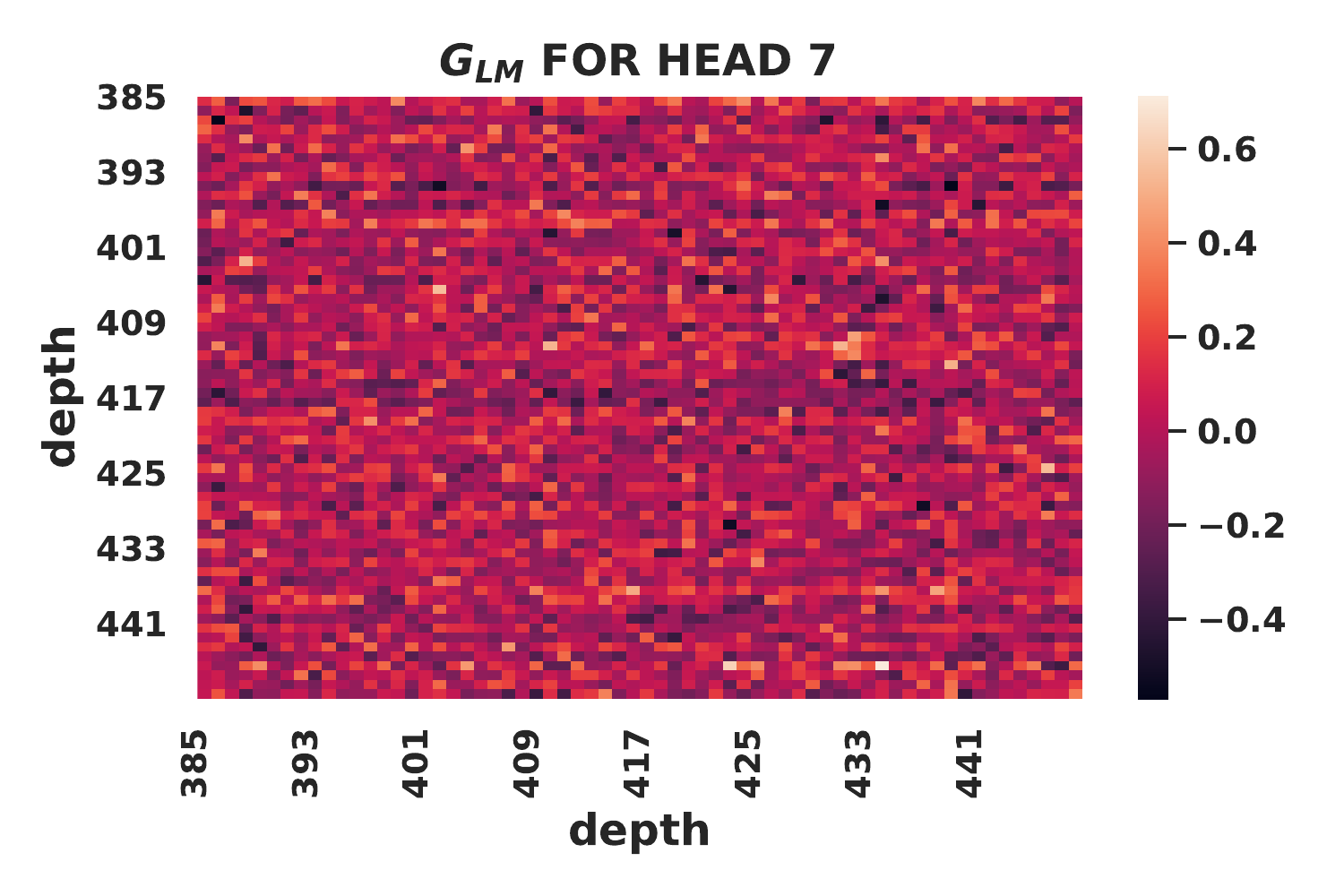}

\end{subfigure}
\hfill
\begin{subfigure}[b]{0.6\textwidth}
	\centering
	\includegraphics[width=1.1\textwidth]{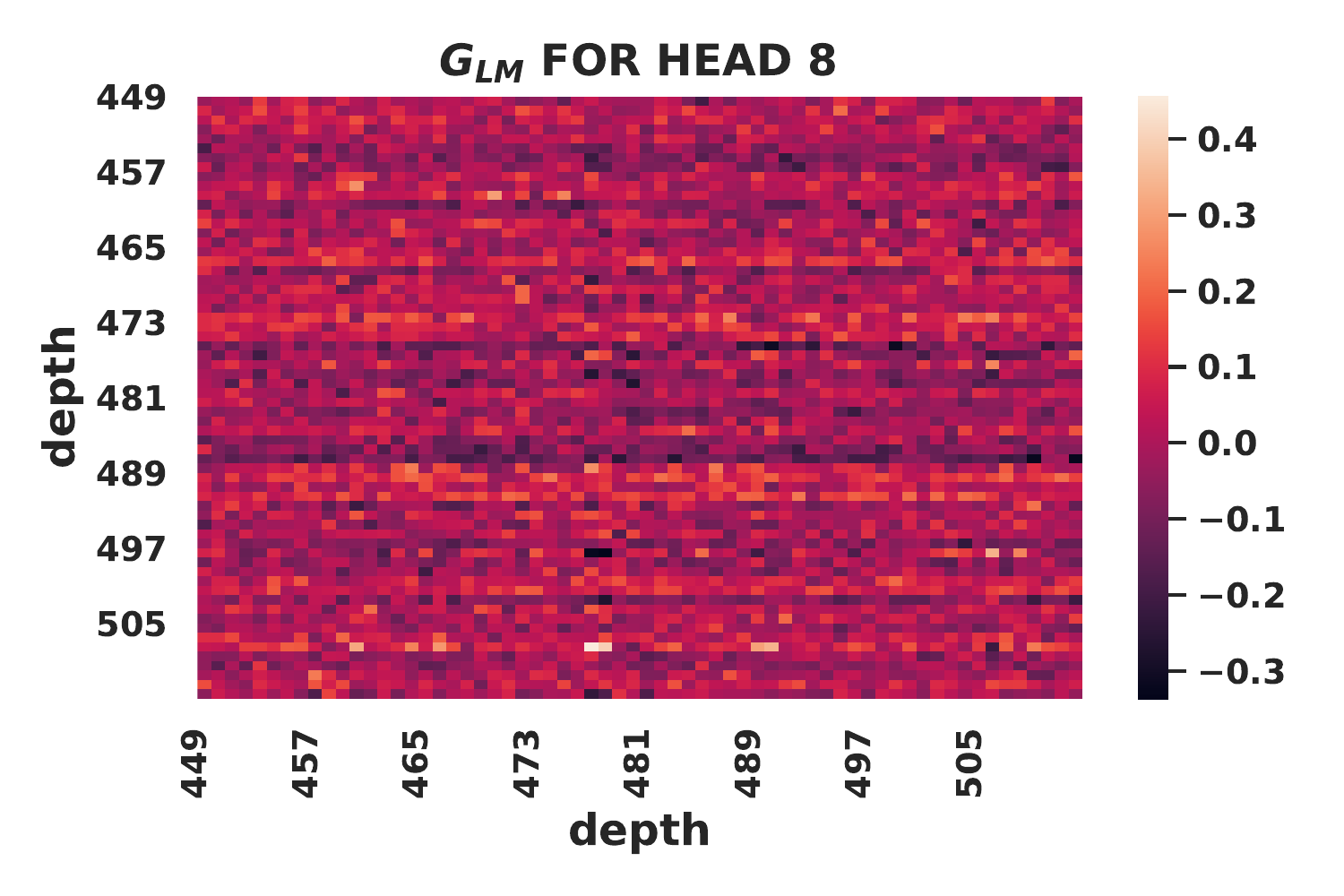}

\end{subfigure}
\caption{$\mG_{LM}$ heatmap plots for all heads from TLM attention stage from graph transformer model \#2 for PT-EN translation task.}
\label{fig32apx}
\end{adjustwidth}
\end{figure}

\clearpage

\thispagestyle{headings}
\begin{figure}
\begin{adjustwidth}{-5em}{-5em}
\centering
\begin{subfigure}[b]{0.6\textwidth}
	\centering
	\includegraphics[width=1.1\textwidth]{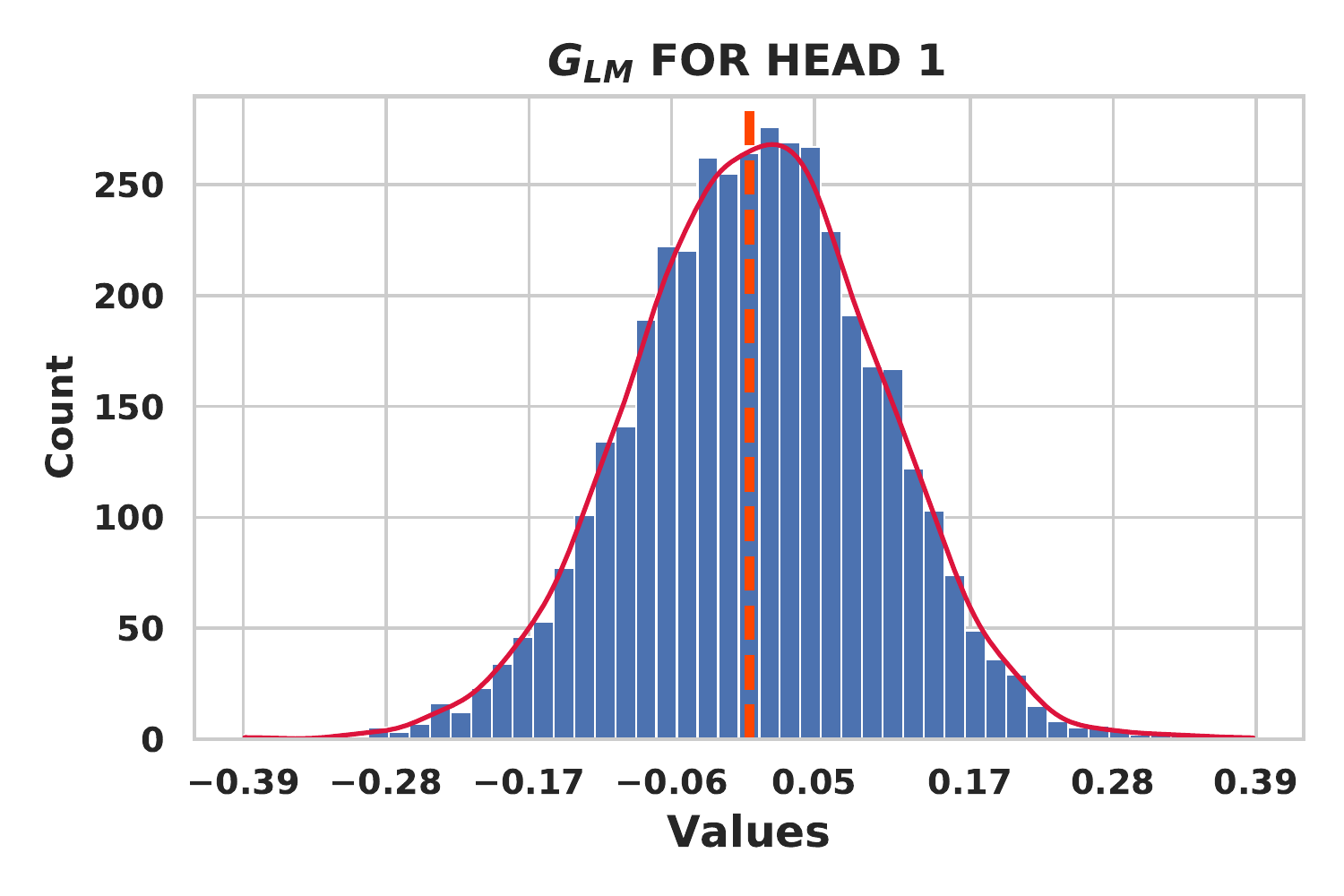}

\end{subfigure}
\hfill
\begin{subfigure}[b]{0.6\textwidth}
	\centering
	\includegraphics[width=1.1\textwidth]{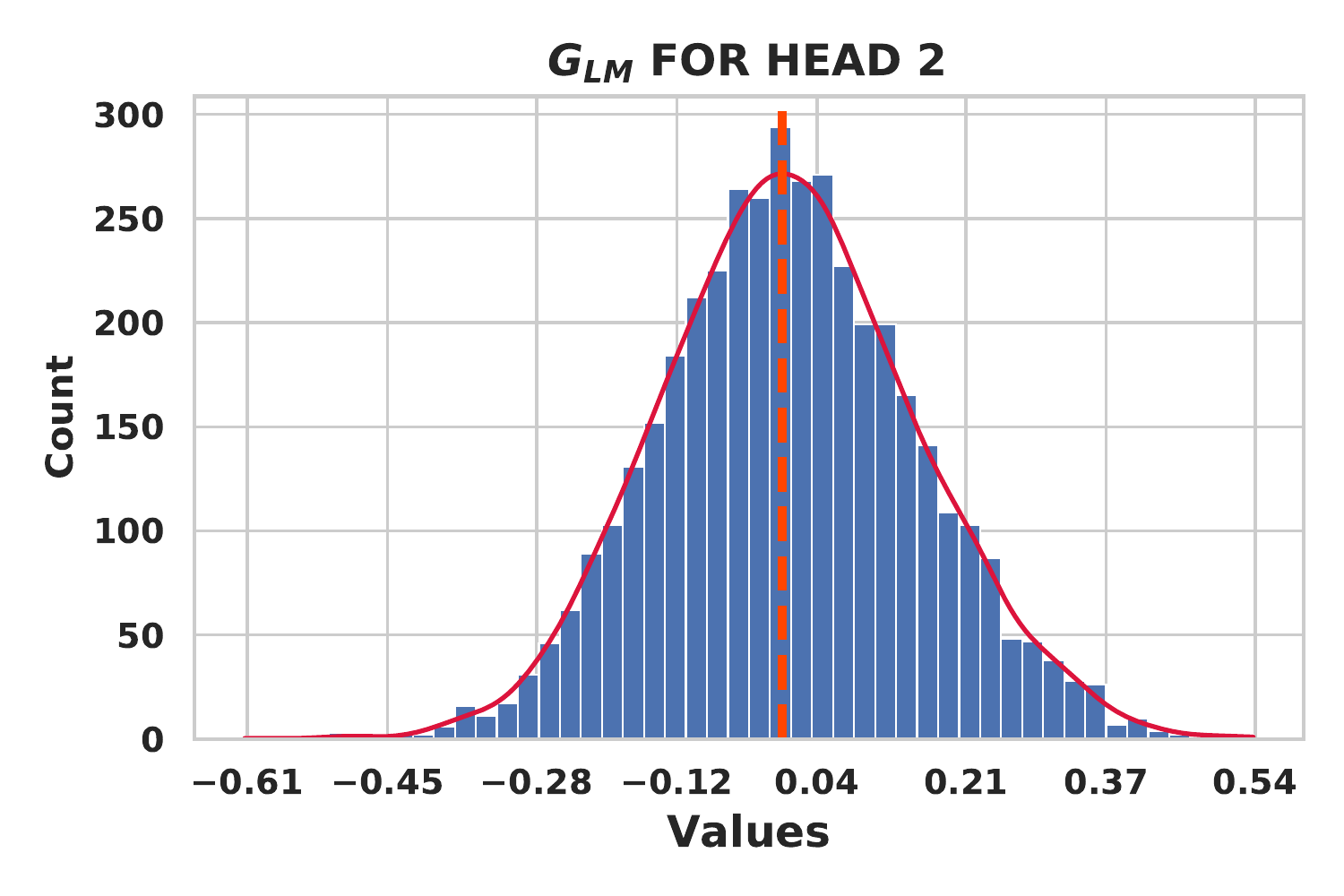}

\end{subfigure}
\hfill
\begin{subfigure}[b]{0.6\textwidth}
	\centering
	\includegraphics[width=1.1\textwidth]{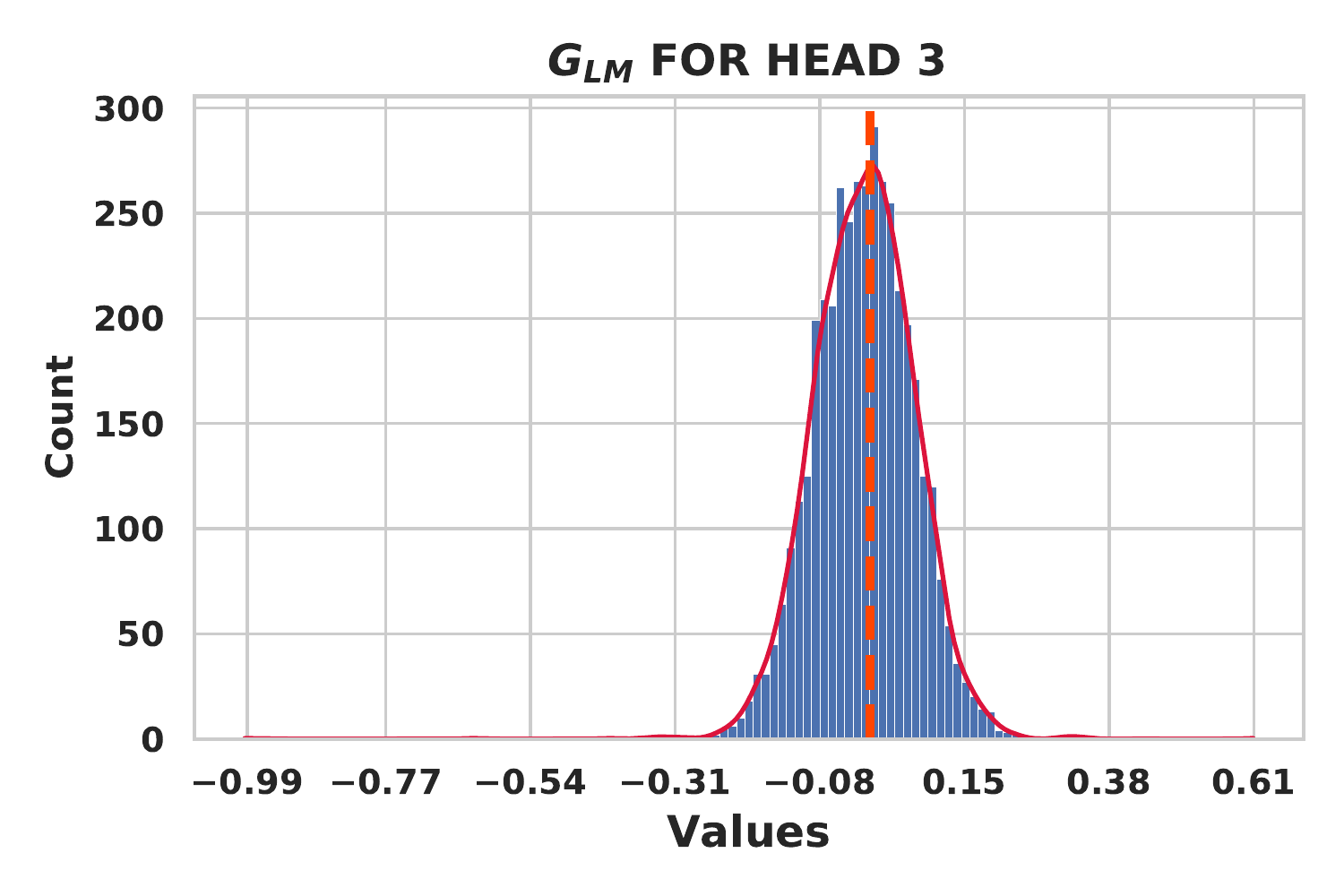}

\end{subfigure}
\hfill
\begin{subfigure}[b]{0.6\textwidth}
	\centering
	\includegraphics[width=1.1\textwidth]{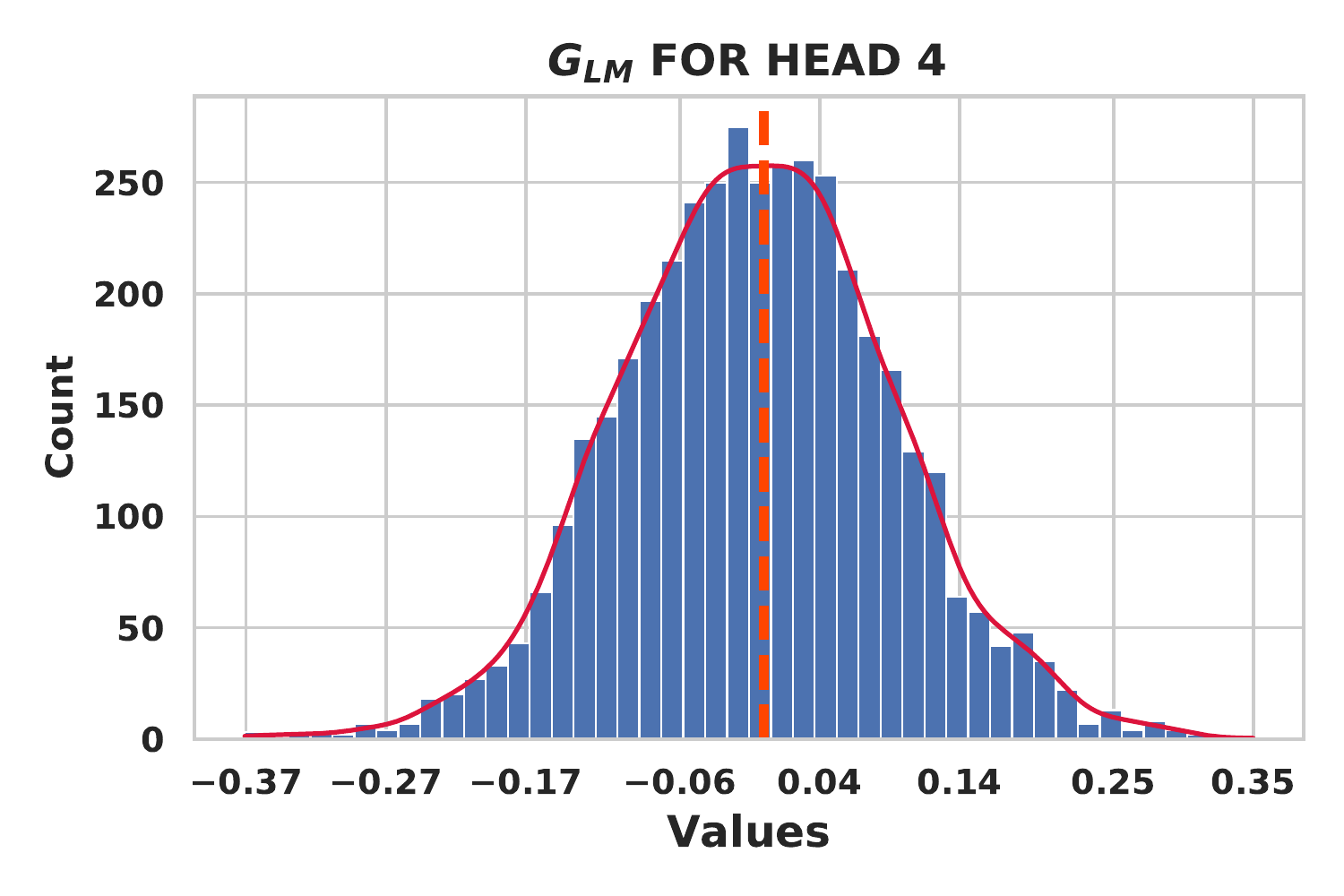}

\end{subfigure}
\centering
\begin{subfigure}[b]{0.6\textwidth}
	\centering
	\includegraphics[width=1.1\textwidth]{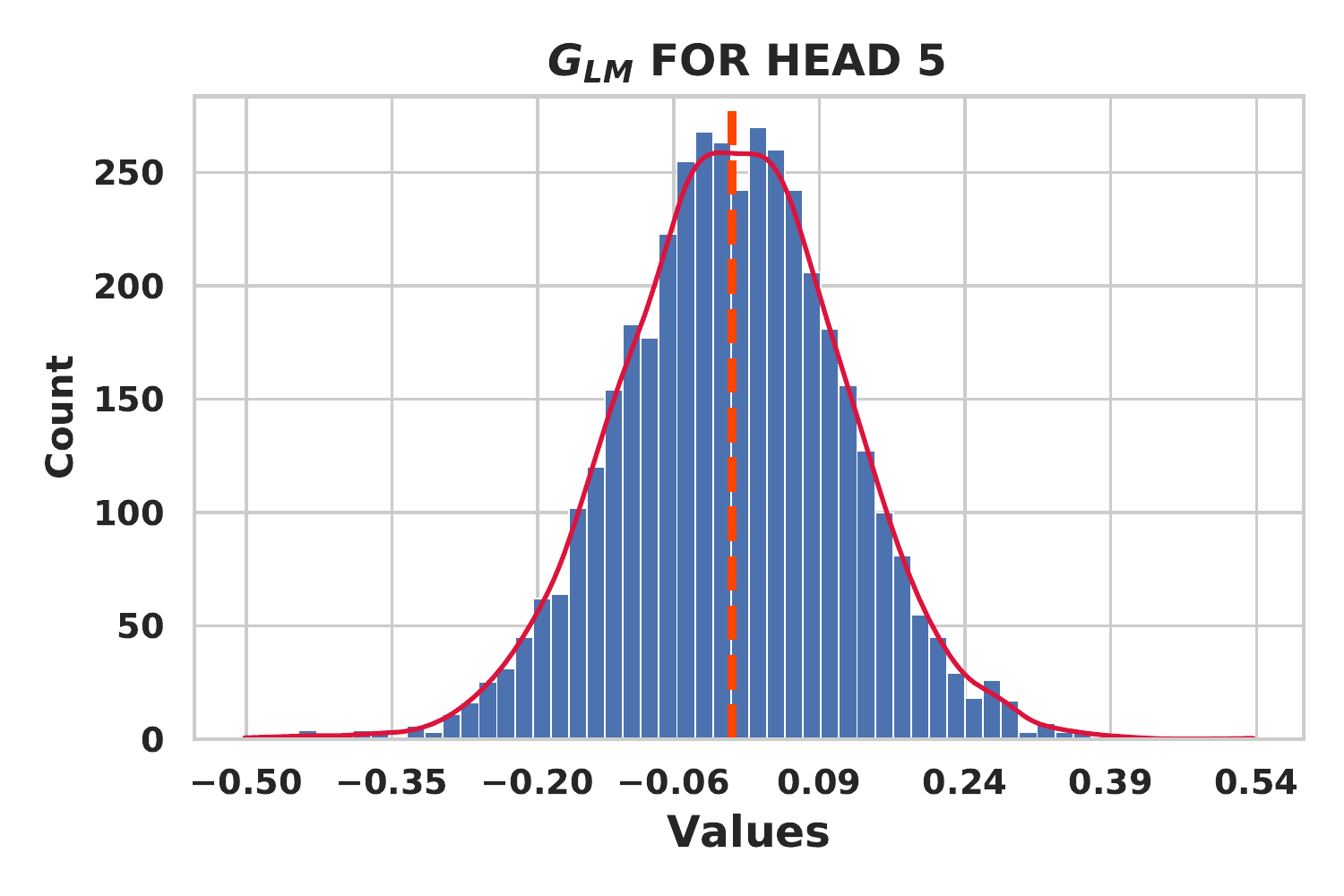}

\end{subfigure}
\hfill
\begin{subfigure}[b]{0.6\textwidth}
	\centering
	\includegraphics[width=1.1\textwidth]{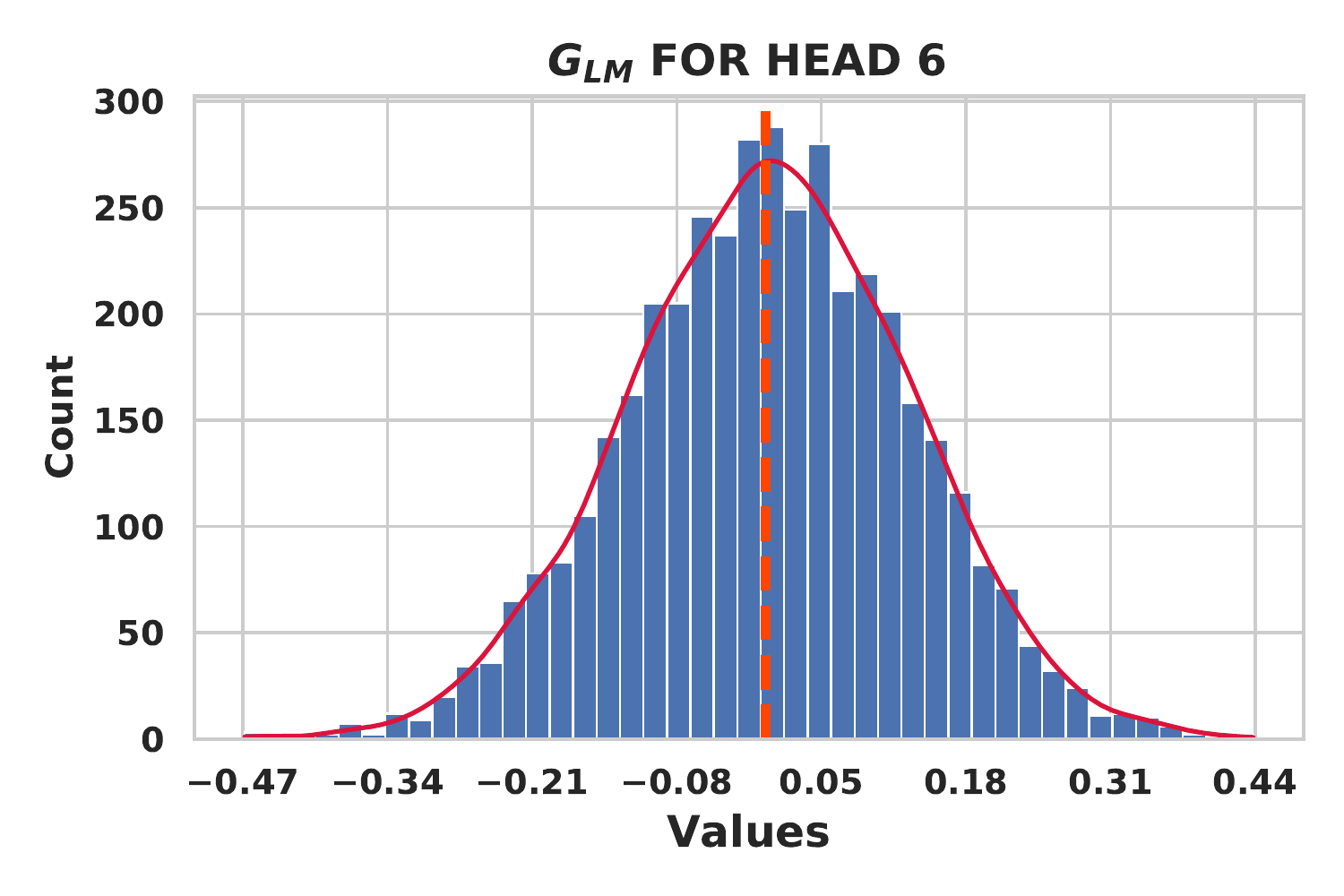}

\end{subfigure}
\hfill
\begin{subfigure}[b]{0.6\textwidth}
	\centering
	\includegraphics[width=1.1\textwidth]{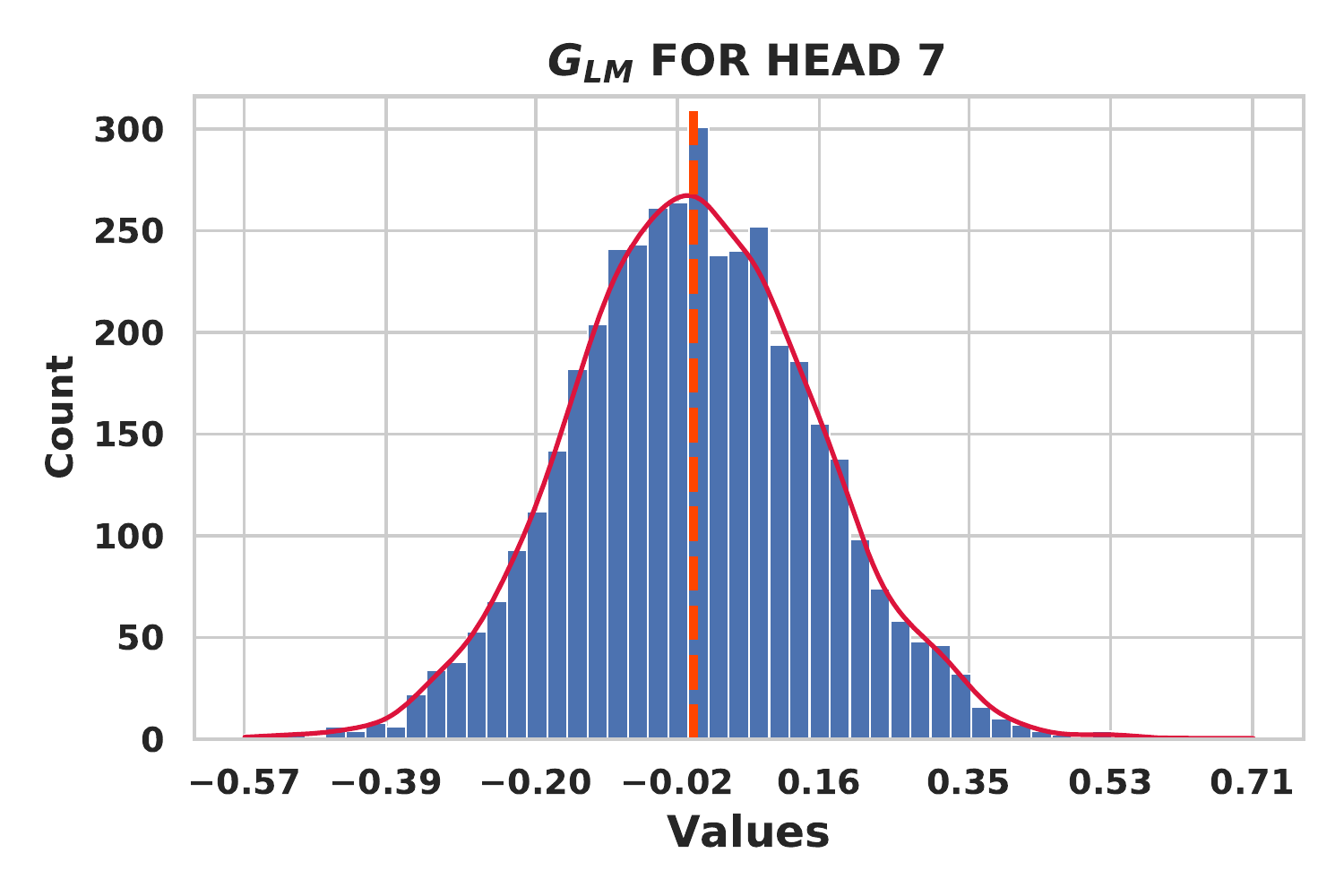}

\end{subfigure}
\hfill
\begin{subfigure}[b]{0.6\textwidth}
	\centering
	\includegraphics[width=1.1\textwidth]{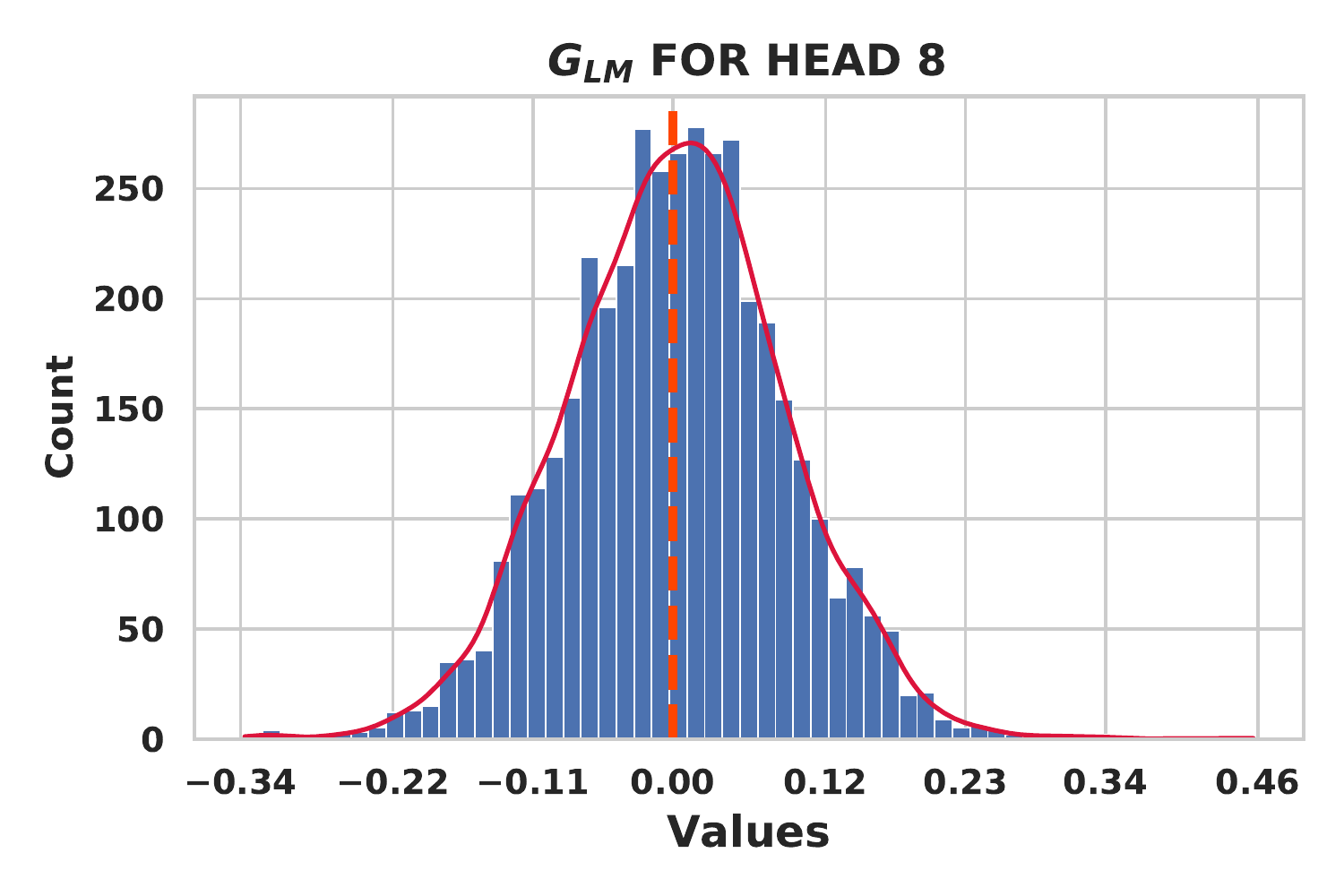}

\end{subfigure}
\caption{$\mG_{LM}$ histogram plots for all heads from TLM attention stage from graph transformer model \#2 for PT-EN translation task. Dashed line in orange marks zero value.}
\label{fig33apx}
\end{adjustwidth}
\end{figure}

\end{document}